\documentclass[10pt,fleqn,twoside,openright]{book}

\usepackage{siunitx}
\usepackage[T1]{fontenc}
\usepackage[utf8]{inputenc}
\usepackage[catalan,spanish,english]{babel}
\usepackage{authblk}
\usepackage{epigraph}
\usepackage[printonlyused,withpage]{acronym}

\usepackage{floatrow}
\usepackage{multirow}
\usepackage{makecell}
\usepackage{etoolbox}

\newfloatcommand{capbtabbox}{table}[][\FBwidth]

\usepackage{lmodern}

\usepackage{fourier} 

\usepackage{setspace} 

\usepackage[usenames, dvipsnames]{xcolor}
\usepackage{graphicx}

\usepackage{subfig}
\usepackage{booktabs}

\usepackage{lipsum}
\usepackage{microtype}
\usepackage{url}
\usepackage[final]{pdfpages}

\usepackage{datasheet}

\usepackage{fancyhdr}

\fancypagestyle{plain}{
	\fancyhf{}

	\fancyfoot[EL,OR]{\thepage}}
\fancypagestyle{addpagenumbersforpdfimports}{
	\fancyhead{}
	
	\fancyfoot{}
	\fancyfoot[RO,LE]{\thepage}
}

\usepackage{listings}
\usepackage[breaklinks=true]{hyperref} 
\hypersetup{pdfborder={0 0 0},
	colorlinks=true,
	linkcolor=black,
	citecolor=black,
	urlcolor=black}
\makeatletter
\def\cleardoublepage{\clearpage\if@twoside \ifodd\c@page\else
    \hbox{}
    \thispagestyle{empty}
    \newpage
    \if@twocolumn\hbox{}\newpage\fi\fi\fi}
\makeatother \clearpage{\pagestyle{plain}\cleardoublepage}

\usepackage{color}
\usepackage{tikz}
\usepackage[explicit]{titlesec}
\newcommand*\chapterlabel{}
\titleformat{\chapter}[display]  
	{\normalfont\bfseries\LARGE} 
	{\gdef\chapterlabel{\thechapter\ }}     
 	{0pt} 
 	  {\begin{tikzpicture}[remember picture,overlay]
    \node[yshift=-4cm] at (current page.north west)
      {\begin{tikzpicture}[remember picture, overlay]
        \draw[fill=black] (0,0) rectangle(35.5mm,15mm);
        \node[anchor=north east,yshift=-3.2cm,xshift=34mm,minimum height=10mm,inner sep=0mm] at (current page.north west)
        {\parbox[top][40mm][t]{15mm}{\raggedleft $\phantom{\textrm{l}}$\color{white}\chapterlabel}};  
        \node[anchor=north west,yshift=-3.2cm,xshift=37mm,text width=\textwidth,minimum height=10mm,inner sep=0mm] at (current page.north west)
              {\parbox[top][10mm][t]{\textwidth}{\color{black}#1}};
       \end{tikzpicture}
      };
   \end{tikzpicture}
   \gdef\chapterlabel{}
  } 

\titlespacing*{\chapter}{0pt}{50pt}{1pt}
\titlespacing*{\section}{0pt}{13.2pt}{10pt}  
\titlespacing*{\subsection}{0pt}{13.2pt}{6pt}
\titlespacing*{\subsubsection}{0pt}{13.2pt}{4pt}

\newcounter{myparts}
\newcommand*\partlabel{}
\titleformat{\part}[display]  
	{\normalfont\bfseries\huge} 
	{\gdef\partlabel{\thepart\ }}     
 	{0pt} 
 	  {\setlength{\unitlength}{20mm}
	  \begin{tikzpicture}[remember picture,overlay]
    \node[anchor=north west,xshift=-65mm,yshift=-6.9cm-\value{myparts}*20mm] at (current page.north east) 
      {\begin{tikzpicture}[remember picture, overlay]
        \draw[fill=black] (0,0) rectangle(62mm,20mm);   
        \node[anchor=north west,yshift=-6.1cm-\value{myparts}*20mm,xshift=-60.5mm,minimum height=30mm,inner sep=0mm] at (current page.north east)
        {\parbox[top][30mm][t]{55mm}{\raggedright \color{white}Part \partlabel $\phantom{\textrm{l}}$}};  
        \node[anchor=north east,yshift=-6.1cm-\value{myparts}*20mm,xshift=-63.5mm,text width=\textwidth,minimum height=30mm,inner sep=0mm] at (current page.north east)
              {\parbox[top][30mm][t]{\textwidth}{\raggedleft \color{black}#1} \thispagestyle{empty}};
       \end{tikzpicture}
      };
   \end{tikzpicture}
  \addtocounter{myparts}{1}
   \gdef\partlabel{}
  } 

\newcommand{\cmark}{\ding{51}}%
\newcommand{\xmark}{\ding{55}}%

\usepackage{rotating}
\usepackage[export]{adjustbox}

\newcommand*{\headformat}[1]{#1}
\newlength{\maxlen}
\newcommand*{\head}[1]{%
	\begin{sideways}
		\makebox[\maxlen][l]{\headformat{#1}}
\end{sideways}}

\usepackage{amsmath}
\makeatletter
\def\resetMathstrut@{%
  \setbox\z@\hbox{%
    \mathchardef\@tempa\mathcode`\(\relax
      \def\@tempb##1"##2##3{\the\textfont"##3\char"}%
      \expandafter\@tempb\meaning\@tempa \relax
  }%
  \ht\Mathstrutbox@1.2\ht\z@ \dp\Mathstrutbox@1.2\dp\z@
}
\makeatother

\usepackage{epigraph}

\usepackage{enumitem}		
\usepackage{colortbl}
\usepackage{nccmath}
\usepackage{bibentry}
\usepackage{filecontents}
\usepackage{epstopdf}
\usepackage{ragged2e}
\usepackage[utf8]{inputenc}
\usepackage{textcomp}
\usepackage[english]{babel}
\usepackage{bm}  
\usepackage{amsfonts}   
\usepackage{verbatim}
\usepackage{amssymb}
\usepackage{graphics}
\usepackage{algorithm}
\usepackage{algorithmic}
\usepackage{color}
\usepackage{geometry}
\usepackage{longtable}
\usepackage{amsmath}
\usepackage{pifont}
\usepackage{lscape}
\usepackage{xspace}
\usepackage{amsopn}
\usepackage{mathtools}
\usepackage{etextools}
\usepackage{alphabeta}

\usepackage{float}
\usepackage{type1cm}         
\usepackage{makeidx}         
\usepackage{multicol}        
\usepackage[bottom]{footmisc}
\usepackage{amsmath}
\usepackage{bm}
\usepackage{mathtools}
\usepackage{multirow}
\usepackage{array}
\usepackage{url}
\usepackage{lipsum}
\usepackage{dblfloatfix}
\usepackage{setspace}
\usepackage{tabulary}
\usepackage[sort,numbers]{natbib}
\usepackage{xstring}

\newcommand{\UAB}{Universitat Autònoma de Barcelona}
\newcommand{\CVC}{Centre de Visió per Computador}
\newcommand{\DCC}{Dept. Ciències de la computació}
\newcommand{\DCCCVC}{\DCC{} \& \CVC{}}
\newcommand{\printer}{Ediciones Gráficas Rey, S.L.}

\newcommand{\ISBN}{978-84-126409-6-0}

\newtheorem{thm}{Theorem}

\newtheorem{defn}[thm]{Definition}

\newlength{\reducedwidth}
\newenvironment{abstract}{\begin{center}\begin{minipage}{\reducedwidth}
\hrulefill\vspace{3pt}\bf\\\small}{\par\hrulefill\\\end{minipage}\end{center}}

\DeclareMathOperator*{\argmax}{\arg\!\max}

\definecolor{light-gray}{gray}{0.4}

\def\ontop#1#2{\setbox0\hbox{#2}\copy0\llap{\raise\ht0\hbox{#1}}}

\graphicspath{{figures/}}

\newtheorem{theorem}{Theorem}[section]

\newtheorem{proposition}[theorem]{Proposition}
\newtheorem{corollary}[theorem]{Corollary}

\newcommand{\qed}{\nobreak \ifvmode \relax \else
      \ifdim\lastskip<1.5em \hskip-\lastskip
      \hskip1.5em plus0em minus0.5em \fi \nobreak
      \vrule height0.75em width0.5em depth0.25em \fi}

\DeclareTextFontCommand{\emph}{\em}

\definecolor{light-gray}{gray}{0.4}

\def\ontop#1#2{\setbox0\hbox{#2}\copy0\llap{\raise\ht0\hbox{#1}}}

\def\1n{\mathbf{1}_n}
\def\0{\mathbf{0}}
\def\1{\mathbf{1}}

\newcommand{\UP}[1]{(\IfSubStr{#1}{-}{ \textcolor{red}{\bf #1} }{ \textcolor{blue}{\bf #1}})}

\makeatletter
\newsavebox\saved@arstrutbox
\newcommand*{\setarstrut}[1]{%
  \noalign{%
    \begingroup
      \global\setbox\saved@arstrutbox\copy\@arstrutbox
      #1%
      \global\setbox\@arstrutbox\hbox{%
        \vrule \@height\arraystretch\ht\strutbox
               \@depth\arraystretch \dp\strutbox
               \@width\z@
      }%
    \endgroup
  }%
}
\newcommand*{\restorearstrut}{%
  \noalign{%
    \global\setbox\@arstrutbox\copy\saved@arstrutbox
  }%
}
\makeatother

\title{Towards Unsupervised Representation Learning: Learning, Evaluating and Transferring Visual Representations}
\author{\textbf{Bonifaz Stuhr}}
\date{September 26, 2023}

\begin{document}
	
\frontmatter
\makeatletter
\begin{titlepage}
	
\begin{center}
\sffamily

\vspace*{0.9cm}
\includegraphics[width=6.5cm]{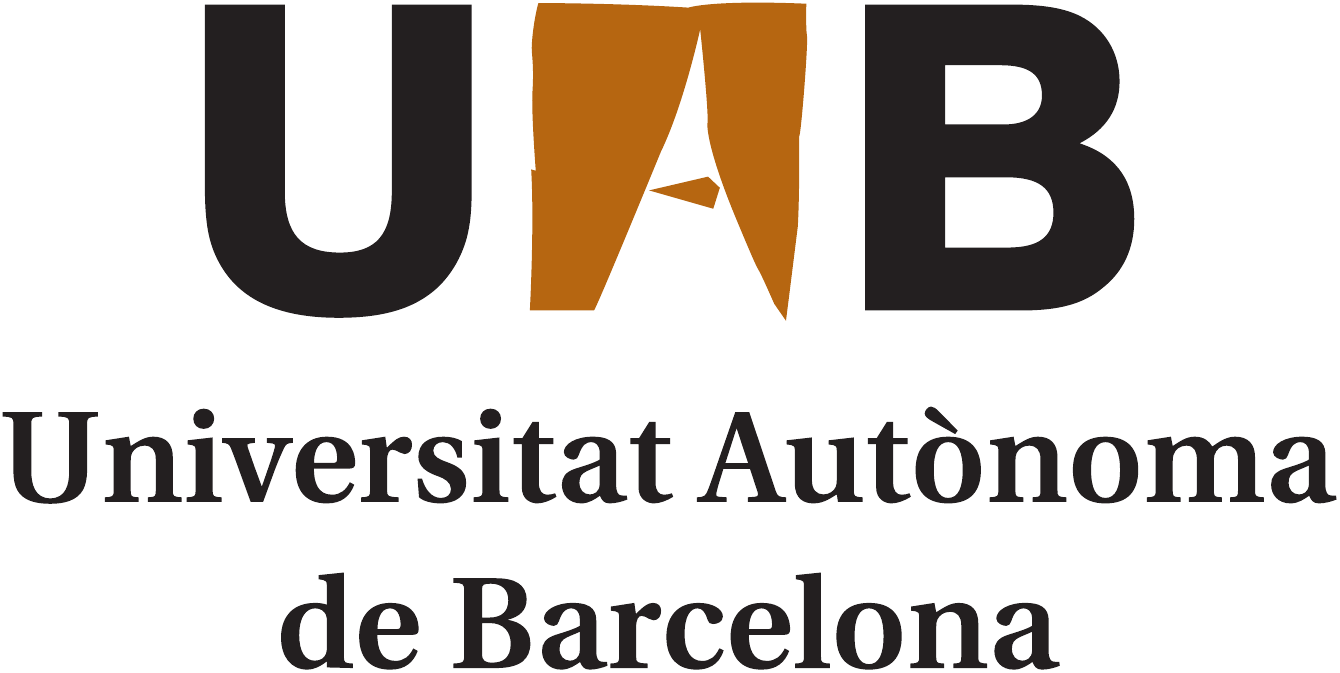}
\null\vspace{3.275cm}

{\par
	\centering 
	\Huge Towards Unsupervised\\[-1mm]Representation Learning:\\[-1mm]Learning, Evaluating and\\[-1mm]Transferring Visual\\[-1mm]Representations
	\par
}
\vfill
\end{center}

\vfill
\begin{flushright}
  \parbox{7.5cm}{A dissertation submitted by {\bf \@author} at \UAB~to fulfil the degree of \textbf{Doctor of Philosophy}.\par
   \vspace*{2mm}
   \hfill Bellaterra, \@date
  }
\end{flushright}
  
\end{titlepage}

\newpage
  \thispagestyle{empty}
  \noindent\begin{tabular}{p{2cm}|l}
   Co-Director  & \textbf{Dr. Jordi Gonzàlez Sabaté} \\
   			 & \DCCCVC \\
   			 & \UAB \\

  \multicolumn{2}{c}{}\\
  
Co-Director  & \textbf{Dr. Jürgen Brauer} \\
             & Department of Computer Science \\
             & University of Applied Sciences Kempten \\

  \multicolumn{2}{c}{}\\
               
Thesis       & \textbf{Dr. Xavier Roca Marva} \\
committee    & \DCCCVC \\
             & \UAB \\   
             & \\                          
             & \textbf{Dr. Ulrich Göhner}  \\
             & Department of Computer Science \& IFM \\ 
             & University of Applied Sciences Kempten \\
             & \\   
             & \textbf{Dr. Wenjuan Gong} \\
             & College of Computer Science and Technology \\
             & China University of Petroleum (East China) \\

  \multicolumn{2}{c}{}\\
    

  \end{tabular}
  
  \vfill
  \noindent\begin{minipage}{\textwidth}
  \setlength{\parskip}{4pt}

  \noindent \rule{80mm}{0.4mm} 
  \includegraphics[trim=40mm 60mm 60mm 100mm,width=3.0cm]{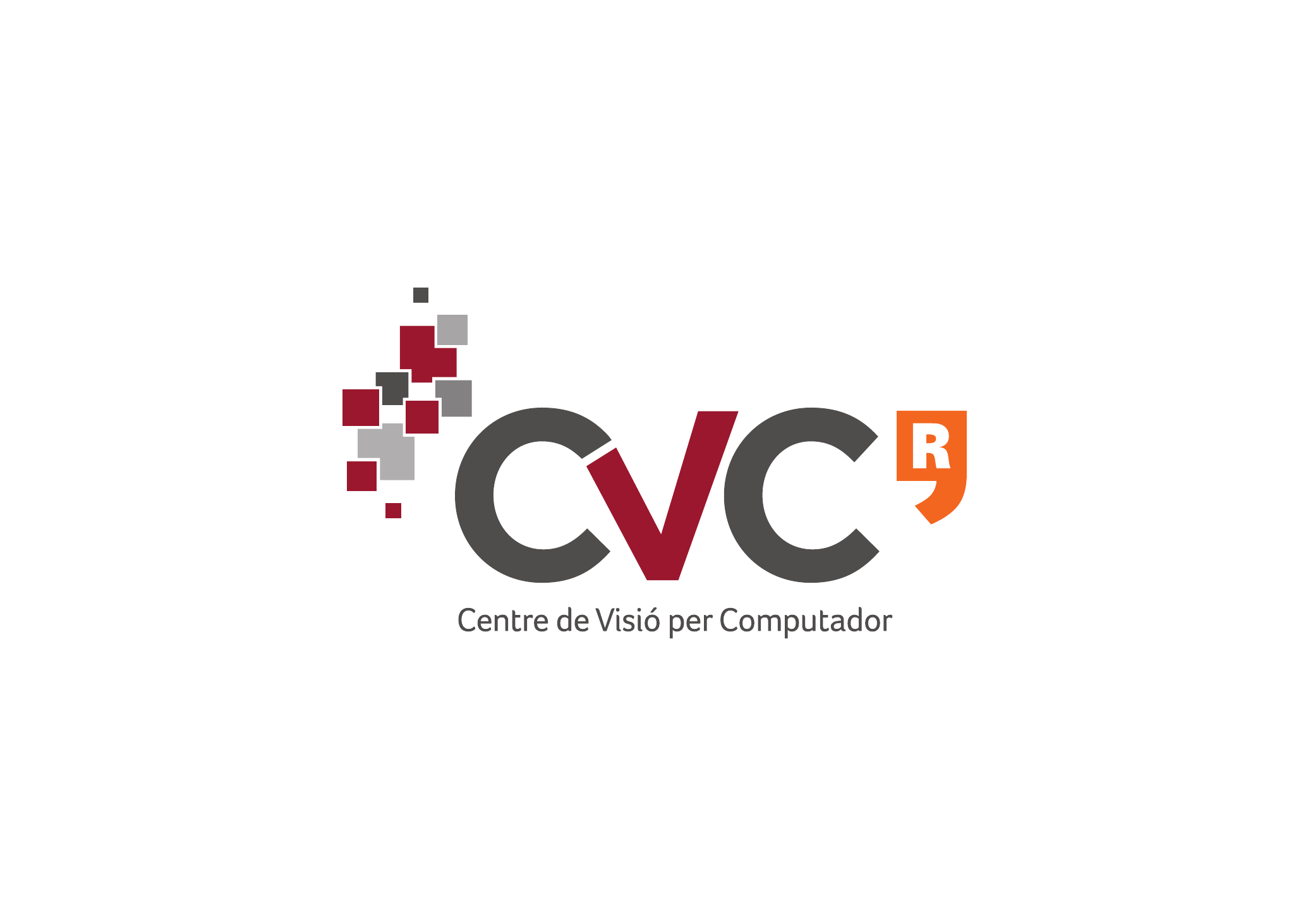}
  \small
  \par
  This document was typeset by the author using \LaTeXe.\par
  A dissertation submitted by {\bf \@author} at \DCC, \UAB~to fulfil the degree of \textbf{Doctor of Philosophy} within the Computer Science Doctorate Program. Copyright \textcopyright 2023 by \@author. All rights reserved. No part of this publication may be reproduced or transmitted in any form or by any means, electronic or mechanical, including photocopy, recording, or any information storage and retrieval system, without permission in writing from the author.\par  
  ISBN: \ISBN\par
  Printed by \printer
  \vspace*{2mm}
  \hfill Bellaterra, \@date
  \end{minipage}
  \setcounter{page}{0}

\makeatother

 \setlength{\parskip}{1em}
\cleardoublepage
\thispagestyle{empty}

\vspace*{3cm}

\begin{raggedleft}
It is important that students bring a certain ragamuffin, barefoot, irreverence to their studies; they are not here to worship what is known, but to question it. \\
     --- Jacob Bronowski\\
\end{raggedleft}

\vspace{8mm}


\vspace{6cm}

\begin{center}
    For my family and friends
\end{center}

\newpage

\setcounter{page}{0}

\setlength{\parindent}{15pt}
\setlength{\parskip}{0em}

\chapter*{Acknowledgements}
\vspace{-14mm}
Looking back, it is all but incomprehensible for me how valuable the last years as a doctoral student have been for my academic and personal development and my life as a whole. I have deeply enjoyed the academic freedom; it has made me more critical in a positive, curious way and has led me to many beautiful discoveries about the scientific community and myself - as a small part of it. I look forward to further paddling my rowing boat through the misty and mysterious sea of science in search of more fascinating wonders beyond the shimmering fog.

For this dissertation and the invaluable opportunity to contribute to the research of new intelligence, I must especially express my deepest gratitude to my supervisors Prof. Dr. Jürgen Brauer and Prof. Dr. Jordi Gonzàlez Sabaté. The freedom you gave me and your wise and on-point advice were the best things that could happen to me during my Ph.D. years. I can not thank you enough for your trust and kindness. Jürgen, I will miss our long meetings where we talked about science, teaching, and technical innovations. The memories of our conference trip to Florida always bring a smile to my face. Jordi, I can not remember any meeting with you that was not enlightening for me. From the first time we met, I knew you were the perfect supervisor for this thesis, both scientifically and organizationally. I will always keep your wise perspective on our papers and the academic field as a whole in my mind.  I am also extremely grateful to Prof. Bernhard Schick from the IFM. Bernhard, you made it possible for me to work with the Adrive team and you gave me the freedom to pursue my ideas in the context of your topics, which led to amazing projects, events, and teaching opportunities. I hope that the future ADC team will continue our legacy! At this point, I would like to thank my colleagues at the CVC and IFM sincerely; our academic, engineering, and recreational discussions frequently made my day. Johann, collaborating with you especially has made for incredible moments and achievements. Without you, research, teaching, and participation in the VDI ADC Cup would only be half as much fun. Julian, thank you for your dedication to your master's thesis and our subsequent project. Huge thanks also go to the CVC and IFM administrative and marketing staff; you keep the research running! 

Furthermore, I thank all my friends, with whom I could easily find distraction and tranquility. A big thank you goes to my uncle for professionally proofreading this thesis. Moreover, I am grateful to my grandparents and brothers for their positive energy, wishes, and support. Finally, I want to thank my parents from the bottom of my heart for their deep love, everlasting support, and for being positive role models. Your influence on me and my life was one of the reasons - if not the reason - that gave me the opportunity and confidence to tackle the tasks of a dissertation.
\setlength{\parskip}{1em}

\addcontentsline{toc}{chapter}{Abstract (Catalan/Spanish/English)} 

\begin{otherlanguage}{catalan}
	\cleardoublepage
	\chapter*{Resum}
	\vspace{-20mm}
	{\small
		L'aprenentatge de representacions no supervisat té com a objectiu trobar mètodes que aprenguin representacions a partir de dades sense senyals basats en anotacions. Abstindre's de les anotacions no només comporta beneficis econòmics, sinó que també pot, i en certa mesura ja ho fa, comportar avantatges en la estructura de la representació, la robustesa i la capacitat de generalització a diferents tasques. A llarg termini, s'espera que els mètodes no supervisats superin les seves contraparts supervisades a causa de la reducció de la intervenció humana i de l'enfocament inherentment més general que no biaixi l'optimització cap a un objectiu que prové de senyals específics basats en anotacions. Tot i que recentment s'han observat avantatges importants de l'aprenentatge de representacions no supervisat en el processament del llenguatge natural, els mètodes supervisats encara dominen en els dominis de la visió per a la majoria de les tasques. En aquesta tesi, contribuïm al camp de l'aprenentatge de representacions (visuals) no supervisades des de tres perspectives: \textit{(i) Aprenentatge de representacions:} Dissenyem Xarxes Neuronals Autoorganitzades Convolucionals (CSNNs) no supervisades i lliures de retropropagació que utilitzen regles d'aprenentatge basades en autoorganització i en Hebb, per aprendre nuclis convolucionals i màscares amb l'objectiu d'assolir models més profunds sense retropropagació. Observem que els mètodes basats en retropropagació i lliures de retropropagació poden patir d'una manca de coincidència de la funció objectiu entre la tasca de pretext no supervisada i la tasca objectiu, la qual cosa pot portar a una disminució en el rendiment per a la tasca objectiu. \textit{(ii) Avaluació de la representació:} Ens basem en el protocol d'avaluació (no) lineal àmpliament utilitzat per definir mètriques independents de la tasca de pretext i la tasca objectiu per a mesurar la manca de coincidència de la funció objectiu. Amb aquestes mètriques, avaluem diverses tasques de pretext i objectiu i revelem les dependències de la manca de coincidència de la funció objectiu en diferents parts de l'entrenament i la configuració del model. \textit{(iii) Transferència de representacions:} Contribuïm amb CARLANE, el primer banc de proves d'adaptació de domini sim-to-real de 3 vies per a la detecció de carrils 2D. Adoptem diversos mètodes coneguts d'adaptació de domini no supervisat com a referència i proposem un mètode basat en l'aprenentatge auto-supervisat prototípic entre dominis. Finalment, ens centrem en l'adaptació de domini no supervisat basada en píxels i contribuïm amb un mètode de traducció d'imatge a imatge no aparellat consistent en contingut que utilitza màscares, discriminadors globals i locals, i mostreig de similitud per mitigar les inconsistències de contingut, així com la denormalització atenta a característiques per fusionar estadístiques basades en contingut en la seqüència del generador. A més, proposem la mètrica cKVD per incorporar inconsistències de contingut específiques de classes en mètriques perceptuals per a mesurar la qualitat de la traducció.
		
		\vspace{1mm}
		\textbf{Paraules clau:} \textit{aprenentatge de representacions no supervisat, adaptació de domini no supervisada, traducció d'imatge a imatge, visió per computador}}	
\end{otherlanguage}

\begin{otherlanguage}{spanish}
	
\chapter*{Resumen}
\vspace{-20mm}

\begin{minipage}{\textwidth}
{\small 
El aprendizaje de representaciones no supervisado tiene como objetivo encontrar métodos que aprendan representaciones a partir de datos sin señales basadas en anotaciones. Abstenerse de las anotaciones no solo conlleva beneficios económicos, sino que también puede, y en cierta medida ya lo hace, resultar en ventajas en cuanto a la estructura de la representación, la robustez y la capacidad de generalización a diferentes tareas. A largo plazo, se espera que los métodos no supervisados superen a sus contrapartes supervisadas debido a la reducción de la intervención humana y al enfoque inherentemente más general que no sesga la optimización hacia un objetivo que proviene de señales específicas basadas en anotaciones. Si bien recientemente se han observado ventajas importantes del aprendizaje de representaciones no supervisadas en el procesamiento del lenguaje natural, los métodos supervisados todavía dominan en los dominios de la visión para la mayoría de las tareas. En esta tesis, contribuimos al campo del aprendizaje de representaciones (visuales) no supervisadas desde tres perspectivas: \textit{(i) Aprendizaje de representaciones:} Diseñamos Redes Neuronales Autoorganizadas Convolucionales (CSNNs) no supervisadas y libres de retropropagación que utilizan reglas de aprendizaje basadas en autoorganización y en Hebb, para aprender núcleos convolucionales y máscaras con el fin de lograr modelos más profundos sin retropropagación. Observamos que los métodos basados en retropropagación y libres de retropropagación pueden sufrir de una falta de coincidencia de la función objetivo entre la tarea de pretexto no supervisada y la tarea objetivo, lo que puede llevar a disminuciones en el rendimiento para la tarea objetivo. \textit{(ii) Evaluación de representación:} Nos basamos en el protocolo de evaluación (no) lineal ampliamente utilizado para definir métricas independientes de la tarea de pretexto y la tarea objetivo para medir la falta de coincidencia de la función objetivo. Con estas métricas, evaluamos varias tareas de pretexto y objetivo y revelamos las dependencias de la falta de coincidencia de la función objetivo en diferentes partes del entrenamiento y la configuración del modelo. Y \textit{(iii) Transferencia de representaciones:} Contribuimos con CARLANE, el primer banco de pruebas de adaptación de dominio sim-to-real de 3 vías para la detección de carriles 2D. Adoptamos varios métodos conocidos de adaptación de dominio no supervisado como referencia y proponemos un método basado en el aprendizaje auto-supervisado prototípico entre dominios. Por último, nos enfocamos en la adaptación de dominio no supervisada basada en píxeles y contribuimos con un método de traducción de imagen a imagen no emparejado consistente en contenido que utiliza máscaras, discriminadores globales y locales, y muestreo de similitud para mitigar las inconsistencias de contenido, así como la denormalización atenta a características para fusionar estadísticas basadas en contenido en la secuencia del generador. Además, proponemos la métrica cKVD para incorporar inconsistencias de contenido específicas de clases en métricas perceptuales para medir la calidad de la traducción.

\vspace{4mm}
\hspace*{4.6mm} \textbf{Palabras clave:} \textit{aprendizaje de representaciones no supervisada, adaptación de dominio no supervisado, traducción de imagen a imagen, visión por computador}}
\end{minipage}

\end{otherlanguage}

\chapter*{Abstract}
\vspace{-20mm}
Unsupervised representation learning aims at finding methods that learn representations from data without annotation-based signals. Abstaining from annotations not only leads to economic benefits but may - and to some extent already does - result in advantages regarding the representation's structure, robustness, and generalizability to different tasks. In the long run, unsupervised methods are expected to surpass their supervised counterparts due to the reduction of human intervention and the inherently more general setup that does not bias the optimization towards an objective originating from specific annotation-based signals. While major advantages of unsupervised representation learning have been recently observed in natural language processing, supervised methods still dominate in vision domains for most tasks. In this dissertation, we contribute to the field of unsupervised (visual) representation learning from three perspectives: \textit{(i) Learning representations:} We design unsupervised, backpropagation-free Convolutional Self-Organizing Neural Networks (CSNNs) that utilize self-organization- and Hebbian-based learning rules to learn convolutional kernels and masks to achieve deeper backpropagation-free models. Thereby, we observe that backpropagation-based and -free methods can suffer from an objective function mismatch between the unsupervised pretext task and the target task. This mismatch can lead to performance decreases for the target task. \textit{(ii) Evaluating representations:} We build upon the widely used (non-)linear evaluation protocol to define pretext- and target-objective-independent metrics for measuring the objective function mismatch. With these metrics, we evaluate various pretext and target tasks and disclose dependencies of the objective function mismatch concerning different parts of the training and model setup. \textit{(iii) Transferring representations:} We contribute CARLANE, the first 3-way sim-to-real domain adaptation benchmark for 2D lane detection. We adopt several well-known unsupervised domain adaptation methods as baselines and propose a method based on prototypical cross-domain self-supervised learning. Finally, we focus on pixel-based unsupervised domain adaptation and contribute a content-consistent unpaired image-to-image translation method that utilizes masks, global and local discriminators, and similarity sampling to mitigate content inconsistencies, as well as feature-attentive denormalization to fuse content-based statistics into the generator stream. In addition, we propose the cKVD metric to incorporate class-specific content inconsistencies into perceptual metrics for measuring translation quality. 

\vspace{1mm}
\textbf{Key words:} \textit{unsupervised representation learning, unsupervised domain adaptation, unpaired image-to-image translation, computer vision}

\tableofcontents
\cleardoublepage
\phantomsection

\addcontentsline{toc}{chapter}{List of Figures} 
\listoffigures
\cleardoublepage
\phantomsection

\addcontentsline{toc}{chapter}{List of Tables} 
\listoftables
\cleardoublepage
\phantomsection

\chapter{List of Abbreviations}

\begin{acronym}[SCLCAE]\itemsep3pt
	\acro{BMU}{Best Matching Unit}
	\acro{CAE}{Convolutional AutoEncoder}
	\acro{CCAE}{Convolutional AutoEncoder for Color restoration}
	\acro{cKVD}{class-specific Kernel VGG Distance}
	\acro{CNN}{Convolutional Neural Network}
	\acro{cOFM}{convergence Objective Function Mismatch}
	\acro{CSNN}{Convolutional Self-organizing Neural Network}
	\acro{cSM3}{convergence Soft Metrics Mismatch}
	\acro{DCAE}{Denoising Convolutional AutoEncoder}
	\acro{FADE}{Feature-ADaptive Denormalization}
	\acro{FATE}{Feature-ATtentive Denormalization}
	\acro{FN}{False Negatives}
	\acro{FP}{False Positives}
	\acro{GAN}{Generative Adversarial Network}
	\acro{GHA}{Generalized Hebbian Algorithm}
	\acro{GPU}{Graphics Processing Unit}
	\acro{KID}{Kernel Inception Distance}
	\acro{LA}{Lane Accuracy}	
	\acro{M3}{hard Metrics MisMatch}
	\acro{MLP}{MultiLayer Perceptron}
	\acro{MMD}{Maximum Mean Discrepancies}
	\acro{MM3}{Mean hard Metrics MisMatch}
	\acro{MOFM}{mean Objective Function Mismatch}
	\acro{MSM3}{Mean Soft Metrics Mismatch}
	\acro{OFM}{Objective Function Mismatch}
	\acro{PCA}{Principal Component Analysis}
	\acro{RCAE}{Convolutional AutoEncoder for Rotation prediction}
	\acro{RQ}{Research Question}
	\acro{SCLCAE}{Convolutional AutoEncoder with the Simple Contrastive learning approach from Chen et al. \cite{chen2020simple}}	
	\acro{sconv}{self-organizing convolution}
	\acro{sKVD}{semantically aligned Kernel VGG Distance}
	\acro{SM3}{Soft Metrics Mismatch}
	\acro{SOM}{Self-Organizing Map}
	\acro{SVM}{Support Vector Machine}
	\acro{t-SNE}{t-Distributed Stochastic Neighbor Embedding}
	\acro{UFLD}{False Negatives}
	\acro{VAE}{Variational AutoEncoder}
	\acro{ViT}{Visual image Transformer}
\end{acronym} 
\cleardoublepage
\phantomsection

\setlength{\parskip}{0em}

\mainmatter
\graphicspath{{./main/2_introduction/figures/}}

\chapter{Introduction}
\label{chap:intro}

\vspace{-10mm}
Unsupervised representation learning \cite{bengio2013representation} aims at finding methods that learn representations from unlabeled data, which can further be utilized to solve various target tasks. This stands in stark contrast to supervised learning, which relies primarily on carefully annotated data to learn target tasks directly from learning signals that pair samples with labels. Learning without explicit guidance from task-specific and biased signals offers several ad hoc benefits. Training data for unsupervised models is abundant in many domains and does not require a costly and time-consuming annotation process. This simplifies the training of large models on large datasets - a trend \cite{sevilla2022compute} that still results in performance gains. Furthermore, unsupervised learning can lead to benefits in setups where only few annotations are available \cite{asano2019critical,su2020does,newell2020useful}. For some target tasks, a large quantity of annotations is nearly impossible to acquire, like in medical image analysis \cite{litjens2017survey}. Additionally, there exist tasks for which it is unknown what to label in the data to solve them correctly and for which it should be considered not to acquire annotations at all, as in the case of anomaly detection \cite{chandola2009anomaly,chalapathy2019deep}. In general, what task or objective is necessary to learn the best representation of an environment is an open research question; this is applicable to both supervised and unsupervised learning. However, unsupervised learning is inherently a more general approach since the representation is learned solely from the environment itself. Taking it a step further, a rich unsupervised representation obviates the need to design complex task-specific mechanisms at all, as shown by representations of large natural language models \cite{min2021recent}. Here, the model is prompted with the task itself at inference time. To put it in engineering terms: The best part is no part. 

From an empirical point of view, experiments show that unsupervised representations are more robust against changes in the environment and adversarial attacks \cite{hendrycks2019using,bordes2021high}, can lead to improved out-of-distribution detection \cite{hendrycks2019using}, and generalize to various target tasks \cite{ericsson2021well,zhao2020makes}. However, especially the empirical analysis of the generalization to various target tasks and domains is still in an early stage.

Regardless of the advantages and promises of unsupervised representation learning, the history of machine learning more tangibly shows that the increasing self-sufficient creation of representations - first by domain-specific human engineering, then by learning rules applied to carefully curated features or datasets and subsequently by learning hierarchical, deep models directly on large data corpora - benefits the performance of machine learning algorithms. Since all of these steps have removed a certain amount of human intervention and each step has led to performance improvements, it is natural to remove even more human intervention to find algorithms that learn representations directly from pure data. 
 \begin{figure}[t]
	\begin{center}
		\includegraphics[width=0.9\linewidth]{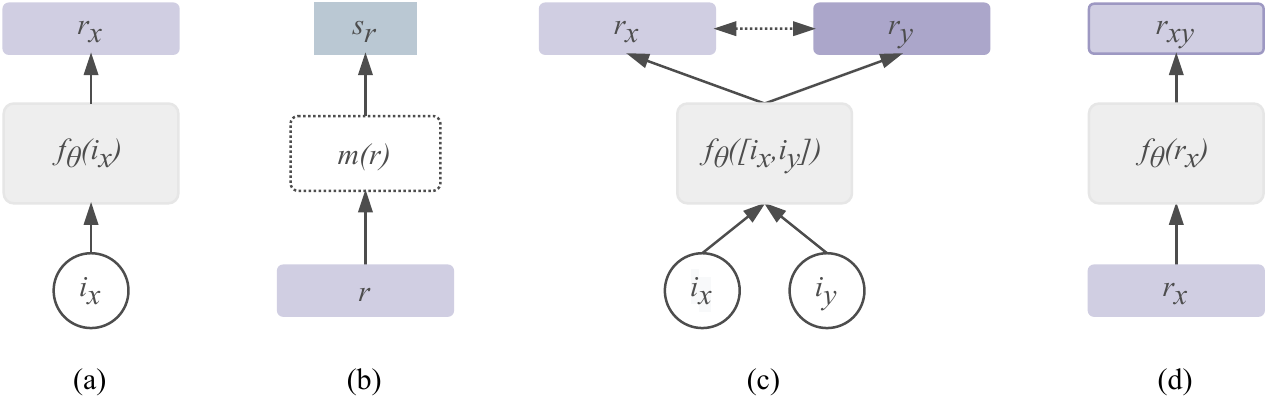}
	\end{center}
	\vspace{-1ex}
	\caption[High-level conceptual overview of examined and proposed methods and metrics.]{High-level conceptual overview of examined and proposed methods and metrics. a) Unsupervised learning of representations. b) Evaluation of unsupervised representation learning. c\&d) Transfer of unsupervised representations either by adapting representations of different domains to another (c) or by translating a representation from one domain to another (d).}
	\label{fig:introduction:scope_overview}
\end{figure}

To date, there is still human intervention and bias in unsupervised learning objectives and architectures. Prominent examples are defined transformations or augmentations for positive and negative pairs in contrastive learning \cite{bromley1993signature, hadsell2006dimensionality, dosovitskiy2014discriminative, wu2018unsupervised, oord2018representation, ye2019unsupervised, Misra_2020_CVPR, he2020momentum, chen2020simple, chen2020big, li2020prototypical, tian2020contrastive, dangovski2021equivariant, kumar2022contrastive} or distillation-based methods \cite{bachman2019learning, grill2020bootstrap, caron2020unsupervised, gidaris2020learning, gidaris2021obow, chen2021exploring, caron2021emerging, zbontar2021barlow, bardes2021vicreg, bardes2022vicregl}. As we show in our experiments in \autoref{chap:01}, biases in the design of methods can lead to a mismatch between the unsupervised learning objective and the desired target tasks, which results in a loss of performance. Recent work has tried to mitigate these human interventions and mismatches, for example, by designing methods that do not rely on hand-designed augmentations of the input, like joint-embedding predictive architectures \cite{assran2023self}. Regardless of these hurdles, unsupervised methods are already superior to supervised methods in some fields - most notably in natural language processing \cite{min2021recent}. Although unsupervised vision models have begun to outperform supervised models in certain situations \cite{hendrycks2019using,newell2020useful,su2020does,ericsson2021well,zhao2020makes}, the decisive breakthrough is still to come. 
\\\\
This work is located in the field of unsupervised learning of visual representations and aims at taking unsupervised learning a few steps further. As shown in \autoref{fig:introduction:scope_overview} and discussed in the following sections, we learn representations from data (a), define measures to evaluate unsupervised representation learning (b), adapt representations for different visual domains (c), and transfer visual representations to other visual domains (d), all without much human intervention in an unsupervised manner. 


\section{Unsupervised Learning of Visual Representations}
\label{sec:intro:ulovr}
Unsupervised visual representation learning aims at finding methods that learn representations of unlabeled data from vision modalities, such as images and videos.
As shown in \autoref{fig:introduction:scope_overview} (a), methods of this field commonly learn a parameterized function $f_\theta(i_x)$ - the model - in an unsupervised manner. This model infers a representation $r_x$ from a visual input $i_x$. $r_x$ is later utilized to solve target tasks. Prominent vision-based target tasks are classification, object detection, or segmentation \cite{jing2020self,chen2022semi}. 

There are several ways to train a model in an unsupervised manner. Recently, especially self-supervised methods have achieved promising results. There exists no standardized definition that distinguishes self-supervised learning from unsupervised learning. However, self-supervised learning is seen as a subfield of unsupervised learning. Even though the line between unsupervised and self-supervised learning is thin, and it can be argued that all unsupervised models are self-supervised in some sense, we follow \cite{ericsson2022self} to distinguish both learning schemes:
\\\\
\textbf{Unsupervised learning} \cite{xu2015comprehensive,espadoto2019toward,ghojogh2021restricted,abukmeil2021survey,ericsson2022self,kumar2022contrastive,chen2022semi,eddahmani2023unsupervised} methods learn directly from unlabeled data. Common learning objectives build generative models, such as
variational autoencoders (\acs{VAE}s) \cite{kingma2013auto} or generative adversarial networks (\acs{GAN}s) \cite{goodfellow2014generative,vincent2008extracting,radford2015unsupervised,pathak2016context,zhang2016colorful,larsson2016learning,larsson2017colorization,kim2018learning,chen2019self,donahue2019large,charte2020analysis,wolf2020instance,he2022masked}, train a density estimator like gaussian mixture models \cite{mclachlan2019finite}, or compress the input into a representation, for example, via autoencoders \cite{rumelhart1985learning} or classical clustering methods \cite{xu2015comprehensive}. 
\\\\
\textbf{Self-supervised learning} \cite{schmidhuber:1987:srl,schmidhuber1990making,jing2020self,fang2022connecting,ericsson2022self,kumar2022contrastive} methods learn from a signal that pairs a self-generated label obtained from the data and the data itself. Early examples learn by predicting rotations or other transformations applied to an input image \cite{gidaris2018unsupervised,chen2019self,zhang2019aet,qi2019avt,lin2021auto} or by solving jigsaw puzzles obtained from an image \cite{noroozi2016unsupervised,kim2018learning}. Recent work has defined objectives most notably based on clustered features \cite{yang2016joint,bautista2016cliquecnn,caron2018deep,asano2019self,zhuang2019local,zhan2020online,gidaris2020learning,gidaris2021obow,caron2020unsupervised,li2022self, assran2022masked}, contrastive learning \cite{bromley1993signature, hadsell2006dimensionality, dosovitskiy2014discriminative, wu2018unsupervised, oord2018representation, ye2019unsupervised, Misra_2020_CVPR, he2020momentum, chen2020simple, chen2020big, li2020prototypical, tian2020contrastive, dangovski2021equivariant, kumar2022contrastive}, distillation \cite{bachman2019learning, grill2020bootstrap, caron2020unsupervised, gidaris2020learning, gidaris2021obow, chen2021exploring, caron2021emerging, zbontar2021barlow, bardes2021vicreg, bardes2022vicregl}, or feature predictive distillation  \cite{assran2022masked, baevski2022data2vec, baevski2022efficient, assran2023self}. Compared to unsupervised learning, self-supervised learning uses a more specific discriminative, but often less general, training objective. Self-supervised learning is a subfield of unsupervised learning.
\\\\
Tasks used to train unsupervised models are often referred to as pretext tasks.
Unsupervised learning aims at finding methods that inherently solve pretext tasks that are general enough to learn a perfect representation of the environment. Although there are efforts to find general methods for various pretext tasks and modalities \cite{baevski2022data2vec,baevski2022efficient}, methods today often exploit knowledge about the data modality used for training and therefore tend to be biased towards this modality and the used exploit. For example, as we show in \autoref{chap:01}, transformation-based pretext tasks tend to retain spatial information in the representation while losing information about other properties like color.

\section{Evaluation of Unsupervised Representation Learning}
Since one goal of unsupervised representation learning is finding methods that learn general representations, evaluating these methods is not trivial. Most work in this area aims at learning "useful" or "good" representations, where these terms are meant to describe various beneficial properties of the representation, such as transferability to different target tasks. For example, good representations are defined to be smooth, sparse, specific, spatially and temporally coherent, transferable to supervised learning, and have multiple hierarchical explanatory factors with simple dependencies that generalize across tasks \cite{bengio2013representation}. Moreover, the features of the representations form categorical, well-separated, variation-robust manifolds for which the concentration of probability mass can lead to a smaller dimensionality than the input space \cite{bengio2013representation}. However, to date, there exists no metric or property that reliably predicts the performance of the representation on different target tasks \cite{locatello2019challenging,locatello2020sober,cabannes2023ssl}. Therefore, one vision is to find a general measure that reliably predicts the quality and performance of the representation for different target tasks with little overhead during the training of the unsupervised model. Since the output of unsupervised representation learning is a representation $r$, a (current) metric $m$ often works on $r$ and produces a measurement $s_r$ for assessment, as shown in \autoref{fig:introduction:scope_overview}. 
\\\\
\textbf{The linear and nonlinear evaluation protocols} are to date the most commonly used protocols for quantitative evaluation. Thereby, an untrained linear or non-linear classifier (probe) is introduced on top of the representations of $f_\theta$. This probe is then trained in an supervised manner on the target task for a part of the training dataset or for the entire training dataset. Thereby, the weights of $f_\theta$ are either frozen or fine-tuned on the target task. Afterwards, the performance on the target task is compared with other methods. To achieve a good measure with these protocols, a careful selection of target tasks is required. When the model is fine-tuned on the target task the unsupervised learning stage is called unsupervised pretraining. This pretraining is often compared with other weight initialization schemes for target task training. 
\\\\
Other ways to evaluate representations is by examining their properties empirically or visually. An example is \cite{hendrycks2019using}, in which ImageNet \cite{deng2009imagenet} is annotated with factors of variation to quantitatively examine if unsupervised models can predict these factors. Another example is \cite{bordes2021high}, in which a generative model is conditioned on the unsupervised representation to generate images for quantitative visual examination of the representation's properties. Quantitatively and qualitatively examining representations aids the research process and increases the interpretabillity of methods.

Most of the recent work has focused on examining the representation at the end of training or at a certain training step. In \autoref{chap:01} we examine the entire unsupervised training process by measuring if the pretext objective is able to generalize to various target tasks during training, or if it mismatches with them after a certain amount of training steps and therefore leads to lower target task performance. 

\section{Unsupervised Adaptation of Visual Representations}
Unsupervised adaptation of visual representations aims at adapting representations to different domains. As shown in \autoref{fig:introduction:scope_overview}, this can be done by training a model $f_\theta$ that learns a common latent space over different domains $x$ and $y$, where representations $r_x$ and $r_y$ of inputs $i_x$ and $i_y$ share properties. However, this is not trivial because different domain statistics lead to a domain gap, also called domain shift, between these domains, which harms training and performance \cite{saenko2010adapting}.
\\\\
\textbf{Unsupervised domain adaptation} \cite{wilson2020survey,farahani2021brief,zhang2021survey,liu2022deep} is a field that aims at adapting representations in an unsupervised manner and is our focus in \autoref{chap:02}. Thereby, a model trained on one or multiple labeled source domain(s) is adapted to one or multiple unlabeled target domain(s). The high-level relation to plain unsupervised learning is three-fold: First, both unsupervised learning and unsupervised domain adaptation methods aim at learning general representations for different target tasks. Second, both unsupervised learning and unsupervised domain adaptation can be seen as special cases of transfer learning. Transfer learning generally aims at utilizing the knowledge gained during training for other tasks and domains \cite{bengio2013representation,pan2010survey,liu2022deep}. Since unsupervised learning aims at learning representations that are general across tasks and domains, it is a special case of transfer learning where either source nor target domains are labeled \cite{bengio2013representation,pan2010survey,liu2022deep}. Unsupervised domain adaptation is a special case of transfer learning where the source is labeled but not the target domain(s). Third, unsupervised domain adaptation performs unsupervised learning in the target domain and often leverages unsupervised auxiliary tasks for both domains \cite{wilson2020survey,farahani2021brief,zhang2021survey,liu2022deep}.

Unsupervised domain adaptation enables the training of domain invariant models, which leads to several benefits similar to unsupervised learning: Unsupervised domain adaptation reduces extensive data annotation, since domains for which a massive amount of labeled data is obtainable without much effort (e.g., simulations) can be utilized to guide the training for the desired task in the target domain. Furthermore, only one model needs to be trained for different domains, which leads to a reduction in overall complexity. In addition, a major assumption of many learning strategies is that training and testing data share the same distribution. This assumption can easily be violated (over time) if there are subtle differences in the subsets, such as different backgrounds or deformations, or variations in sample quality. This is known as covariate shift \cite{shimodaira2000improving}. Therefore, an adaptation or far-ranging generalization may be required at inference time. 

Although there have been significant advances in unsupervised domain adaptation, much of this work focuses on simple classification tasks, the transfer to a single target domain, or do not share a common benchmark due to the lack of these. In \autoref{chap:02}, we propose a benchmark and a method for 3-way sim-to-real domain adaptation for 2D lane detection.

\section{Unsupervised Translation of Visual Representations}
\label{sec:intro:utovr}
As shown in \autoref{fig:introduction:scope_overview}, the unsupervised translation of visual representations focuses on learning a domain mapping $f_\theta(r_x)$ of representations $r_x$ from one domain to another by inferring a translated representation $r_{xy}$ of $r_x$ for the other domain. This can be seen as a special case of unsupervised domain adaptation, which generally aims at learning a domain mapping between different domains, but is allowed to adapt the learned representations of all domains to another \cite{wilson2020survey}. In contrast, the translation of representations preserves certain properties of the source domain, such as the content of the image, while adapting other properties to the target domain, such as the style of the image.
\\\\
\textbf{Unpaired image-to-image translation.} \cite{alotaibi2020deep,pang2021image} In our work in \autoref{chap:03}, we focus on unsupervised image-to-image translation, which is also called unpaired image-to-image translation \cite{alotaibi2020deep,pang2021image}. Thereby, a model is trained to translate images from a source domain to a target domain without seeing concrete pairs in both domains. Therefore, unpaired image-to-image translation can be seen as pixel-level unsupervised domain adaptation \cite{wilson2020survey,farahani2021brief}. There are many prominent applications like photorealism, translation of daytime and seasons, or style transfer \cite{alotaibi2020deep,pang2021image}. In each of these applications, it is either difficult or impossible to collect true pairs of samples showing the same scene with the desired variation to utilize them in a supervised setting. In contrast, paired image-to-image translation focuses on use cases where paired samples can be easily acquired. Examples are super-resolution, where a low-resolution image is translated to a high-resolution image, or image synthesis, where an image is synthesized from basic representations such as segmentation maps \cite{alotaibi2020deep,pang2021image}. We focus on unpaired image-to-image translation in this work. Besides the concrete image-based use cases and domain adaptation use cases, the controllable creation of more data in specific domains is another benefit. However, since unpaired image-to-image translation methods work at a pixel level, it is hard to evaluate the translation quality empirically. Different widely-adopted metrics have been proposed, which use perceptual metrics to compare the target images with the translated source images \cite{alotaibi2020deep,pang2021image}. However, one open research question is the content consistency of translated images. Biases between datasets often lead to inconsistency, like hallucinations in the translated source images \cite{richter2022enhancing}. For example, if there are more trees at the top half of the target domain's images than in the source domain's images, the model will hallucinate trees into the translated source images to fulfill the source-to-target translation objective. Creating non-ill-posed constraints that enforce content consistency and creating metrics that adequately incorporate the negative effect of content-inconsistency are still open research areas. Therefore, we propose a content-consistent unpaired image-to-image translation method and a metric that measures inconsistency based on a widely used perceptual metric in \autoref{chap:03}.




\section{Objectives and Scope}
This dissertation aims at creating novel methods to learn, evaluate, and transfer visual representations in the unsupervised representation learning field. Therewith, we want to contribute to the general goal of developing methods that lead to useful representations from data without much human intervention. The scope of this Ph.D. thesis focuses on image-based visual domains. These domains are often - but not exclusively - located in the fields of driving assistance or autonomous driving. We base the construction of our methods, metrics, and benchmarks on the following research questions (\acs{RQ}s): 
\\\\
\textbf{Learning representations}:
\begin{enumerate}
\item \acs{RQ}-L1: Is it possible to create an unsupervised, backpropagation-free method with modules that are currently considered more biologically plausible than backpropagation-based methods? \label{RQ-L1}
\item \acs{RQ}-L2: Can we extend our unsupervised, backpropagation-free model to a larger model?  \label{RQ-L2}
\end{enumerate}

\noindent \textbf{Evaluating representations}:
\begin{enumerate}
\item \acs{RQ}-E1: How well does our unsupervised backpropagation-free method perform compared to backpropagation-free and backpropagation-based methods? \label{RQ-E1}
\item \acs{RQ}-E2: Can we design metrics to measure the objective function mismatch between unsupervised pretext models and (supervised) target models trained for different target tasks? \label{RQ-E2}
\item \acs{RQ}-E3: Does the objective function mismatch depend on specific parts of the training setup and model? \label{RQ-E3}
\item \acs{RQ}-E4: What effect does the objective function have on target task performance? \label{RQ-E4}
\item \acs{RQ}-E5: Can we create a 3-way sim-to-real domain adaptation benchmark for 2D lane detection to evaluate unsupervised domain adaptation for transferring models trained in simulations to multiple real-world domains? \label{RQ-E5}
\item \acs{RQ}-E6: How well do our unsupervised domain adaptation method and other state-of-the-art methods perform on our 3-way sim-to-real domain adaptation benchmark? \label{RQ-E6}
\item \acs{RQ}-E7: How can we incorporate the content inconsistencies in the translated images into the evaluation of unpaired image-to-image translation methods? \label{RQ-E7} \vspace{-4mm}
\item \acs{RQ}-E8: How well does our unpaired image-to-image translation method perform compared to the state of the art? \label{RQ-E8}
\end{enumerate}

\noindent \textbf{Transferring representations}:
\begin{enumerate}
\item \acs{RQ}-T1: Can we create and train a single-source, multi-target model for our 3-way sim-to-real domain adaptation benchmark and adapt (single) models to multiple real-world domains? \label{RQ-T1}
\item \acs{RQ}-T2: Can we utilize masks to create an unpaired image-to-image translation method that mitigates content inconsistencies when translating images between biased datasets? \label{RQ-T2}
\item \acs{RQ}-T3: Can we improve the incorporation of content features into the generator by extending feature-adaptive denormalization with an attention mechanism? \label{RQ-T3}
\end{enumerate}
With the contributions presented in this dissertation, we hope to shed more light on the answers to these questions and to reduce the uncertainty that surrounds them.
\section{Outline} 
\label{sec:intro:outline}
\textbf{\autoref{chap:related}} describes the related work with respect to our contributions: In \autoref{sec:realtedwork:unsupervisedlearning} the related work for unsupervised visual representation learning is described. In \autoref{sec:realtedwork:unsupervisedlearningevaluation} the related work for the evaluation of (visual) unsupervised learning is described. In \autoref{sec:realtedwork:unsuperviseddomainadaptation} the related work for image-based unsupervised domain adaptation is described, and in \autoref{sec:realtedwork:unsupervisedtranslation} the related work for unpaired image-to-image translation is described. Following that, this dissertation is divided into the following five chapters: 
\\\\
\textbf{\autoref{chap:00}} deals with visual unsupervised representation learning. Backpropagation-free Convolutional Self-Organizing Neural Networks (\acs{CSNN}s) are proposed, which learn representations via self-organization-based clustering and local masks. The learning rules for our models differ strongly from backpropagation-based \acs{CNN}s \cite{lecun1989backpropagation}, while the structures of the networks are similar. The learned representations of our models are tested on various classification downstream tasks to achieve a critical and insightful look into the field of unsupervised representation learning. At the time of publication, the performance of our models was comparable with the state-of-the-art. More notably, we have become aware of the objective function mismatch (\acs{OFM}). 
\\\\
\textbf{\autoref{chap:01}}. Our work in \autoref{chap:00} and other research shows various problems of unsupervised representation learning. One problem is the objective function mismatch, which states that the performance on a desired target task can decrease when the unsupervised pretext task is learned for too long - especially when both tasks are ill-posed. We propose metrics to measure this mismatch in a comparable manner and evaluate state-of-the-art methods at the time of publication. Thereby, we find that each of these methods can suffer from the objective function mismatch. We disclose dependencies of this mismatch across several pretext and target tasks with respect to the pretext model's representation size, target model complexity, pretext and target augmentations, as well as pretext and target task types. 
\\\\
\textbf{\autoref{chap:02}}. Within the framework of a project to build a self-driving model vehicle, we transfer lane detection models from simulation to the real world with unsupervised domain adaptation methods. Although unsupervised domain adaptation has been applied to a variety of complex vision task, we find that not much of this work focuses on lane detection in autonomous driving. This can be attributed to the lack of publicly available datasets. In \autoref{chap:02} we describe CARLANE, a 3-way sim-to-real unsupervised domain adaptation benchmark for 2D lane detection. CARLANE encompasses the single-target datasets MoLane and TuLane and the multi-target dataset MuLane. We evaluate several well-known unsupervised domain adaptation methods as systematic baselines and propose our own method that achieves state-of-the-art performance. 
\\\\
\textbf{\autoref{chap:03}}. In this chapter, we focus on mitigating content inconsistencies arising in unpaired image-to-image translation. We show that masking the discriminator's input based on content is sufficient to reduce these inconsistencies significantly. Our masking procedure leads to more minor artifacts, which we significantly reduce by introducing a local discriminator and a similarity sampling technique. We also design a feature-based denormalization block that incorporates content features into the generator stream by attending to features correlating with a specific property, such as features of shadows. To enable a finer-grained measure of translation quality, we propose the cKVD metric to examine translated images at the class or category level. Our approach results in state-of-the-art performance for photo-realistic sim-to-real and weather translation and performs well for day-to-night translation. 
\\\\
\textbf{\autoref{chap:end}} concludes the dissertation with a summary of our contributions alongside a list of publications, a list of contributed code and datasets, and a list of awards received during the Ph.D. period. Furthermore, we discuss developments, current limitations, and ethics of unsupervised visual representation learning and give future perspectives on this field.

\graphicspath{{./main/3_related_work/figures/}}

\chapter{Related Work}
\label{chap:related}
\vspace{-10mm}

\section{Unsupervised Learning of Visual Representations}
\label{sec:realtedwork:unsupervisedlearning}
Various works categorize unsupervised representation learning by the underlying model architecture \cite{assran2023self}, the definition of the objective function, or a generic description of the learning objective \cite{ericsson2022self, bardes2021vicreg, kumar2022contrastive,chen2022semi}. However, since this field is very active, there is a constant change in categorization. Furthermore, categorization depends on the perspective of the respective work. Here, we categorize unsupervised methods by generic descriptions of image-based learning objectives since this results in a useful structure for our work.
\\\\
\textbf{Spatial patch-based methods} \cite{doersch2015unsupervised,noroozi2016unsupervised,kim2018learning,dai2021up,zhai2022position,caron2022location} try to leverage the knowledge obtained by predicting the spatial relationship of patches. Early examples define pretext tasks by predicting the relative position of a patch in a grid around another patch \cite{doersch2015unsupervised} or by solving jigsaw puzzles \cite{noroozi2016unsupervised,kim2018learning}, where the relative position of each patch is predicted. In \cite{dai2021up} a transformer-based model is trained to localize a random query patch in an image, given the entire image as context. Building upon this work, a transformer-based model is subsequently trained to predict the position of all patches in an image, given a ratio of random context patches from the image as input \cite{zhai2022position}. Recent work has proposed to predict the relative position of each patch in a view of the image with respect to a different reference view of the same image \cite{caron2022location}. 
\\\\
\textbf{Transformation prediction} \cite{gidaris2018unsupervised,chen2019self,zhang2019aet,qi2019avt,lin2021auto}: Spatial context can also be encoded by predicting transformations, which has led to a line of research focusing on autoencoding transformations rather than data. Early work predicts simple rotations of an input image \cite{gidaris2018unsupervised, chen2019self}, while more recent work has focused on complex transformations by minimizing different geometric distances between target and predicted transformation matrices \cite{zhang2019aet,qi2019avt,lin2021auto}.
\\\\
\textbf{Generation-based methods} \cite{vincent2008extracting,radford2015unsupervised,pathak2016context,zhang2016colorful,larsson2016learning,larsson2017colorization,kim2018learning,chen2019self,donahue2019large,charte2020analysis,wolf2020instance,he2022masked} examine the generation of an arbitrary output from a learned representation of the given input. One line of work improves on autoencoders \cite{6302929} and variational autoencoders \cite{kingma2013auto} by defining generation-based pretext tasks, which lead to representations valuable for target tasks. Examples are denoising \cite{vincent2008extracting,charte2020analysis}, colorization \cite{zhang2016colorful,larsson2016learning,larsson2017colorization, kim2018learning}, or inpainting of images \cite{pathak2016context,wolf2020instance, kim2018learning}. A second line of work based on GANs \cite{goodfellow2014generative} adjusts their latent space for representation learning, for example, by constraining \cite{radford2015unsupervised} or changing \cite{donahue2019large} the architecture. Recently, large masked autoencoders have achieved promising results with an asymmetric encoder-decoder architecture and a high masking ratio \cite{he2022masked}.
\\\\
\textbf{Cluster-based methods} \cite{yang2016joint,bautista2016cliquecnn,caron2018deep,asano2019self,zhuang2019local,zhan2020online,gidaris2020learning,gidaris2021obow,caron2020unsupervised,li2022self, assran2022masked} cluster representations and use their cluster assignments as pseudo labels for the learning signal. Thereby, the clustering objective is used as a proxy task to calculate pseudo labels. In contrast to classical clustering methods, the goal is to learn a good representation instead of cluster assignments. Early approaches train the model by alternating between two steps: First, the cluster assignments of the model's representations are optimized by a clustering method. Second, the model is trained with the pseudo labels obtained from the assignments. Well-known clustering objectives use agglomerative clustering \cite{yang2016joint}, k-means \cite{caron2018deep,gidaris2020learning,gidaris2021obow}, anchor neighborhood discovery \cite{huang2019unsupervised}, or maximize the information between data indices and equipartitioned pseudo labels \cite{asano2019self}. To avoid the costly two-step procedure, recent work has utilized online clustering methods which simultaneously update the clusters while the model is training. Most notably, methods have achieved online learning by classifying clustered batches \cite{bautista2016cliquecnn}, utilizing memory modulus for representations \cite{zhuang2019local, zhan2020online} and centroids \cite{zhan2020online}, assigning view-invariant codes to learned prototypes \cite{caron2020unsupervised, assran2022masked}, or with self-organizing layers \cite{li2022self}. 
\\\\
\textbf{Contrastive methods} \cite{bromley1993signature, hadsell2006dimensionality, dosovitskiy2014discriminative, wu2018unsupervised, oord2018representation, ye2019unsupervised, Misra_2020_CVPR, he2020momentum, chen2020simple, chen2020big, li2020prototypical, tian2020contrastive, dangovski2021equivariant, kumar2022contrastive} utilize context information between negative and positive pairs. Thereby different views are sampled from the inputs and the model tries to predict which views belong to each other (positives) and which views are different (negatives). Representations of positive views are brought together, while representations of negative views are pushed away. The definition of negatives and positives depends on the method. Examples of positive and negative pairs are augmentations of the same images as positives and of different images as negatives \cite{dosovitskiy2014discriminative, ye2019unsupervised, Misra_2020_CVPR, he2020momentum, chen2020simple, chen2020big, chen2020improved}, different sensory views of an image as positives and a random image as a negative \cite{tian2020contrastive}, as well as corresponding prototypes of samples as positives and vice versa \cite{li2020prototypical}. Furthermore, methods explore, e.g., different similarity and objective functions \cite{bromley1993signature, hadsell2006dimensionality, dosovitskiy2014discriminative, oord2018representation, ye2019unsupervised, Misra_2020_CVPR, li2020prototypical}, different transformations to create the views \cite{hadsell2006dimensionality, dosovitskiy2014discriminative, chen2020simple, li2020prototypical, dangovski2021equivariant}, or different model architectures \cite{dosovitskiy2014discriminative, oord2018representation, he2020momentum, chen2020simple, chen2020big}. Siamese networks, in which the encoder for positives and negatives shares weights, are the most commonly used architecture \cite{bromley1993signature, hadsell2006dimensionality,dosovitskiy2014discriminative, ye2019unsupervised, Misra_2020_CVPR, he2020momentum, chen2020simple, chen2020big}. Contrastive methods require large numbers of contrastive pairs to work well; therefore many techniques use large batch sizes \cite{chen2020simple, chen2020big} or memory banks \cite{wu2018unsupervised, ye2019unsupervised, he2020momentum, chen2020improved, chen2020big}. The downsides of high memory requirements has led to the exploration of alternatives like distillation-based methods.
\\\\
\textbf{Distillation-based methods} \cite{bachman2019learning, grill2020bootstrap, caron2020unsupervised, gidaris2020learning, gidaris2021obow, chen2021exploring, caron2021emerging, zbontar2021barlow, bardes2021vicreg, bardes2022vicregl} are inspired by knowledge distillation \cite{hinton2015distilling} and build upon the success of contrastive methods. Distillation-based methods make use of a student and a teacher model. Both models get different augmented views of the same image as input, and the student model tries to predict the representation of the teacher model. Often the student model is updated via backpropagation \cite{linnainmaa1970representation,werbos1974beyond,rumelhart1986learning}, while the teacher model is updated with a moving average of the student model's parameters \cite{grill2020bootstrap, caron2021emerging, gidaris2021obow}. There is work that shows that the moving average update is not needed when simple weight-sharing between both models is combined with a stop-gradient operation for the teacher branch \cite{chen2021exploring} or when both branches are updated with backpropagation simultaneously \cite{caron2020unsupervised, zbontar2021barlow, bardes2021vicreg, bardes2022vicregl}. It has been shown that distillation-based methods, unlike most contrastive methods, do not require large batch sizes to perform well \cite{grill2020bootstrap, chen2021exploring}. One reason for this is that distillation-based methods capture only positive pairs. However, the student and teacher model need to be prevented from collapsing to a static representation (mode collabs). There are different tricks to mitigate collabs, like mutual information maximization with a global summary feature vector \cite{bachman2019learning}, a predictor head on top of the student model \cite{grill2020bootstrap, chen2021exploring}, a stop-gradient operation for the teacher model \cite{chen2021exploring}, clustering constraints \cite{caron2020unsupervised}, or by redundancy-reduction in the representations of different views via regularization \cite{zbontar2021barlow, bardes2021vicreg, bardes2022vicregl}. It was hypothesized that batch normalization is needed to prevent collapse \cite{tian2020understanding}, but it has been shown that batch normalization can be replaced by other normalizations \cite{richemond2020byol} or can be left out entirely for the prediction head \cite{chen2021exploring}. In \cite{caron2021emerging} it has been shown that only the centering and sharpening of the teacher output is enough to prevent collapse for visual image transformers (\acs{ViT}s) \cite{dosovitskiy2020image}. However, besides the effectiveness of these methods, there is no clear understanding of how they avoid collapse and why they perform so well. In further work, different forms of representations are predicted by the student model, such as codes obtained from prototypes \cite{caron2020unsupervised}, a bag of visual words \cite{gidaris2020learning, gidaris2021obow}, or local and global features \cite{bardes2022vicregl}. 
\\\\
\textbf{Feature predictive distillation} \cite{assran2022masked, baevski2022data2vec, baevski2022efficient, assran2023self} builds upon the success of distillation-based methods and borrow their setup. Given a masked view, the student model tries to predict the teacher model's representation of a different, potentially masked view of the same input. Recent methods differ most notably in how they mask the input, what part of the teacher model's representation they predict, and whether other augmentations are applied to the input besides masking. Examples of masking strategies are randomly masking patches \cite{assran2022masked} and different block masking techniques \cite{baevski2022data2vec, baevski2022efficient, assran2023self}. Methods predict the representation of the entire input of the teacher model \cite{assran2022masked, baevski2022data2vec} or different target blocks \cite{baevski2022efficient, assran2023self}. Most of these methods do not use hand-crafted data augmentations in addition to masking to create a view; this removes much human intervention \cite{baevski2022data2vec,baevski2022efficient,assran2023self}. In \cite{assran2023self} a predictor head conditioned on (latent) variables is introduced to facilitate feature prediction.
\\\\
\textbf{Backpropagation-free methods}. While there have been many advances in unsupervised methods that learn with backpropagation, in comparison, there have been only a few efforts to propose backpropagation-free alternatives. Early classical clustering algorithms \cite{xu2015comprehensive} like variants of k-means \cite{macqueen1967classification}, Expectation Maximization \cite{dempster1977maximum}, or Self-Organizing Maps (\acs{SOM}s) \cite{kohonen1982self} are a form of unsupervised representation learning. Dimensionality reduction methods \cite{espadoto2019toward} like Principal Component Analysis (\acs{PCA}) \cite{pearson1901liii} and t-Distributed Stochastic Neighbor Embedding (\acs{t-SNE}) \cite{van2008visualizing} can also be seen as unsupervised representation learning methods. Furthermore, different variants and learning rules for Boltzmann machines \cite{hinton1983optimal,ackley1985learning,holyoak1987parallel,ghojogh2021restricted} have been explored. 

Building up on these classical methods, weights are learned as convolutional filters in \acs{CNN}-like architectures with one or few learnable layers. In \cite{coates2011analysis} different algorithms (e.g., k-means) and hyperparameters have been studied for single-layer networks, and general performance gains with an increasing representation size have been observed. Other examples of single-layer models use \acs{SOM} variants \cite{kohonen1982self,zheng2008fast,hankins2018towards} to learn convolutional filters or a spatial image pyramid to learn hierarchical representations with single-layer methods \cite{lazebnik2006beyond,agarwal2006hyperfeatures,boureau2010learning}. Examples of two-layer methods stack \acs{PCA} \cite{chan2015pcanet}, k-means \cite{dong2017cunet}, or SOM \cite{hankins2018somnet,hankins2018towards} convolutional layers. In \cite{grinberg2019local} two layers are trained in a convolutional manner with the Hebbian-like learning rule proposed in \cite{doi:10.1073/pnas.1820458116}. Some of these methods use nonlinear activation functions \cite{jarrett2009best, dong2017cunet, grinberg2019local}, normalization \cite{jarrett2009best,grinberg2019local}, and pooling layers between the two convolutional layers \cite{jarrett2009best, dong2017cunet, hankins2018towards,grinberg2019local}. Both one- and two-layer methods often use an output stage based on binary hashing and histograms to reduce dimensionality and increase nonlinearity, invariance, and robustness in the representation \cite{jarrett2009best,chan2015pcanet,dong2017cunet,hankins2018somnet,hankins2018towards}. Furthermore, there exists work on two-layer unsupervised spiking neural network models using spike-timing dependent plasticity learning rules \cite{saunders2019locally}.

In \cite{lin2014stable} k-means representations are encoded to a sparse representation to achieve a three-layer network. Furthermore, Boltzmann Machines can be stacked to Deep Boltzmann Machines with up to three non-convolutional layers \cite{pmlr-v5-salakhutdinov09a}. 

However, it has been found that simple stacking of these backpropagation-free layers with potential pooling, normalization, and encoding layers brings little to no benefit. Therefore, recent work has tried to find methods to connect layers efficiently or has searched directly for stackable learning rules. Most successfully, variants of biologically-inspired Hebbian-like learning rules \cite{hebb2005organization} are used in these methods to build multi-level models \cite{stuhr2019csnns, miconi2021hebbian, illing2021local, journe2022hebbian}. In our work \cite{stuhr2019csnns}, presented in \autoref{chap:00}, three multi-headed SOM layers are stacked with Hebbian-like learning rules in a convolutional manner. To the best of our knowledge, this is the first approach to combine the successful principles of CNNs, SOMs, and Hebbian Learning into a single architecture. In subsequent work, inspired by local contrastive learning \cite{lowe2019putting}, five-layer models are learned with a Hebbian-like learning rule to predict features of saccades \cite{illing2021local}. In \cite{miconi2021hebbian} up to three Hebbian-like layers have been stacked with a channel-wise hard winner-take-all selection for each layer. Further work improves over these methods by building three- to five-layer networks with a soft, softmax-based winner-take-all selection \cite{moraitis2022softhebb} as well as with other learning and architecture improvements like an adaptive learning rate per neuron \cite{journe2022hebbian}. 
\\\\
\textbf{Other methods} use meta-learning \cite{Schmidhuber95onlearning,hsu2018unsupervised}, self-supervised relational reasoning \cite{patacchiola2020self}, mutual information maximization \cite{hjelm2018learning,bachman2019learning,wu2020mutual,kumar2022contrastive}, or task-specific unsupervised setups \cite{godard2017unsupervised}. Furthermore, many methods link \cite{wu2020mutual, caron2020unsupervised} or combine multiple self-supervised approaches \cite{kim2018learning,chen2019self,gidaris2020learning,yan2020clusterfit,li2020prototypical,dangovski2021equivariant,gidaris2021obow,assran2022masked}.

\section{Evaluating Visual Unsupervised Representation\\Learning}
\label{sec:realtedwork:unsupervisedlearningevaluation}
\textbf{The Objective Function Mismatch in unsupervised learning} states that the performance on a desired target task can decrease over the course of training when the pretext task and the target tasks are ill-posed. Some work directly or indirectly observes that learning a pretext task too long may hurt target task performance but makes no further investigations on this topic \cite{locatello2019challenging,kolesnikov2019revisiting,wallace2020extending,locatello2020sober,chen2021ssl++}. Some early work shows performances of linear target models over training epochs but does not examine or define the objective function mismatch in detail \cite{zhai2019large,stuhr2019csnns}. Instead, unsupervised multi-task learning, meta-learning, or complementary features approaches have been proposed to lower the objective function mismatch \cite{doersch2017multi,metz2018meta,chen2021ssl++}. In contrast, in our work \cite{stuhr2022don} presented in \autoref{chap:01}, we focus solely on defining simple, general protocols for measuring mismatches from metrics over the course of pretext task training when a target task is trained afterward on top of the pretext model's representations. To the best of our knowledge, this has not been done before. Furthermore, we highlight important findings and properties of our evaluation protocols. 
\\\\
\textbf{Changing the underlying model} is a common theme when comparing unsupervised learning techniques \cite{goyal2019scaling,kolesnikov2019revisiting,chen2020simple,newell2020useful,cabannes2023ssl}. Thereby, the effects of changing the entire model architecture or its parts are examined. Various works observe that scaling the underlying model capacity leads to performance improvements \cite{goyal2019scaling, kolesnikov2019revisiting,chen2020simple,newell2020useful,cabannes2023ssl}, which seems crucial for larger dataset sizes \cite{goyal2019scaling}. A well-known finding is that a larger representation size consistently increases the quality of the learned visual representations for different target tasks \cite{kolesnikov2019revisiting,chen2020simple,cabannes2023ssl}. However, for the joint-embedding framework, it has been argued that a reduced model complexity through simple architectures like CNNs and regularization can promote "simple" representations (e.g., smooth) with a lower effective dimension \cite{cabannes2023ssl}. In our work in \autoref{chap:01}, we also change the underlying model to examine the effects on the objective function mismatch.
\\\\
\textbf{Varying the amount of data samples} leads to interesting observations as well \cite{goyal2019scaling,asano2019critical,newell2020useful,su2020does}. Several works observe that more training data does not necessarily lead to target task performance improvements if the underlying model has a low capacity or the pretext task is not suitable \cite{newell2020useful,goyal2019scaling}. Furthermore, unsupervised learning can improve performance for setups with few labels \cite{newell2020useful,su2020does}, even for small datasets \cite{su2020does}, but improvements diminish with a growing amount of annotations \cite{newell2020useful}. Notably, \cite{asano2019critical} shows that unsupervised learning is capable of learning early-layer features from a single image.
\\\\
\textbf{Analyzing self-supervised learning across target tasks} is another way to define and evaluate benchmarks for unsupervised approaches \cite{doersch2017multi,newell2020useful,wallace2020extending,zhai2019large,goyal2019scaling,locatello2019challenging,ericsson2021well,zhao2020makes,gwilliam2022beyond,gwilliam2022beyond}. In our work in \autoref{chap:01} we compare the objective function mismatch across various target domains with our proposed metrics. Good representations are often defined as those that adapt to diverse, unseen tasks with few target examples \cite{zhai2019large,goyal2019scaling}. However, various works show in different setups that the performance of unsupervised models is not consistent across different domains and depends on the pretext task and that there is no best unsupervised method across all target tasks \cite{doersch2017multi,newell2020useful,wallace2020extending,zhai2019large,goyal2019scaling,locatello2019challenging,gwilliam2022beyond,gwilliam2022beyond}. By transferring unsupervised methods from one domain to various target tasks, it has also been found that no single method dominates, but unsupervised methods outperform the supervised baseline in most tasks \cite{ericsson2021well,zhao2020makes}. Similarly, unsupervised visual representations seem to be superior in transferring to unseen concepts, e.g., transferring from cat to tiger cat \cite{sariyildiz2021concept}.
\\\\
\textbf{Other empirical evaluations of unsupervised representations} directly or indirectly measure target task performance. It has been found that unsupervised learning can improve robustness to adversarial examples as well as input and label corruptions and can exceed supervised performance in out-of-distribution detection \cite{hendrycks2019using}. With a small subset of ImageNet \cite{deng2009imagenet} annotated with sixteen factors of variation like background or lighting, it has been observed that the examined unsupervised models are consistently failing to predict all these factors \cite{idrissi2022imagenet}. This underscores the findings of our preceding work in \autoref{chap:01}. Furthermore, augmentations can have a positive and negative effect on the prediction of these factors \cite{stuhr2022don,idrissi2022imagenet,gwilliam2022beyond}, and effects of augmentations can also depend on the unsupervised method \cite{appalaraju2020towards,stuhr2022don}. In addition, it has been observed that different unsupervised and supervised methods learn somewhat similar representations in middle layers, which vary strongly in the layers near the loss function \cite{grigg2021self,gwilliam2022beyond}. With aggregated representations of joint-embedding models for fixed-scale image patches, which result in on-par or even better performance, it has been shown that these models mainly learn a distributed representation of image patches containing local information shared among similar patches \cite{chenintra2023}. For joint-embedding methods it has been observed that mini-batch training is a detrimental prior for learning features of class-imbalanced data that can be utilized to cluster the data uniformly \cite{assran2022hidden}. However, contrastive methods (and maybe others) learn more balanced feature spaces compared to supervised learning \cite{kang2021exploring}. This prior can be mitigated by prior-matching \cite{assran2022hidden}.

By indirectly investigating the disentanglement of unsupervised representation with six metrics, it has been found that none of the examined unsupervised methods reliably learn disentangled representations, nor does the disentanglement of these representations directly correspond with target task performance \cite{locatello2019challenging,locatello2020sober}. To date, there exists no indirect measurement with a clear correlation between metrics score and the performance on a variety of target tasks \cite{locatello2019challenging,locatello2020sober,cabannes2023ssl}. 
\\\\
\textbf{Visualizing unsupervised representations} to understand their inherent structure has recently led to interesting findings \cite{appalaraju2020towards,caron2021emerging,bordes2021high,chenintra2023}. An early work uses three techniques - activation maximization, sampling, and linear combinations of filters - to visually examine the responses of individual units of stacked denoising autoencoders \cite{vincent2008extracting} and deep belief networks \cite{hinton2006fast,erhan2009visualizing}. It has been found that these models learn hierarchical representations that combine into meaningfully more complicated representations in deeper layers \cite{erhan2009visualizing}. By using feature inversion \cite{ulyanov2018deep} before and after the projection head of joint-embedding methods, it has been found that layers closer to the contrastive loss function lose information due to invariances to augmentations \cite{appalaraju2020towards,grigg2021self}. A similar effect has been observed in \cite{gwilliam2022beyond} for other methods. In \cite{bordes2021high} a diffusion model is conditioned on unsupervised representations to visualize them in image space. Based on observations backed up by empirical evidence, it has been also observed that projector representations are invariant to augmentations used during training. In contrast, the encoder representations contain contextualized local information \cite{bordes2021high}. Similar observations have been made in \cite{zhao2020makes} and \cite{chenintra2023}. Another work shows that the self-attention maps of an unsupervised trained transformer relate to class-specific feature maps analogous to segmentations \cite{caron2021emerging}. Furthermore, visual observations have indicated that representations are more robust to small adversarial attacks and retain more detailed information (e.g., scale or color) of the image in a structured manner than supervised ones \cite{caron2021emerging}.
\section{Unsupervised Adaptation of Visual Representations}
\label{sec:realtedwork:unsuperviseddomainadaptation}
Our work in \autoref{chap:02} relates to unsupervised visual domain adaptation, which has been extensively studied in recent years \cite{wilson2020survey,farahani2021brief,zhang2021survey,liu2022deep}.
\\\\
\textbf{Data generation for sim-to-real lane detection.} In recent years, much attention has been paid to lane detection benchmarks in the real world, such as CULane \cite{pan2018spatial}, TuSimple \cite{TuSimple2017}, LLAMAS \cite{behrendt2019unsupervised}, and BDD100K \cite{yu2020bdd100k}. Despite the popularity of these benchmarks, there is little research that focuses on sim-to-real lane detection datasets. One work proposed a method for generating synthetic images with 3D lane annotations in the open-source engine blender \cite{garnett20193d}. Their \textit{synthetic-3D-lanes} dataset contains 300K train, 1,000 validation and 5,000 test images, while their real-world \textit{3D-lanes} dataset consists of 85K images, which are annotated in a semi-manual manner. Utilizing the data generation method from \cite{garnett20193d}, in \cite{garnett2020synthetic} 50K labeled synthetic images have been collected to perform sim-to-real domain adaptation for 3D lane detection. At this point, the source domain of the dataset is not publicly available. Recently, unsupervised domain adaptation techniques for 2D lane detection have been investigated in \cite{hu2022sim}. The proposed data generation method relies on CARLA's built-in agent to automatically collect 16K synthetic images \cite{hu2022sim}. However, the dataset is not publicly available at this point. In comparison, our work \cite{stuhr2022carlane} presented in \autoref{chap:01} leverages an efficient and configurable waypoint-based agent in CARLA to collect simulation data. Furthermore, in contrast to the aforementioned works, considering only single-source single-target unsupervised domain adaptation, we additionally focus on multi-target unsupervised domain adaptation with data collected from a 1/8th model vehicle and a cleaned version of highway drives from the TuSimple \cite{TuSimple2017} dataset.
\\\\
\textbf{Discrepancy-based methods} employ a distance metric to measure the discrepancy between the source and target domain \cite{long2015learning,sun2016return,zhu2020deep,zhang2021efficient}. A prominent example is DAN \cite{long2015learning} which uses maximum mean discrepancies (\acs{MMD}) \cite{gretton2006kernel, gretton2012kernel} to match embeddings of different domain distributions. DSAN \cite{zhu2020deep} builds upon DAN with local MMD and exploits fine-grained features to align subdomains accurately. In \cite{sun2016return} the second-order statistics of source and target distributions are aligned by re-coloring whitened source features with the covariance of the target distribution. Furthermore, there is work that takes domain-specific statistics into account by re-normalizing the model with domain statistics \cite{mancini2018boosting,xu2019self} or by training separate batch norm layers for each domain \cite{chang2019domain}. Other work, for example, minimizes divergences \cite{meng2018adversarial,jiang2020resource}, and entropy \cite{sohn2019unsupervised}, aligns mutual information \cite{gholami2020unsupervised}, uses a contrastive loss \cite{kang2019contrastive}, or mitigates optimization inconsistencies by minimizing the gradients discrepancy of the source samples and target samples \cite{du2021cross}. 
\\\\
\textbf{Adversarial discriminative methods} employ a domain classifier or discriminator besides the target model on the feature extractor(s), which tries to classify the domain of the input's representation; this encourages the feature extractor to produce domain-invariant representations \cite{ganin2015unsupervised,ganin2016domain,tzeng2017adversarial,zhang2018collaborative,chen2019progressive,sohn2019unsupervised,xu2019self,wei2021toalign,akkaya2021self}. In \cite{ganin2015unsupervised,ganin2016domain} the domain classifier is connected with a gradient reversal layer to the features extractor, which leads to optimization for features indistinguishable for the domain classifier. Another work employs multiple domain classifiers on early blocks of the model to learn domain-informative features and multiple domain classifiers on later blocks with a gradient rehearsal layer to learn domain invariant features \cite{zhang2018collaborative}. In \cite{wang2020classes} a fine-grained discriminator is used to not only distinguish between domains but also to distinguish between complex structures by splitting the discriminator output and softening the discriminator's labels. ADDA \cite{tzeng2017adversarial} aligns a target feature encoder with a pre-trained, frozen source encoder utilizing a discriminator. While precedent methods mainly rely on feature-level alignment, adversarial generative methods solely \cite{bousmalis2017unsupervised,xie2020self} or additionally \cite{hoffman2018cycada} operate on pixel-level. Our work in \autoref{chap:02} combines DANN \cite{ganin2016domain} and ADDA \cite{tzeng2017adversarial} with the UFLD \cite{qin2020ultra} method and adopts them for row-based 2D lane detection to evaluate them on the proposed CARLANE benchmark.
\\\\
\textbf{Pseudo-labeling-based methods} use different strategies to create pseudo labels for the target domains and utilize these labels during training. There are three common strategies: 1) Directly selecting pseudo labels based on the model output \cite{zhang2018collaborative,zhang2021efficient,saito2017asymmetric}. 2) Training separate blocks or layers for pseudo-labeling during domain adaptation \cite{saito2017asymmetric,xie2018learning,chen2019progressive,zhang2020label}. 3) Using a pre-trained model to create the pseudo labels \cite{chang2019domain}.
In \cite{zhang2018collaborative} and \cite{akkaya2021self} pseudo labels are selected based on the scores of an image classifier and a domain discriminator. Another work iteratively extends the source training set with pseudo-labeled target samples with a high confidence score of the classifier \cite{zhang2021efficient}. In \cite{saito2017asymmetric} the agreement and confidence of two models is utilized to estimate target pseudo labels. In \cite{pan2019transferrable} and \cite{chen2019progressive} target predictions are compared with source prototypes to estimate the pseudo label and a distance measure or similarity score is used to select the pseudo labels. Another work utilizes weighted combinations of soft cluster assignments to create pseudo-labeled virtual instances \cite{zhang2020label}. In \cite{chang2019domain} a pre-trained model is used to estimate pseudo labels for a fully-supervised classifier network, which is utilized to refine the pseudo labels interactively. Inspired by \cite{akkaya2021self} in our work in \autoref{chap:02}, we propose a pseudo labels selection mechanism for row-based 2D lane detection utilizing classifier and discriminator scores. Furthermore, we combine SGADA \cite{akkaya2021self} with our pseudo label selection mechanism and with the UFLD \cite{qin2020ultra} method for evaluation on our CARLANE benchmark. 
\\\\
\textbf{Indirect feature-matching approaches} match features of the source and target domain indirectly, for example, through prototypes \cite{xie2018learning,tanwisuth2021prototype,yue2021prototypical}, decomposed features \cite{wei2021toalign}, or mirror samples \cite{zhao2021reducing}. This avoids variabilities and biases in sampling and class distribution \cite{tanwisuth2021prototype}. In \cite{xie2018learning} source prototypes are directly aligned with target prototypes computed from pseudo-labeled target samples. Instead of averaging sample features, another work learns prototypes with a cross-entropy loss between prototype and source features \cite{tanwisuth2021prototype}. PCS \cite{yue2021prototypical} creates source and target prototypes from memory banks via k-means clustering and utilizes a contrastive loss between the sample's feature and the prototypes for in-domain feature learning. Furthermore, the source sample's features are aligned with target prototypes via a similarity score for cross-domain feature learning. Another work constructs mirror samples - the ideal counterparts in the other domain - and aligns them across the domains \cite{zhao2021reducing}. In \cite{wei2021toalign} source domain features are decomposed to task-related features for domain adaptation and task-irrelenvant features. Our work in \autoref{chap:02} combines PCS \cite{yue2021prototypical} with the UFLD \cite{qin2020ultra} method as well as our pseudo label selection method and adopts PCS for row-based 2D lane detection. 
\\\\
\textbf{Self-supervised auxiliary tasks} are leveraged to improve domain adaptation effectiveness by capturing in-domain \cite{ghifary2016deep,sun2019unsupervised,xu2019self} or cross-domain \cite{bousmalis2016domain,wu2018unsupervised,xu2019self,xie2020self} structures. In \cite{ghifary2016deep} a simple reconstruction objective is used to learn in-domain target features. Rotation, flip, and patch location prediction are utilized as auxiliary tasks in the source and target domain and can lead to a reduction of the domain gap \cite{sun2019unsupervised}. Another work has assessed different setups for early unsupervised tasks to improve domain adaptation and has found that simple rotation prediction results in significant performance gains \cite{xu2019self}. In \cite{bousmalis2016domain} shared encoders and private encoders are used in an autoencoder setup to learn in-domain and cross-domain features. Similarly, instance discrimination \cite{wu2018unsupervised} is utilized for in-domain feature learning together with a cross-domain self-supervision based on similarity to align both domains \cite{kim2020cross}. Additionally, in-domain contrastive learning is performed between the sample's features and prototypes in \cite{yue2021prototypical} and our work in \autoref{chap:02}. 
\\\\
Furthermore, other work, for example, applies a meta-learning scheme between the domain alignment and the targeted classification task \cite{wei2021metaalign} or uses attention to focus on relevant source samples \cite{moon2017completely}.  
\section{Unsupervised Translation of Visual Representations}
\label{sec:realtedwork:unsupervisedtranslation}
Our work in \autoref{chap:03} relates to unpaired image-to-image translation, which is a special case of unsupervised domain adaptation \cite{wilson2020survey} and has been extensively studied in recent years \cite{alotaibi2020deep,pang2021image}.
\\\\
\textbf{Unpaired image-to-image translation.} Following the success of GANs \cite{goodfellow2014generative}, the conditional GAN framework \cite{mirza2014conditional} enables image generation based on an input condition. Pix2Pix \cite{isola2017image} uses images from a source domain as a condition for the generator and discriminator to translate them to a target domain. Since Pix2Pix relies on a regression loss between generated and target images, translation can only be performed between domains where paired images are available. To achieve unpaired image-to-image translation, methods like CycleGAN \cite{zhu2017unpaired}, UNIT \cite{liu2017unsupervised}, and MUNIT \cite{huang2018multimodal} utilize a second GAN to perform the translation in the opposite direction and impose a cycle-consistency constraint or weight-sharing constraint between both GANs. However, these methods require additional parameters for the second GAN, which are used to learn the unpaired translation and are omitted when inferring a one-sided translation. In works such as TSIT \cite{jiang2020tsit} and CUT \cite{park2020contrastive}, these additional parameters are completely omitted at training time by either utilizing a perceptual loss \cite{johnson2016perceptual} between the input image of the generator and the image to be translated or by patchwise contrastive learning. Recently, additional techniques have achieved promising results, like pseudo-labeling \cite{hao2021gancraft} or a conditional discriminator based on segmentations created with a robust segmentation model for both domains \cite{richter2022enhancing}. Furthermore, there are recent efforts to adapt diffusion models to unpaired image-to-image translation \cite{su2022dual,zhao2022egsde,wu2022unifying}.
\\\\
\textbf{Content consistency in unpaired image-to-image translation.} Due to biases between unpaired datasets, the content of translated samples can not be trivially preserved \cite{richter2022enhancing}. There are ongoing efforts to preserve the content of an image when it is translated to another domain by improving various parts of the training pipeline: Several consistency constraints have been proposed for the generator, which operate directly on the translated image \cite{zhu2017unpaired, benaim2017one, lin2020multimodal}, on a transformation of the translated image \cite{taigman2016unsupervised, zhang2019harmonic, fu2019geometry, yang2020phase, wang2020classes}, or on distributions of multi-modal translated images \cite{zhao2020unpaired}. The use of a perceptual loss \cite{johnson2016perceptual} or LPIPS loss \cite{zhang2018unreasonable} between input images and translated images, as in \cite{jiang2020tsit} and \cite{richter2022enhancing}, can also be considered a consistency constraint between transformed images. In \cite{xie2020self} content consistency is enforced with self-supervised in-domain and cross-domain patch position prediction. There is work that enforces consistency by constraining the latent space of the generator \cite{liu2017unsupervised, huang2018multimodal, sendik2020crossnet}. Semantic scene inconsistencies can be mitigated with a separate segmentation model \cite{yang2020phase, lin2020multimodal}. To avoid inconsistency arising from style transfer, features from the generator stream are masked before AdaIN \cite{huang2017arbitrary,ma2018exemplar}. Another work exploits small perturbations in the input feature space to improve semantic robustness \cite{jia2021semantically}. However, if the datasets of both domains are unbalanced, discriminators can use dataset biases as learning shortcuts, which leads to content inconsistencies. Therefore, only constraining the generator for content consistency still results in an ill-posed unpaired image-to-image translation setup. Constraining discriminators to achieve content consistency is currently underexplored, but recent work has proposed promising directions. There are semantic-aware discriminator architectures \cite{liang2018generative, liu2019learning, richter2022enhancing, hao2021gancraft} that enforce discriminators to base their predictions on semantic classes, or VGG discriminators \cite{richter2022enhancing}, which additionally operate on abstract features of a frozen VGG model instead of the input images. Training discriminators with small patches \cite{richter2022enhancing} is another way to improve content consistency. To mitigate dataset biases during training for the whole model, sampling strategies can be applied to sample similar patches from both domains \cite{kao2019patch, richter2022enhancing}. Furthermore, in \cite{theiss2022unpaired} a model is trained to generate a hyper-vector mapping between source and target images with an adversarial loss and a cyclic loss for content consistency. In contrast, our work \cite{stuhr2023masked} in \autoref{chap:03} utilizes a robust semantic mask to mask global discriminators with a large field of view, which provide the generator with the gradients of the unmasked regions. This leads to a content-consistent translation while preserving the global context. We combine this discriminator with an efficient sampling method that uses robust semantic segmentations to sample similar crops from both domains.
\\\\
\textbf{Attention in image-to-image translation.}
Previous work utilizes attention for different parts of the GAN framework. A common technique is to create attention mechanisms that allow the generator or discriminator to focus on important regions of the input \cite{alami2018unsupervised, tang2021attentiongan, yang2019show, kim2020u,zhang2022region} or to capture the relationship between regions of the input(s) \cite{yao2019attention, tang2020dual, hu2022qs}. Other work guides a pixel loss with uncertainty maps computed from attention maps \cite{tang2019multi}, exploits correlations between channel maps with scale-wise channel attention \cite{tang2020dual}, disentangles content and style with diagonal attention \cite{kwon2021diagonal}, or merges features from multiple sources with an attention block before integrating them into the generator stream \cite{liu2021liquid}. In \cite{lin2021attention}, an attention-based discriminator is introduced to guide the training of the generator with attention maps. Furthermore, \acs{ViT}s \cite{dosovitskiy2020image} are adapted for unpaired image-to-image translation \cite{torbunov2023uvcgan, zheng2022ittr}, and the computational complexity of their self-attention mechanism is reduced for high-resolution translation \cite{zheng2022ittr}. In contrast, our work in \autoref{chap:03} proposes an attention mechanism to selectively integrate statistics from the content stream of the source image into the generator stream. This allows the model to focus on statistical features from the content stream that are useful for the target domain.

\part{Unsupervised Learning of\\Visual Representations}
\graphicspath{{./main/4_chapter00/sections/figures/}}

\chapter{Self-Organizing Convolutional Neural Networks }
\label{chap:00}
\vspace{-8mm}
\begin{abstract}
In this chapter, we combine Convolutional Neural Networks (\acs{CNN}s), clustering via Self-Organizing Maps (\acs{SOM}s), and Hebbian learning to propose the building blocks of Convolutional Self-Organizing Neural Networks (\acs{CSNN}s), which learn representations in an unsupervised and backpropagation-free manner. Our approach replaces the learning of traditional convolutional layers from CNNs with the competitive learning procedure of SOMs and simultaneously learns local masks between these layers with separate Hebbian-like learning rules to mitigate the problem of disentangling factors of variation when filters are learned through clustering. We investigate the learned representation by designing two simple models with our building blocks, achieving comparable performance to many methods which use backpropagation. Furthermore, we reach comparable performance on Cifar10 and give baseline performances on Cifar100, Tiny ImageNet, and a small subset of ImageNet for backpropagation-free methods. 
\end{abstract}
\section{Motivation}
A well-known downside of many successful deep learning approaches like backpro-pagation-based \acs{CNN}s is their need for large, labeled datasets. Obtaining these datasets can be costly and time-consuming, or the required quantity of samples is unavailable due to restrictive conditions. In fact, the availability of big datasets and the computational power to process this information are two of the main reasons for the success of deep learning techniques \cite{krizhevsky2017imagenet}. This raises the question of why models like \acs{CNN}s need that much data: Is it the huge amount of model parameters, are training techniques like backpropagation \cite{linnainmaa1970representation,werbos1974beyond,rumelhart1986learning} just too inefficient, or is the cause a combination of both?

The fact that neural networks have become both large and deep in recent years, often consisting of millions of parameters, suggests that the cause is a combination of both.
On the one hand, large amounts of data are needed because there are so many model parameters to optimize. On the other hand, the data may not yet be optimally utilized. The suboptimal utilization of data to date could be related to the following problem, which we call the \textit{Scalar Bow Tie Problem}:
\begin{figure}[t]
	\begin{center}
		\includegraphics[width=0.35\linewidth]{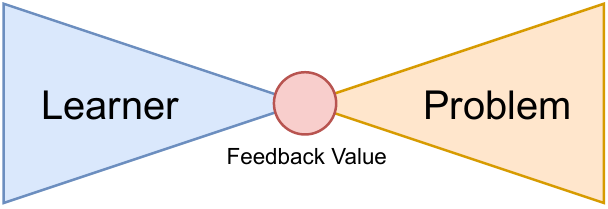}
	\end{center}
	\vspace{-1ex}
	\caption[The Scalar Bow Tie Problem.]{The Scalar Bow Tie Problem. With this term, we describe the current problematic situation in deep learning in which most training approaches for neural networks and agents depend only on a single scalar loss or feedback value. This single value is used to subsume the performance of an entire neural network or agent, even for very complex problems.}
	\label{fig:csnns:scalar_bow_tie_problem}
\end{figure}
Most training strategies for neural networks rely just on a single scalar value (the loss value) which is computed for each training step and then used as a feedback signal for optimizing millions of parameters. In contrast, humans utilize multiple feedback types to learn, for example, different sensor modalities when learning representations. With the Scalar Bow Tie Problem, we refer to the current situation for training neural networks. Similar to a bow tie, both ends of the learning setting (the learner and the task to be learned) are complex, and the only connection between these complex parts is one single point (the scalar loss or feedback value).

There exists a current effort of backpropagation-based methods to overcome the Scalar Bow Tie Problem, e.g., by using multiple loss functions at different output locations of the network (a variant of multi-task learning, i.a. \cite{doersch2017multi}). Furthermore, there exists a debate to which degree backpropagation is biologically plausible on a network level (e.g., \cite{bengio2015towards,lillicrap2016random,whittington2019theories}), which often highlights that layer-dependent weight updates in the whole model are less likely than more independent updates, like layer-wise ones, and beyond that would require an exact symmetric copy $W^T$ of the upstream weight $W$. In \cite{grossberg1987competitive} this problem is described in detail and referred to as the weight transport problem.

In this chapter we do not aim at solving all these problems, but want to take a small step away from common deep learning methods while keeping some benefits of these models, such as modularity and hierarchical structure.
\section{Contributions}
We propose a CNN variant with building blocks that learn in an unsupervised, self-organizing, and backpropagation-free manner. Within these blocks, we combine methods from \acs{CNN}s \cite{hubel1959receptive,lecun1989backpropagation}, SOMs \cite{kohonen1982self}, and Hebbian learning \cite{hebb2005organization}. 
\\\\\\
Our main contributions are as follows:
\begin{itemize}[]
	\item We propose an unsupervised, backpropagation-free learning algorithm that utilizes two learning rules to update the weights layer-wise without using an explicitly defined loss function, thereby reducing the Scalar Bow Tie Problem. (\hyperref[RQ-L1]{RQ-L1})
	\item This learning algorithm is used to train \acs{CSNN} models, with which we achieve comparable performance to many models trained in an unsupervised manner. (\hyperref[RQ-E1]{RQ-E1}) 
	\item We overcome a fundamental problem of \acs{SOM}s trained on image patches by presenting two types of weight masks to mask input and neuron activities. (\hyperref[RQ-L1]{RQ-L1})
	\item We propose a multi-headed version of our building blocks to further improve performance. (\hyperref[RQ-L2]{RQ-L2})
	\item To the best of our knowledge, our approach is the first to combine the successful principles of \acs{CNN}s, \acs{SOM}s, and Hebbian learning into a single model.
\end{itemize}
\section{CSNN}
\begin{figure}[t]
	\begin{center}
		\includegraphics[width=0.75\linewidth]{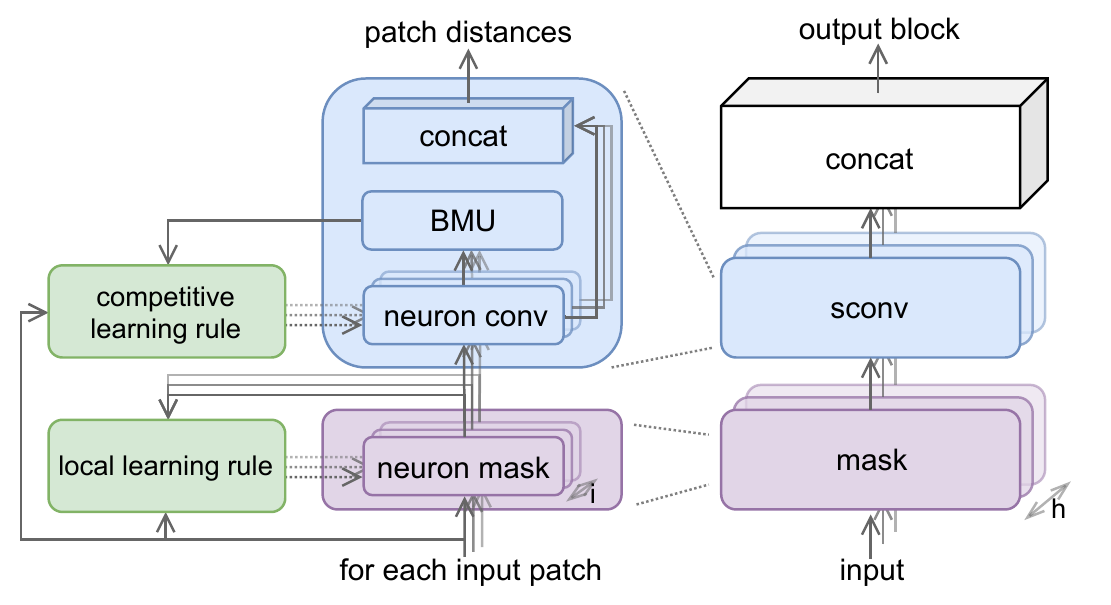}
	\end{center}
	\vspace{-1ex}
	\caption[Architecture of a \acs{CSNN} layer.]{Architecture of a \acs{CSNN} layer. (left) mask and \acs{sconv} layer computations for a single patch: First, each patch is masked for each neuron $i$. The masked outputs and the input patch are subsequently used in the local learning process. Second, the \acs{sconv} layer computes the convolution distances and the \acs{BMU}. The \acs{BMU} and the input patch are subsequently used in competitive learning. The distances are concatenated to the patch's spatial activation vector. Analogous to \acs{CNN}s, the spatial activation vectors of all patches form the convolutional output block. (right) Multi-Head masked \acs{sconv}: Multiple masking and \acs{sconv} layers $h$ are applied to the entire input, and the resulting output blocks are concatenated in the feature dimension.}
	\label{fig:csnns:csnn_layer}
\end{figure}

This section describes the key methods of \acs{CSNN}s. The proposed blocks for each method and their interactions are summarized in \autoref{fig:csnns:csnn_layer}.

\subsection{Convolutional, Self-Organizing Layer}
The convolutional, self-organizing layer (\acs{sconv}) closely follows the convolutional layer of standard \acs{CNN}s. The entire input is convolved patchwise with each learned filter:
\begin{ceqn}
	\begin{equation}
	y_{m,n,i}=\mathbf{p}_{m,n}\cdot \mathbf{w}_{i}
	\label{eqn:conv}
	\end{equation}
\end{ceqn}
where $\mathbf{p}_{m,n}$ is the image patch centered at position $m, n$ of the input, $\mathbf{w}_i$ is the $i$-th filter of the layer, and $y_{m,n,i}$ is the activation at position $m,n,i$ of the output. The only difference to a standard convolutional layer is that the weights are learned through a competitive, self-organizing optimization procedure. The layer's outputs for one image patch are the distances for each neuron in a \acs{SOM}. Each neuron of the \acs{SOM} corresponds to one filter in a \acs{CNN}, and each distance output of a \acs{SOM} neuron corresponds to a specific feature $y_{m,n,i}$ of the feature map $i$ for the patch $m,n$ after the convolution operation was applied. The main difference to a standard CNN is that the \acs{CSNN} only uses local learning rules. Furthermore, \acs{CSNN}s are trained unsupervised in a bottom-up manner, in contrast to the top-down supervised backpropagation approach.
Analogous to \acs{CNN}s, the proposed modular layers can be combined with other layers to form deep learning architectures.

\subsection{Competitive Learning Rule}
Most learning rules for \acs{SOM}s require a best matching unit (\acs{BMU}) to compute the weight change $\Delta w$. Since we use a convolution as a distance metric, the index of the \acs{BMU} for one patch is defined as:
\begin{ceqn}
	\begin{equation}
		c_{m,n}=\argmax_{i}\{y_{m,n,i}\}
		\label{eqn:BMU}
	\end{equation}
\end{ceqn}
where $c_{m,n}$ is the index of the \acs{BMU} in a 2D \acs{SOM}-grid for patch $m,n$. 

To allow for self-organization, learning rules of \acs{SOM}s require a neighborhood function to compute the neighborhood coefficients from all the other neurons in the grid to the \acs{BMU}. A common type of neighborhood function is the Gaussian:
\begin{ceqn}
	\begin{equation}
		h_{m, n, i}(t)=\exp(\frac{-d_{A}(k_{m,n,c_{m, n}}, k_{m,n,i})^2}{2\delta(t)^2})
		\label{eqn:nfunction}
	\end{equation}
\end{ceqn}
where $k_{m,n,c_{m, n}}$ and $k_{m,n,i}$ are the coordinates of the \acs{BMU} and the $i$-th neuron in the SOM grid, and $\delta(t)$ is a hyperparameter to control the radius of the Gaussian, which could change with the training step $t$. For the distance function $d_A$, we utilize the Euclidean distance between the \acs{BMU} and the $i$-th neuron. Then $h_{m, n, i}(t)$ is the neighborhood coefficient of neuron $i$ to the center (the BMU) $c_{m,n}$ for the patch $m,n$. Now the weight update for one patch can be defined as:
\begin{ceqn}
	\begin{gather}
		\Delta \mathbf{w}_{m,n,i}(t)=a(t)h_{m, n, i}(t)\mathbf{p}_{m,n}\\
		\mathbf{w}_{m,n,i}(t)=\frac{\mathbf{w}_{m,n,i}(t)+\Delta \mathbf{w}_{m,n,i}(t)}{\left\| \mathbf{w}_{m,n,i}(t)+\Delta \mathbf{w}_{m,n,i}(t)\right\|}
	\end{gather}
\end{ceqn}
where $\left\|...\right\|$ is the Euclidean norm used to obtain the positive effects of normalization, such as preventing the weights from growing too much, and $a(t)$ is the learning rate, which could change over the course of the training. For an entire image, the final weight change for the weight vector of a \acs{SOM} neuron is calculated by averaging over all patch weight updates:
\begin{ceqn}
	\begin{equation}
		\Delta \mathbf{w}_{i}(t)=\frac{a(t)}{mn}\sum_{m,n}h_{m, n, i}(t)\mathbf{p}_{m,n} 
		\label{eqn:slearning}
	\end{equation}
\end{ceqn}
In batch training, the average of all patches in a batch is taken.

Using other distance metrics (such as the L1 or L2 norm) may require changing the BMU computation (e.g., to argmin) and the learning rule. 

\subsection{Mask Layers}\label{masklayer}
We argue that given the learning rule \ref{eqn:slearning}, a \acs{SOM} neuron is unable to disentangle factors of variation, which is a key property of deep learning architectures \cite{goodfellow2016deep}. In other words, the \acs{SOM} neuron cannot pay more or less attention to a part of the input or ignore that part entirely. The \acs{SOM} simply tries to shift its neurons to best fit the dataset. This can lead to poor performance because a neuron is unable to learn abstract features that can be considered concepts. Even at higher layers, a \acs{SOM} neuron processes as input only the output from a collection of \acs{SOM} neurons of the previous layer in its receptive field. Therefore, the network forms a hierarchy of collections rather than a hierarchy of concepts. Furthermore, a neuron in a deeper layer with a small receptive field, e.g., $3\times3$, needs to compute the distance between its $3\times3\times256$ vector and the input patch if there are $256$ \acs{SOM} neurons in the previous layer. Since the weight update shifts all the values of the weights from the \acs{BMU} and its neighbors in a direction to the input at once, it seems unlikely that the \acs{SOM} neurons learn distinct enough features for data with a high amount of information, such as images. This argument goes hand in hand with another work, where similar observations were made for k-means clustering \cite{dundar2015convolutional}. 

\begin{figure}[t]
	\begin{center}
		\includegraphics[width=0.75\linewidth]{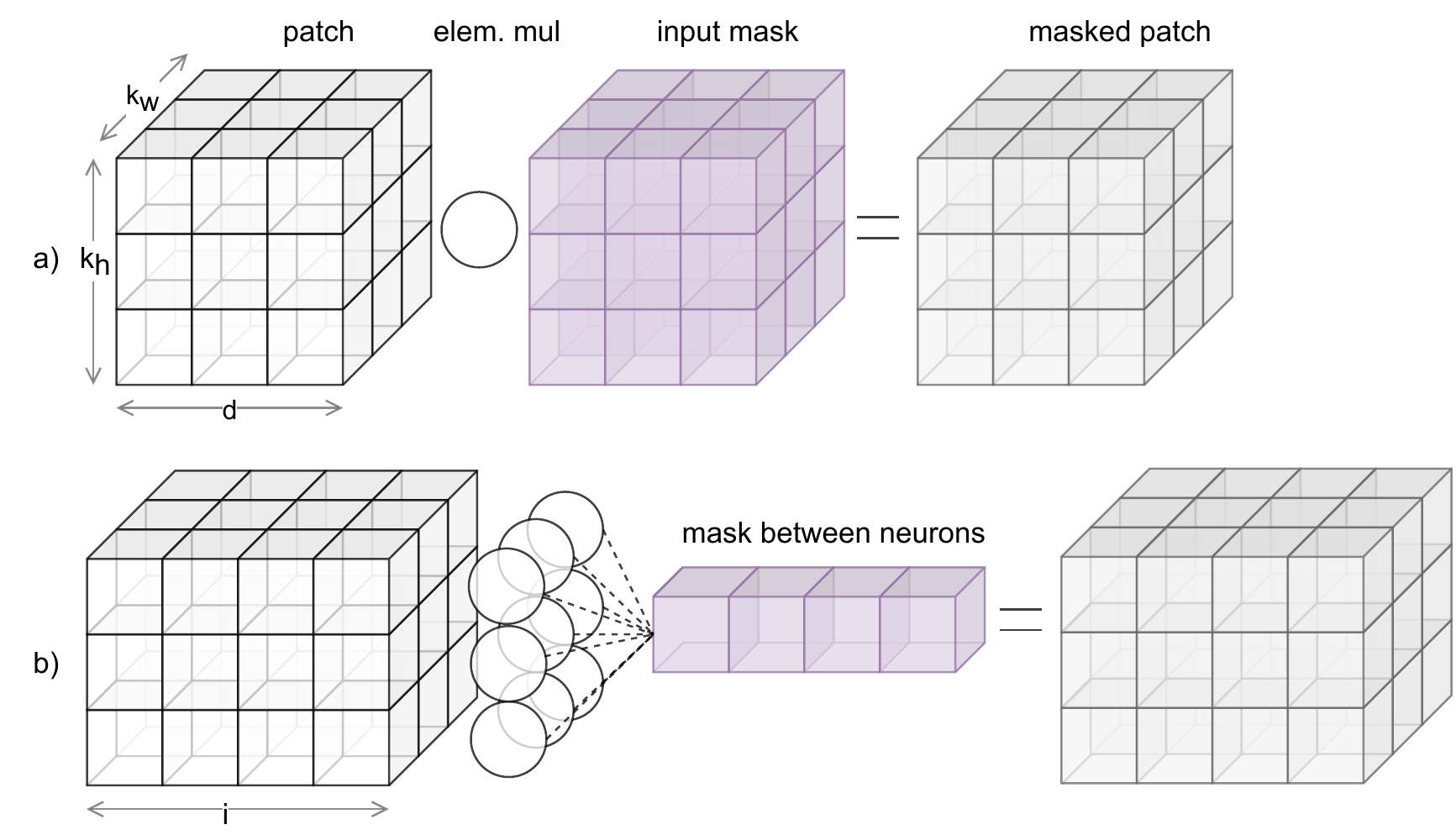}
	\end{center}
	\vspace{-1ex}
	\caption[\acs{CSNN} mask types.]{\acs{CSNN} mask types. a) Input mask: The Hadamard product between an input patch and a mask of the same size. b) Mask between layers: The mask, with its length equal to the number of filters in the previous layer, is multiplied element-wise with every spatial activation vector from the patch.}
	\label{fig:csnns:mask_types}
\end{figure}

To allow the network to focus on certain parts of the input, we propose two types of separately-learned mask vectors, which are shown in \autoref{fig:csnns:mask_types}. The first mask (a) enables each neuron to mask its input values with a vector of the same dimension as the input patch and the SOM neuron's weight vector. Each CSNN neuron multiplies the image patch with its mask and convolves the masked patch with the weight vector of the SOM neuron:
\begin{ceqn}
	\begin{gather}
		\mathbf{\hat{y}}_{m,n,i}=\mathbf{p}_{m,n}\circ \mathbf{m}_{i}
		\label{eqn:inputmask}\\
		y_{m,n,i}=\mathbf{\hat{y}}_{m,n,i}\cdot \mathbf{w}_{i}
		\label{eqn:maskpatch}
	\end{gather}
\end{ceqn}
where $\mathbf{m}_{i}$ is the mask of neuron $i$, $\mathbf{\hat{y}}_{m,n,i}$ is the masked patch for neuron $i$, and $\circ$ denotes the Hadamard product.

The second mask (b) is located between the neurons of two layers. Therefore, it requires $k_w\times k_h$ fewer parameters than the first (input) mask when $k_w$ and $k_h$ are the kernel's sizes. The mask is defined as:
\begin{ceqn}
	\begin{equation}
		\mathbf{\hat{y}}_{m_h,n_w,i}=\mathbf{p}_{m_h,n_w}\circ \mathbf{n}_{i}
		\label{eqn:neuronmask}
	\end{equation}
\end{ceqn}
where $\mathbf{n}_{i}$ is the mask element-wise multiplied in the feature (depth) dimension $k_w\times k_h$ times. With this mask, we want to enable each \acs{SOM} neuron to learn from which neuron of the previous layer to receive information. Mask (a) is used for the input and mask (b) between each sconv layer.
\\\\
We note that we still utilize the unmasked input for training the \acs{SOM} weight vectors - the mask is only used to calculate the convolutional distances and to determine the \acs{BMU}. Moving the \acs{BMU} and its neighbors to the unmasked input direction leads to a more stable learning process since the self-organization does not depend too much on the higher mask coefficients. High mask coefficients could drive the weight vector of a single \acs{SOM} neuron into a direction from which it can hardly adapt to other new but similar input patches. Thereby the neuron could get stuck in a dead end. Furthermore, this allows the mask to learn new regions typical for a particular input type, even when the mask tends to ignore these regions of the input. This is because the \acs{SOM} moves towards the unmasked input, which enables the mask to increase its corresponding coefficients.

\subsection{Local Learning Rules}
To learn the input mask of each neuron, we propose two simple local learning rules inspired by the Generalized Hebbian Algorithm \cite{sanger1989optimal} (\acs{GHA}). These rules show good results in our experiments, but there are certainly other learning rule alternatives that could guide the training process of the masks. Below we derive our final learning rules by discussing the following equations within the context of \acs{CSNN}s, where $e$ indicates that we present the element-wise form of the learning rules. For the original inviolate versions we refer to the literature.
\begin{ceqn}
	\begin{gather}
	\Delta h_e(\mathbf{p}_{m, n}, \mathbf{\hat{y}}_{m, n, i}) = \mathbf{\hat{y}}_{m, n, i} \circ \mathbf{p}_{m, n} 
	\label{eqn:simple_hebbian_learning}\\
	\Delta o_e(\mathbf{p}_{m, n}, \mathbf{\hat{y}}_{m, n, i}, \mathbf{m}_{i}) = \mathbf{\hat{y}}_{m, n, i} \circ (\mathbf{p}_{m, n} - \mathbf{\hat{y}}_{m, n, i} \circ \mathbf{m}_{i})
	\label{eqn:ojas_rule}\\
	\Delta m_{es}(\mathbf{p}_{m, n}, \mathbf{\hat{y}}_{m, n, i}, \mathbf{m}_{i}) = h_e([\mathbf{p}_{m, n}-\gamma\sum_{k}\mathbf{\hat{y}}_{m,n,k}\circ \mathbf{m}_{k}], \mathbf{\hat{y}}_{m, n, i}, \mathbf{m}_{i})
	\label{eqn:similar_to_gha}\\
	\Delta m_{ec}(\mathbf{p}_{m, n}, \mathbf{\hat{y}}_{m, n, i}, \mathbf{m}_{i}) = o_e([\mathbf{p}_{m, n}-\gamma\sum_{k<i}\mathbf{\hat{y}}_{m,n,k}\circ \mathbf{m}_{k}], \mathbf{\hat{y}}_{m, n, i}, \mathbf{m}_{i})
	\label{eqn:similar_to_gha_with_restricted_summation}\\
	\Delta \mathbf{m}_{i}(t)=\frac{a(t)}{mn}\sum_{m,n}[\Delta m_{es}(\mathbf{p}_{m, n}, \mathbf{\hat{y}}_{m, n, i}, \mathbf{m}_{i})]
	\label{eqn:final_update_formula_for_input_masks}\\
	\Delta \mathbf{n}_{i}(t)=\frac{a(t)}{mnhw}\sum_{m,n, h, w}[\Delta m_{es}(\mathbf{p}_{m_h,n_w}, \mathbf{\hat{y}}_{m_h,n_w,i}, \mathbf{n}_{i})]
	\label{eqn:final_update_formula_for_masks_between_neurons}
	\end{gather}
\end{ceqn}
\autoref{eqn:simple_hebbian_learning} describes simple Hebbian learning in vector notation. \autoref{eqn:simple_hebbian_learning} applies the principle ``neurons that fire together, wire together'' where $\mathbf{p}_{m, n}$ is the presynaptic neuron activities vector and $\mathbf{\hat{y}}_{m, n, i}$ is the postsynaptic neuron activities vector. This principle per se is useful for mask learning, because the procedure could find mask weights, which indicate the connection strengths between mask input and output. However, Hebb's rule is unstable. If there are any dominant signals in the network - and in our case this is likely due to \acs{BMU}s - mask weights rapidly approach numerical positive or negative infinity. In fact, it has been shown that the instability of Hebbian learning accounts for every neuron model \cite{principe1999neural}.

\autoref{eqn:ojas_rule}, known as Oja's rule \cite{oja1982simplified}, tries to prevent this problem through multiplicative normalization, where the resulting additional negative term can be seen as the forgetting term to control weight growth. This restricts the magnitude of weights to lie between 0 and 1.0, where the squared sum of weights tends to 1.0 in a single-neuron fully-connected network \cite{oja1982simplified}. For that reason, using Oja's rule in our case would lead to mask weights approaching 1.0, since each mask weight has its own input and output due to the element-wise multiplication and is therefore independent of all other mask weight updates — each mask weight forms a one-weight, single-neuron fully-connected network. To prevent this tendency towards 1.0, we make the mask weight updates dependent on each other. 

The \acs{GHA} makes its updates dependent on each other through input modification to approximate eigenvectors for networks with more than one output. In contrast to Oja's rule, which approximates the first eigenvector for networks with only one output, GHA performs a neural \acs{PCA} \cite{sanger1989optimal}.
However, in \cite{sanger1989optimal} it has been shown that the algorithm is equivalent to performing Oja learning using a modified version of the input; a particular weight is dependent on the training of the other weights only through the modifications of the input. If Oja's algorithm is applied to the modified input, it causes the $i-$th output to learn the $i$-th eigenvector \cite{sanger1989optimal} but in our case, each mask weight has its own individual input-output pair, and the update of this weight is independent of other mask weights in the same mask since a single mask weight does not see its surroundings.

Therefore, in \autoref{eqn:similar_to_gha} we use a modified input patch for mask $i$ by simply subtracting the input patch from the sum over masked outputs, which sums the filtered information of all masks when $k$ iterates over all masks. Similar to the \acs{GHA}, we additionally multiply the mask with the output before summing up, which leads to a normalization effect and drives the mask coefficients to the initialization interval $[-1, 1]$. Now each mask update tries to incorporate the missing input information in the output in a competitive manner, where the mask weight growth is restricted. This can be seen as a self-organization process for the masks. In our experiments we show that due to the summation over all masks the updates are stable enough to use Hebbian learning (\autoref{eqn:simple_hebbian_learning}) to learn the masks, which saves computation time.

In \autoref{eqn:similar_to_gha_with_restricted_summation} we subtract the sum of all $k<i$ masked postsynaptic neuron activity vectors from the presynaptic neuron activity vector. Therefore, each next mask $\mathbf{m}_i$ sees an output from which the filtered information of the previous masks $\{\mathbf{m}_{1},...,\mathbf{m}_{i-1}\}$ in the grid is subtracted. The hyperparameter $\gamma$ controls  how much information we want to subtract from the input. A smaller $\gamma < 1.0$ value enables the masks to share more input information. 

\autoref{eqn:final_update_formula_for_input_masks} shows the final update formula for the input masks (a), where we compute the mean over every patch $\mathbf{p}_{m,n}$. For the second mask type (b), the final update formula is shown in \autoref{eqn:final_update_formula_for_masks_between_neurons}, where we instead update the mask $\mathbf{n}_i$ with the mean over each spatial activation vector $\mathbf{p}_{m_h,n_w}$ of every patch. Experiments show that the update rules \ref{eqn:final_update_formula_for_input_masks} and \ref{eqn:final_update_formula_for_masks_between_neurons} lead to slightly better performance for smaller models when using the input modification of \autoref{eqn:similar_to_gha_with_restricted_summation} and $\gamma=0.5$. However, \autoref{eqn:similar_to_gha_with_restricted_summation} requires Oja's rule and sometimes larger neighborhood coefficients - especially when $\gamma$ is close to $1.0$ - because of the unequal learning opportunities of the masks, which can lead to poorer efficiency. With the other input modification (\autoref{eqn:similar_to_gha}) and $\gamma=1.0$, we achieve better performance for our deeper models.

Furthermore, we want to note that the mask weights are initialized uniformly in $[-1,1]$ in our experiments to encourage the model to learn negative mask weights, which allow the masks to flip their input. For stability reasons, it is recommended that the input is normalized. It is also possible to multiply $\mathbf{m}_{i}$ by $h_{m, n, i}(t)$ to obtain similar masks for neighboring neurons, but experiments show that the resulting loss in diversity decreases performance. Therefore we only update the \acs{BMU} mask.

\subsection{Multi-Head Masked SConv-Layer}
To further improve the performance of our models, we use multiple smaller \acs{SOM} maps per layer instead of one big \acs{SOM} map per layer. For the output of a layer, the outputs of the \acs{SOM}s are concatenate in the feature dimension. We use multiple maps for multiple reasons: First, the neuron's mask should lead to specialization, and in combination with the self-organization process, a single \acs{SOM} is possibly still unable to learn enough distinct features. Second, multiple smaller maps seem to learn more distinct features more efficiently than scaling up a single map.
This approach is flexible; experiments show similar performance between \acs{SOM}s using $3$ big maps or $12$ small maps per layer. To further increase diversity, we update only the best \acs{BMU} of all maps per patch. This could result in maps that never update or maps that stop updating too early during training. We do not use a procedure to prevent these dead maps, as we have not had any problems with these during our experiments. Furthermore, dead maps can be prevented to some extent by scaling up the coefficient of the neighborhood function \ref{eqn:nfunction}. It is a task for future research to investigate dead maps and whether the use of multiple \acs{SOM}s in the \acs{CSNN} model results in redundant information
\\\\
All the presented learning methods can be applied layer-wise. On the one hand, this brings benefits like the opportunity to learn big models by computing layer by layer on a \acs{GPU}, and more independent weight updates compared to gradient descent, which reduces the Scalar Bow Tie Problem. On the other hand, this independence could lead to fewer ``paths'' through the network, reducing the capability of the network to disentangle factors of variation. Examining this problem by creating and comparing layer-wise dependent and independent learning rules is an interesting direction for future research, and some work has already been done (e.g., \cite{lillicrap2016random}).

\subsection{Other Layers and Methods}
Since we are in the regime of convolutional deep learning architectures, there are plenty of other methods we could test within the context of our modules. In our experiments, we investigate batch normalization \cite{ioffe2015batch} (with no trainable parameters) and max pooling \cite{krizhevsky2017imagenet}.
In addition, \acs{SOM}s are well studied, and there are many improvements over the standard \acs{SOM} formulation that could be explored in future work (e.g., the influence of different distance metrics).
\section{Experiments}
To assess our proposed methods, we design two \acs{CSNN} models for our experiments. During the training of these models, no additional strategies, such as learning rate schedules or regularization, are utilized. We simply normalize the datasets to zero mean and unit variance. Our implementation is in TensorFlow and Keras and is available at \href{https://github.com/BonifazStuhr/CSNN}{https://github.com/BonifazStuhr/CSNN}.

\subsection{Datasets}
\textit{(1) Cifar10} \cite{krizhevsky2009learning}, with its $32\times32$ images and 10 classes, is used to ablate the importance of the proposed building blocks. We split the validation set into $5000$ evaluation and test samples.
\\\\
\textit{(2) Cifar100} \cite{krizhevsky2009learning}, with its $32\times32$ images and 100 classes, is used to test the capability of our representation to capture a higher number of classes. We split the validation set into $5000$ evaluation and test samples.
\\\\
\textit{(3) Tiny ImageNet} \cite{le2015tiny}, with its $64\times64$ images and 200 classes, is used to investigate the capability of our models to learn on a dataset with a higher number of more complex classes.
\\\\
\textit{(4) SOMe ImageNet} is a subset of ImageNet \cite{deng2009imagenet} containing $10100$ training, $1700$ evaluation, and $1700$ test samples of 10 classes to test our performance on larger $256\times256$ images. We refer to our implementation for details about this dataset.

\subsection{Evaluation Metrics}

\textbf{Quantitative evaluation.} We use the linear and nonlinear evaluation protocols to evaluate our learning rules, building blocks, and models. Thereby, we use the representations of the frozen models to train linear and nonlinear classifiers for the target task. Additionally, we train a few-shot classifier on our frozen models to test the few-shot capabilities of the learned representations. We test generalization capabilities by transferring frozen \acs{CSNN} models trained on one dataset to another by training linear, nonlinear, and few-shot classifiers for the other dataset. To further evaluate our models, we compute the batch-wise neuron utilization - the percentage of used neurons per batch. In addition, we provide random (R) baseline performances for our models. At this point, we want to highlight the importance of random baselines since it has been shown that models with random weights can often achieve performances just a few percent worse than sophisticated unsupervised learning methods (e.g., \cite{saxe2011random}). Furthermore, we prove the stability of our learning processes by reporting 10-fold cross-validation results.
\\\\
\textbf{Qualitative evaluation.} To visually examine the capabilities of our representations, we train a decoder on the frozen \acs{CSNN} models to reconstruct the input. Furthermore, we compare the average representations for each class by visualizing the \acs{CSNN} layers in image space. In addition, we examine neuron activities with respect to specific classes of the input image by converting the weights of \acs{BMU}s for each input patch into image space for different \acs{CSNN} layers during training.

\subsection{Model Architectures}
(1) \textit{S-CSNN.} Our small model consists of two \acs{CSNN}-layers, each is followed by a batch normalization and a max-pooling layer to halve the spatial dimension. The first layer uses a $10\times10\times1$ \acs{SOM} grid (1 head with 100 \acs{SOM} neurons), stride $2\times2$, and input masks (\autoref{eqn:final_update_formula_for_input_masks}) the second layer a $16\times16\times1$ \acs{SOM} grid, stride $1\times1$ and masks between neurons (\autoref{eqn:final_update_formula_for_masks_between_neurons}). Both layers use a kernel size of $3\times3$ and the padding type ``same''.
\\\\
(2) \textit{D-CSNN.} Our deeper model consists of three \acs{CSNN}-layers, each is followed by a batch normalization and a max-pooling layer to halve the spatial dimension. The first layer uses a $12\times 12\times 3$ \acs{SOM} grid and input masks (\autoref{eqn:final_update_formula_for_input_masks}), all remaining layers use masks between neurons (\autoref{eqn:final_update_formula_for_masks_between_neurons}). The second layer consist of a $14\times 14\times 3$ \acs{SOM} grid, and the third layer of a $16\times 16\times 3$ \acs{SOM} grid. All layers use a kernel size of $3\times3$, a stride of $1\times1$, and the padding type ``same''. For Tiny ImageNet the layer 1 stride is $2\times2$ and for SOMe ImageNet the layer 1 and 2 strides are $3\times3$ to keep the representation sizes the same. 

\subsection{Training and Evaluation Setup}
\textbf{Training.} We set the learning rate of the \acs{SOM}s to $0.1$ and the learning rate of the local mask weights to $0.005$ for all layers. The neighborhood coefficients for each layer are set to $(1.0,1.25)$ for the S-CSNN and $(1.0,1.5,1.5)$ for the D-CSNN. The individual layers are learned bottom up, one after another, since its easier for deeper layers to learn, when the representations of the previous layers are fully learned. The training steps for each individual layer can be defined in a flexible training interval. 
\\\\
\textbf{Classifiers.} Our nonlinear classifier is a three-layer multilayer perceptron \acs{MLP} (512, 256, num-classes) with batch normalization and dropout $(0.5,0.3)$ between the layers. Our linear and few-shot classifier is simply a fully-connected layer (num-classes). All classifiers use elu activation functions between layers and are trained using the Adam optimizer \cite{kingma2014adam} with standard parameters and a batch size of $512$. To infer the results on the test set, we use the model of the training step where the evaluation accuracy is the best (pocket algorithm approach).

\begin{figure}[t]
	\begin{center}
		\includegraphics[width=1.0\linewidth]{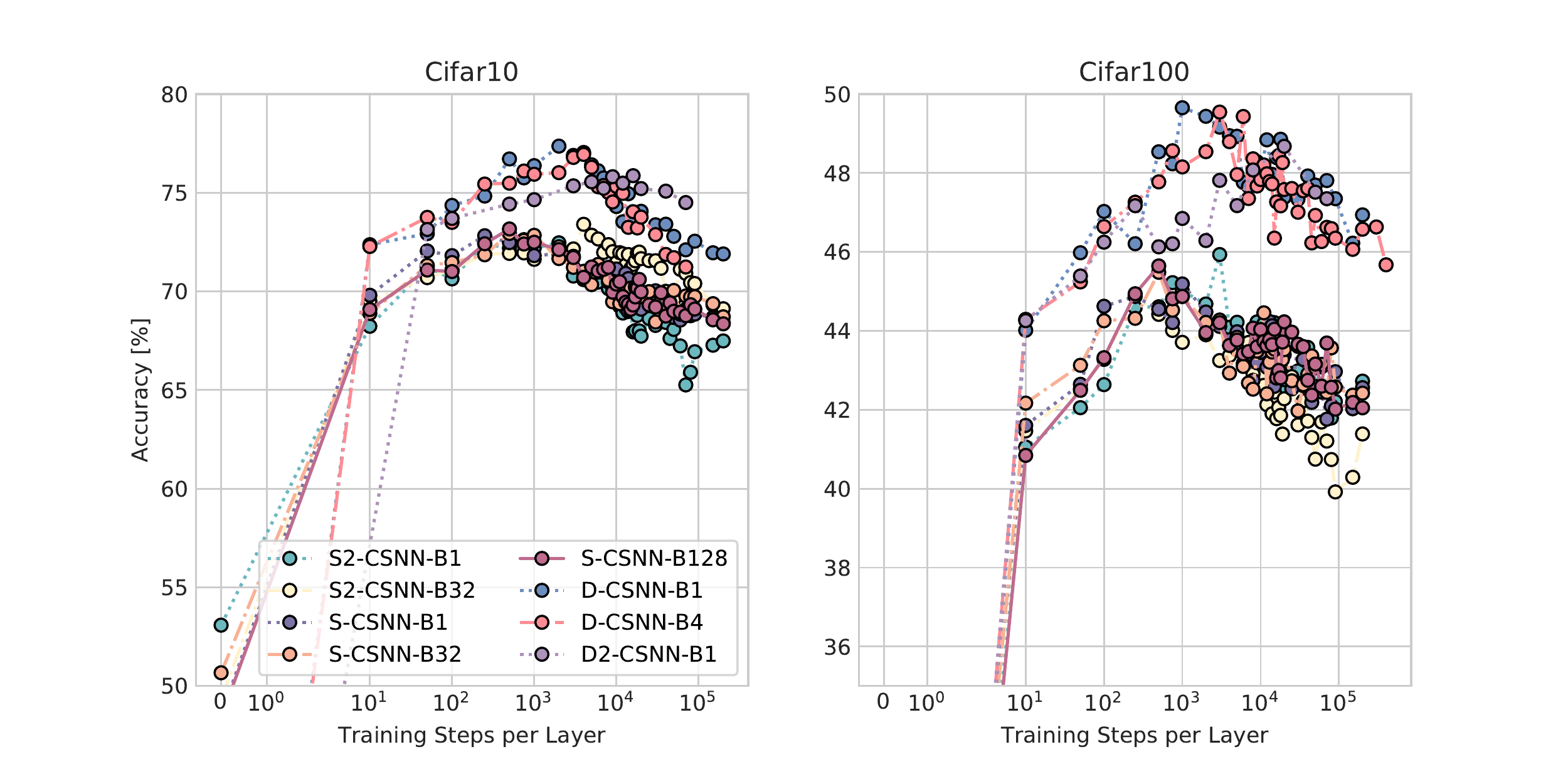}
	\end{center}
	\vspace{-3ex}
	\caption[Accuracies of our models trained with our local mask learning rules and different batch sizes.]{Accuracies of our models trained with our local mask learning rules (S/D for \autoref{eqn:similar_to_gha} and S2/D2 for \autoref{eqn:similar_to_gha_with_restricted_summation}) and different batch sizes (B). Every point corresponds to the test dataset accuracy of a nonlinear classifier that has been trained on the representation of the \acs{CSNN} for every sample in the training dataset. The layers of the \acs{CSNN} have been trained layer-wise for the steps shown on the x-axis. Best viewed in color.}
	\label{fig:csnns:ofm}
\end{figure}

\subsection{Objective Function Mismatch}
\autoref{fig:csnns:ofm} shows that all models reach their accuracy peak early in training. Since this peak is often reached before seeing the entire training set and the models show ongoing SOM weight convergence (see \autoref{fig:csnns:app:weight_changes} of \autoref{app:00}), it seems likely that the decrease in performance is due to an objective function mismatch between the self-organization process and the objective function to learn features for classification. This is a common problem in unsupervised learning techniques \cite{metz2018meta}. On the positive side, it could be argued that our models require fewer samples to learn representations (e.g., 2000 samples ($\approx. 3\%$) from Cifar10 for the D-CSNN-B1 peak). Even more surprising is the significant jump in accuracy (by up to 62\%) after the models have only seen $10$ examples. This could be a positive effect of reducing the Scalar Bow Tie Problem. However, we would also like to point out that better initialization and no dropout in the nonlinear classifier lead to better initial performance. Furthermore, we can see that the mask learning rule \ref{eqn:similar_to_gha} is superior in this setting for the deeper model and vice versa.

\begin{table}[h]
	\RawFloats
	\caption[Ablation study on the Cifar10 dataset.]{Ablation study on the Cifar10 dataset.} 
	\vspace{-1ex}
	\label{tab:csnns:ablation_lr_aug_som_other}
	\begin{center}
		\scalebox{0.8}{%
			\setlength{\tabcolsep}{0.4em}
			\begin{tabular}{lccclccc}
				\toprule
				\multirow{2}[2]{*}{Method} & \multirow{2}[2]{*}{Model} &   \multicolumn{2}{c}{Cifar10} & \multirow{2}[2]{*}{Method} & \multirow{2}[2]{*}{Model} &   \multicolumn{2}{c}{Cifar10}   \\
				\cmidrule(lr){3-4}
				\cmidrule(lr){7-8}
				{} & {} & linear & nonlinear & {} & {} & linear & nonlinear  \\
				\midrule
				\multicolumn{8}{l}{\textbf{\textit{Learning rule ablations}}:}\\  
				w/ M	&S&   $\mathbf{66.43\substack{+0.96 \\ -1.02}}$ &$72.79\substack{+0.59 \\ -0.84}$  &w/ M&  D &$72.66\substack{+0.40 \\ -1.07}$ &$77.21\substack{+0.62 \\ -0.56}$    \\
				w/ M &S2	&$66.18\substack{+1.17 \\ -0.95}$    &$72.82\substack{+1.02 \\ -1.22}$ &	 	w/ M&D2 &$71.74\substack{+0.99 \\ -1.42}$  &$76.18\substack{+0.77 \\ -1.61}$  \\
				w/ RM	&S2	&$62.89$    &$70.42$    &w/ RM	&D	&$69.86$    &$73.69$ \\  
				w/ NOM	&S2	&$56.84$    &$14.83$    &w/ NOM	&D	&$65.53$   	&$20.22$ \\
				w/o M	&S2	&$40.37$    &$22.32$    &w/o M	&D	&$43.33$    &$12.13$ \\
				RS w/ M 	&S2	&$65.20$    &$67.75$    &RS w/ M&D	&$69.69$    &$69.74$ \\
				RS w/o M &S2		&$54.64$    &$15.32$   &RS w/o M	&D	&$61.86$    &$16.01$ \\  
				RS w/ RM	&S2	&$56.42\substack{+1.17 \\ -1.46}$   &$52.23\substack{+2.16 \\ -2.48}$  &RS w/ RM&D	&$63.36\substack{+1.28 \\ -1.34}$     &$13.44\substack{+5.25 \\ -3.22}$ \\ 
				\midrule
				\multicolumn{8}{l}{\textbf{\textit{Augmentation ablations}}:}\\
				w/ aug &S2   &$66.32$   &$\mathbf{76.90}$  	& w/ aug &D	& $\mathbf{73.73}\substack{+1.95 \\ -1.17}$  	 &$\mathbf{80.03\substack{+1.52 \\ -1.05}}$     \\
				\midrule	
				\multicolumn{8}{l}{\textbf{\textit{SOM ablations}}:}\\
				1$\times$SOM& D 	&$69.69$    &$71.86$  & 12$\times$SOM &M 	&$71.28$     &$75.77$ \\
				2$\times$SOM &D &	$72.05\substack{+0.67 \\ -0.99}$   &$76.11\substack{+0.49 \\ -0.57}$ &w/ a.u. &  D &  $73.14\substack{+0.73 \\ -0.70}$     &$76.88\substack{+0.59 \\ -0.90}$   \\
				\midrule	
				\multicolumn{8}{l}{\textbf{\textit{Other ablations}}:}\\
				w/ M n.u.  & D& $71.70$     &$75.08$  &	w/o bn&  D & $71.67$ 	      &$76.16$ \\
				\bottomrule
		\end{tabular}}
	\end{center}
	\vspace{-2ex}
\end{table}

\subsection{Ablation Study}
In \autoref{tab:csnns:ablation_lr_aug_som_other}, we show the influence on target task performance of the proposed learning rules and building blocks for Cifar10. 
\\\\
\textbf{Learning rule ablations.} The first two rows (w/ M) show the performance of our small models (S/S2) and deep models (D/D2) with fully trained \acs{SOM}s and masks. We observe that the deeper model outperforms the smaller model and that the Hebbian mask learning rule \ref{eqn:similar_to_gha} (S/D) slightly outperforms the Oja mask learning rule \ref{eqn:similar_to_gha_with_restricted_summation} (S2/D2). When we successively remove learning procedures from the small and big models, we observe the following: Not training the mask and initializing them randomly at the beginning of training (w /RM) or treating them as random noise for each training step (w /NOM) decreases performance notably. We generated these mask from a uniform distribution in $[-1, 1]$. However, when we train the model completely without masks (w/o M), the performance decreases significantly compared to the models with trained and untrained masks. This reassures the argumentation that \acs{SOM} convolutional filters may need masks to focus on or to ignore certain parts of the input - even adding untrained masks leads to performance improvements.

When the \acs{SOM} remains untrained, but the masks are trained (RS w/ M), we observe a performance decrease as well compared to our fully trained model (w/ M). This shows that our \acs{SOM} convolutional filters can learn useful features during training. When the \acs{SOM} weights remain untrained and no masks are used (RS w/o M), the performance decreases further, again showing the importance of the masks.

Our overall learning increases accuracy by about 10\% for the linear classifier and 20-65\% for the nonlinear classifier compared to the uniform (in $[-1, 1]$) random baseline with randomly initialized masks (RS /w RM). The low accuracy of the nonlinear classifier for random weights can be attributed to the use of dropout. Furthermore, the overall learning increases accuracy by about 20-26\% for the linear classifier and 50-65\% for the nonlinear classifier compared to the model without masks but with trained \acs{SOM} convolutional filters (w/o M). Without masks, the \acs{SOM} learning procedure results in a performance decrease compared to the random baseline (RS /w RM). Only when utilizing random masks, noise masks, or learned masks does \acs{SOM} learning provide an advantage to the overall training process for our simple \acs{CNN}-like architecture that does not use further clustering-specific or \acs{SOM}-specific tricks.
\\\\
\textbf{Augmentation ablations.} Using data augmentation (rotations, shifts, and horizontal flips) during training to create a larger dataset of representations for the linear and nonlinear classifiers increases performance in most cases (w/ aug).
\\\\
\textbf{SOM ablations.} Increasing the number of \acs{SOM} maps (1$\times$\acs{SOM} to 2$\times$\acs{SOM} for model D) improves performance. Furthermore, we can construct a smaller model with 12 \acs{SOM} maps per layer that results in comparable performance to our larger model with two \acs{SOM} maps. Analogous to \cite{kolesnikov2019revisiting}, we observe that increasing the representation size improves performance, but in contrast, the classifier does not take longer to converge on larger representation sizes in our case. In addition, we find that in contrast to bottom-up training, updating all maps per step (w/ a.u.) increases performance slightly for the linear classifier and decreases performance slightly for the nonlinear classifier. 
\\\\
\textbf{Other ablations.} When we not only update the BMU masks but also update neighboring masks, we observe a performance decrease. Again SOM-neuron-specific masks seem to be crucial. Not using batch normalization (w/o bn) also decreases performance. We note that for deeper models, batch normalization is necessary to prevent too large values in deeper layers (exploding weights).

\begin{table}[t]
	\RawFloats
	\caption[Quantitative comparison with previous work.]{Quantitative comparison with previous work. With BH, we refer to binary hashing, with H to histograms, and with W to whitening.}
	\vspace{-1ex} 
	\label{tab:csnns:quantitativ_results_comparison}
	\begin{center}
		\scalebox{0.8}{%
			\setlength{\tabcolsep}{0.4em}
			\begin{tabular}{lcccc}
				\toprule
				\multirow{2}[2]{*}{Method} &   \multicolumn{2}{c}{Cifar10} & Cifar100 & Tiny ImageNet \\
				\cmidrule(lr){2-3}
				\cmidrule(lr){4-4}
				\cmidrule(lr){5-5}
				{} 												&linear 				&nonlinear 			&nonlinear 			&nonlinear\\
				\midrule
				\textbf{\textit{Backpropagation}}:&&&&\\
				
				VAE \cite{hjelm2018learning}					&$54.45$ (SVM)    		&$60.71$    		&$37.21$    		&$18.63$\\
				BiGAN \cite{hjelm2018learning}					&$57.52$ (SVM)    		&$62.74$    		&$37.59$     		&$24.38$\\
				DIM(L) \cite{hjelm2018learning} 				&$64.11$ (SVM) 			&$80.95$  			&$49.74$ 			&$\mathbf{38.09}$\\
				AET \cite{zhang2019aet} &$\mathbf{83.35}$ (FC) 	&$\mathbf{90.59}$ 	&- 					&-  	\\	
				\midrule
				\textbf{\textit{Backpropagation-free}}:&&&&\\
				K-means Triangle+W \cite{coates2011analysis} 	&$79.60$ (SVM) 			&- 					&- 					&-   	\\
				SOMNet+BH+H \cite{hankins2018somnet}		&$71.81$ (SVM)   		&-      			&-  				&-      \\
				D-CSNN-B1 (ours)								&$73.73$ (FC)   		&$77.83$    		&$49.80$   			&$3.56$ \\
				D-CSNN-B1-Aug (ours) 							&$75.68$ (FC)  			&$81.55$   			&$\mathbf{52.66}$   &$5.20$ \\
				\bottomrule
		\end{tabular}}
	\end{center}
	\vspace{-2ex}
\end{table}
\subsection{Comparison to the State of the Art}
\autoref{tab:csnns:quantitativ_results_comparison} shows the performance of our best models compared to other methods. For few-shot learning, 50 samples per class are used. The CSNN models used for Cifar100, Tiny ImageNet, and SOMe ImageNet classification have not been tuned for these datasets and have been trained with the same hyperparameters as our Cifar10 model (except the training steps for Cifar100). We show that the performance of our D-CSNN-B1 model is comparable to many state-of-the-art methods at the time of publication. However, our models lack performance for Tiny ImageNet. One possible explanation could be that more complex datasets may require deeper models or improved learning techniques. Overall, we achieve an improvement over the SOMNet baseline without any additional architectural tricks such as binary hashing, histograms, or whiting. Furthermore, it should be noted that we report linear performance with a simple, fully-connected layer instead of an SVM \cite{cortes1995support} to follow the recent trend. SVMs are considered linear models, but they can also solve nonlinear problems using the kernel trick \cite{cortes1995support}.

\begin{table}[t]
	\RawFloats
	\scriptsize
	\caption[Quantitative results of models transferred to other datasets.]{Quantitative results of models transferred to other datasets. (top rows) Performance of the models trained directly on the target datasets. (bottom rows) Performance on models trained on Cifar10 and Cifar100 transferred to the other target datasets by re-training the classifiers for each target dataset.} 
	\vspace{-1ex}
	\label{tab:csnns:model_transfer_quantitativ_results}
	\begin{center}
		\scalebox{0.9}{%
			\setlength{\tabcolsep}{0.4em}
			\begin{tabular}{lcccccccccccc}
				\toprule
				\multirow{2}[2]{*}{Method} &   \multicolumn{3}{c}{Cifar10} & \multicolumn{3}{c}{Cifar100} & \multicolumn{3}{c}{Tiny ImageNet} & \multicolumn{3}{c}{SOMe ImageNet} \\
				\cmidrule(lr){2-4}
				\cmidrule(lr){5-7}
				\cmidrule(lr){8-10}
				\cmidrule(lr){11-13}
				{} 				&fs(50) &linear &nonlinear 			&fs(50) &linear &nonlinear 	&fs(50) &linear &nonlinear 	&fs(50) &linear &nonlinear 	\\
				\midrule
				D-CSNN-B1		&$\mathbf{41.59}$&$73.73$&$77.83$ 	&$27.29$&$45.17$&$49.80$     &$13.70$&$13.52$ &$3.56$   &$\mathbf{68.92}$ &$70.69$ &$49.99$\\
				D-CSNN-B1-Aug&- &$\mathbf{75.68}$  &$\mathbf{81.55}$       &-    &$\mathbf{46.82}$      &$\mathbf{52.66}$      &-   &$13.32$    &$\mathbf{5.20}$      &- 	&$\mathbf{70.80}$    &$\mathbf{75.29}$    \\
				\midrule
				D-CSNN-B1-Cifar10&$\mathbf{41.59}$ 	&$73.73$    &$77.83$      &$\mathbf{27.96}$     &$45.68$    &$49.70$     &$13.46$     &$\mathbf{14.36}$    &$2.03$     &$47.65$  &$65.34$    &$65.59$  	\\
				D-CSNN-B1-Cifar100 &$41.43$ 	&$71.60$    &$76.46$      &$27.29$     &$45.17$    &$49.80$    &$\mathbf{13.72}$     &$13.57$    &$1.61$      &$45.68$  &$63.79$    &$65.13$      \\
				\bottomrule
		\end{tabular}}
	\end{center}
	\vspace{-6ex}
\end{table}
\subsection{Generalization}
To test generalizability, we train models on one dataset and retrain the classifiers for each dataset to which we transfer the models. Following that, we report the test set performance of the dataset for which we transferred the models. As shown in \autoref{tab:csnns:model_transfer_quantitativ_results}, our models achieve surprising generalization capability, where differences in accuracy between the model trained and the model transferred to the dataset lie in the low percentage range. Furthermore, we observe good performance on the larger  $256\times256$ SOMe ImageNet images. 

\begin{figure}[H]
	\begin{center}
		\includegraphics[width=1.0\linewidth]{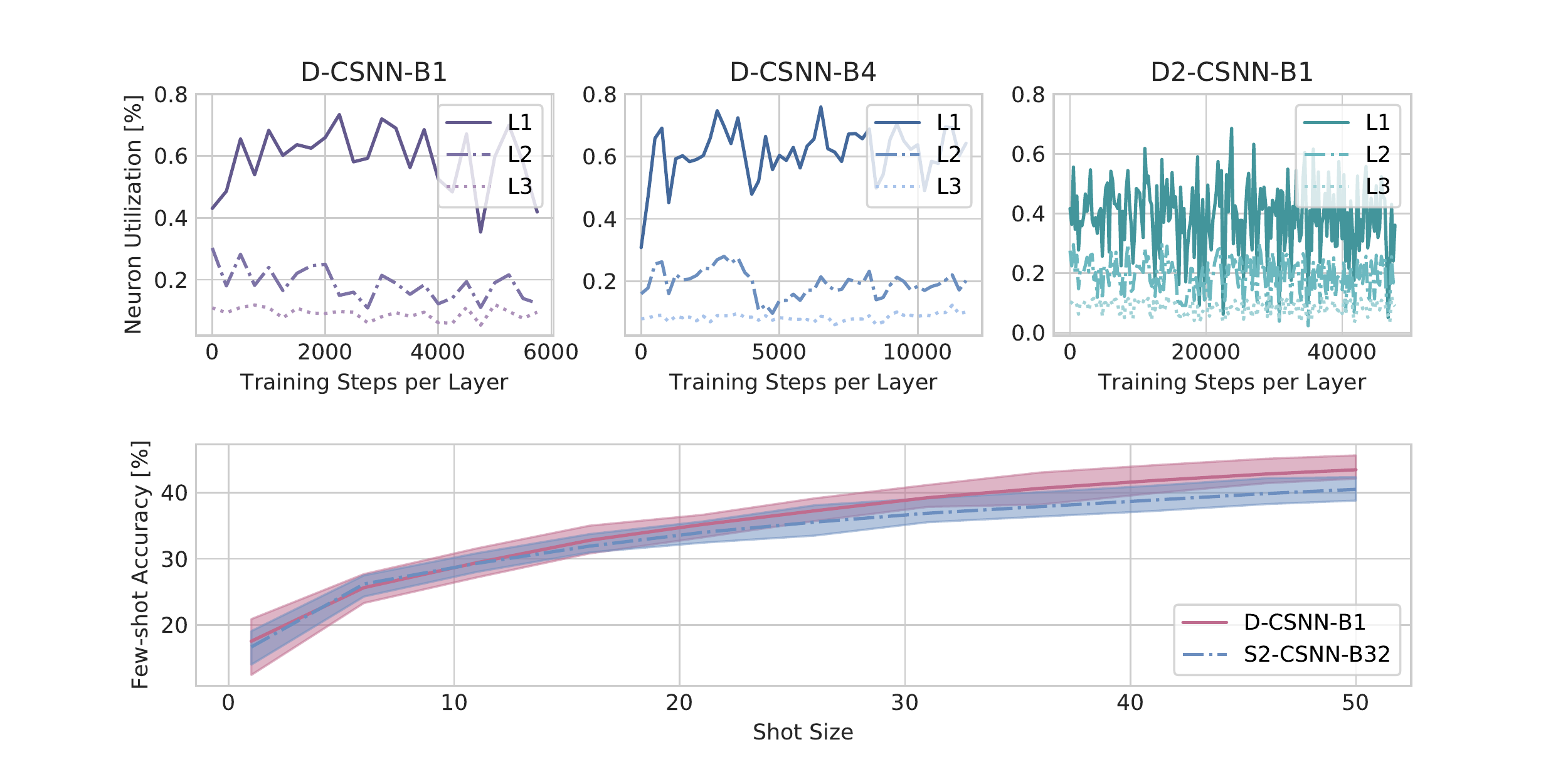}
	\end{center}
	\vspace{-3ex}
	\caption[Neuron utilization and few-shot learning.]{Neuron utilization and few-shot learning. (top) Neuron utilization of models trained with various batch sizes and mask learning rules on Cifar10. (bottom) 10-fold few-shot classifier accuracy of our models with increasing shot sizes from $1$ to $50$ on Cifar10.}
	\label{fig:csnns:neuron_utilization_fs_learning}
\end{figure}

\subsection{Neuron Utilization}
In the upper part of \autoref{fig:csnns:neuron_utilization_fs_learning}, we observe that the neuron utilization — the percentage of used neurons per input — is getting smaller for deeper layers, indicating that the SOM neurons of deeper layers specialize to certain inputs. Compared to learning rule \ref{eqn:similar_to_gha_with_restricted_summation}, learning rule \ref{eqn:similar_to_gha} leads to a higher neuron utilization, which may be explained by the forgetting term of Oja's rule, which is not present in Hebb's rule. We want to note that a higher neuron utilization does not necessarily lead to higher performance since a lower neuron utilization indicates sparser representations that may be more distinguishable.

\subsection{Few-shot Performance}
The lower part of \autoref{fig:csnns:neuron_utilization_fs_learning} shows the increase in accuracy of the linear classifier as the number of representations per class used to train the linear classifier increases. Our models reach an accuracy of 41.49\% on Cifar10, 27.29\% on Cifar100, 13.70\% on Tiny ImageNet, and 68.92\% on SOMe ImageNet when trained on 50 examples per class. The comparatively higher accuracy in SOMe ImageNet indicates that the model can utilize the additional patches obtained from the larger input images.

\begin{figure}[t]
	\begin{center}
		\includegraphics[width=1.0\linewidth]{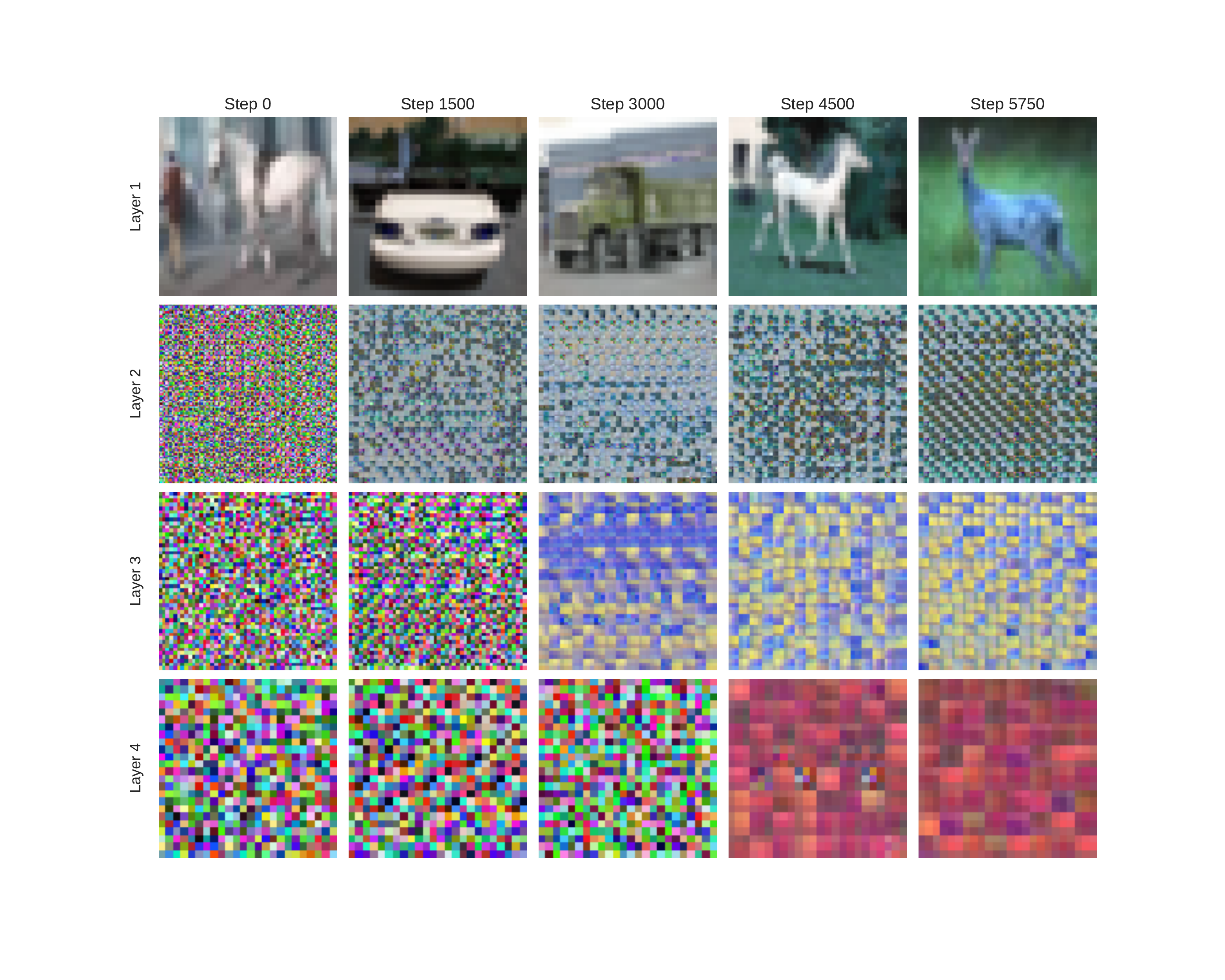}
	\end{center}
	\vspace{-8ex}
	\caption[BMU images for each layer of the D-CSNN model during training.]{\acs{BMU} images for each layer of the D-CSNN model during training. Images have been created by replacing each patch with the \acs{SOM} weight of the BMU, which is reshaped into the 3D patch shape (RGB image). For deeper layers, we only show a slice of depth three through the BMU weight, since its hard to visualize kernel depths larger then three. Since the D-CSNN-B1 model contains three maps, we use the best BMU from all maps of the layer. Best viewed in color.}
	\label{fig:csnns:app:bmu_images_1}
\end{figure}
\begin{figure}[t]
	\begin{center}
		\includegraphics[width=1.0\linewidth]{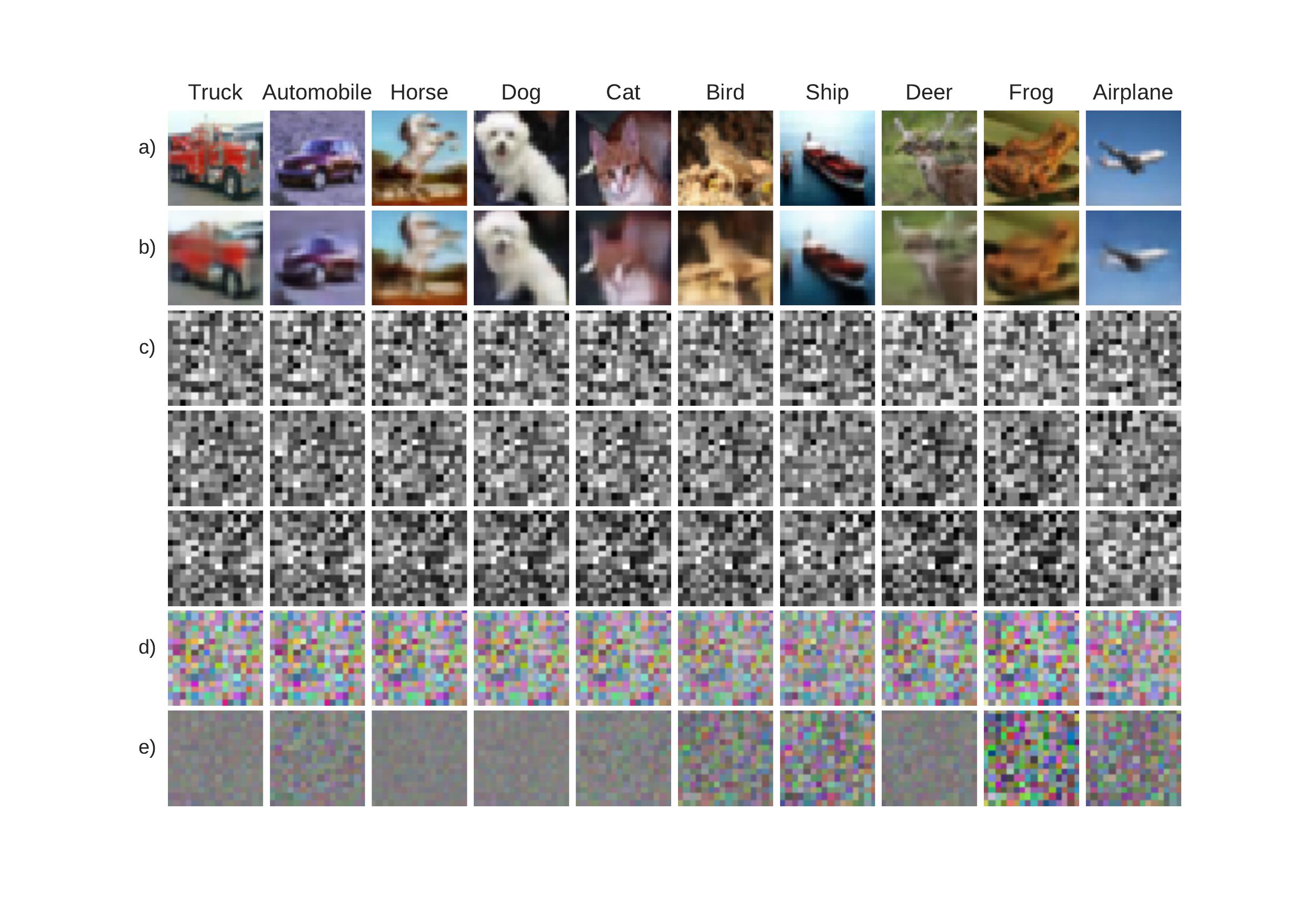}
	\end{center}
	\vspace{-7ex}
	\caption[Representations and reconstructions.]{Representations and reconstructions. a) The Cifar10 input image. b) The reconstruction of the input image using the representation of the D-CSNN-B1 model. c) The average representation per map for each class created by taking the mean over the spatial dimension of each test sample's representation and reshaping the resulting vector to \acs{SOM} grid shape. d) RGB image of the average representation (three maps lead to an RGB image). e) Differences of the average representations calculated by element-wise subtracting the average representation of the respective class from the average representation of the class on the right-hand side. Best viewed in color.}
	\label{fig:csnns:representations_reconstructions}
\end{figure}
\subsection{Visual Interpretations}
In \autoref{fig:csnns:app:bmu_images_1} we show BMU images for each layer during the training of the D-CSNN model. These images are created by replacing the image patch with the reshaped BMU weights or a slice of these weights. For earlier training steps, \acs{BMU}s that have been calculated from randomly initialized neurons can be seen for layers 2 - 4. After training the individual layers, patterns emerge. For example, for layer 1, we observe patterns corresponding to the content of specific regions of the input image. These patterns become less interpretable at deeper layers. Moreover, we observe that the \acs{SOM} weights are not only sensitive to colors since different backgrounds result in similar \acs{BMU} patterns, for example, in image 5 of layer 2.
\\\\
As shown in row b) of \autoref{fig:csnns:representations_reconstructions}, the input images can be reconstructed with a high level of detail from the representation of our unsupervised CSNN models. This is surprising and shows the amount of information present in the representation of our unsupervised CSNN models. Moreover, in rows c), d), and e), it can be observed that "similar" classes lead to similar average representations: 1) In rows c) and d) exist similar spots with high average neuron activities. 2) The classes have been sorted according to $L_1$ distance of their average representation from the average representation of the truck class. The average car class representation has the shortest distance to the average truck class representation, followed by the average representations of "similar" looking animal classes. 3) Row e) shows images where the average representation of the right-hand class has been subtracted element-wise from the average representation of the respective class. For example, the first column of e) equals the truck's average representation minus the car's average representation. Therefore, row e) shows the remaining distinguishable differences between average representations of the respective class and the right-hand class. We observe that this remaining information is lower when a "similar" class is on the right-hand side (e.g., dog minus cat).
\\\\
Additional quantitative and qualitative results can be found in \autoref{app:00}.

\section{Conclusion}
In this chapter, we have introduced the modular building blocks of CSNNs to learn representations in an unsupervised manner without backpropagation. With the proposed CSNN modules and learning rules — which combine CNNs, SOMs and Hebbian learning of masks — a new, alternative way for learning unsupervised feature hierarchies has been explored. Along the way, we have discussed the Scalar Bow Tie Problem and the objective function mismatch: two problems in the field of deep learning that we believe can potentially be solved together and provide an interesting direction for future research. 

\part{Evaluating Visual\\Unsupervised Repre-\\sentation Learning} 
\graphicspath{{./main/5_chapter01/sections/figures/}}

\chapter{Investigating the Objective Function Mismatch}
\label{chap:01}
\vspace{-8mm}
\begin{abstract}
Finding general evaluation metrics for unsupervised representation learning techniques is a challenging open research question, which has recently become more and more necessary due to the increasing interest in unsupervised methods. Even though these methods promise beneficial representation characteristics, most approaches currently suffer from the objective function mismatch. This mismatch states that the performance on a desired target task can decrease when the unsupervised pretext task is learned too long - especially when both tasks are ill-posed. In this chapter, we build upon the widely used linear evaluation protocol and define new general evaluation metrics to quantitatively capture the objective function mismatch and the more generic metrics mismatch. We discuss the usability and stability of our protocols on a variety of pretext and target tasks and study mismatches in a wide range of experiments. Thereby we disclose dependencies of the objective function mismatch across several pretext and target tasks with respect to the pretext model's representation size, target model complexity, pretext and target augmentations, as well as pretext and target task types. In our experiments, we find that the objective function mismatch reduces performance by $\sim$0.1-5.0\% for Cifar10, Cifar100, and PCam in many setups and up to $\sim$25-59\% in extreme cases for the 3dshapes dataset.
\end{abstract}
\section{Motivation}
Unsupervised representation learning is a promising approach to learn useful features from huge amounts of data without human annotation effort. Thereby a common evaluation pattern is to train an unsupervised pretext model on different datasets and then test its performance on several target tasks. Because of the huge variety of target tasks and preferred representation characteristics, the evaluation of these methods is challenging. In preceding work, several evaluation metrics have been proposed \cite{locatello2019challenging,hjelm2018learning,palacio2019evaluation,lorena2019complex}, but due to the fast changes in unsupervised learning methodologies, only a few of them can be used across the wide spectrum of promising approaches. This is one reason why the linear evaluation protocol is commonly used \cite{gidaris2018unsupervised,kolesnikov2019revisiting,zhang2019aet,donahue2019large,patacchiola2020self,he2020momentum,chen2020simple}, which trains a linear model for a target task on-top of the representations of an unsupervised pretext model. In this chapter, we show that simply training a target model for different layers of the pretext model does not yield the entire picture of the training process and leads to a loss of useful temporal information about learning. It is already known in literature that succeeding in a pretext task can be the reason why the model fails on the target task. Here we propose that the linear evaluation protocol does not capture this properly. 
\section{Contributions}
\begin{figure}[t]
	\begin{center}
		\includegraphics[width=0.8\linewidth]{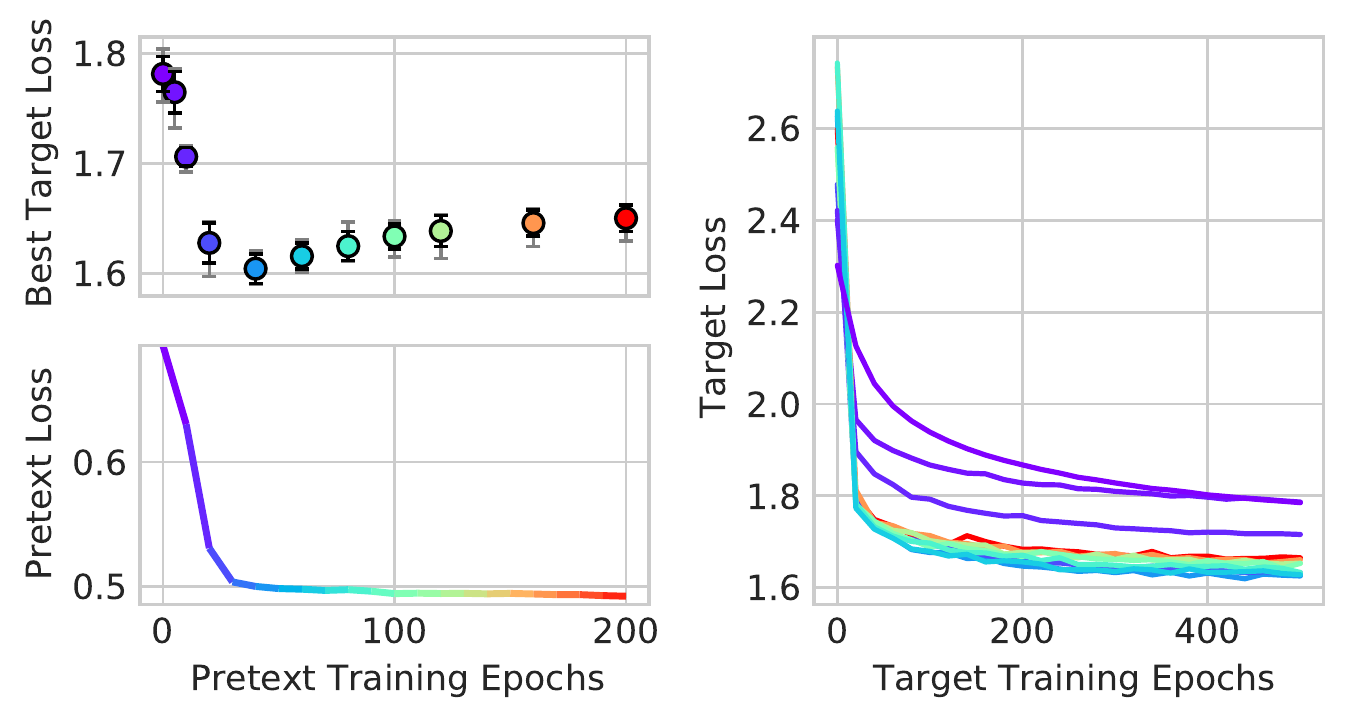}
	\end{center}
	\vspace{-2ex}
	\caption[The objective function mismatch.]{The objective function mismatch. (bottom left) Evaluation loss of a pretext autoencoder trained on Cifar10. (top left) Best evaluation losses of linear target models trained for classification on the representations of the pretext autoencoder from different pretext training epochs. (right) Evaluation loss curves from each linear target model. Colors correspond to the current epoch of pretext task training, and each value is obtained by 5-fold cross-validation. An objective function mismatch occurs around pretext training epoch 40, from which the target loss increases. Best viewed in color.}
	\label{fig:ofm:ofm}
\end{figure}
We extend the linear and nonlinear evaluation protocols and address the question of when succeeding in a pretext task hurts performance and how much. We train target models on representations obtained from different training steps or epochs of the pretext model and plot target and pretext model metrics in comparison, as shown in \autoref{fig:ofm:ofm}. Thereby we observe that training an unsupervised pretext model too long can lead to an objective function mismatch \cite{metz2018meta,stuhr2019csnns} between the objectives used to train both models. This mismatch leads to a drop in performance on the target task, while the pretext model and the target models still converge correctly, which can be seen in \autoref{fig:ofm:ofm}. To quantify our results, we define soft and hard versions for two simple and general evaluation metrics - the \textit{metrics mismatch} and the \textit{objective function mismatch} - formally. With these metrics, we then evaluate different image-based pretext task types for self-supervised learning by using the linear evaluation protocol.
\\\\
Our contributions can be summarized as follows: 
\begin{itemize}
	\item We propose hard and soft versions of general metrics to measure and compare mismatches of (unsupervised) representation learning methods across different target tasks (Sections \ref{sec:metrics_mismatch} and \ref{sec:soft_metrics_mismatch}). To the best of our knowledge, this has not been done before. (\hyperref[RQ-E2]{RQ-E2})
	\item We discuss the usability and stability of our protocols on a variety of pretext and target tasks (Section \ref{sec:stability}). (\hyperref[RQ-E2]{RQ-E2}) 
	\item In our experiments, we qualitatively show dependencies of the objective function mismatch with respect to the pretext model's representation size (Section \ref{sec:repsize}), target model complexity (Section \ref{sec:tcomplex}), pretext and target augmentations (Section \ref{sec:aug}), as well as pretext and target task types (Section \ref{sec:targettask}). (\hyperref[RQ-E3]{RQ-E3}) 
	\item We find that the objective function mismatch can reduce performance on various benchmarks. Specifically, we observe a performance decrease of $\sim$0.1-5.0\% for Cifar10, Cifar100, and PCam, and up to $\sim$25-59\% in extreme cases for the 3dshapes dataset (Section \ref{sec:evaluation}). (\hyperref[RQ-E4]{RQ-E4}) 
\end{itemize}
\section{Hard Metrics Mismatch}
\label{sec:metrics_mismatch}
\begin{figure}[t]
	\begin{center}
		\includegraphics[width=0.8\linewidth]{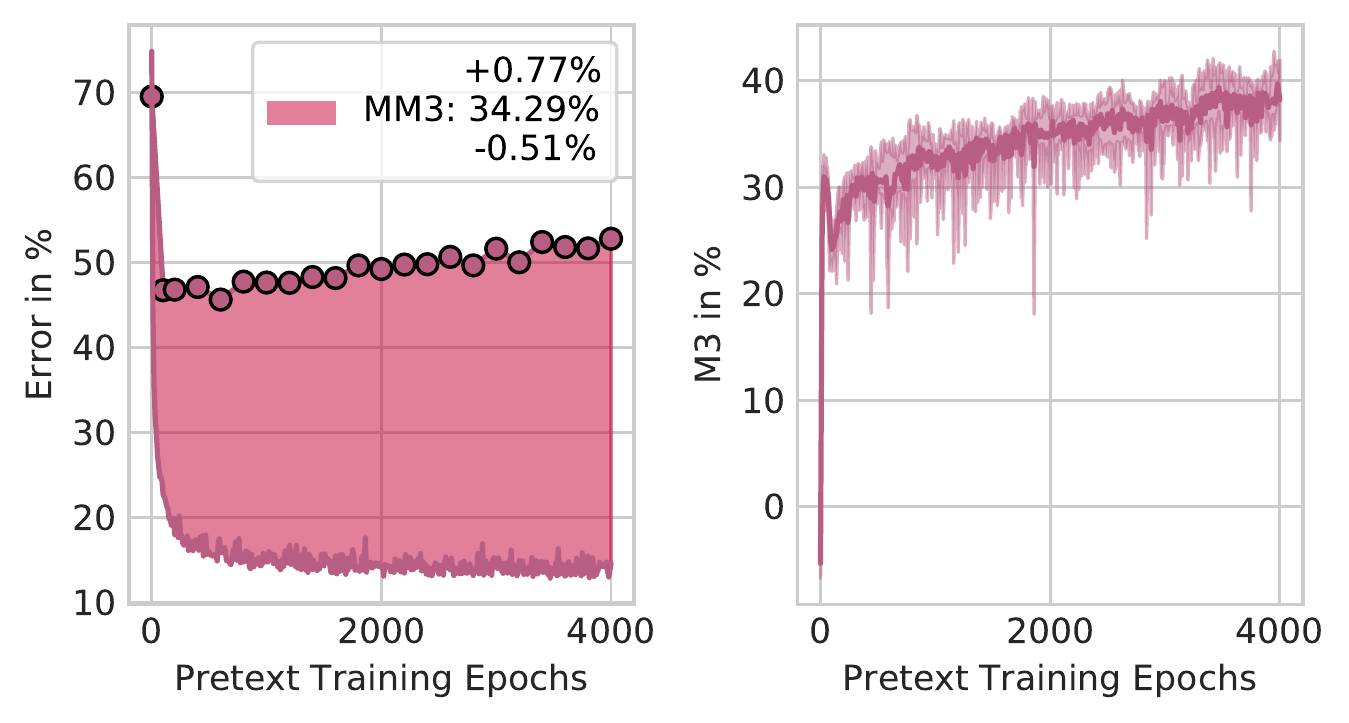}
	\end{center}
	\vspace{-2ex}
	\caption[$\mathrm{M3}$ and $\mathrm{MM3}$.]{$\mathrm{M3}$ and $\mathrm{MM3}$. (left) The intuition behind $\mathrm{MM3}$: In this case, both metrics measure a classification error in \%. The pretext metric (solid curve) is measured on the pretext task of predicting rotations with a ResNet18 model, and the target metric (dotted curve) on the fully trained Cifar10 classification task. When divided by the number of measurements, the discrete area enclosed by the target and pretext task curves corresponds to the $\mathrm{MM3}$ of the entire training process. (right) The Metrics Mismatch $\mathrm{M3}$ plotted during training: We observe a common behavior where the mismatch increases as training progresses. Additionally, we show the stability ($+,-$) of $\mathrm{M3}$, and $\mathrm{MM3}$ across a 5-fold cross-validation.}
	\label{fig:ofm:m3_and_mm3}
\end{figure}

With the objective function mismatch, we want to measure the mismatch of two objectives while training a model on a (unsupervised) pretext task and using its representations to train another model on a target task. In general we can measure the mismatch of two comparable metrics, if one metric is captured during training of a single pretext model and the other is captured for each target model fully trained on the representations of different steps or epochs of the pretext model. Two comparable metrics, for example, are classification accuracies for the pretext and target task because they use the same measurement unit and scale.
As illustrated in \autoref{fig:ofm:m3_and_mm3}, the metric values of the target models form a curve over the course of learning. Between a metric value on this curve and the corresponding metric value on the pretext model curve we can define the \textit{metrics mismatch} ($\mathrm{M3}$) for a certain step (or epoch) in training by calculating their distance.

More formally, let $M^{P}=(m_{1}^{P}, ... , m_{n}^{P})$ denote an $n$-tuple of values from a metric used to measure pretext model $P$ for different steps $S=(s_1, ...,  s_n)$. The length $n$ of the tuple is usually given by a convergence criterion $C$ on the metric of model $P$ during training. Furthermore, let $M^{T}=(m_{1}^{T}, ..., m_{n}^{T})$ denote an $n$-tuple of values from a comparable metric used to measure target model $T$. $M^{T}$ is of the same length and order as $M^{P}$ and all values are calculated at the same training steps $S$ of $M^{P}$. Thereby the target model $T$ is fully retrained for every step $s_i$ in $S$ on the representations of model $P$ at this step before we measure $m_{i}^{T}$. 

\begin{defn}
	\label{M3}
	The hard {\normalfont Metrics MisMatch} ($\mathrm{M3}$) between $m_{i}^{T}$ and $m_{i}^{P}$ at step $s_i$ is defined as:
	\begin{ceqn}	
		\begin{equation}
			\mathrm{M3}(m_{i}^{T}, m_{i}^{P}) \coloneqq m_{i}^{T}-m_{i}^{P}
			\label{eqn:M3}
		\end{equation}
	\end{ceqn}
	where $m_{i}^{T}$ and $m_{i}^{P}$ are single values measured with comparable metrics at step $s_i$.\footnote{Note that we define our measurements only for the case where lower metric values correspond to better performance. The definition for the opposite case arises naturally by changing maximum and minimum operations and/or subtraction orders.}  
\end{defn}
If $\mathrm{M3}>0$, the performance of the target model is lower then the performance of the pretext model at step $s_i$. In contrast, $\mathrm{M3}\leq0$ represents the desired case in unsupervised representation learning, where the target model performance is the same or above the pretext model performance at step $s_i$. In our case, we measure $m_{i}^{T}$ and $m_{i}^{P}$ over the entire evaluation dataset for every step $s_i$ in $S$. We plot $\mathrm{M3}(m_{i}^{T}, m_{i}^{P})$ for the pretext task of predicting rotations and the target task of Cifar10 classification during training in \autoref{fig:ofm:m3_and_mm3}. This shows that our metric captures the behavior of the target task performance regarding the pretext task performance, and we observe an increasing mismatch as training progresses. 
To capture the mismatch of the entire training procedure with respect to the target task in a single value, we can now define the \textit{mean hard metrics mismatch} ($\mathrm{MM3}$) as the mean bias error between $M^{T}$ and $M^{P}$.
\begin{defn}
	\label{MM3}
	The {\normalfont Mean hard Metrics MisMatch} ($\mathrm{MM3}$) between $M^{T}$ and $M^{P}$ is defined as:	
	\begin{ceqn}
		\begin{equation}
			\mathrm{MM3}({M}^{T}, M^{P}) \coloneqq 
			\frac{1}{n}\sum\limits_{0<i\leq n} (m_{i}^{T}-m_{i}^{P}) 
			\label{eqn:MM3}
		\end{equation}
	\end{ceqn}
	where $M^{T}$ and $M^{P}$ are tuples measured with comparable metrics until the pretext model converges at step $s_n$.
\end{defn}
$\mathrm{MM3}$ measures the bias of the target model metric to the pretext model metric. For positive or negative values of $\mathrm{MM3}$, we can make similar observations as for $\mathrm{M3}$, but they now account for the tendency of the entire training process and not for a single step $s_i$. In general, the mean bias error can convey useful information, but it should be interpreted cautiously because there are special cases where positive and negative values cancel each other out. In our case, this can happen, for example, when learning the pretext task is very useful for the target task early in training but hurts the target performance equally strong later on when the pretext task is sufficiently solved. We capture this behavior simply by measuring and plotting $\mathrm{M3}$ individually for the metric values of each step, as in \autoref{fig:ofm:m3_and_mm3}, analogous to the way a loss is measured and plotted during training. 

\subsection{Hard Objective Function Mismatch}
Naively we could compare the objective functions of the target and pretext task by using $\mathrm{M3}$, which we define as the \textit{hard objective function mismatch}. In most cases, however, the objective functions used to train the pretext model and the target models are not directly comparable. This is due to the usage of different objective functions for both model types, which, i.a., use different (non-)linearities. But for some pretext tasks simple comparable metrics can be defined. These metrics can be used as a proxy to measure the objective function mismatch in a general and comparable manner. A well-known example is the accuracy metric, which can be used on the self-supervised tasks of predicting rotations \cite{gidaris2018unsupervised} and the state-of-the-art approach of contrastive learning \cite{chen2020simple}. But comparable metric pairs can not always be found easily. For example, if we train a variational autoencoder and later use its representation for a classification target task, it does not make sense to define a pixel-wise error between the given and generated images as a comparable pretext task metric. To achieve a comparable measurement for this situation and on the loss curves in general, one could think of individual normalization techniques between objective function pairs. However, we want to be practical and define a measure that can be used independently of the objective function pairs for every pretext and target model combination. Furthermore, in practice we might be especially interested in how much the target task mismatches with the pretext task if a mismatch decreases target performance. This is why we define soft versions of our measurements.

\begin{figure}[t]
	\begin{center}
		\includegraphics[width=0.8\linewidth]{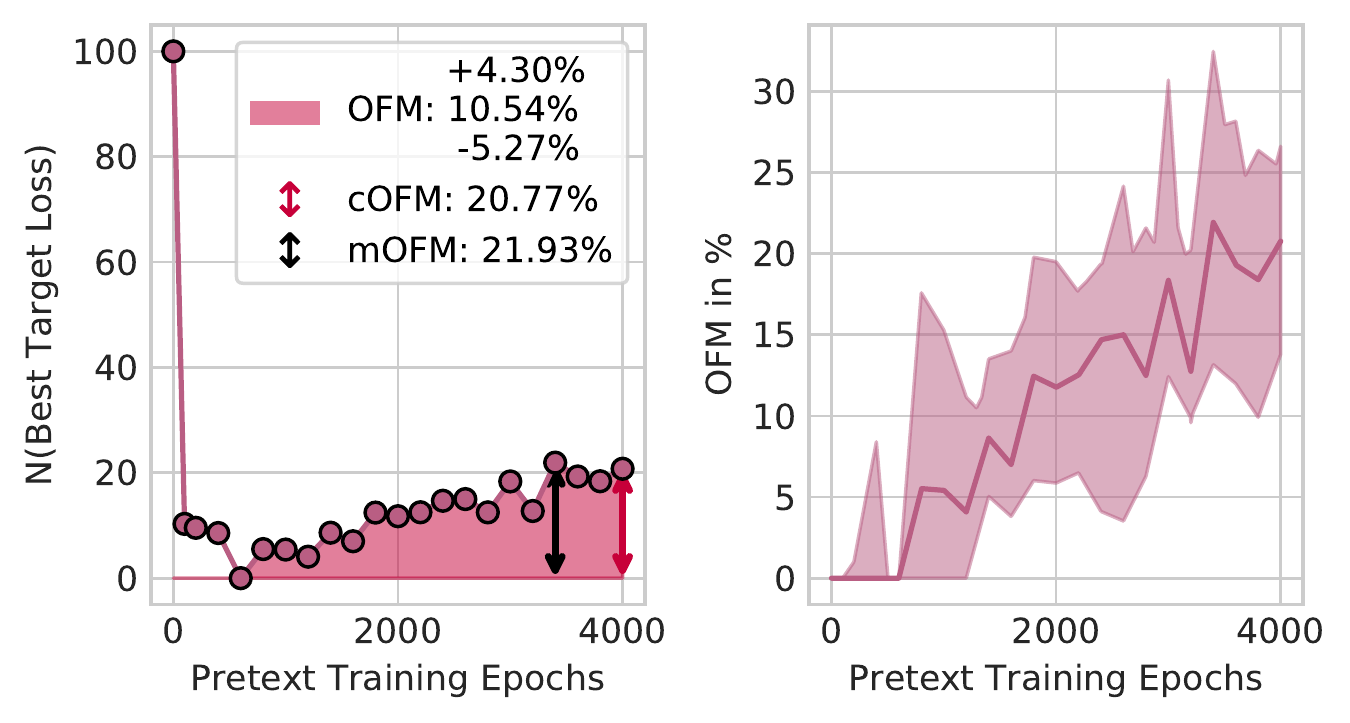}
	\end{center}
	\vspace{-2ex}
	\caption[$\mathrm{OFM}$, $\mathrm{cFM}$, $\mathrm{mOFM}$, and $\mathrm{MOFM}$.]{$\mathrm{OFM}$, $\mathrm{cFM}$, $\mathrm{mOFM}$, and $\mathrm{MOFM}$. We measure the $\mathrm{OFM}$ instead of $\mathrm{SM3}$ by normalizing the metric values with \autoref{eqn:norm}. For visualization, we additionally shift the normalized metric values such that they lie in $[0, 100]$ by subtracting the minimal measurement. (left) Intuition for the $\mathrm{MOFM}$: When divided by the number of measurements, the discrete area enclosed by the target metric values and their previous minimal target metric values correspond to the $\mathrm{OFM}$ of the entire training process. The red arrow shows the $\mathrm{cOFM}$, and the black arrow the $\mathrm{mOFM}$. In this case, the target metric measures the cross-entropy loss of each fully trained target model on a Cifar10 validation set. (right) The $\mathrm{OFM}$ plotted during training: We observe an increasing mismatch starting around epoch 600. Additionally, we show the range ($+,-$) of the $\mathrm{OFM}$ and $\mathrm{MOFM}$ across a 5-fold cross-validation.}
	\label{fig:ofm:sm3_and_msm3}
\end{figure}
\section{Soft Metrics Mismatch}
\label{sec:soft_metrics_mismatch}
To bypass objective function pair normalization we define the \textit{soft metrics mismatch} ($\mathrm{SM3}$) directly on the target metric. Thereby we no longer take the exact improvement of the pretext metric into account, we only care about its convergence. Since we now have no exact information about the pretext metric curve, we define $\mathrm{SM3}$ for the current step $s_i$ between the current target metric value and the previously or currently occurred minimal target metric value:
\begin{defn}
	\label{SM3}
	The {\normalfont Soft Metrics Mismatch} ($\mathrm{SM3}$) between $M^{T}$ and $M^{P}$ at step $s_i$ is defined as:
	\begin{ceqn}	
		\begin{equation}
			\mathrm{SM3}(m_{i}^{T}) \coloneqq m_{i}^{T} - \min_{0<j\leq i}(m_{j}^{T})
			\label{eqn:SM3}
		\end{equation}
	\end{ceqn}
	where $\min_{0<j\leq i}(m_{j}^{T})$ is the previously or currently occurred minimal target metric value.
\end{defn}

$\mathrm{SM3}$ has a slightly different meaning compared to $\mathrm{M3}$: It equals zero if $m_{i}^{T}$ is a minimal metric value and is positive if $m_{i}^{T}$ is higher than the previously occurred minimal metric value. We want to point out that the only way we incorporate the pretext metrics into this measurement is by making sure that the pretext model does not overfit and has not yet converged. Again, we measure $m_{i}^{T}$ and $m_{i}^{P}$ over the entire evaluation dataset for every step $s_i$ in $S$ and plot $\mathrm{SM3}(m_{i}^{T})$. A common case is shown in \autoref{fig:ofm:sm3_and_msm3}, which captures the behavior of pretext model training with respect to the target model. Here we observe zero soft mismatch early in training followed by increasing soft mismatch until pretext model convergence. Again, we can capture the mismatch of the pretext task with respect to the target task for the entire training process until pretext model convergence as the mean bias error of every metric value $m_{i}^{T}$ and its minimal metric value:
\begin{defn}
	\label{MSM3}
	The {\normalfont Mean Soft Metrics Mismatch} ($\mathrm{MSM3}$) between $M^{T}$ and $M^{P}$ is defined as:	
	\begin{ceqn}
		\begin{equation}
			\mathrm{MSM3}(M^{T}) \coloneqq \frac{1}{n}\sum\limits_{0<i\leq n} \left(m_{i}^{T} - \min_{0<j\leq i}(m_{j}^{T}) \right)
			\label{eqn:MSM3}
		\end{equation}
	\end{ceqn}
	when the pretext model convergences at step $s_n$.
\end{defn}
$\mathrm{MSM3}$ can either be zero if no mismatch occurs or positive if there is a mismatch. Therefore, using $\mathrm{MSM3}$ brings the benefit that positive values can not be canceled out by negative values. 
Furthermore, we define the maximum occurring mismatch $\mathrm{mSM3}$ and the mismatch at pretext model convergence $\mathrm{cSM3}$. We are especially interested in $\mathrm{cSM3}$ since it measures the representations one would naively take for the target task:
\begin{ceqn}
	\begin{gather}
		\mathrm{cSM3}(M^{T}) \coloneqq \mathrm{SM3}(m_{n}^{T})
		\label{eqn:CSM3}\\
		\mathrm{mSM3}(M^{T}) \coloneqq \max_{0<i\leq n}\left(\mathrm{SM3}(m_{i}^{T})\right)
		\label{eqn:mSM3}
	\end{gather}
\end{ceqn}
\subsection{Soft Objective Function Mismatch}
Now we can use $\mathrm{SM3}$ to measure a soft form of the objective function mismatch on the loss curve obtained by the target models. However, the values of these measurements lie in a range that depends on the target objective function. Therefore, they are not directly comparable to the measurements on loss curves from other target tasks. This is why we normalize the measurements of the target metric to the percentage range and define the \textit{objective functions mismatch} ($\mathrm{OFM}$) as follows:
\begin{defn}
	\label{OFM}
	The {\normalfont Soft Objective Function Mismatch } ($\mathrm{OFM}$) between $M^{T}$ and $M^{P}$ at step $s_i$ is defined as:
	\begin{ceqn}	
		\begin{gather}
			\mathrm{OFM}(m^{T}_i) \coloneqq \mathrm{SM3}\left(\mathrm{N}(m^{T}_i)\right)
			\label{eqn:OFM}\\
			\mathrm{N}(x) \coloneqq \left\{\begin{array}{ll}
				\frac{100\times x}{m_{1}^{T}-m_{b}^{T}} & \quad m_{1}^{T} > x \geq m_{b}^{T}\\
				0 & \quad m_{1}^{T} = m_{b}^{T} = x\\
				\infty\,(special\,case)\, & \quad m_{1}^{T} = m_{b}^{T} < x \\
			\end{array}
			\right.
			\label{eqn:norm}
		\end{gather}
	\end{ceqn}
	where $m_{1}^{T}$ is the loss value of the target model trained on an untrained pretext model ($s_1=0$) and $b=\mathrm{argmin}_{0<i\leq n}(m^{T}_i)$ denotes the index of the minimal target loss value. We then use $\hat{M}^{T} = (\mathrm{N}(m^{T}_1),...,\mathrm{N}(m^{T}_n))$ to calculate the $\mathrm{OFM}$.
\end{defn} 
The intuition behind this normalization is that we declare $m_{b}^{T}$ as the value where the pretext model has learned all of the target objective it was able to learn (with this setting) and $m_{1}^{T}$ as the value where the model has learned nothing of the target objective. Now we measure with $\mathrm{OFM}(m^{T}_i)$ for what percentage the learning of a pretext objective hurts the maximum achieved target performance at step $s_i$. An example is illustrated in \autoref{fig:ofm:sm3_and_msm3}. Furthermore, we can normalize the other soft measurements from \autoref{eqn:CSM3} and \ref{eqn:mSM3} analog to \autoref{eqn:OFM}.\\
\\
The $\mathrm{OFM}$ is a general measure that can be used for pretext and target models where no good proxy metrics can be defined. With the $\mathrm{OFM}$, we are able to compare mismatches across different pretext and target task objectives and their combinations. We propose these measurements to obtain quantitative and therefore comparable results for individual pretext tasks. To get the best information about the training process we encourage to plot the curves formed by our metrics as well. We want to point out that our metrics are not intended to measure target task performance, they measure how much the performance on a target task can decrease when an (ill-posed) pretext task is learned too long. Now, to understand the $\mathrm{OFM}$ further, we take a look at some cases:
\\\\
$\mathrm{OFM}(m^{T}_i)=0$: In this case, solving the pretext task objective has not hurt the performance of the target task objective at this point in training.
\\\\
$\mathrm{OFM}(m^{T}_i)=x$: Solving the pretext objective has hurt the performance of the target objective at this point in training by $x\%$ of what the model has learned. Therefore, we should have stopped training earlier. It is not guaranteed that longer training would hurt performance even more, but a growing $\mathrm{OFM}$ curve or $\mathrm{MOFM}$ is a good indicator of that.
\\\\
$\mathrm{OFM}(m^{T}_i)>100$: The target objective performance is worse than for the untrained model at this point in training.
\\\\
$\mathrm{MOFM}(M^{T})=\infty$: Solving the pretext objective hurts the performance of the target objective from the point of initialization. Because we have learned essentially 0\% about the target objective in the training process, there is no interval to be used for normalization. Therefore, we interpret this case as if the model has an infinite mismatch as soon as the model forgets something about the target objective.

\section{Experimental Setup}
In our experiments, we focus on image-based self-supervised learning. However, it is likely that other target domains show mismatches as well, e.g., \cite{doersch2017multi}. 
\\\\
\textbf{Pretext Tasks.} For generation-based self-supervision, we evaluate the approaches of autoencoding data from autoencoders (CAE) and color restoration (CCAE) as suggested in \cite{chen2020simple}. To evaluate context structure generation, we use denoising autoencoders (DCAE). Transformation prediction (spatial context structure) is evaluated via autoencoding tranformations by predicting rotations \cite{gidaris2018unsupervised} (RCAE). For contrastive methods (context-based similarity), we evaluate SimCLR \cite{chen2020simple} (SCLCAE). We refer to the literature for first glances into mismatches for VAEs \cite{metz2018meta}, and meta-learning \cite{metz2018meta} and to \autoref{chap:00} for self-organization \cite{stuhr2019csnns}. 
\\\\
\textbf{Pretext Models.} Unless stated otherwise, we use a four-layer CNN as the encoder. For the autoencoding data approaches, we use a four-layer decoder with transposed convolutions, for rotation prediction a single dense layer, and for contrastive learning a nonlinear head, as suggested in \cite{chen2020simple}. We show that mismatches account for other architectures as well by carrying out additional evaluations using ResNets \cite{he2016deep} in \autoref{tab:ofm:quantitativ_results_resnets} and \autoref{app:01}.
\\\\
\textbf{Target Tasks.} We evaluate our metrics on image-based target tasks. For coarse-grained classification, we use Cifar10, Cifar100 \cite{krizhevsky2009learning}, and the coarse-grained labels of 3dshapes \cite{3dshapes18}. For fine-grained classification, we use the PCam dataset \cite{veeling2018rotation} and the fine-grained labels of 3dshapes. 
\\\\
\textbf{Target Models.}: Following the linear evaluation protocol, we use a single, linear dense layer (FC) as the target model with a softmax activation. To evaluate our metrics for other target models, we use a two-layer MLP (2FC) and a three-layer MLP (3FC). 
\\\\
\textbf{Augmentations.} We make sure not to compare augmentations instead of pretext tasks by following \cite{chen2020simple} for our base augmentations, to which we add the pretext task-specific augmentations for pretext task training and evaluation. For the target task, we use the base training and evaluation augmentations of \cite{chen2020simple}.
\\\\
\textbf{Optimization.} Our models are trained using the Adam optimizer \cite{kingma2014adam} with standard parameters and a batch size of $2048$ without any regularization instead of batch normalization. For our ResNets, we additionally use a weight decay of $1\mathrm{e}{-4}$. 
\\\\
\textbf{Mismatch Evaluation.} All reported values are determined by 5-fold cross-validation. We use standard early stopping (from tf.keras) as a convergence criterion on the pretext evaluation curve with a minimum delta (threshold) of $0$ and patience of $3$. We change the patience in some experiments of Tables \ref{tab:ofm:ablation_rep_tm_augs} and \ref{tab:ofm:quantitativ_results_resnets} to get a reasonable convergence epoch. For more details we refer to \autoref{app:01}. When calculating our metrics, we estimate target values of missing epochs with linear interpolation to save computation time. In our case, $\mathrm{SM3}$ and $\mathrm{MM3}$ are measured on the target task accuracy.
\\\\
\textbf{Implementation.} Our implementation is available at \\ \url{https://github.com/BonifazStuhr/OFM}. 

\section{Evaluation}
\label{sec:evaluation}
In the following, we show the results of most pretext and target tasks we have evaluated. We refer to \autoref{app:01} for additional, more detailed evidence. Since we capture our metrics during training, all mismatches are measured on the evaluation dataset.

\subsection{Mismatch and Convergence}
\label{sec:mandc}
For our measurements we make sure to use metric value pairs from models that do not overfit. We achieve this by applying a convergence criterion on the pretext task and by using the best metric values from each target model evaluation curve. As shown in \autoref{app:01}, most observations in our experiments are independent of the use of a convergence criterion if pretext models are trained long enough and without overfitting. Furthermore, we observe a common behavior in \autoref{fig:ofm:ofm}: Target models trained on higher epochs of the pretext model tend to converge faster. \textit{This indicates that longer training of the pretext task tends to create easier separable representations, which may mismatch with the class label}.
\begin{figure}[t]
	\begin{center}
		\includegraphics[width=1.0\linewidth]{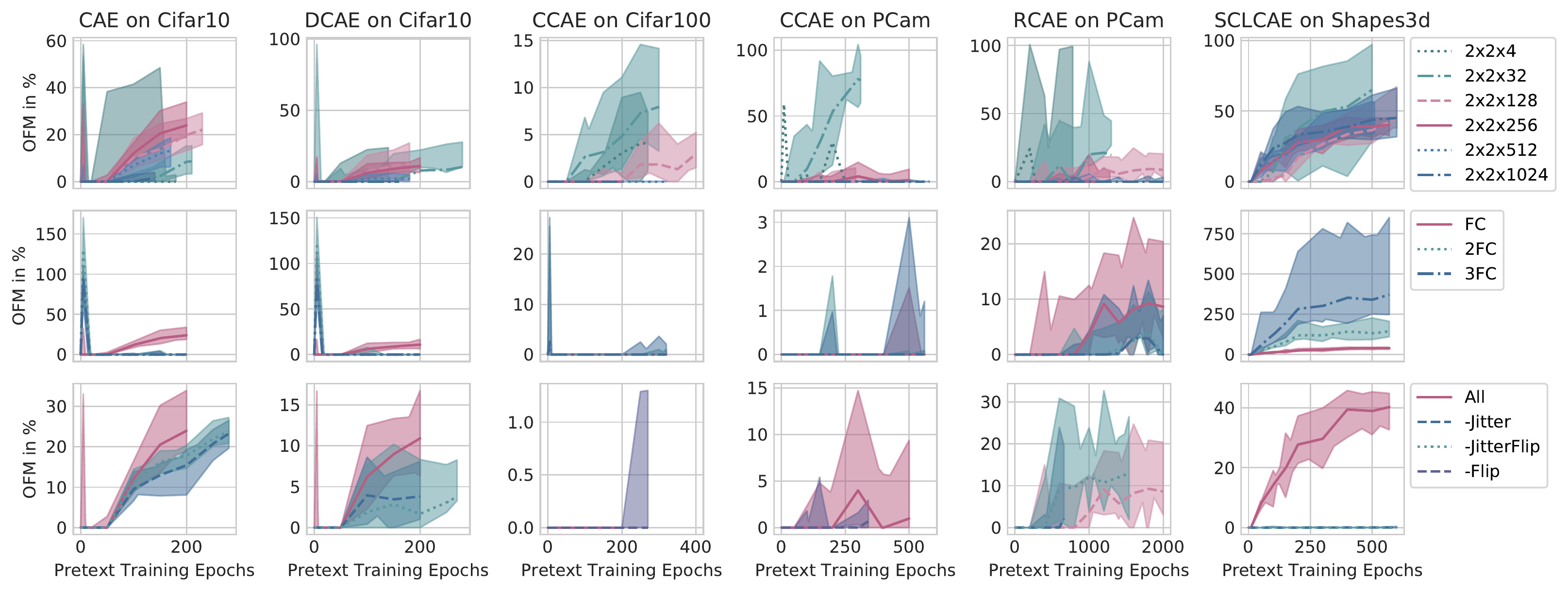}
	\end{center}
	\vspace{-2ex}
	\caption[$\mathrm{OFM}$ ablation study.]{$\mathrm{OFM}$ ablation study. (top) Impact of different pretext model representation sizes on the $\mathrm{OFM}$. (middle) $\mathrm{OFM}$ for the linear target model and nonlinear target models trained on our pretext model. (bottom) $\mathrm{OFM}$ for the linear target model and the pretext models trained on fewer augmentations. First, we removed the color jitter and then the vertical flip from the augmentations. The target models of SCLCAE were trained on 3dshapes to predict the object hue. Best viewed in color.}
	\label{fig:ofm:ofm_ablation_study}
\end{figure}
\begin{table}[t]
	\RawFloats
	\scriptsize
	\caption[$\mathrm{MOFM}$, $\mathrm{cSM3}$ and $\mathrm{MM3}$ of the models from \autoref{fig:ofm:ofm_ablation_study}.]{$\mathrm{MOFM}$, $\mathrm{cSM3}$ and $\mathrm{MM3}$ of the models from \autoref{fig:ofm:ofm_ablation_study}. $\mathrm{MM3}$ is measured on the target and pretext task classification error. $\mathrm{cSM3}$ is measured on the target task classification error and corresponds to the accuracy we would lose when we naively train the target model after pretext model convergence. Values are obtained by 5-fold cross-validation. We show the stability of each measurement in Tables \ref{tab:ofm:app:ablation_rep_tm_augs_details_1}, \ref{tab:ofm:app:ablation_rep_tm_augs_details_2}, and \ref{tab:ofm:app:ablation_rep_tm_augs_details_3} of \autoref{app:01}.}
	\vspace{-1ex} 
	\label{tab:ofm:ablation_rep_tm_augs}
	\begin{center}
		\scalebox{0.875}{%
			\setlength{\tabcolsep}{0.4em}
			\begin{tabular}{lcccccccccccccc}
				\toprule
				&\multicolumn{2}{c}{\acs{CAE} (Cifar10)} &\multicolumn{2}{c}{\acs{DCAE} (Cifar10)} 	&\multicolumn{2}{c}{\acs{CCAE} (Cifar100)} 	&\multicolumn{2}{c}{\acs{CCAE} (PCam)}  &\multicolumn{3}{c}{\acs{RCAE} (PCam)}&\multicolumn{3}{c}{\acs{SCLCAE} (3dshapes)} \\	
				\cmidrule(lr){2-3}
				\cmidrule(lr){4-5}
				\cmidrule(lr){6-7}
				\cmidrule(lr){8-9}
				\cmidrule(lr){10-12}
				\cmidrule(lr){13-15}
				& $\mathrm{cSM3}$ & $\mathrm{MOFM}$ &   $\mathrm{cSM3}$ & $\mathrm{MOFM}$ &  $\mathrm{cSM3}$ & $\mathrm{MOFM}$ &  $\mathrm{cSM3}$ &  $\mathrm{MOFM}$ & $\mathrm{cSM3}$ & $\mathrm{MOFM}$  & $\mathrm{MM3}$ &   $\mathrm{cSM3}$ & $\mathrm{MOFM}$ & $\mathrm{MM3}$\\
				\midrule
				\textbf{\textit{Rep. Size}}: &&&&&&&&&&&&& \\ 
				2x2x4   	&\textbf{0.00}	&\textbf{0.00}  
				&	0.05	&\textbf{0.00}  
				&0.28 &1.54   	
				&4.98  &9.28
				&5.38 &4.15  &-22.26
				&26.34 &$\infty$ &-7.99\\
				2x2x32   &	0.07	&1.99  
				&	 \textbf{0.00}     &3.20 	
				&0.65& 3.64
				&5.17 	&34.30 
				&3.34 &7.92 & -21.09  
				&12.98 &36.39 &-57.67\\
				2x2x128 &  0.20		&10.10  
				&   0.06  &5.51   	
				&0.51 & 0.81
				&0.32 	&0.10  
				&1.03 &4.04  &-23.47
				&8.14 &\textbf{22.65} &\textbf{-66.19}\\
				2x2x256   &	0.75	&11.14  
				& 0.69   &5.17   	
				&0.17& \textbf{0.00}	
				&0.43 	&0.87 
				&0.44 &\textbf{0.00}  &-27.60  
				&6.52 &27.65 &-65.77\\
				2x2x512   &	0.43	&5.28  
				&	 0.36    &1.25 	
				&\textbf{0.00} &\textbf{0.00}	
				&0.20 	&0.07 
				&0.18 &\textbf{0.00}  &\textbf{-28.03}  
				&5.96 &27.78 &-63.61\\
				2x2x1024  & 	\textbf{0.24}	&\textbf{0.25}
				& \textbf{0.03}    &\textbf{0.00} 	
				& \textbf{0.00}	&\textbf{0.00}
				&\textbf{0.00}	&\textbf{0.00}  
				&\textbf{0.09} &\textbf{0.00}  &-26.56
				&\textbf{4.76} &32.70 &-57.92 \\
				\midrule
				\textbf{\textit{Target Model}}:&&&&&&&&&&&&&  \\
				FC   	&	0.75	&11.14    
				&  0.69 &5.17  	
				& \textbf{0.00} & \textbf{0.00} 	
				&\textbf{0.00}	&\textbf{0.00}  
				&1.03 &4.04  &-23.47 
				&6.52 &\textbf{27.65} &-65.77\\
				2FC   &	\textbf{0.03}		&5.68  
				&	\textbf{0.00}     &5.14  	
				& 0.08 &\textbf{0.00}  
				&\textbf{0.00}	&\textbf{0.00} 
				&\textbf{0.31} &0.76  &-28.56  
				&1.84 &103.30 &-70.49\\
				3FC   	&	\textbf{0.03}	&\textbf{3.94}  
				&  \textbf{0.00}   &\textbf{3.17}	
				& 0.12 & 0.02	
				&0.08 	&\textbf{0.00} 
				&0.37 &\textbf{0.61}  &\textbf{-29.61}   
				&\textbf{0.91}& 258.18 &\textbf{-71.26} \\
				\midrule
				\textbf{\textit{Augmentations}}: &&&&&&&&&&&&& \\
				All   &	\textbf{0.75}		&11.14 
				& 0.69   &5.17   
				& \textbf{0.17}	 & \textbf{0.00}	
				&0.43 	&0.87  
				&1.03 &4.04  &-23.47   
				&6.52 &27.65 &\textbf{-65.77}\\
				NoJitter  &  	0.99	&\textbf{10.99} 
				& \textbf{0.50}  &2.33 	
				&  - 	&  -
				&-	& -
				&\textbf{0.58} &\textbf{0.10}  &\textbf{-28.60}   
				&\textbf{0.00} &0.01 &-37.92 \\
				NoJitterNoFlip & 1.00 &12.51   
				&0.55  &\textbf{1.73} 	
				&-		&  -
				&-	&- 	
				&1.51 &7.33 &-10.60
				&\textbf{0.00} &\textbf{0.00}  &-35.95  \\
				NoFlip   		&-		& -     
				& -    &-  	
				&0.20	 & \textbf{0.00}		  
				&\textbf{0.41}	&\textbf{0.04}  
				&- & - & - 
				&- &- & -\\
				\bottomrule
		\end{tabular}}
	\end{center}
	\vspace{-8ex}
\end{table}

\subsection{Stability}
\label{sec:stability}
To evaluate the stability of our measurements, we show the mismatches of the entire training process and their range ($+,-$) using 5-fold cross-validation in Tables \ref{fig:ofm:m3_and_mm3} and \ref{fig:ofm:sm3_and_msm3}. The range of all other models we have trained is shown in \autoref{app:01}. We observe that $\mathrm{M3}$ generally seems more stable than $\mathrm{SM3}$ or the $\mathrm{OFM}$ since it does not rely so heavily on the target metric values, which can be quite unstable. The instability of the target task mismatch is captured in $\mathrm{M3}$ but does not matter much in the overall measurement for most cases. This is favorable if a stable value is desired and unfavorable if one wants to capture the instability of the target task training process explicitly. Furthermore, $\mathrm{M3}$ is able to compensate target fluctuations with pretext fluctuations. In general, we observe that as long as we calculate the $\mathrm{OFM}$ across a fair amount of cross-validations (in our case $5$), we can make statements about the mismatch. We measure our metrics on the mean losses during 5-fold cross-validation instead of calculating them five times and taking the average. For $\mathrm{M3}$, both variants are equivalent, and for the $\mathrm{OFM}$ measuring on the mean losses leads to a lower bound in the case where all models converge at step $s_n$ (see \autoref{app:01} for the simple proofs). We prefer to measure our metrics on the mean losses since this avoids mismatches occurring only in some validation cycles due to small fluctuations in the underlying training procedure. An example is shown in translucent red in \autoref{fig:ofm:sm3_and_msm3} at the beginning of training. We want to point out that the training and validation data differ slightly in every training round because of the cross-validation setup. This increases the instability but shows the general behavior of the metrics for the underlying data distribution. In Figures \ref{fig:ofm:app:cae_xfold_full_stability} and \ref{fig:ofm:app:cae_xfold_partial_stability} of \autoref{app:01}, we compare the instability of partially measured mismatches using linear interpolation with mismatches measured for every pretext training epoch and observe a similar instability. However, when using the $\mathrm{OFM}$ in practice to compare models on a finer scale, we recommend searching for the actual minimal target metric value since the $\mathrm{OFM}$ relies on this value at each step. When tuning a model for maximum performance, one searches for this value. Thereby looking at the $\mathrm{OFM}$ curve gives good indications of which interval one should search. This makes this protocol useful for performance tuning if enough computational power is available.
\begin{figure}[t]
	\begin{center}
		\includegraphics[width=1.0\linewidth]{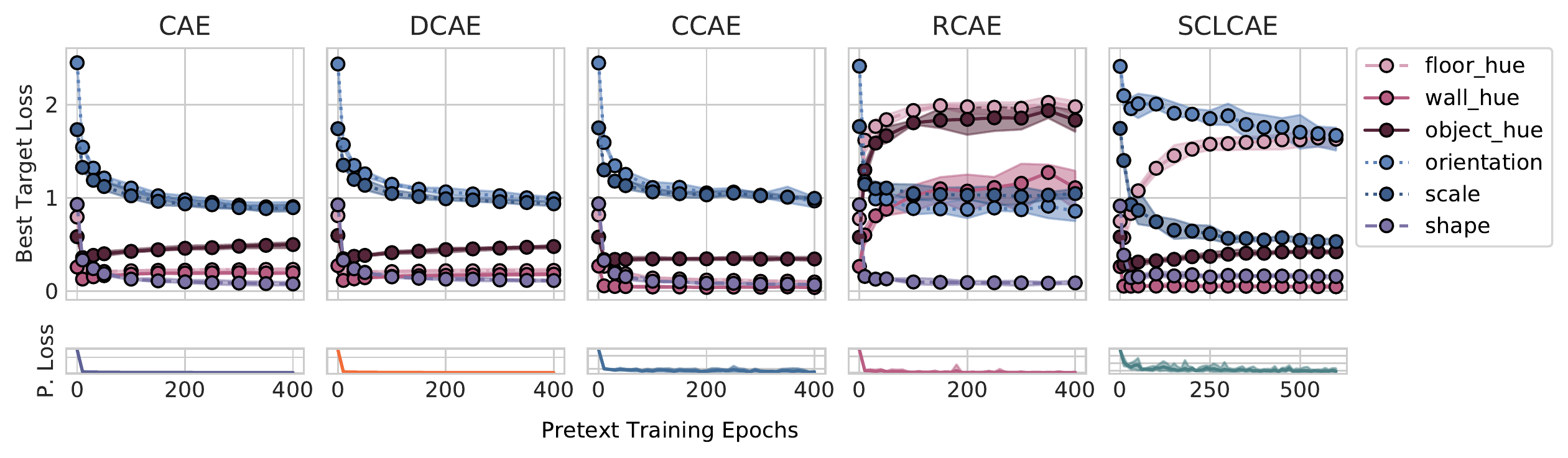}
	\end{center}
	\vspace{-2ex}
	\caption[Pretext and target losses on the 3dshapes dataset.]{Pretext and target losses on the 3dshapes dataset. (bottom) Pretext losses of our model trained for color restoration (CCAE), prediction rotations (RCAE), and contrastive learning (SCLCAE). (top) Best target losses of linear models trained for the different prediction tasks of 3dshapes. Best viewed in color.}
	\label{fig:ofm:pretext_target_losses_3dshapes_big}
\end{figure}
\begin{table}[t]
	\RawFloats
	\scriptsize
	\caption[$\mathrm{cSM3}$, $\mathrm{MOFM}$ and $\mathrm{MM3}$ on the 3dshapes dataset.]{$\mathrm{cSM3}$, $\mathrm{MOFM}$ and $\mathrm{MM3}$ on the 3dshapes dataset. $\mathrm{SM3}$ and $\mathrm{MM3}$ are measured on the target task accuracy. Values are obtained by 5-fold cross-validation. We show the stability of each measurement in Tables  \ref{tab:ofm:app:quantitativ_results_3dshapes_details_1}, \ref{tab:ofm:app:quantitativ_results_3dshapes_details_2}, and \ref{tab:ofm:app:quantitativ_results_3dshapes_details_3} of \autoref{app:01}.}
	\vspace{-1ex} 
	\label{tab:ofm:quantitativ_results_3dshapes}
	\begin{center}
		\scalebox{1.08}{%
			\setlength{\tabcolsep}{0.4em}
			\begin{tabular}{lcccccccccccc}
				\toprule
				&\multicolumn{2}{c}{\acs{CAE}} &\multicolumn{2}{c}{\acs{DCAE}} &\multicolumn{2}{c}{\acs{CCAE}}   &\multicolumn{3}{c}{\acs{RCAE}} 	&\multicolumn{3}{c}{\acs{SCLCAE}} 
				\\
				\cmidrule(lr){2-3}
				\cmidrule(lr){4-5}
				\cmidrule(lr){6-7}
				\cmidrule(lr){8-10}
				\cmidrule(lr){11-13}	
				& $\mathrm{cSM3}$ & $\mathrm{MOFM}$
				& $\mathrm{cSM3}$ & $\mathrm{MOFM}$
				&  $\mathrm{cSM3}$ & $\mathrm{MOFM}$
				&  $\mathrm{cSM3}$ & $\mathrm{MOFM}$& $\mathrm{MM3}$
				& $\mathrm{cSM3}$ & $\mathrm{MOFM}$&$\mathrm{MM3}$\\
				\midrule
				floor\_hue 	&0.01&0.95
				&\textbf{0.00}&1.28
				&\textbf{0.02}&\textbf{0.00}   
				&56.68&$\infty$&44.67
				&28.18&268.27&-48.38\\  
				wall\_hue 
				&0.02&32.03
				&\textbf{0.00}&24.43
				&0.10&\textbf{0.00}
				&25.17&$\infty$&7.80
				&\textbf{0.29}&0.46&\textbf{-76.40}\\
				object\_hue 	
				&0.38&22.71   
				&0.43&24.55
				&1.55&0.63
				&59.65&$\infty$&40.1
				&2.87&8.69&-73.17\\
				scale 
				&0.41&\textbf{0.00}   
				&0.27&\textbf{0.00}
				&0.10&\textbf{0.00}    
				&2.60&0.13&31.78
				&2.43&\textbf{0.00}&-44.80\\ 
				shape 
				&0.07&\textbf{0.00}  
				&0.08&\textbf{0.00}
				&0.03&\textbf{0.00}  
				&\textbf{0.20}&0.06&\textbf{-2.48}
				&1.67&2.16&-67.54\\	
				orientation 
				&\textbf{0.00}&\textbf{0.00}
				&\textbf{0.00}&\textbf{0.00}
				&0.23&\textbf{0.00}
				&0.48&\textbf{0.00}&22.26
				&2.50&6.68&-9.11\\ 
				\midrule
				average
				&0.15&9.28
				&\textbf{0.13}&8.21
				&0.34&\textbf{0.11}
				&24.13&$\infty$&24.02
				&6.32&47.71&\textbf{-53.23}\\ 
				\bottomrule
		\end{tabular}}
	\end{center}
	\vspace{-2ex}
\end{table}

\subsection{Dependence on Representation Size}
\label{sec:repsize}
We hypothesize that large representation sizes tend to lower the $\mathrm{OFM}$, which could be one reason why representation sizes are large in unsupervised learning. To affirm this hypothesis empirically, we train our pretext models with varying representation sizes on different target tasks while fixing all other model parameters. Figures \ref{fig:ofm:ofm_ablation_study} and \ref{tab:ofm:ablation_rep_tm_augs} show that the $\mathrm{OFM}$ tends to decrease when we enlarge the representation size. A reason for that might be that target models can exploit the high dimensional space of large representations to find better-fitting clusters for their target task. We found an exception to this behavior, where we use larger representations of SCLCAE for the easy task of object hue prediction. Here, the target models trained on the untrained pretext models with larger representation sizes already achieve high performance due to a larger number of color-selective random features. Further learning of the pretext model does not lead to a high performance gain in this case, and forgetting these sensitive random features during training leads to a high mismatch. Additionally, we observe that mismatches decrease when we decrease the representation size for generation-based methods. A reason could be that the pretext models are forced to generalize to solve the target task for small representation sizes due to the limited amount of features in the bottleneck or simply underfit on the pretext task.

\subsection{Dependence on Target Model Complexity}
\label{sec:tcomplex}
In Figures \ref{fig:ofm:ofm_ablation_study} and \ref{tab:ofm:ablation_rep_tm_augs}, we observe an $\mathrm{OFM}$ spike early in training for more complex target models. This spike occurs probably because nonlinear target models make better sense of specific random features at pretext task initialization, in contrast to the linear target model. \textit{Besides early spikes, mismatches tend to decrease when we add complexity to the target model}. A model with increased nonlinearity has more freedom to disentangle representations that do not fit properly with the target task. Again we found an exception where the $\mathrm{MOFM}$ is lower for linear models when predicting the object hue after contrastive learning, which can be appointed to the color-selective, random features of the untrained pretext model. 

\subsection{Dependence on Augmentations}
\label{sec:aug}
We vary the augmentations used for the pretext and target model by removing the color jitter and the image flip from our base augmentations successively. \autoref{fig:ofm:ofm_ablation_study} shows that \textit{augmentations can have a positive or negative impact on the mismatch.} E.g., when predicting the object's hue, the ill-posed color jitter augmentation increases the mismatch significantly. 

\subsection{Dependence on Target Task Type}
\label{sec:targettask}
Here we use our metrics to examine findings stated in \cite{doersch2017multi}, \cite{kolesnikov2019revisiting}, \cite{zhai2019large}, and \cite{wallace2020extending}, where it is argued that some pretext tasks are better suited for different target tasks. We fix the underlying data distribution by using the 3dshapes dataset and train our target models for the different tasks. These tasks require a generic understanding of the scene like coarse-grained knowledge about the object's type and hue and fine-grained knowledge about shapes, positions, and scales. In Figures \ref{fig:ofm:pretext_target_losses_3dshapes_big} and \ref{tab:ofm:quantitativ_results_3dshapes}, we observe that \textit{pretext models tend to learn pretext task-specific features and discard features that are not needed to solve the pretext task during training. Therefore, these models mismatch with ill-posed target tasks.} For example, rotation prediction discards features corresponding to the hue while it learns much about the orientation of the object.

\begin{table}[h]
	\RawFloats
	\caption[Mismatches of ResNets with convergence criterium.]{Mismatches of ResNets with convergence criterium. $\mathrm{ACC}$ stands for the best accuracy on the target task of all target models trained on the pretext model. $\mathrm{cSM3}$ corresponds to the accuracy we would lose when we naively train the target model after pretext model convergence. Values are obtained by 5-fold cross-validation. The mismatches are measured with a convergence criterium.} 
	\vspace{-1ex}
	\label{tab:ofm:quantitativ_results_resnets}
	\begin{center}
		\scalebox{0.8}{%
			\setlength{\tabcolsep}{0.4em}
			\begin{tabular}{lcccc}
				\toprule
				&\multicolumn{4}{c}{ResNet}\\
				\cmidrule(lr){2-5}
				& $\mathrm{ACC}$& $\mathrm{cSM3}$& $\mathrm{MOFM}$ & $\mathrm{MM3}$ \\
				\midrule
				\acs{RCAE} (Cifar10) &$54.64^{+1.80}_{-2.01}$&$3.98^{+1.74}_{-3.60}$&$4.87^{+4.42}_{-3.11}$&$31.82^{+0.75}_{-0.68}$ \\
				\acs{SCLCAE} (PCam) &$96.25^{+0.44}_{-0.23}$&$0.37^{+0.44}_{-0.37}$&$0.86^{+1.00}_{-0.60}$&$-53.26^{+0.52}_{-0.38}$ \\ 
				\bottomrule
		\end{tabular}}
	\end{center}
	\vspace{-2ex}
\end{table}

\begin{figure}[t]
	\begin{center}
		\begin{tabular}{cc}
			\includegraphics[width=0.45\columnwidth]{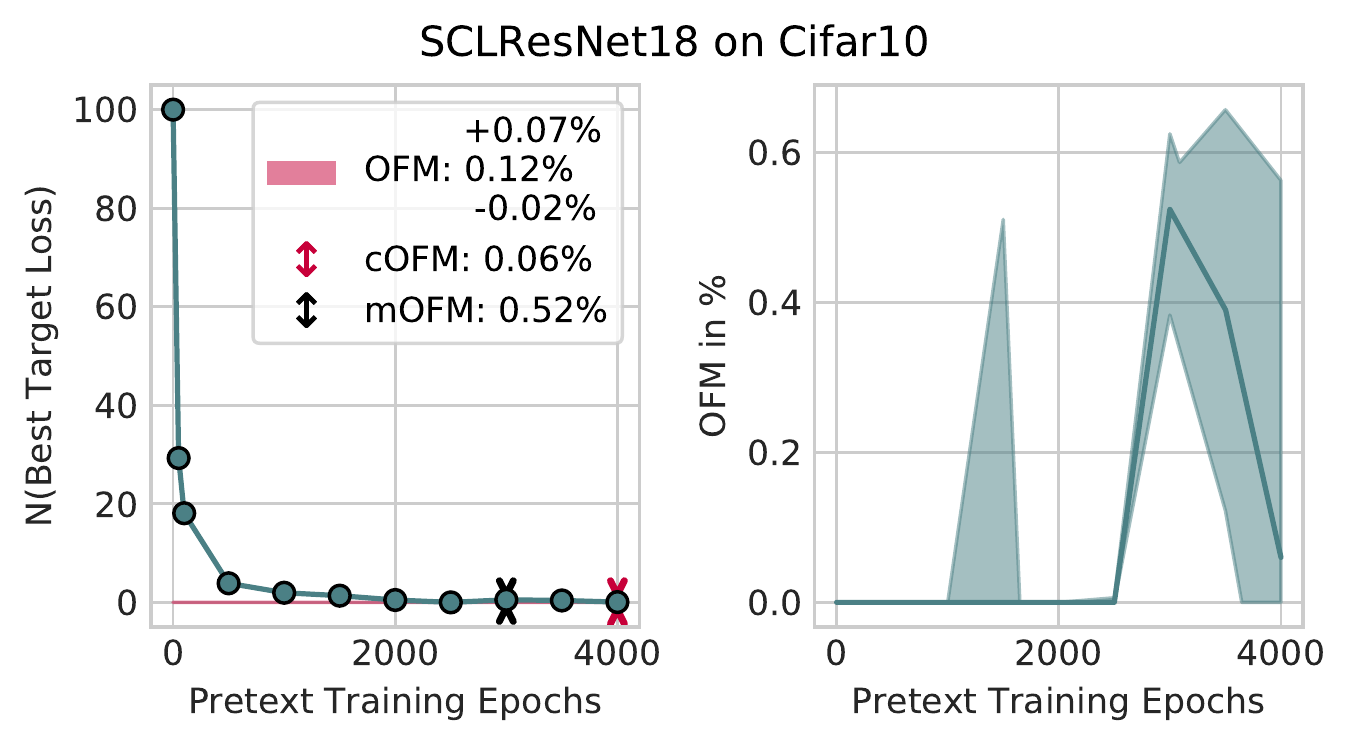} &    
			\includegraphics[width=0.45\columnwidth]{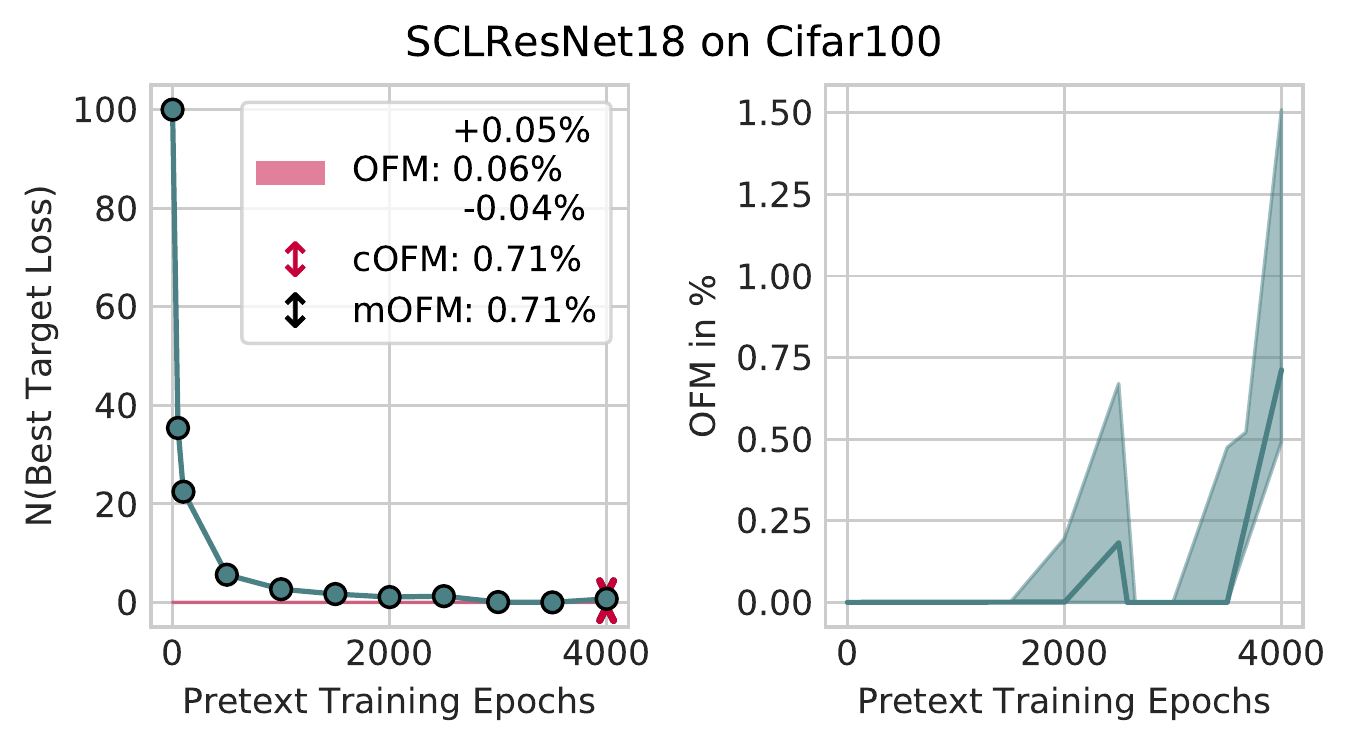} \\    
			\includegraphics[width=0.45\columnwidth]{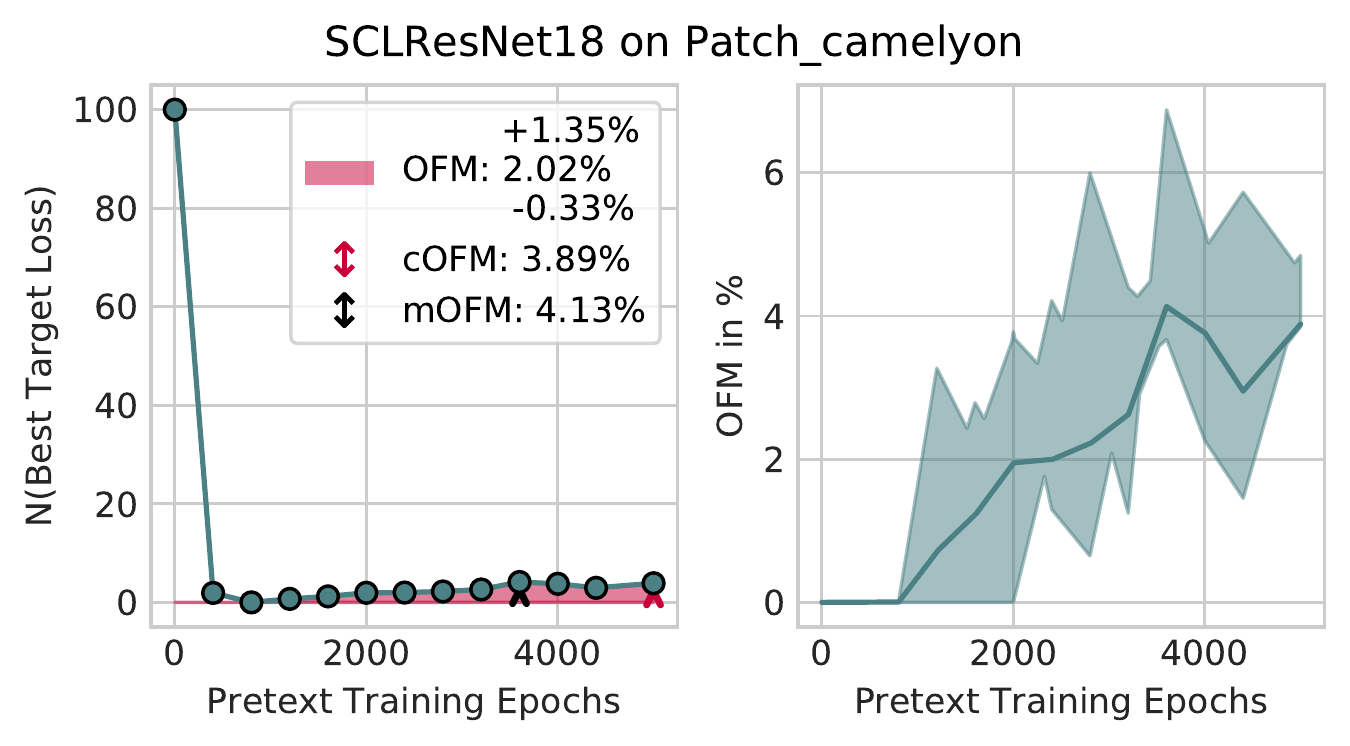} &    
			\includegraphics[width=0.45\columnwidth]{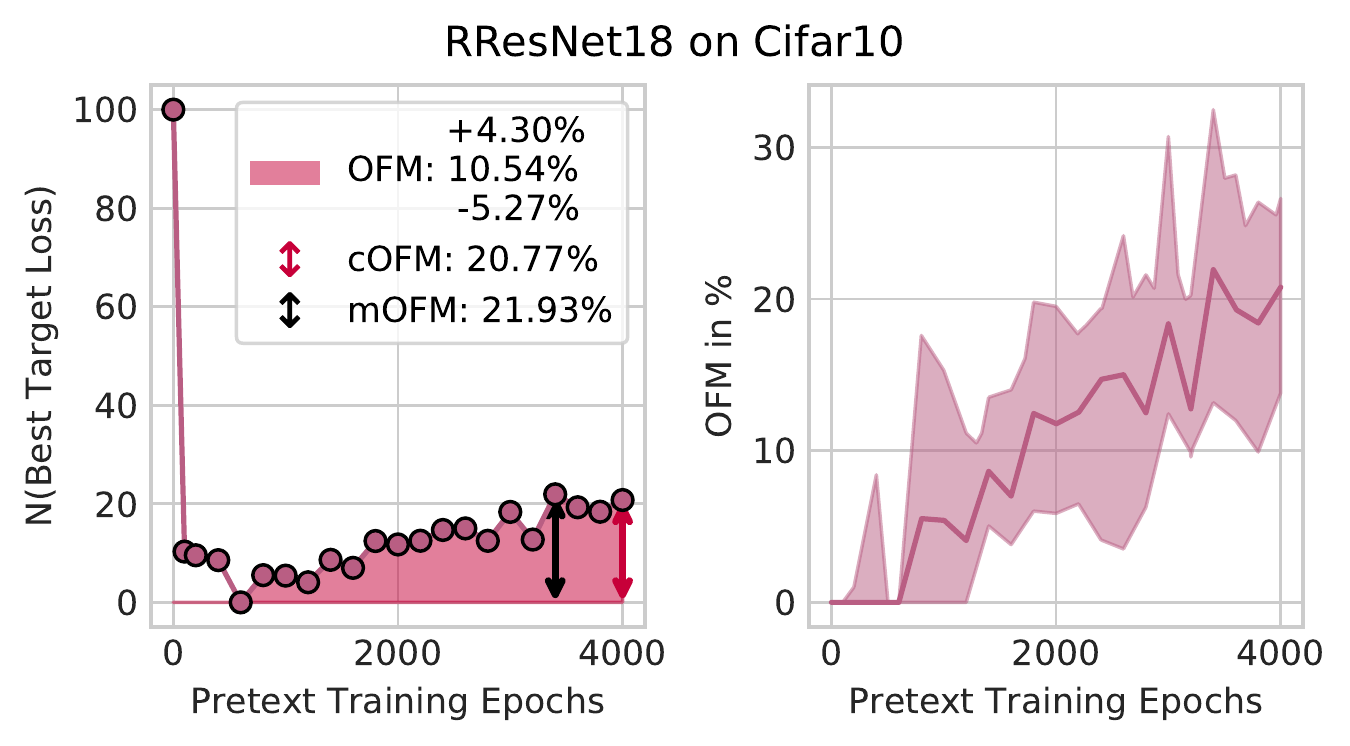} \\
		\end{tabular}
	\end{center}
	\vspace{-3ex}
	\caption[$\mathrm{OFM}$ for different pretext tasks trained with a ResNet18 model as backbone.]{$\mathrm{OFM}$ for different pretext tasks trained with a ResNet18 model as the backbone. The mismatches are shown for the entire training process.}
	\label{fig:ofm:resnet_mismatches}
\end{figure}
\subsection{Applying our Metrics to ResNet Models}
In \autoref{fig:ofm:resnet_mismatches}, we apply our metrics to ResNet models for several pretext and target tasks. For contrastive learning, we observe a small $\mathrm{OFM}$ for Cifar10 and Cifar100, which occurs late in training after pretext model convergence. However, when we use contrastive learning as a pretext task for fine-grained tumor detection on the PCam dataset, we observe a mismatch before pretext model convergence. For the well-known rotation prediction pretext task, we observe a high mismatch in Cifar10 classification early in pretext training. In \autoref{tab:ofm:quantitativ_results_resnets}, we show the corresponding mismatches measured until pretext model convergence.

\section{Future Work}
In future work, our metrics can be used to create, tune and evaluate (self-supervised) representation learning methods for different target tasks and datasets. These metrics make it possible to quantify the extent to which a pretext task matches a target task and to determine whether the pretext task learns the right kind of representation throughout the entire training process. This enables a comparison of methods on benchmarks across different pretext tasks and models. The dependencies of the objective function mismatch on different parts of the self-supervised setup (e.g., representation size) can be explored in future work in more detail to evaluate our findings further and to create pretext tasks and model architectures that are robust against mismatches. Our metrics are defined for setups where the target models are trained on pretext model representations in general. Therefore, they can also be applied to other representation learning areas such as supervised, semi-supervised, few-shot, or biological plausible representation learning.

\section{Conclusion}
In this work, we have used the linear evaluation protocol as the basis to define and discuss metrics to measure the metrics mismatch and the objective function mismatch. With soft and hard versions of our metrics, we have collected evidence of how these mismatches relate to the pretext model's representation size, target model complexity, pretext and target augmentations, as well as pretext and target task types. Furthermore, we have observed that the epoch of target task peak performance varies strongly for different datasets and pretext tasks. This highlights the importance of the protocol and shows that comparing approaches after a fixed number of epochs does not yield the entire picture of their capability. Our protocols make it possible to define benchmarks across different target tasks, where the goal is not to mismatch with the target metrics while achieving the best possible performance.

\part{Unsupervised Transfer of\\Visual Representations} 
\graphicspath{{./main/6_chapter02/sections/figures/}}

\chapter{A Lane Detection Benchmark for Multi-Target Domain Adaptation}
\label{chap:02}
\vspace{-8mm}
\begin{abstract}
Unsupervised domain adaptation demonstrates great potential to mitigate domain shifts by transferring models from labeled source domains to unlabeled target domains. While unsupervised domain adaptation has been applied to a wide variety of complex vision tasks, only few works focus on lane detection for autonomous driving. This can be attributed to the lack of publicly available datasets. To facilitate research in these directions, we propose CARLANE, a 3-way sim-to-real domain adaptation benchmark for 2D lane detection. CARLANE encompasses the single-target datasets MoLane and TuLane and the multi-target dataset MuLane. These datasets are built from three different domains, which cover diverse scenes and contain a total of 163K unique images, 118K of which are annotated. In addition, we evaluate and report systematic baselines, including our own method, which builds upon prototypical cross-domain self-supervised learning. We find that false positive and false negative rates of the evaluated domain adaptation methods are high compared to those of fully supervised baselines. This affirms the need for benchmarks such as CARLANE to further strengthen research in unsupervised domain adaptation for lane detection. CARLANE, all evaluated models, and the corresponding implementations are publicly available at \href{https://carlanebenchmark.github.io}{\textbf{https://carlanebenchmark.github.io}}
\end{abstract}
\section{Motivation}
Vision-based deep learning systems for autonomous driving have made significant progress in the past years \cite{pan2018spatial,garnett20193d,qin2020ultra,munir2021sstn,hu2022sim}. Recent state-of-the-art methods achieve remarkable results on public, real-world benchmarks but require labeled, large-scale datasets. Annotations for these datasets are often hard to acquire, mainly due to the high expenses of labeling in terms of cost, time, and difficulty.
Instead, simulation environments for autonomous driving, such as CARLA \cite{dosovitskiy2017carla}, can be utilized to generate abundant labeled images automatically. 
However, models trained on data from simulation often experience a significant performance drop in a different domain, i.e., the real world, mainly due to the domain shift \cite{saenko2010adapting}. Unsupervised domain adaptation methods \cite{ganin2015unsupervised,wilson2020survey,long2015learning,zhu2020deep,ganin2016domain,tzeng2017adversarial,xu2019self,sun2019unsupervised} try to mitigate the domain shift by transferring models from a fully-labeled source domain to an unlabeled target domain. 
This eliminates the need for annotating images but assumes that the target domain is accessible at training time. 
While unsupervised domain adaptation has been applied to complex tasks for autonomous driving, such as object detection \cite{munir2021sstn,xu2021spg} and semantic segmentation \cite{zhao2019multi,wu2021dannet}, only few works focus on lane detection \cite{garnett2020synthetic,hu2022sim}. This can be attributed to the lack of public unsupervised domain adaptation datasets for lane detection.
\section{Contributions}
\begin{figure}[t]
	\small
	\begin{center}
		\begin{tabular}{rc@{}c@{\hskip 0.2cm}c@{}c}
			~ & \multicolumn{2}{c}{MoLane} & \multicolumn{2}{c}{TuLane} \\
			%
			Source & 
			\includegraphics[width=0.2\linewidth,valign=m]{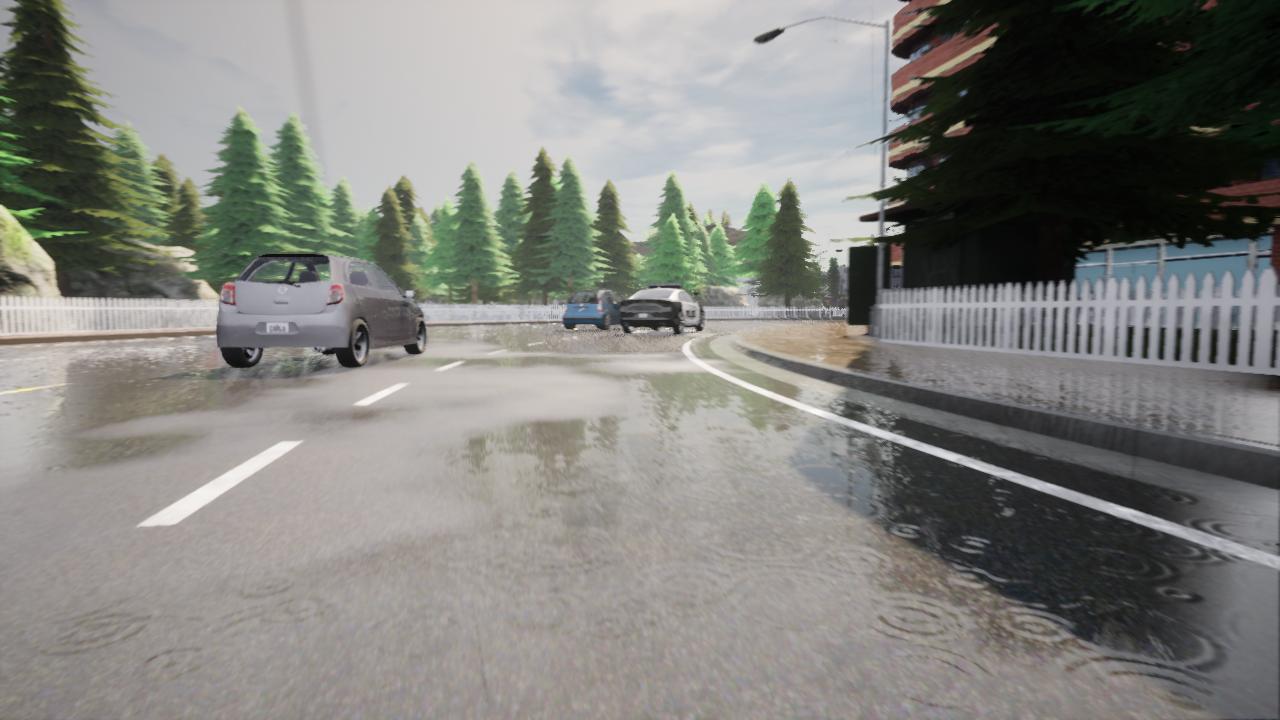} & 
			\includegraphics[width=0.2\linewidth,valign=m]{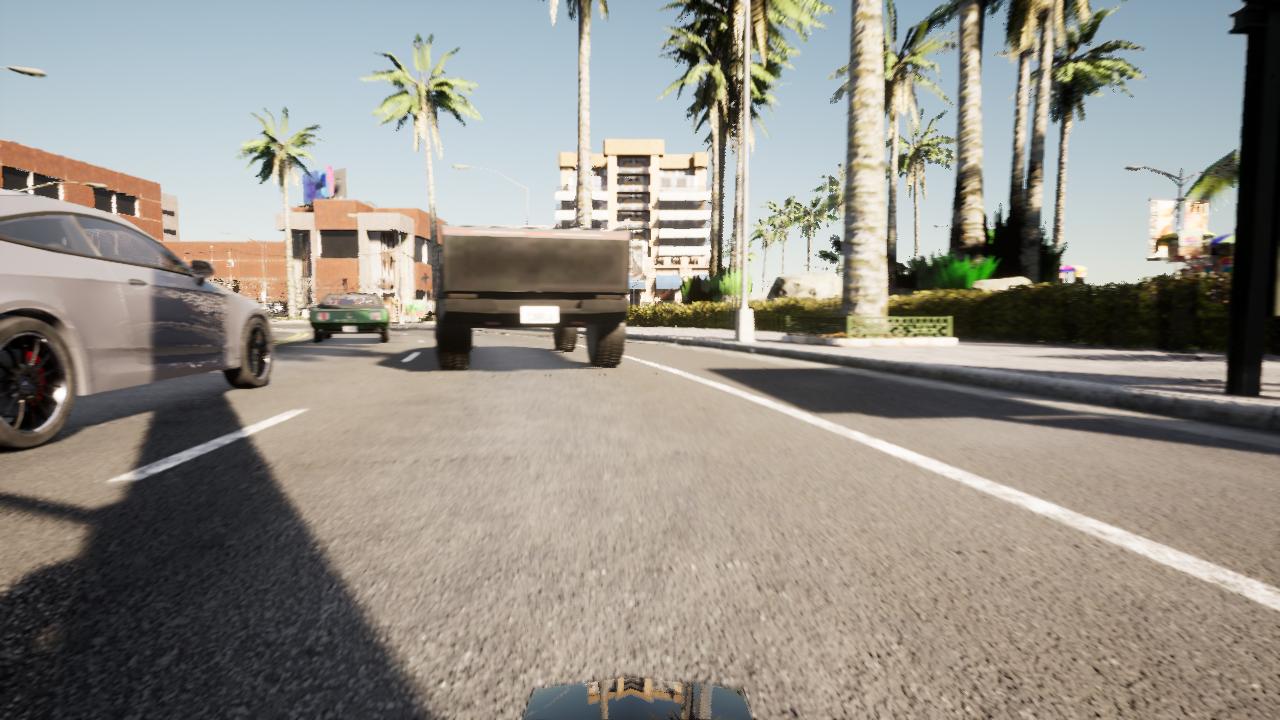} &
			\includegraphics[width=0.2\linewidth,valign=m]{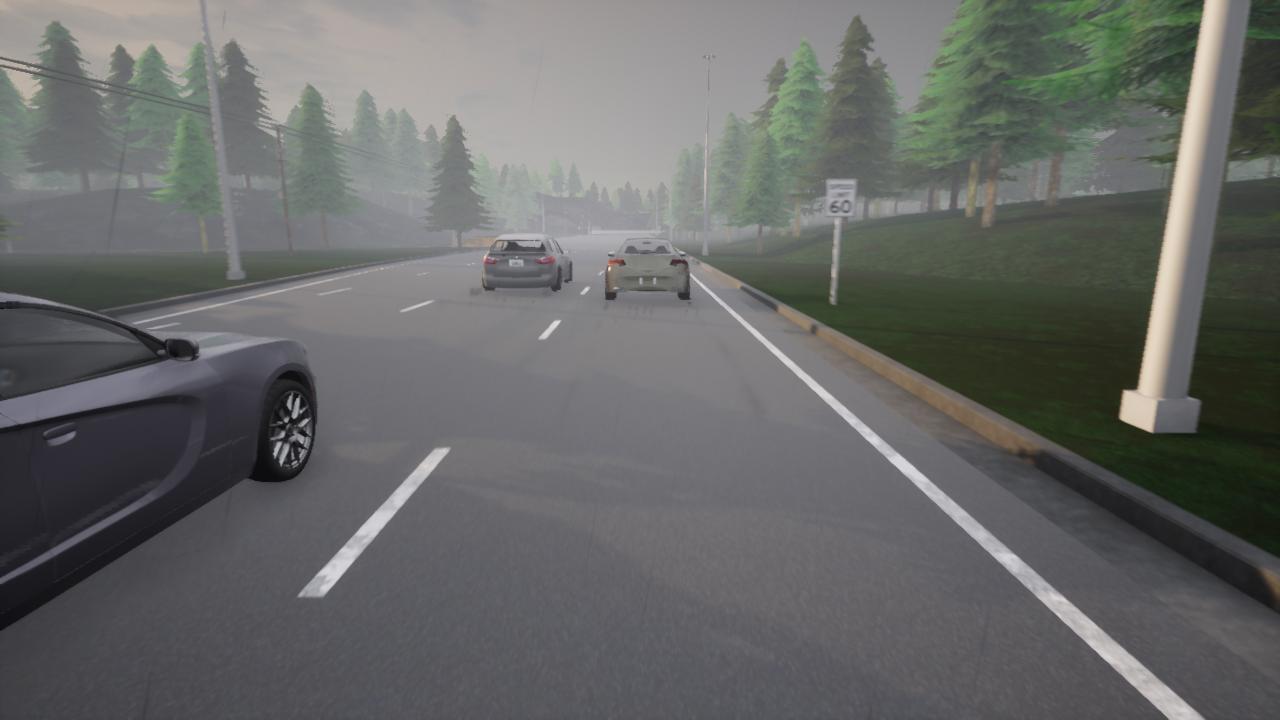} & 
			\includegraphics[width=0.2\linewidth,valign=m]{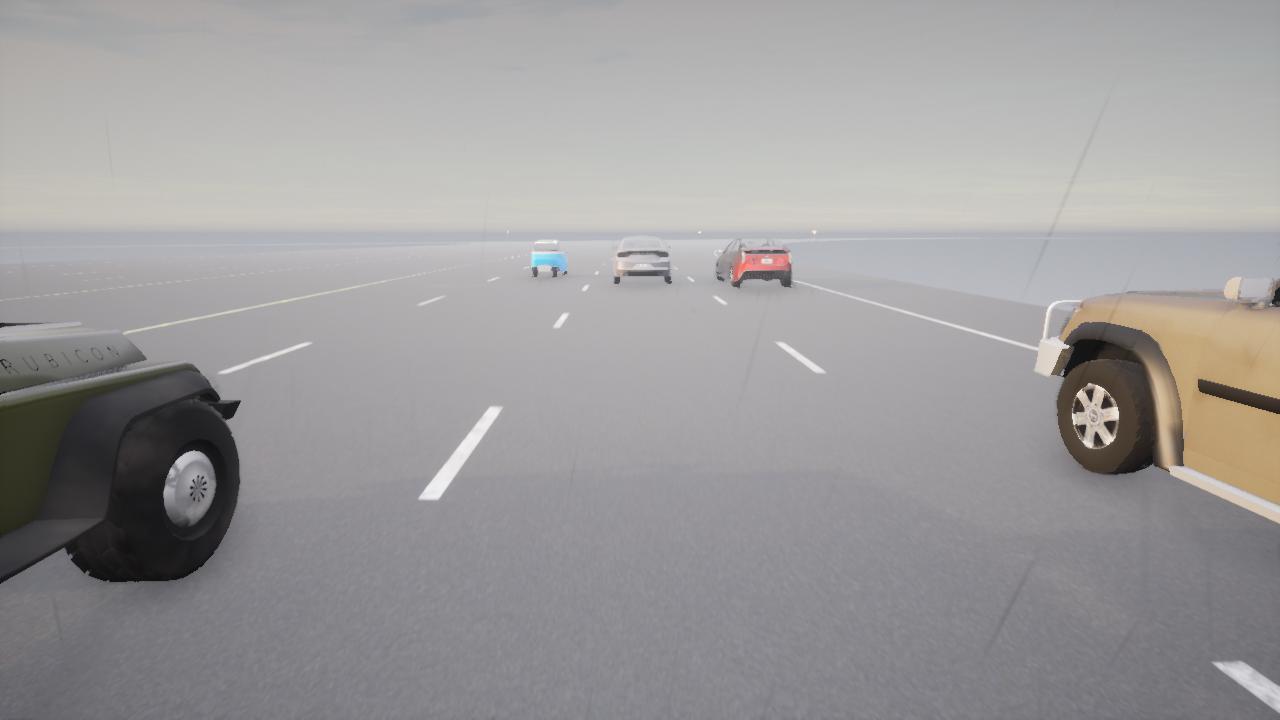}\\
			%
			Target & 
			\includegraphics[width=0.2\linewidth,valign=m]{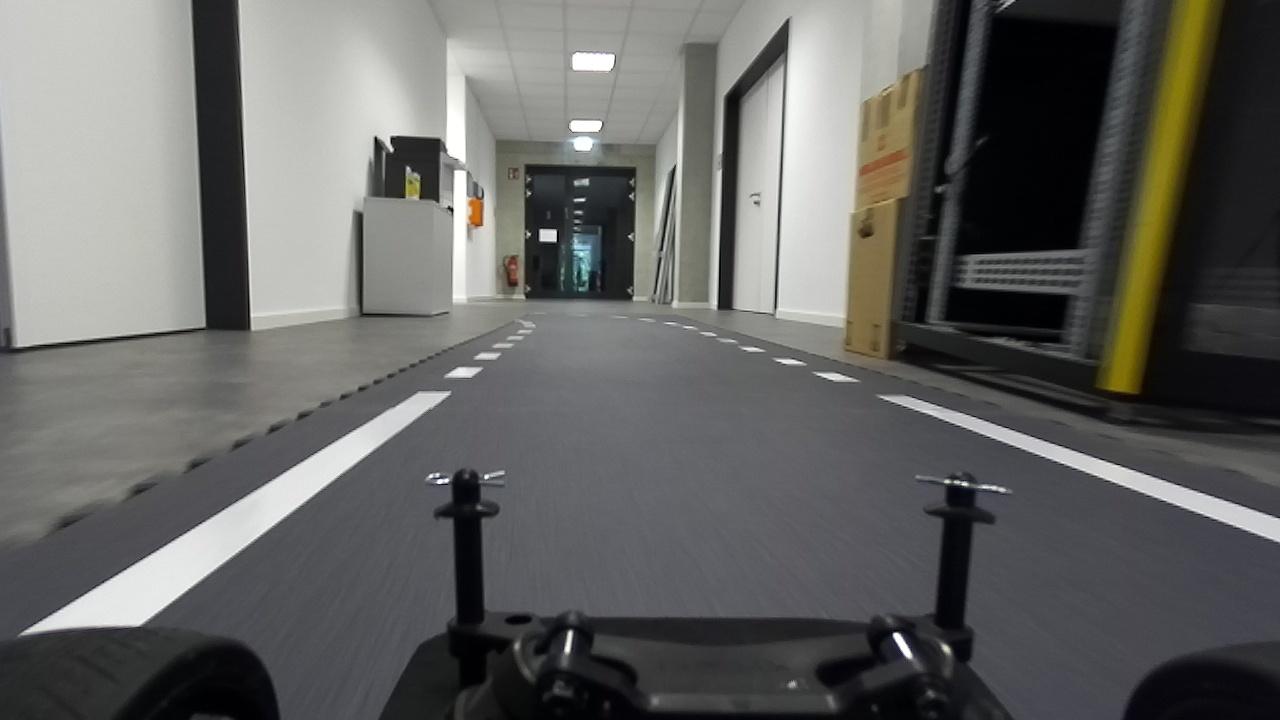} & 
			\includegraphics[width=0.2\linewidth,valign=m]{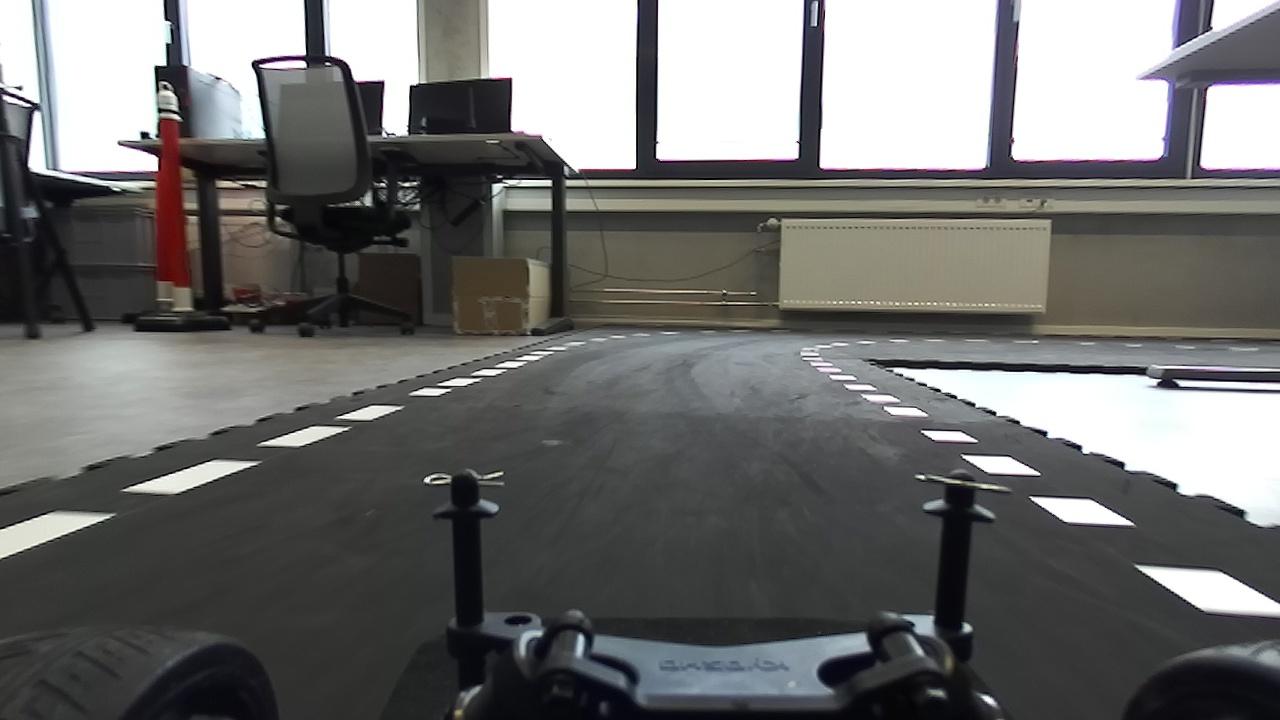} &
			\includegraphics[width=0.2\linewidth,valign=m]{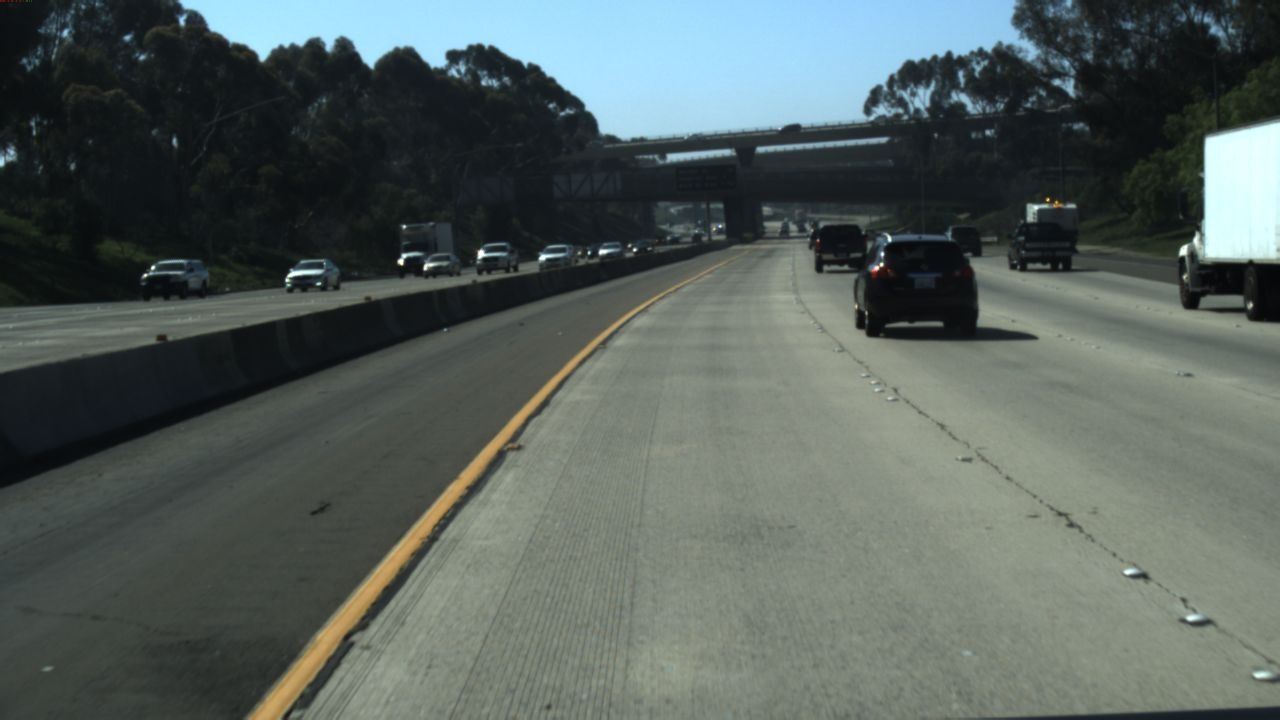} & 
			\includegraphics[width=0.2\linewidth,valign=m]{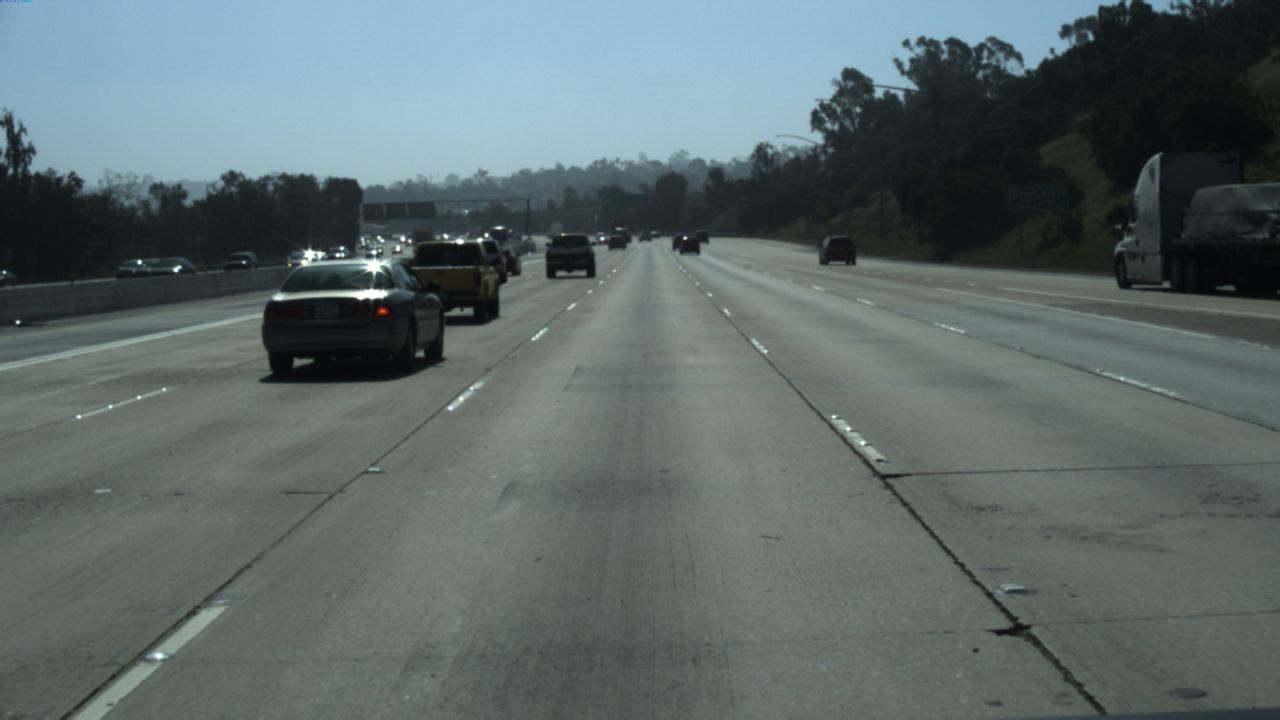}\\
		\end{tabular}
	\end{center}
	\vspace{-1ex}
	\caption[Images sampled from our CARLANE Benchmark.]{Images sampled from our CARLANE Benchmark. Best viewed in color.}
	\label{fig:carlane:intro}
\end{figure}

To compensate for this data scarcity and encourage future research, we introduce CARLANE, a sim-to-real domain adaptation benchmark for lane detection. 
We use the CARLA simulator for data collection in the source domain with a free-roaming waypoint-based agent and data from two distinct real-world domains as target domains. This enables us to construct a benchmark that consists of three datasets:
\vspace{-2pt}
\\\\
\textit{(1) MoLane} focuses on abstract lane markings in the domain of a 1/8th \textit{Mo}del vehicle. We collect 80K labeled images from simulation as the source domain and 44K unlabeled real-world images from several tracks with two lane markings as the target domain. Further, we apply domain randomization as well as data balancing. For evaluation, we annotate 2,000 validation and 1,000 test images with our labeling tool.
\vspace{-2pt}
\\\\
\textit{(2) TuLane} incorporates 24K balanced and domain-randomized images from simulation as the source domain and the well-known \textit{Tu}Simple \cite{TuSimple2017} dataset with 3,268 real-world images from U.S. highways with up to four labeled lanes as the target domain. The target domain of MoLane is a real-world abstraction from the target domain of TuLane, which may result in interesting insights about unsupervised domain adaptation.
\vspace{-2pt}
\\\\
\textit{(3) MuLane} is a balanced combination of \textit{M}oLane and T\textit{u}Lane with two target domains. For the source domain, we randomly sample 24K images from MoLane and combine them with TuLane's synthetic images. For the target domains, we randomly sample 3,268 images from MoLane and combine them with TuSimple. This allows us to investigate multi-target unsupervised domain adaptation for lane detection.
\vspace{-2pt}
\\\\
To establish baselines and investigate unsupervised domain adaptation on our benchmark, we evaluate several adversarial discriminative methods, such as DANN \cite{ganin2016domain}, ADDA \cite{tzeng2017adversarial}, and SGADA \cite{akkaya2021self}. Additionally, we propose SGPCS, which builds upon PCS \cite{yue2021prototypical} with a pseudo-labeling approach to achieve state-of-the-art performance. 
\\\\
In summary, our contributions are three-fold: (1) We introduce CARLANE, a 3-way sim-to-real benchmark, allowing single- and multi-target unsupervised domain adaptation. (\hyperref[RQ-E5]{RQ-E5}) (2) We provide several dataset tools, i.e., an agent to collect images with lane annotations in CARLA and a labeling tool to annotate the real-world images manually. (3) We evaluate several well-known unsupervised domain adaptation methods - as well as our own SGPCS method - to establish baselines and discuss results on both single- and multi-target unsupervised domain adaptation. (\hyperref[RQ-T1]{RQ-T1} and \hyperref[RQ-E6]{RQ-E6}) To the best of our knowledge, we are the first to adapt a lane detection model from simulation to multiple real-world domains. 
\section{Data Generation}
To construct our benchmark, we gather image data from a real 1/8th model vehicle and the CARLA simulator \cite{dosovitskiy2017carla}. Ensuring the verification of results and transferability to real driving scenarios, we extend our benchmark with the TuSimple dataset \cite{TuSimple2017}. This enables gradual testing, starting from simulation, followed by model cars, and ending with full-scale real word experiments. Data variety is achieved through domain randomization in all domains. However, naively performing domain randomization might lead to an imbalanced dataset. Therefore, similar driving scenarios are sampled across all domains, and a bagging approach is utilized to uniformly collect lanes by their curvature with respect to the camera position. We strictly follow TuSimple's data format \cite{TuSimple2017} to maintain consistency across all our datasets. 

\subsection{Real-World Environment}
\begin{figure}[t]
	\begin{center}
		\subfloat[]{\includegraphics[trim={-1.5cm 0 -1.5cm 0}, clip, scale=0.06]{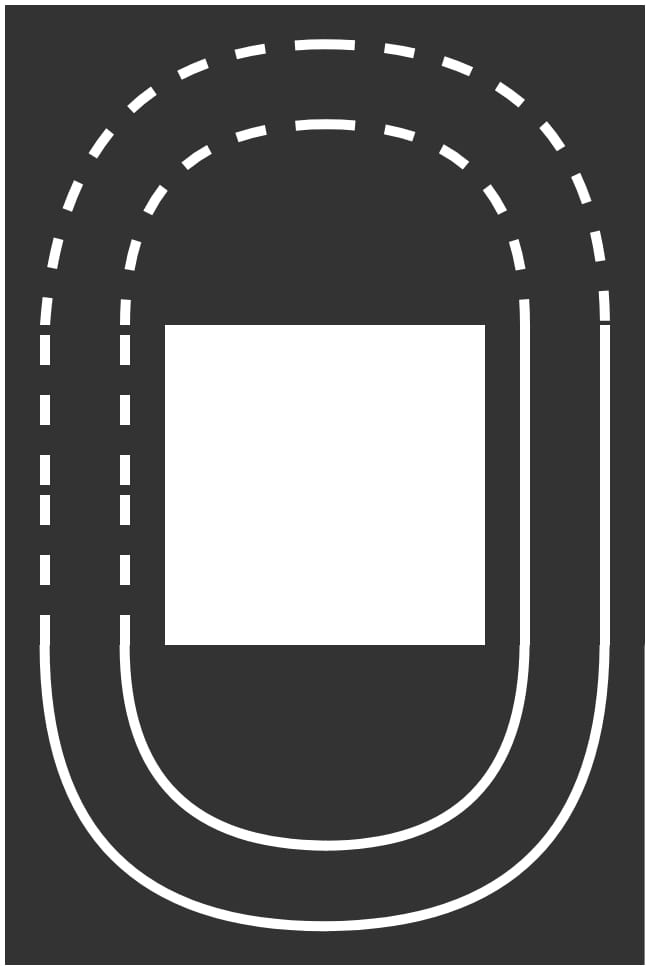}}
		\subfloat[]{\includegraphics[trim={0 0 0 0}, clip, scale=0.06]{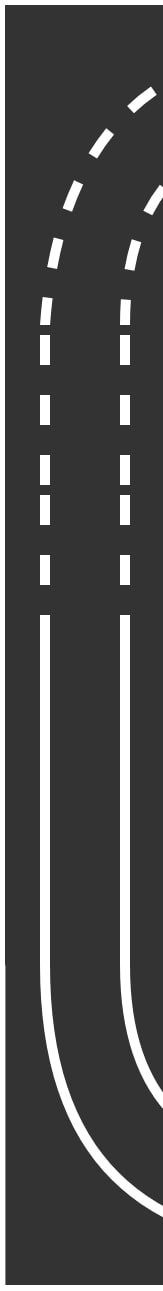}}
		\hspace*{0.1cm}
		\subfloat[]{\includegraphics[trim={0 0 0 0}, clip, scale=0.06]{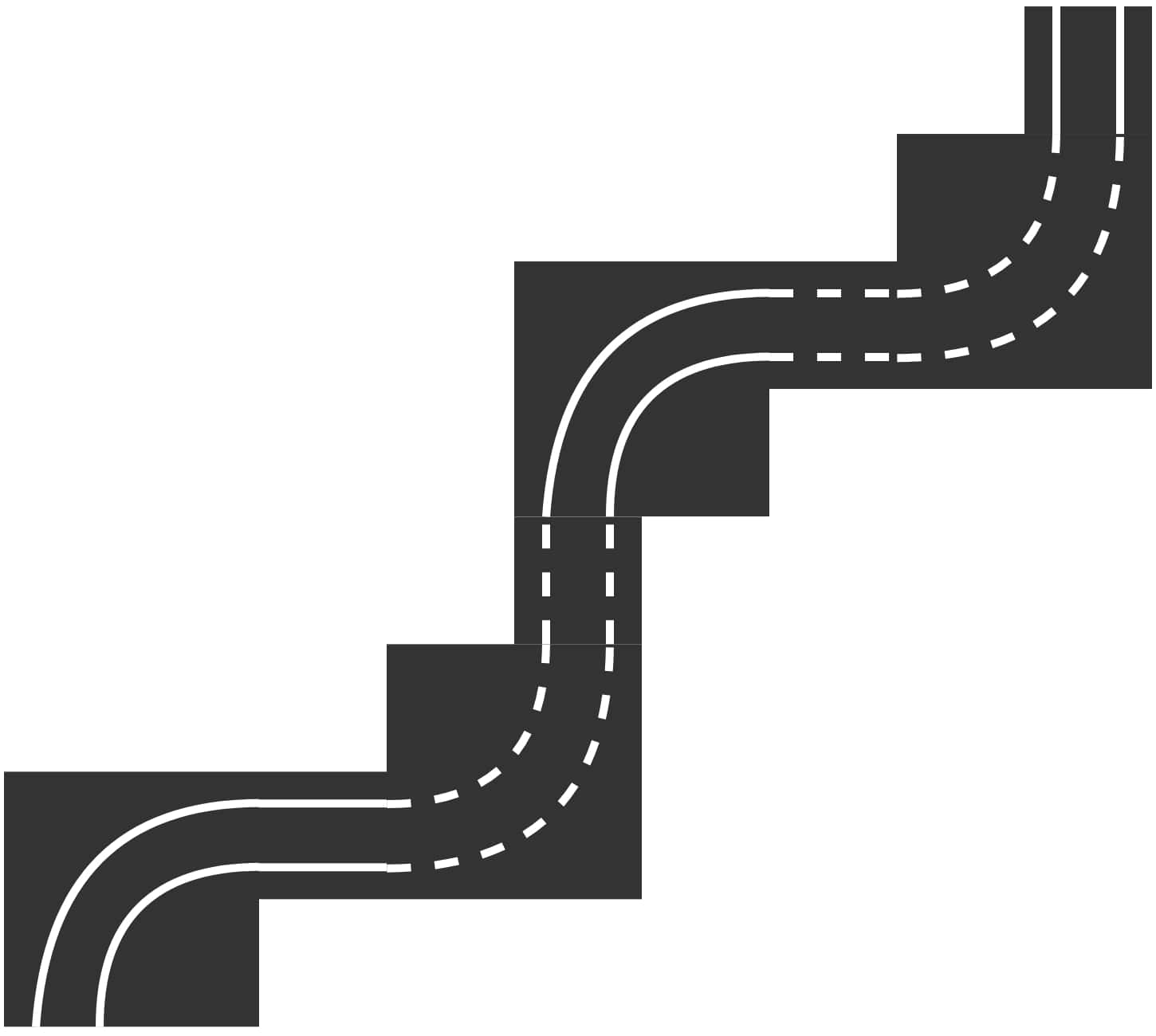}}
		\hspace*{-1.1cm}
		\subfloat[]{\includegraphics[scale=0.06]{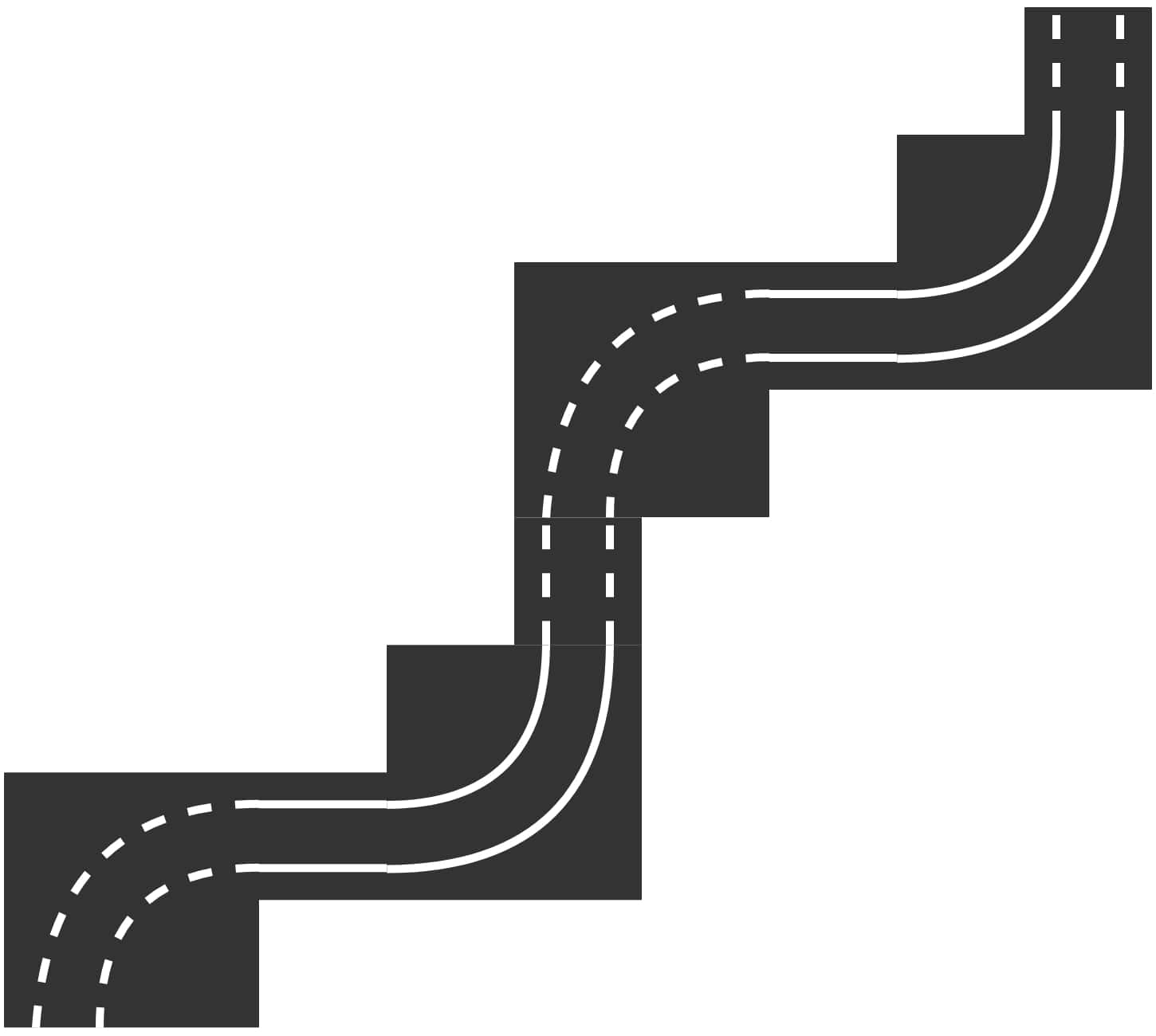}}
		\hspace*{0.1cm}
		\subfloat[]{\includegraphics[scale=0.06]{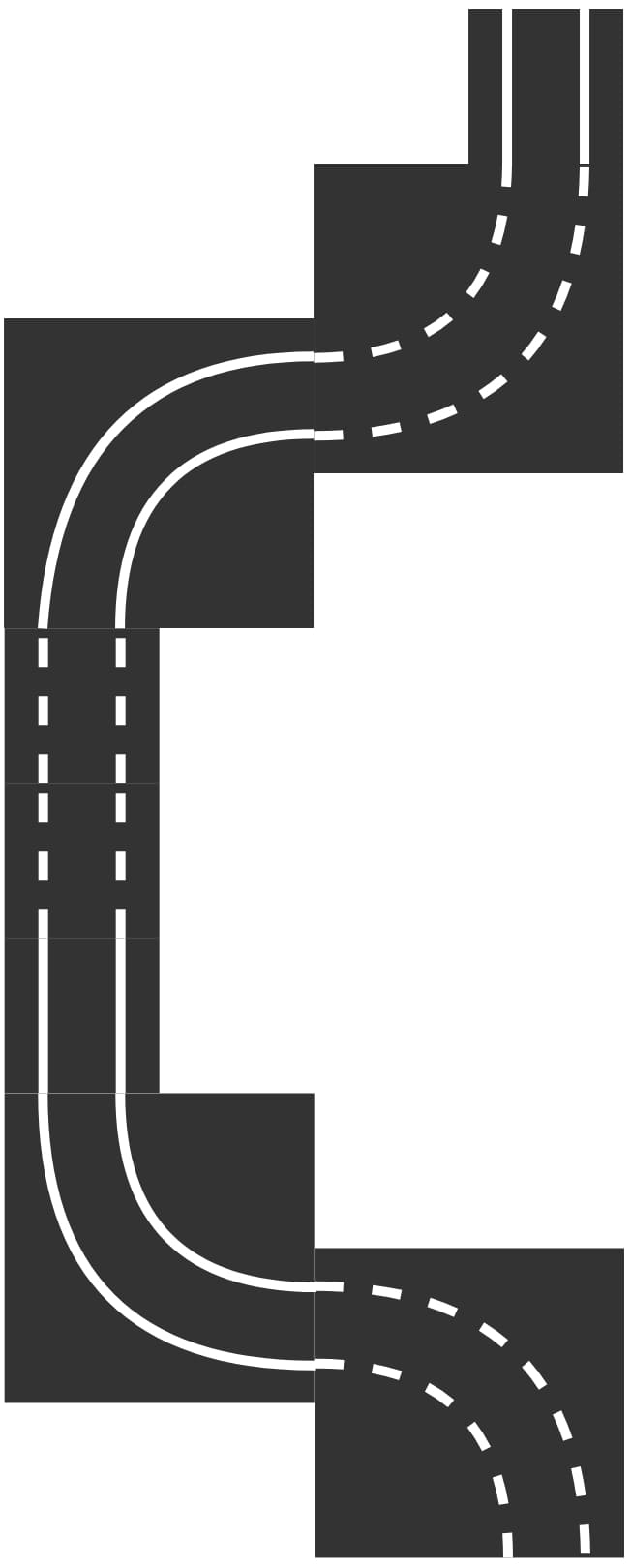}}
		\hspace*{0.1cm}
		\subfloat[]{\includegraphics[scale=0.06]{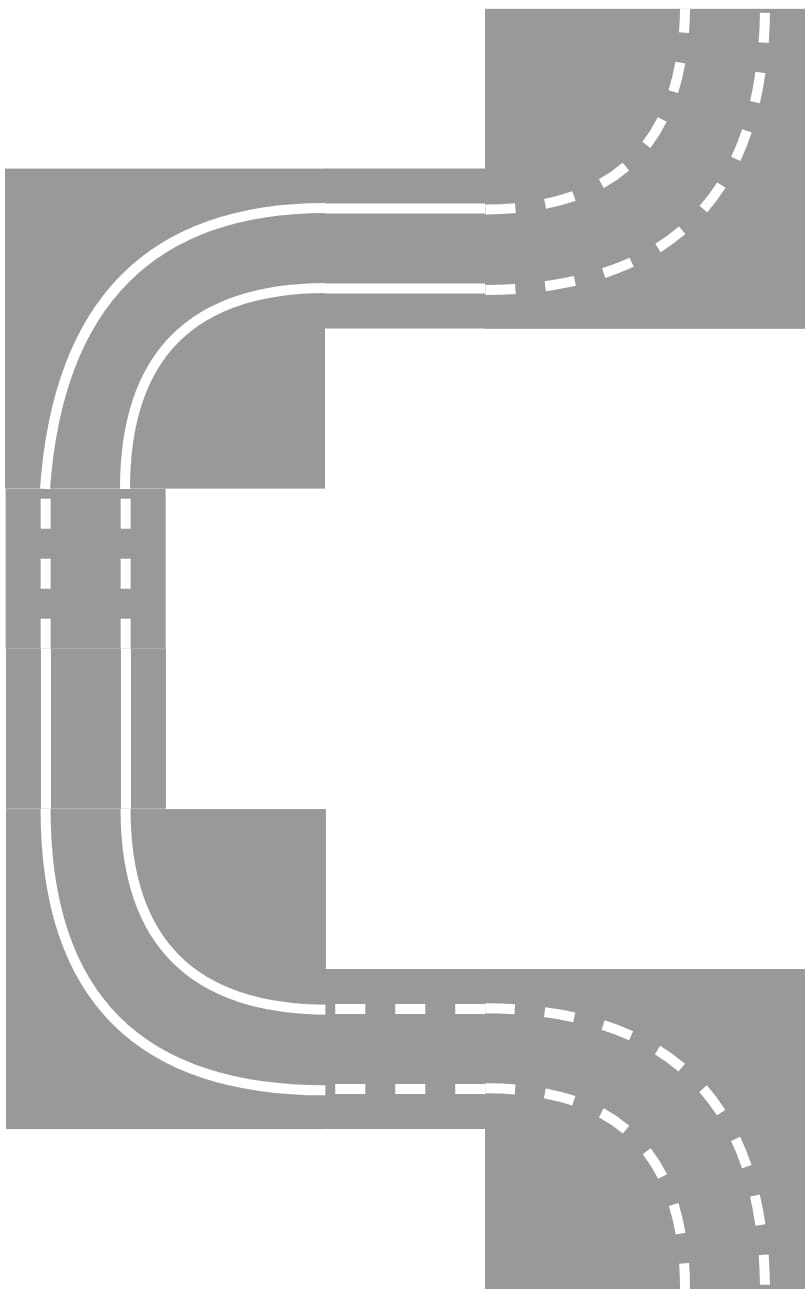}}
	\end{center}
	\vspace{-1ex}
	\caption[Overview of our track types for MoLane.]{Overview of our track types for MoLane. (a) - (d) show the black version of the training and validation tracks. These tracks are also constructed using a light gray surface material. (e) and (f) depict our test tracks.}
	\label{fig:carlane:tracks_dark}
\end{figure}

As shown in \autoref{fig:carlane:tracks_dark}, we build six different 1/8th race tracks, where each track is available in two different surface materials (dark and light gray). We vary between dotted and solid lane markings, which are pure white and \SI{50}{\milli\metre} thick. The lanes are constantly \SI{750}{\milli\metre} wide, and the smallest inner radius is \SI{250}{\milli\metre}. The track layouts are designed to roughly contain the same proportion of straight and curved segments to obtain a balanced label distribution. We construct these tracks in four locations with alternating backgrounds and lighting conditions. 

\subsection{Real-World Data Collection}
Raw image data is recorded from a front-facing Stereolabs ZEDM camera with 30 FPS and a resolution of $1280 \times 720$ pixels. A detailed description of the 1/8th car can be found in \autoref{app:02}. The vehicle is moved with a quasi-constant velocity clockwise and counter-clockwise to cover both directions of each track. All collected images from tracks (e) and (f) are used for the test subset. In addition, we annotate lane markings with our labeling tool for validation and testing, which is made publicly available.

\subsection{Simulation Environment}
We utilize the built-in APIs from CARLA to randomize multiple aspects of the agent and environment, such as weather, daytime, ego vehicle position, camera position, distractor vehicles, and world objects (i.a., walls, buildings, and plants). Weather and daytime are varied systematically by adapting parameters for cloud density, rain intensity, puddles, wetness, wind strength, fog density, sun azimuth, and sun altitude. For further details, we refer to our implementation. To occlude the lanes similar to real-world scenarios, up to five neighbor vehicles are spawned randomly in the vicinity of the agent. We consider five different CARLA maps in urban and highway environments (Town03, Town04, Town05, Town06, and Town10) to collect our dataset, as the other towns' characteristics are not suitable for our task (i.a., mostly straight lanes). In addition, we collect data from the same towns without world objects to strengthen the focus on lane detection, similar to our model vehicle target domain.

\subsection{Simulation Data Agent}
We implement an efficient agent based on waypoint navigation, which roams randomly and reliably in the aforementioned map environments and collects $1280\times720$ images. In each step, the waypoint navigation stochastically traverses the CARLA road map with a fixed lookahead distance of one meter. 
In addition, we sample offset values $\Delta y_k$ from the center lane within the range \SI{\pm 1.20}{\metre}.

To avoid saturation at the lane borders, which would occur with a sinusoidal function, we use the triangle wave function:
\begin{ceqn}
	\begin{equation}
		\Delta y_k =\frac{2m}{\pi}\arcsin(\sin(i_k))
	\end{equation}
\end{ceqn}
where $m$ is the maximal offset and $i_k$ is incremented by $0.08$ for each simulation step $k$. Per frame, our agent moves to the next waypoint with an increment of one meter, enabling the collection of highly diverse data in a fast manner. We use a bagging approach for balancing, which allows us to define lane classes based on their curvature.

\section{The CARLANE Benchmark}
\begin{table}[t]
	\RawFloats
	\caption[Dataset overview.]{Dataset overview. Unlabeled images are denoted by *, partially labeled images are denoted by **.} 
	\vspace{-1ex}
	\label{tab:carlane:dataset_overview}
	\begin{center}
		\scalebox{0.8}{%
			\setlength{\tabcolsep}{0.4em}
			\begin{tabular}{lcccccc}
				\toprule
				Dataset                  & domain             & total images & train   & validation  & test  & lanes       \\ \midrule
				\multirow{2}{*}{MoLane}  & CARLA simulation   & 84,000       & 80,000  & 4,000       & -     & \(\leq\) 2  \\ 
				& model vehicle      & 46,843       & 43,843* & 2,000       & 1,000 & \(\leq\) 2  \\ \midrule
				\multirow{2}{*}{TuLane}  & CARLA simulation   & 26,400       & 24,000  & 2,400       & -     & \(\leq\) 4  \\ 
				& TuSimple \cite{TuSimple2017} & 6,408        & 3,268   & 358         & 2,782 & \(\leq\) 4  \\ \midrule
				\multirow{2}{*}{MuLane}  & CARLA simulation   & 52,800       & 48,000  & 4,800       & -     & \(\leq\) 4  \\ 
				& model vehicle + TuSimple \cite{TuSimple2017} & 12,536      & 6,536** & 4,000       & 2,000 & \(\leq\) 4  \\ 
				\bottomrule
		\end{tabular}}
	\end{center}
	\vspace{-2ex}
\end{table}
\label{sec:carlane:CARLANE}
The CARLANE Benchmark consists of three distinct sim-to-real datasets, which we build from our three different domains. The details of the individual subsets can be found in \autoref{tab:carlane:dataset_overview}.
\\\\
\textit{MoLane} consists of images from CARLA and the real 1/8th model vehicle. For the abstract real-world domain, we collect 46,843 images with our model vehicle, of which 2,000 validation and 1,000 test images are labeled. For the source domain, we use our simulation agent to gather 84,000 labeled images. To match the label distributions between both domains, we define five lane classes based on the relative angle $\beta$ of the agent to the center lane for our bagging approach: strong left curve ($\beta\leq$\ang{-45}), soft left curve (\ang{-45} $ < \beta \leq $ \ang{-15}), straight (\ang{-15} $ < \beta <$ \ang{15}), soft right curve (\ang{15} $ \leq \beta < $ \ang{45}), and strong right curve (\ang{45}$\leq \beta$). In total, MoLane encompasses 130,843 images. 
\\\\
\textit{TuLane} consists of images from CARLA, and a cleaned version of the TuSimple dataset \cite{TuSimple2017}, which is licensed under the Apache License, Version 2.0. To clean test set annotations, we utilize our labeling tool to ensure that the up to four lanes closest to the car are correctly labeled. We adapt the bagging classes to align the source dataset with TuSimple's lane distribution: left curve (\ang{-12} $ < \beta \leq$ \ang{5}), straight (\ang{-5} $ < \beta <$ \ang{5}), and right curve (\ang{5} $ \leq \beta < $ \ang{12}). 
\\\\
\begin{figure}[t]
	\small
	\begin{center}
		\begin{tabular}{rccc}
			~ & MoLane & TuLane & MuLane \\
			Source & 
			\includegraphics[width=0.22\linewidth,valign=m,trim={6.1cm 3.9cm 12.3cm 4cm},clip]{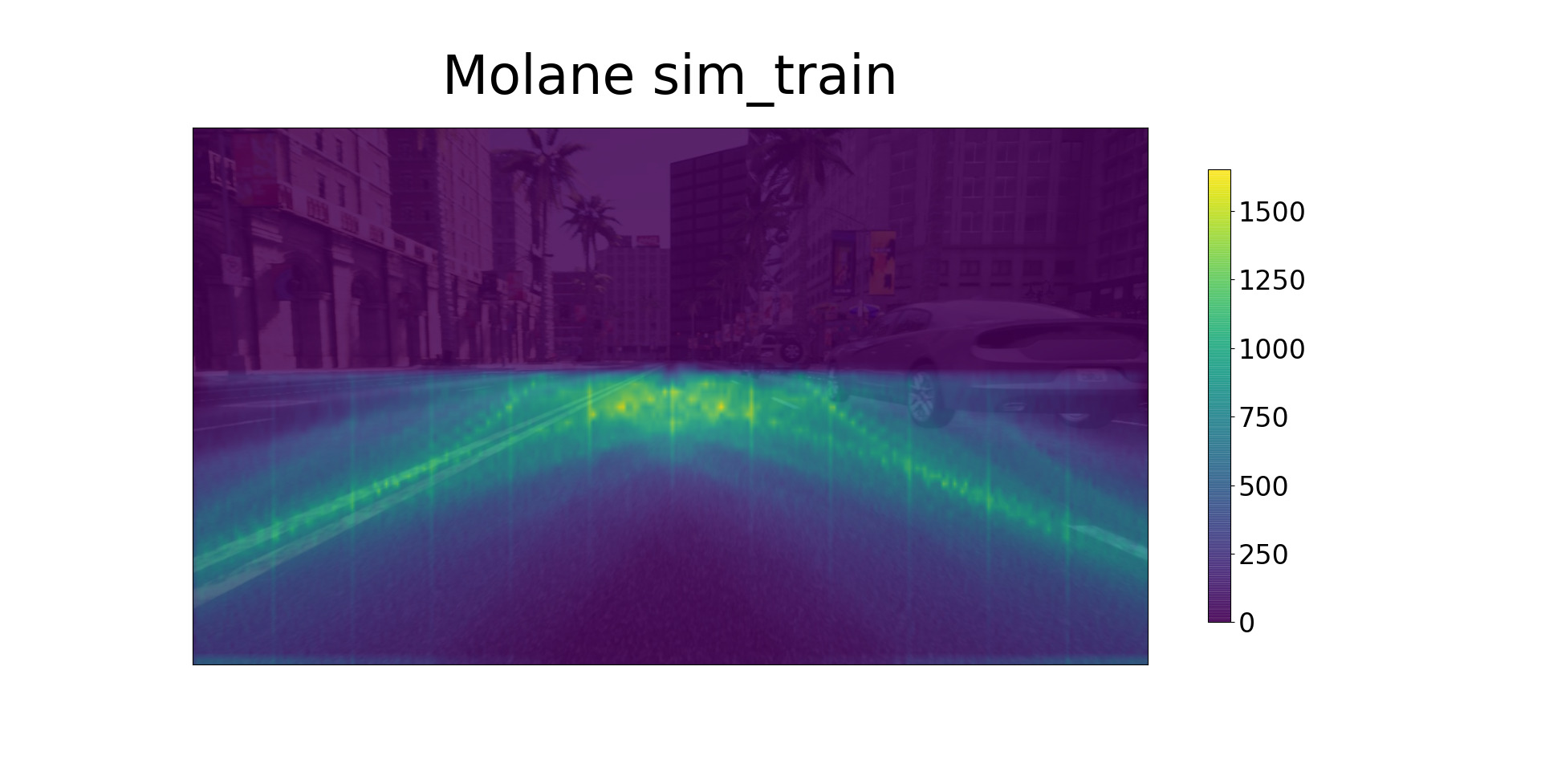} & \includegraphics[width=0.22\linewidth,valign=m,trim={6.1cm 3.9cm 12.3cm 4cm},clip]{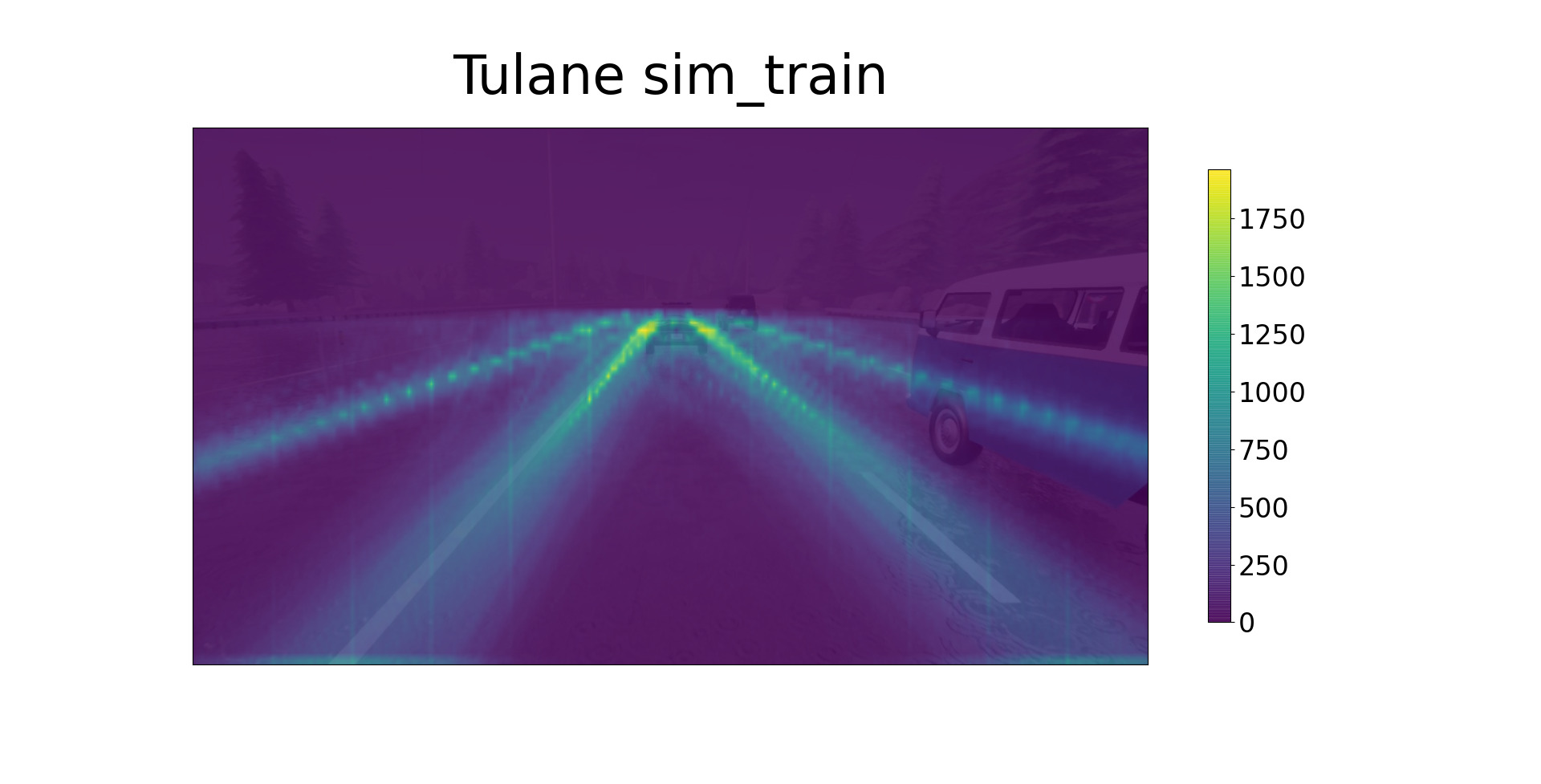} & \includegraphics[width=0.22\linewidth,valign=m,trim={6.1cm 3.9cm 12.3cm 4cm},clip]{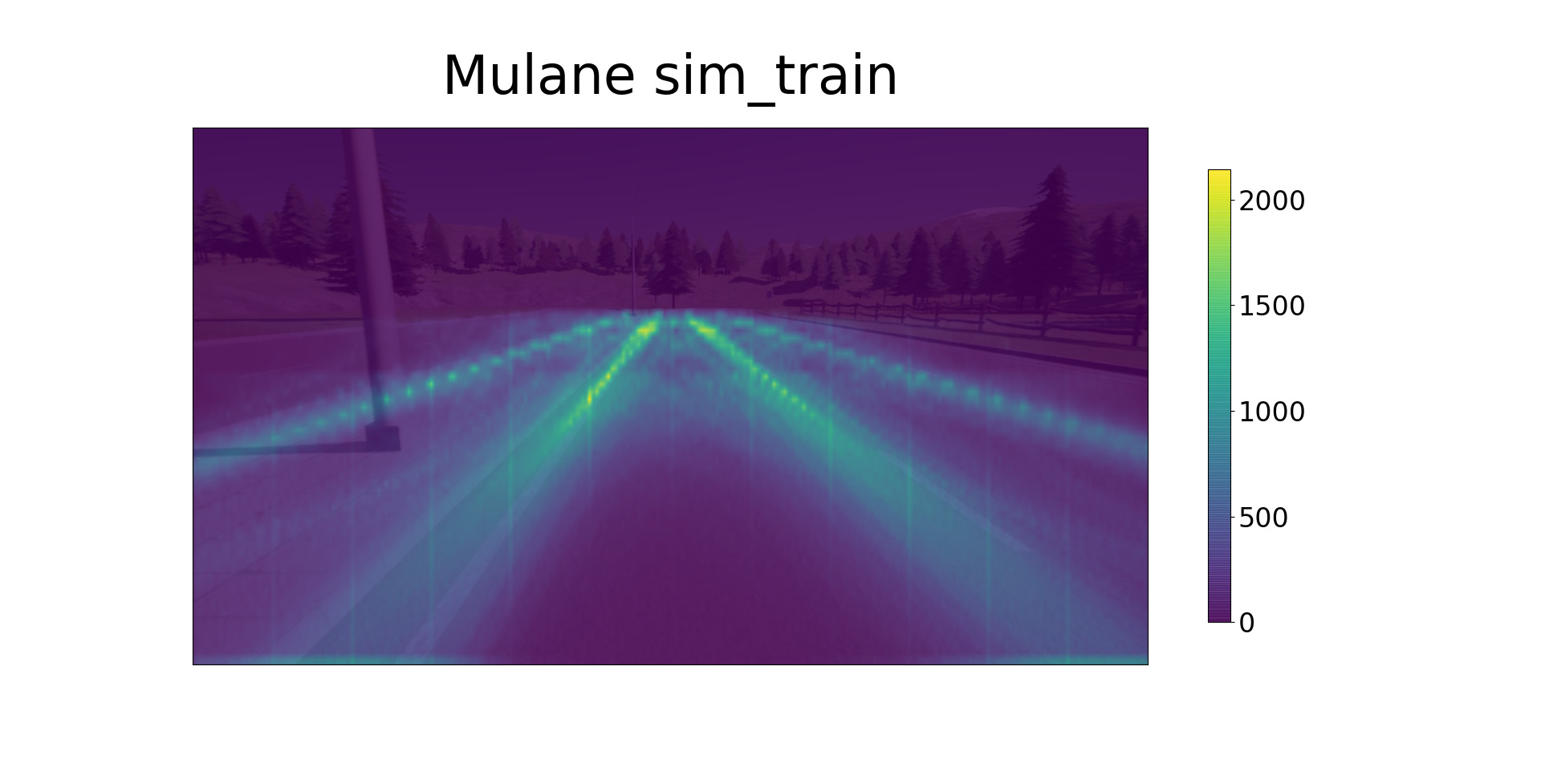}\\
			~ & ~ & ~ \\
			Target & 
			\includegraphics[width=0.22\linewidth,valign=m,trim={6.1cm 3.9cm 12.3cm 4cm},clip]{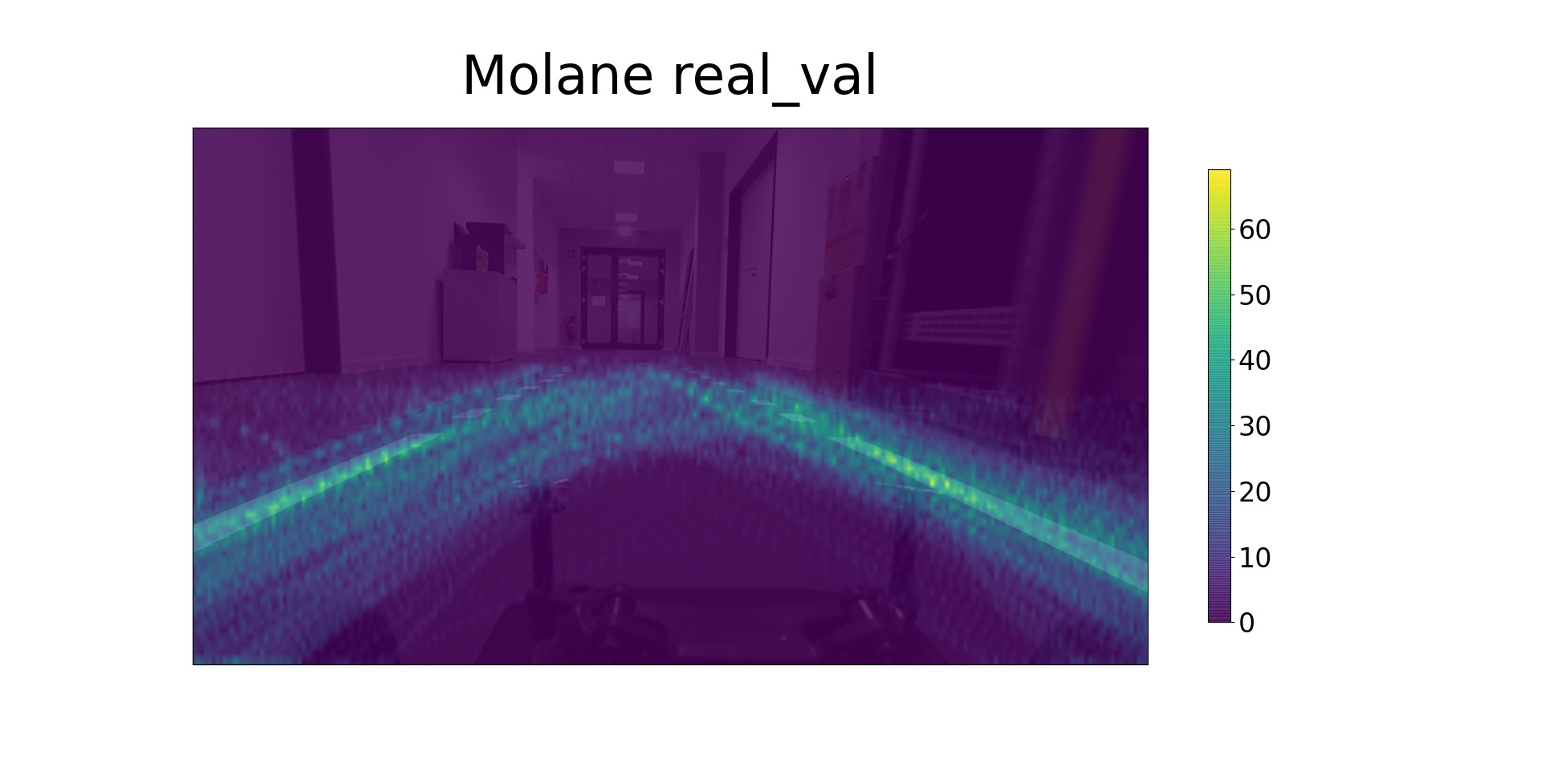} & \includegraphics[width=0.22\linewidth,valign=m,trim={6.1cm 3.9cm 12.3cm 4cm},clip]{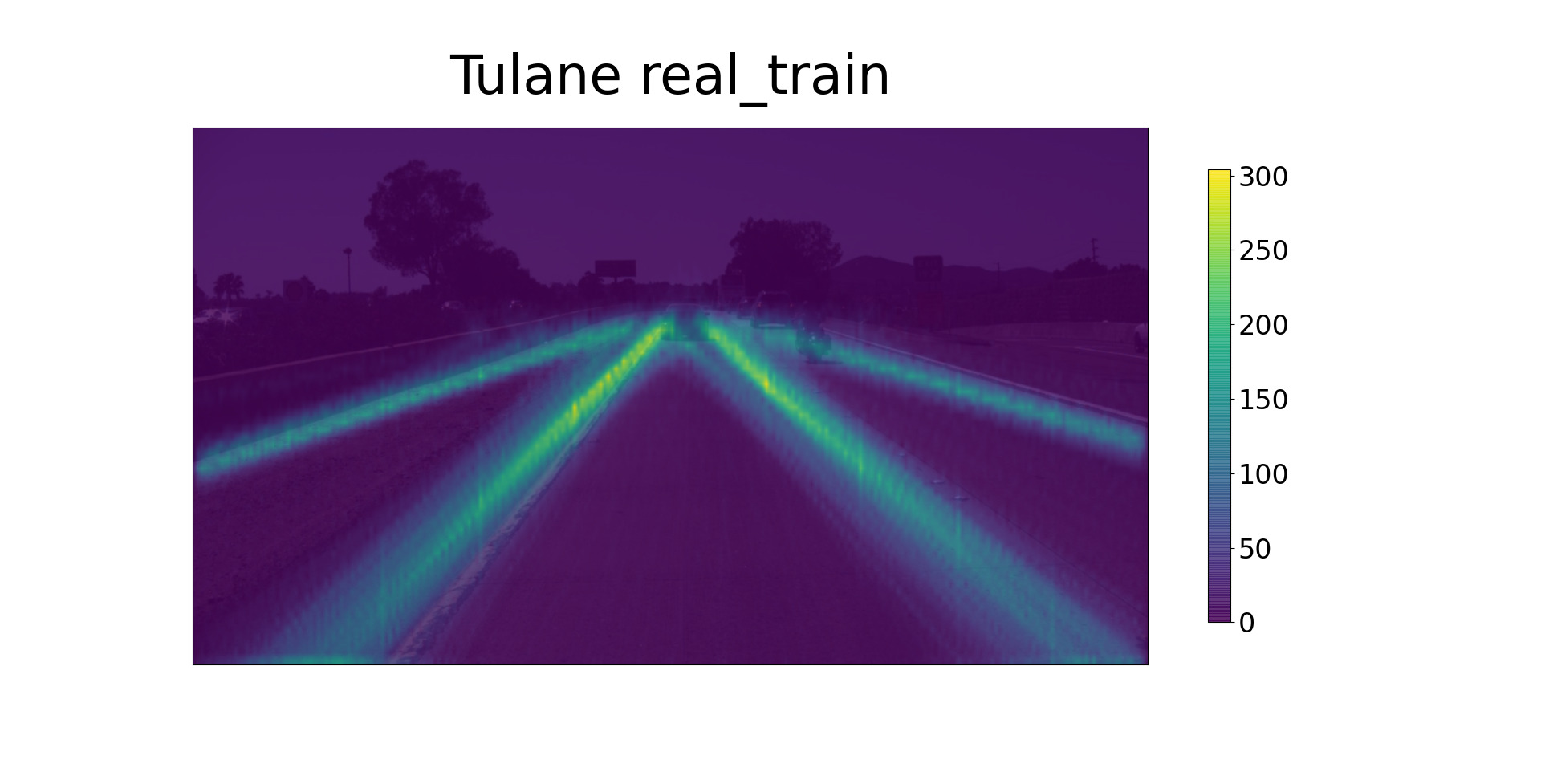} & \includegraphics[width=0.22\linewidth,valign=m,trim={6.1cm 3.9cm 12.3cm 4cm},clip]{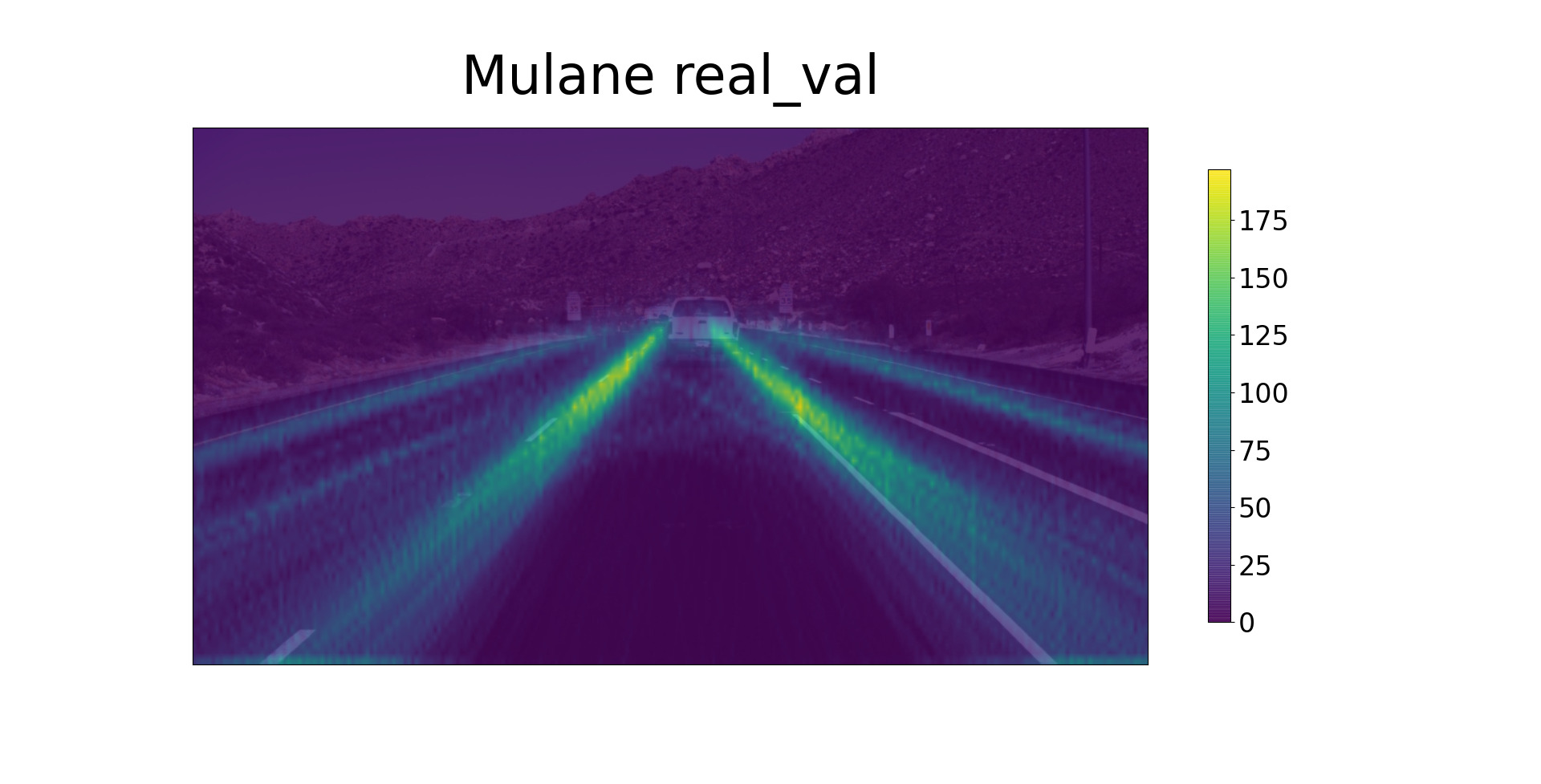}\\
		\end{tabular}
	\end{center}
	\vspace{-1ex}
	\caption[Lane annotation distributions of the three subsets of CARLANE.]{Lane annotation distributions of the three subsets of CARLANE. Since the real-world training data of MoLane and MuLane is unlabeled, we utilize their validation data for visualization.}
	\label{fig:carlane:dataset_distribution}
\end{figure}
\textit{MuLane} is a multi-target unsupervised domain adaptation dataset and is a balanced mixture of images from MoLane and TuLane. For MuLane's entire training set and its source domain validation and test set, we use all available images from TuLane and sample the same amount of images from MoLane. We adopt the 1,000 test images from MoLane's target domain and sample 1,000 test images from TuSimple to form MuLane's test set. For the validation set, we use the 2,000 validation images from MoLane and 2,000 of the remaining validation and test images of TuLane's target domain. In total, MuLane consists of 65,336 images.
\\\\
To further analyze CARLANE, we visualize the ground truth lane distributions in \autoref{fig:carlane:dataset_distribution}. We observe that the lane distributions of source and target data from our datasets are well aligned. 
\\\\
MoLane, TuLane, and MuLane are publicly available at\\ \href{https://carlanebenchmark.github.io}{https://carlanebenchmark.github.io} and licensed under the Apache License, Version 2.0.

\subsection{Dataset Format}
For each dataset, we split training, validation, and test samples into source and target subsets. Lane annotations are stored within a \emph{.json} file containing the lanes' y-values discretized by raw anchors, the lanes' x-values, and the image file path following the data format of TuSimple \cite{TuSimple2017}. Additionally, we adopt the method from \cite{qin2020ultra} to generate \emph{.png} lane segmentations and a \emph{.txt} file containing the linkage between the raw images and their segmentation as well as the presence and absence of a lane.  

\subsection{Dataset Tasks}
The main task of our datasets is unsupervised domain adaptation for lane detection, where the goal is to predict lane annotations $Y_{t} \in \mathbb{R}^{R \times G \times N}$ given the input image $X_{t} \in \mathbb{R}^{H \times W \times 3}$ from the unlabeled target domain $\mathcal{D}_{\mathcal{T}} = \{(X_{t})\}_{t\in \mathcal{T}}$. $R$ defines the number of row anchors, $G$ the number of griding cells, and $N$ the number of lane annotations available in the dataset, where the definition of $Y_{t}$ follows \cite{TuSimple2017}. During training time, the images $X_{s} \in \mathbb{R}^{H \times W \times 3}$, corresponding labels $Y_{s} \in \mathbb{R}^{H \times W \times C}$ from the source domain $\mathcal{D}_{\mathcal{S}} = \{(X_{s},Y_{s})\}_{s\in \mathcal{S}}$, and the unlabeled target images $X_{t}$ are available. Additionally, MuLane focuses on multi-target unsupervised domain adaptation, where $\mathcal{D}_{\mathcal{T}} = \{(X_{t_1})\cup(X_{t_2})\}_{t_1\in \mathcal{T}_1, t_2\in \mathcal{T}_2}$.
\\\\
Although we focus on sim-to-real unsupervised domain adaptation, our datasets can be used for unsupervised and semi-supervised tasks and partially for supervised learning tasks. Furthermore, a real-to-real transfer can be performed between the target domains of our datasets.

\section{Benchmark Experiments}
\label{sec:carlane:experiments}
We conduct experiments on our CARLANE Benchmark for several unsupervised domain adaptation methods from the literature and our proposed method. Additionally, we train fully supervised baselines on all domains.

\subsection{Metrics}
For evaluation, we use the following metrics:
\\\\
\textit{(1) Lane Accuracy} (\acs{LA}) \cite{qin2020ultra} is defined by $\textrm{LA} = \frac{p_{c}}{p_{y}}$, where $p_c$ is the number of correctly predicted lane points and $p_{y}$ is the number of ground truth lane points. Lane points are considered as correct if their $L_1$ distance is smaller than the given threshold $t_{pc}=\frac{20}{\cos(a_{yl})}$, where $a_{yl}$ is the angle of the corresponding ground truth lane.
\\\\
\textit{(2) False Positives} (\acs{FP}) and \textit{False Negatives} (\acs{FN}) \cite{qin2020ultra}: To further determine the error rate and to draw more emphasis on mispredicted or missing lanes, we measure false positives with $\textrm{FP}=\frac{l_{f}}{l_p}$ and false negatives with $\textrm{FN}=\frac{l_m}{l_{y}}$, where $l_{f}$ is the number of mispredicted lanes, $l_p$ is the number of predicted lanes, $l_{m}$ is the number of missing lanes, and $l_{y}$ is the number of ground truth lanes. Following \cite{qin2020ultra}, we classify lanes as mispredicted if the $LA < 85\%$. 

\subsection{Baselines}
We use Ultra Fast Structure-aware Deep Lane Detection (\acs{UFLD}) \cite{qin2020ultra} as baseline and strictly adopt its training scheme and hyperparameters. \acs{UFLD} treats lane detection as a row-based classification problem and utilizes the row anchors defined by TuSimple \cite{TuSimple2017}.
To achieve a lower bound for the evaluated unsupervised domain adaptation methods, we train \acs{UFLD} as a supervised baseline on the source simulation data (\acs{UFLD}-SO). Furthermore, we train our baseline on the labeled real-world training data for a surpassable fully-supervised performance in the target domain (\acs{UFLD}-TO). Since the training images from MoLane and MuLane have no annotations, we train \acs{UFLD}-TO in these cases on the labeled validation images and validate our model on the entire test set.

\subsection{Compared Unsupervised Domain Adaptation Methods}
We evaluate the following feature-level unsupervised domain adaptation methods on the CARLANE Benchmark by adopting their default hyperparameters and tuning them accordingly. Each model is initialized with the pre-trained feature encoder of our baseline model (\acs{UFLD}-SO). The optimized hyperparameters can be found in \autoref{tab:carlane:method_hyperparameters}. 
\\\\
\textit{(1) DANN} \cite{ganin2016domain} is an adversarial discriminative method that utilizes a shared feature encoder and a dense domain classifier connected via a gradient reversal layer.
\\\\
\textit{(2) ADDA} \cite{tzeng2017adversarial} employs a feature encoder for each domain and a dense domain discriminator. Following ADDA, we freeze the weights of the pre-trained classifier of UFLD-SO to obtain final predictions.
\\\\
\textit{(3) SGADA} \cite{akkaya2021self} builds upon ADDA and utilizes its predictions as pseudo labels for the target training images. Since \acs{UFLD} treats lane detection as a row-based classification problem, we reformulate the pseudo label selection mechanism. For each lane, we select the highest confidence value from the griding cells of each row anchor. Based on their griding cell position, the confidence values are divided into two cases: absent lane points and present lane points. Thereby, the last griding cell represents absent lane points as in \cite{qin2020ultra}. For each case, we calculate the mean confidence over the corresponding lanes. We then use the thresholds defined by SGADA to decide whether the prediction is treated as a pseudo label.
\\\\
\textit{(4) SGPCS} (ours) builds upon PCS \cite{yue2021prototypical} and performs in-domain contrastive learning and cross-domain self-supervised learning via cluster prototypes. Our overall objective function comprises the in-domain and cross-domain loss from PCS, the losses defined by UFLD, and our adopted pseudo loss from SGADA. We adjust the momentum for memory bank feature updates to $0.5$ and use spherical k-means \cite{johnson2019billion} with $K=2,500$ to cluster them into prototypes.

\begin{table}[t]
	\RawFloats
	\scriptsize
	\caption[Optimized hyperparameters to achieve the reported results.]{Optimized hyperparameters to achieve the reported results. $C$ denotes domain classifier parameters, $D$ denotes domain discriminator parameters, adv the adversarial loss from \cite{tzeng2017adversarial}, cls the classifier loss, sim the similarity loss, and aux the auxiliary loss from \cite{qin2020ultra}. Loss weights are set to $1.0$ unless stated otherwise.} 
	\vspace{-1ex}
	\label{tab:carlane:method_hyperparameters}
	\begin{center}
		\scalebox{0.9}{%
			\setlength{\tabcolsep}{0.4em}
			\begin{tabular}{lccccll}
				\toprule
				Method  	& Initial Learning Rate    	             & Scheduler     & Batch Size  & Epochs    &Losses    & Other Changes \\   
				\midrule	
				UFLD-SO                &$4e^{-4}$                 & Cosine Annealing           & 4           &150  & cls, sim, aux & - \\       
				DANN                   & $1e^{-5}$, $C$: $1e^{-3}$   & $\frac{1e^{-5}}{(1 + 10p)^{0.75}}$   & 4  & 30 & cls, sim, aux, adv \cite{ganin2016domain}  & $C$: 3 fc layers (1024-1024-2)  \\   
				ADDA                    & $1e^{-6}$, $D$: $1e^{-3}$  & Constant           & 16          & 30 & map \cite{tzeng2017adversarial}, adv \cite{tzeng2017adversarial}& $D$: 3 fc layers (500-500-2) \\    
							
				\multirow{2}{*}{SGADA}  &\multirow{2}{*}{$1e^{-6}$, $D$: $1e^{-3}$} & \multirow{2}{*}{Constant} & \multirow{2}{*}{15} & \multirow{2}{*}{10} & map \cite{tzeng2017adversarial}, adv
				\cite{tzeng2017adversarial}, & \multirow{2}{*}{Pseudo label selection} \\
				&&&&& pseudo: 0.25 & \\
				 
				\multirow{3}{*}{SGPCS}  &\multirow{3}{*}{$4e^{-4}$} & \multirow{3}{*}{Cosine Annealing} & \multirow{3}{*}{16} & \multirow{3}{*}{10} & in-domain \cite{yue2021prototypical}, & \multirow{3}{*}{-} \\
				&&&&& cross-domain \cite{yue2021prototypical}, & \\
				&&&&& cls, sim, aux, pseudo: 0.25 & \\
				UFLD-TO                   &$4e^{-4}$  & Cosine Annealing           & 4          & 300 & cls, sim, aux & - \\	   
				\bottomrule	
		\end{tabular}}
	\end{center}
	\vspace{-2ex}
\end{table}

\subsection{Implementation Details}
\label{sec:carlane:implementation_details}
We implement all methods in PyTorch 1.8.1 and train them on a single machine with four RTX 2080 Ti GPUs. Tuning all methods has required a total amount of compute of approximately 3.5 petaflop/s-days. The training times for each model range from 4-13 days for \acs{UFLD} baselines and 6-44 hours for domain adaption methods. In addition, we have found that applying output scaling on the last linear layer of the model yields slightly better results. Therefore, we divide the models' output by 0.5. Our implementation is publicly available at \href{https://carlanebenchmark.github.io}{https://carlanebenchmark.github.io}.

\subsection{Evaluation}
\textbf{Quantitative Evaluation.}
In \autoref{tab:carlane:quantitativ_results_comparison}, we report the results on MoLane, TuLane, and MuLane across five different runs. We observe that \acs{UFLD}-SO is able to generalize to a certain extent to the target domain. This is mainly due to the alignment of semantic structure from both domains. ADDA, SGADA, and our proposed SGPCS manage to adapt the model to the target domain slightly and consistently. However, DANN suffers from negative transfer \cite{wang2019characterizing} when trained on MoLane and MuLane. The negative transfer of DANN for complex domain adaptation tasks is also observed in other works \cite{ tanwisuth2021prototype,fan2022self,wang2019characterizing,kim2020cross} and can be explained by the source domain's data distribution and the model complexity \cite{wang2019characterizing}. In our case, the source domain contains labels not present in the target domain, as shown in \autoref{fig:carlane:dataset_distribution}, which is more pronounced in MoLane and MuLane.

\begin{figure}[t]
	\small
	\begin{center}
		\begin{tabular}{rc@{}c@{}c@{}c}
			~ & MoLane & TuLane & \multicolumn{2}{c}{MuLane} \\
			UFLD-SO & 
			\includegraphics[width=0.18\linewidth,valign=m]{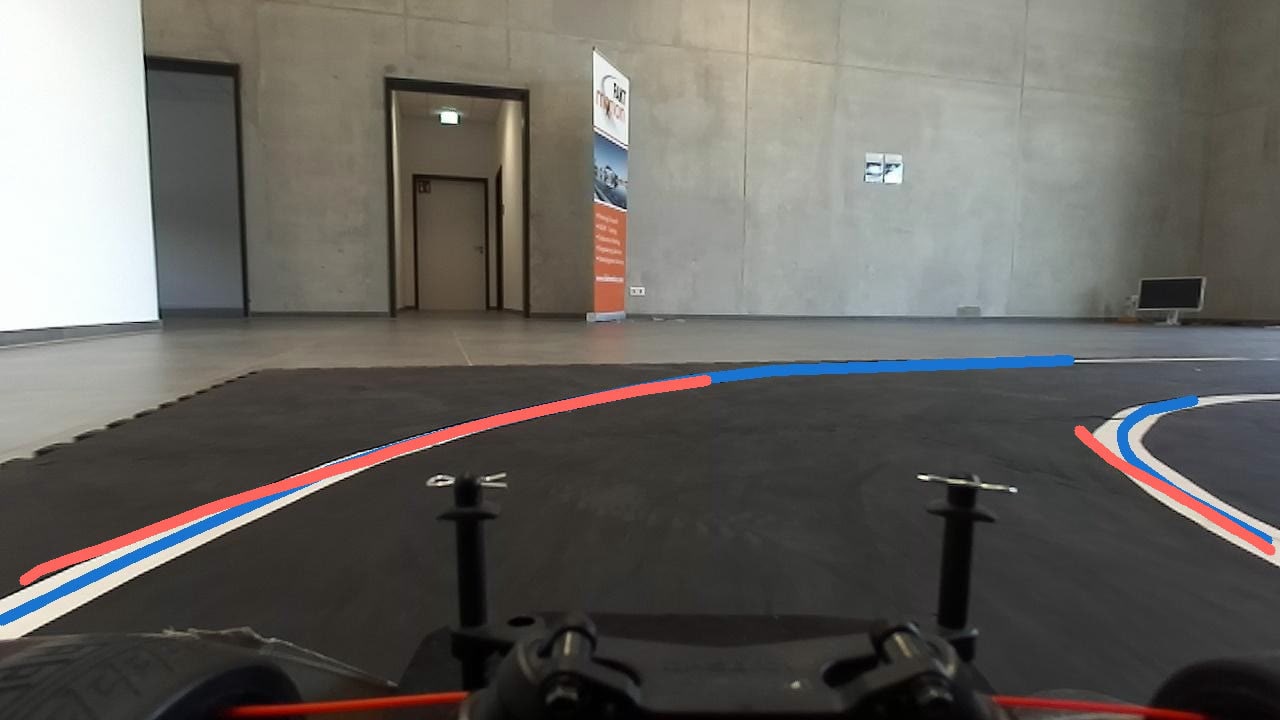} & \includegraphics[width=0.18\linewidth,valign=m]{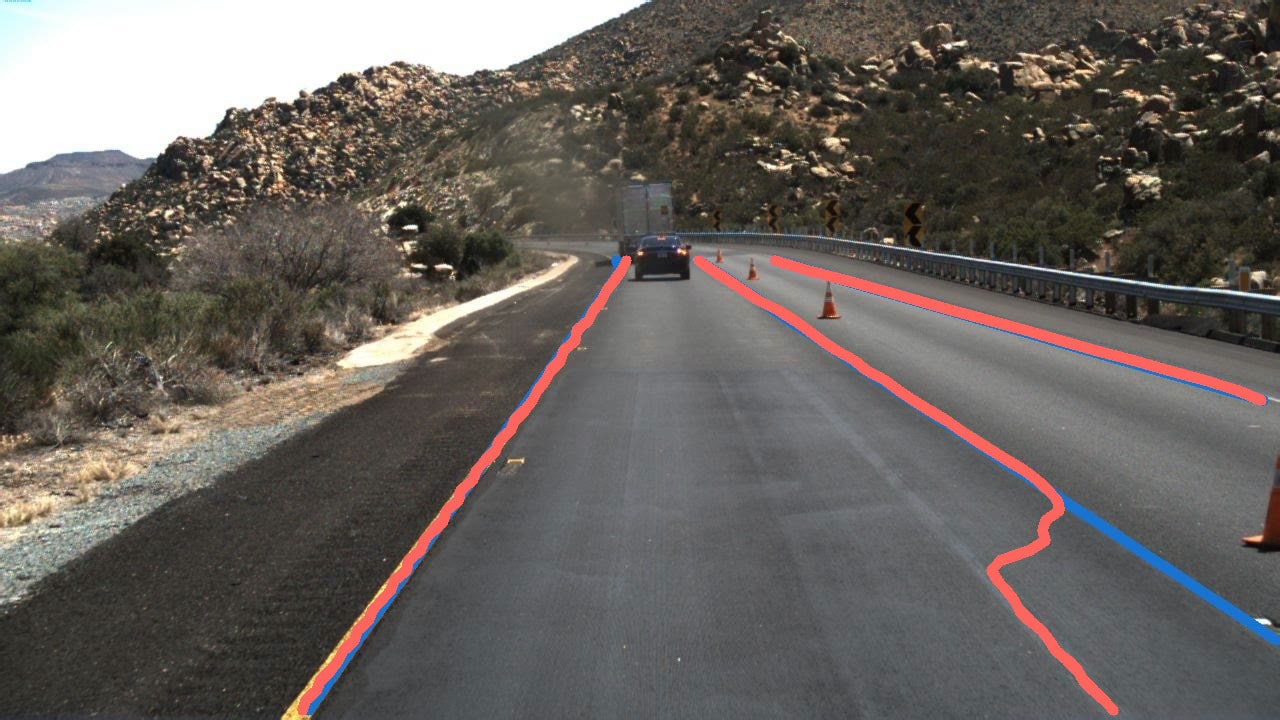} &
			\includegraphics[width=0.18\linewidth,valign=m]{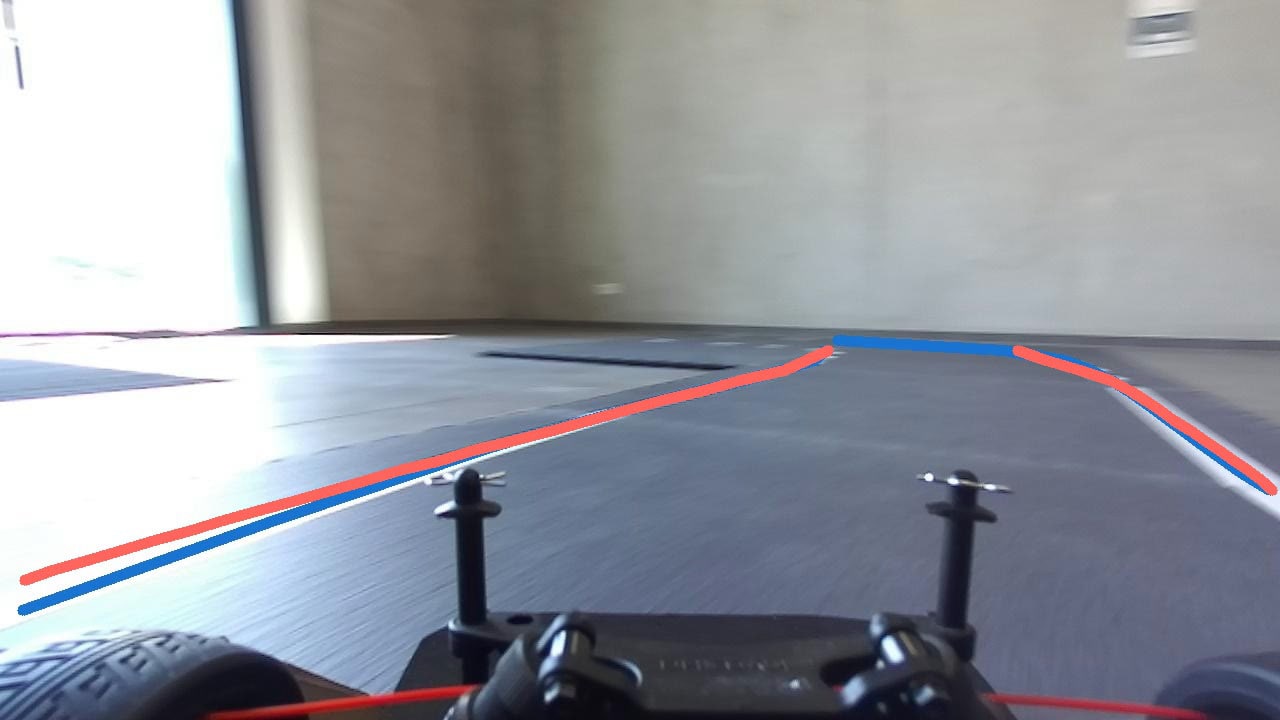} & \includegraphics[width=0.18\linewidth,valign=m]{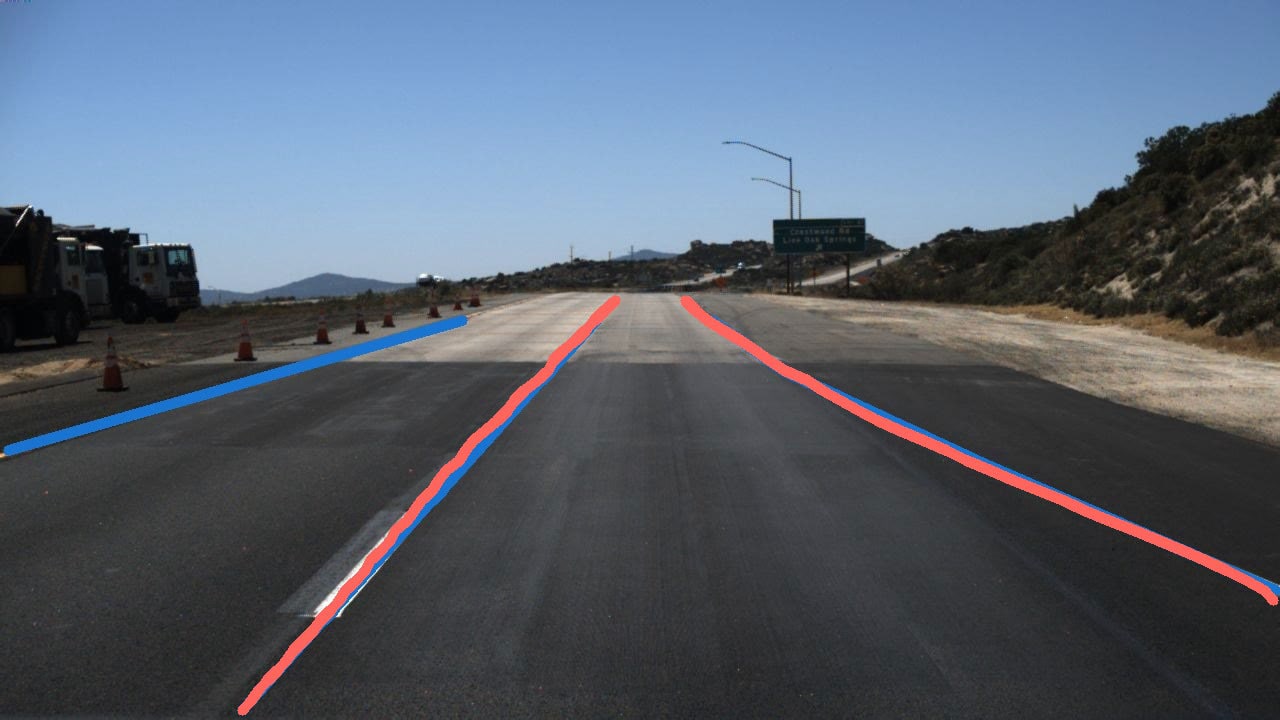}\\
			DANN & 
			\includegraphics[width=0.18\linewidth,valign=m]{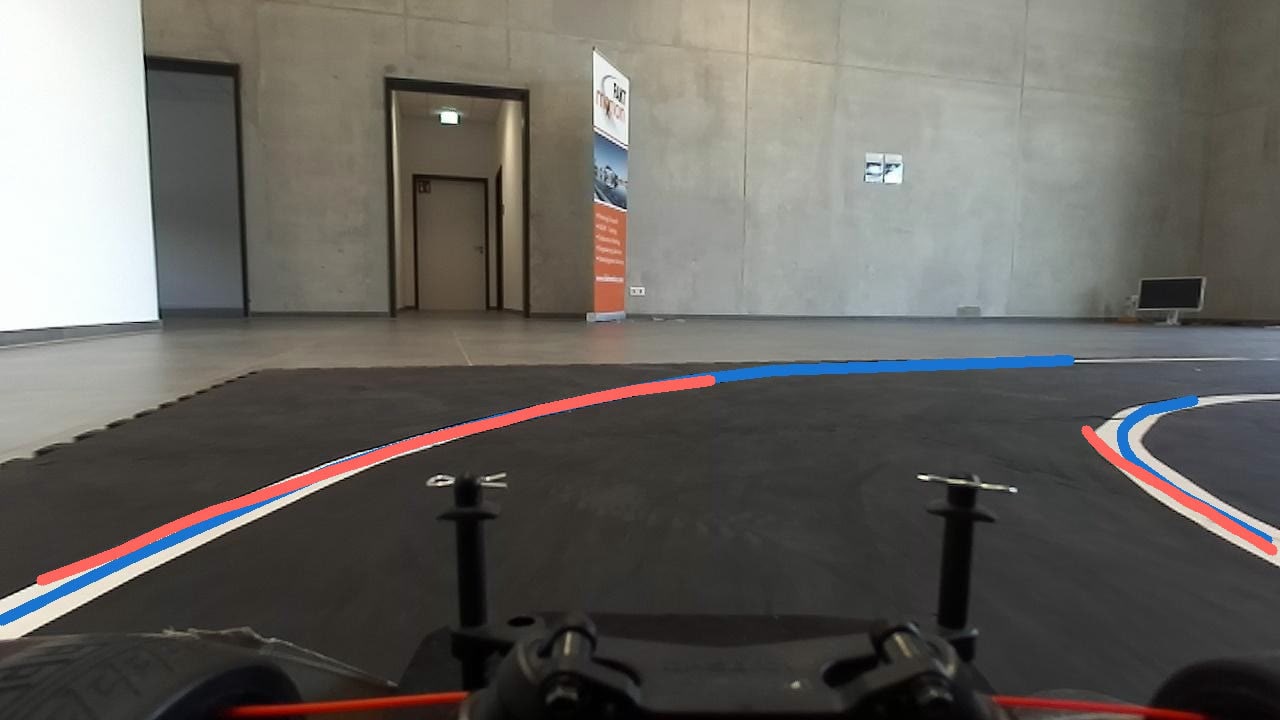} & 
			\includegraphics[width=0.18\linewidth,valign=m]{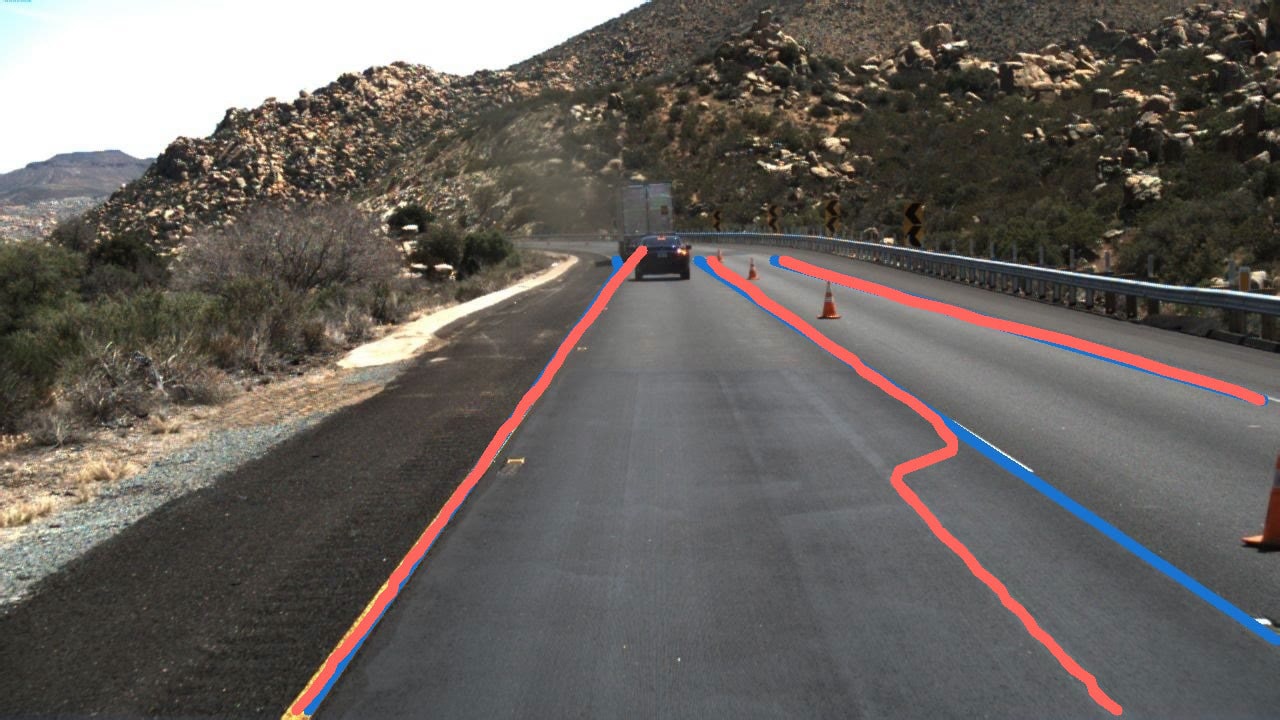} &
			\includegraphics[width=0.18\linewidth,valign=m]{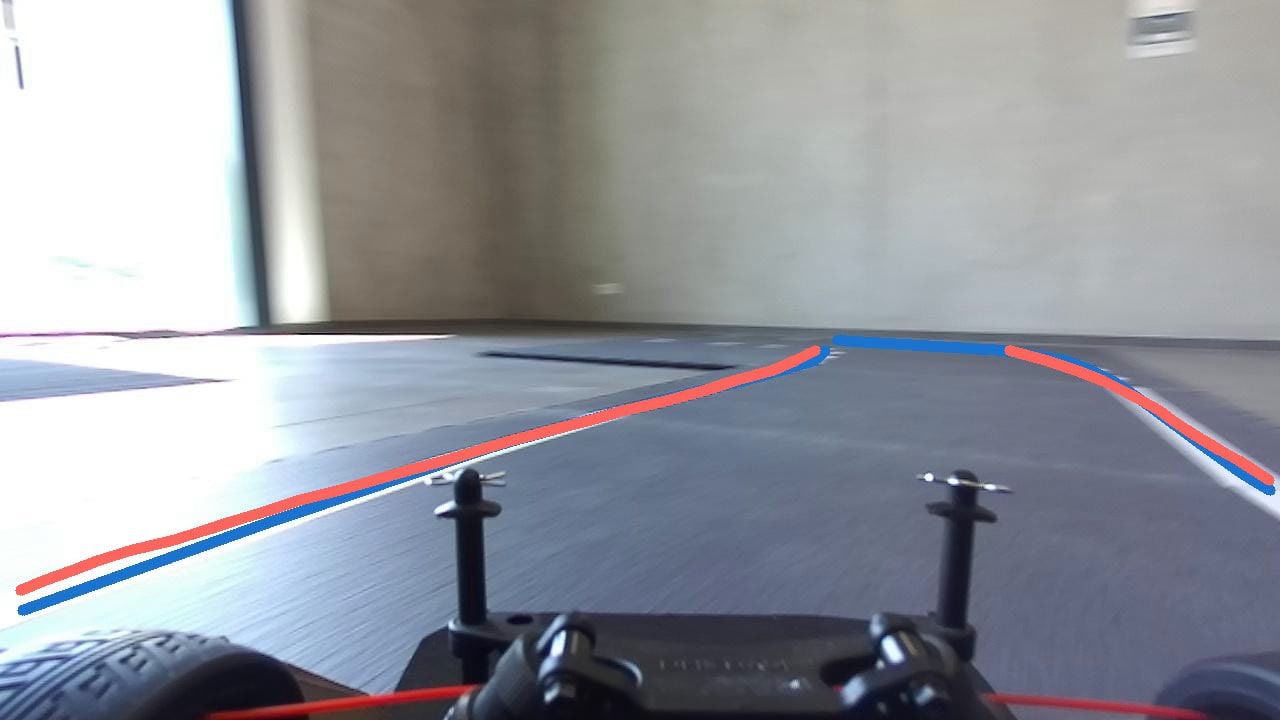} & \includegraphics[width=0.18\linewidth,valign=m]{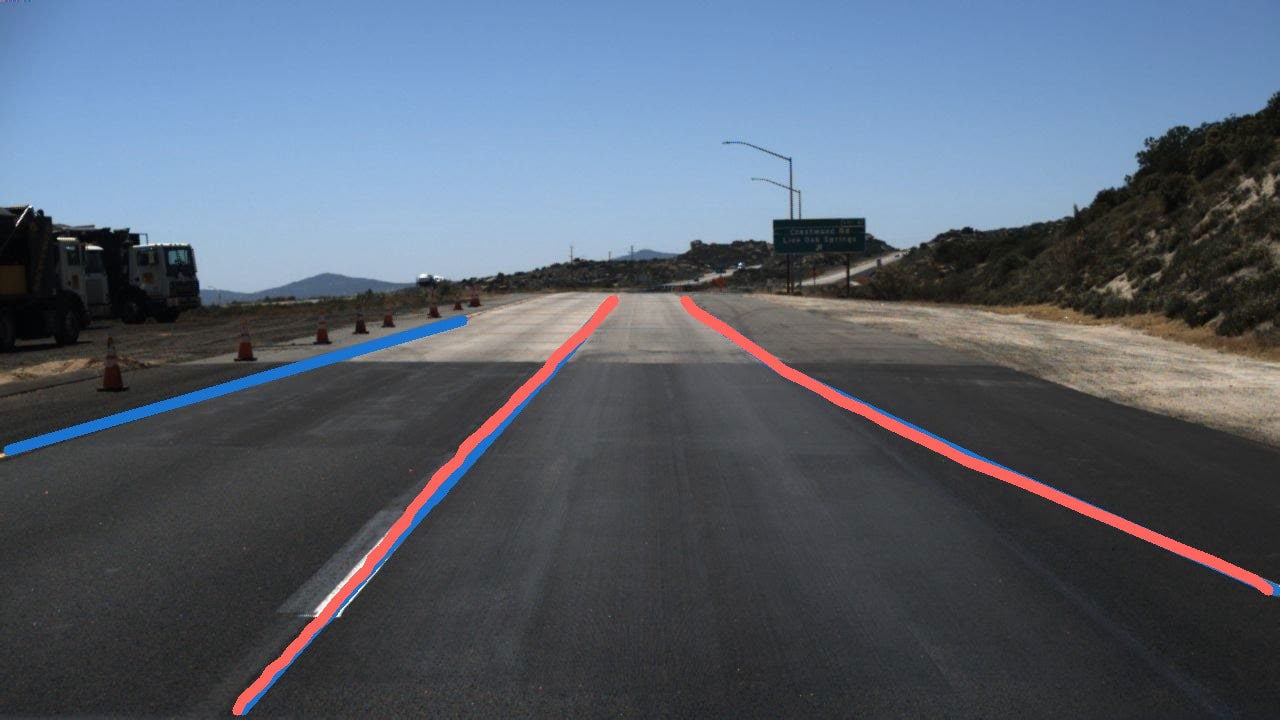}\\
			ADDA & 
			\includegraphics[width=0.18\linewidth,valign=m]{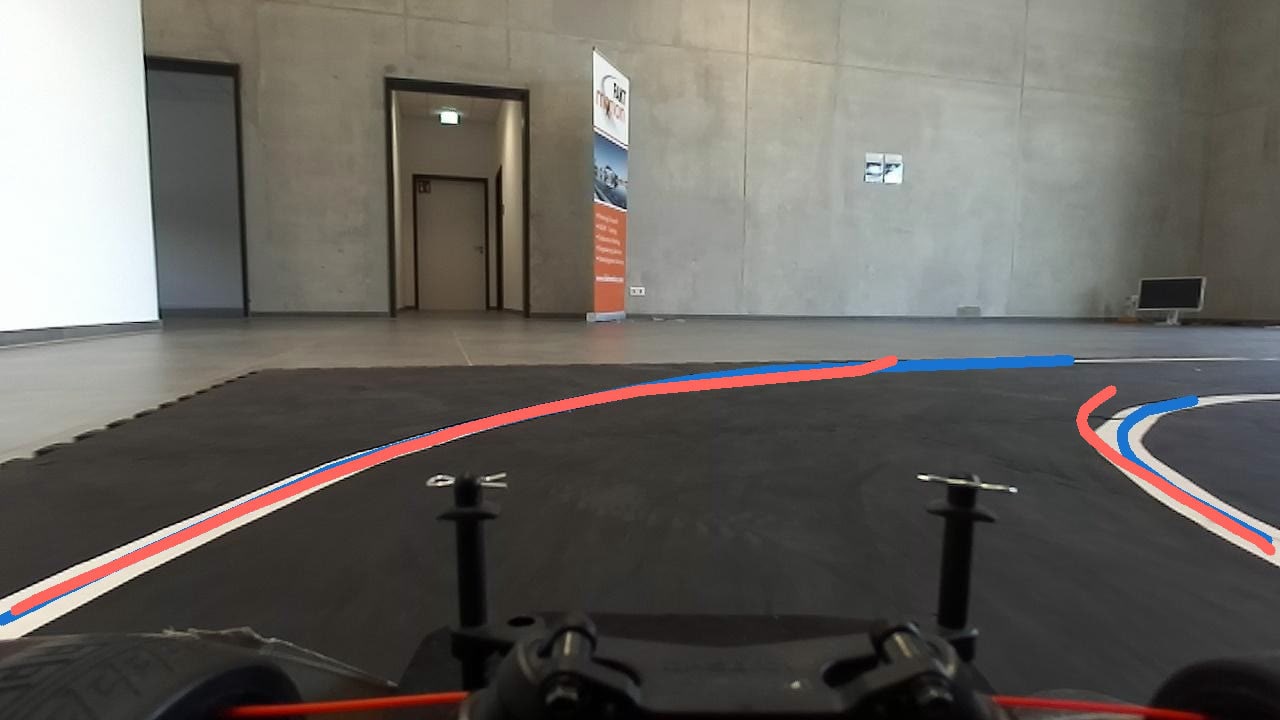} & 
			\includegraphics[width=0.18\linewidth,valign=m]{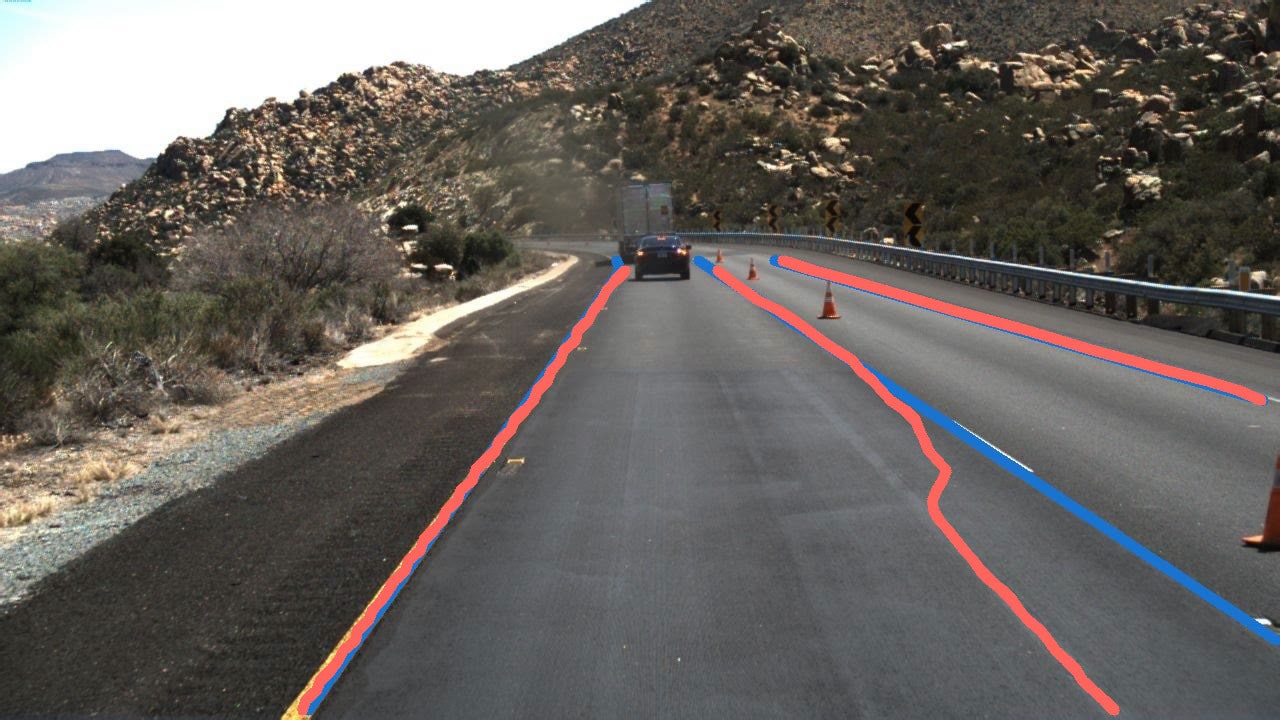} &
			\includegraphics[width=0.18\linewidth,valign=m]{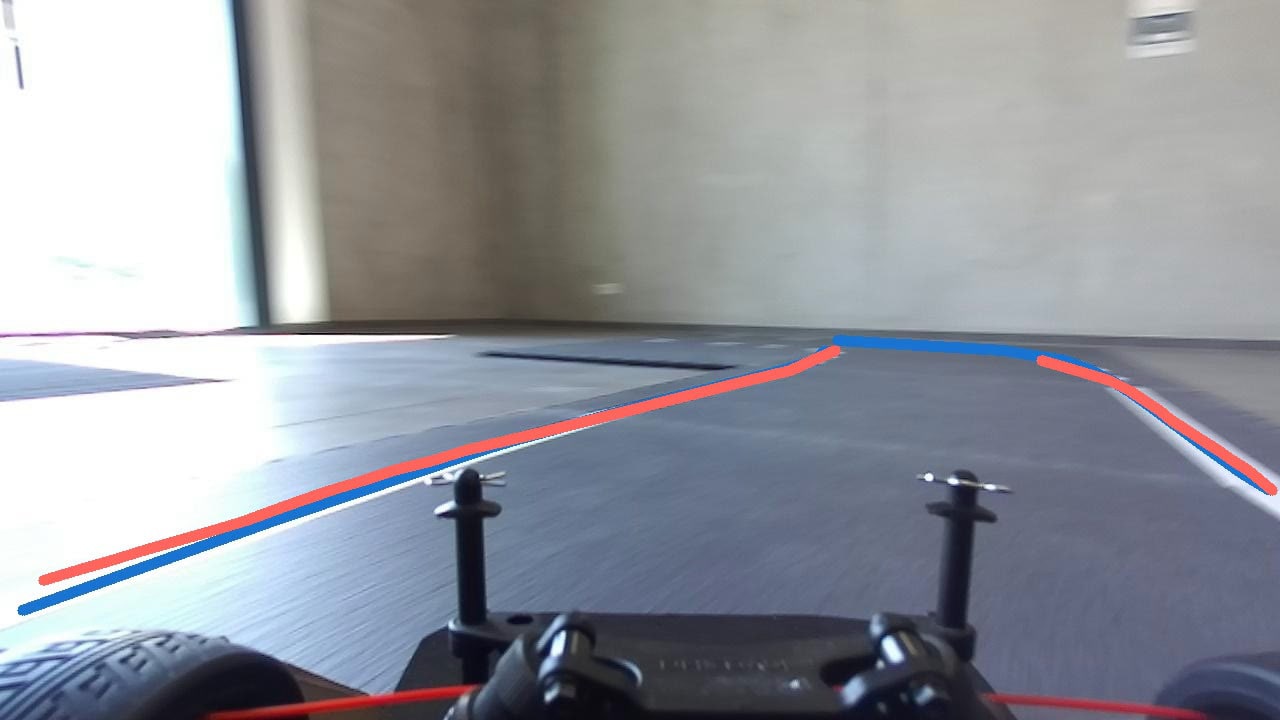} & \includegraphics[width=0.18\linewidth,valign=m]{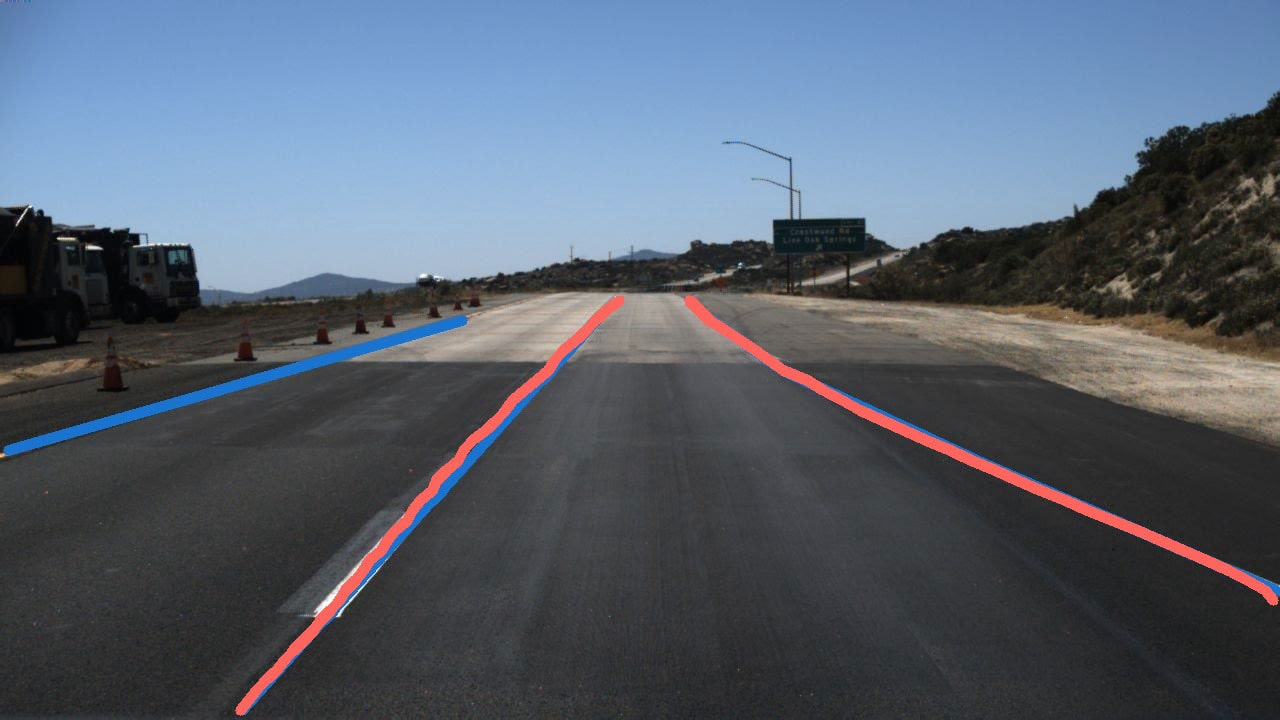}\\
			SGADA & 
			\includegraphics[width=0.18\linewidth,valign=m]{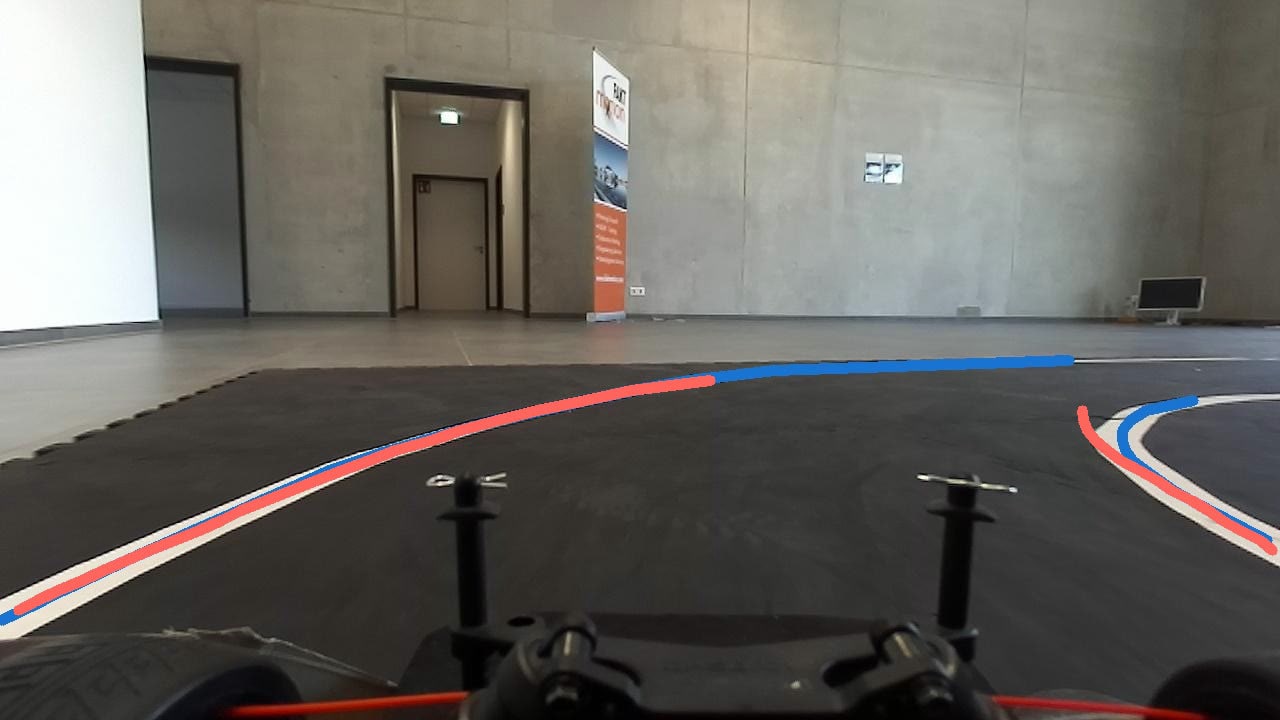} & 
			\includegraphics[width=0.18\linewidth,valign=m]{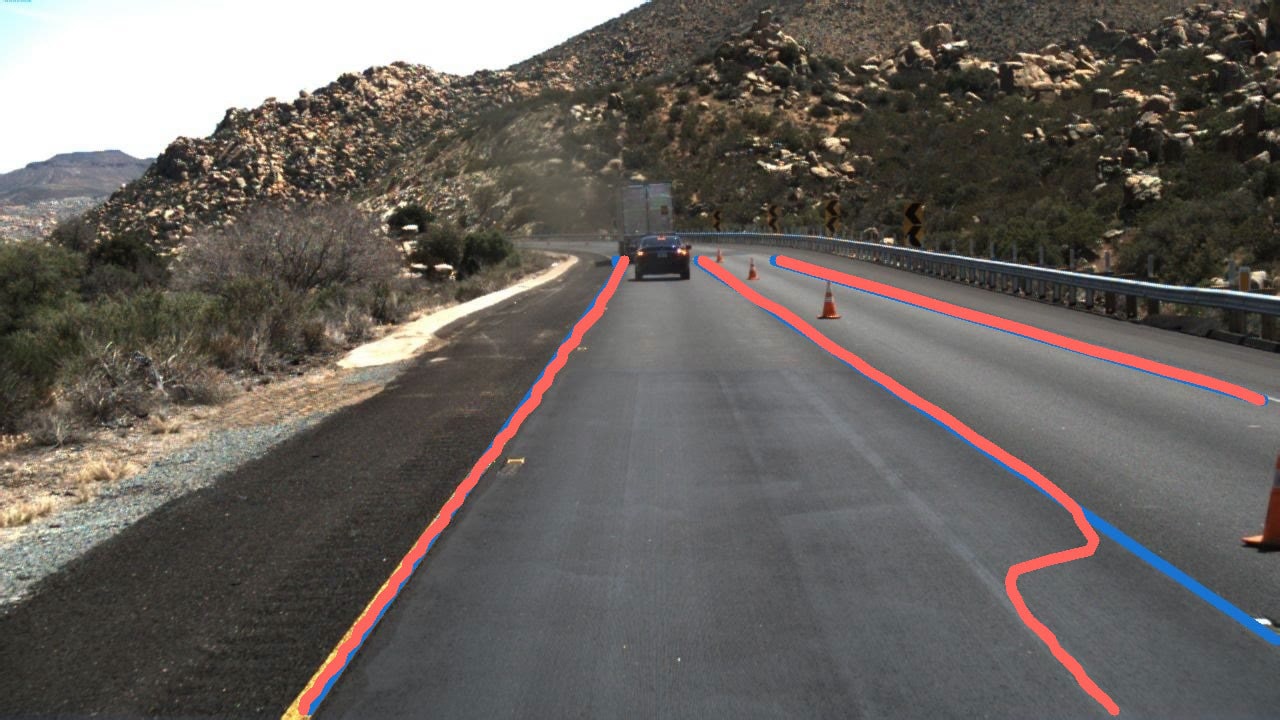} &
			\includegraphics[width=0.18\linewidth,valign=m]{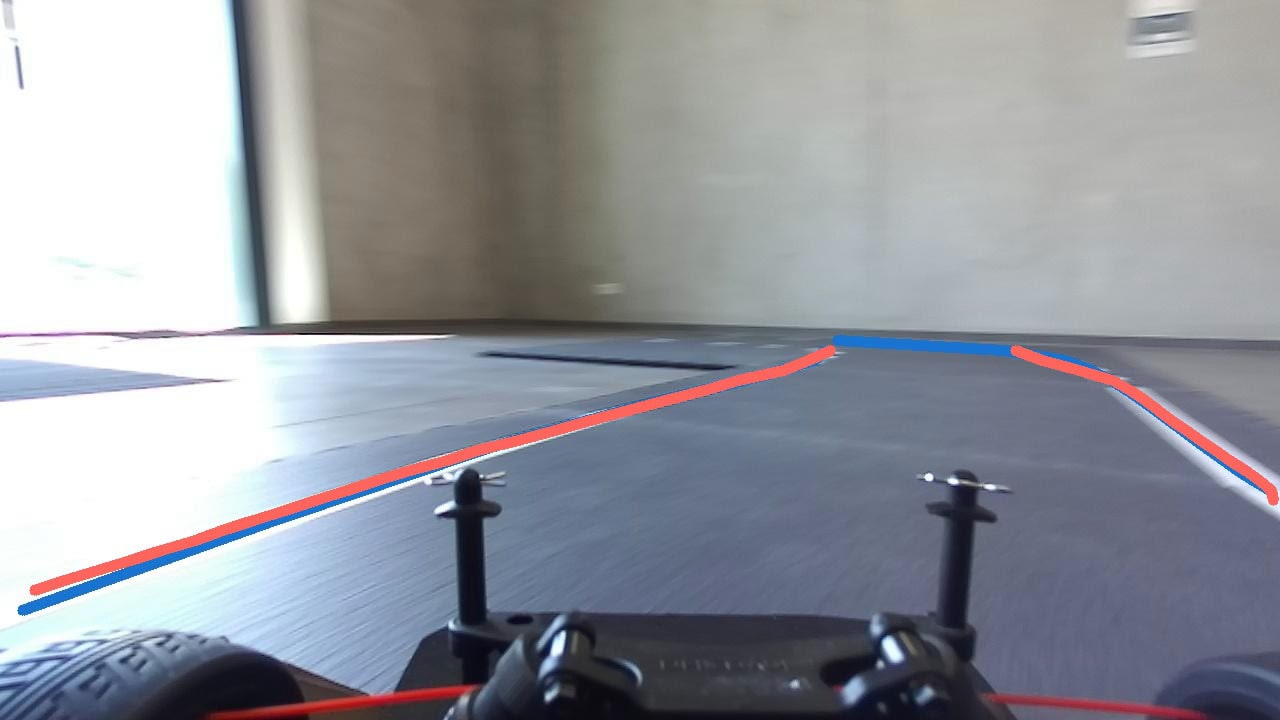} & \includegraphics[width=0.18\linewidth,valign=m]{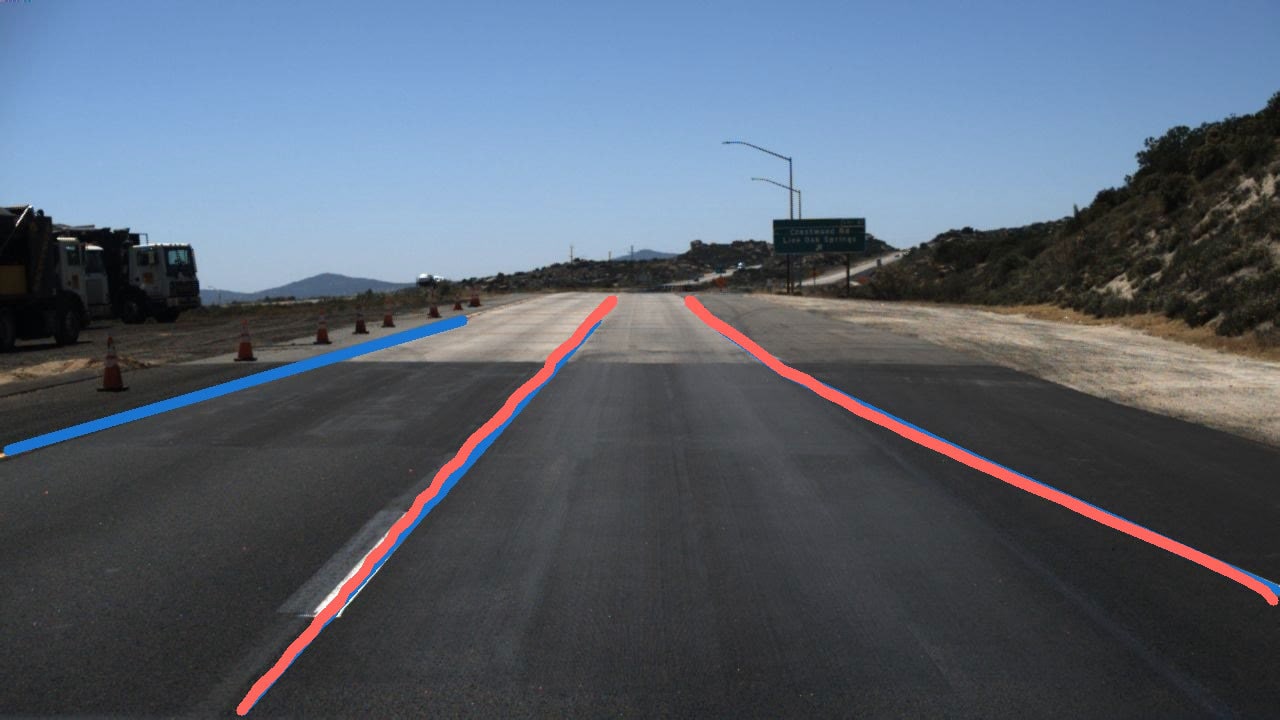}\\
			SGPCS & 
			\includegraphics[width=0.18\linewidth,valign=m]{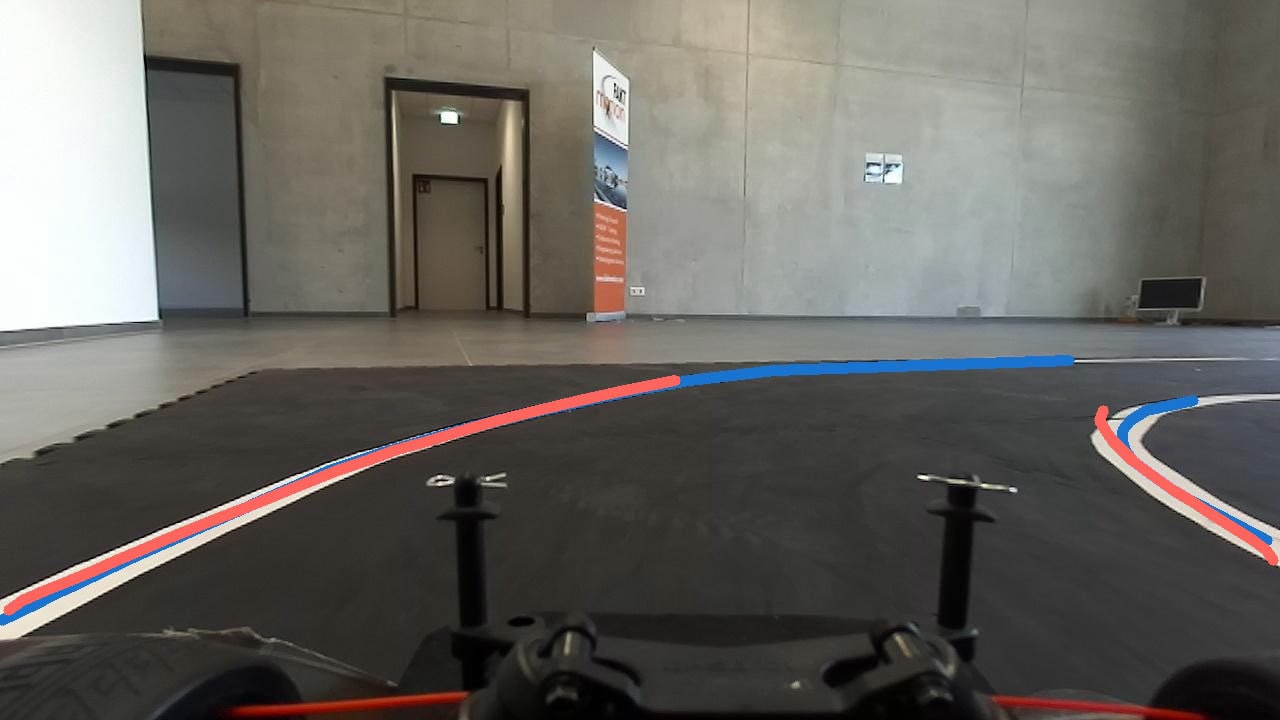} & \includegraphics[width=0.18\linewidth,valign=m]{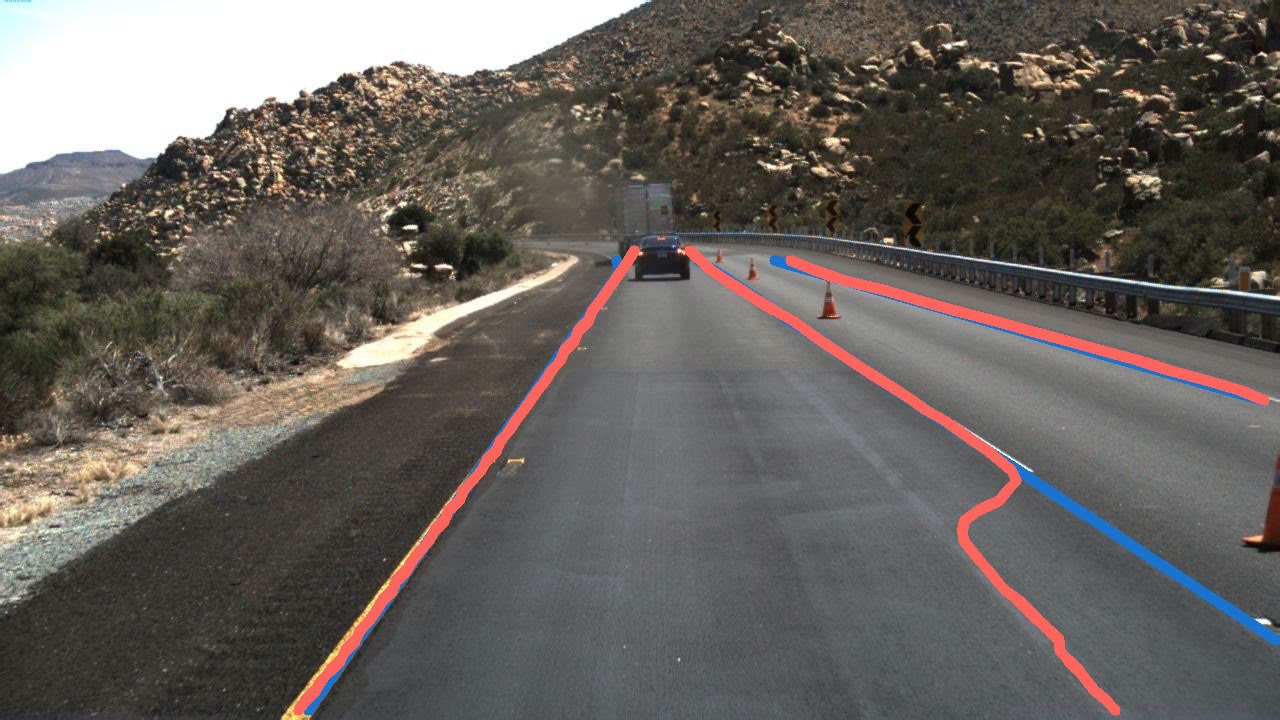} &
			\includegraphics[width=0.18\linewidth,valign=m]{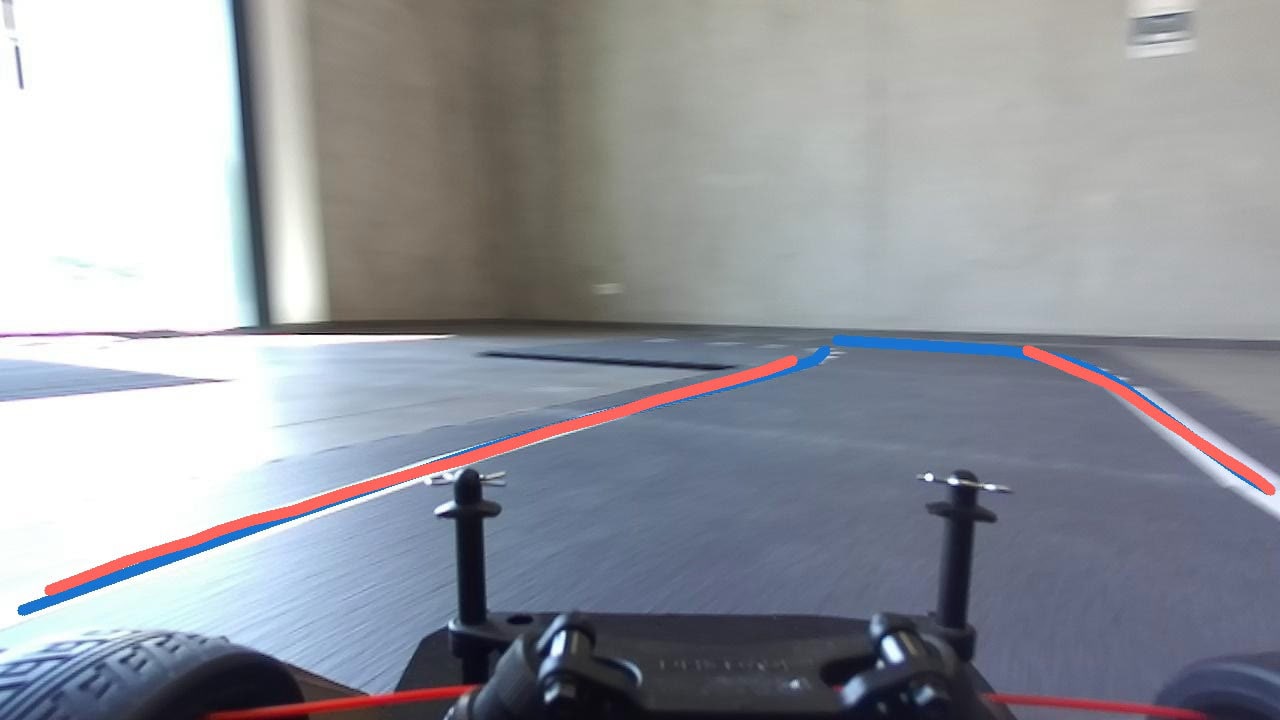} & \includegraphics[width=0.18\linewidth,valign=m]{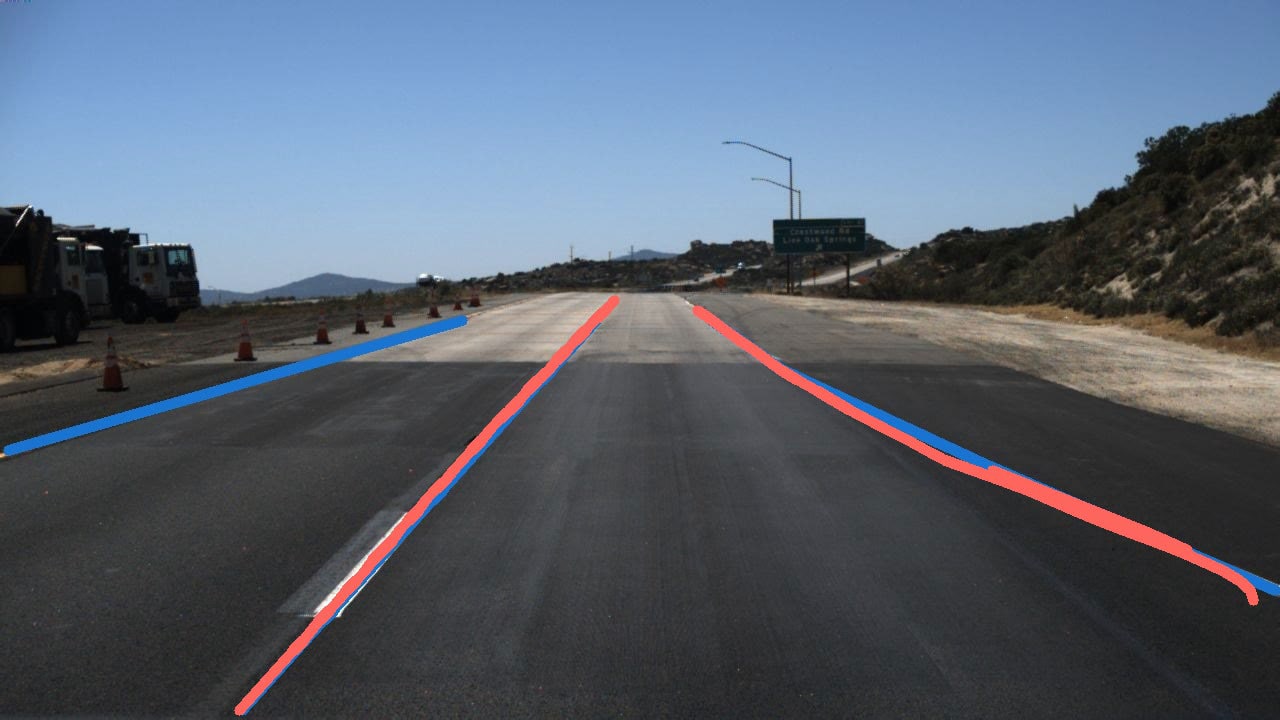}\\
			UFLD-TO & 
			\includegraphics[width=0.18\linewidth,valign=m]{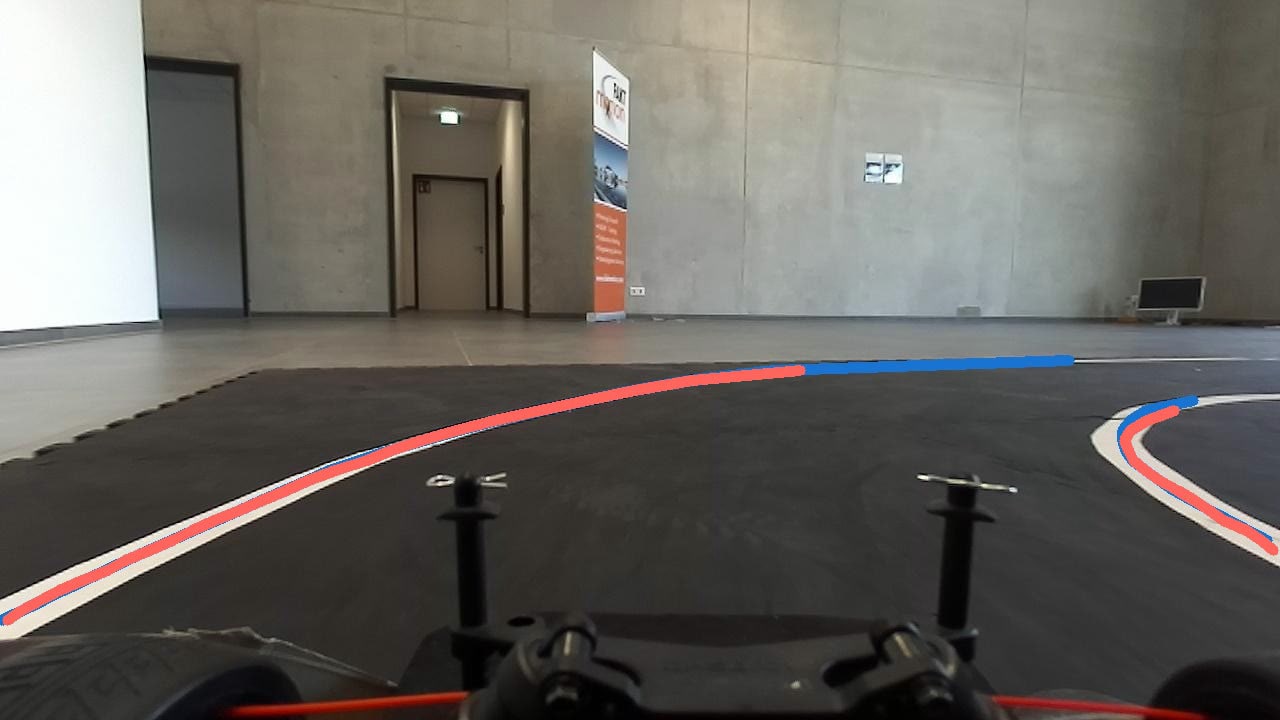} & \includegraphics[width=0.18\linewidth,valign=m]{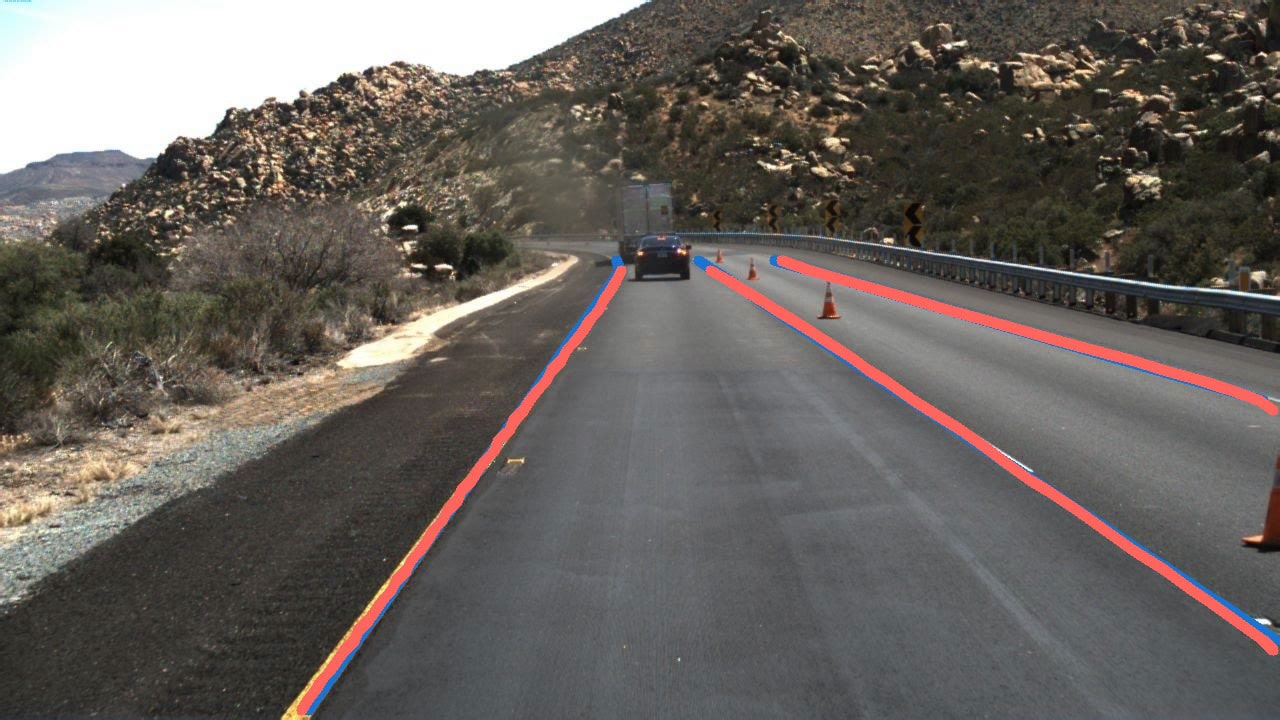} &
			\includegraphics[width=0.18\linewidth,valign=m]{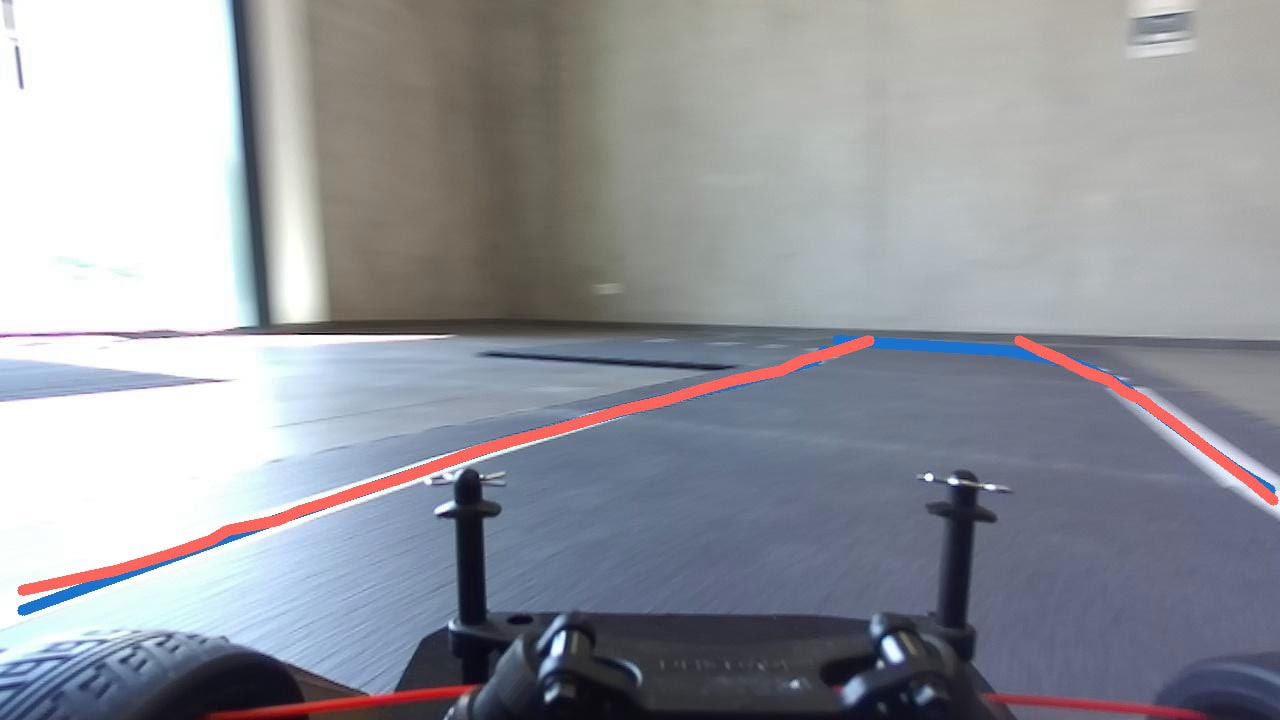} & \includegraphics[width=0.18\linewidth,valign=m]{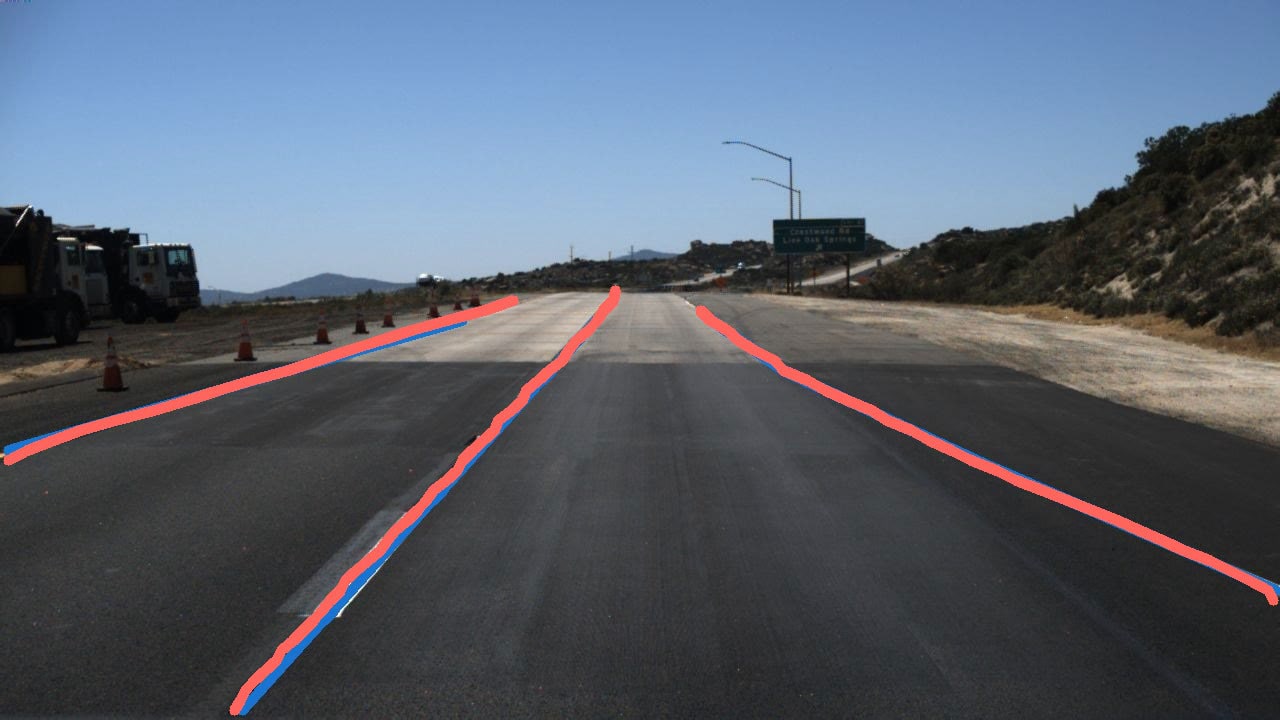}\\
		\end{tabular}
	\end{center}
	\vspace{-1ex}
	\caption[Qualitative results of target domain predictions.]{Qualitative results of target domain predictions. Ground truth lane annotations are marked in blue, and predictions in red. Best viewed in color.}
	\label{fig:carlane:qualitative_resuls}
\end{figure}

We want to emphasize that with an accuracy gain of a maximum of 5.14\% (SGPCS) and high false positive and false negative rates, the domain adaptation methods are not able to achieve comparable results to the supervised baselines (\acs{UFLD}-TO). Furthermore, we observe that false positive and false negative rates increase significantly on MuLane, indicating that the multi-target dataset forms the most challenging task. False positives and false negatives represent wrongly detected and missing lanes which can lead to crucial impacts on autonomous driving functions. These results affirm the need for the proposed CARLANE Benchmark to further strengthen the research in unsupervised domain adaptation for lane detection.
\begin{table}[t]
	\RawFloats
	\scriptsize
	\caption[Performance on the test set.]{Performance on the test set. Lane accuracy (\acs{LA}), false positives (\acs{FP}), and false negatives (\acs{FN}) are reported in \%.} 
	\vspace{-1ex}
	\label{tab:carlane:quantitativ_results_comparison}
	\begin{center}
		\scalebox{0.878}{%
			\setlength{\tabcolsep}{0.4em}
			\begin{tabular}{c|ccc|ccc|ccc}
				\toprule
				\multirow{2}{*}{ResNet-18} & \multicolumn{3}{c|}{MoLane}       & \multicolumn{3}{c|}{TuLane}                       & \multicolumn{3}{c}{MuLane} \\
				& LA           & FP & FN         & LA            & FP    &  FN                      & LA       & FP    &  FN   \\
				\midrule
				UFLD-SO                   & 88.15          & 34.35  &  28.45        & 87.43           & 34.21 & 23.48                    & 79.61      & 44.78 & 33.36 \\
				DANN \cite{ganin2016domain}     & 85.25$\pm$0.49 & 39.07$\pm$1.21  & 36.18 $\pm$1.50 & 88.74$\pm$0.32 & 32.71$\pm$0.52  & 21.64$\pm$0.65  & 78.25$\pm$0.62 & 48.67$\pm$1.17 & 41.69$\pm$1.80  \\
				ADDA \cite{tzeng2017adversarial} & 92.10$\pm$0.21 & 19.71$\pm$0.77  & 11.46$\pm$0.92 & 90.72$\pm$0.15 & 29.73$\pm$0.36  & 17.67$\pm$0.42  & 81.25$\pm$0.33 & 41.69$\pm$0.40 & 28.58$\pm$0.57   \\
				SGADA \cite{akkaya2021self}    & 93.15$\pm$0.12 & 16.23$\pm$0.22 & 7.68$\pm$0.26 & \textbf{91.70}$\pm$0.13 & \textbf{28.42}$\pm$0.34 & \textbf{16.10}$\pm$0.43 & 82.05$\pm$0.10 & \textbf{40.37}$\pm$0.33 & \textbf{25.95}$\pm$0.36\\
				SGPCS (ours)                & \textbf{93.29}$\pm$0.07 & \textbf{16.22}$\pm$0.17 & \textbf{7.29}$\pm$0.13  & 91.55$\pm$0.13 & 28.52$\pm$0.21 & 16.16$\pm$0.26 & \textbf{82.91}$\pm$0.19 & 40.76$\pm$0.57 & 26.14$\pm$0.75\\
				\midrule
				UFLD-TO                   & 97.15           & 0.96  &    0.05       &  94.97          & 18.05           & 3.84          & 87.64  & 29.48 & 11.52 \\ 
				
				\midrule
				\midrule
				
				ResNet-34                & LA            & FP & FN        & LA            & FP     & FN                       & LA        & FP    & FN    \\ 
				\midrule
				UFLD-SO                  & 88.76          & 31.30     &   26.70    & 89.42           & 32.35  & 21.19                    & 80.70      & 43.63 & 31.40 \\
				DANN \cite{ganin2016domain}    & 89.58$\pm$0.63  & 28.75$\pm$1.53 & 23.24$\pm$2.16  & 91.06$\pm$0.14 & 30.17$\pm$0.20 & 18.54$\pm$0.25    & 80.40$\pm$0.15 & 43.52$\pm$0.37 & 31.53$\pm$0.66\\
				ADDA \cite{tzeng2017adversarial}& 91.76$\pm$0.29  & 21.04$\pm$0.81 & 13.06$\pm$1.02  & 91.39$\pm$0.16 & 28.76$\pm$0.30 & 16.63$\pm$0.36    & 81.64$\pm$0.34 & 40.74$\pm$0.48 & 27.50$\pm$0.78\\
				SGADA \cite{akkaya2021self}   & 92.59$\pm$0.18  & 18.32$\pm$0.24& 9.88$\pm$0.29   & 92.04$\pm$0.09 & 28.18$\pm$0.20 & 15.99$\pm$0.24    & \textbf{82.93}$\pm$0.03 & \textbf{39.45}$\pm$0.11 & \textbf{24.98}$\pm$0.13\\
				SGPCS (ours)               & \textbf{92.82}$\pm$0.27 & \textbf{17.10}$\pm$0.73 & \textbf{8.77}$\pm$0.95 & \textbf{93.29}$\pm$0.18 & \textbf{25.68}$\pm$0.48 & \textbf{12.73}$\pm$0.59 & 82.87$\pm$0.17 & 40.13$\pm$0.28 & 25.38$\pm$0.45\\
				\midrule
				UFLD-TO                  & 96.92           & 0.94   &   0.03     & 94.43          & 20.74      & 7.20  & 87.62  &  29.19  & 11.08 \\
				\bottomrule	
		\end{tabular}}
	\end{center}
	\vspace{-2ex}
\end{table}
\begin{figure}[t]
	\small
	\begin{center}
		\begin{tabular}{c@{}c@{}c@{}c@{}c}
			UFLD-SO & DANN & ADDA & SGADA & SGPCS \\
			\includegraphics[width=0.2\linewidth,valign=m]{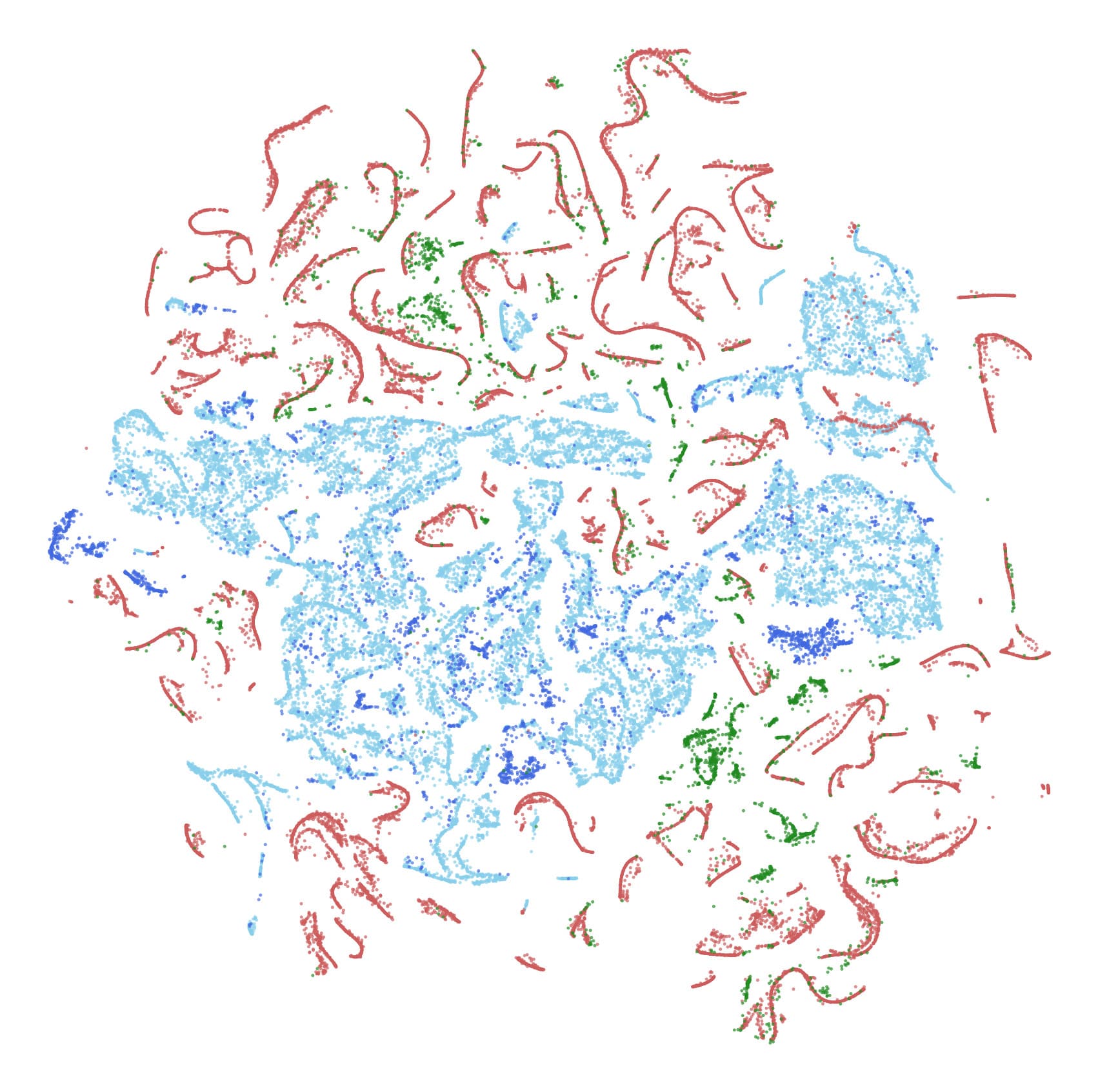} & \includegraphics[width=0.2\linewidth,valign=m]{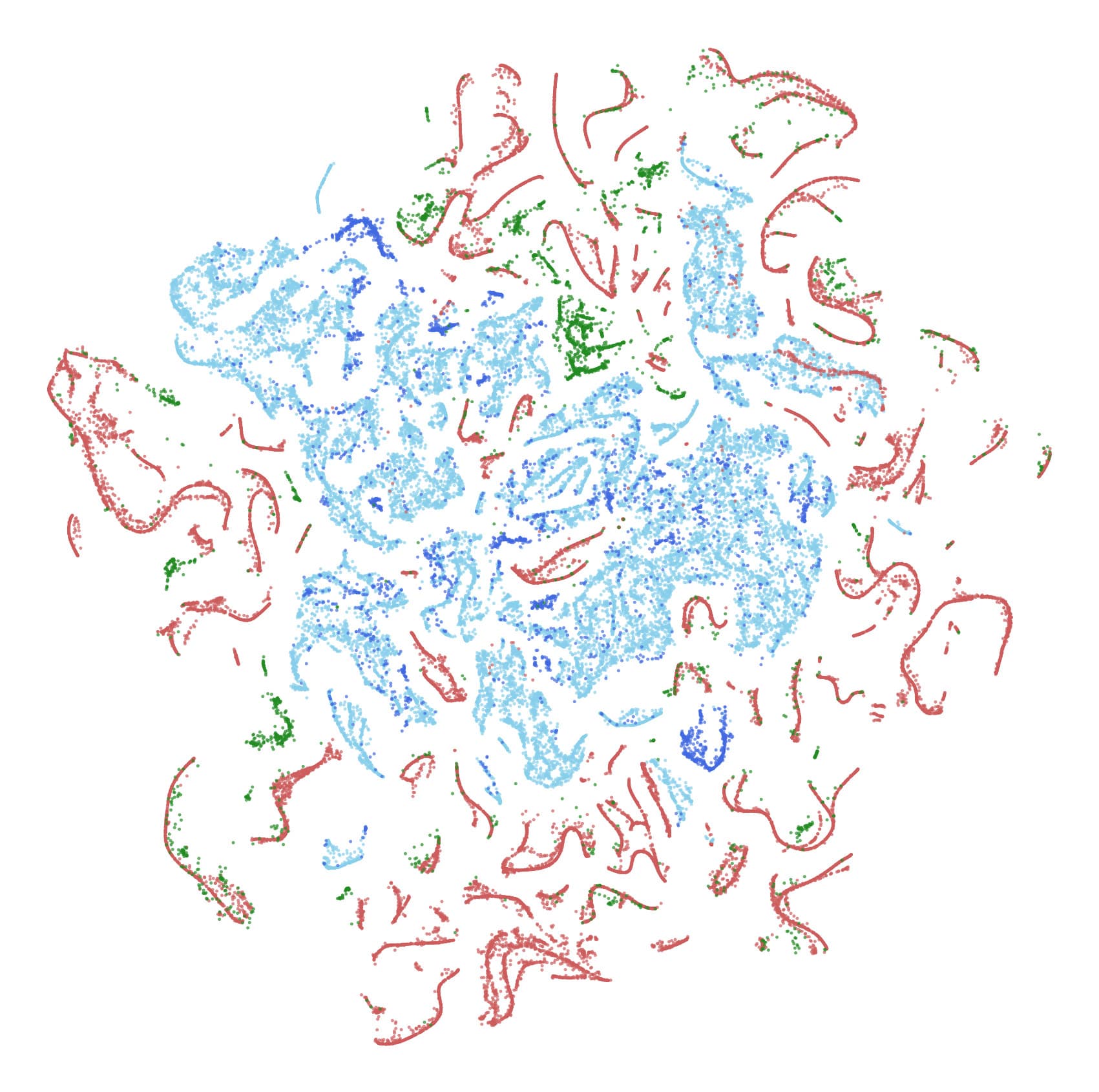} &
			\includegraphics[width=0.2\linewidth,valign=m]{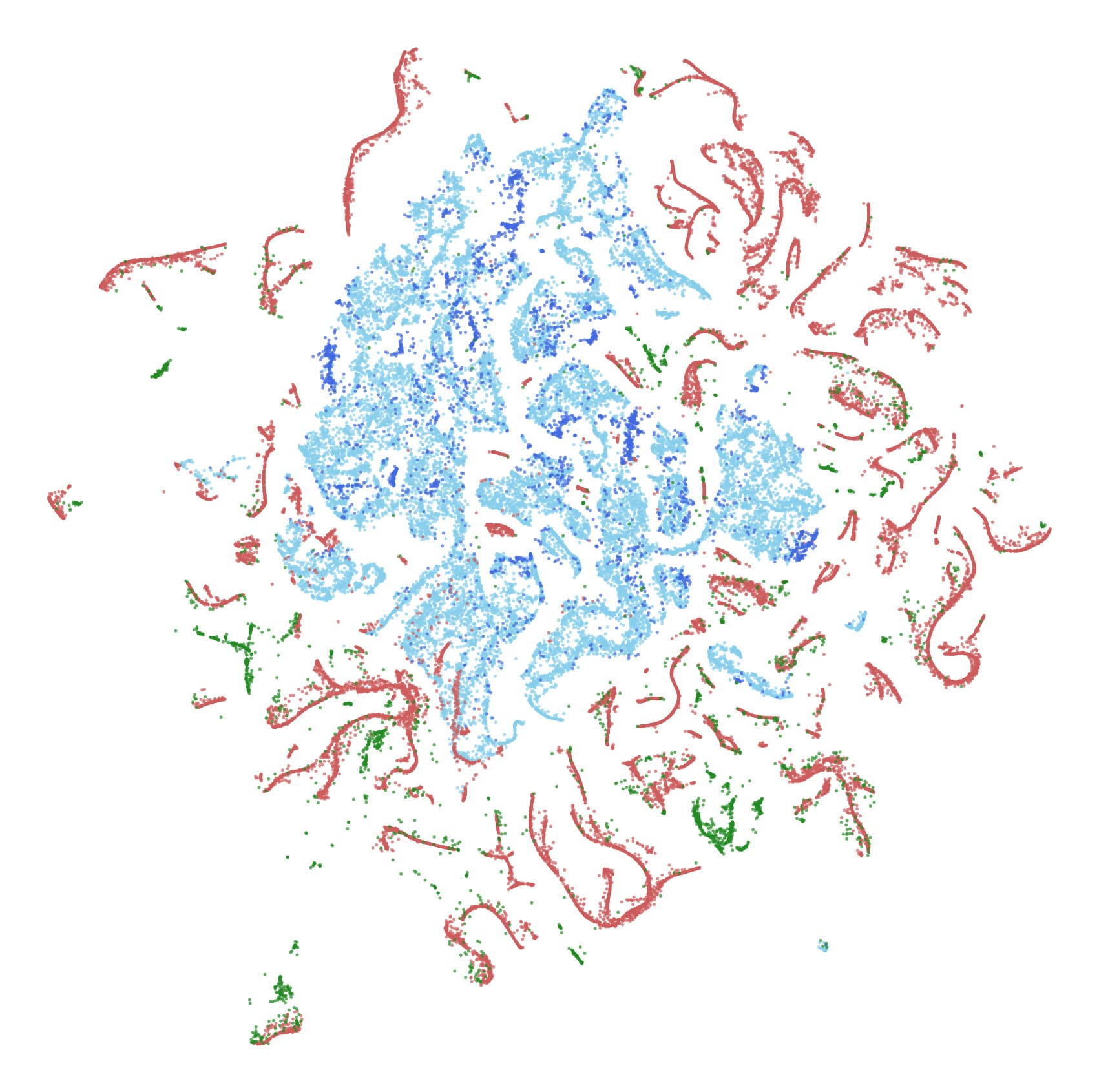} &
			\includegraphics[width=0.2\linewidth,valign=m]{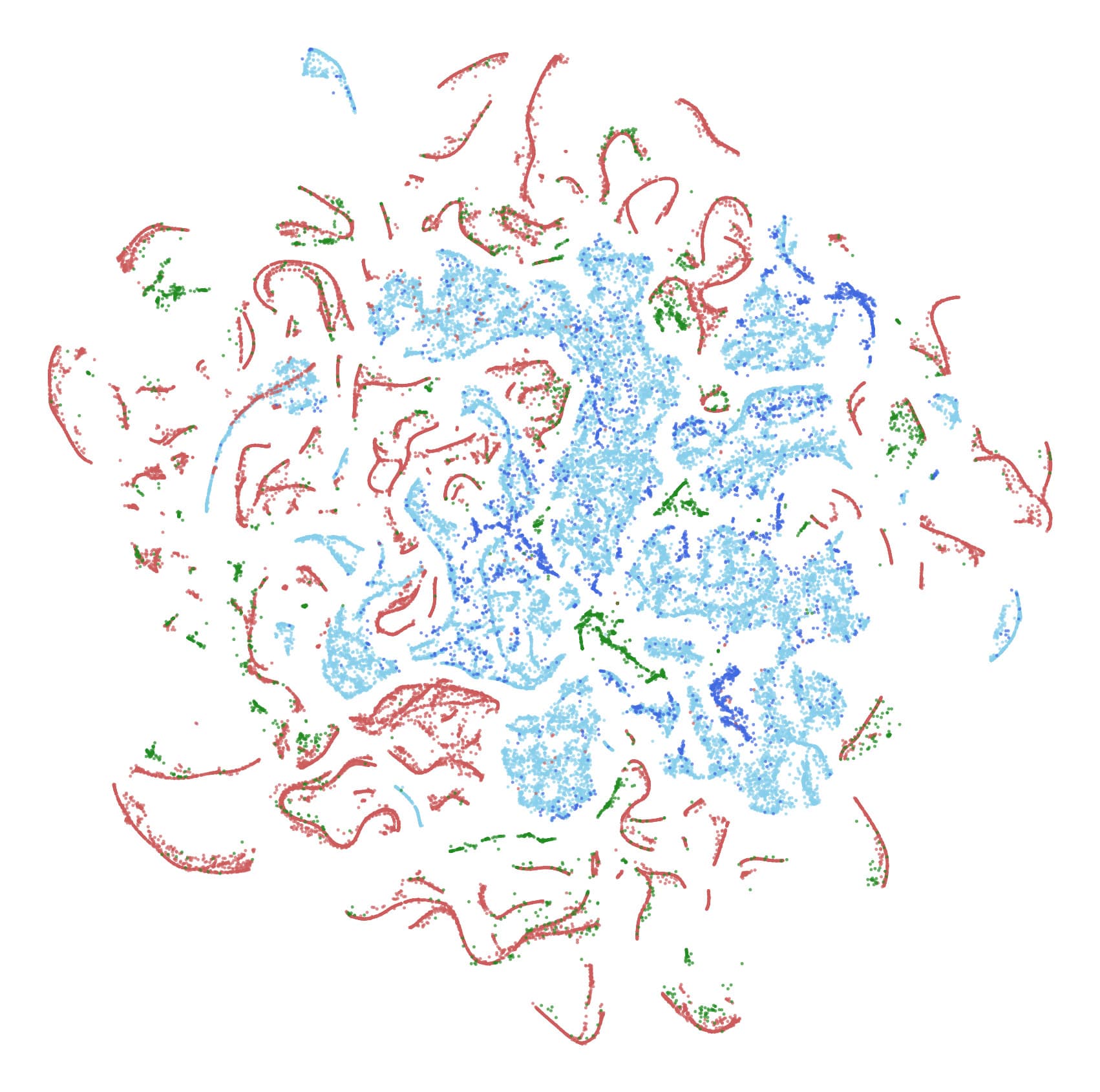} & \includegraphics[width=0.2\linewidth,valign=m]{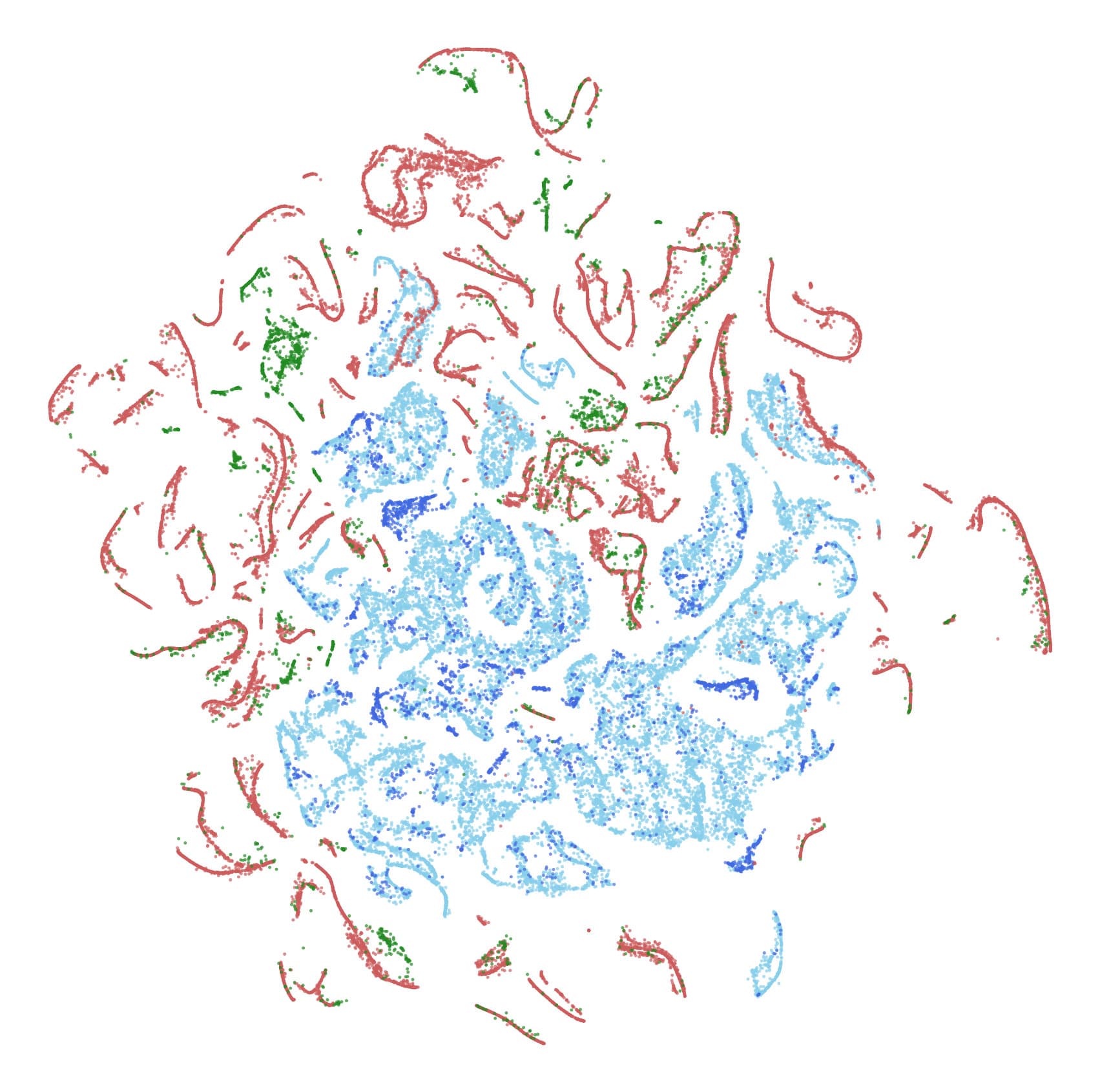}\\
		\end{tabular}
	\end{center}
	\vspace{-1ex}
	\caption[\acs{t-SNE} visualization of MuLane dataset.]{\acs{t-SNE} visualization of MuLane dataset. The source domain is marked in blue, the real-world model vehicle target domain in red, and the TuSimple domain in green. Best viewed in color.}
	\label{fig:carlane:TSNE_plot_mulane}
\end{figure}
\\\\
\textbf{Qualitative Evaluation.}
We use \acs{t-SNE} \cite{van2008visualizing} to visualize the features of the features encoders for the source and target domains of MuLane in \autoref{fig:carlane:TSNE_plot_mulane}. \acs{t-SNE} visualizations of MoLane and TuLane can be found in \autoref{app:02}. In accordance with the quantitative results, we observe only a slight adaptation of the source and target domain features for ADDA, SGADA, and SGPCS compared to the supervised baseline \acs{UFLD}-SO. Consequently, the examined well-known domain adaptation methods have no significant effect on feature alignment. In addition, we show results from the evaluated methods in \autoref{fig:carlane:qualitative_resuls} and observe that the models are able to predict target domain lane annotations in many cases but are not able to achieve comparable results to the supervised baseline (\acs{UFLD}-TO).
\\\\
In summary, we find quantitatively and qualitatively that the examined domain adaptation methods do not significantly improve the performance of lane detection and feature adaptation. For this reason, we believe that the proposed benchmark could facilitate the exploration of new domain adaptation methods to overcome these problems. 

\section{Conclusion}
\label{sec:carlane:conclusion}
We present CARLANE, the first unsupervised domain adaptation benchmark for lane detection. CARLANE was recorded in three domains and consists of three datasets: the single-target datasets MoLane and TuLane and the multi-target dataset MuLane, which is a balanced combination of both. Based on the \acs{UFLD} model, we have conducted experiments with different unsupervised domain adaptation methods on CARLANE and found that the selected methods are able to adapt the model to target domains slightly and consistently. However, none of the methods achieve comparable results to the supervised baselines. The most significant performance differences are noticeable in the high false positive and false negative rates of the unsupervised domain adaptation methods compared to the target-only baselines, which is even more pronounced in the MuLane multi-target task. These false-positive and false-negative rates can negatively impact autonomous driving functions since they represent misidentified and missing lanes. Furthermore, as shown in the t-SNE plots of \autoref{fig:carlane:TSNE_plot_mulane}, the examined well-known domain adaptation methods have no significant effect on feature alignment. The current difficulties of the examined unsupervised domain adaptation methods to align the source and target domains adequately confirm the need for the proposed CARLANE benchmark. We believe that CARLANE eases the development and comparison of unsupervised domain adaptation methods for lane detection. In addition, we open-source all tools for dataset creation and labeling and hope that CARLANE facilitates future research in these directions. 
\\\\
\textbf{Limitations.} One limitation of our work is that we only use a fixed set of track elements within our 1/8th scaled environment. These track elements represent only a limited number of distinct curve radii. Furthermore, neither buildings nor traffic signs exist in MoLane's model vehicle target domain.
Moreover, the full-scale real-world target domain of TuLane is derived from TuSimple. TuSimple's data was predominantly collected under good and medium conditions and lacks variation in weather and time of day. In addition, we want to emphasize that collecting data for autonomous driving is still an ongoing effort and that datasets such as TuSimple do not cover all possible real-world driving scenarios to ensure safe, practical use. For the synthetically generated data, we limited ourselves to using existing CARLA maps without defining new simulation environments. Despite these limitations, CARLANE serves as a supportive dataset for further research in the field of unsupervised domain adaptation.
\\\\
\textbf{Ethical and Responsible Use.} Considering the limitations of our work, unsupervised domain adaptation methods trained on TuLane and MuLane should be tested with care and under the right conditions on a full-scale car. However, real-world testing with MoLane in the model vehicle domain can be carried out in a safe and controlled environment. Additionally, TuLane contains open-source images with unblurred license plates and people. This data should be treated with respect and in accordance with privacy policies. In general, our work contributes to the research in the field of autonomous driving, in which a lot of unresolved ethical and legal questions are still being discussed. The step-by-step testing possibility across three domains makes it possible for our benchmark to include an additional safety mechanism for real-world testing.
\chapter{Content-Consistent Translation with Masked Discriminators}
\label{chap:03}
\vspace{-8mm}
\vspace{-8ex}
\begin{figure}[H] 
	\captionsetup[subfigure]{labelformat=empty}
	\begin{center}
		\subfloat[PFD$\rightarrow$Cityscapes]
		{\includegraphics[width=0.249\textwidth]{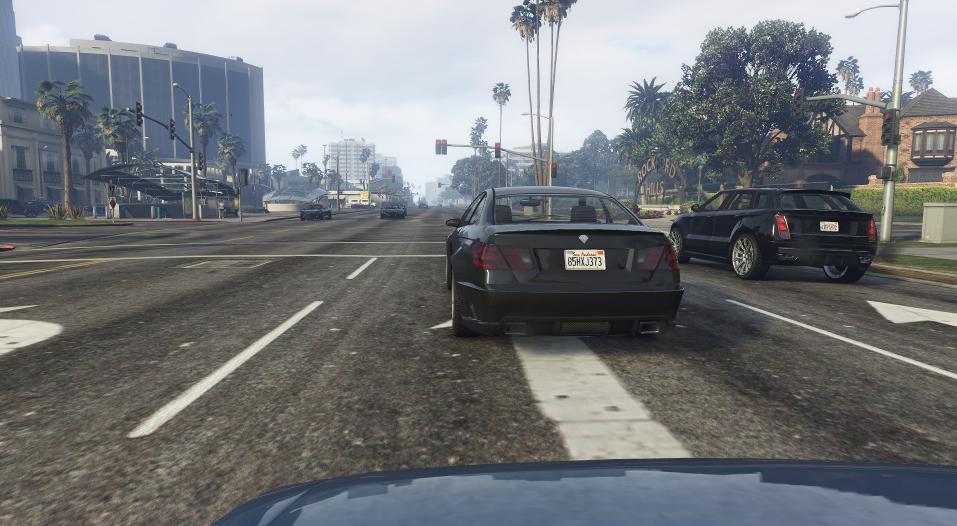}{\includegraphics[width=0.249\textwidth]{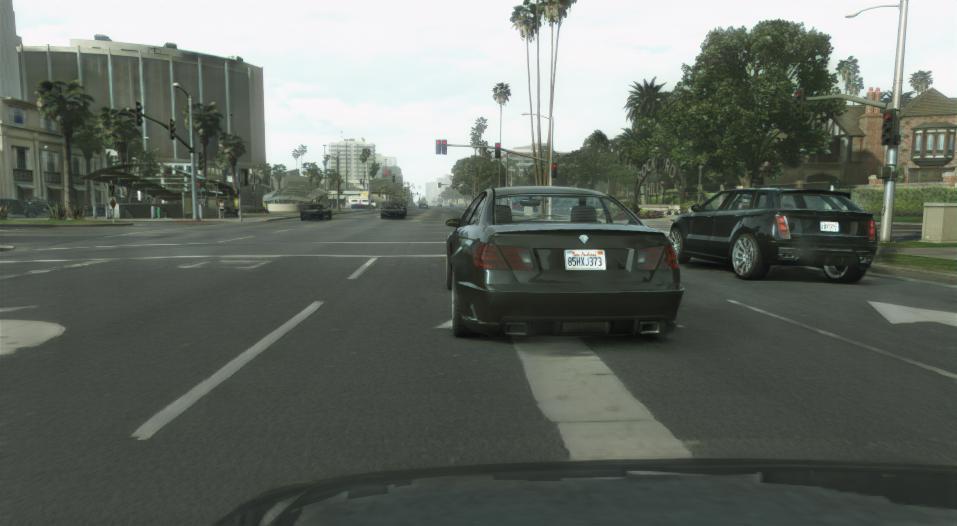}}}\hfill 
		\subfloat[Day$\rightarrow$Night]
		{\includegraphics[width=0.249\textwidth]{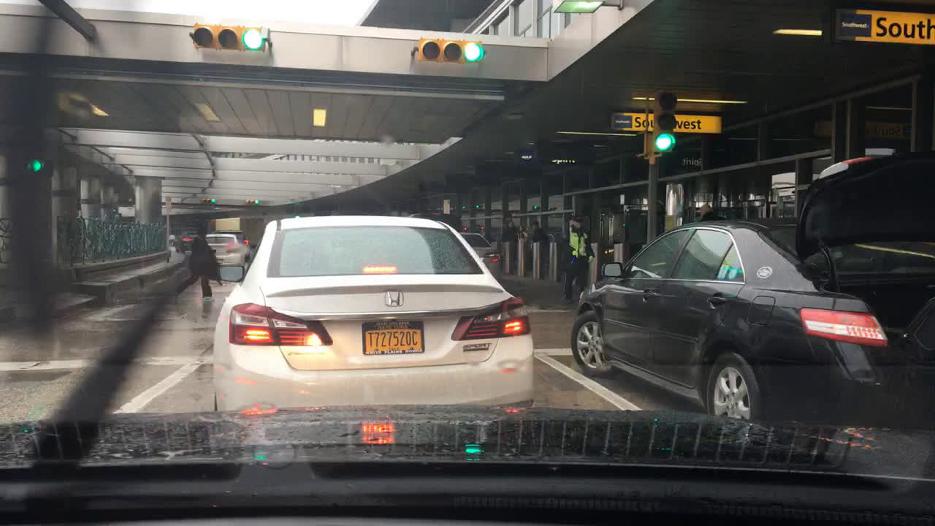}{\includegraphics[width=0.249\textwidth]{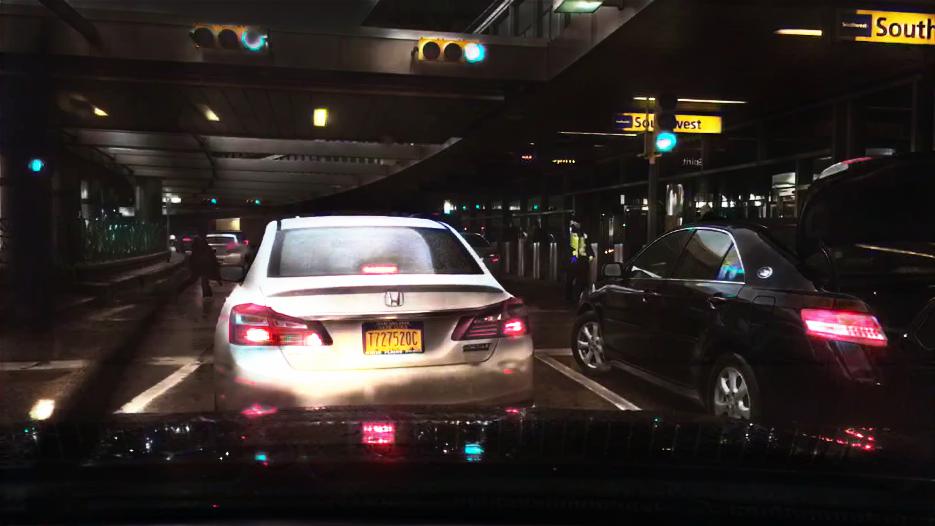}}}\hfill\\
		\vspace{-6pt}
		\subfloat[Viper$\rightarrow$Cityscapes]
		{\includegraphics[width=0.249\textwidth]{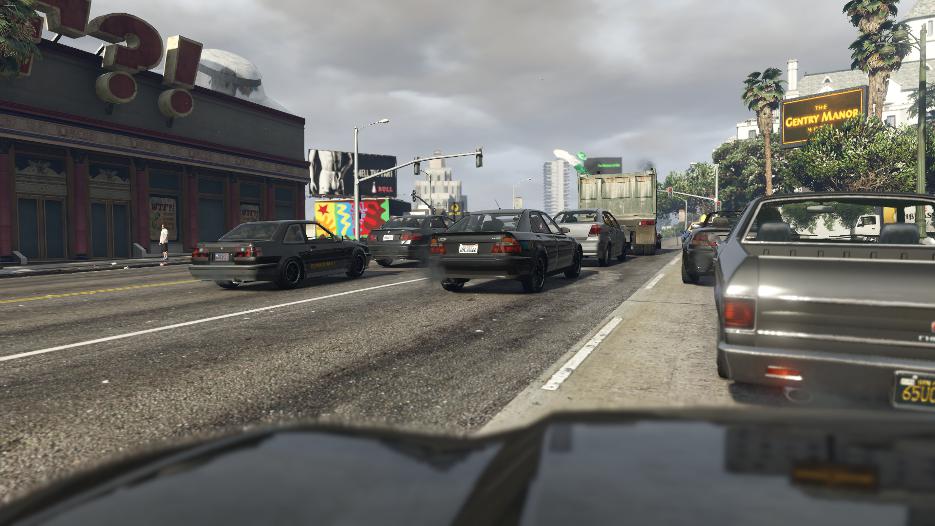}{\includegraphics[width=0.249\textwidth]{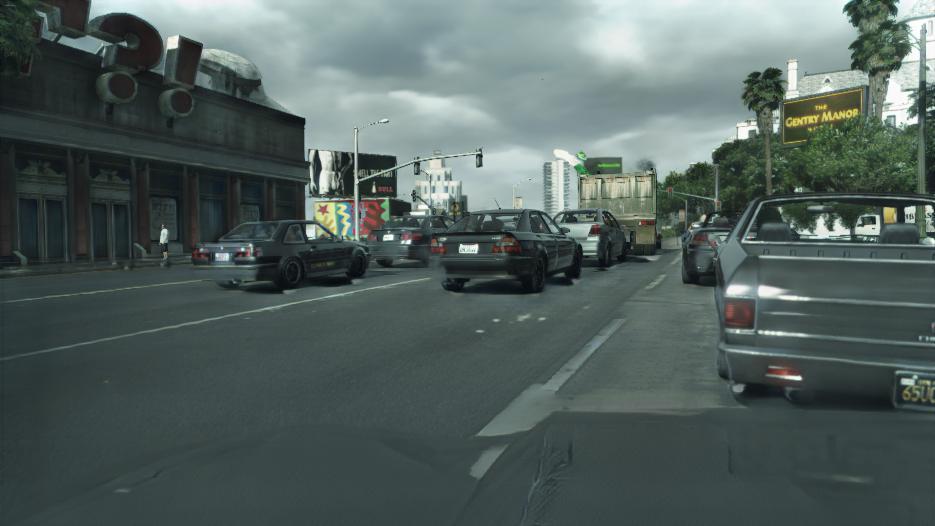}}}\hfill	 
		\subfloat[Clear$\rightarrow$Snowy]
		{\includegraphics[width=0.249\textwidth]{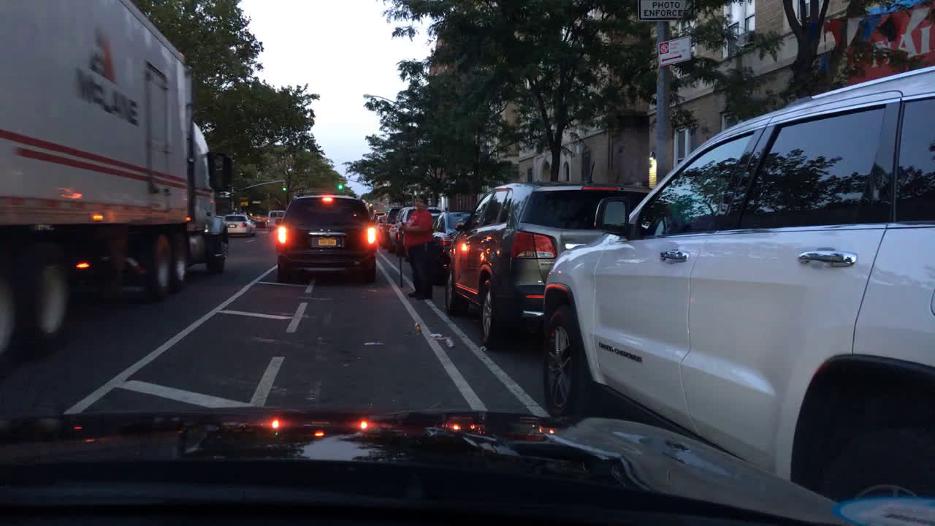}{\includegraphics[width=0.249\textwidth]{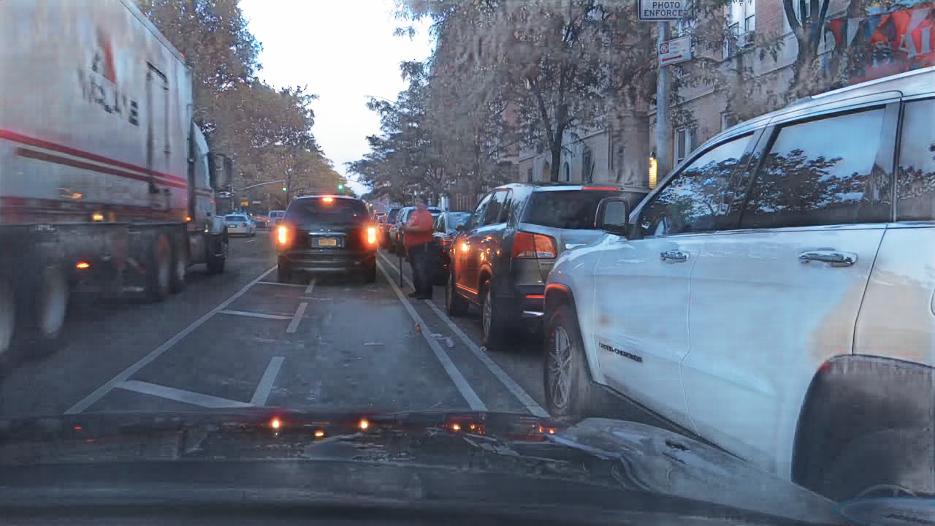}}}\hfill
	\end{center}
	\vspace{-1ex}
	\caption[Results of our method.]{Results of our method. Best viewed in color.}
	\label{fig:feamgan:results}
\end{figure}
\begin{abstract}
A common goal of unpaired image-to-image translation is to preserve content consistency between source images and translated images while mimicking the style of the target domain. Due to biases between the datasets of both domains, many methods suffer from inconsistencies caused by the translation process. Most approaches introduced to mitigate these inconsistencies do not constrain the discriminator, leading to an even more ill-posed training setup. Moreover, none of these approaches is designed for larger crop sizes. In this work, we show that masking the inputs of a global discriminator for both domains with a content-based mask is sufficient to reduce content inconsistencies significantly. However, this strategy leads to artifacts that can be traced back to the masking process. To reduce these artifacts, we introduce a local discriminator that operates on pairs of small crops selected with a similarity sampling strategy. Furthermore, we apply this sampling strategy to sample global input crops from the source and target dataset. In addition, we propose feature-attentive denormalization to selectively incorporate content-based statistics into the generator stream. In our experiments, we show that our method achieves state-of-the-art performance in photorealistic sim-to-real translation and weather translation and also performs well in day-to-night translation. Additionally, we propose the cKVD metric, which builds on the sKVD metric and enables the examination of translation quality at the class or category level.
\end{abstract}
\section{Motivation}
Unpaired image-to-image translation aims at transferring images from a source domain to a target domain when no paired examples are given. Recently, this field has attracted increasing interest and has advanced several use cases, such as photorealism \cite{pizzati2021comogan, richter2022enhancing, jia2021semantically}, neural rendering \cite{hao2021gancraft}, domain adaptation \cite{hoffman2018cycada, roy2021trigan}, the translation of seasons or daytime \cite{jiang2020tsit, jeong2021memory, pizzati2021comogan}, and artistic style transfer \cite{huang2017arbitrary, yao2019attention, kim2020u}. Current work has primarily focused on improving translation quality \cite{nederhood2021harnessing,jeong2021memory}, efficiency \cite{liang2021high, shaham2021spatially}, multi-modality \cite{huang2018multimodal,lin2020multimodal}, and content consistency \cite{richter2022enhancing, jia2021semantically}. Due to the ill-posed nature of the unpaired image-to-image translation task and biases between datasets, content consistency is difficult to achieve. To mitigate content inconsistencies, several methods have been proposed that constrain the generator of GANs \cite{zhu2017unpaired,benaim2017one,fu2019geometry,lin2020multimodal,zhang2019harmonic,zhao2020unpaired,liu2017unsupervised,huang2018multimodal,sendik2020crossnet,yang2020phase}. However, only constraining the generator leads to an unfair setup, as biases in the datasets can be detected by the discriminator: The generator tries to achieve content consistency by avoiding biases in the output, while the discriminator is still able to detect biases between both datasets and, therefore, forces the generator to include these biases in the output, for example, through hallucinations. Constraining the discriminator \cite{liang2018generative,richter2022enhancing, theiss2022unpaired} or improving the sampling of training pairs \cite{kao2019patch,richter2022enhancing} is currently underexplored, especially for content consistency on a global level, where the discriminator has a global view on larger image crops instead of a local view on small crops.
\section{Contributions}
In this work, we propose \textit{masked conditional discriminators}, which operate on masked global crops of the inputs to mitigate content inconsistencies. We combine these discriminators with an efficient sampling strategy based on a pre-trained robust segmentation model to sample similar global crops. Furthermore, we argue that when transferring feature statistics from the content stream of the source image to the generator stream, content-unrelated feature statistics from the content stream could affect image quality if the generator is unable to ignore this information since the output image should mimic the target domain. Therefore, we propose a \textit{feature-attentive denormalization (\acs{FATE})} block that extends feature-adaptive denormalization (\acs{FADE}) \cite{jiang2020tsit} with an attention mechanism. This block allows the generator to selectively incorporate statistical features from the content stream into the generator stream. In our experiments, we find that our method achieves state-of-the-art performance on most of the benchmarks shown in \autoref{fig:feamgan:results}.
\\\\
Our contributions can be summarized as follows:
\begin{itemize}
	\item We propose an efficient sampling strategy that utilizes robust semantic segmentations to sample similar global crops. This reduces biases between both datasets induced by semantic class misalignment. (\hyperref[RQ-T2]{RQ-T2}) 
	\item We combine this strategy with masked conditional discriminators to achieve content consistency while maintaining a more global field of view. (\hyperref[RQ-T2]{RQ-T2})
	\item We extend our method with an unmasked local discriminator. This discriminator operates on local, partially class-aligned patches to minimize the underrepresentation of frequently masked classes and associated artifacts. (\hyperref[RQ-T2]{RQ-T2})
	\item We propose a feature-attentive denormalization (\acs{FATE}) block, which selectively fuses statistical features from the content stream into the generator stream. (\hyperref[RQ-T3]{RQ-T3})
	\item We propose the class-specific Kernel VGG Distance (\acs{cKVD}) that builds upon the semantically aligned Kernel VGG Distance (\acs{sKVD}) \cite{richter2022enhancing} and uses robust segmentations to incorporate class-specific content inconsistencies in the perceptual image quality measurement. (\hyperref[RQ-E7]{RQ-E7})
	\item In our experiments, we show that our method achieves state-of-the-art performance on photo-realistic sim-to-real transfer and the translation of weather and performs well for daytime translation. (\hyperref[RQ-E8]{RQ-E8})
\end{itemize}
\section{Method}
\begin{figure}[t]
	\begin{center}
		\includegraphics[width=1.0\linewidth]{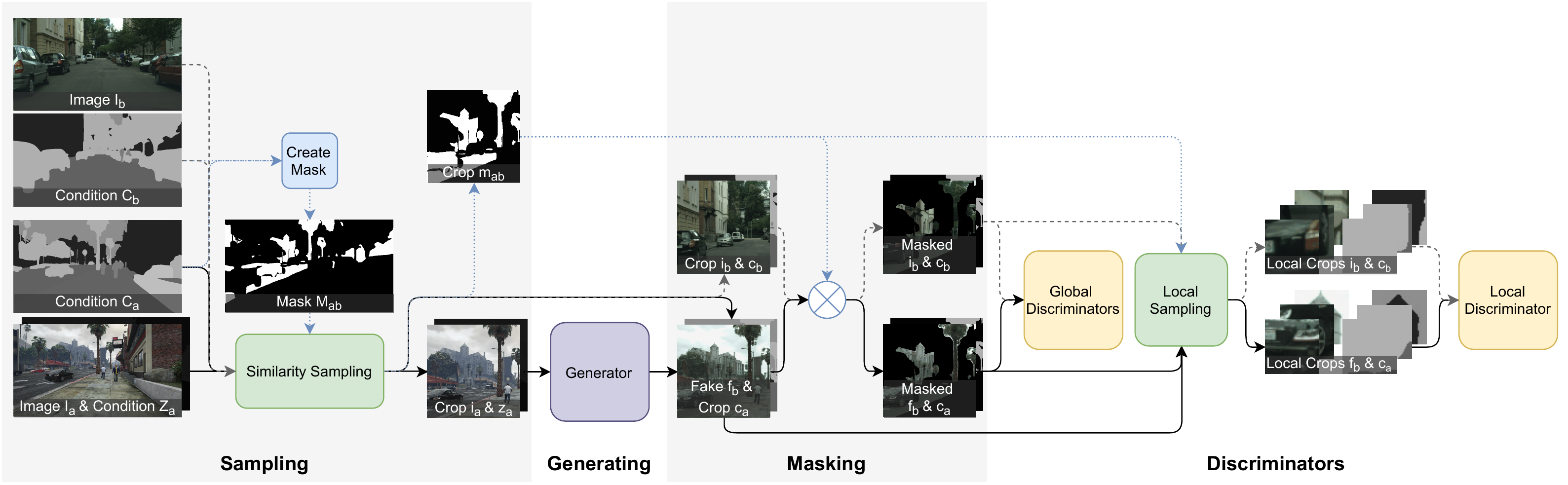}
	\end{center}
	\vspace{-1ex}
	\caption[Method overview.]{Method overview. In our method, similar image crops from both domains ($i_a$, $i_b$) and their corresponding conditions ($c_a$, $c_b$, $z_a$) are selected via a sampling procedure. In this sampling procedure, a mask $M_{ab}$ is created from the conditions $C_a$ and $C_b$. This mask is used to sample crops from both datasets for which the semantic classes align by at least 50\%. The cropped mask $m_{ab}$ is also used to mask the generated fake image $f_b$, the real images $i_b$, and the corresponding conditions for the global conditional discriminators. Through the mask, these discriminators can only see the parts of the crop where the semantic classes align. To further improve image quality, a local discriminator is introduced that works on a batch of small patches selected from the crop using our sampling technique. This discriminator is not masked and works on patches where the semantic classes do not fully align. Best viewed in color.}
	\label{fig:feamgan:method_overview}
\end{figure}
We propose an end-to-end framework for unpaired image-to-image translation that transfers an image $I_a \in \mathbb{R}^{3\times h\times w}$ from a source domain $a$ to an image $F_b \in \mathbb{R}^{3\times h\times w}$ from a target domain $b$. Our goal is to design a method for content-consistent translations that utilizes a simple masking strategy for the global crops seen by the discriminators. We achieve this by combining an efficient segmentation-based sampling method that samples large crops from the input image with a masked discriminator that operates on these global crops. This is in contrast to EPE \cite{richter2022enhancing}, which achieves content-consistent translation at the local level by sampling small, similar image crops from both domains. To further improve image quality, we use a local discriminator that operates on a batch of small image patches sampled from the global input crops utilizing our sampling method. An overview of our method is shown in \autoref{fig:feamgan:method_overview}.
Furthermore, we propose a feature-attentive denormalization (FATE) block that extends feature-adaptive denormalization (FADE) \cite{jiang2020tsit} with an attention mechanism, allowing the generator to selectively incorporate statistical features from the content stream of the source image into the generator stream.

\subsection{Contend-based Similarity Sampling}
\label{sec:similaritySampling}
To minimize the bias between both datasets in the early stage of our method, we sample similar image crops with an efficient sampling procedure. This procedure uses the one-hot encoded semantic segmentations $C_a \in \mathbb{R}^{d\times h\times w}$ and $C_b \in \mathbb{R}^{d\times h\times w}$ of both domains, where $d$ is the channel dimension of the one-hot encoding. In our case, these segmentations are created with the robust pre-trained MSeg model \cite{lambert2020mseg}. First, a mask $M_{ab} \in \mathbb{R}^{1\times h\times w}$ is computed from the segmentations:
\begin{ceqn}
	\begin{equation}
		\label{eqn:mask}
		M_{ab} = \operatorname*{max}_d (C_a\circ C_b),
	\end{equation}
\end{ceqn}
where $\circ$ denotes the Hadamard product. 
We can now sample semantically aligned image crops $i_a$ and $i_b$ from the images $I_a$ and $I_b$ with the crop $m_{ab}$ from mask $M_{ab}$. Thereby, we calculate the percentage of overlap of semantic classes between both image crops as follows:
\begin{ceqn}
	\begin{equation}
		\label{eqn:similaritysampling}
		\mathcal{P}_{match}(i_a)=\{i_b\mid\operatorname*{mean}(m_{ab}) > t\} ,
	\end{equation}
\end{ceqn}
where $t$ is the similarity sampling threshold. In our case, we sample crops where more than $50\%$ of the semantic classes align ($t > 0.5$). We use this procedure to sample crops $c_a$, $c_b$, and $z_b$ from the discriminator conditions $C_a$, $C_b$, and the generator condition $Z_b$ as well. The cropped mask $m_{ab}$ is also used for our masked conditional discriminator.s

\subsection{Contend-based Discriminator Masking}
To train a discriminator with a global field of view that facilitates the usage of global properties of the scene, while maintaining content consistency, we mask the discriminator input from both domains with a content-based mask $m_{ab}$. This mask erases all pixels from the discriminator input where the semantic classes do not align. This removes the bias between both datasets caused by the underlying semantic class distribution of the two domains without directly restricting the generator. The objective function of a conditional GAN with a masked discriminator that transfers image crops $i_a$ to domain $b$ can be then defined as follows:
\begin{ceqn}
	\begin{equation}
		\label{eqn:masked_adv_obj}
		\begin{split}
			&\mathcal{L}_{madv}= ~\mathbb{E}_{i_b,c_b,m_{ab}}[\log D(i_b\circ m_{ab}| c_b\circ m_{ab})] \\ &+ \mathbb{E}_{i_a,z_a,c_a, m_{ab}}[\log (1 - D(G(i_a| z_a)\circ m_{ab}| c_a\circ m_{ab}))].
		\end{split}
	\end{equation} 
\end{ceqn}
To ensure that the discriminator does not use the segmentation maps as learning shortcuts, we follow \cite{richter2022enhancing} and create the segmentations of both datasets using a robust segmentation model such as MSeg \cite{lambert2020mseg}. With this setting, we are able to train discriminators with large crop sizes with significantly reduced hallucinations in the translated image.

\subsection{Local Discriminator}
Masking the input of the discriminator may lead to the underrepresentation of some semantic classes. Therefore, we additionally train a local discriminator that operates on a batch of small patches sampled from the global crop. Our local discriminator is not masked but only sees patches where a certain amount of the semantic classes align. In our case, we sample patches with 1/8th the size of the global input crop where more than $50\%$ of the semantic classes align. We use our sampling procedure from \autoref{sec:similaritySampling} to sample these patches. Using small, partially aligned patches ensures that semantic classes are less underrepresented while maintaining content consistency. 

\subsection{Feature-attentive Denormalization (FATE)}
\begin{figure}[t]
	\begin{center}
		\includegraphics[width=0.75\linewidth]{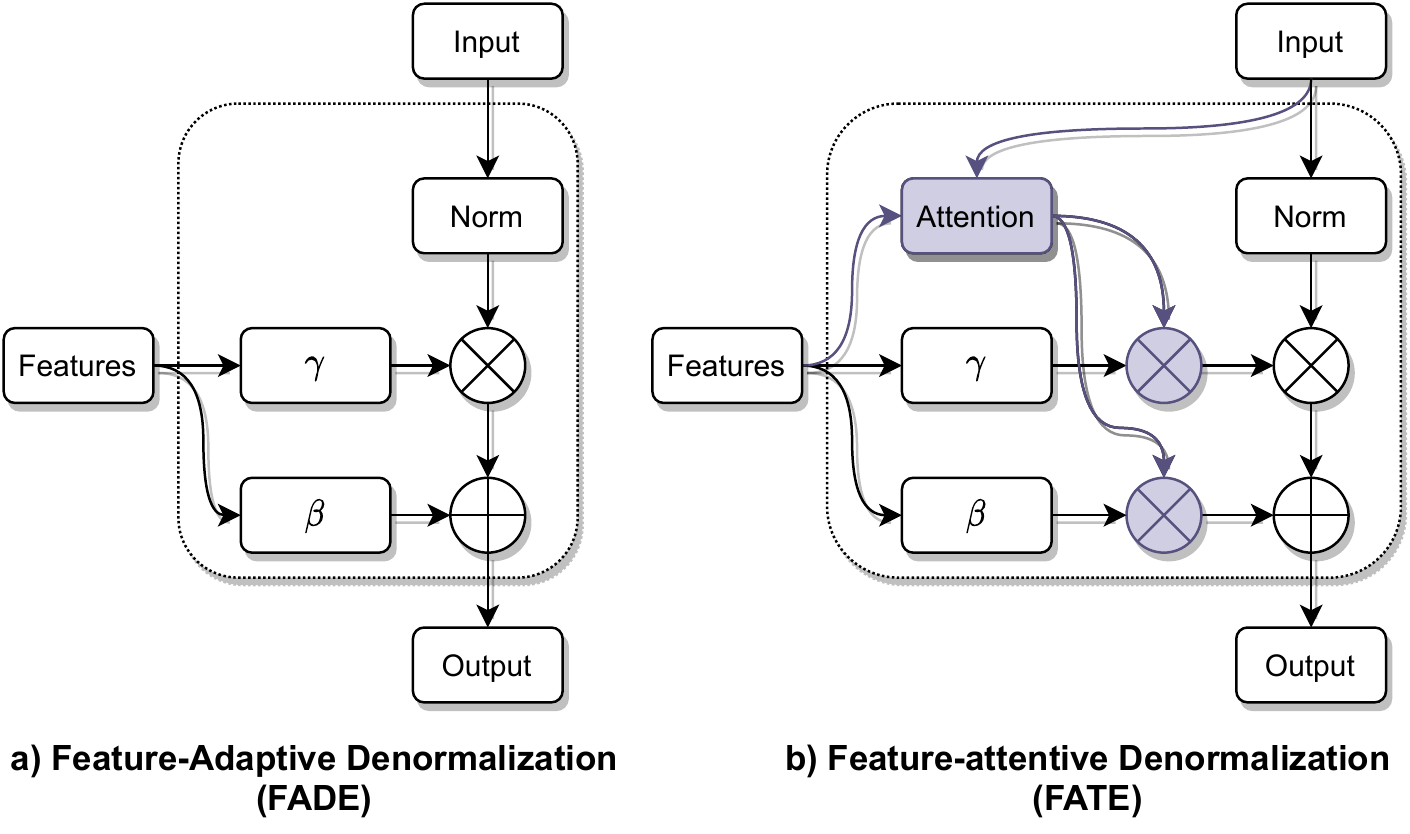}
	\end{center}
	\vspace{-1ex}
	\caption[FADE and FATE.]{FADE and FATE.}
	\label{fig:feamgan:FATE}
\end{figure}

Spatially adaptive denormalization (SPADE) \cite{park2019semantic} fuses resized semantic segmentation maps as content into the generator stream. Feature-adaptive denormalization (FADE) \cite{jiang2020tsit} generalizes SPADE to features learned through a content stream. As shown in \autoref{fig:feamgan:FATE}, the normalized features $N(h)$ of the generator are modulated with the features $f$ of the content stream using the learned functions $\gamma$ and $\beta$ as follows:
\begin{ceqn}
	\begin{equation}
		\label{eqn:FADE}
		\operatorname*{FADE}(h,f) = N(h)\circ\gamma(f) + \beta(f) ,
	\end{equation}
\end{ceqn}
where $\gamma$ and $\beta$ are one-layer convolutions. This denormalization is applied in several layers of the generator. However, we argue that denormalization with content features is not always appropriate for transferring images to another domain because, as shown in \cite{gatys2016image,li2016combining,li2017demystifying,huang2017arbitrary}, image feature statistics contain not only content information but also style information. When transferring feature statistics from the content stream of the source image to the generator stream, style information from the source image could affect the final image quality if the generator cannot ignore this information since the output image should mimic the style of the target domain. Therefore, we propose an additional attention mechanism to selectively incorporate statistics from the content stream into the generator stream. This allows the model to only fuse the statistical features from the source image into the generator stream that are useful for the target domain. As shown in \autoref{fig:feamgan:FATE}, this attention mechanism relies on the features of the content stream and the features of the generator stream and attends to the statistics $\gamma$ and $\beta$. With this attention mechanism, we can extend FADE to feature-attentive denormalization (FATE) as follows:
\begin{ceqn}
	\begin{equation}
		\label{eqn:FATE}
		\operatorname*{FATE}(h,f) = N(h)\circ A(h,f)\circ\gamma(f)+ A(h,f)\circ \beta(f),
	\end{equation}
\end{ceqn}
where $A$ is the attention mechanism and $A(h,f)$ is the attention map for the statistics. We use a lightweight two-layer CNN with sigmoid activation in the last layer as the attention mechanism. More details can be found in \autoref{app:03}.

\subsection{Training Objective}
Our training objective consists of three losses: a global masked adversarial loss $\mathcal{L}_{madv}^{global}$, a local adversarial loss $\mathcal{L}_{adv}^{local}$, and the perceptual loss $\mathcal{L}_{perc}$ used in \cite{jiang2020tsit}. We define the final training objective as follows:
\begin{ceqn}
	\begin{equation}
		\label{eqn:overall_obj_gen}
		\mathcal{L}=\lambda_{madv}^{global}\mathcal{L}_{madv}^{global} + \lambda_{adv}^{local}\mathcal{L}_{adv}^{local} + \lambda_{perc}\mathcal{L}_{perc},
	\end{equation}
\end{ceqn}
where we use a hinge loss to formulate the adversarial losses and $\lambda_{madv}^{global}$, $\lambda_{madv}^{local}$, $\lambda_{perc}$ are the corresponding loss weights.
\section{Experiments}
\subsection{Experimental Settings}
\noindent \textbf{Implementation details.} 
Our method is implemented in PyTorch 1.10.0 and trained on an A100 GPU (40 GB) with batch size $1$. For training, we initialize all weights with the Xavier normal distribution \cite{glorot2010understanding} with a gain of $0.02$ and use an Adam optimizer \cite{kingma2014adam} with $\beta_1=0.9$ and $\beta_2=0.999$. The initial learning rates of the generator and discriminators are set to $0.0001$ and halved every $d_e$ epochs. Learning rate decay is stopped after reaching a learning rate of $0.0000125$. We formulate our adversarial objective with a hinge loss \cite{lim2017geometric} and weight the individual parts of our loss function as follows: $\lambda_{madv}^{global}=1.0, \lambda_{madv}^{local}=1.0$, $\lambda_{perc}=1.0$. In addition, we use a gradient penalty on target images \cite{gulrajani2017improved, mescheder2018training} with $\lambda_{rp}=0.03$. The images of both domains are resized and cropped to the same size and randomly flipped before the sampling strategy is applied. In our experiments, we show that we achieve the best performance by cropping global patches of size 352$\times$352. We crop local patches with 1/8th the size of the global crop (i.a., 44$\times$44). The global discriminators are used on two scales. Crops are scaled down by a factor of two for the second scale. We train all our models for $\sim\!400$K iterations. Training a model takes 4-8 days, depending on the dataset, model, and crop size. We report all results as an average across five different runs. We refer to \autoref{app:03} for more details regarding the training and model. Our implementation is publicly available at \href{https://github.com/BonifazStuhr/feamgan}{https://github.com/BonifazStuhr/feamgan}.
\\\\
\noindent \textbf{Memory usage.} Our best model requires $\sim$25 GB of VRAM at training time and performs inference using $\sim$12 GB for an image of size 957$\times$526. Our small model, with a slight performance decrease, runs on consumer graphic cards with $\sim$9 GB of VRAM at training time and performs inference using $\sim$8 GB for an image of size 957$\times$526. 
\\\\
\noindent \textbf{Datasets.} We conduct experiments on four translation tasks across four datasets. For all datasets, we compute semantic segmentations with MSeg \cite{lambert2020mseg}, which we use as a condition for our discriminator and to calculate the discriminator masks.
\\\\
(1) \textit{PFD} \cite{richter2016playing} consists of images of realistic virtual world gameplay. Each frame is annotated with pixel-wise semantic labels, which we use as additional input for our generator. We use the same subset as \cite{richter2022enhancing} to compare with recent work.\\\\
(2) \textit{Viper} \cite{richter2017playing} consists of sequences of realistic virtual world gameplay. Each frame is annotated with different labels, where we use the pixel-wise semantic segmentations as additional input for our generator. Since Cityscapes does not contain night sequences, we remove them from the dataset.\\\\
(3) \textit{Cityscapes} \cite{cordts2016cityscapes} consists of sequences of real street scenes from 50 different German cities. We use the sequences of the entire training set to train our models. We use datasets (1-3) for the sim-to-real translation tasks \textit{PFD$\rightarrow$Cityscapes} and \textit{Viper$\rightarrow$Cityscapes}. \\\\
(4) \textit{BDD100K} \cite{yu2020bdd100k} is a large-scale driving dataset.  We use subsets of the training and validation data for the following translation tasks: \textit{Day$\rightarrow$Night}, \textit{Clear$\rightarrow$Snowy}.
\\\\
\noindent \textbf{Compared methods.}
We compare our work with the following methods.
\vspace{-4pt}
\begin{itemize}
	\itemsep0em 
	\item Color Transfer (CT) \cite{reinhard2001color} performs color correction by transferring statistical features in lαβ space from the target to the source image.	 
	\item MUNIT \cite{huang2018multimodal} achieves multimodal translation by recombining the content code of an image with a style code sampled from the style space of the target domain. It is an extension of CycleGAN \cite{zhu2017unpaired} and UNIT \cite{liu2017unsupervised}.
	\item CUT \cite{park2020contrastive} uses a patchwise contrastive loss to achieve one-sided unsupervised image-to-image translation.	
	\item TSIT \cite{jiang2020tsit} achieves one-sided translation by fusing features from the content stream into the generator on multiple scales using FADE and utilizing a perceptual loss between the translated and source images.
	\item QS-Attn \cite{hu2022qs} builds upon CUT \cite{park2020contrastive} with an attention module that selects significant anchors for the contrastive loss instead of features from random locations of the image.
	\item EPE \cite{richter2022enhancing} relies on a variety of gbuffers as input. Techniques such as similarity cropping, utilizing segmentations for both domains generated by a robust segmentation model as input to the conditional discriminators, and small patch training are used to achieve content consistency.
\end{itemize}
Since EPE \cite{richter2022enhancing} provides inferred images of size 957$\times$526 for the \textit{PFD$\rightarrow$Cityscapes} task, comparisons are performed on this resolution. For the \textit{Viper$\rightarrow$Cityscapes}, \textit{Day$\rightarrow$Night}, and \textit{Clear$\rightarrow$Snowy} tasks, we train the models using their official implementations. Furthermore, we retrain models as additional baselines for the \textit{PFD$\rightarrow$Cityscapes} task.
\\\\
\noindent \textbf{Evaluation metrics.} Following prior work \cite{richter2022enhancing}, we use the Fréchet Inception Distance (FID) \cite{heusel2017gans}, the Kernel Inception Distance (KID) \cite{binkowski2018demystifying}, and the semantically aligned Kernel VGG Distance (sKVD) \cite{richter2022enhancing} to evaluate image translation quality quantitatively. The sKVD metric was introduced in \cite{richter2022enhancing} and improved over previous metrics for mismatched layouts in source and target data. In addition, we propose the class-specific Kernel VGG Distance (cKVD), where a robust segmentation model is used before the sKVD calculation to mask input crops by class (or category). Thereby, for each given class, all source and target image crops are filtered using their segmentations by erasing the pixels of all other classes. We select crops where more then $5$\% of the pixels belong to the respective class. Then, the sKVD is calculated class-wise on the filtered crops. Afterward, we can report the cKVD as an average over all classes or separately for each class to achieve a more fine-grained measurement. We follow \cite{richter2022enhancing} and use a crop size of $1/8$ and sample source and target crop pairs with an similarity threshold  of $0.5$ between unmasked source and target segmentation crops. More information on the classes used in the cKVD metric can be found in \autoref{tab:feamgan:app:sKVD_class_mapping} of \autoref{app:03}. For the KID, sKVD, and cKVD metrics, we multiply the measurements by $1000$ to improve the readability of results.
\begin{figure}[H] 
	\captionsetup[subfigure]{labelformat=empty}
	\begin{center}
		{\includegraphics[width=0.33\textwidth]{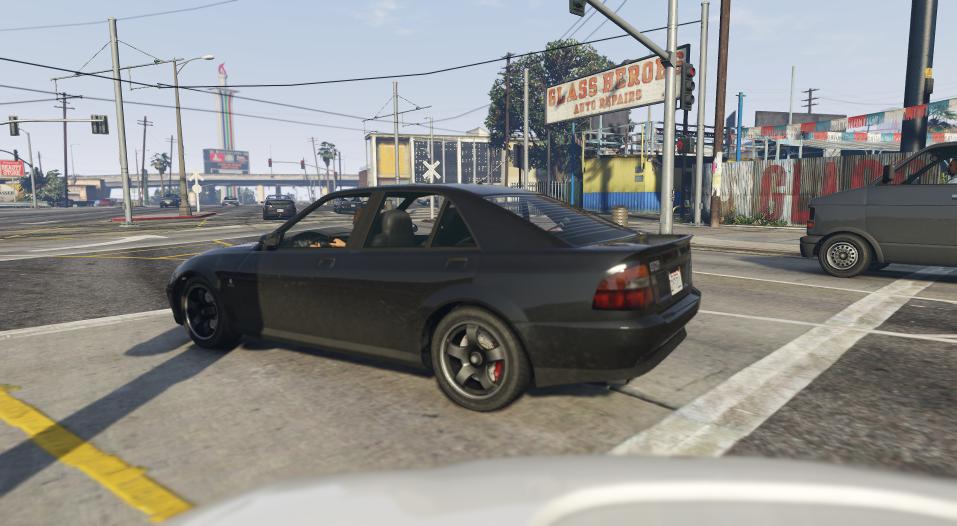}}\hfill
		{\includegraphics[width=0.33\textwidth]{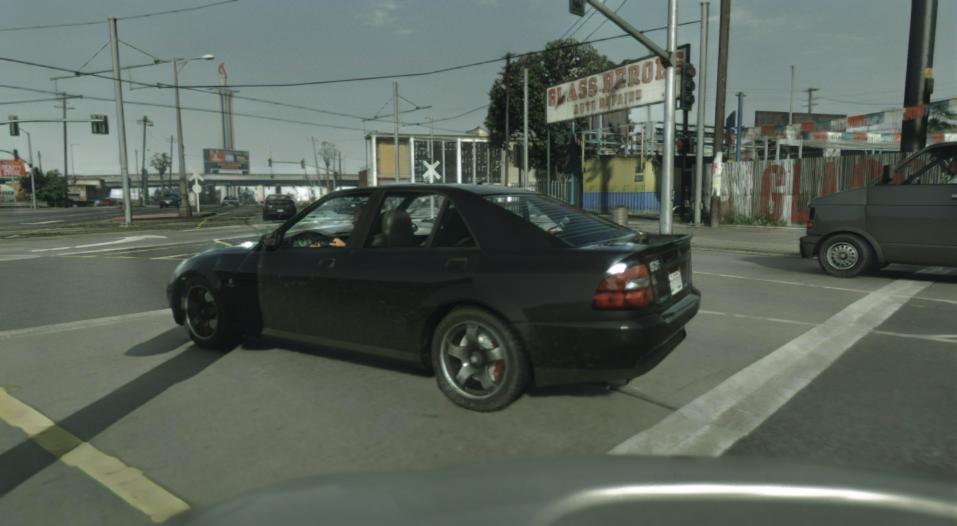}}\hfill
		{\includegraphics[width=0.33\textwidth]{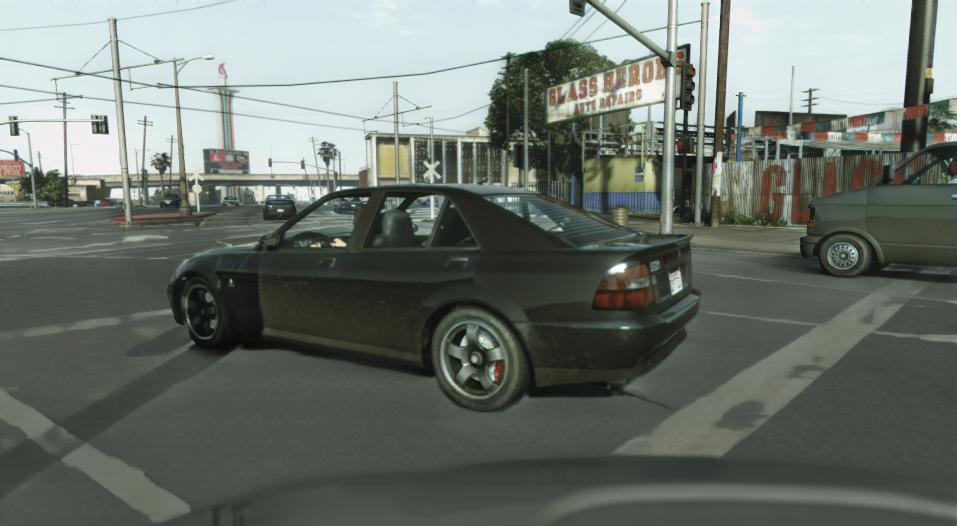}}\hfill\\
		\vspace{-9pt}
		\subfloat[Input]
		{\includegraphics[width=0.33\textwidth]{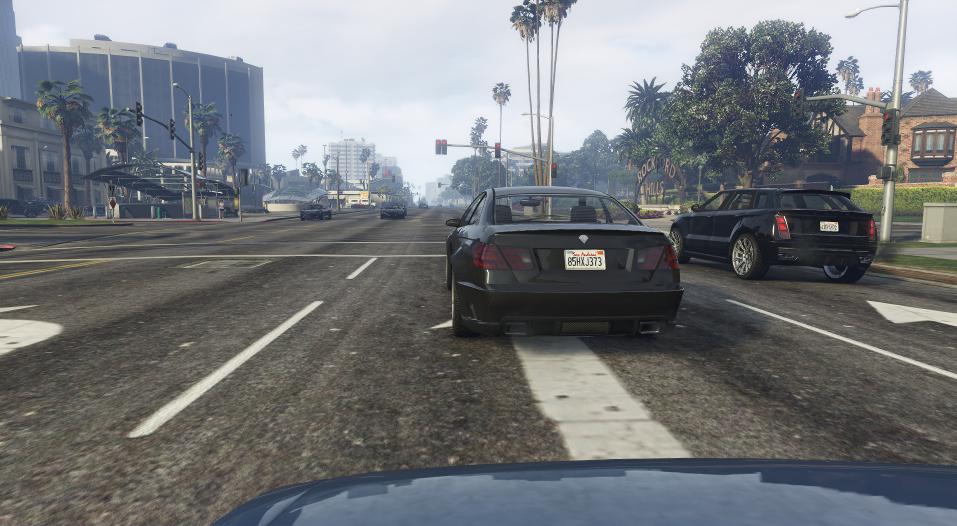}}\hfill
		\subfloat[EPE]
		{\includegraphics[width=0.33\textwidth]{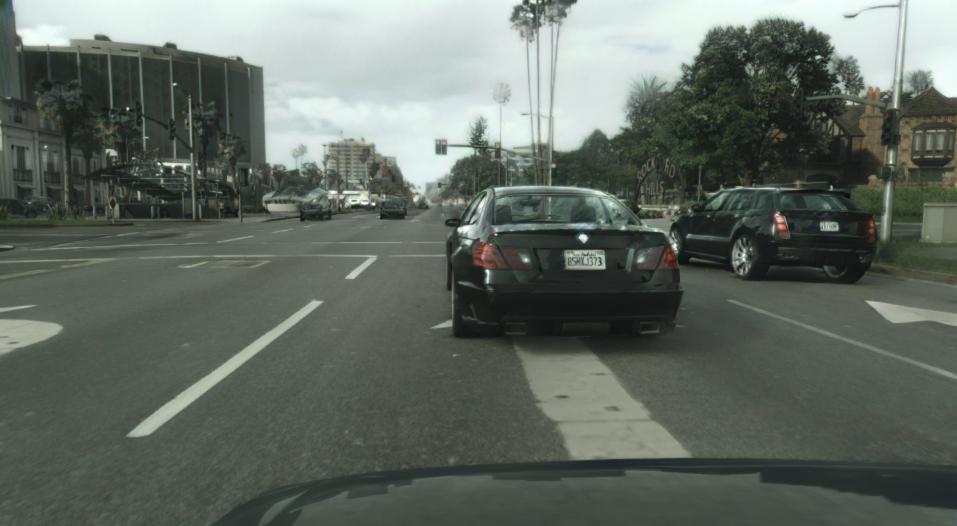}}\hfill
		\subfloat[FeaMGAN (ours)]
		{\includegraphics[width=0.33\textwidth]{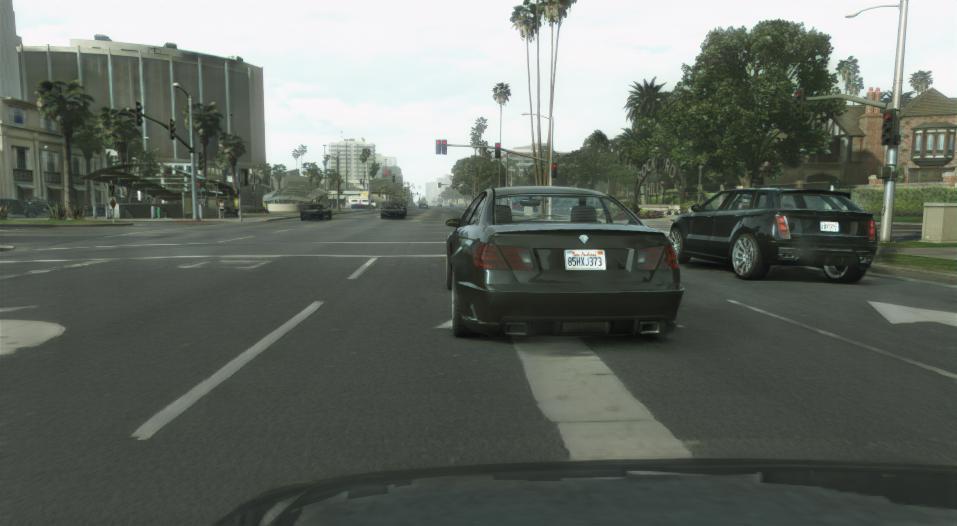}}\hfill\\
		\vspace{-4pt}
		\subfloat[Generator Input Types]
		{\includegraphics[width=1.0\linewidth]{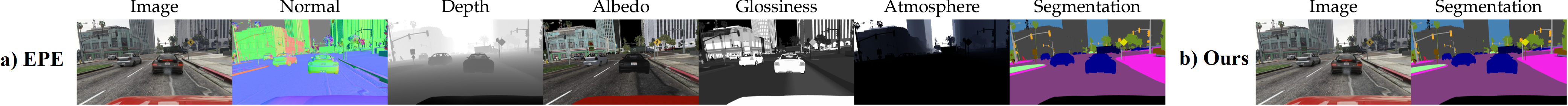}}\\
	\end{center}
	\vspace{-1ex}
	\caption[Qualitative comparison to EPE.]{Qualitative comparison to EPE. We compare our method with the provided inferred images of EPE \cite{richter2022enhancing}. Best viewed in color.}
	\label{fig:feamgan:qualitative_comparison_epe_baselines}
\end{figure}
\begin{table}[H]
	\RawFloats
	\scriptsize
	\caption[Quantitative comparison to the baselines provided by EPE.]{Quantitative comparison to the baselines provided by EPE. We calculate all metrics on the provided inferred images of EPE and its baselines \cite{richter2022enhancing}.} 	
	\vspace{-3ex}
	\label{tab:feamgan:quantitative_comparison_epe_baselines}
	\begin{center}
		\setlength{\tabcolsep}{0.25em}
		\scalebox{0.82}{
		\begin{tabular}{lccccccccccccccc}
			\toprule
			\multirow{2}{*}{Method}&\multirow{2}{*}{FID}&\multirow{2}{*}{KID} &\multirow{2}{*}{sKVD}&\multicolumn{12}{c}{cKVD}\\
			\cmidrule(lr){5-16}
			&    &  &  & AVG& AVG$_{sp}$&		sky& 	ground&	road&	terrain&	vegetation&	building&	roadside-obj.&	person&	vehicle&rest\\
			\midrule	
			ColorTransfer& 84.34 & 88.17 & 16.65 & 36.01 & 33.12& 32.40 & \textbf{12.97} &16.13 & 20.94 & 19.24 & 29.92 & 74.79 & 62.78 & 41.79 & \textbf{49.16}\\
			MUNIT  & 45.00  & 35.05 & 16.51  & 38.57 &34.81 &29.80 & 16.93	& 17.62 & 29.52& 19.29	&\textbf{24.28} & 79.14 & 77.34 & 40.13 & 51.61 \\
			CUT & 47.71 & 42.01   & 18.03   & 35.31 & 33.26 &\textbf{25.96} & 15.32 &	17.87 & \textbf{20.09} & 22.72 & 25.00 & 74.02 & \textbf{60.99} & 41.71	&49.37 \\
			EPE & 44.06 & 33.66  & 13.87  & \textbf{35.22} &\textbf{30.21}&27.14	&13.54	&\textbf{13.56}&24.77&20.77	&26.75&\textbf{50.58} &	83.34&41.29&50.45 \\
			\midrule			
			FeaMGAN-S (ours) & \textbf{43.27} & \textbf{32.59} & \textbf{12.98}  &40.23	&32.69	
			& 38.10	&13.29	&15.34	&26.29	&20.17	&27.32	&61.57&102.65	&42.83	&54.73 \\
			FeaMGAN (ours) & \textbf{40.32}  & \textbf{28.59} & \textbf{12.94} & 40.02  &31.78&46.70	&13.72	&15.60	&23.23	&\textbf{17.69}	&25.57	&66.65	&99.24	&\textbf{39.38}	&52.40 \\	
			\bottomrule
		\end{tabular}}
	\end{center}
	\vspace{-2ex}
\end{table}

\subsection{Comparison to the State of the Art}
We compare our models quantitatively and qualitatively with different baselines. First, we compare our results with EPE and the baselines provided by EPE \cite{richter2022enhancing}. Then, we train our own baselines on the four translation tasks for further comparison. \\\\
\noindent \textbf{Comparison to EPE.}
A set of inferred images is provided for EPE and each of the baselines \cite{richter2022enhancing}. Therefore, we train our models on the same training set and use the inferred images from our best models for this comparison. We select our best models based on scores of various visual metrics and visual inspections of translated images. As shown in \autoref{fig:feamgan:qualitative_comparison_epe_baselines} a) and b), our model relies solely on segmentation maps as additional input compared to EPE, which uses a variety of gbuffers. In addition, our model is trained with significantly fewer steps ($\sim\!400$K iterations) compared to EPE and the baselines ($1$M iterations). As shown in \autoref{tab:feamgan:quantitative_comparison_epe_baselines}, our model outperforms the baselines and EPE in all commonly used metrics (FID and KID) and the sKVD metric. More surprisingly, our small model, which can be trained on consumer GPUs, outperforms all baselines and EPE as well. 

However, our cKVD metric shows that our models have difficulty with the person and sky classes. Therefore, the average cKVD values are high and become low when we remove both classes from the average calculation (AVG$_{sp}$). A possible reason for the weaker performance on the person class is our masking procedure. Since the masking procedure requires overlapping samples in both domains, the person class is not seen frequently during training. This can lead to inconsistencies (a glow) around the person class, as seen in \autoref{fig:feamgan:limitations} of our limitations. The masking procedure also leads to a drop in performance in the sky class, as seen in \autoref{tab:feamgan:quantitative_ablation} of our ablation study.

As shown in the first row of \autoref{fig:feamgan:qualitative_comparison_epe_baselines} and the results of Figures \ref{fig:feamgan:app:qualitative_comparison_epe_additional} and \ref{fig:feamgan:app:qualitative_comparison_epe_additional_random} of \autoref{app:03}, our model translates larger structures, such as lane markings, more consistently, but fails to preserve some in-class characteristics from the source dataset. This is evident, for example, in the structure of translated streets and the corresponding cKVD value (road). As shown in the second row and \autoref{app:03}, EPE achieves visually superior modeling of the reflective properties of materials (e.g., the car) but suffers from inconsistencies (erased objects) regarding the vegetation, which can be seen in the palm trees and the corresponding cKVD value (vegetation). The superior modeling of reflective properties can be attributed to the availability of gbuffers (i.a., glossines) in EPE's input.

By surpassing EPE in all commonly used quantitative metrics while maintaining content consistency, we are able to show that our model improves overall quantitative translation performance. However, our method has specific drawbacks that we discussed with the help of the cKVD metric and visual comparisons. \\
\begin{figure}[H] 
	\captionsetup[subfigure]{labelformat=empty}
	\begin{center}	
		{\scriptsize PFD$\rightarrow$Cityscapes}\hfill\\\vspace{1pt}
		{\includegraphics[width=0.165\textwidth]{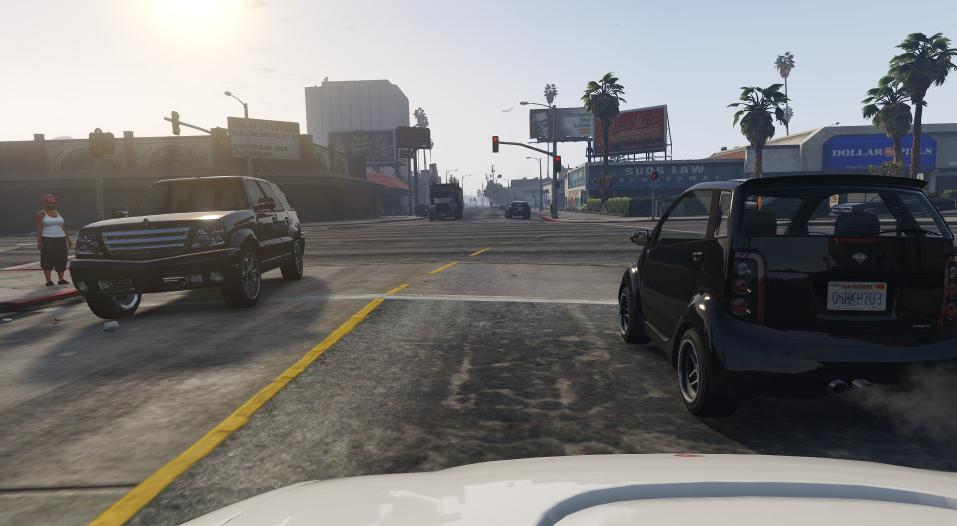}}\hfill
		{\includegraphics[width=0.165\textwidth]{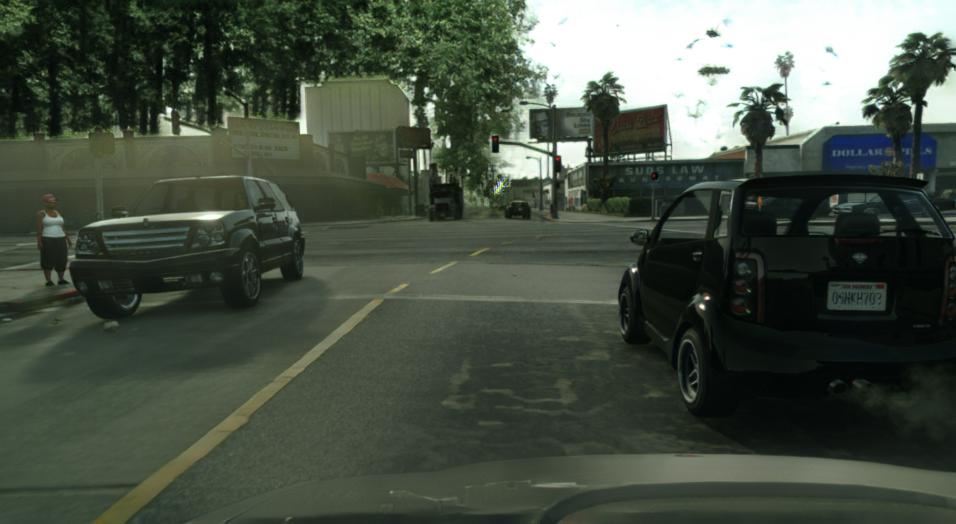}}\hfill
		{\includegraphics[width=0.165\textwidth]{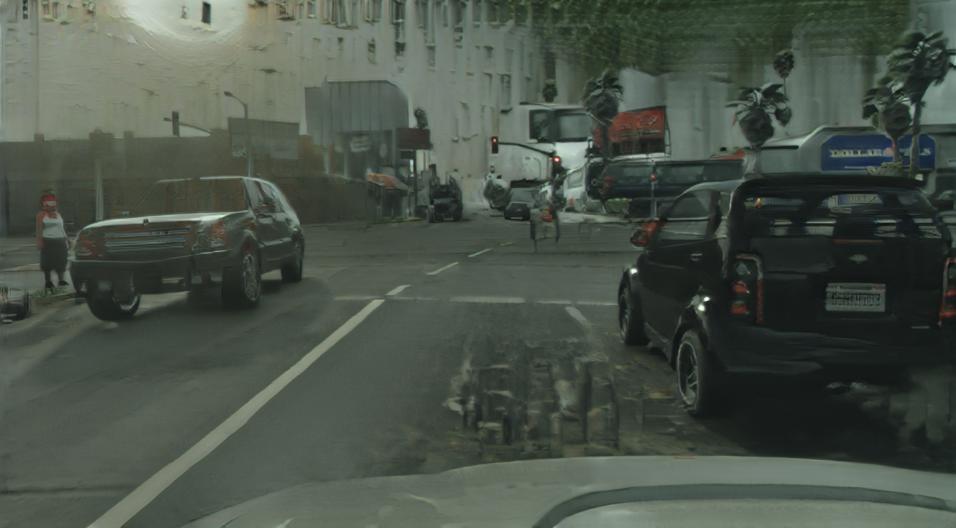}}\hfill
		{\includegraphics[width=0.165\textwidth]{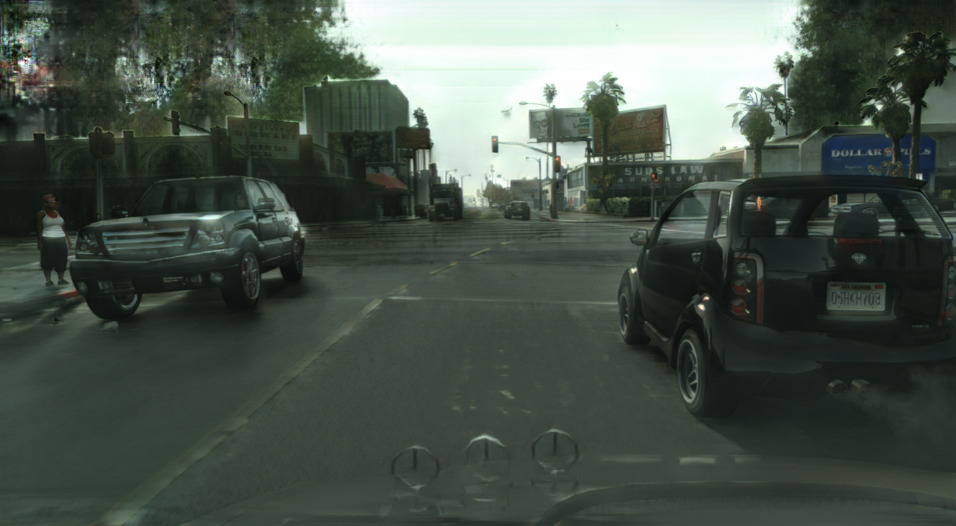}}\hfill
		{\includegraphics[width=0.165\textwidth]{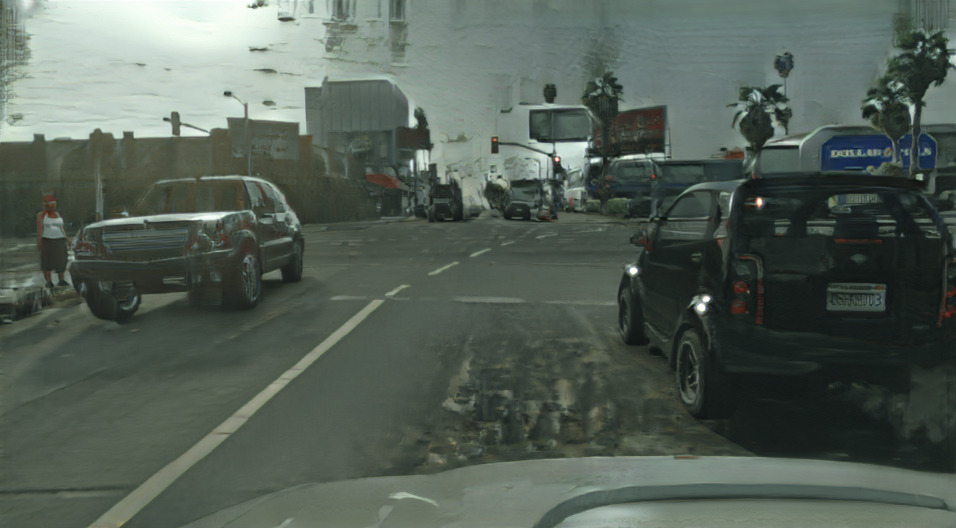}}\hfill
		{\includegraphics[width=0.165\textwidth]{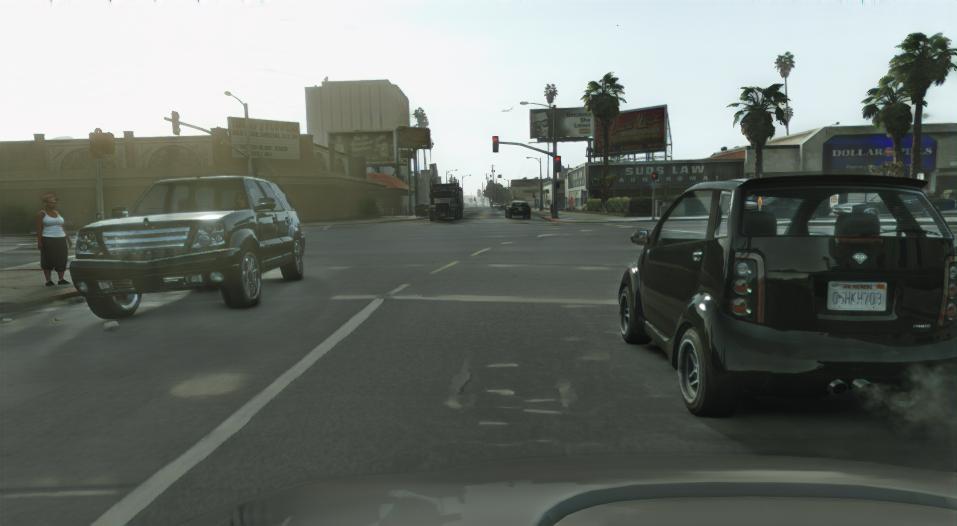}}\hfill \\ \vspace{-2.2pt}
		{\scriptsize Viper$\rightarrow$Cityscapes} \hfill\\\vspace{1pt}
		{\includegraphics[width=0.165\textwidth]{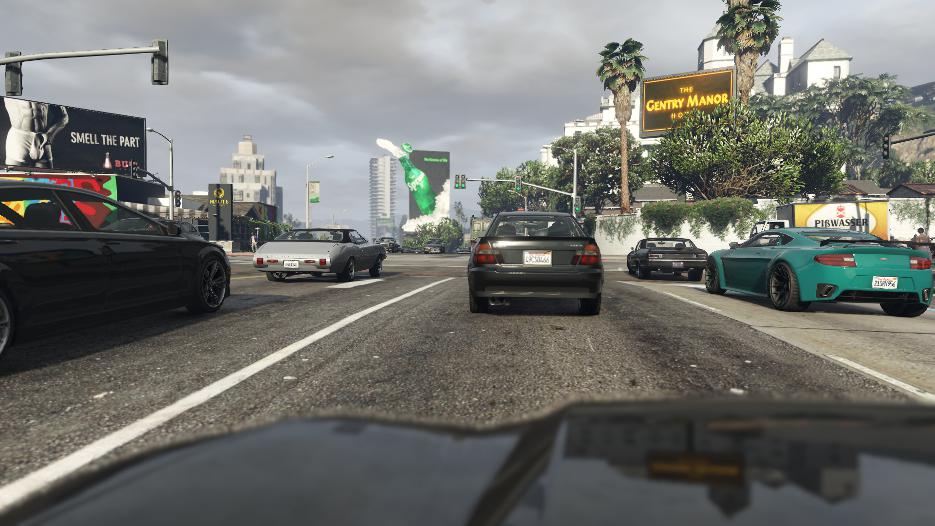}}\hfill
		{\includegraphics[width=0.165\textwidth]{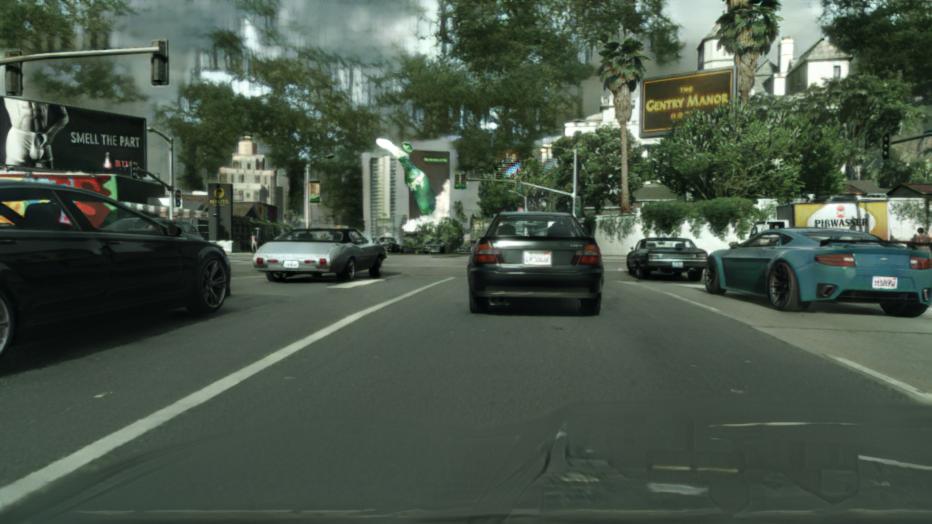}}\hfill
		{\includegraphics[width=0.165\textwidth]{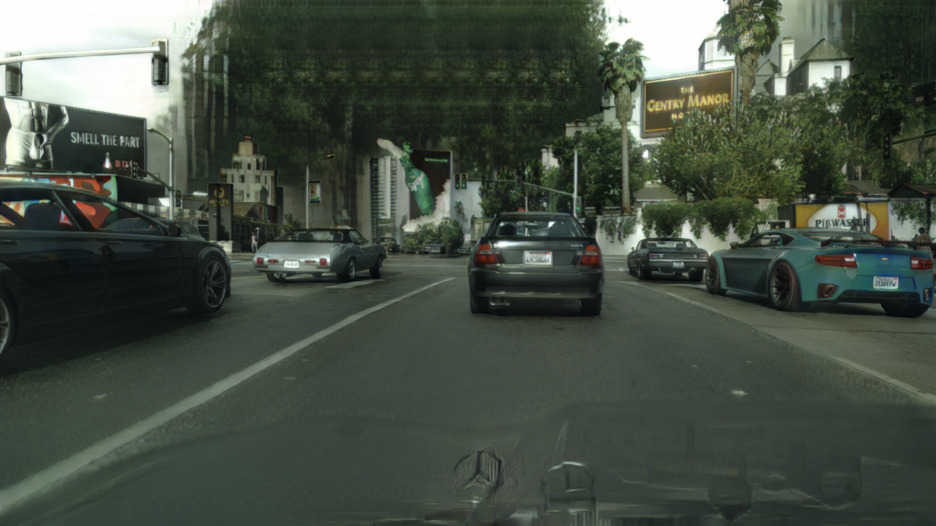}}\hfill
		{\includegraphics[width=0.165\textwidth]{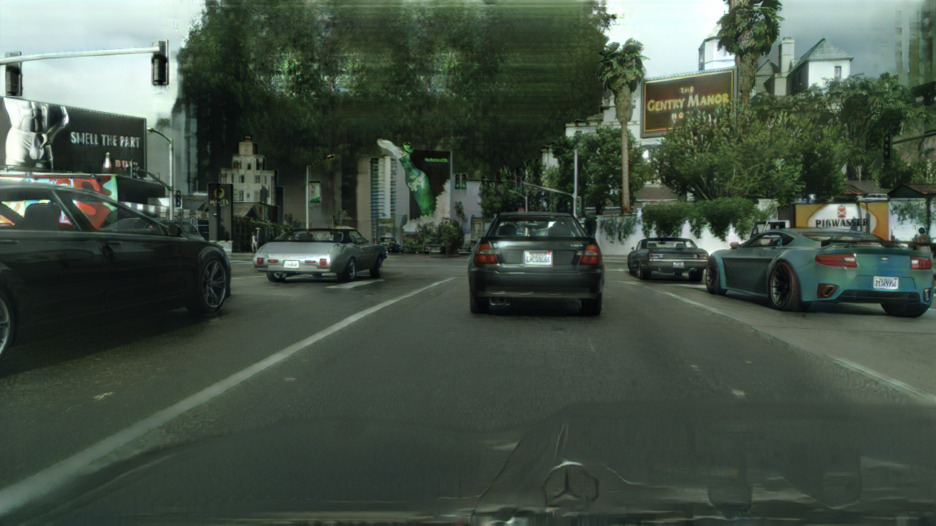}}\hfill
		{\includegraphics[width=0.165\textwidth]{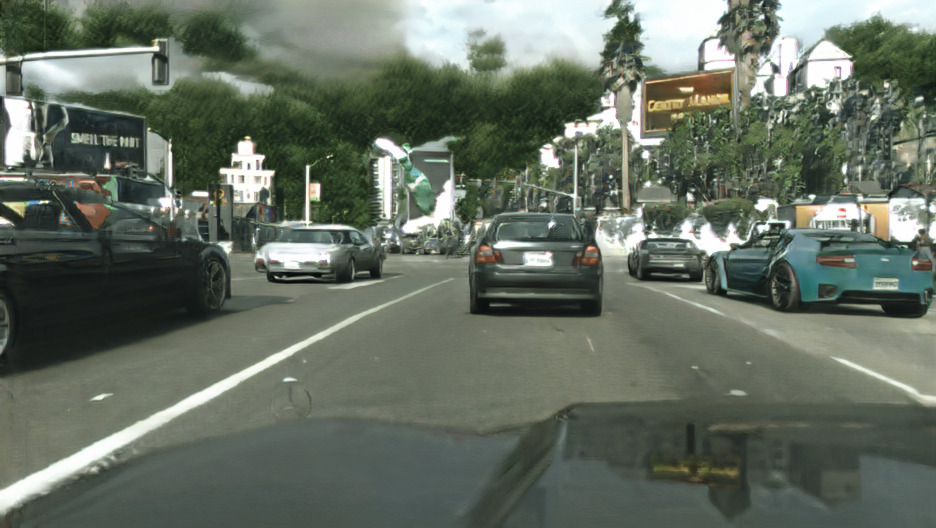}}\hfill
		{\includegraphics[width=0.165\textwidth]{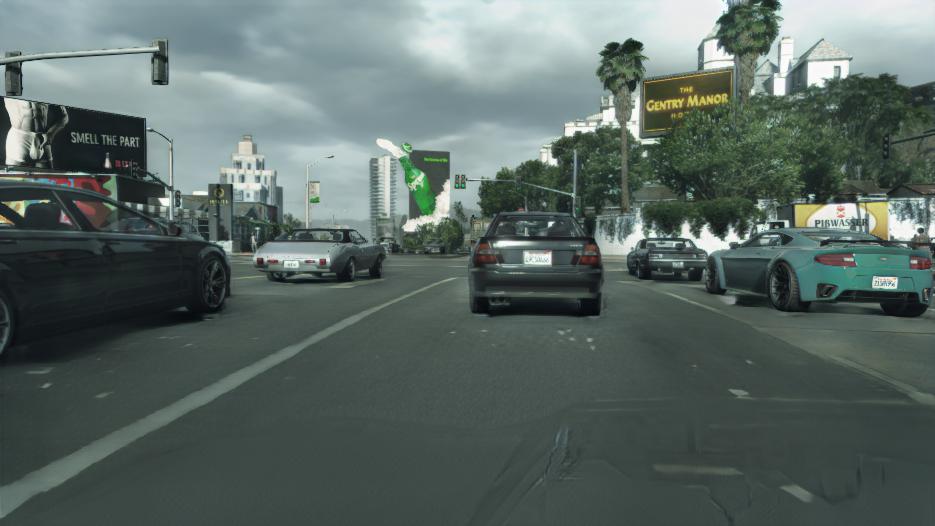}}\hfill \\ \vspace{-2.2pt}
		{\scriptsize Day$\rightarrow$Night} \hfill\\\vspace{1pt}
		{\includegraphics[width=0.165\textwidth]{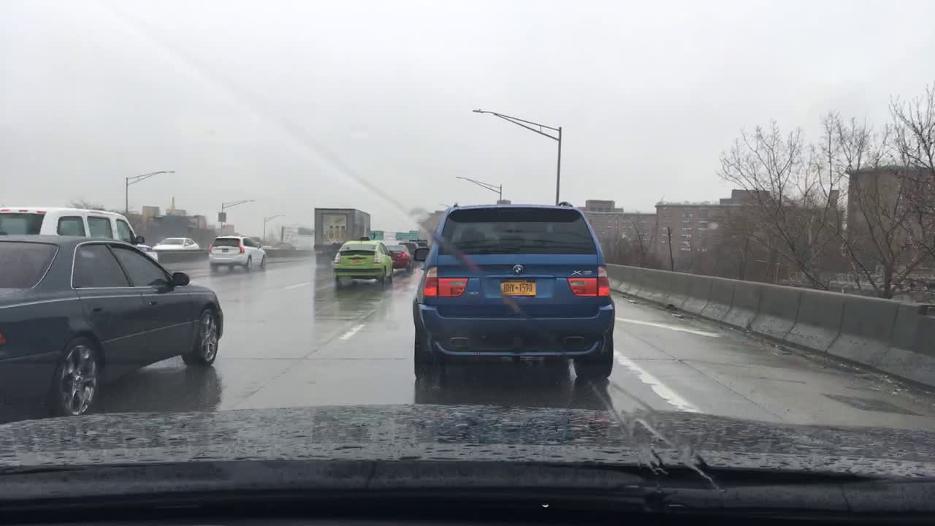}}\hfill
		{\includegraphics[width=0.165\textwidth]{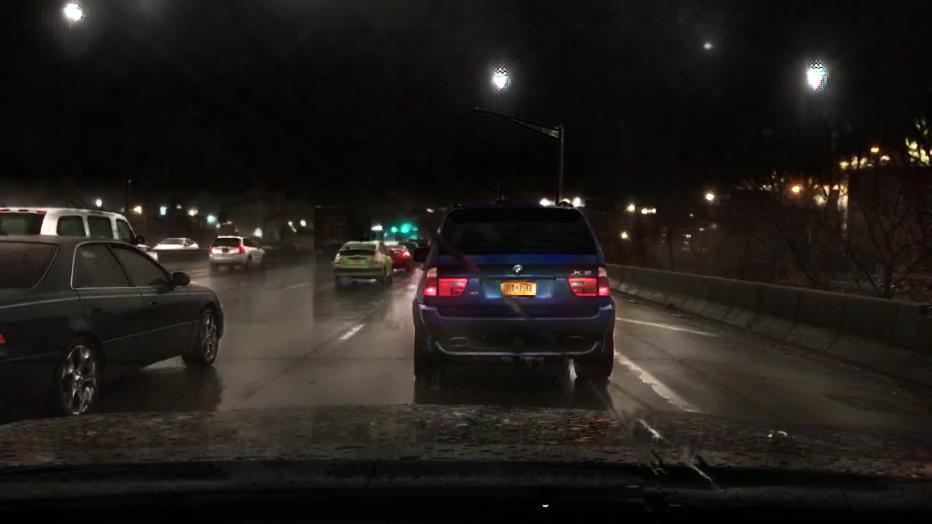}}\hfill
		{\includegraphics[width=0.165\textwidth]{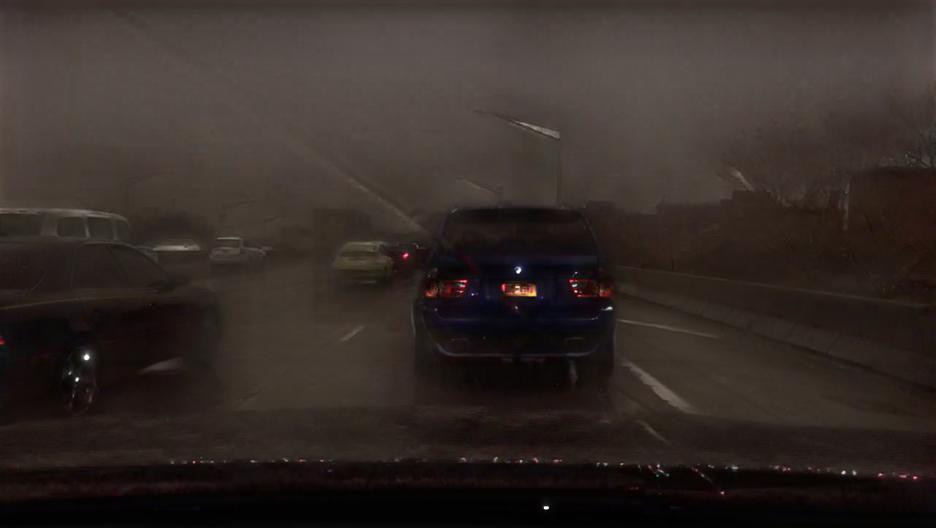}}\hfill
		{\includegraphics[width=0.165\textwidth]{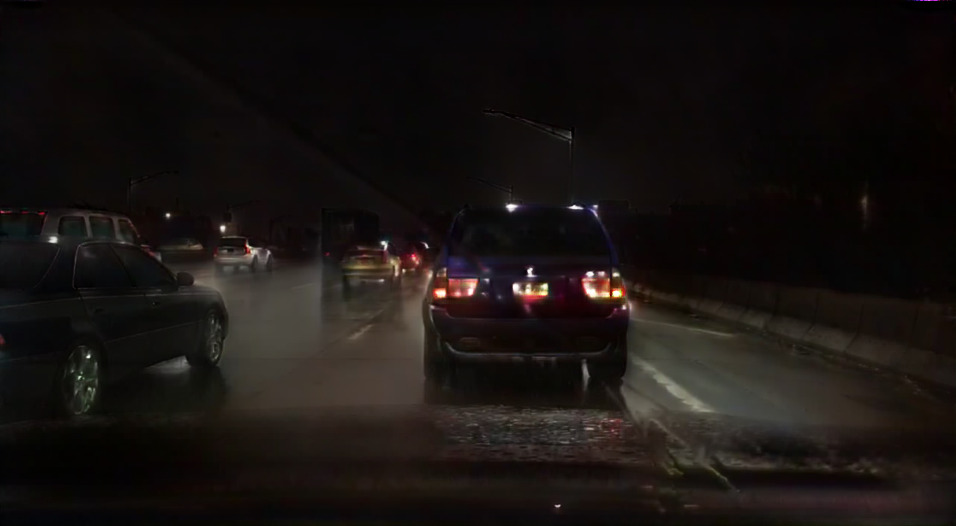}}\hfill
		{\includegraphics[width=0.165\textwidth]{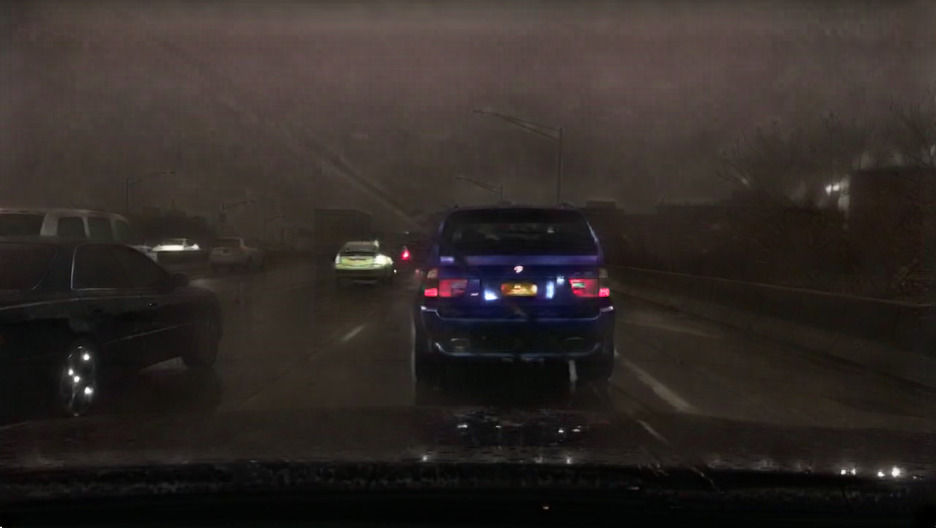}}\hfill
		{\includegraphics[width=0.165\textwidth]{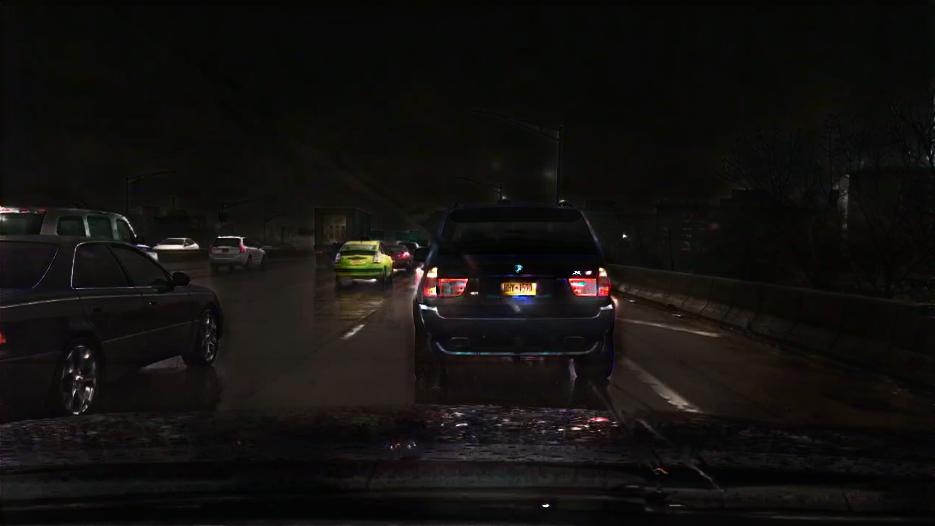}}\hfill \\ \vspace{-2.2pt}
		{\scriptsize Clear$\rightarrow$Snowy} \hfill\\\vspace{-9pt}
		\subfloat[Input]
		{\includegraphics[width=0.165\textwidth]{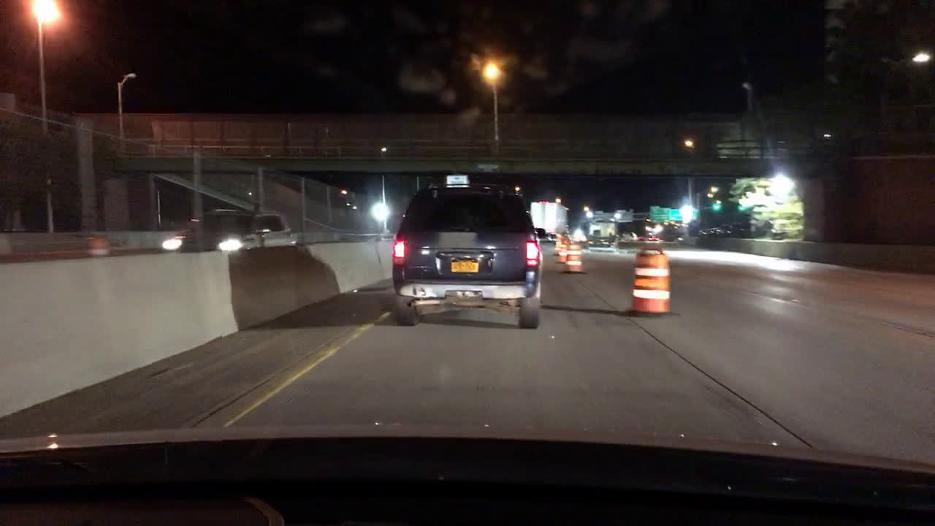}}\hfill
		\subfloat[MUNIT]
		{\includegraphics[width=0.165\textwidth]{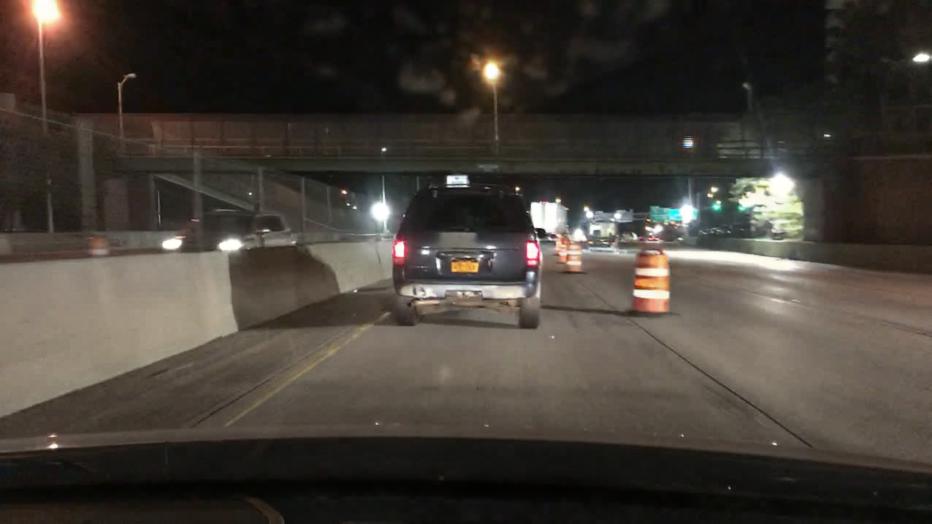}}\hfill
		\subfloat[CUT]
		{\includegraphics[width=0.165\textwidth]{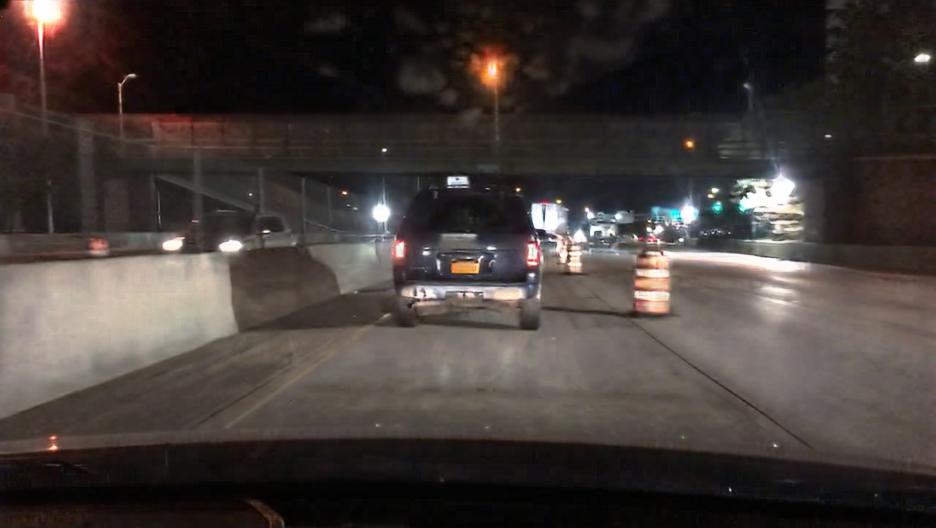}}\hfill
		\subfloat[TSIT]
		{\includegraphics[width=0.165\textwidth]{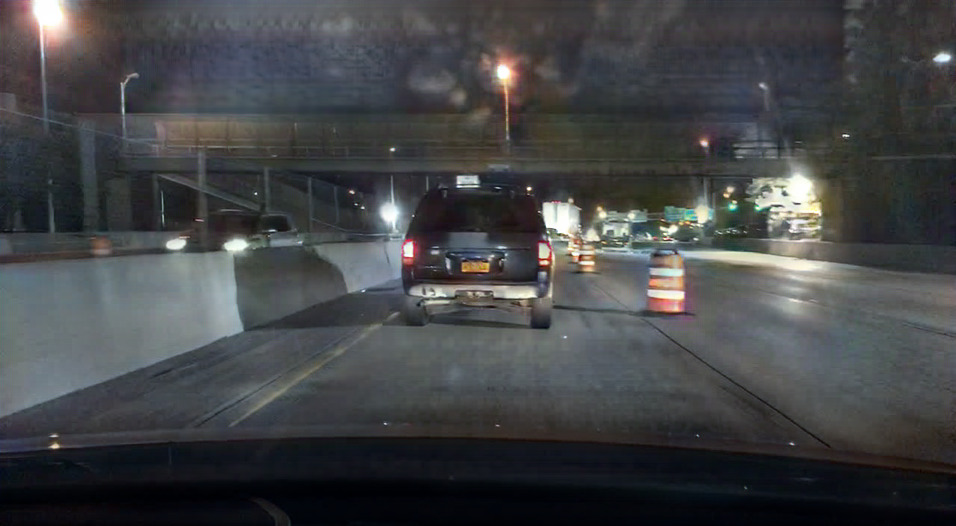}}\hfill
		\subfloat[QS-Attn]
		{\includegraphics[width=0.165\textwidth]{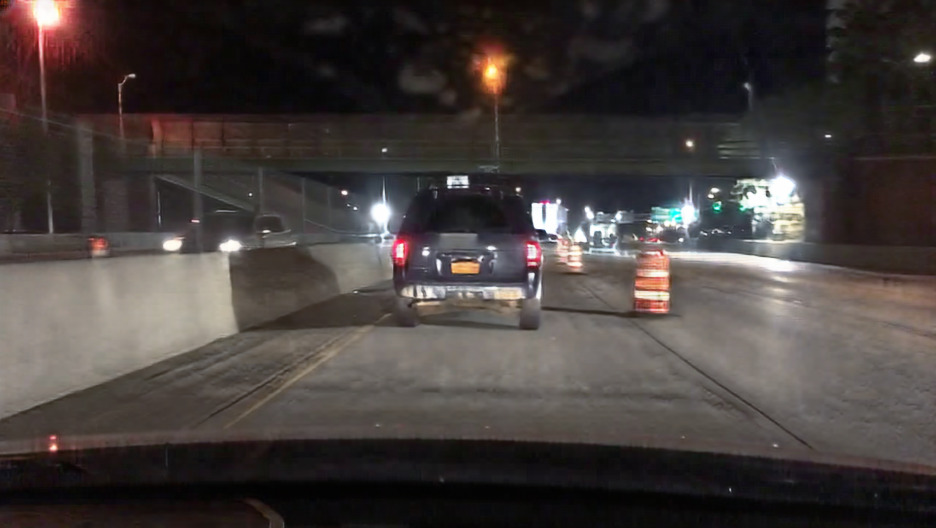}}\hfill
		\subfloat[FeaMGAN (ours)]
		{\includegraphics[width=0.165\textwidth]{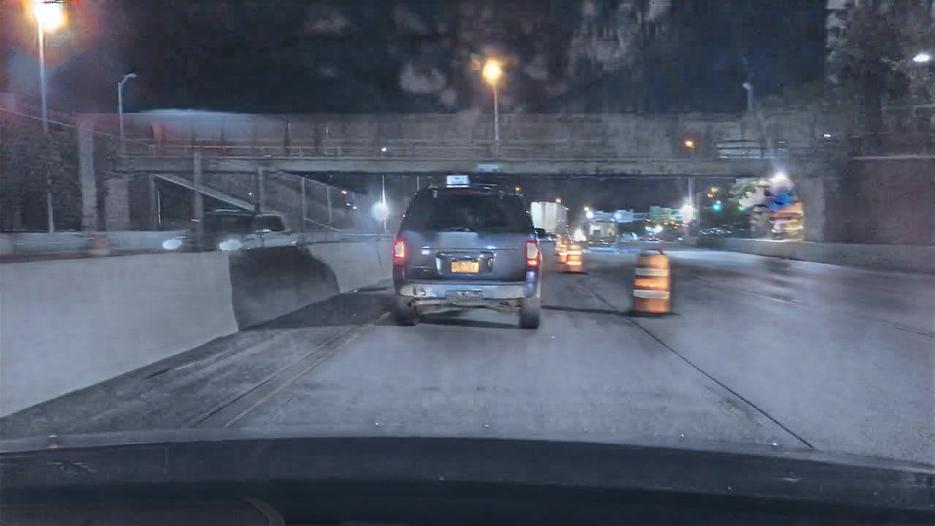}}\hfill
	\end{center}
	\vspace{-1ex}
	\caption[Qualitative comparison to prior work.]{Qualitative comparison to prior work. Models were trained using their official implementations. Randomly sampled results can be found in \autoref{fig:feamgan:app:qualitative_comparison_additional_random} of \autoref{app:03}. Best viewed in color.}
	\label{fig:feamgan:qualitative_comparison}
\end{figure}

\begin{table}[H]
	\RawFloats
	\scriptsize
	\caption[Quantitative comparison to prior work.]{Quantitative comparison to prior work. Models were trained using their official implementations. Results are reported as the average across five runs. We refer to \autoref{tab:feamgan:app:quantitative_comparison_extended} of \autoref{app:03} for an extended version of this table.}
	\vspace{-3ex}
	\label{tab:feamgan:quantitative_comparison}
	\begin{center}
		\setlength{\tabcolsep}{0.25em}
		\scalebox{0.905}{
			\begin{tabular}{lcccccccccccccccc}
				\toprule
				\multirow{2}{*}{Method}&\multicolumn{4}{c}{PFD$\rightarrow$Cityscapes}&\multicolumn{4}{c}{Viper$\rightarrow$Cityscapes}&\multicolumn{4}{c}{Day$\rightarrow$Night}&\multicolumn{4}{c}{Clear$\rightarrow$Snowy}\\
				\cmidrule(lr){2-5}
				\cmidrule(lr){6-9}
				\cmidrule(lr){10-13}
				\cmidrule(lr){14-17}
				& FID  & KID & sKVD& cKVD& FID  & KID & sKVD& cKVD& FID  & KID & sKVD& cKVD& FID  & KID & sKVD& cKVD\\
				\midrule
				Color Transfer
				& 91.01  & 94.82  & 18.16 & 50.87	
				& 89.30  & 83.51  & 20.20 & 51.23
				& 125.90 & 140.60 & 32.58 & 56.52
				& 46.85  & 19.44  & 14.91 & 42.89 \\				
				MUNIT
				& 40.36 & 29.98 & 14.99 & 43.24
				& 47.96 & 30.35 & 14.14 & 59.62
				& 42.53 & 31.83 & 15.02 & 50.83
				& \textbf{44.74} & 17.48 & 11.65 & 48.10 \\
				CUT
				& 49.55	& 44.25	& 16.85 & \textbf{37.53}
				& 60.35	& 49.48	& 16.80 & 51.02
				& \textbf{34.36} & \textbf{20.54}& 10.16 & 53.55
				& 46.03	& 15.70	& 14.71 & 43.91 \\					
				TSIT
				& \textbf{38.70} & \textbf{28.70} & \textbf{10.80} & 42.35
				& \textbf{45.26} & \textbf{28.40} &\textbf{8.47} & 50.03
				& 54.96 & 33.21 & 12.71 & 57.91
				& 79.28 & 40.02 & 12.97 & 41.52 \\	
				QS-Attn
				& 49.41 & 42.87 & 14.01 & 38.57
				& 55.62 & 39.31 & 12.99 & 63.22
				& 46.67 & 21.47 &  \textbf{7.58} & 52.02	
				& 60.91 & 18.85 & 14.19 & 44.00 \\	
				\midrule
				FeaMGAN-S (ours) 
				& 45.16	& 34.93 & 13.87 & 40.50
				& 52.79	& 35.92 & 14.34 & \textbf{45.38}
				& 70.40	& 51.30 & 14.68 & \textbf{46.66}
				& 57.93	& 16.24 & 11.88 & \textbf{38.28} \\	
				FeaMGAN (ours)
				& 46.12 & 36.56 & 13.69 & 41.19
				& 51.56 & 34.63 & 14.01 & \textbf{47.21}
				& 66.39 & 46.96 & 13.14 & \textbf{46.88} 
				& 56.78 & \textbf{14.77} & \textbf{11.36}  & 41.72 \\
				\bottomrule			
		\end{tabular}}
	\end{center}
	\vspace{-2ex}
\end{table}
\noindent \textbf{Comparisons to retrained baselines.}
We find that retraining the baselines with their original training setup for the PFD$\rightarrow$Cityscapes task significantly improves their performance on commonly used metrics compared to the baselines provided by EPE, as can be seen in \autoref{tab:feamgan:quantitative_comparison}. However, as shown in \autoref{fig:feamgan:qualitative_comparison} and the random results of \autoref{fig:feamgan:app:qualitative_comparison_additional_random} of \autoref{app:03}, content-consistency problems remain. This indicates again that simply relying on commonly used metrics does not provide a complete picture if content consistency is taken into account. When qualitatively comparing our model to the baselines for the PFD$\rightarrow$Cityscapes and Viper$\rightarrow$Cityscapes tasks in \autoref{fig:feamgan:qualitative_comparison}, we observe that our method significantly reduces content inconsistencies. However, a limitation of our masking strategy are class boundary artifacts, which are particularly evident in the Day$\rightarrow$Night translation task (\autoref{fig:feamgan:limitations}). Since masking allows our method to focus on specific classes, we achieve state-of-the-art performance for the Clear$\rightarrow$Snowy translation task.

\begin{figure}[t] 
	\captionsetup[subfigure]{labelformat=empty}
	\begin{center}	
		{\includegraphics[width=0.198\textwidth]{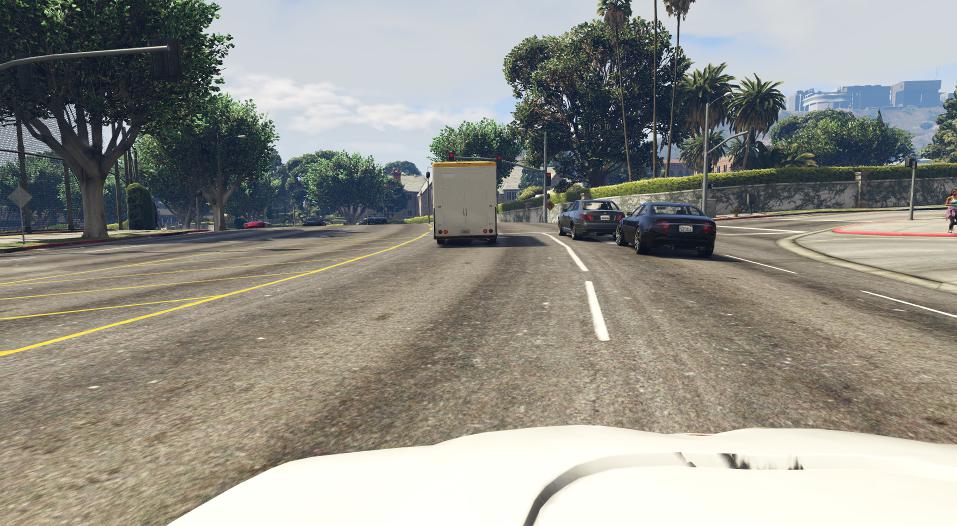}}\hfill
		{\includegraphics[width=0.198\textwidth]{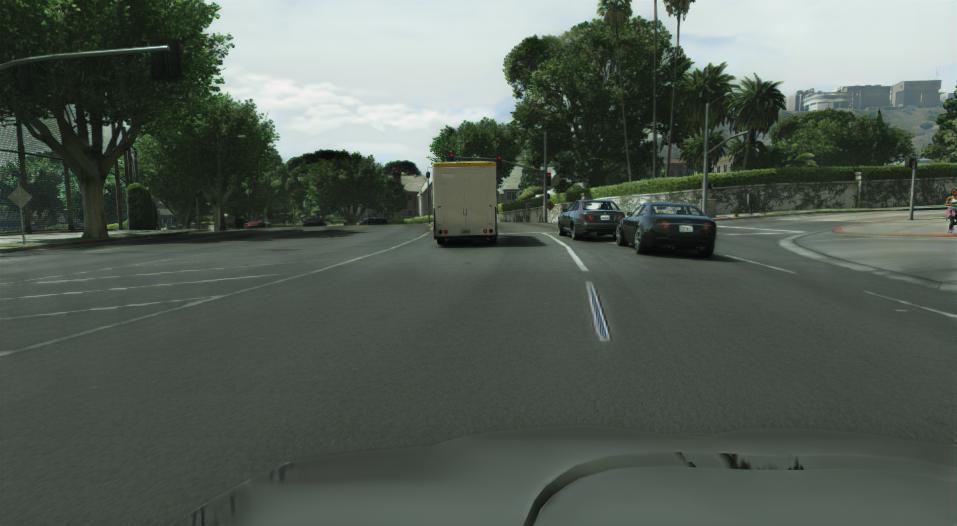}}\hfill
		{\includegraphics[width=0.198\textwidth]{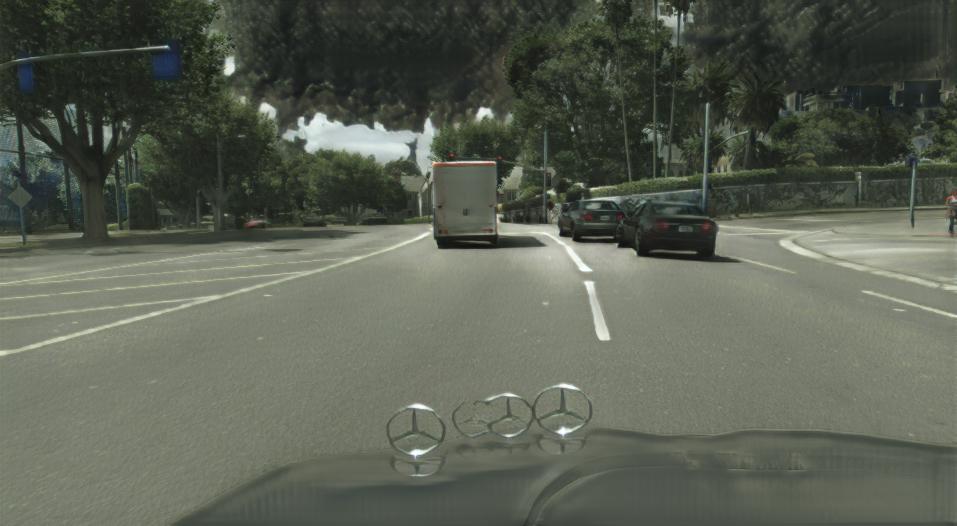}}\hfill
		{\includegraphics[width=0.198\textwidth]{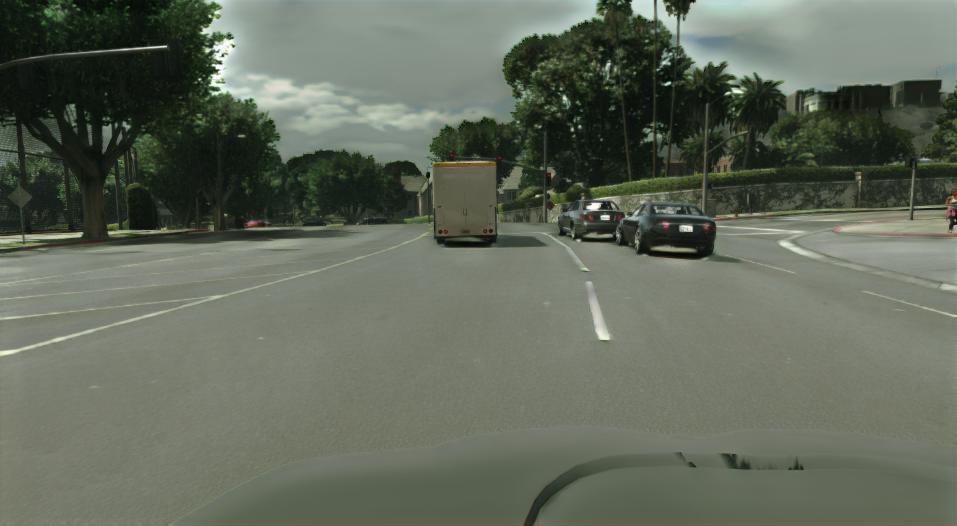}}\hfill
		{\includegraphics[width=0.198\textwidth]{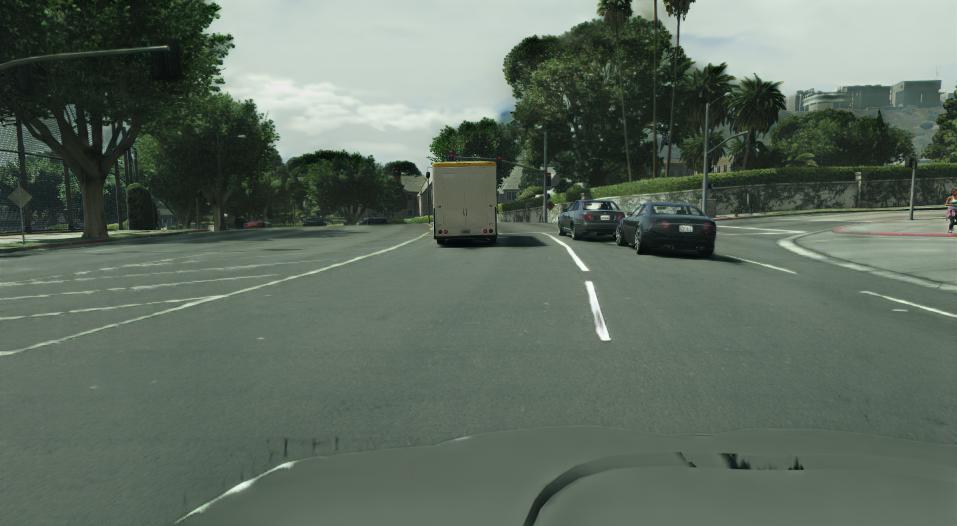}}\hfill \\ 
		{\includegraphics[width=0.198\textwidth]{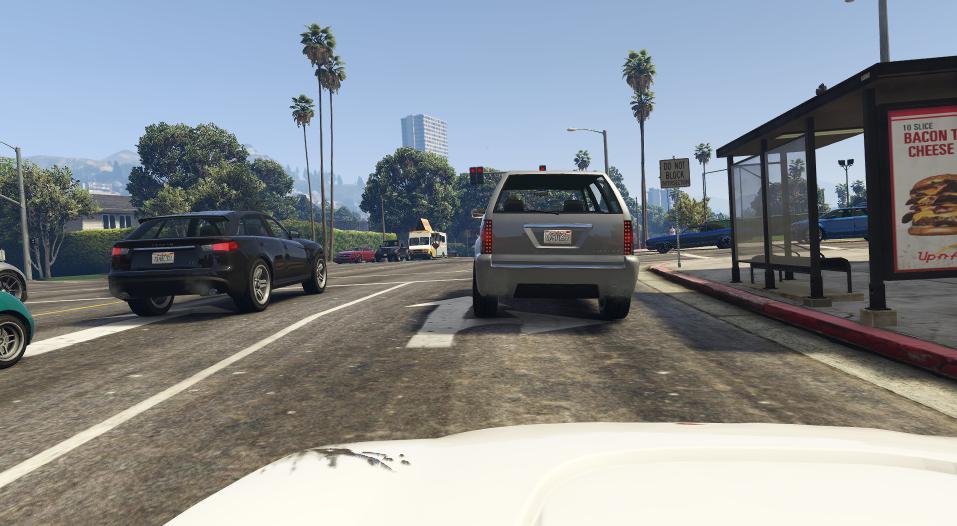}}\hfill
		{\includegraphics[width=0.198\textwidth]{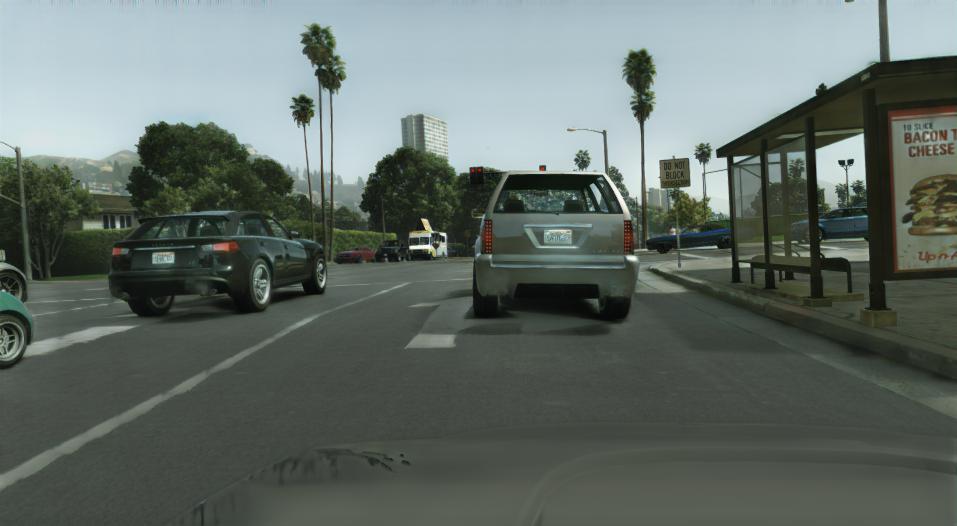}}\hfill
		{\includegraphics[width=0.198\textwidth]{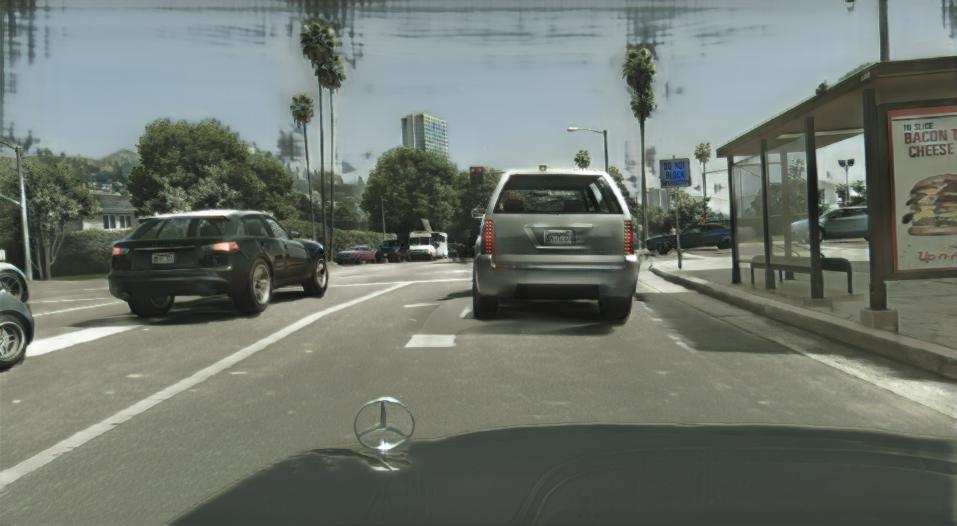}}\hfill
		{\includegraphics[width=0.198\textwidth]{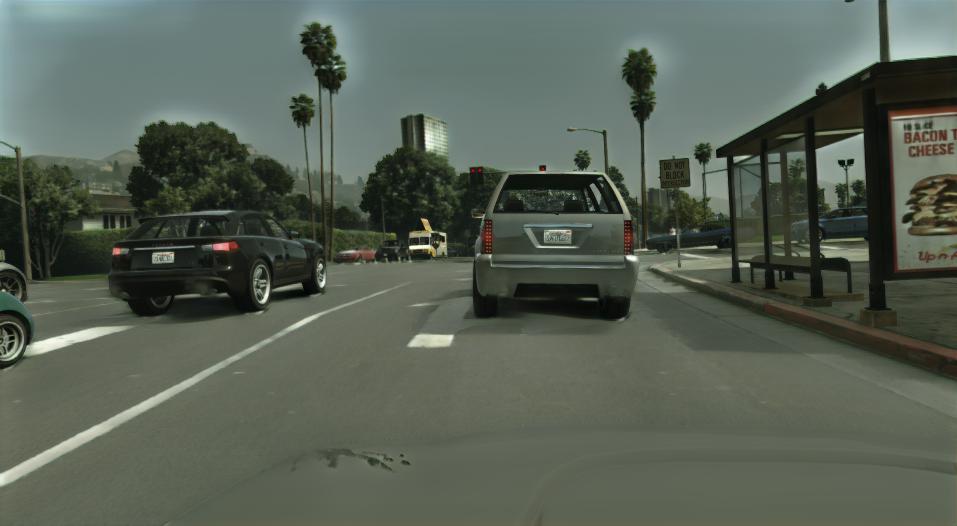}}\hfill
		{\includegraphics[width=0.198\textwidth]{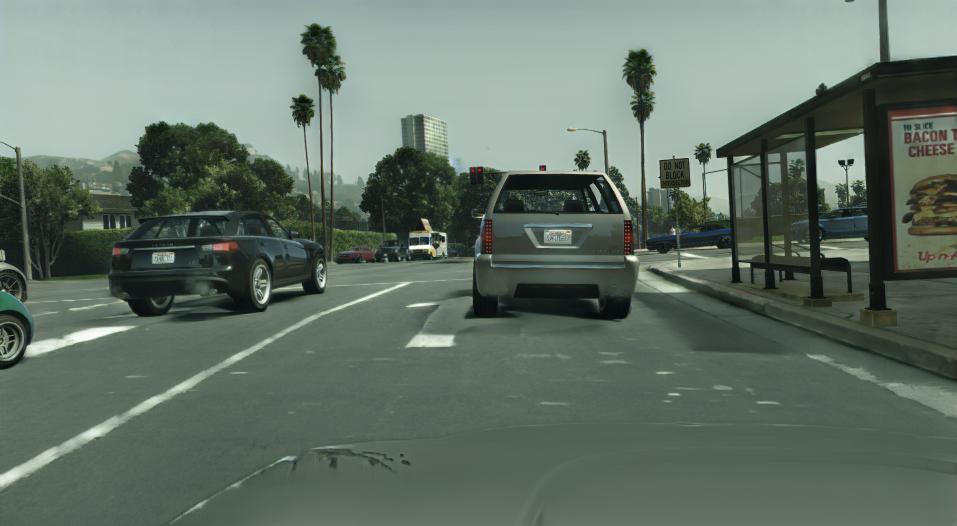}}\hfill \\ 
		\vspace{-10pt}
		\subfloat[Input]
		{\includegraphics[width=0.198\textwidth]{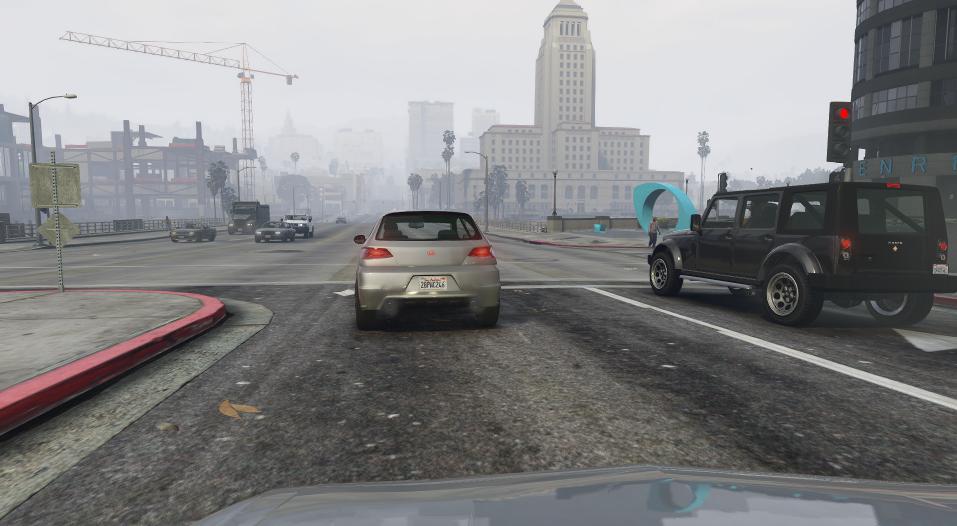}}\hfill
		\subfloat[Full]
		{\includegraphics[width=0.198\textwidth]{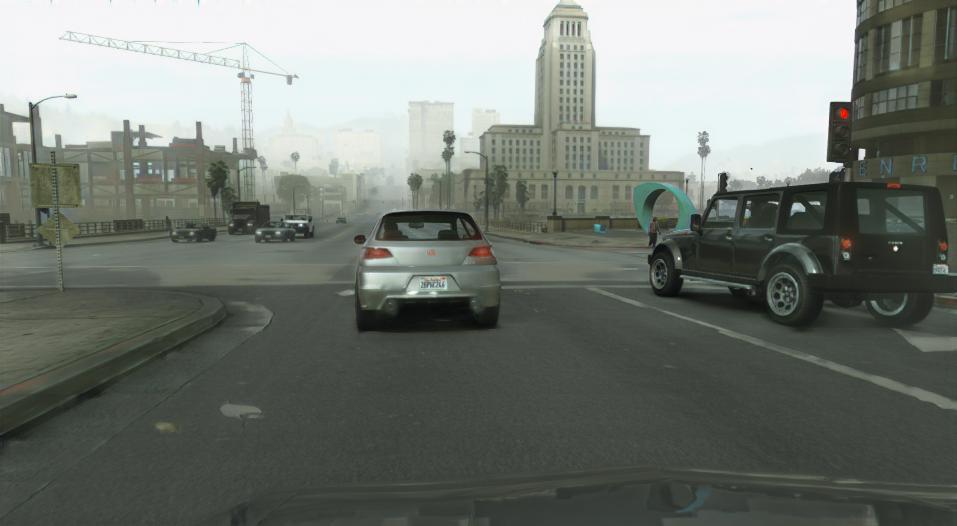}}\hfill
		\subfloat[w/o Dis. Mask]
		{\includegraphics[width=0.198\textwidth]{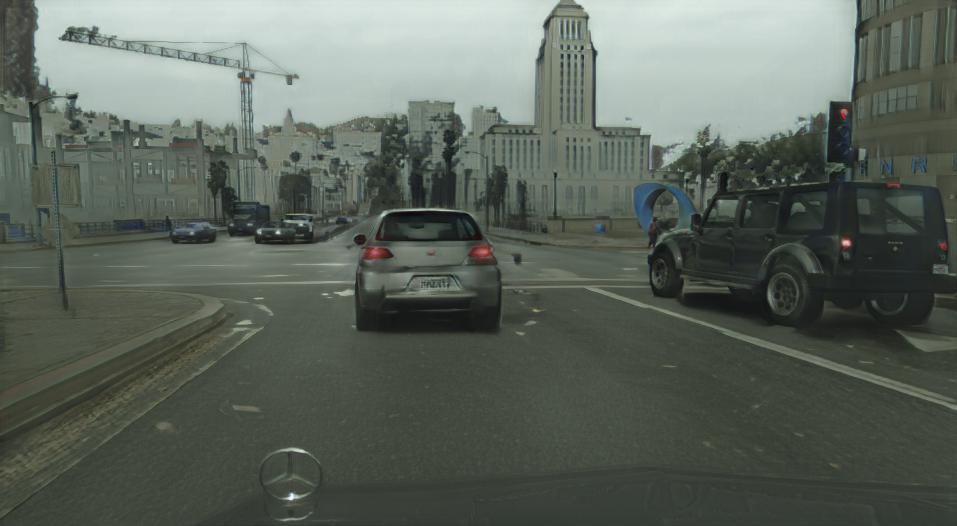}}\hfill
		\subfloat[w/o Local Dis.]
		{\includegraphics[width=0.198\textwidth]{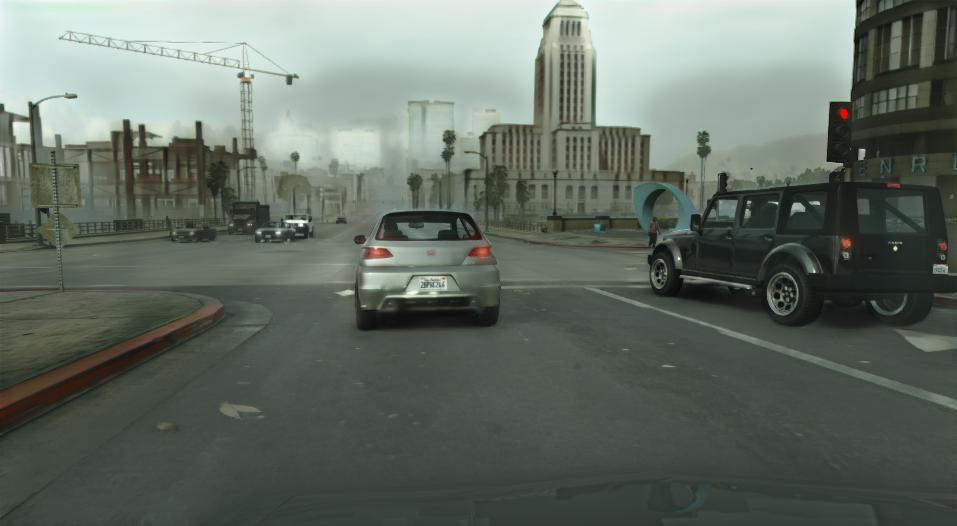}}\hfill
		\subfloat[w/ FADE w/o FATE ]
		{\includegraphics[width=0.198\textwidth]{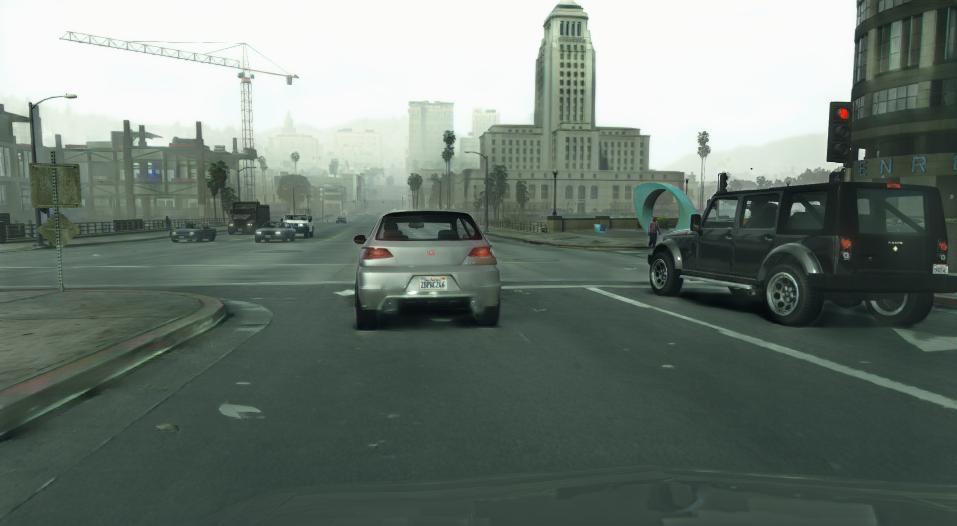}}\hfill 
	\end{center}
	\vspace{-1ex}
	\caption[Qualitative ablations.]{Qualitative ablations. Results are selected from the best model. Randomly sampled results can be found in \autoref{fig:feamgan:app:qualitative_ablations_additional_random} of \autoref{app:03}. Best viewed in color.}
	\label{fig:feamgan:qualitative_ablations}
\end{figure}

\begin{table}[t]
	\RawFloats
	\scriptsize
	\vspace{-3ex}
	\caption[Quantitative evaluation for ablation study.]{Quantitative evaluation for ablation study. Results are reported as the average across five runs. We refer to \autoref{tab:feamgan:app:quantitative_ablation_extended} of \autoref{app:03} for an extended version of this table.}
	\label{tab:feamgan:quantitative_ablation}
	\begin{center}
		\setlength{\tabcolsep}{0.25em}
		\scalebox{0.86}{
			\begin{tabular}{lcccccccccccccc}
				\toprule
				\multirow{2}{*}{Method}&\multirow{2}{*}{FID}&\multirow{2}{*}{KID} &\multirow{2}{*}{sKVD}&\multicolumn{11}{c}{cKVD}\\
				\cmidrule(lr){5-15}
				&    &  &  & AVG& 		sky& 	ground&	road&	terrain&	vegetation&	building&	roadside-obj.&	person&	vehicle&	rest	\\
				\midrule	
				FeaMGan (Full)& 46.12 &	36.56 & 13.69 & 41.19 &42.69	&14.97	&17.35	&26.51	&\textbf{20.25}	&26.34	&64.64	&102.23	&\textbf{42.38}	&54.52 \\
				w/o Dis. Mask  & \textbf{37.10} & \textbf{25.88} & 14.73 &\textbf{39.65}	&\textbf{26.70}	&15.81	&16.65	&31.02	&22.97	&\textbf{25.39}	&67.01	&\textbf{93.78}	&44.23	&\textbf{52.91}
				\\
				w/ FADE w/o FATE & 45.46 & 35.73 & \textbf{13.17} &40.90	&41.49	&13.78	&16.78	&25.30	&20.58	&27.21	&\textbf{63.12}	&104.43	&42.44	&53.83\\
				w/ Random Crop & 47.88	&38.48&	13.37	& 40.18	&39.88	&\textbf{12.90}	&\textbf{14.65}	&\textbf{25.09}	&21.89	&27.32	&64.32	&98.81	&43.08	&53.86 \\
				w/ VGG Crop& 51.23&	42.46&	13.56 & 40.62	&40.32	&13.38	&15.67	&26.47	&21.09	&27.28	&65.23	&99.61	&43.19	&53.94 \\
				\midrule
				w/o Local Dis. &   &  &  &  &  &  &   &  &  &  &    &  &  &    \\
				- w/ 256$\times$256 Crop& 48.57 & 38.89 & \textbf{12.89} & 41.26 &	42.31 &	13.57	& 15.98 &	\textbf{25.28} &	22.18	& \textbf{26.56} &	\textbf{61.13} &	107.48 & 42.44&	55.62\\
				- w/ 352$\times$352 Crop& 47.26 & 37.75 & 14.38 & 39.30 &34.44	&\textbf{13.09}	&15.84& 25.83	&21.50	&27.20	&61.24	&98.25	&42.24	&53.38\\
				- w/ 464$\times$464 Crop& \textbf{46.61} & \textbf{37.25} & 15.04 &\textbf{38.62}	&\textbf{31.60}	&13.13	&\textbf{15.38}	&27.06	&22.23	&29.67	&63.38	&\textbf{87.51}	&44.41	&51.77\\
				- w/ 512$\times$512 Crop& 55.89	& 49.12 & 15.94 & 39.35	&36.48 &14.68 & 16.06	&26.87	&\textbf{19.61}	&27.37	&62.40	&98.90	&\textbf{40.32}	&\textbf{50.86}
				\\	
				\bottomrule
		\end{tabular}}
	\end{center}
	\vspace{-2ex}
\end{table}

\subsection{Ablation Study}
\noindent \textbf{Effectiveness of masked discriminator.}
As shown in \autoref{fig:feamgan:qualitative_ablations} and the random samples in \autoref{fig:feamgan:app:qualitative_ablations_additional_random} of \autoref{app:03}, our masking strategy for the discriminator positively impacts content consistency. Without masking, inconsistencies occur that correlate with biases between the class distributions of the source and target domains. As shown in \cite{richter2022enhancing}, the distributions of certain classes in the spatial image dimension vary greatly between the PFD dataset and the Cityscapes dataset. For example, trees in Cityscapes appear more frequently in the top half of the image, resulting in hallucinated trees when the images are translated without accounting for biases. In the first and second row of \autoref{fig:feamgan:qualitative_ablations}, we show that our masking strategy (Full) prevents these inconsistencies in contrast to our model trained without masking (w/o Dis. Mask). However, as shown in \autoref{tab:feamgan:quantitative_ablation}, this comes with a quantitative tradeoff in performance on commonly used metrics.\\

\noindent \textbf{Effectiveness of local discriminator.}
We compare our model trained with a local discriminator (Full) to the model trained without a local discriminator (w/o Local Dis. 352x352). As shown in \autoref{fig:feamgan:qualitative_ablations}, the local discriminator leads to an increase in quantitative performance. Furthermore, we show the qualitative effects of the local discriminator in \autoref{fig:feamgan:qualitative_ablations}, where we observe a decrease in glowing objects and a significant decrease of erased objects in the translation. An example of a glowing object is the palm tree in row two of \autoref{fig:feamgan:qualitative_ablations}. An example of erased objects are the missing houses in the background of the images from row three. In addition, small inconsistencies near object boundaries are reduced, as shown by the randomly sampled results in \autoref{fig:feamgan:app:qualitative_ablations_additional_random} of \autoref{app:03} (e.g., the wheels of the car in row one and three). Overall, we can conclude that local discriminators can reduce local inconsistencies, which might arise from the robust but not flawless segmentation maps used for masking.\\

\noindent \textbf{Effectiveness of segmentation-based sampling.}
We compare our segmentation-based sampling method with random sampling and sampling based on VGG features. For the sampling strategy based on VGG features, we follow EPE \cite{richter2022enhancing} to calculate scores for 352$\times$352 crops of the input images. Crops with a similarity score higher than $0.5$ are selected for training. As shown in \autoref{tab:feamgan:quantitative_ablation}, our segmentation-based sampling strategy (Full) slightly outperforms the other sampling strategies in overall translation performance.\\
\begin{figure}[t] 
	\captionsetup[subfigure]{labelformat=empty}
	\begin{center}
		{\includegraphics[width=0.164\textwidth]{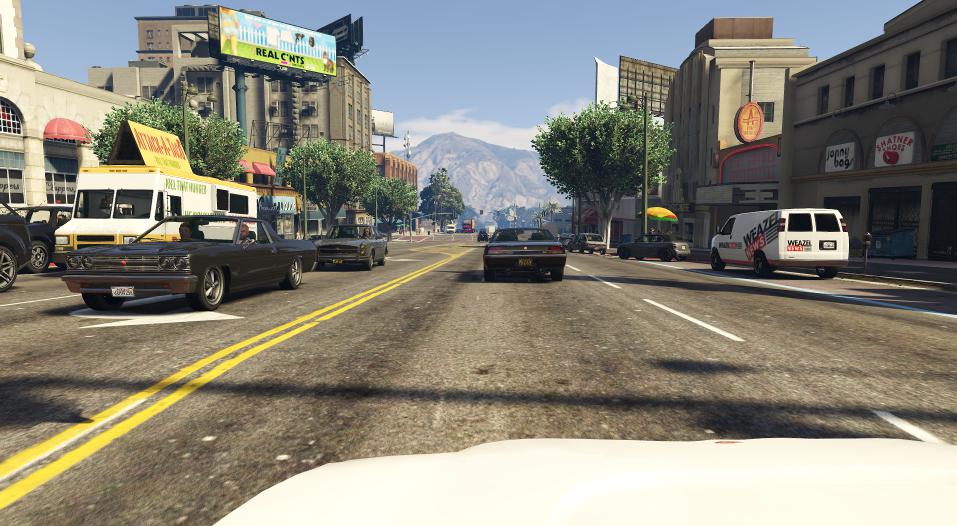}}\hfill
		{\includegraphics[width=0.164\textwidth]{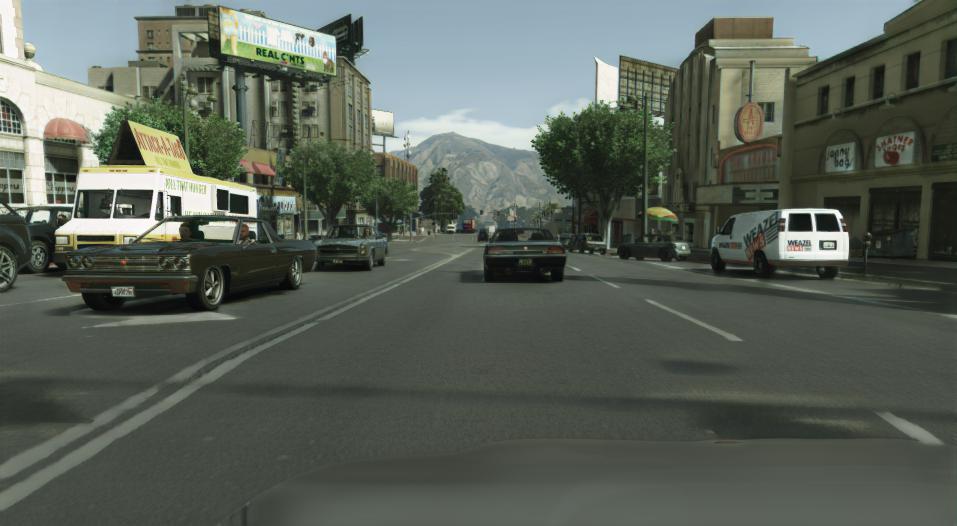}}\hfill
		{\includegraphics[width=0.164\textwidth]{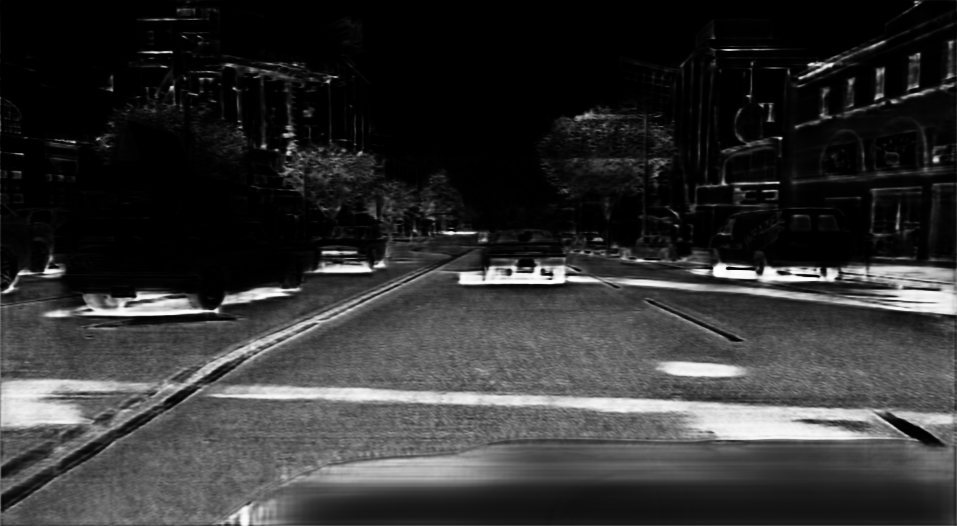}}\hfill
		{\includegraphics[width=0.164\textwidth]{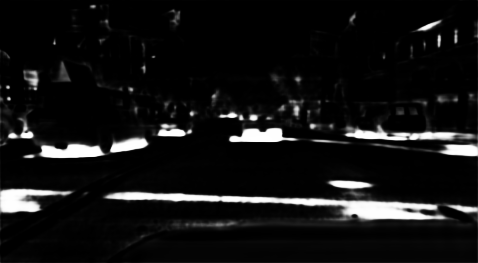}}\hfill
		{\includegraphics[width=0.164\textwidth]{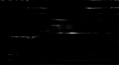}}\hfill
		{\includegraphics[width=0.164\textwidth]{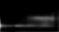}}\hfill \\ 
		{\includegraphics[width=0.164\textwidth]{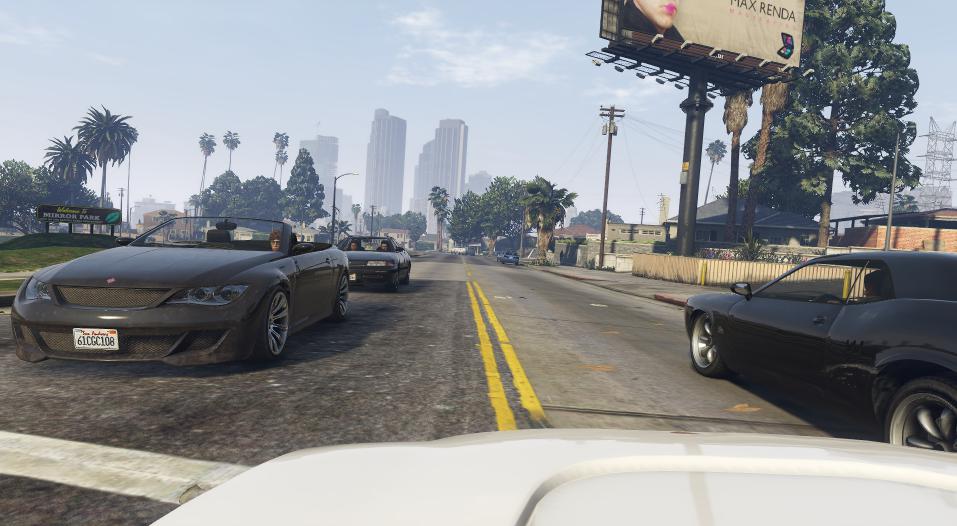}}\hfill
		{\includegraphics[width=0.164\textwidth]{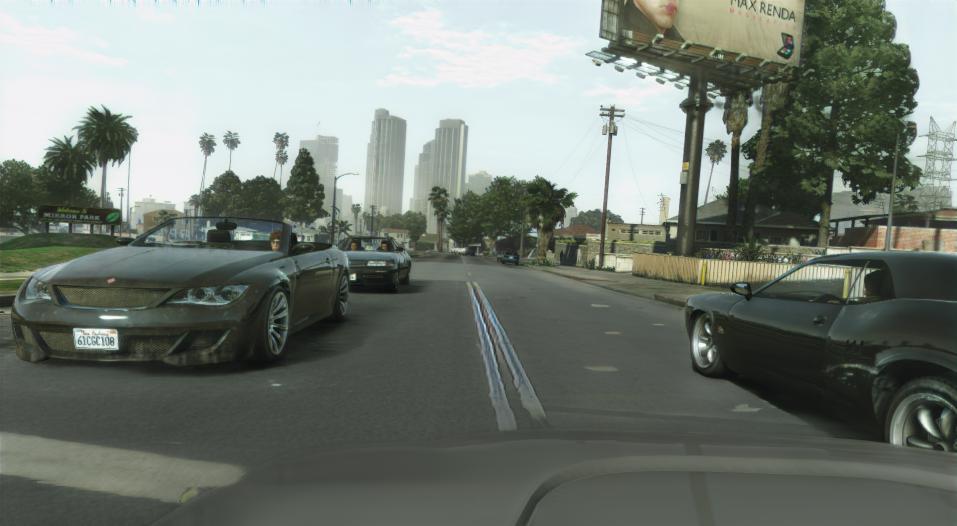}}\hfill
		{\includegraphics[width=0.164\textwidth]{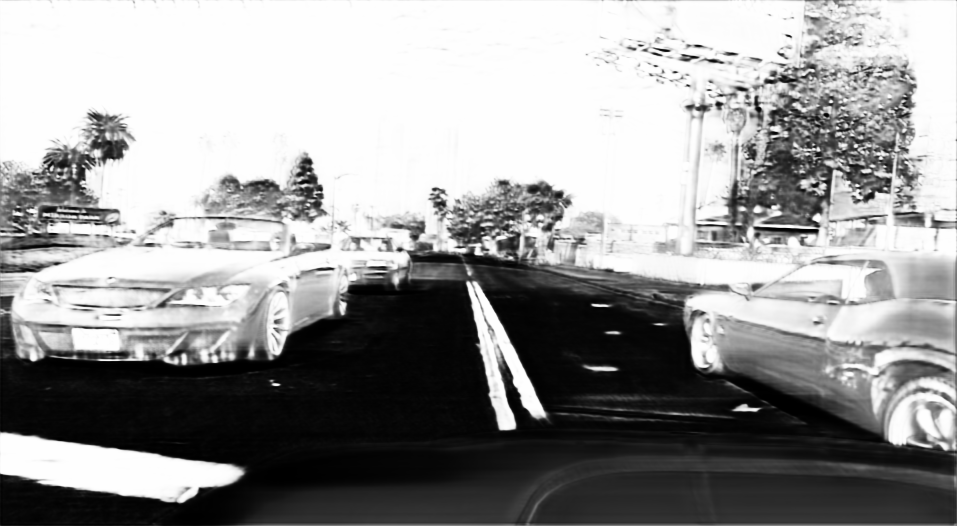}}\hfill
		{\includegraphics[width=0.164\textwidth]{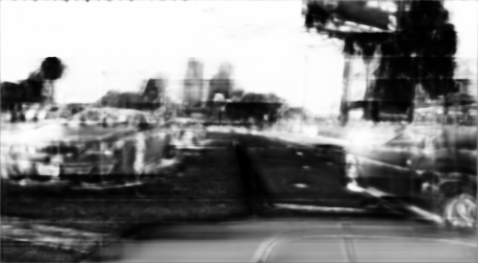}}\hfill
		{\includegraphics[width=0.164\textwidth]{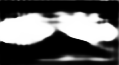}}\hfill
		{\includegraphics[width=0.164\textwidth]{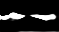}}\hfill \\ 
		\vspace{-10pt}
		\subfloat[957$\times$526 Input]
		{\includegraphics[width=0.164\textwidth]{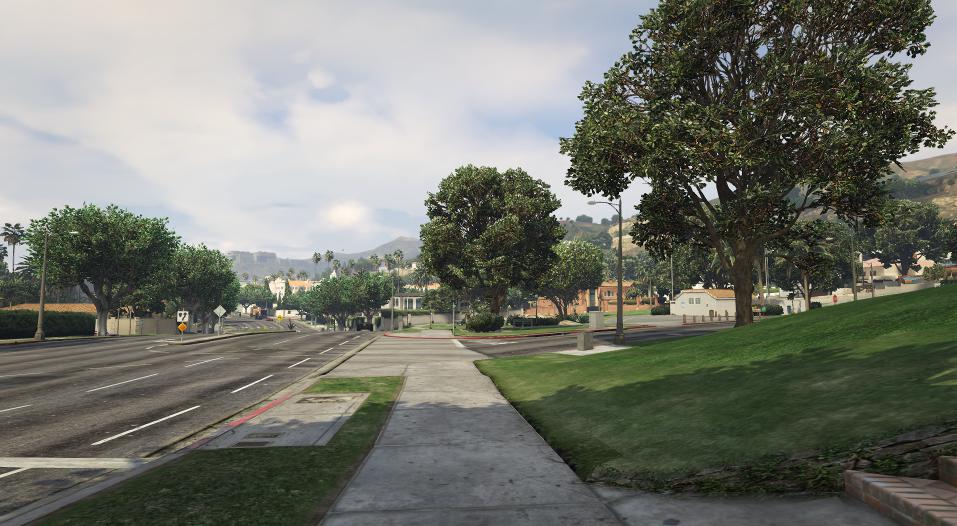}}\hfill
		\subfloat[957$\times$526 Output]
		{\includegraphics[width=0.164\textwidth]{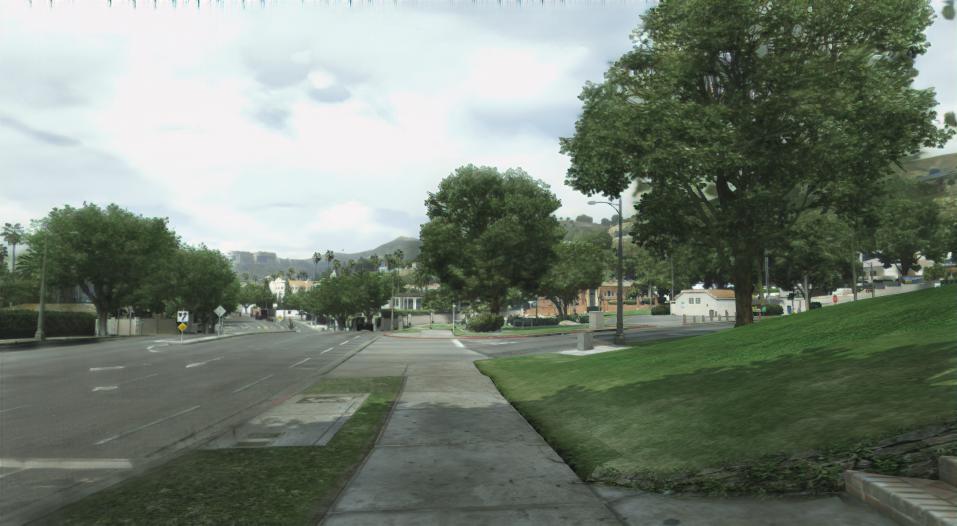}}\hfill
		\subfloat[957$\times$526 Layer]
		{\includegraphics[width=0.164\textwidth]{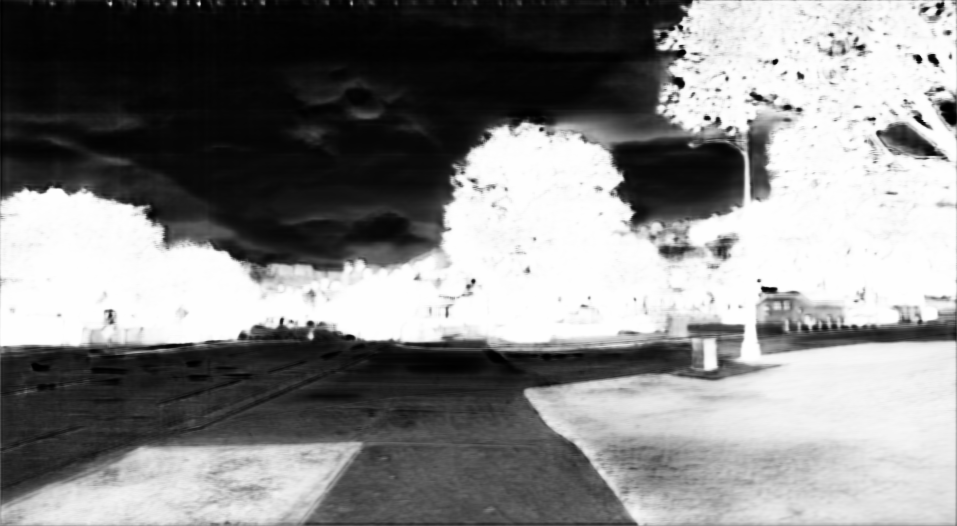}}\hfill
		\subfloat[478$\times$263 Layer]
		{\includegraphics[width=0.164\textwidth]{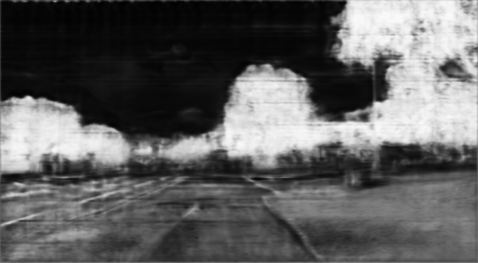}}\hfill
		\subfloat[119$\times$65 Layer]
		{\includegraphics[width=0.164\textwidth]{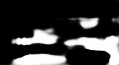}}\hfill
		\subfloat[59$\times$32 Layer]
		{\includegraphics[width=0.164\textwidth]{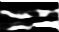}}\hfill 
	\end{center}
	\vspace{-1ex}
	\caption[FATE attention maps.]{FATE attention maps. Results are selected from the best model. Best viewed in color.}
	\label{fig:feamgan:FATE_attention_maps}
\end{figure}

\noindent \textbf{Effectiveness of FATE.}
For each spatial point ("pixel") in the input feature map, our feature-attentive denormalization block selects the features in the feature dimension to be incorporated into the output stream of the generator by denormalization. We show the attention values of our feature-attentive denormalization block in \autoref{fig:feamgan:FATE_attention_maps} by visualizing all attention values for a single feature across the entire feature map. Since a single feature represents a property of the input, a spatial pattern should emerge. This is expected especially in earlier layers, where the spatiality of the convolutional model's feature map is best preserved. As shown in \autoref{fig:feamgan:FATE_attention_maps}, our attention mechanism learns to attend to features that correlate with a property. Examples are the shadows of a scene (row 1), cars and their lighting (row 2), and vegetation (row 3). In addition, we find increasingly more white feature maps in deeper layers. This can be interpreted positively as an indication that the learned content (source) features in deeper layers are important for the translation task and that more shallow content features of earlier layers are increasingly ignored. However, this can also be interpreted negatively and could indicate that our simple attention mechanism is not able to separate deeper features properly.

Comparing FATE to FADE, we find that FATE leads to a subtle increase in training instability, resulting in slightly worse average performance over the five runs per model. However, FATE also leads to our best models. Therefore, we select the FATE block as the standard configuration for our model. The deviation from the average values for all runs can be found in \autoref{tab:feamgan:app:quantitative_ablation_extended} of \autoref{app:03}. The slight increased instability suggests that the attention mechanism of FATE can be further improved. \\
\begin{figure}[h] 
	\captionsetup[subfigure]{labelformat=empty}
	\begin{center}
		{\includegraphics[width=0.198\textwidth]{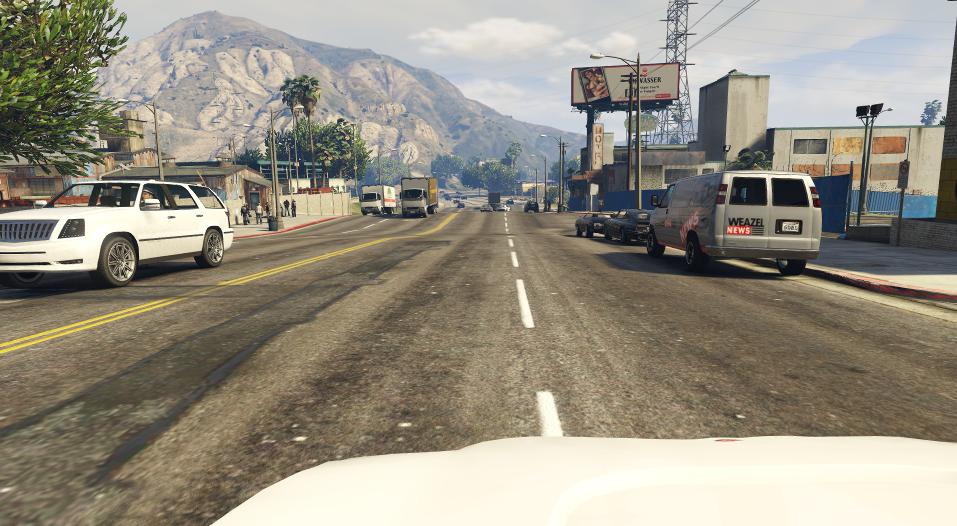}}\hfill
		{\includegraphics[width=0.198\textwidth]{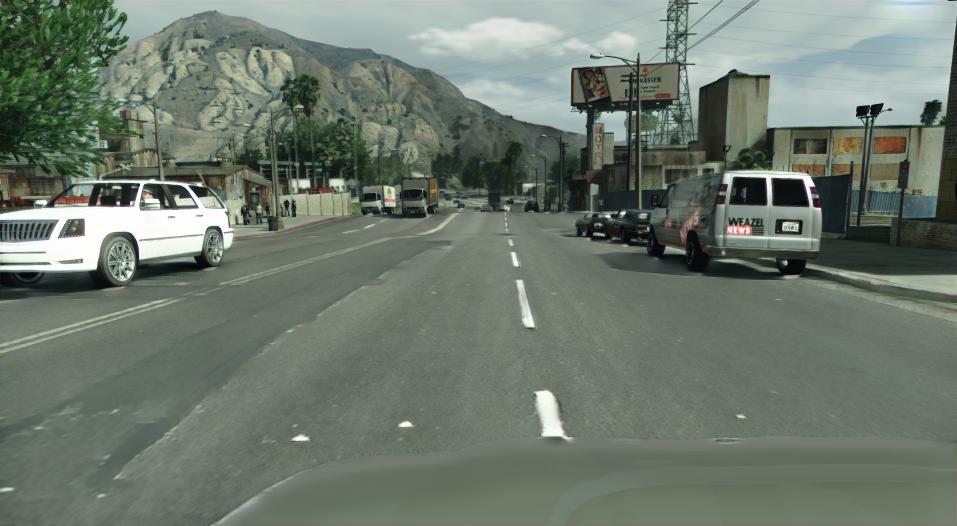}}\hfill
		{\includegraphics[width=0.198\textwidth]{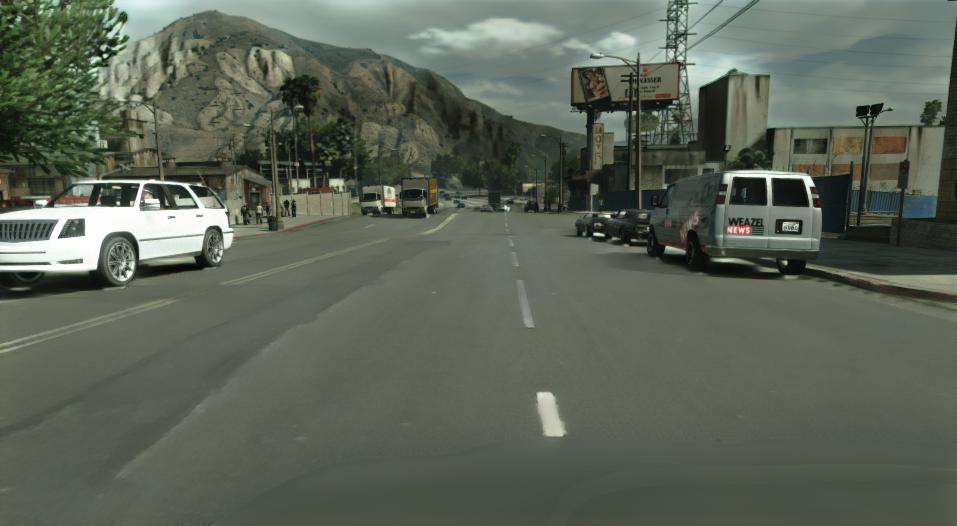}}\hfill
		{\includegraphics[width=0.198\textwidth]{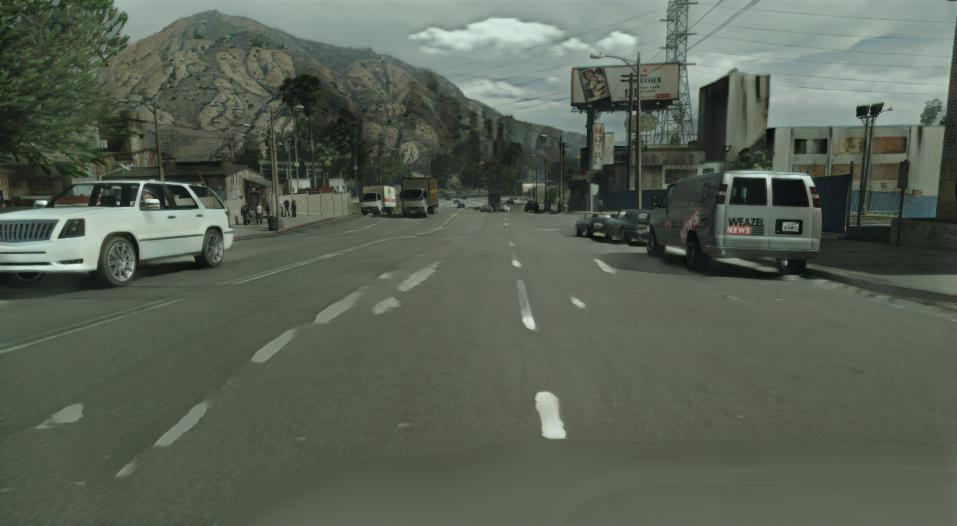}}\hfill
		{\includegraphics[width=0.198\textwidth]{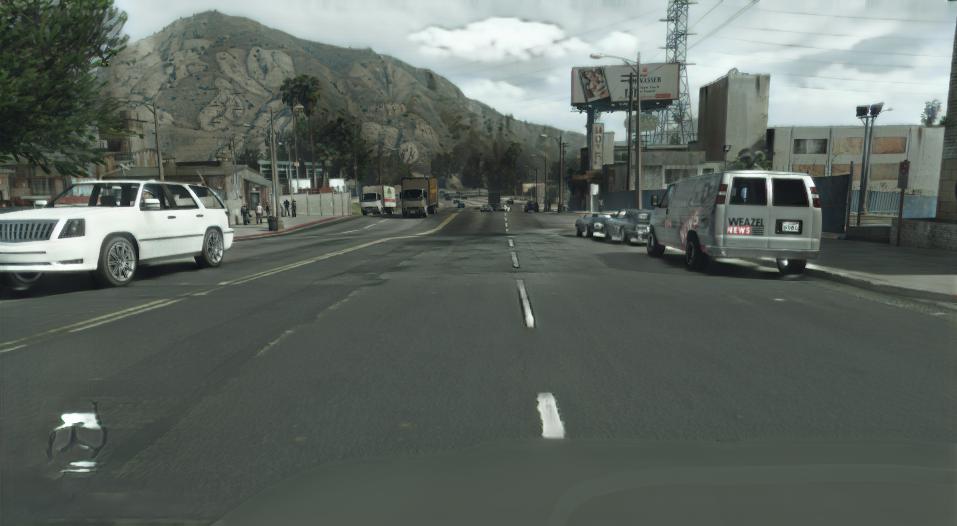}}\hfill \\ 
		{\includegraphics[width=0.198\textwidth]{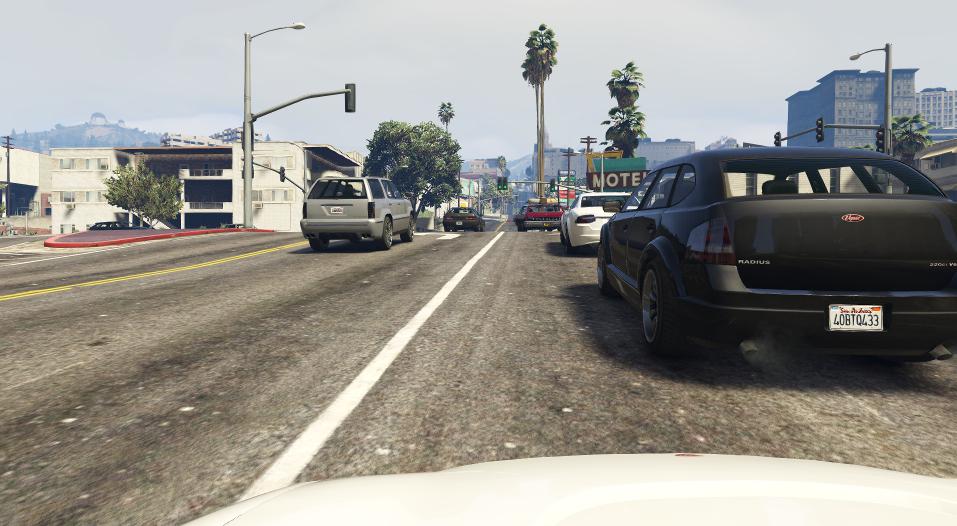}}\hfill 
		{\includegraphics[width=0.198\textwidth]{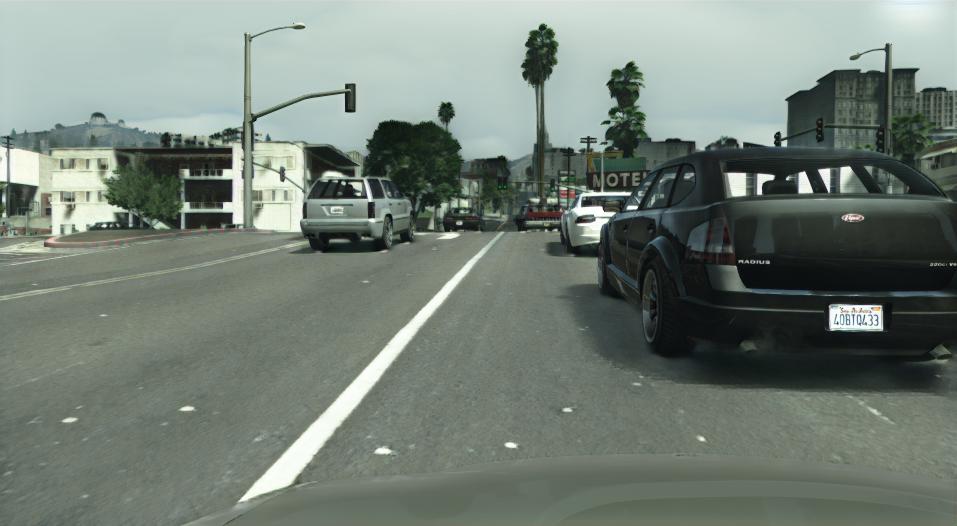}}\hfill
		{\includegraphics[width=0.198\textwidth]{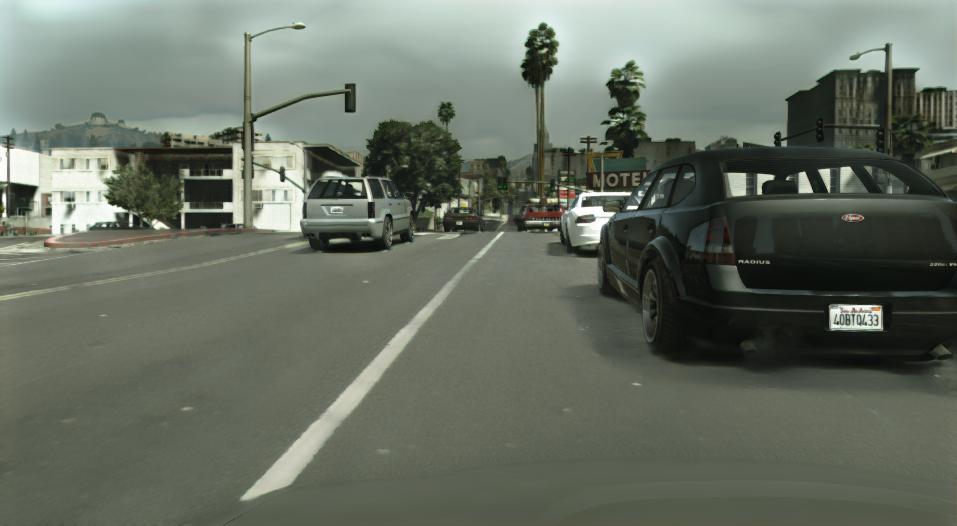}}\hfill
		{\includegraphics[width=0.198\textwidth]{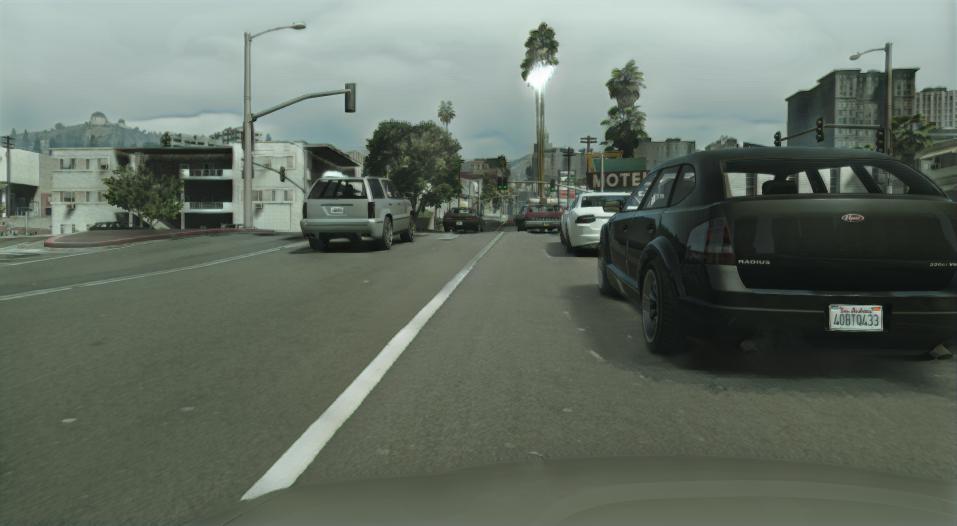}}\hfill
		{\includegraphics[width=0.198\textwidth]{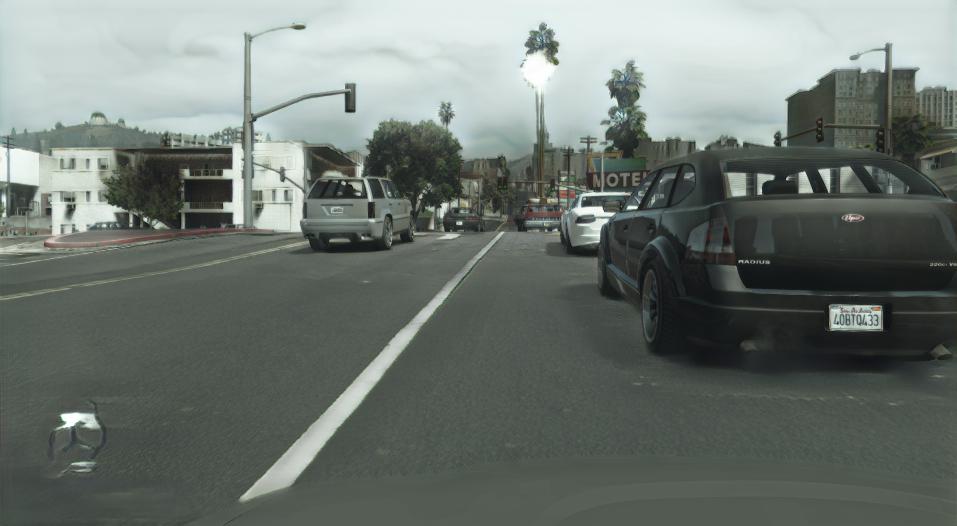}}\hfill \\ 
		\vspace{-10pt}
		\subfloat[Input]
		{\includegraphics[width=0.198\textwidth]{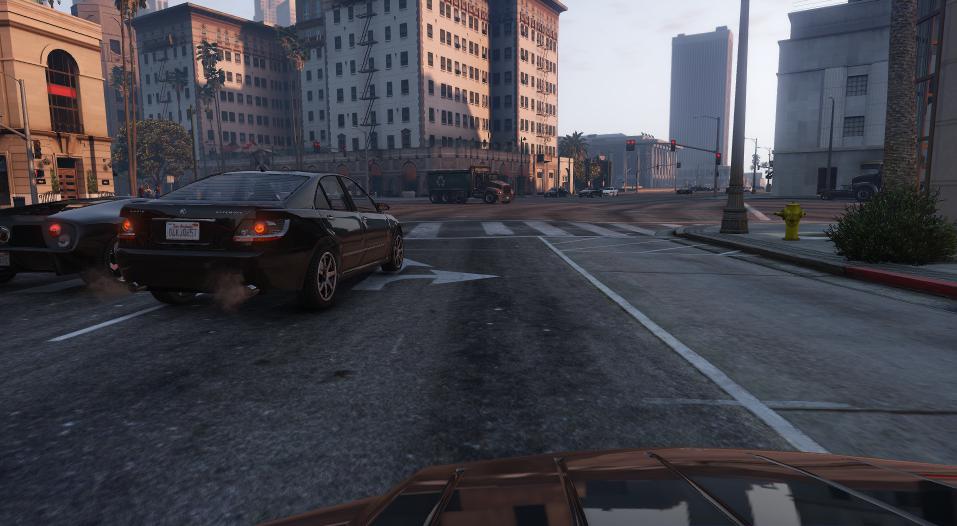}}\hfill
		\subfloat[252$\times$252]
		{\includegraphics[width=0.198\textwidth]{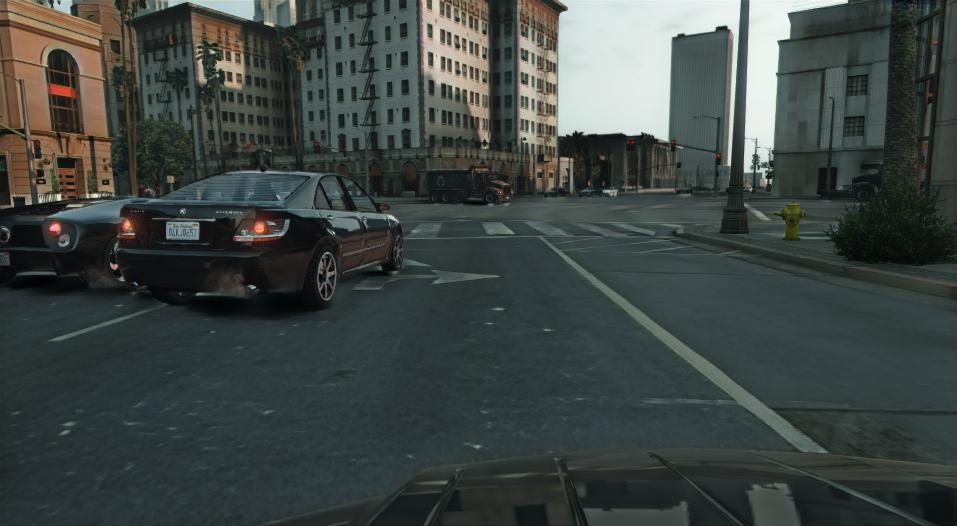}}\hfill
		\subfloat[352$\times$352]
		{\includegraphics[width=0.198\textwidth]{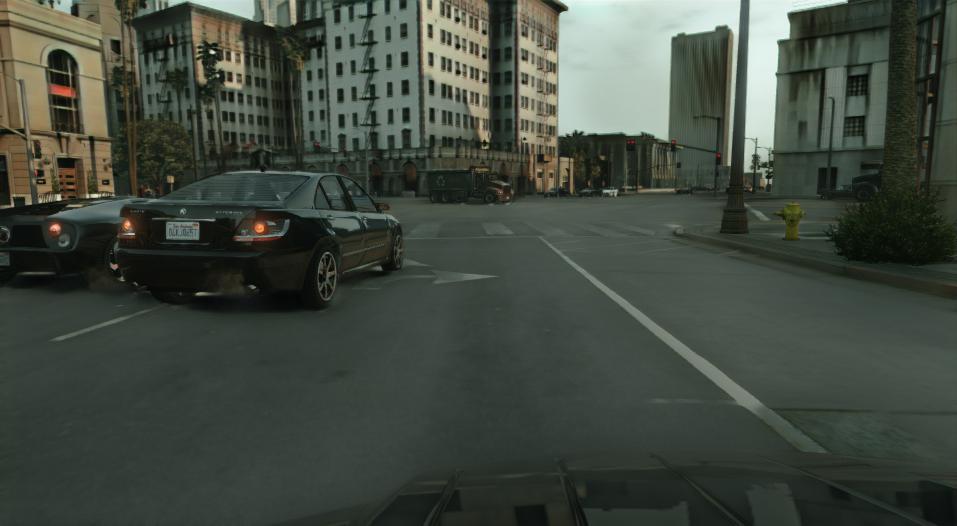}}\hfill
		\subfloat[464$\times$464]
		{\includegraphics[width=0.198\textwidth]{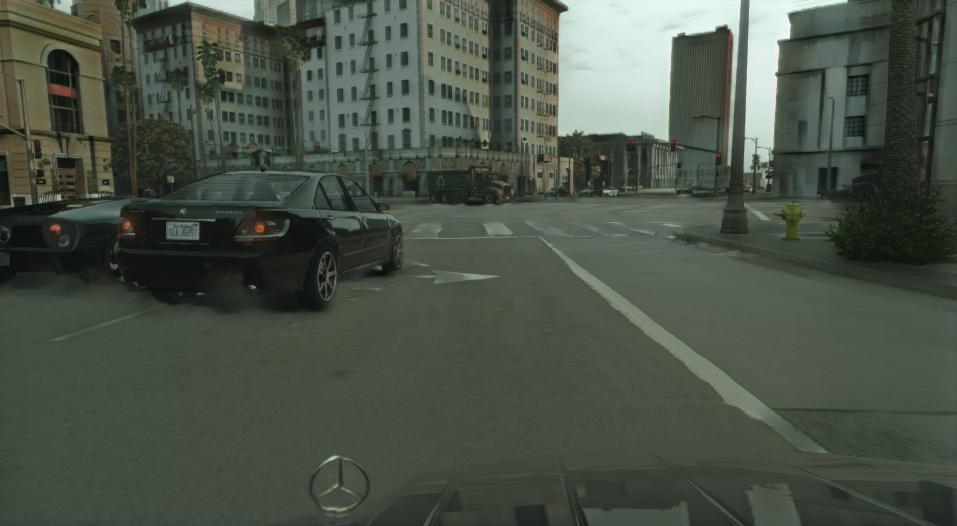}}\hfill
		\subfloat[512$\times$512]
		{\includegraphics[width=0.198\textwidth]{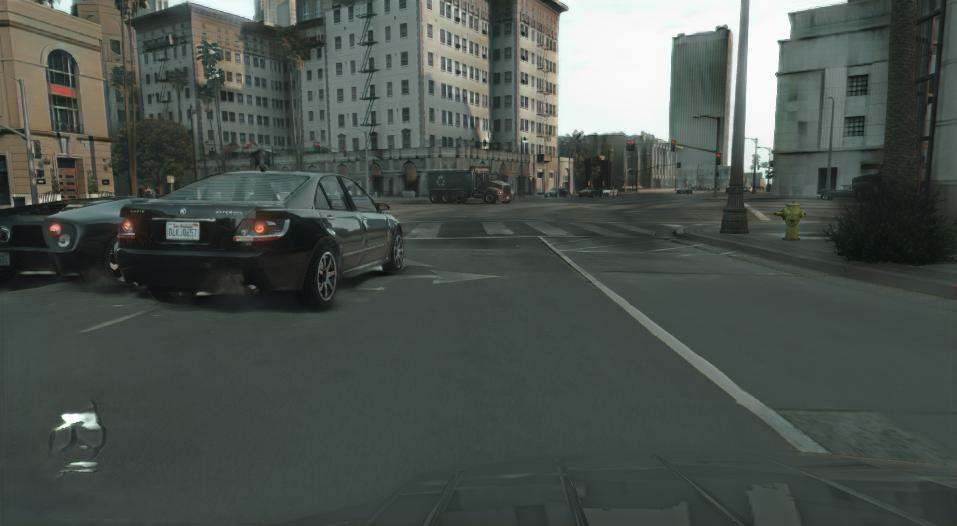}}\hfill 
	\end{center}
	\vspace{-1ex}
	\caption[Qualitative ablation of crop sizes.]{Qualitative ablation of crop sizes. For each crop size, results are selected from the best model. Randomly sampled results can be found in \autoref{fig:feamgan:app:qualitative_ablation_crop_size_additional_random} of \autoref{app:03}. Best viewed in color.}
	\label{fig:feamgan:qualitative_ablation_crop_size}
\end{figure}
\begin{figure}[h]
	\begin{center}
		\includegraphics[width=\linewidth]{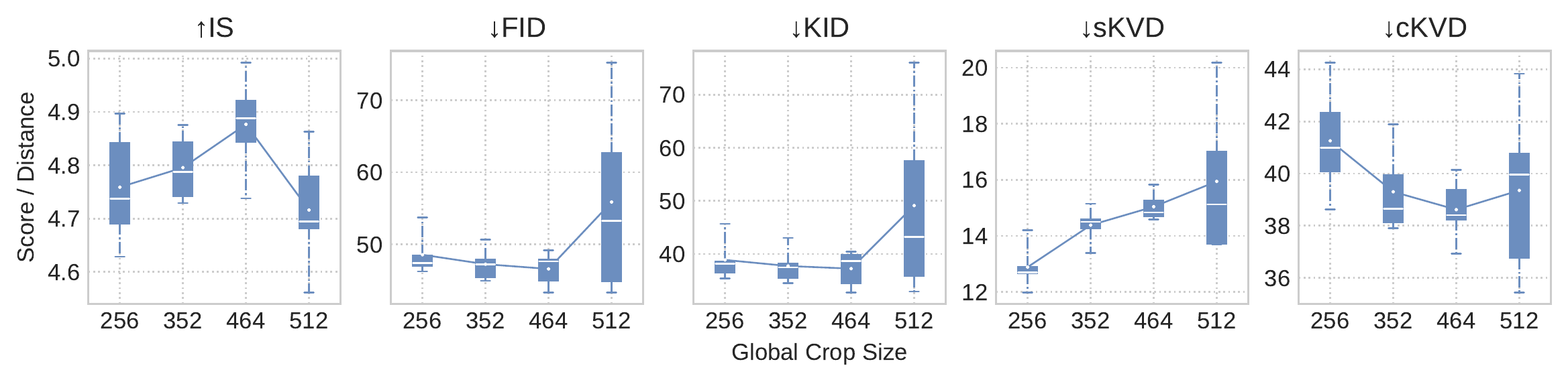} 
	\end{center}
	\vspace{-1ex}
	\caption[Quantitative ablation of crop sizes.]{Quantitative ablation of crop sizes.}
	\label{fig:feamgan:quantitative_ablation_crop_size}
\end{figure} 

\noindent \textbf{Effect of global crop size.}
We successively increase the global crop size of the generator and discriminators from 256$\times$256 to 512$\times$512 and examine the effects on translation performance. As shown in \autoref{fig:feamgan:qualitative_ablations}, increasing the global crop size results in a better approximation of the target domain style. However, increasing the global crop size also leads to an increasing number of artifacts in the translated image. In \autoref{fig:feamgan:quantitative_ablation_crop_size}, we report the score of various metrics with respect to the global crop size. The commonly used metrics for measuring translation quality (IS, FID, and KID) show that translation quality increases steadily up to a global crop size of 464$\times$464, after which the results become unstable. The cKVD metric also shows an increase in average performance up to a crop size of 464$\times$464, mainly because translation quality for the underrepresented person class increases. This is intuitive since a larger crop size leads to a more frequent appearance of underrepresented classes during training. Furthermore, the sKVD metric shows a steady decline in consistency as the global crop size increases. Therefore, we choose a tradeoff between approximation of the target domain style, artifacts, and computational cost, and select 352$\times$352 as the global crop size for our model.
\begin{figure}[h] 
	\captionsetup[subfigure]{labelformat=empty}
	\begin{center}
		{\includegraphics[width=0.249\textwidth]{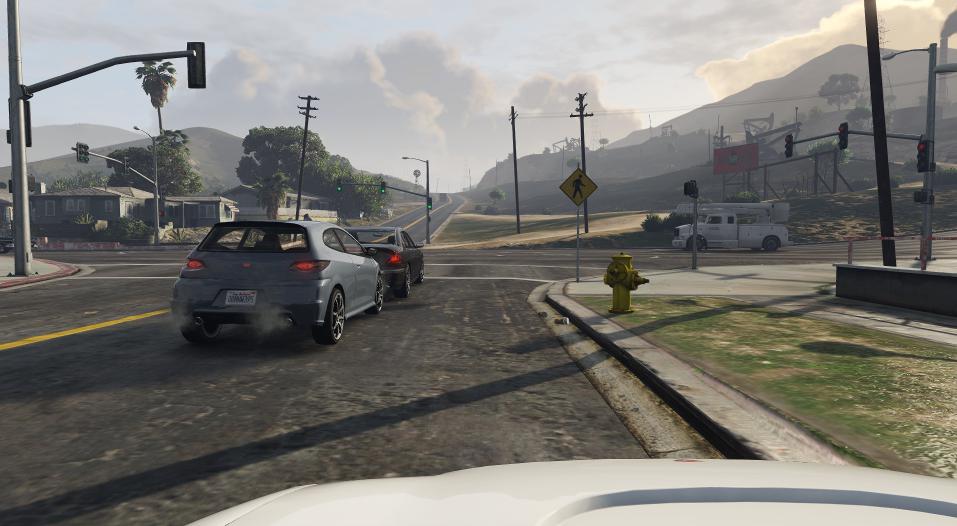}{\includegraphics[width=0.249\textwidth]{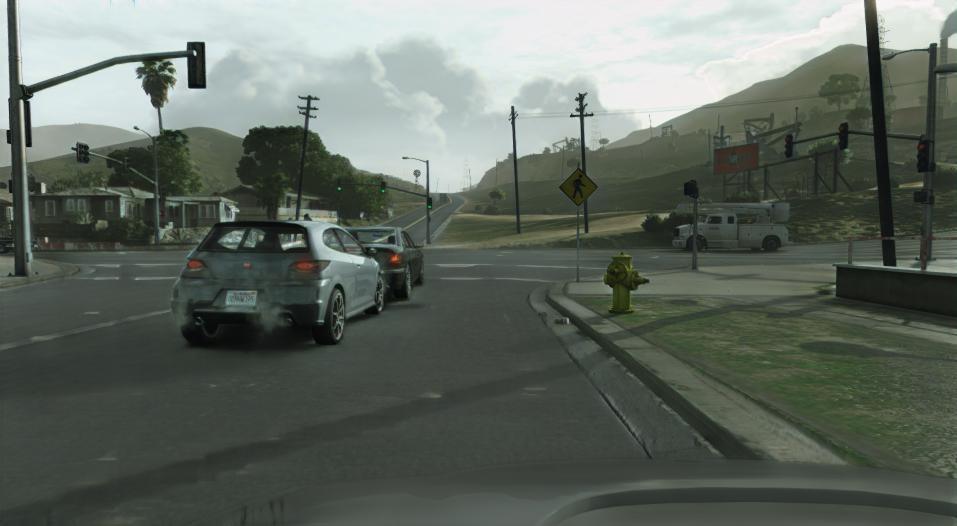}}}\hfill
		{\includegraphics[width=0.249\textwidth]{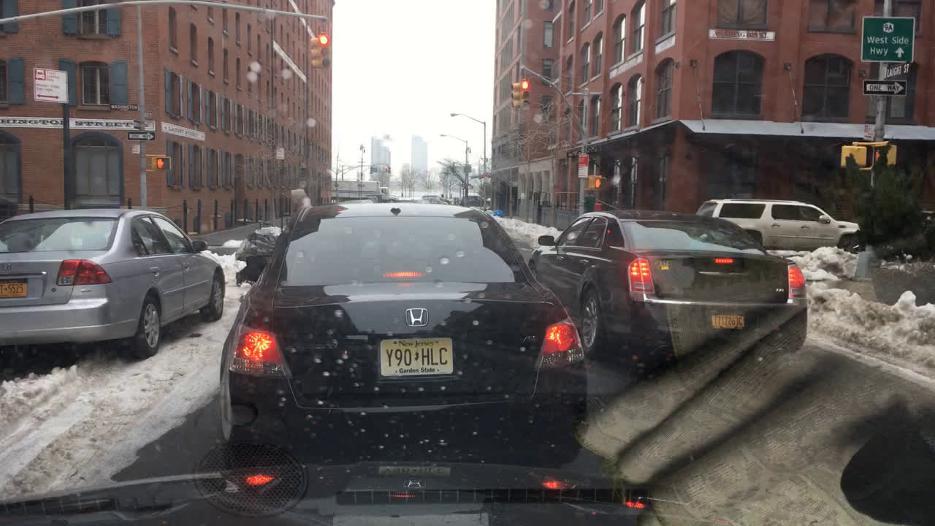}{\includegraphics[width=0.249\textwidth]{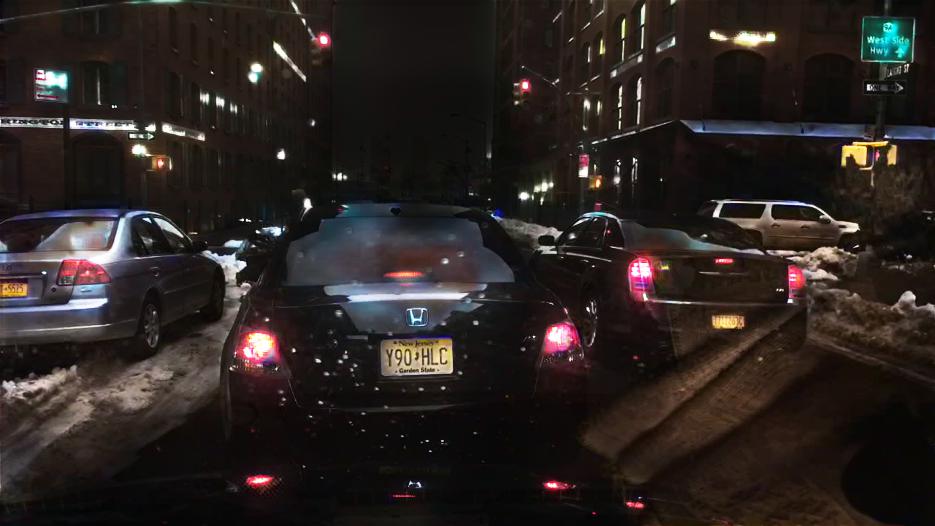}}}\hfill\\  
		{\includegraphics[width=0.249\textwidth]{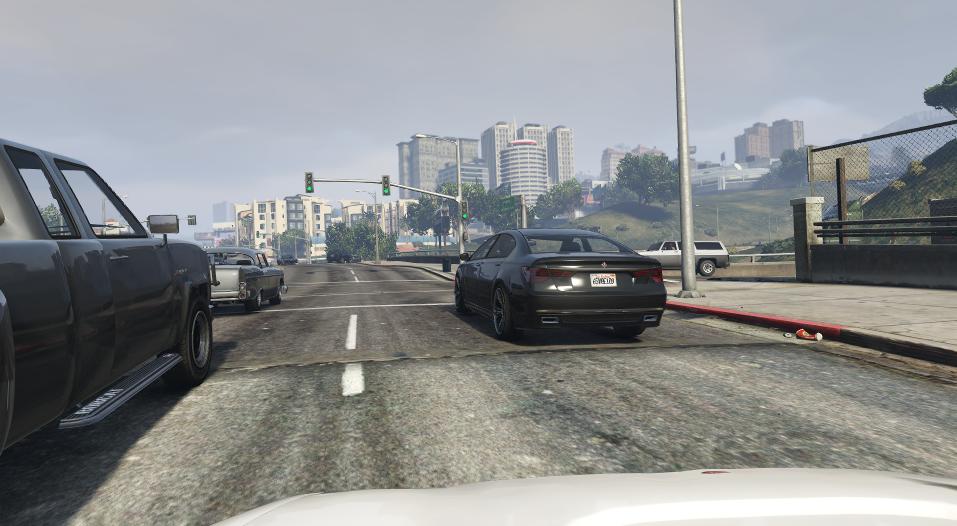}{\includegraphics[width=0.249\textwidth]{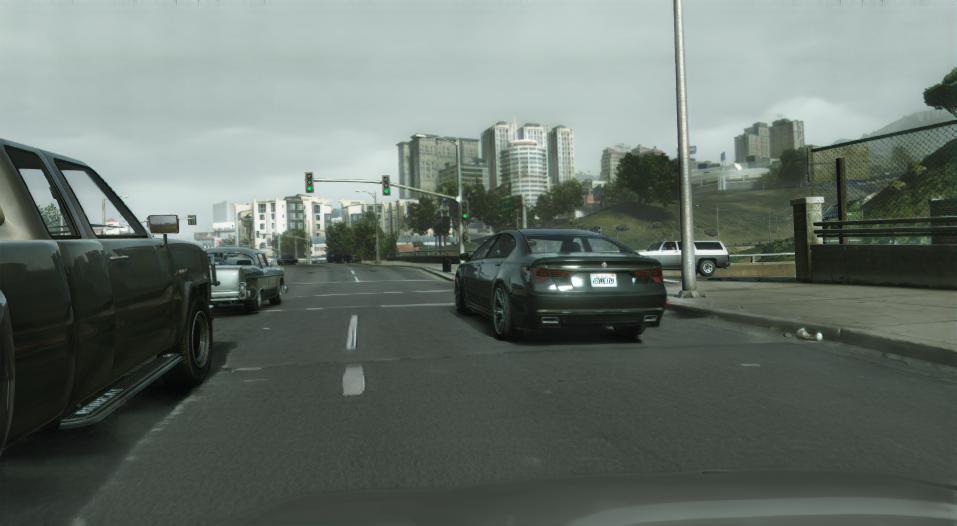}}}\hfill
		{\includegraphics[width=0.249\textwidth]{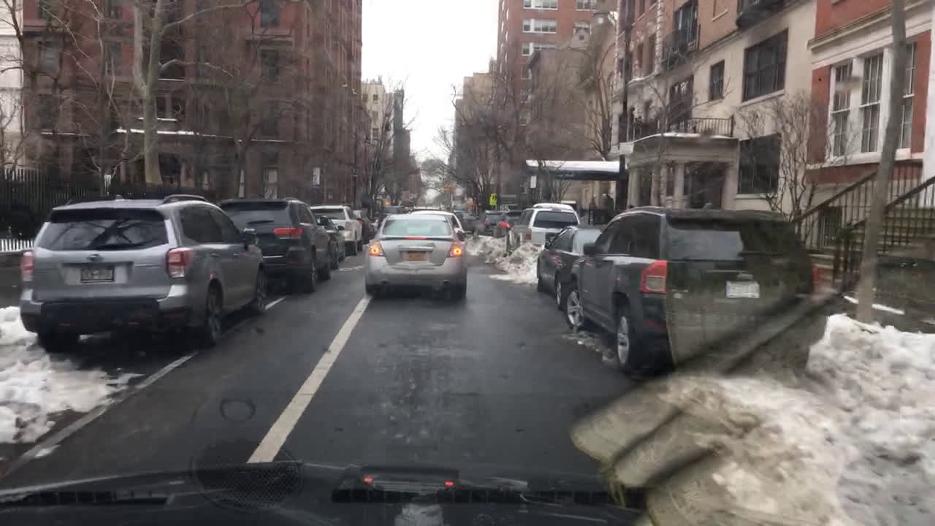}{\includegraphics[width=0.249\textwidth]{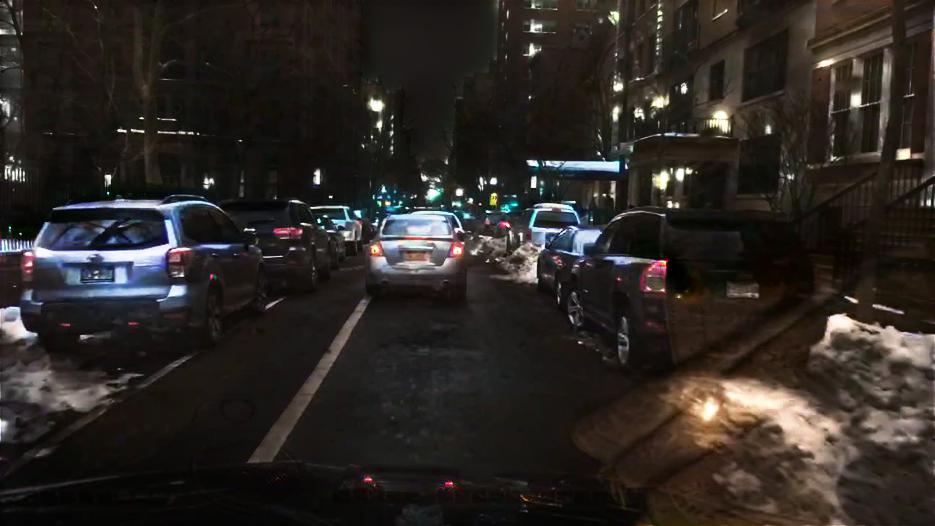}}}\hfill \\  
		{\includegraphics[width=0.249\textwidth]{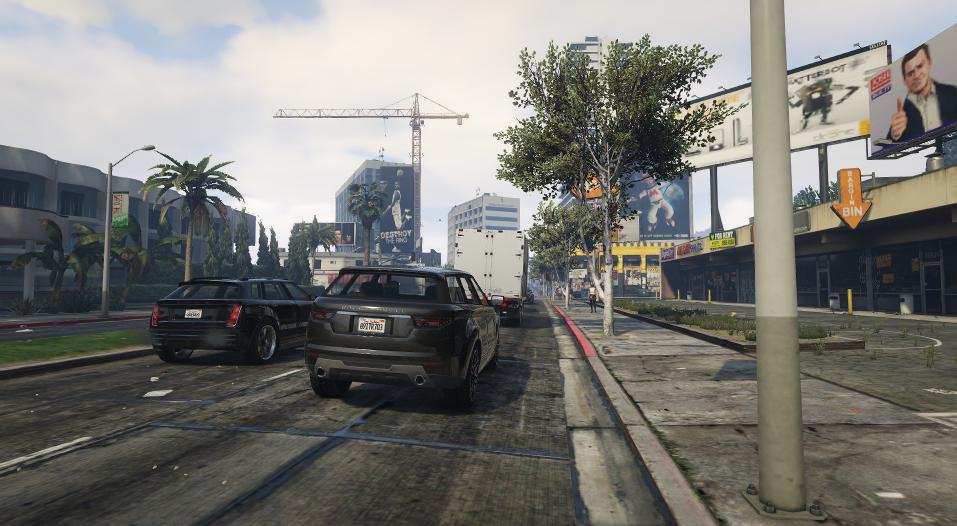}{\includegraphics[width=0.249\textwidth]{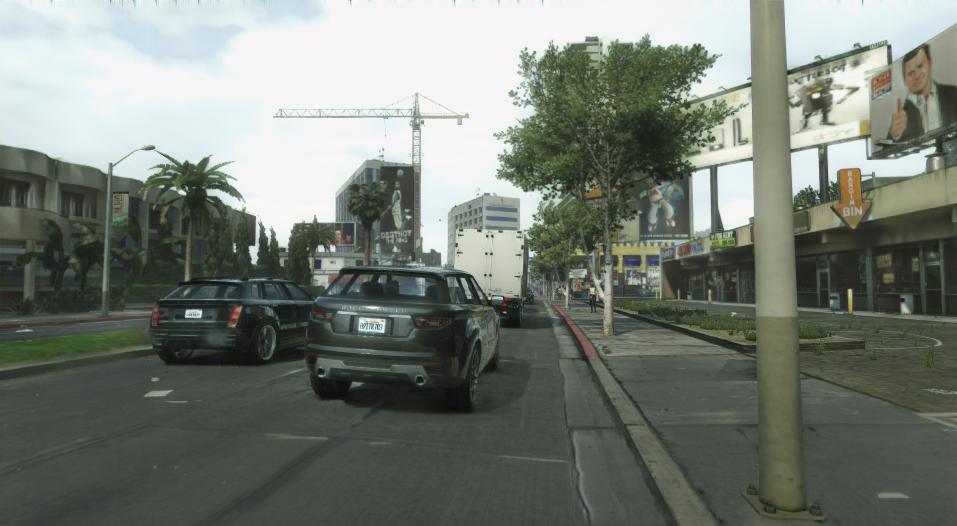}}}\hfill
		{\includegraphics[width=0.249\textwidth]{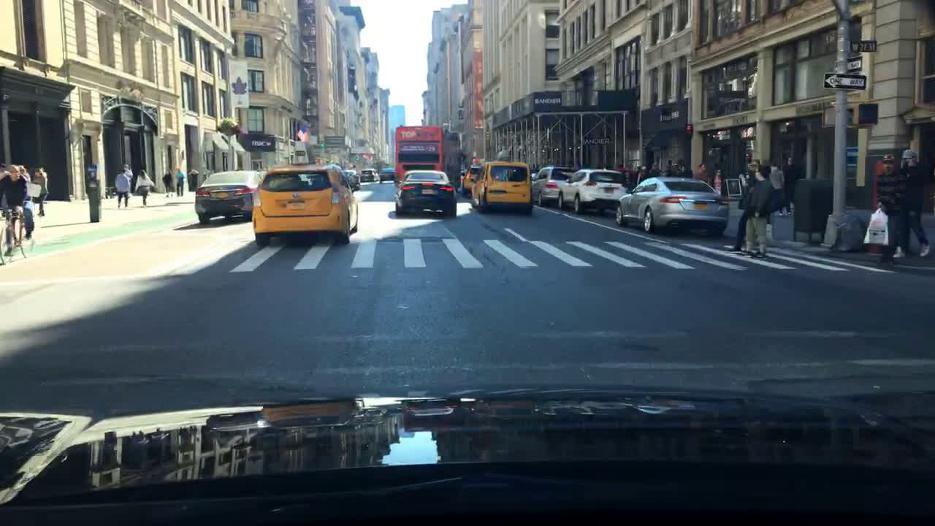}{\includegraphics[width=0.249\textwidth]{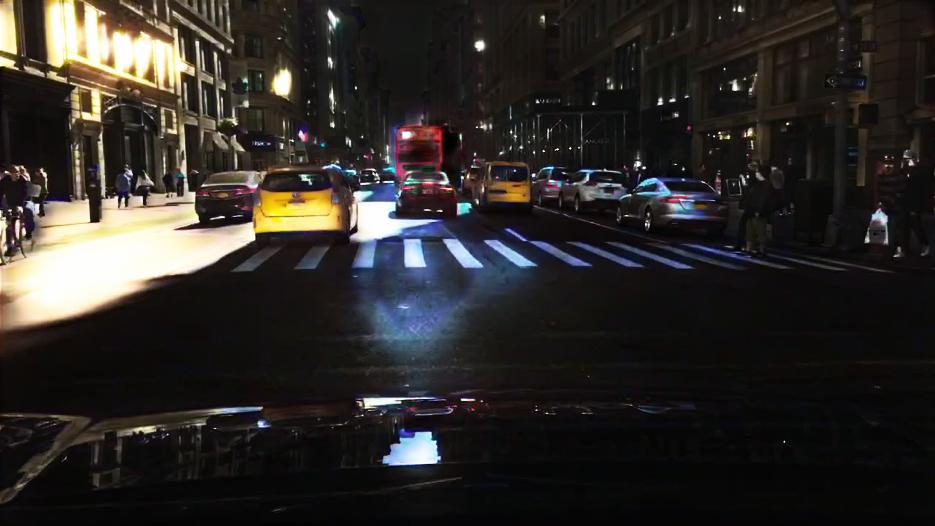}}}\hfill\\ 
		\vspace{-10pt}
		\subfloat[PFD$\rightarrow$Cityscapes]
		{\includegraphics[width=0.249\textwidth]{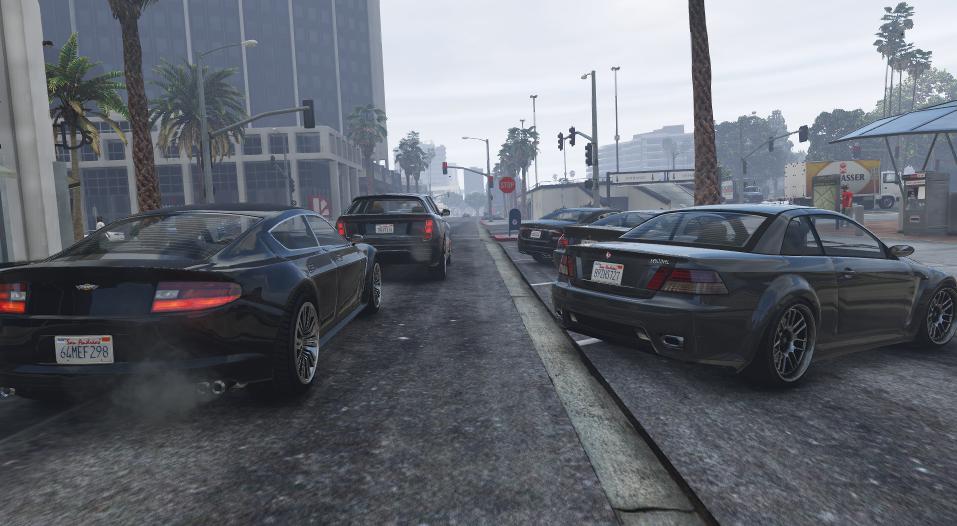}{\includegraphics[width=0.249\textwidth]{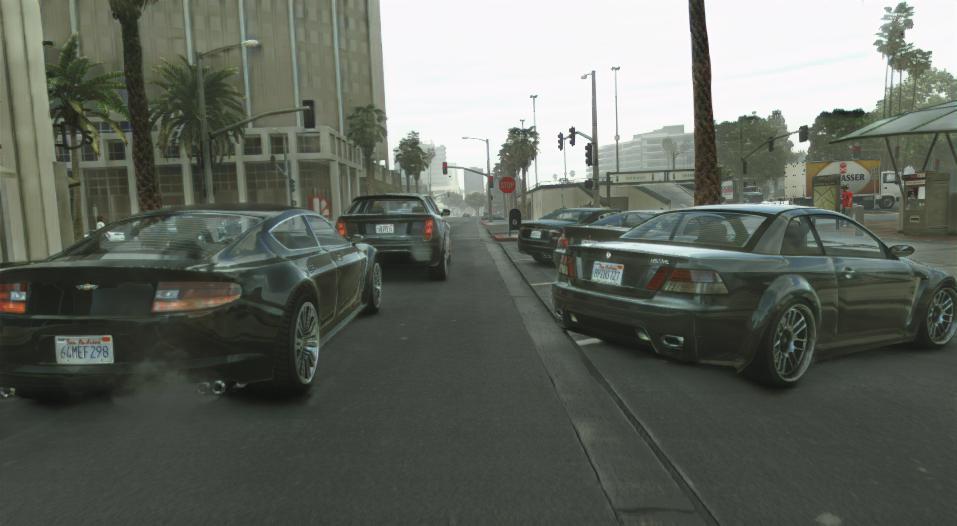}}}\hfill
		\subfloat[Day$\rightarrow$Night]
		{\includegraphics[width=0.249\textwidth]{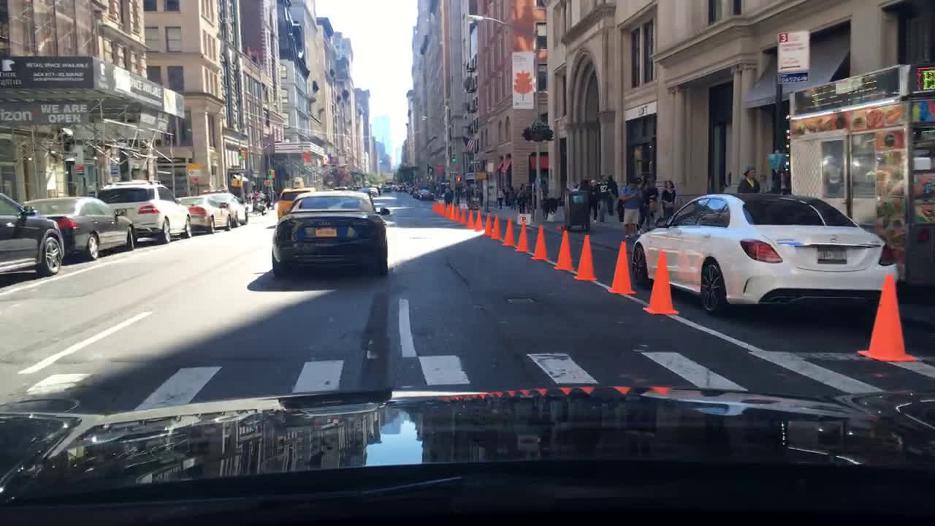}{\includegraphics[width=0.249\textwidth]{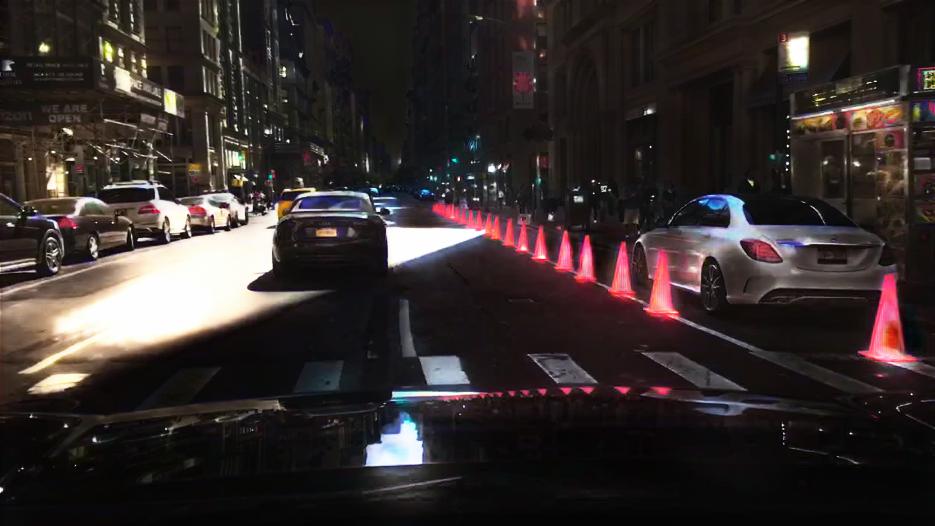}}}\hfill\\ 
		\vspace{+5pt}
		
		{\includegraphics[width=0.249\textwidth]{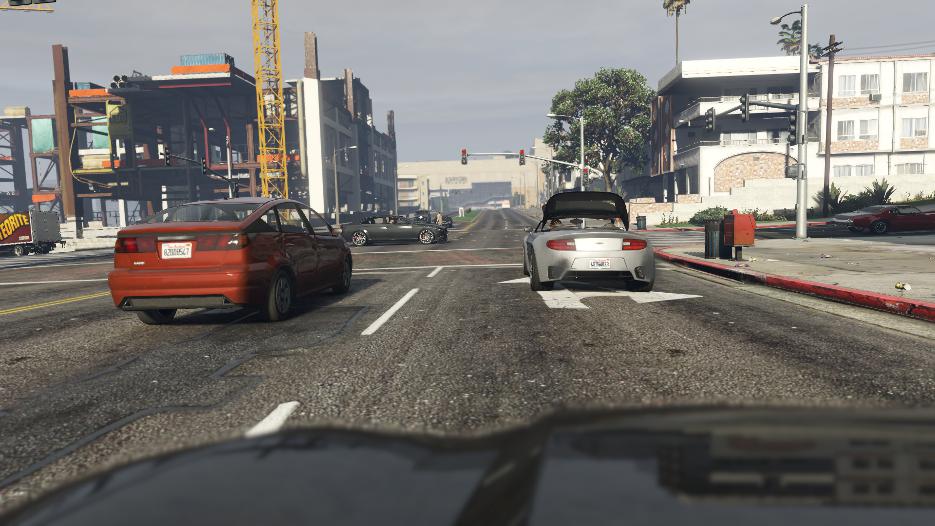}{\includegraphics[width=0.249\textwidth]{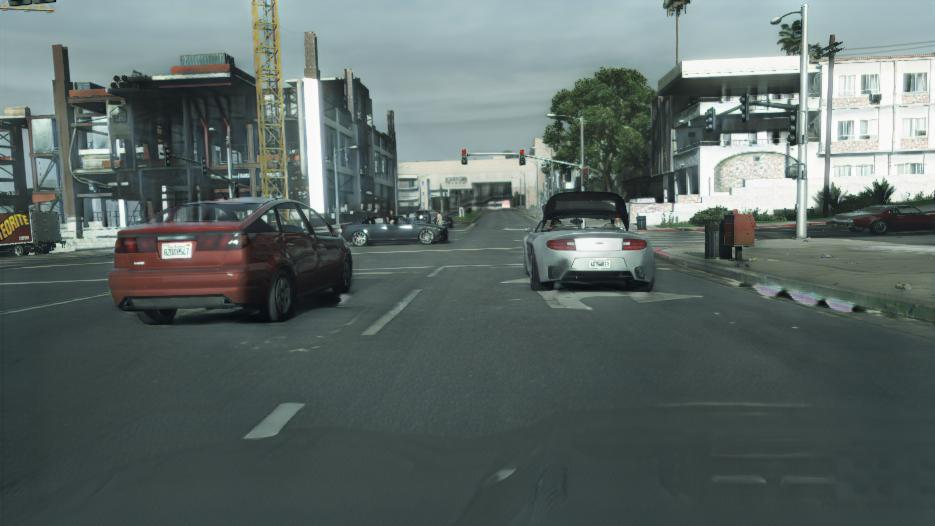}}}\hfill
		{\includegraphics[width=0.249\textwidth]{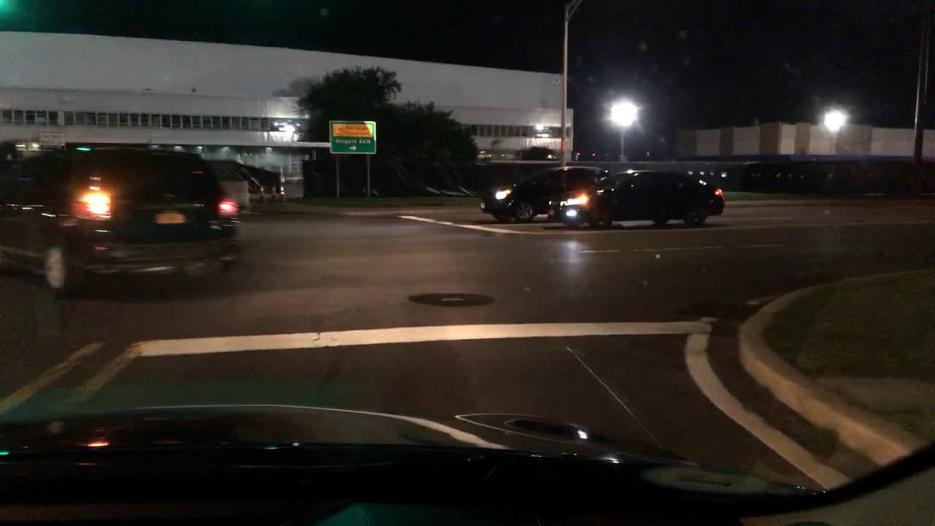}{\includegraphics[width=0.249\textwidth]{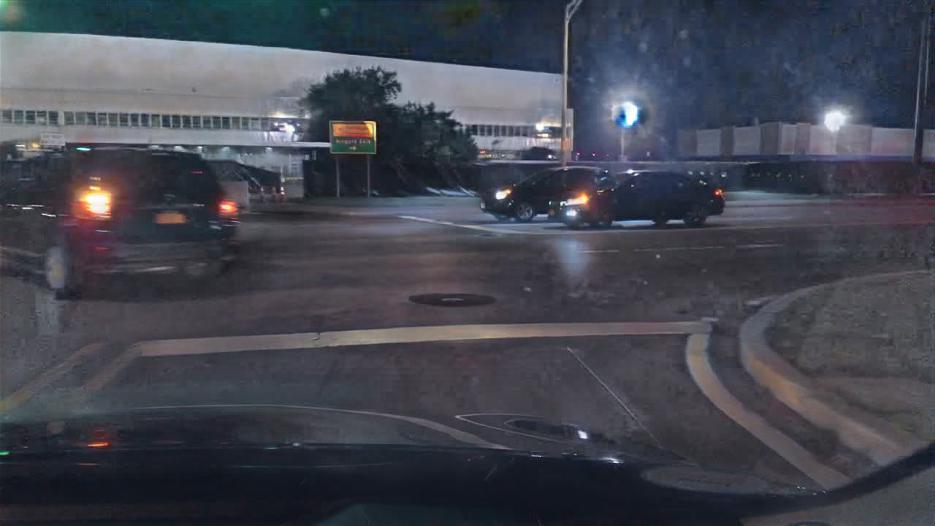}}}\hfill\\  
		{\includegraphics[width=0.249\textwidth]{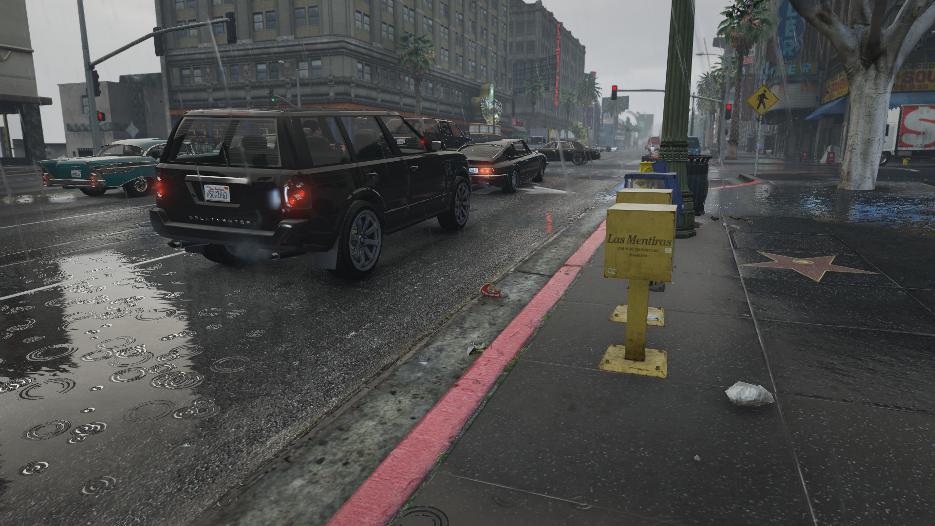}{\includegraphics[width=0.249\textwidth]{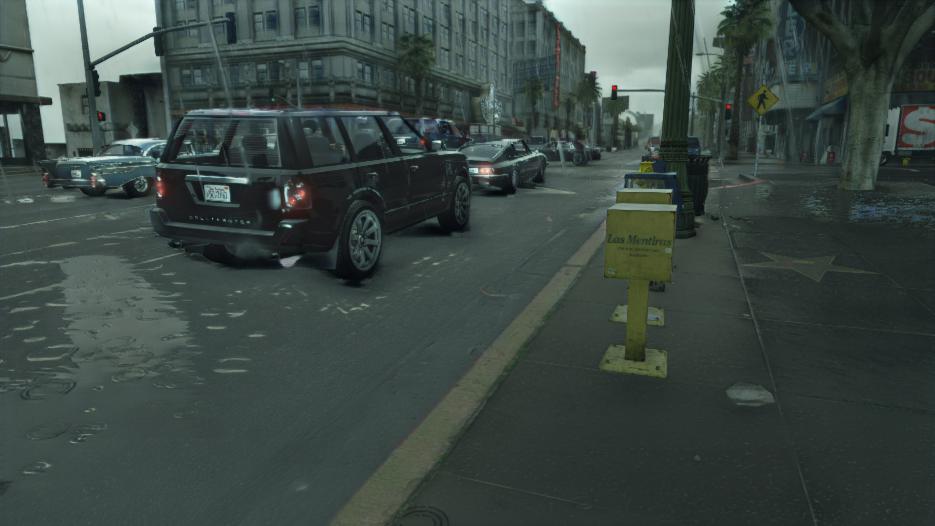}}}\hfill
		{\includegraphics[width=0.249\textwidth]{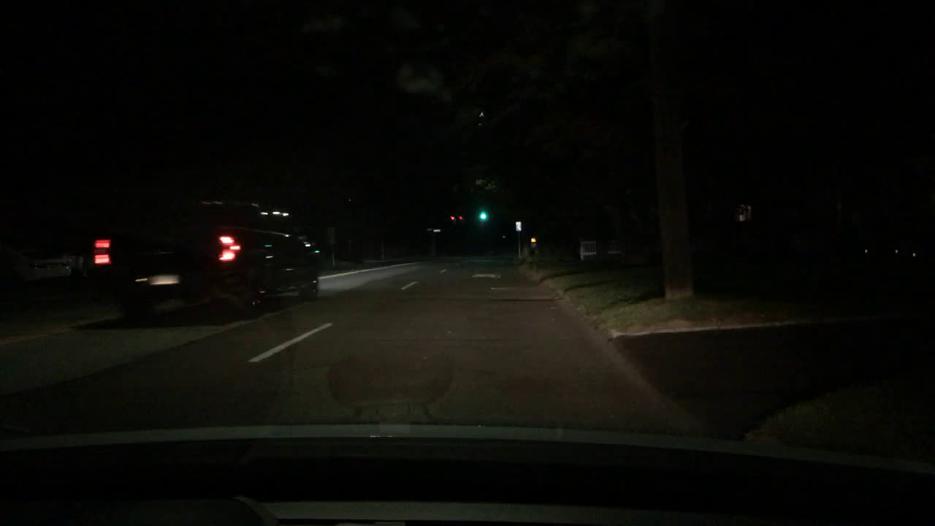}{\includegraphics[width=0.249\textwidth]{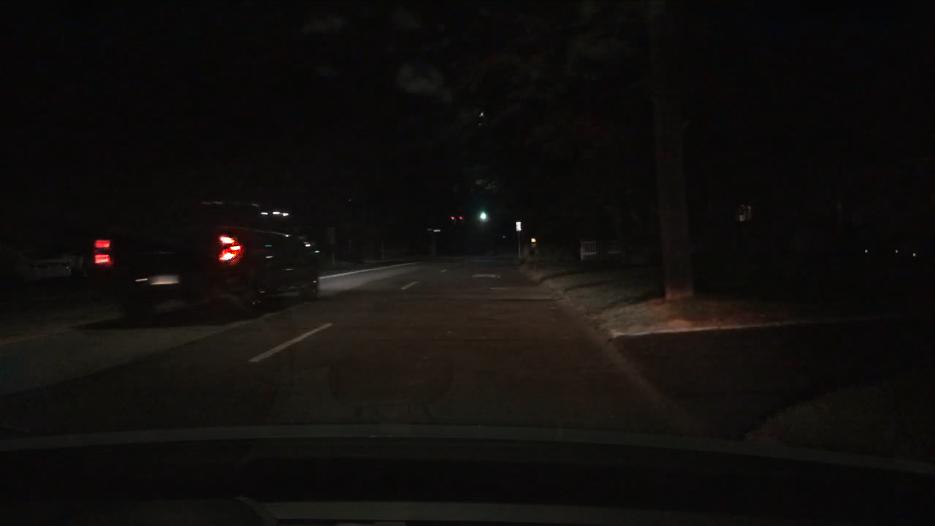}}}\hfill \\  
		{\includegraphics[width=0.249\textwidth]{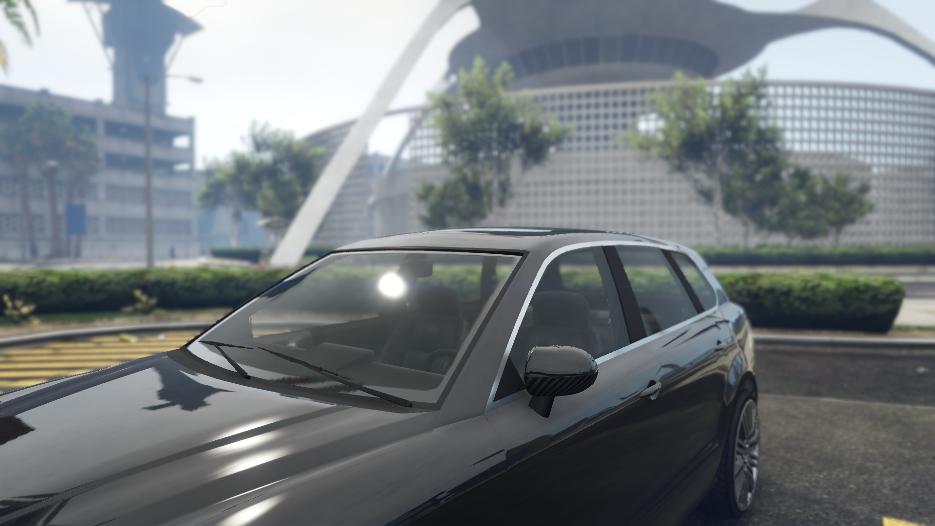}{\includegraphics[width=0.249\textwidth]{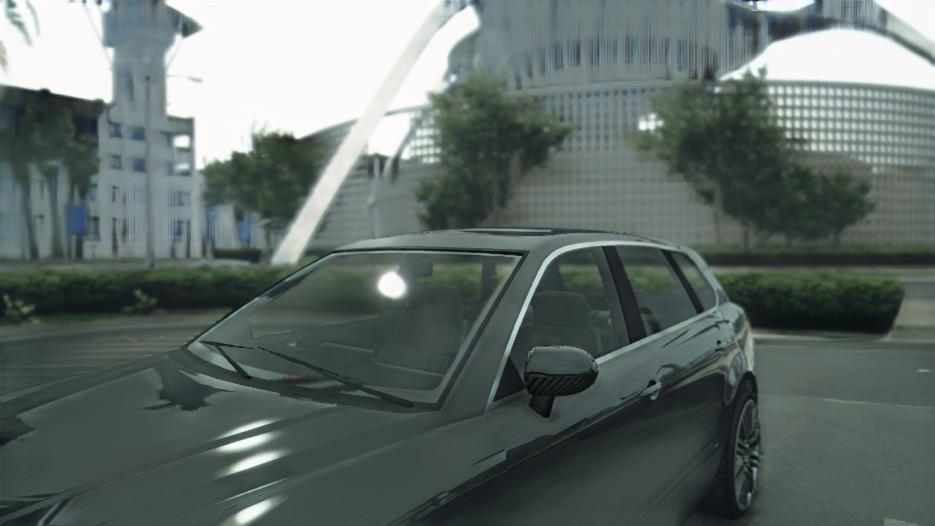}}}\hfill
		{\includegraphics[width=0.249\textwidth]{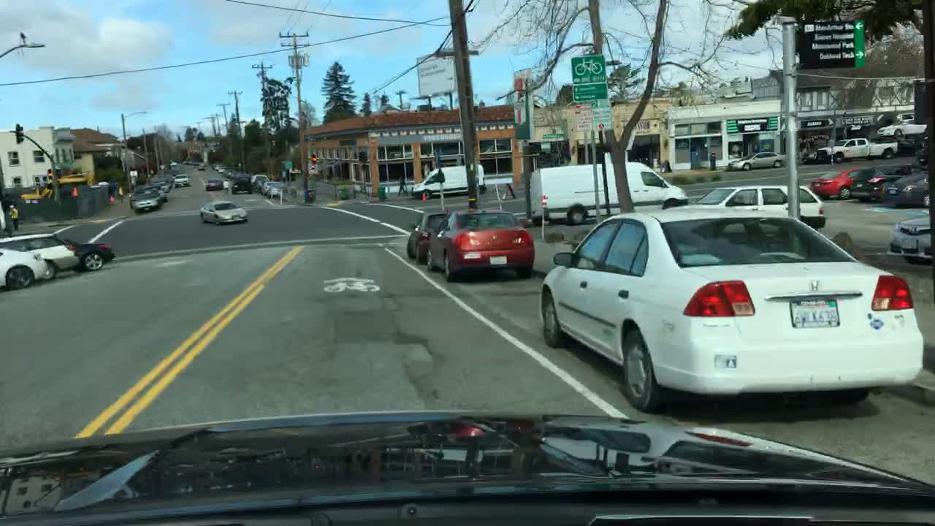}{\includegraphics[width=0.249\textwidth]{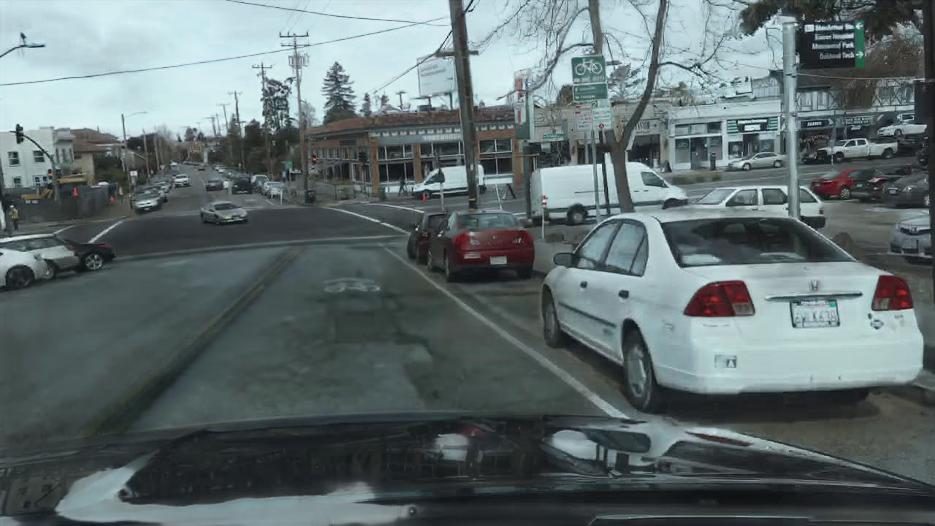}}}\hfill\\ 
		\vspace{-10pt}
		\subfloat[Viper$\rightarrow$Cityscapes]
		{\includegraphics[width=0.249\textwidth]{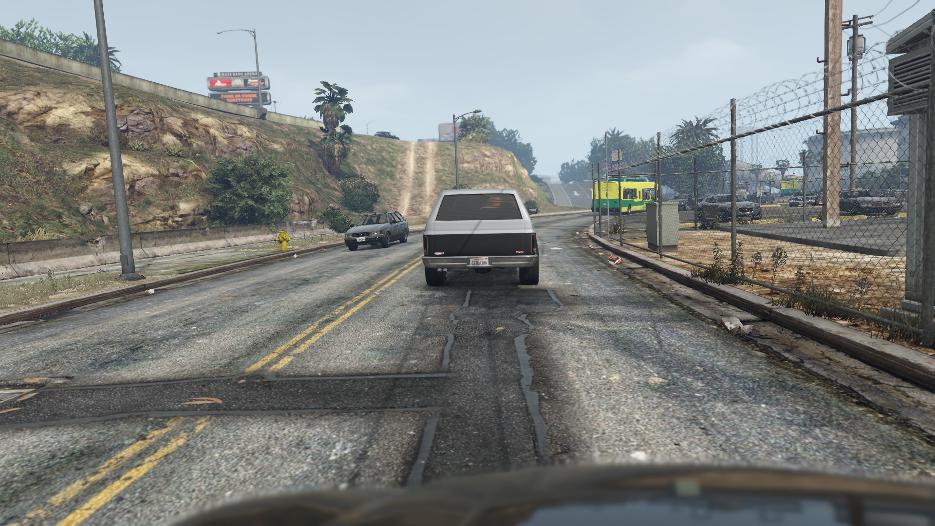}{\includegraphics[width=0.249\textwidth]{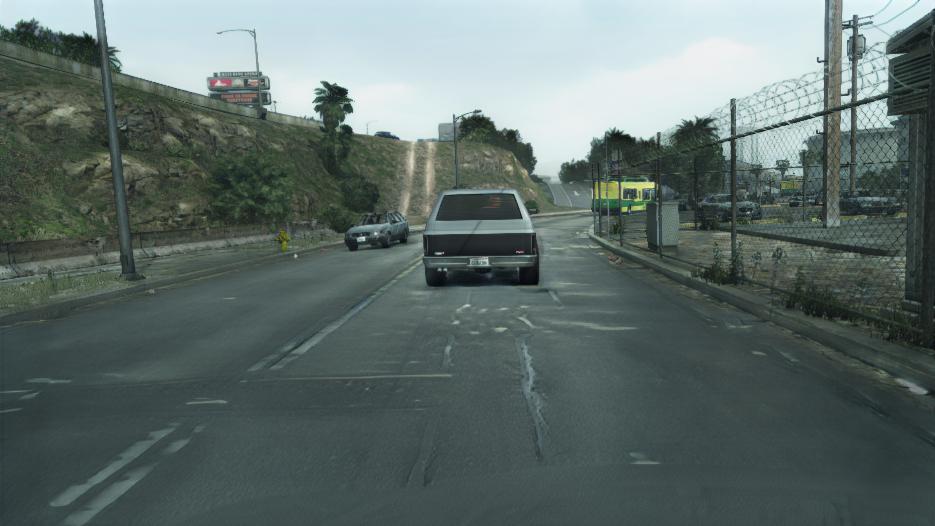}}}\hfill
		\subfloat[Clear$\rightarrow$Snowy]
		{\includegraphics[width=0.249\textwidth]{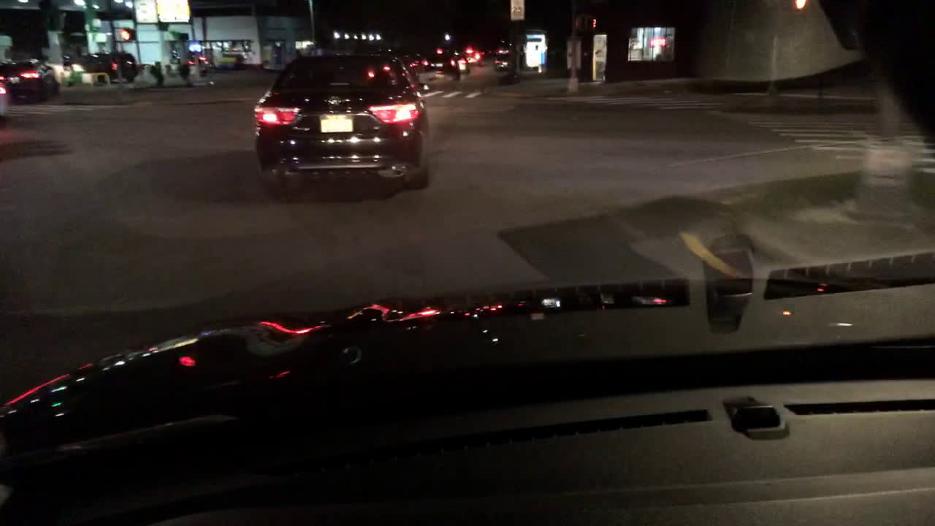}{\includegraphics[width=0.249\textwidth]{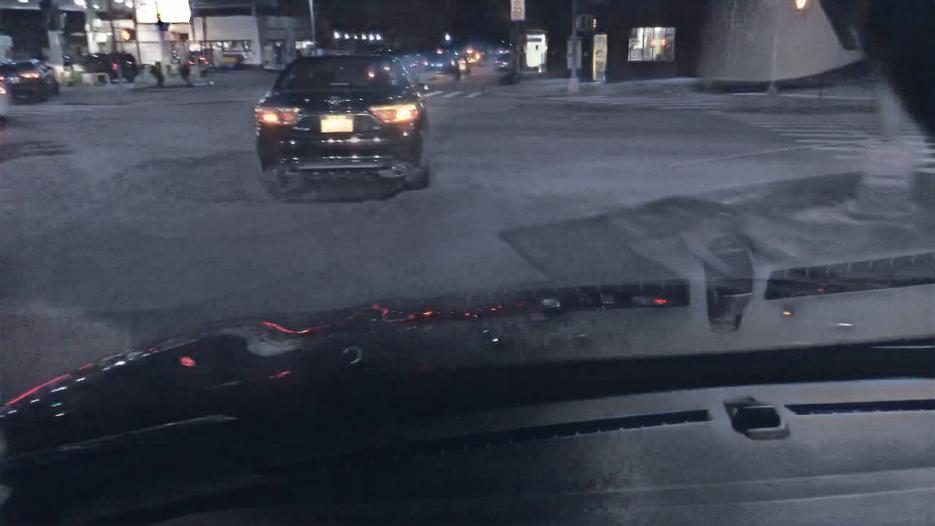}}}\hfill
	\end{center}
	\vspace{-1ex}
	\caption[Additional qualitative results.]{Additional qualitative results. Best viewed in color.}
	\label{fig:feamgan:qualitatative_additional_results}
\end{figure}

\section{Conclusion}
\begin{figure}[h] 
	\captionsetup[subfigure]{labelformat=empty}
	\begin{center}
		{\includegraphics[width=0.248\textwidth]{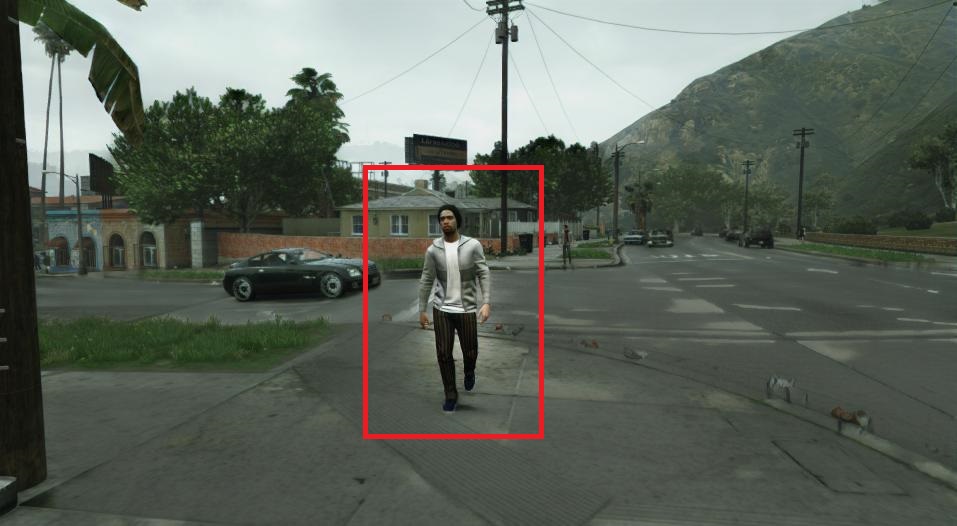}}\hfill
		{\includegraphics[width=0.248\textwidth]{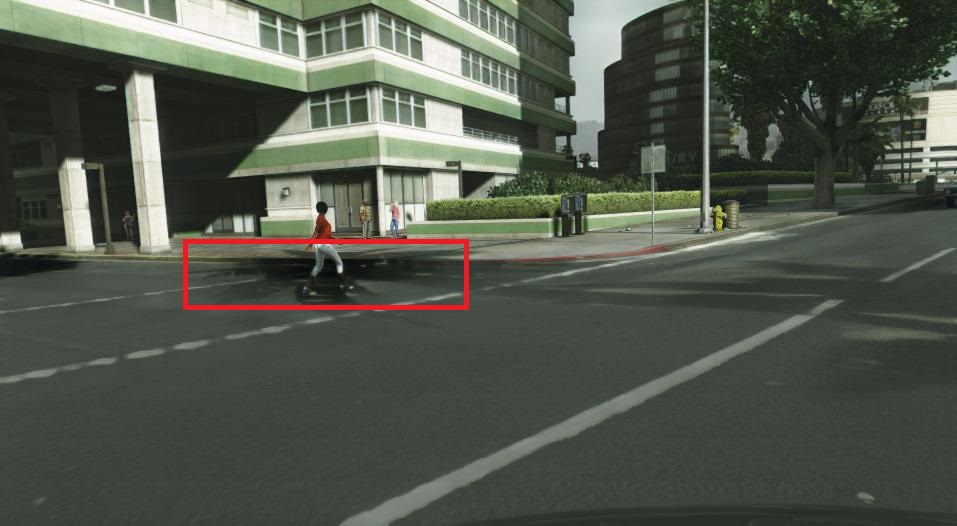}}\hfill
		{\includegraphics[width=0.248\textwidth]{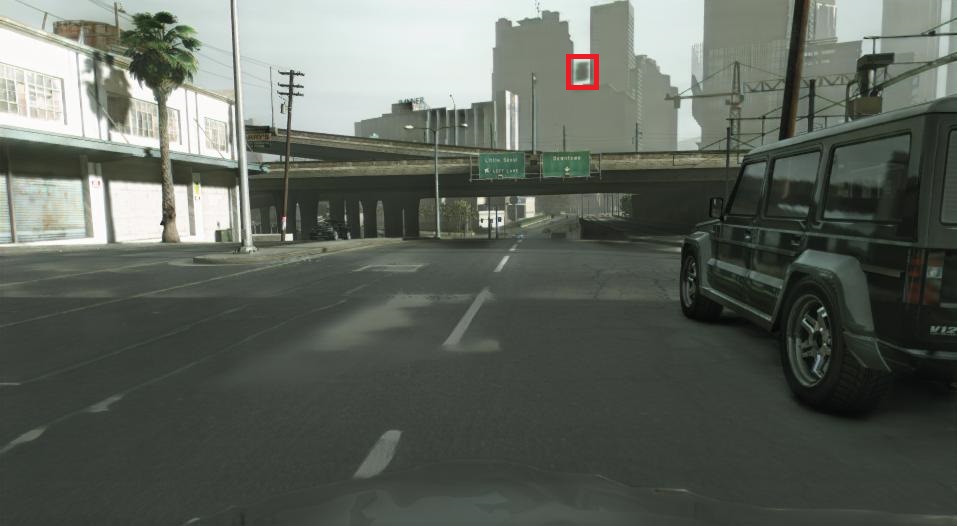}}\hfill
		{\includegraphics[width=0.248\textwidth]{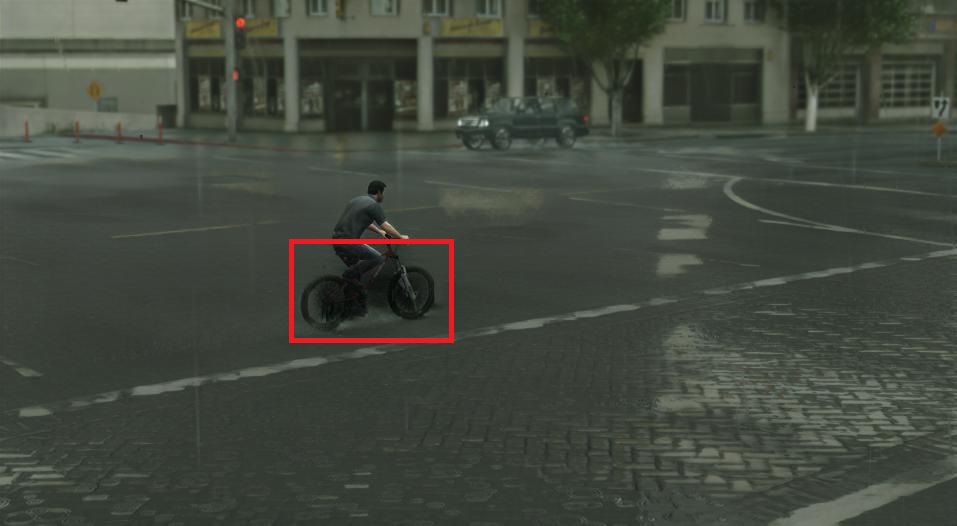}}\hfill \\ 
		{\includegraphics[width=0.248\textwidth]{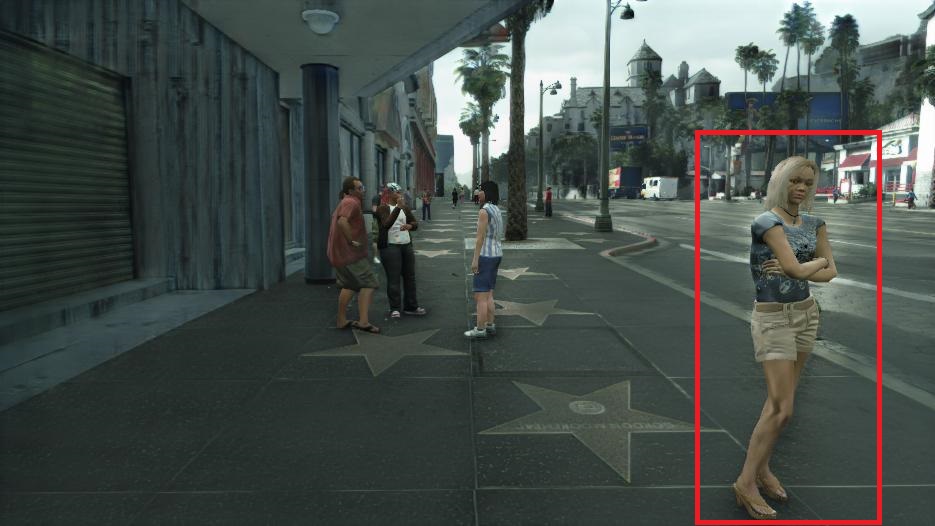}}\hfill
		{\includegraphics[width=0.248\textwidth]{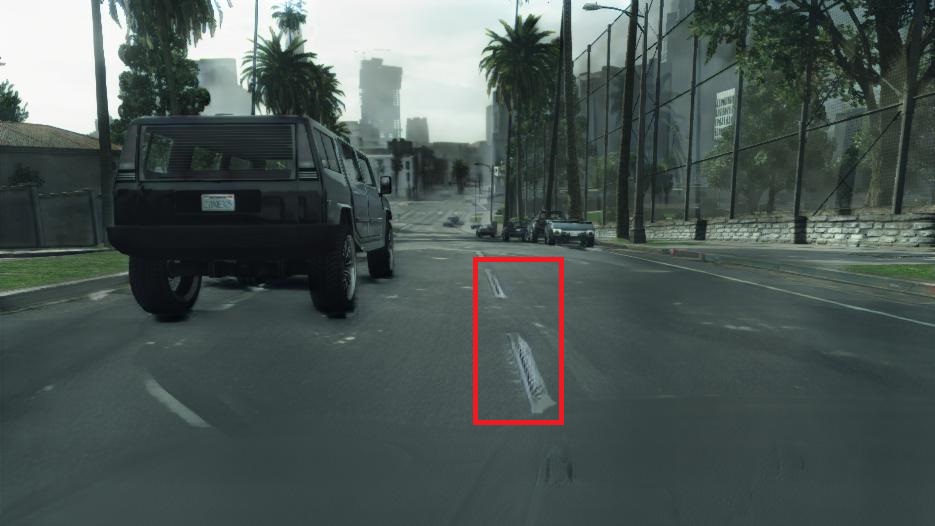}}\hfill
		{\includegraphics[width=0.248\textwidth]{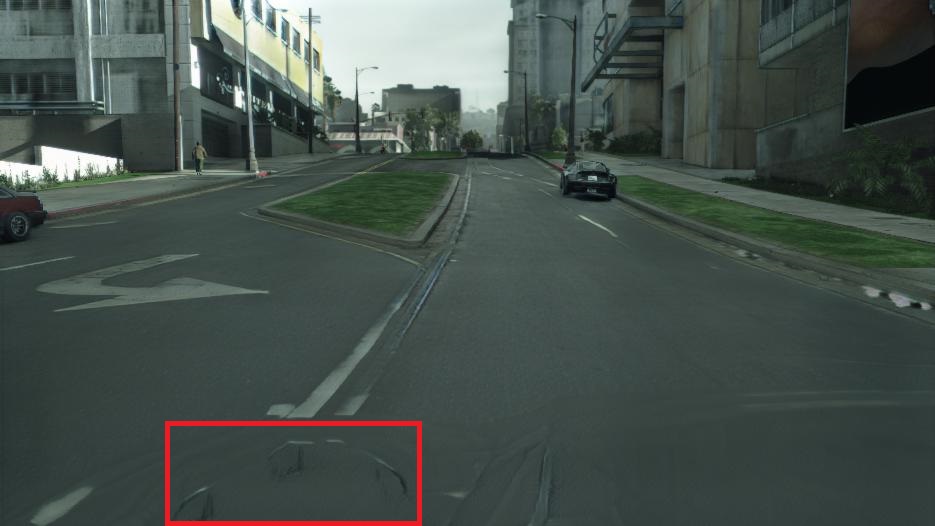}}\hfill
		{\includegraphics[width=0.248\textwidth]{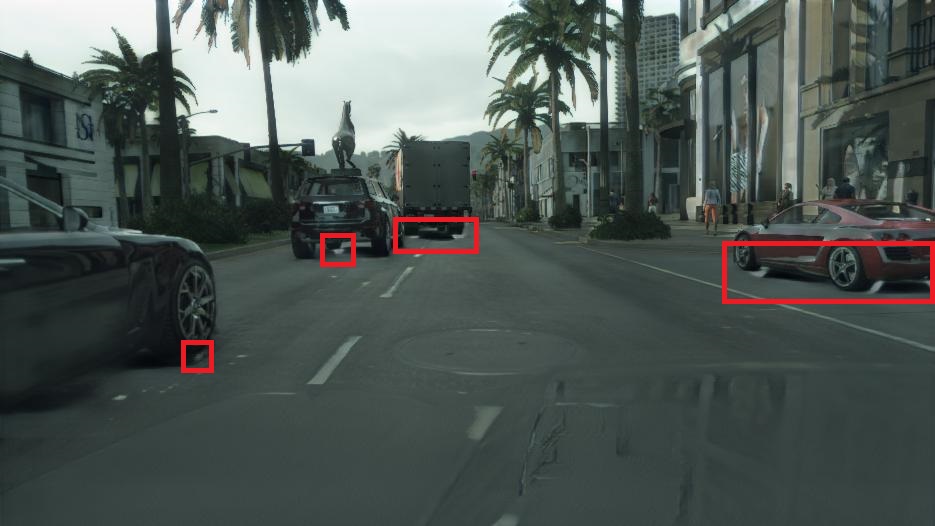}}\hfill \\ 
		{\includegraphics[width=0.248\textwidth]{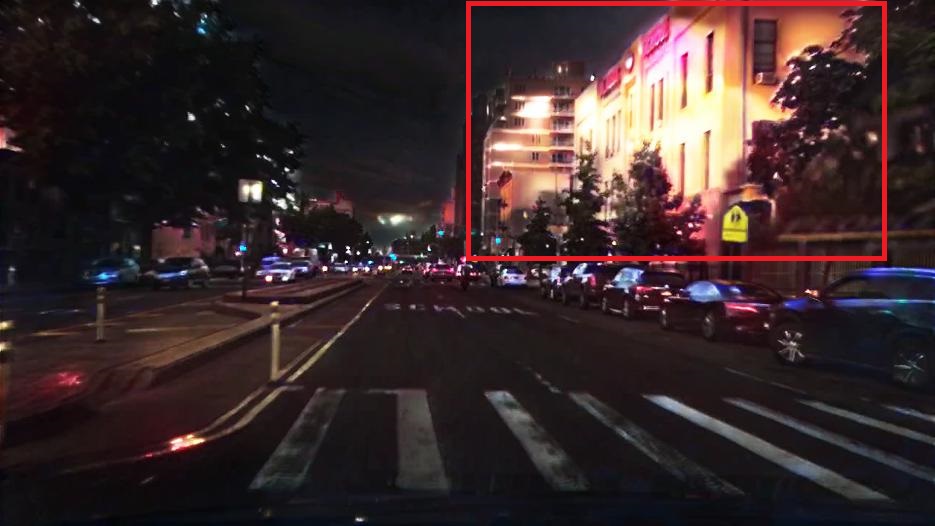}}\hfill
		{\includegraphics[width=0.248\textwidth]{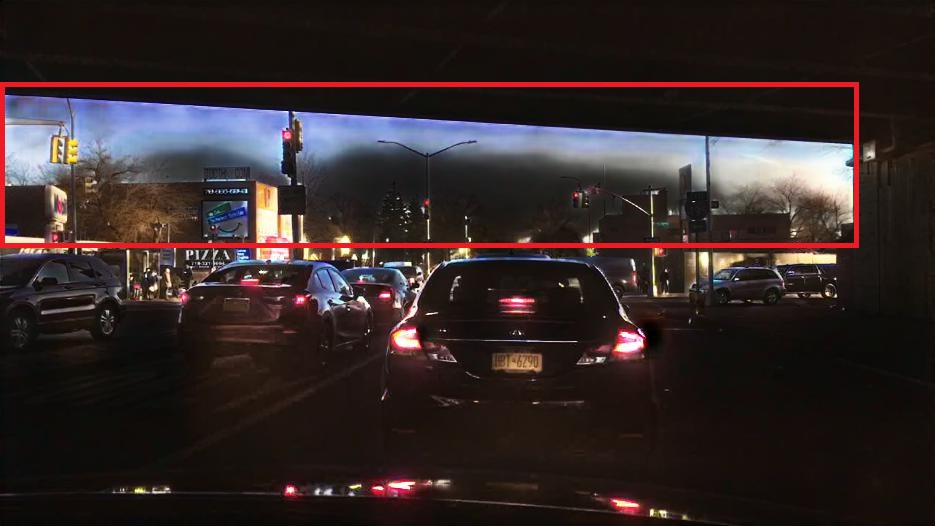}}\hfill
		{\includegraphics[width=0.248\textwidth]{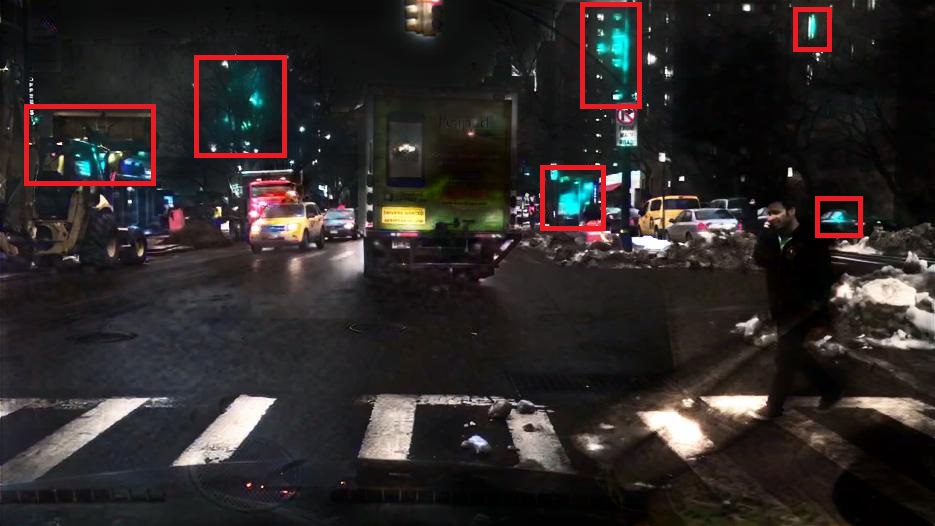}}\hfill 
		{\includegraphics[width=0.248\textwidth]{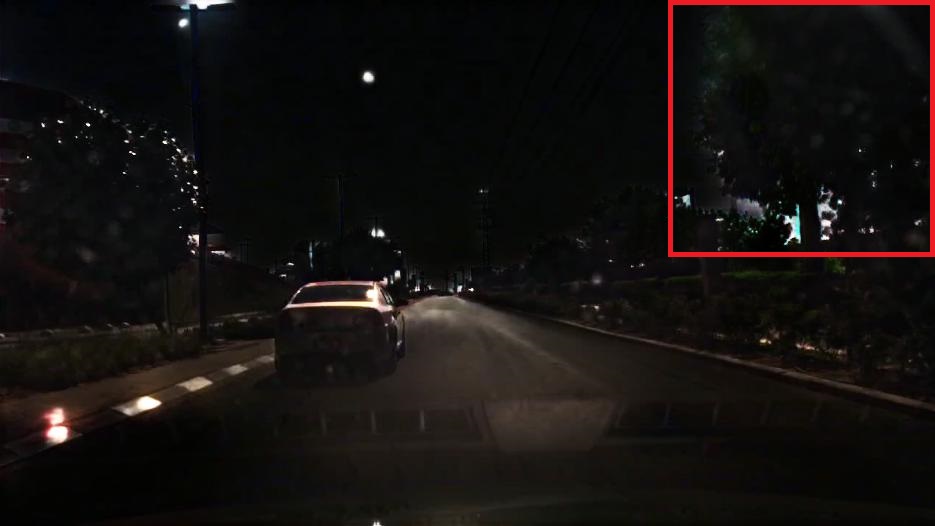}}\hfill  \\
		\vspace{-10pt}
		\subfloat[Glowing objects]
		{\includegraphics[width=0.248\textwidth]{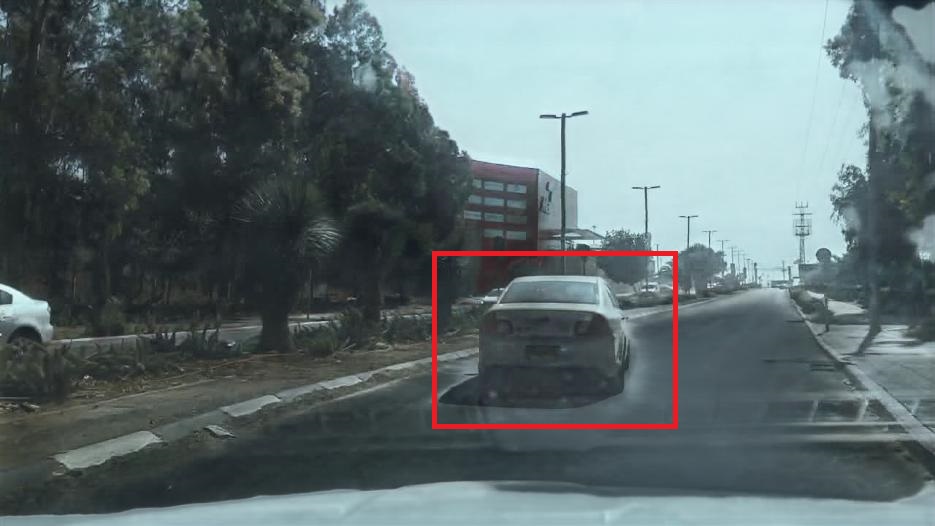}}\hfill
		\subfloat[Intra-class inconsistency]
		{\includegraphics[width=0.248\textwidth]{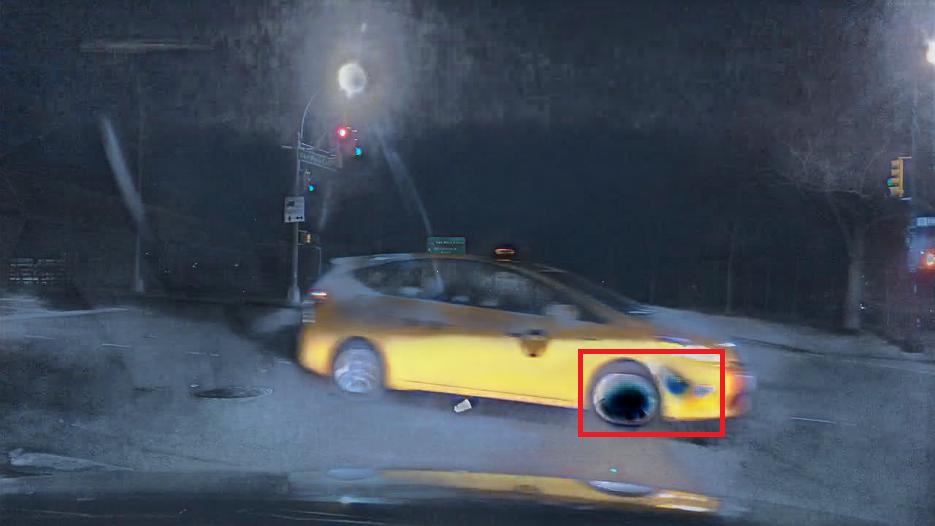}}\hfill
		\subfloat[Minor hallucinations]
		{\includegraphics[width=0.248\textwidth]{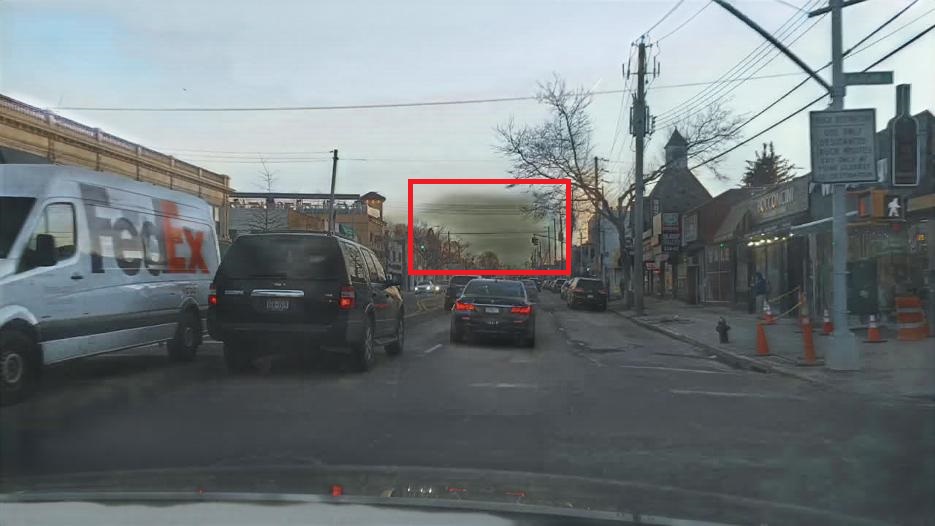}}\hfill
		\subfloat[Class boundary artifiacts]
		{\includegraphics[width=0.248\textwidth]{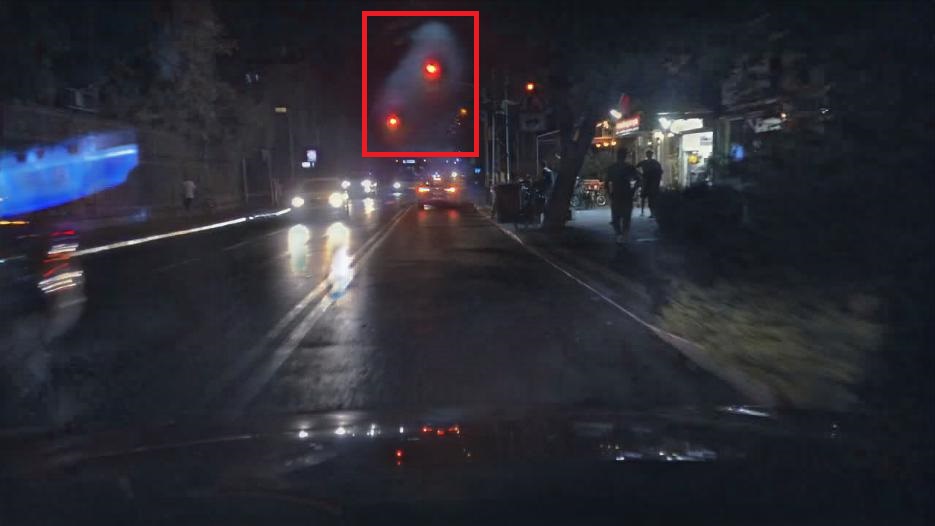}}\hfill
	\end{center}
	\vspace{-1ex}
	\caption[Limitations.]{Limitations. Best viewed in color.}
	\label{fig:feamgan:limitations}
\end{figure}

In this work, we have shown that content-based masking of the discriminator is sufficient to significantly reduce content inconsistencies that arise in unpaired image-to-image translation. Furthermore, artifacts caused by the masking procedure can be significantly reduced by introducing a local discriminator that utilizes a segmentation-based similarity sampling technique. Moreover, our similarity sampling technique leads to a further increase in performance when applied to global input crops. We have also shown that our feature-based denormalization block is able to attend to specific content features, such as features of shadows, but can slightly increase training instability. In addition, we have proposed the cKVD metric to examine translation quality at the class or category level. In our experiments, we have found that these techniques lead to state-of-the-art performance on photo-realistic sim-to-real transfer and the translation of weather. Although our method performs well in Day$\rightarrow$Night translation, the remaining limitations of our approach are especially evident in this task.
\\\\
\textbf{Limitations.} We remark on limitations regarding the dataset, sampling, method, and implementation. Probably the most significant limitations are the complex public datasets currently available and in use, as they are not specifically designed for unpaired translation. Collection strategies and datasets that mitigate biases between source and target domains would be beneficial. Furthermore, our sampling strategy only works on an image basis and could be extended across the entire dataset to sample more significant pairs for training. Although our method works for large crops, there is still a crop size limit that must be taken into account when tuning the hyperparameters. In addition, our method for mitigating content inconsistencies depends on the segmentation model. In theory, the number of classes could be used to control how fine-grained the content consistency should be, which leads to flexibility but allows for errors depending on the segmentation quality. This can result in artifacts such as glowing objects, as shown in \autoref{fig:feamgan:limitations}. Intra-class inconsistencies that may arise from intra-class biases ignored by the loss, such as small textures, represent another problem. Intra-class inconsistencies are currently underexplored in unpaired image-to-image translation and are an interesting direction for future research. Finally, we would like to point out that the efficiency of our implementation could be further improved. Apart from these limitations, our method achieves state-of-the-art performance in complex translation tasks while mitigating inconsistencies through a masking strategy that works by applying few tricks. Simple masking strategies have proven to be very successful in other fields. Therefore, we believe that masking strategies for unpaired image-to-image translation represent a promising direction for further research.
\\\\
\textbf{Ethical and responsible use.} Considering the limitations of current methods, unpaired image-to-image translation methods should be trained and tested with care, especially for safety-critical domains like autonomous driving. A major concern is that it is often unclear or untested whether the transferred content can still be considered consistent for subsequent tasks in the target domain. Even though measures exist for content-consistent translation, they do not allow for the explainability of what exactly is being transferred and changed by the model on a fine-grained level. With our proposed cKVD metric we contribute to this field by allowing class-specific translation measurements - a direction that we hope is the right one. However, even if the content is categorically consistent at a high (class) level, subcategories (like parts of textures) may still be interchanged. At a lower level, content consistency and style consistency are intertwined (e.g., a yellow stop sign). Another privacy and security question is whether translation methods are (or will) be able to (indirectly) project sensitive information from the target domain to the translated images (e.g., exchange faces from simulation with faces of existing persons during the translation). A controllable (class-level and in-class-level) consistency method could help to resolve such issues.\\

\part{Clausula}

\chapter{Conclusions}
\label{chap:end}
\section{Conclusion and Discussion}
\label{sec:conclusion_discussion}
This Ph.D. thesis addresses the field of unsupervised representation learning from the perspectives of learning, evaluating, and transferring visual representations. After an introduction to the field as a whole in \autoref{chap:intro}, we have provided a high-level overview and an introduction of the subfields we have studied and contributed to in sections \ref{sec:intro:ulovr} through \ref{sec:intro:utovr}. Subsequently, the objectives and scope of this dissertation have been defined, along with several research questions focused on the learning, evaluation, and transfer of unsupervised visual representations. Following an outline of the thesis in \autoref{sec:intro:outline}, we have extensively discussed the related work with respect to our contributions in \autoref{chap:related}. In the proceeding chapters \ref{chap:00} to \ref{chap:03}, we have described our contributed backpropagation-free unsupervised representation learning method, metrics for measuring objective function mismatches, CARLANE Benchmark with a new unsupervised domain adaptation method, and content-consistent unpaired image-to-image translation method. Each aforementioned chapter consists of a description of the motivation for the proposed work, a listing of the contributions with respect to the research questions, a method description, the conducted experiments, and a conclusion. A summary of our contributions is provided in the subsequent \autoref{sec:end:summary_of_contributions}.
\\\\
\textbf{Developments of unsupervised visual representation learning}. In retrospect, unsupervised learning of visual representations has made significant progress in recent years. Therefore, we discuss our work as a small part of these (positive) developments: 

In unsupervised representation learning, performance gains for various visual target tasks have been significant. Self-supervised pretext tasks have evolved from data-indirect approaches such as rotation prediction towards more data-direct prediction tasks such as contrastive learning, distillation, or feature predictive distillation. Thus the trend towards the main paradigm of unsupervised representation learning - reducing human intervention - shows promising and steadily improving results. However, there is still comparatively little focus on creating a multitude of unsupervised learning rules apart from backpropagation. Nevertheless, several works - such as \cite{miconi2021hebbian,journe2022hebbian} and ours \cite{stuhr2019csnns} - have contributed to the field of backpropagation-free and/or biologically plausible (unsupervised) learning, leading to increasingly deep models. 

For the evaluation of unsupervised models, there has been an ongoing effort to create novel methods for explaining the inner workings of unsupervised representations and models. This has resulted in promising visualization techniques and benchmarks. In recent years, the development of novel metrics that do not rely on specific datasets and/or benchmarks has received little attention but has not gone unnoticed either. Some works have led to several insights into unsupervised learning setups by defining such new metrics, like our metrics for measuring objective function mismatches \cite{stuhr2022don}, or by examining existing metrics for unsupervised representation learning \cite{grigg2021self,gwilliam2022beyond}.
 
For the transfer of unsupervised visual representations, unsupervised domain adaptation is increasingly being applied to more complex tasks such as semantic segmentation \cite{csurka2021unsupervised} or the CARLANE Benchmark \cite{stuhr2022carlane}. However, with the emergence of large models trained on vast amounts of cross-domain data, the most basic approach to domain adaptation - simply training the model concurrently on all these domains - may lead to surprising performance gains. The problem of content consistency in the translation of unpaired images has recently been recognized and addressed with less ill-posed GAN-based methods - including \cite{richter2022enhancing,theiss2022unpaired}, and our own method \cite{stuhr2023masked} - which have yielded promising results. The evaluation of these methods in the context of content inconsistency still lacks a metric that is widely adopted. Nevertheless, there are ongoing efforts to define novel metrics, as can be seen in \cite{richter2022enhancing} or our work \cite{stuhr2023masked}.
\\\\
\textbf{Limitations of unsupervised visual representation learning}. Now that we have highlighted recent developments related to our work, we discuss some insightful limitations that persist in these subfields.
First, we would like to emphasize limitations that apply to all of these subfields: In particular, multimodality, availability of multimodal curated datasets, continuous learning of models, the efficiency of models and training, and the explainability of methods are still limited. Especially, the unavailability of multidomain curated open-source datasets has been a relevant limitation with regard to recent trends. 

In backpropagation-free unsupervised representation learning, a major limitation is the scalability of these methods to deeper models. In the case of biologically plausible unsupervised learning, methods often rely on longstanding neuroscientific insights without integrating the latest neuroscientific findings. Furthermore, beyond the scientific aspects - such as understanding the brain - it is still unclear whether biologically-plausible, backpropagation-free approaches lead to significant practical benefits for many tasks, as the trajectory of backpropagation-based learning still leads to performance gains. As for the current state of the art of unsupervised representation learning, there is still a performance gap compared to supervised learning. However, this gap is continually closing. Although data-indirect objectives with specific signals, such as rotation prediction, have been succeeded by novel data-direct objectives, such as feature-predictive distillation, these data-direct objectives do not account for the possible plausible variations of masked features. Often there could be more than one plausible solution for the predictive features of a masked region, which is not taken into account by the utilized objectives. For large vision models trained on carefully curated data, limitations lie in their trajectory: For example, the utilized datasets do not cover all vision domains (or tasks), and the current combination of models and training objectives requires additional regularization to stabilize training. Furthermore, there is currently no insight into whether these models exploit in-dataset statistics of the datasets from which the curated dataset is built.

For the evaluation of unsupervised models, a representation is mostly measured by transferring models to target tasks and by visualizing specific properties. This is the only widely adopted consensus that defines good representations. Other empirical measurements are barely adopted and underexplored. For example, the robustness analysis of these models is limited to a few works and lacks further studies. Moreover, to date, no measurement can reliably predict the performance of models on a variety of target tasks. 

For unsupervised domain adaptation, there is a lack of large multi-domain datasets for complex tasks that strongly indicate the direct usability of methods in the real word.  Furthermore, stability, especially in the temporal dimension of the adapted models, has not yet been sufficiently explored. Another limitation is that only longstanding, but no recent unsupervised auxiliary tasks have been explored in the context of unsupervised domain adaptation.

In unpaired image-to-image translation, the unavailability of suitable bias-reduced datasets is a major limitation that can have a strong impact on performance. Given the benefits of real-time, high-resolution inference, there is an especially strong need to improve the efficiency and memory utilization of these models. Furthermore, a major concern is that it is often not clear what exactly is being transferred and changed by the model and whether the transferred content is still consistent. Even if the content is categorically consistent at a high level, subcategories (like parts of textures) may still be interchanged. At a lower level, content consistency and style consistency are intertwined (e.g., a yellow stop sign). 
\\\\
\textbf{Ethics of unsupervised visual representation learning}. With the steady integration of machine learning approaches into the society, ethical considerations regarding these approaches are becoming ever more necessary. These considerations apply especially to backbone models that are or will be used for a multitude of tasks. As unsupervised representation learning methods rapidly improve and the first large models trained on curated data from multiple domains show promising results \cite{oquab2023dinov2}, analysis for fairness and biases, such as analysis for geographic-, gender-, skintone-, and age-based potentially harmful label associations, are rapidly gaining importance. In addition, the environmental impact of such large models should be taken into account. Here we believe that an open-source culture with releases of datasets, implementations, and pre-trained models can reduce the environmental impact since environmentally expensive model pretraining and dataset curation are shared across the community. As models trained on vast curated datasets containing billions of images proliferate, another increasingly relevant ethical consideration is the potential encoding of personal information in pre-trained models that could be indirectly obtained from their representations of those models. The aforementioned considerations also apply to the transfer of representation into other domains.

\section{Future Perspectives on Unsupervised Representation Learning} 
Based on the discussed developments, limitations, and ethics in \autoref{sec:conclusion_discussion}, current trends in unsupervised visual representation learning, and the paradigm shift in natural language processing, we want to provide future perspectives for unsupervised representation learning. In particular, large unsupervised vision models, stably-trained on curated data, appear to be the future of unsupervised representation learning. These models could encompass or at least assist all the subfields discussed in this dissertation as well as others. Carefully curated data and stable, scalable training objectives may be critical to the development of these models. Going a step further, continual unsupervised learning of carefully monitored backbones from curated data streams could be a possible future. From an ethical perspective, aligning these backbones with the right intent for the target task may become relevant, which brings various scientific challenges, legal questions, and ethical considerations. In addition, multimodal integration or training beyond the vision domain could lead to more powerful models. A more specific future perspective is the development of training objectives (based on discriminators) that integrate plausible variability in masked feature prediction. Another specific perspective is to replace the prediction head of latent-conditioned joint-embedding predictive architectures with GAN-based or diffusion-based models, which seems natural due to the latent conditioning. These specific perspectives could complement the development of large vision models trained on curated data. Future perspectives regarding backpropagation-free unsupervised learning include continued examinations and developments regarding the scalability of these models. Of particular interest appears to be the successful integration of skip connections and the incorporation of the advantages of image transformers into these models. For biologically plausible unsupervised representational learning, the integration of the latest neuroscience findings, such as new neuron formulations, in combination with scaling efforts is promising. 

For the evaluation of unsupervised visual representation learning, in addition to continuing the current directions, a relevant direction is to further investigate the temporal dimension of the training setup and its effects on representations and models. This could also assist the process of creating large vision models. Furthermore, with the rise of curated datasets, metrics to prevent the model from exploiting in-database statistics (of the datasets from which the curated dataset is built) may be relevant. For evaluations regarding unpaired image-to-image translation, novel metrics to measure content consistency and style consistency correctly on multiple consistency levels could increase the trust in these methods. 

For the unsupervised adaptation of visual representations, it is plausible that the advent of large vision models, trained on curated data, could set a new hard-to-beat state of the art and finally combine unsupervised representation learning with unsupervised domain adaptation. However, utilizing adaptation techniques for the alignment of large vision models appears to be an interesting future research direction. For unpaired image-to-image translation, adopting and investigating unpaired diffusion models and large, carefully pre-trained (text-to-image) generation models in the context of content consistency and style consistency is intriguing. A further appealing direction could be the integration of large vision models into this setup, for example, as encoders for an unpaired latent-to-latent translation.

\section{Summary of Contributions}
\label{sec:end:summary_of_contributions}
In this dissertation, we contribute to the field of unsupervised representation learning by developing and evaluating methods to learn, adapt and translate visual representations. We have designed a convolutional, backpropagation-free method for unsupervised visual representation learning with modules that can be stacked deeper than previous backpropagation-free baselines. During the evaluation of our backpropagation-free models, we have encountered objective function mismatches between our unsupervised pretext task and the evaluated target tasks. This has led to the design of metrics for measuring objective function mismatches between unsupervised pretext models and (supervised) target models. With these metrics, we have investigated several well-known unsupervised representation learning approaches and target tasks and have found dependencies of the objective function mismatch with respect to different aspects of the training setup and model. Furthermore, we have highlighted the impact of the objective function mismatch on target task performance.

A parallel line of research on an autonomous model vehicle has led to the CARLANE Benchmark, the first 3-way sim-to-real domain adaptation benchmark for 2D lane detection. To achieve baseline performances for this benchmark, we have combined well-known unsupervised domain adaptation methods with the ultra-fast lane detection method and have performed evaluations on single-source single-target and single-source multi-target unsupervised domain adaptation. In addition, we have constructed a new unsupervised domain adaptation method, which has reached state-of-the-art performance. We then have focused on unpaired, one-sided domain adaptation of images and have constructed a content-consistent unpaired image-to-image translation method. Our method utilizes masked global and local discriminators to mitigate content inconsistencies during the translation of images between biased datasets, which are a common problem in the ill-posed setups of unpaired image-to-image translation. In addition, we have proposed feature-attentive denormalization to selectively fuse feature statistics into the generator stream, which allows the model to ignore certain statistics of the source domain in different layers. For the quantitative comparison, we have used well-known perceptual metrics and have designed a new metric that incorporates content inconsistency based on segmentation maps. Both quantitatively and qualitatively, our method has outperformed the state of the art on the evaluated translation tasks.
\\\\
Overall, this dissertation has led to the following publications, code and dataset contributions, and awards:
\subsection{List of Publications}
\begin{itemize}
	\item \textbf{Bonifaz Stuhr}, Jürgen Brauer, Bernhard Schick, and Jordi Gonzàlez. "Masked Discriminators for Content-Consistent Unpaired Image-to-Image Translation.", under review, 2023
	\item \textbf{Bonifaz Stuhr}, Johann Haselberger, and Julian Gebele. "CARLANE: A Lane Detection Benchmark for Unsupervised Domain Adaptation from Simulation to multiple Real-World Domains." Advances in Neural Information Processing Systems (NeurIPS), 2022
	\item \textbf{Bonifaz Stuhr} and Jürgen Brauer. "Don’t miss the mismatch: investigating the objective function mismatch for unsupervised representation learning." Neural Computing and Applications (\textbf{Q1 journal}), 2022
	\item \textbf{Bonifaz Stuhr} and Jürgen Brauer. "Csnns: Unsupervised, backpropagation-free convolutional neural networks for representation learning." 18th IEEE International Conference On Machine Learning And Applications (ICMLA) (\textbf{oral}), 2019
\end{itemize}

\subsection{Contributed Code and Datasets}
\begin{itemize}
	\item \textbf{FeaMGan}: A framework for training and validating our FeaMGan method, containing configuration files and links to pre-trained model weights and inferred images for all evaluated translation tasks. All results of the publication  "Masked Discriminators for Content-Consistent Unpaired Image-to-Image Translation" can be reproduced with the included configuration files. \href{https://github.com/BonifazStuhr/feamgan}{https://github.com/BonifazStuhr/feamgan}
	\item \textbf{CARLANE Benchmark}: The official home of CARLANE, a 3-way sim-to-real domain adaptation benchmark for 2D lane detection. This homepage contains links to download the benchmark datasets (MoLane, TuLane, and MuLane), a tutorial, the paper, an introductory video, the CARLANE Baselines and CARLANE Dataset Tools repositories, a leaderboard, as well as all model weights. Furthermore, this homepage shows the results of the baselines and our SGPCS method. CARLANE Baselines is a framework containing the implementations and configuration files of all CARLANE Benchmark baseline models and our own SGPCS method. All experiments of the publications  "CARLANE: A Lane Detection Benchmark for Unsupervised Domain Adaptation from Simulation to multiple Real-World Domains" can be reproduced with the included configurations. CARLANE Dataset Tools is a repository containing all the tools created to collect, clean, and label the datasets for the CARLANE Benchmark. \href{https://carlanebenchmark.github.io}{https://carlanebenchmark.github.io}
	\item \textbf{OFM}: A framework containing our metrics to measure the objective function mismatch between unsupervised models and target tasks. Furthermore,  the implementations and configuration files of all measured unsupervised methods and target tasks are included. All experiments of the publications  "Don’t miss the mismatch: investigating the objective function mismatch for unsupervised representation learning" can be reproduced with the included configurations. \href{https://github.com/BonifazStuhr/OFM}{https://github.com/BonifazStuhr/OFM}
	\item \textbf{CSNN}: A framework to create, train, evaluate and test backpropagation-free, unsupervised CSNN models, which includes CSNN layers, learning rules, models, and configuration files. With these configuration files, all experiments of the publication "Csnns: Unsupervised, backpropagation-free convolutional neural networks for representation learning" can be reproduced.\\ \href{https://github.com/BonifazStuhr/CSNN}{https://github.com/BonifazStuhr/CSNN}
\end{itemize}

\subsection{Awards}
\begin{itemize}
	\item \textbf{Outstanding Reviewer Award} for NeurIPS 2022\\
	\href{https://neurips.cc/Conferences/2022/DatasetBenchmarkProgramCommittee}{	https://neurips.cc/Conferences/2022/DatasetBenchmarkProgramCommittee}
	\item \textbf{1st Place} at the VDI Autonomous Driving Challenge 2022\\ \href{https://www.vdi-adc.de}{https://www.vdi-adc.de}
	\item \textbf{1st Place} at the VDI Autonomous Driving Challenge 2020\\
	\href{https://www.vdi-adc.de}{https://www.vdi-adc.de}
\end{itemize}

\appendix
\vspace{-8mm}
\chapter{Supplementary Material:\\Self-Organizing Convolutional Neural Networks}
\label{app:00}

\begin{figure}[h]
	\begin{center}
		\includegraphics[width=1.0\linewidth]{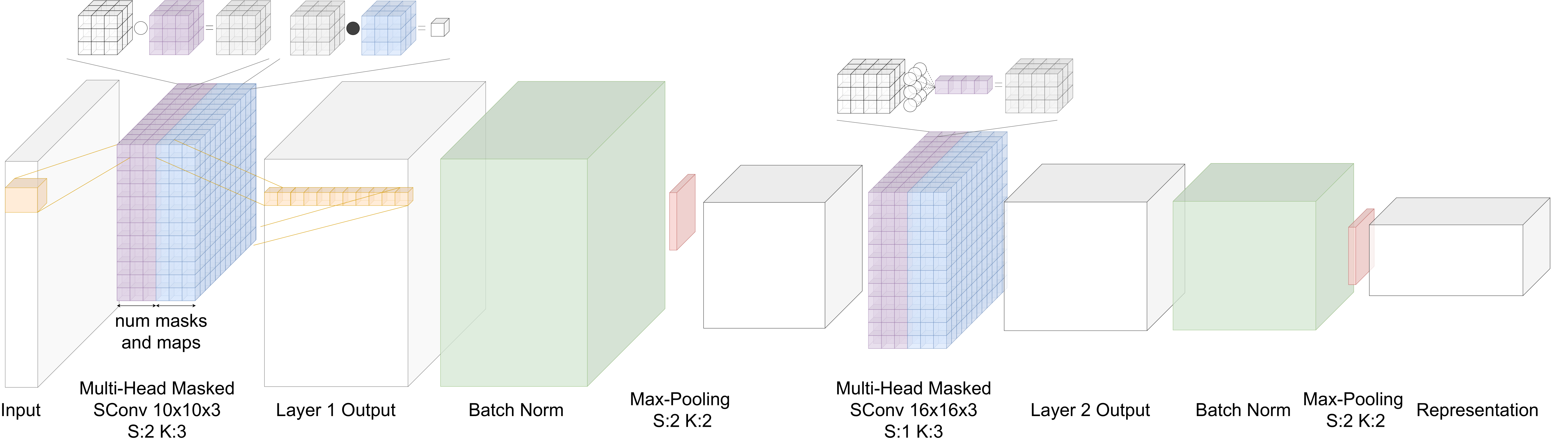}
	\end{center}
	\vspace{-1ex}
	\caption[Detailed architecture of the S-\acs{CSNN} model with three \acs{SOM} maps per layer.]{Detailed architecture of the S-CSNN model with three SOM maps per layer. The model consists of two \acs{CSNN} layers, each followed by a batch normalization and a max-pooling layer to halve the spatial dimension of the output tensor. The first layer uses a $10\times10\times3$ \acs{SOM} grid (3 heads with 100 \acs{SOM} neurons), stride $2\times2$, and input mask weights. The second layer uses a $16\times16\times3$ \acs{SOM} grid, stride $1\times1$,  and mask weights between neurons. Both layers use a kernel size of $3\times3$ and padding type ``same''. Best viewed in color.}
	\label{fig:csnns:app:scsnn_architecture}
\end{figure}

\begin{figure}[t]
	\begin{center}
		\includegraphics[width=1.0\linewidth]{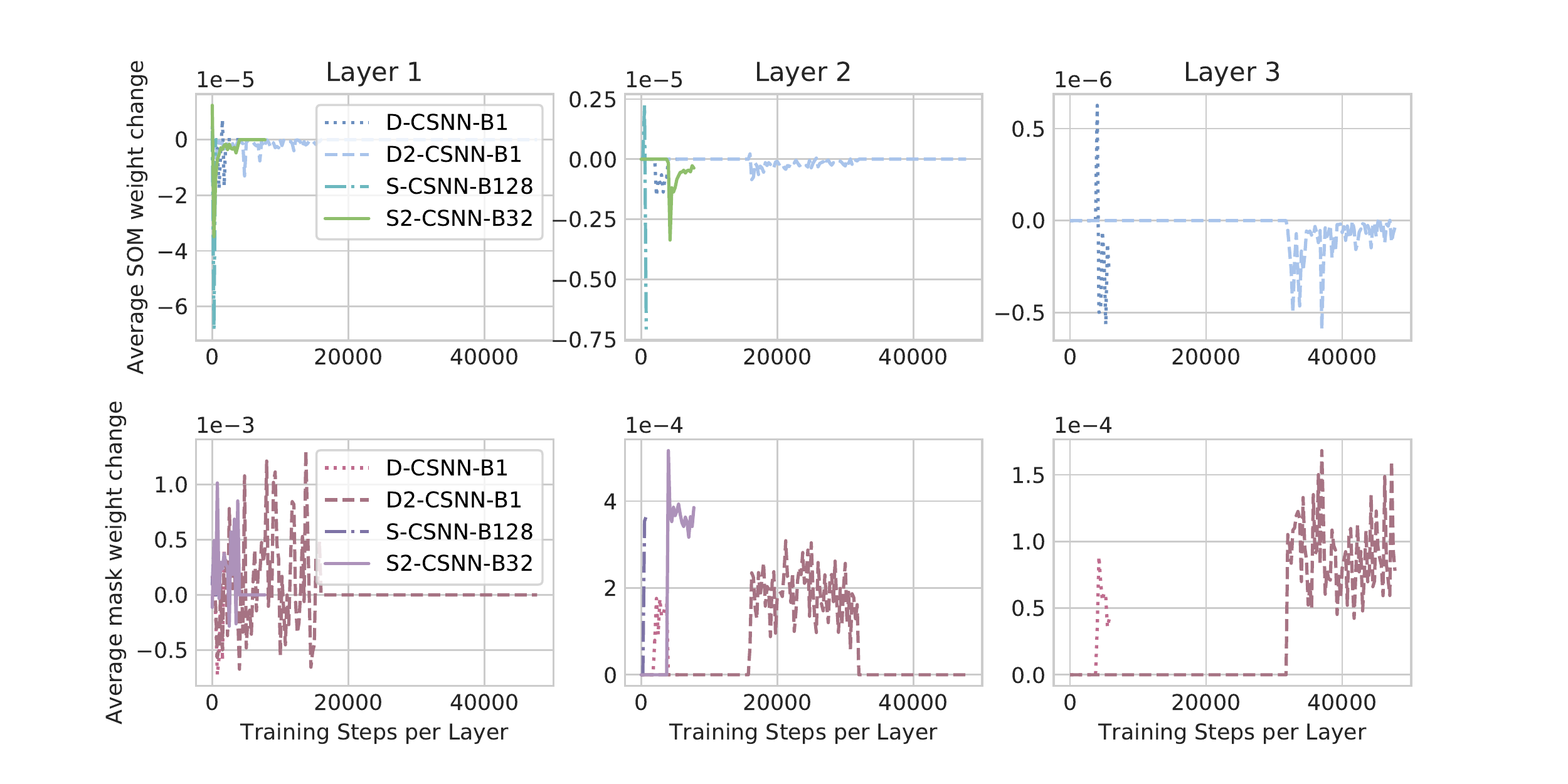}
	\end{center}
	\vspace{-3ex}
	\caption[Mask and \acs{SOM} weight changes.]{Mask and \acs{SOM} weight changes. The first row shows the average \acs{SOM} weight changes during the training of each layer. We can see that each layer was trained bottom-up in its defined training interval. The \acs{SOM} weights show convergence, which takes longer in deeper layers (and possibly could be improved when learning rates and neighborhood coefficients change over the course of the training). We suspect that the reasons for the slower convergence of deeper layers are the increased kernel and \acs{SOM} map sizes. The second row shows average mask weight changes. Best viewed in color.}
	\label{fig:csnns:app:weight_changes}
\end{figure}

\begin{figure}[t]
	\begin{center}
		\includegraphics[width=1.0\linewidth]{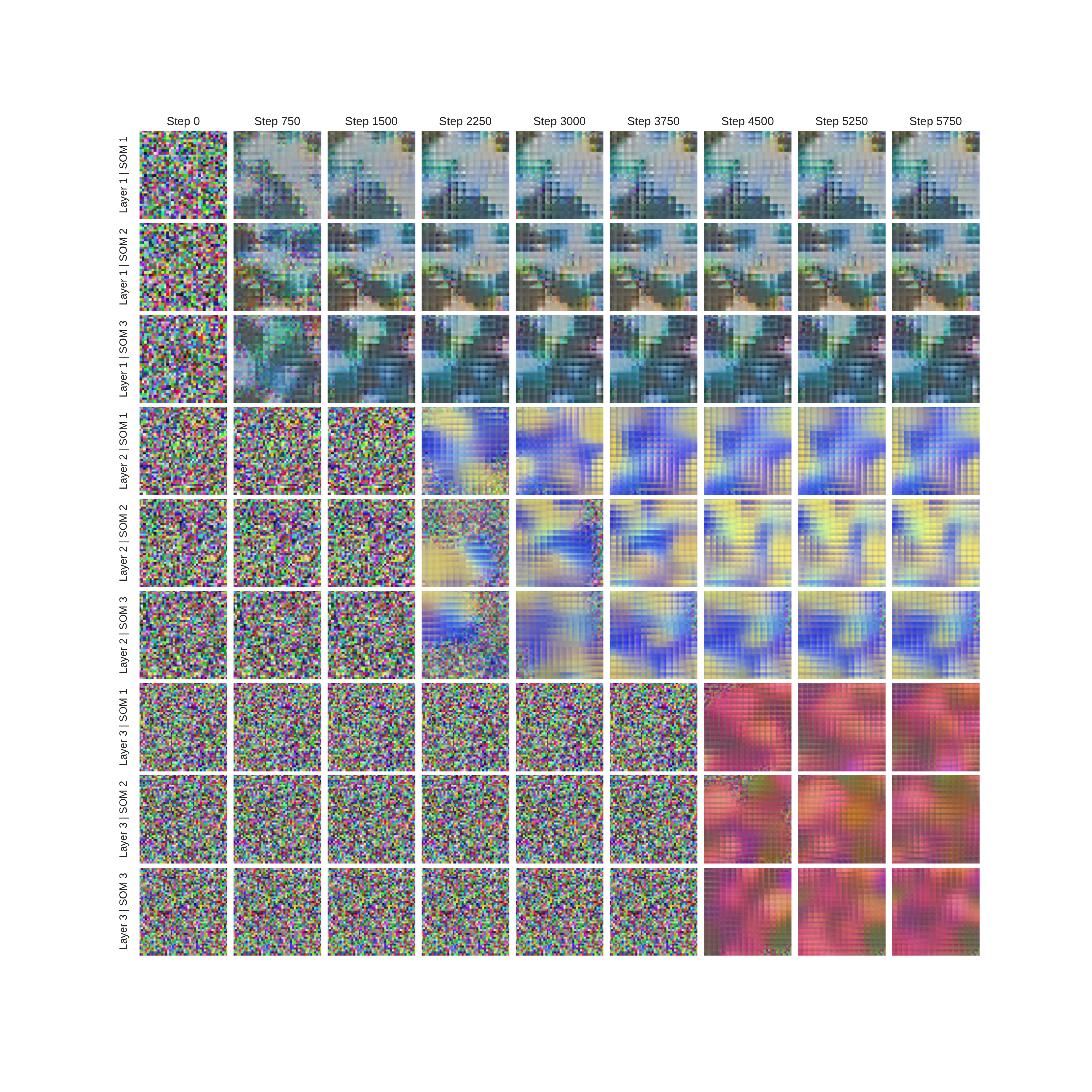}
	\end{center}
	\vspace{-10ex}
	\caption[\acs{SOM} weights during training.]{\acs{SOM} weights during training. Here we show the layer's \acs{SOM} weights at their \acs{SOM} grid and map position throughout the training process of the D-CSNN. Each \acs{SOM} image is formed by taking the weights from all \acs{SOM} neurons of the map at the current training step, reshaping them to the patch shape (3D), and forming an image by placing them according to the corresponding \acs{SOM} neuron coordinates in the \acs{SOM} grid. For deeper layers, we only show a slice of depth three through the \acs{SOM} weights since it is hard to visualize kernel (\acs{SOM} weight) depths larger than three. We can see that all \acs{SOM} weights have been trained bottom-up in their defined training intervals. Furthermore, this figure clearly shows self-organization. We want to note that the individual \acs{SOM} weights in layer 1 do not necessarily represent color due to the masks, as shown in the following \autoref{fig:csnns:app:bmu_images_1}, \autoref{fig:csnns:app:bmu_images_2}, and \autoref{fig:csnns:app:bmu_images_3}; the reason for that are the mask weights. Best viewed in color.}
	\label{fig:csnns:app:som_weights_during_training}
\end{figure}

\begin{figure}[t]
	\begin{center}
		\includegraphics[width=1.0\linewidth]{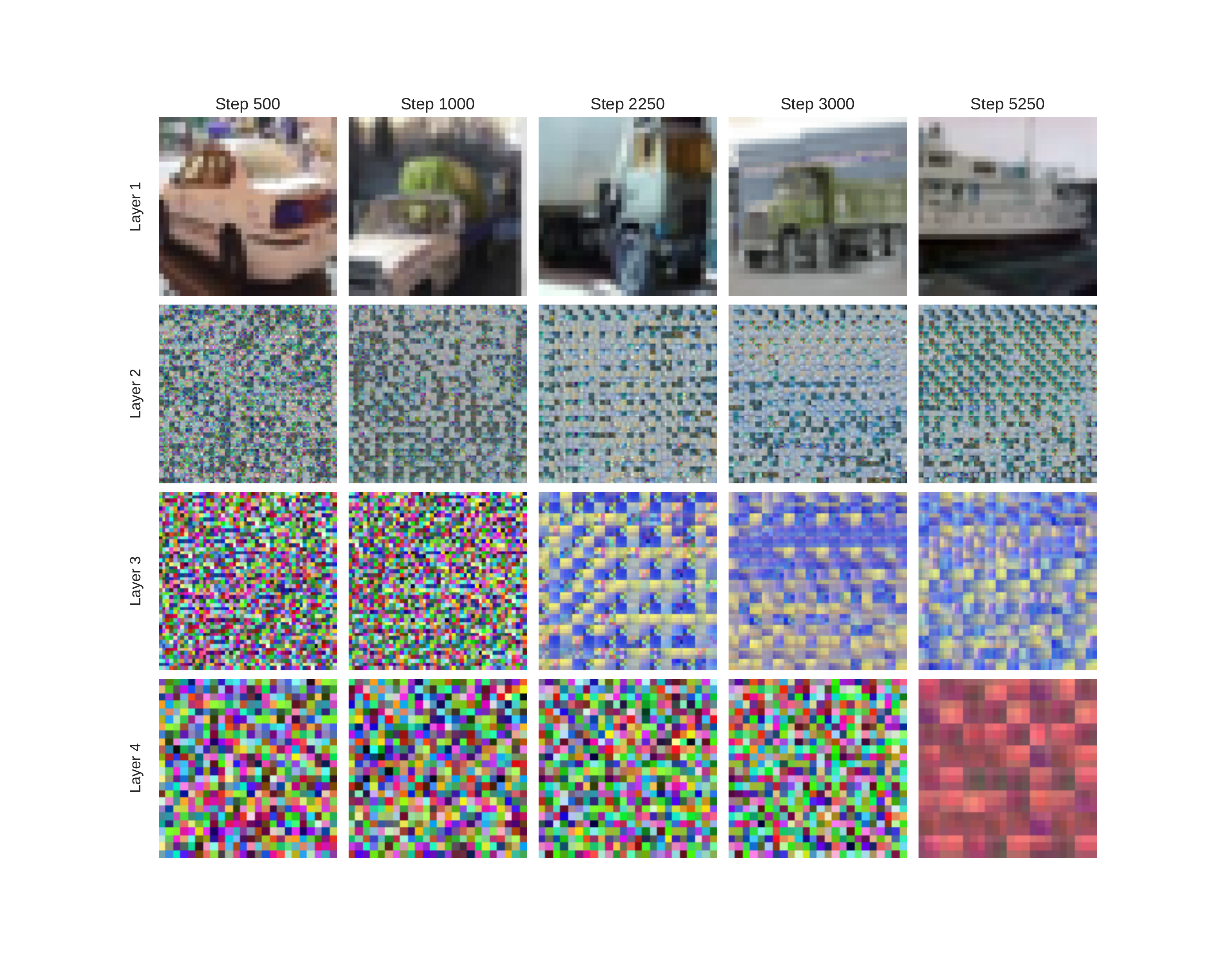}
	\end{center}
	\vspace{-8ex}
	\caption[More \acs{BMU} images for each layer of the D-CSNN.]{More \acs{BMU} images for each layer of the D-CSNN. Best viewed in color.}
	\label{fig:csnns:app:bmu_images_2}
\end{figure}

\begin{figure}[t]
	\begin{center}
		\includegraphics[width=1.0\linewidth]{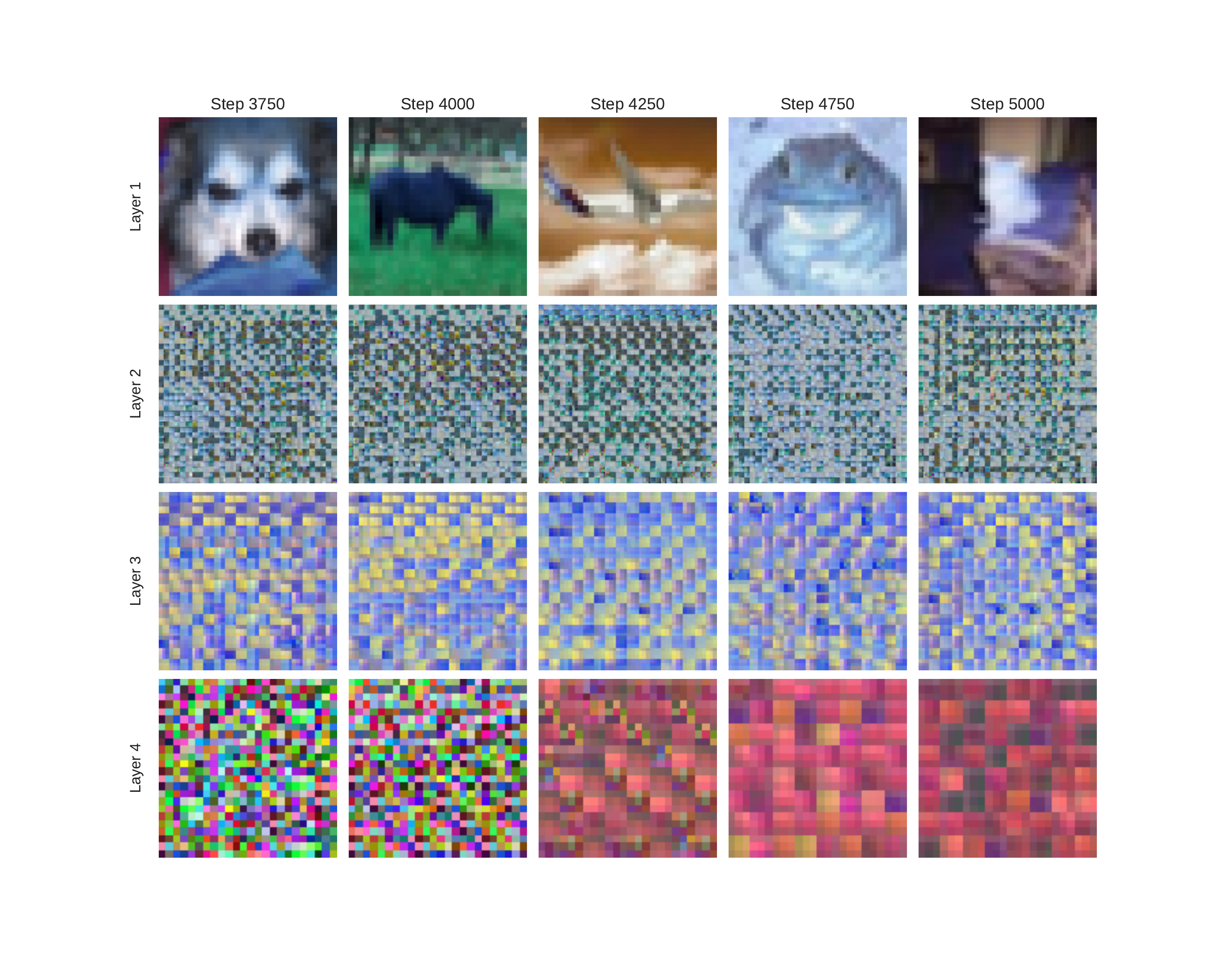}
	\end{center}
	\vspace{-8ex}
	\caption[More \acs{BMU} images for each layer of the D-CSNN.]{More \acs{BMU} images for each layer of the D-CSNN. Best viewed in color.}
	\label{fig:csnns:app:bmu_images_3}
\end{figure}

\begin{figure}[t]
	\begin{center}
		\includegraphics[width=1.0\linewidth]{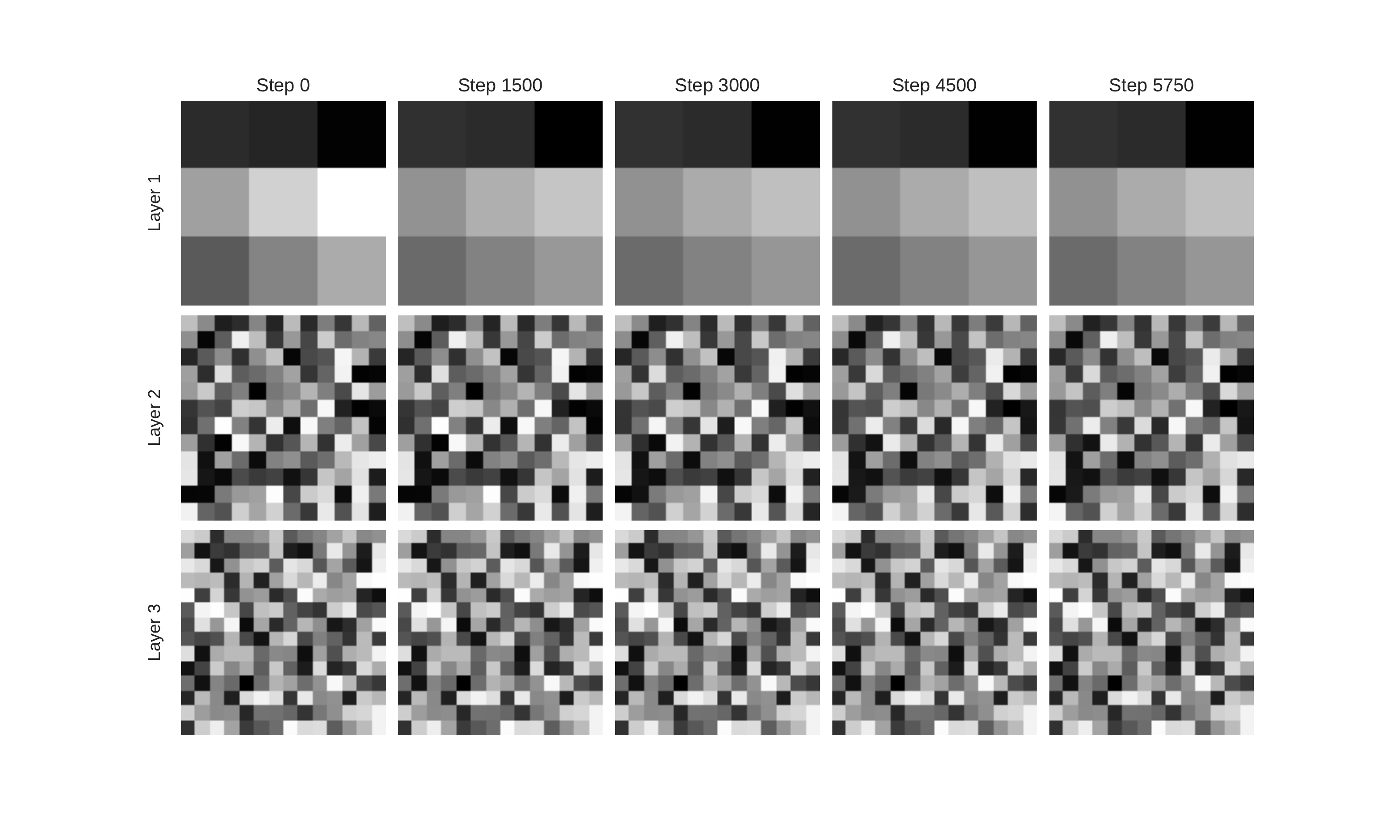}
	\end{center}
	\vspace{-6ex}
	\caption[Mask images.]{Mask images. Here we show a slice of depth one through the mask values of one mask for each layer, where white corresponds to $1.0$ and black to $-1.0$. One can see that throughout the training process, these values change slightly, indicating that the local mask learning chooses preferred inputs.}
	\label{fig:csnns:app:mask_images}
\end{figure}

\begin{figure}[t]
	\begin{center}
		\includegraphics[width=1.0\linewidth]{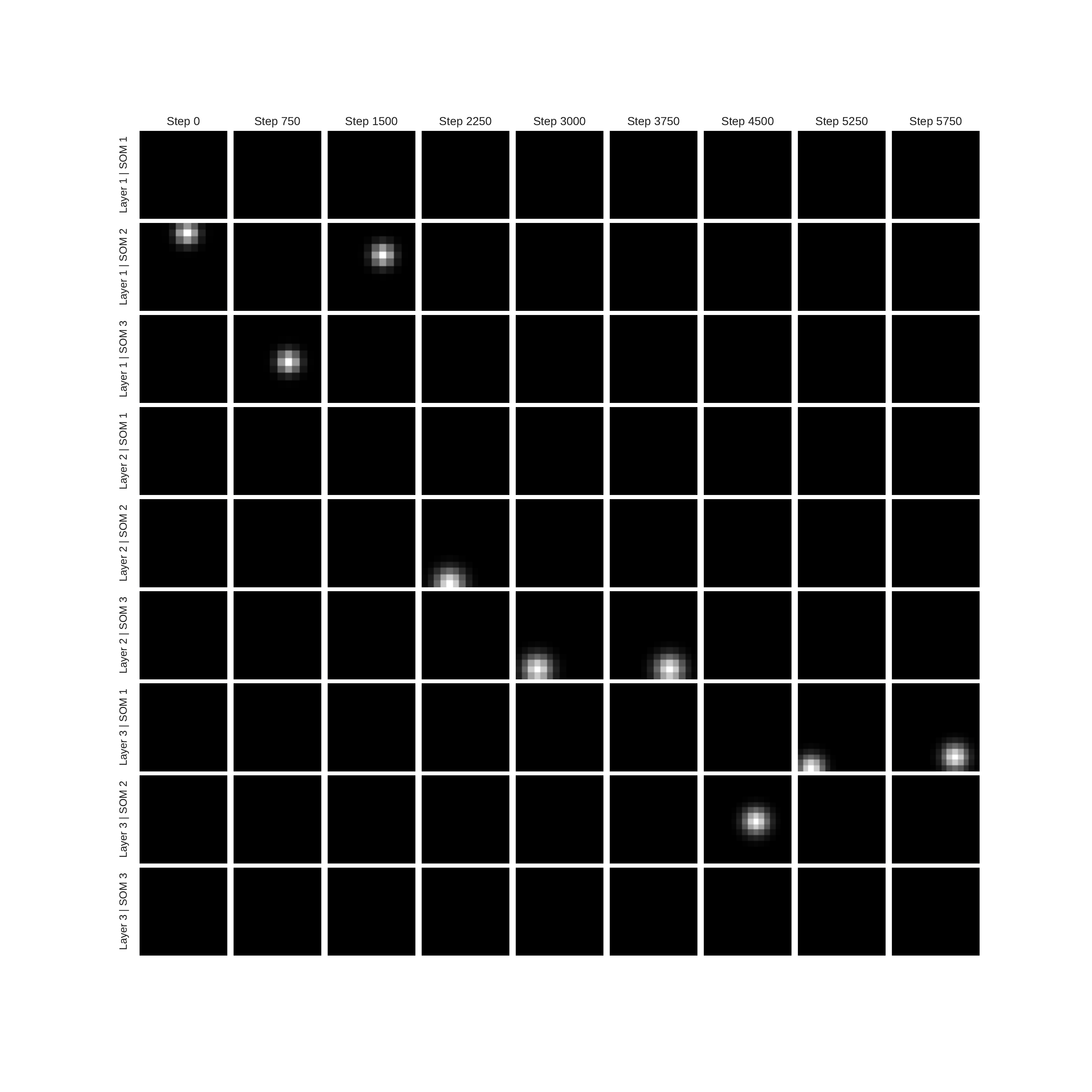}
	\end{center}
	\vspace{-10ex}
	\caption[Neighborhood coefficients.]{Neighborhood coefficients. Here we show the neighborhood coefficients of one spatial activation for each layer and map, where white corresponds to $1.0$ and black to $0.0$. To create one image, the flat vector containing the neighborhood coefficients is reshaped into the shape of the 2D \acs{SOM} grid. It can be seen that the neighborhood is Gaussian and that only the map with the best \acs{BMU} have been updated. Furthermore, it can be seen that layers are only trained in their defined training intervals $[y, x]$.}
	\label{fig:csnns:app:neighborhood_coefficients}
\end{figure}
\chapter{Supplementary Material:\\Investigating the Objective Function Mismatch}
\label{app:01}

\section{Proofs}
We measure our metrics on the mean losses during cross-validation instead of calculating the metrics for each round and taking the average. We prove that both variants are equivalent for $\mathrm{M3}$ while measuring $\mathrm{SM3}$ on the mean losses leads to a lower bound, given that all models converge at step $s_n$. 
\begin{proposition} 
	The $\mathrm{MM3}$ of the average metric value tuples $\frac{1}{h}\sum\limits_{0<c\leq h}{M}^{P}_c$ and $\frac{1}{h}\sum\limits_{0<c\leq h}{M}^{T}_c$ with $0<c\leq h$ is equivalent to the average $\mathrm{MM3}$ of the individual tuples $\frac{1}{h}\sum\limits_{0<c\leq h} \mathrm{MM3}({M}^{T}_c, M^{P}_c)$, given that the tuples are measured for the same steps $S$ and converge at the same step $s_n$.
\end{proposition}
\textit{Proof.}
\begin{align}
	\footnotesize
	\small
	\frac{1}{h}\sum\limits_{0<c\leq h} \mathrm{MM3}({M}^{T}_c, M^{P}_c) &= \frac{1}{h}\sum\limits_{0<c\leq h} \frac{1}{n}\sum\limits_{0<i\leq n} (m_{ic}^{T}-m_{ic}^{P})\\
	&=\frac{1}{n} \sum\limits_{0<i\leq n} \frac{1}{h}\sum\limits_{0<c\leq h}(m_{ic}^{T}-m_{ic}^{P}) \\
	&=\frac{1}{n} \sum\limits_{0<i\leq n} \left(\frac{1}{h}\sum\limits_{0<c\leq h}m_{ic}^{T}-\frac{1}{h}\sum\limits_{0<c\leq h}m_{ic}^{P}\right)\\ 
	&=\mathrm{MM3}\left(\frac{1}{h}\sum\limits_{0<c\leq h}{M}^{T}_c, \frac{1}{h}\sum\limits_{0<c\leq h}M^{P}_c\right)
\end{align}

\begin{corollary}
	$\frac{1}{h}\sum\limits_{0<c\leq h} \mathrm{M3}({m}^{T}_c, m^{P}_c) = \mathrm{M3}( \frac{1}{h}\sum\limits_{0<c\leq h}{m}^{T}_c,  \frac{1}{h}\sum\limits_{0<c\leq h}m^{P}_c)$
\end{corollary}
\newpage
\begin{proposition} 
	The $\mathrm{MSM3}$ of the average metric value tuples $\frac{1}{h}\sum\limits_{0<c\leq h}{M}^{P}_c$ and $\frac{1}{h}\sum\limits_{0<c\leq h}{M}^{T}_c$ with $0<c\leq h$ is a lower bound of the average $\mathrm{MSM3}$ of the individual tuples $\frac{1}{h}\sum\limits_{0<c\leq h} \mathrm{MSM3}({M}^{T}_c)$, given that the tuples are measured for the same steps $S$ and converge at the same step $s_n$.
\end{proposition}
\textit{Proof.}	
\begin{align}
	\footnotesize
	\small
	\frac{1}{h}\sum\limits_{0<c\leq h} \mathrm{MSM3}({M}^{T}_c)  &= \frac{1}{h}\sum\limits_{0<c\leq h} \frac{1}{n}\sum\limits_{0<i\leq n} \left(m_{ic}^{T} - \min_{0<j\leq i}(m_{jc}^{T}) \right)\\
	&=\frac{1}{n}\sum\limits_{0<i\leq n} \frac{1}{h}\sum\limits_{0<c\leq h} \left(m_{ic}^{T} - \min_{0<j\leq i}(m_{jc}^{T}) \right)\\
	&=\frac{1}{n}\sum\limits_{0<i\leq n} \left( \frac{1}{h}\sum\limits_{0<c\leq h} m_{ic}^{T} - \frac{1}{h}\sum\limits_{0<c\leq h}  \min_{0<j\leq i}(m_{jc}^{T}) \right)\\ 
	& \geq\frac{1}{n}\sum\limits_{0<i\leq n} \left( \frac{1}{h}\sum\limits_{0<c\leq h} m_{ic}^{T} - \min_{0<j\leq i}(\frac{1}{h}\sum\limits_{0<c\leq h}  m_{jc}^{T}) \right) \\ & \text{\tiny since $\frac{1}{h}\sum\limits_{0<c\leq h} \min_{0<j\leq i}(m_{jc}^{T}) \leq \min_{0<j\leq i}(\frac{1}{h}\sum\limits_{0<c\leq h} m_{jc}^{T})$}\\
	&\geq\mathrm{MSM3}\left(\frac{1}{h}\sum\limits_{0<c\leq h}{M}^{T}_c \right) 
\end{align}

\begin{corollary}
	$\frac{1}{h}\sum\limits_{0<c\leq h} \mathrm{SM3}({m}^{T}_c, m^{P}_c) \geq \mathrm{SM3}(\frac{1}{h}\sum\limits_{0<c\leq h}{m}^{T}_c, \frac{1}{h}\sum\limits_{0<c\leq h}m^{P}_c)$
\end{corollary}

\newpage
\section{Additional Model and Training Details}
\noindent \textbf{\acs{CNN} encoders}: We consider a family of convolutional encoders with four Conv-BatchNorm-ReLU layers. Filter widths are $[32, 64, 128, f]$ and paddings are $"valid"$. For input sizes of $32\times32$ (Cifar10, Cifar100), kernel sizes are $[3,3,3,2]$ and strides are $[2,2,2,1]$; for input sizes of $64\times64$ (3dshapes, PCam), kernel sizes are $[4,4,4,3]$ and strides are $[2,2,2,2]$. Weights are initialized with the standard TensorFlow initialization (kernel\_initializer="glorot\_uniform", bias\_initializer="zeros"). We vary $f$ in $[4,32,128,256,512,1024]$ for our experiments on representation sizes. For all other experiments $f=256$. 
\\
\\
\textbf{\acs{CNN} image decoders}:
We consider a family of decoders with transposed convolutions with three TranConv-BatchNorm-ReLU layers followed by a TranConv-BatchNorm-Sigmoid layer. Filter widths are $[128, 64, 32, 3]$ and paddings are $"valid"$. For input sizes of $32\times32$ (Cifar10, Cifar100), kernel sizes are $[4,4,4,3]$ and strides are $[2,2,2,1]$; for input sizes of $64\times64$ (3dshapes, PCam), kernel sizes are $[4,4,5,4]$ and strides are $[2,2,2,2]$. Weights are initialized with the standard TensorFlow initialization (kernel\_initializer="glorot\_uniform", bias\_initializer="zeros").
\\
\\
\textbf{\acs{CNN}/ResNet head for rotation prediction}: For our CNN encoder, we use a fully-connected layer with $4$ neurons and softmax activation as the head to predict the four different rotations. We use the standard TensorFlow initialization (kernel\_initializer="glorot\_uniform", bias\_initializer="zeros").
For our ResNet decoder, we initialize with (kernel\_initializer=RandomNormal(stddev=.01), bias\_initializer=\\"zeros").
\\
\\
\textbf{\acs{CNN}/ResNet heads for contrastive learning}: For our CNN encoder, we use a two-layer \acs{MLP} with an FC-BatchNorm-ReLU layer followed by an FC-BatchNorm-Softmax layer as the projection head for contrastive learning. The numbers of neurons are $[f,128]$. We use the standard TensorFlow initialization (kernel\_initializer=\\"glorot\_uniform", bias\_initializer="zeros"). We vary $f$ in $[4,32,128,256,512,1024]$ for our experiments on representation sizes. For all other experiments $f=256$. For our ResNet head, the numbers of neurons are $[512,128]$, and we initialize as in \cite{chen2020simple}.
\\
\\	
\textbf{Target models}: For our linear target model, we use a fully-connected layer with $num\_classes$ neurons and a softmax activation. For our two- and three-layer nonlinear models, we add layers consisting of [256] and [512, 256] hidden units with batch normalization followed by ReLu activations respectively. Weights are initialized with the standard TensorFlow initialization (kernel\_initializer=="glorot\_uniform", bias\_initializer="zeros").
\\
\\
\textbf{Hardware}: We run our experiments on two servers, each containing four Nvidia GeForce RTX 2080 Ti GPUs. 
\\
\\	
\textbf{Mismatch evaluation}: In \autoref{tab:ofm:app:train_eval_details}, we show additional details about training, evaluation, and measurements.
\begin{table}[h]
	\RawFloats
	\scriptsize
	\caption[Information about measurements and training.]{Information about measurements and training. Under (Rep. Size, TMC, Augs), we refer to all models that have been trained with different representations, target model complexities, and augmentations.} 
	\vspace{-1ex}
	\label{tab:ofm:app:train_eval_details}
	\begin{center}
		\scalebox{0.875}{%
			\setlength{\tabcolsep}{0.4em}
			\begin{tabular}{lccccc}
				\toprule
				&Measurement Epochs &\begin{tabular}{@{}c@{}}Convergence\\Criterium	\end{tabular} &\begin{tabular}{@{}c@{}}Pretext Model\\Training Epochs\end{tabular}	&\begin{tabular}{@{}c@{}}Target Model\\Training Epochs\end{tabular}	&Validation\\
				\midrule
				\textbf{\textit{Rep. Size, TMC, Augs}}: &&&&& \\ 
				\acs{CAE} (Cifar10)   &(0,5,20,50,100,...,400)&Patience: 3&400&500&5-fold cross-validation\\
				\acs{DCAE} (Cifar10)  &(0,5,20,50,100,...,400)&Patience: 3&400&500&5-fold cross-validation\\
				\acs{CCAE} (Cifar100)  &(0,5,20,50,100,...,400)&Patience: 6&400&500&5-fold cross-validation\\
				\acs{CCAE} (PCam)  &(0,10,50,100,150,200,300,...,800)&Patience: 10&800&500&5-fold cross-validation\\	
				\acs{RCAE} (PCam)   &(0,200,400,...,2000)&Patience: 30&2000&500&5-fold cross-validation\\
				\acs{SCLCAE} (3dshapes)  &(0,10,50,100,150,200,300,...,600)&Patience: 15&60 0&100&5-fold cross-validation\\
				\midrule
				\textbf{\textit{Target Task Type}}:&&&&& \\
				\acs{CAE} (3dshapes)  &(0,10,30,50,100,...,400)&Patience: 3&400&100&5-fold cross-validation\\
				\acs{DCAE} (3dshapes)  &(0,10,30,50,100,...,400)&Patience: 3&400&100&5-fold cross-validation\\
				\acs{CCAE} (3dshapes)  &(0,10,30,50,100,...,400)&Patience: 3&400&100&5-fold cross-validation\\
				\acs{RCAE} (3dshapes)  &(0,10,30,50,100,...,400)&Patience: 3&400&100&5-fold cross-validation\\
				\acs{SCLCAE} (3dshapes)  &(0,10,30,50,100,...,600)&Patience: 3&600&100&5-fold cross-validation\\
				\midrule
				\textbf{\textit{Stability}}: &&&&& \\ 
				CAE100E (Cifar10)   &(0,1,...,100)&Epoch 100&100&500&5-fold cross-validation\\
				\acs{CAE} (Cifar10)   &(0,5,20,50,100,...,400)&Epoch 400&400&500&5-fold cross-validation\\
				CAENoCrossVal (Cifar10)   &(0,5,20,50,100,...,400)&Epoch 400&400&500&5 $\times$ same split\\
				\midrule
				\textbf{\textit{ResNets}}: &&&&& \\ 
				RResNet18 (Cifar10)   &(0,50,100,200,400,...,4000)&Patience: 30&4000&600&5-fold cross-validation\\
				SCLResNet18 (Cifar10)   &(0,50,100,200,400,,...,4000)&Epoch 4000&4000&600&3-fold cross-validation\\
				SCLResNet18 (Cifar100)   &(0,50,100,200,400,...,4000)&Epoch 4000&4000&600&3-fold cross-validation\\
				SCLResNet18 (PCam)   &(0,400,800,...,5000)&Patience: 60&5000&500&5-fold cross-validation\\
				\midrule
				\textbf{\textit{ResNets Rep. Size}}: &&&&& \\ 
				RResNet18(Cifar10)   &(0,100,200,...,1000,1200,...,3000)&Patience: 30&3000&700&5-fold cross-validation\\
				\bottomrule
		\end{tabular}}
	\end{center}
	\vspace{-2ex}
\end{table}

\newpage
\section{Additional Evidence}

\begin{table}[t]
	\RawFloats
	\caption[Detailed version of \acs{CAE} (Cifar10) and \acs{DCAE} (Cifar10) from \autoref{tab:ofm:ablation_rep_tm_augs}.]{Detailed version of \acs{CAE} (Cifar10) and \acs{DCAE} (Cifar10) from \autoref{tab:ofm:ablation_rep_tm_augs}.} 
	\vspace{-1ex}
	\label{tab:ofm:app:ablation_rep_tm_augs_details_1}
	\begin{center}
		\scalebox{0.8}{%
			\setlength{\tabcolsep}{0.4em}
			\begin{tabular}{lcccccc}
				\toprule
				&\multicolumn{3}{c}{\acs{CAE} (Cifar10)} &\multicolumn{3}{c}{\acs{DCAE} (Cifar10)} 
				\\	
				\cmidrule(lr){2-4}
				\cmidrule(lr){5-7}
				& $\mathrm{ACC}$ & $\mathrm{cSM3}$ & $\mathrm{MOFM}$ & $\mathrm{ACC}$  & $\mathrm{cSM3}$ & $\mathrm{MOFM}$\\
				\midrule
				\textbf{\textit{Rep. Size}}: &&&& \\ 
				2x2x4   	
				&$27.94^{+1.47}_{-0.76}$  &$\mathbf{0.00}^{+0.00}_{-0.00}$	&$\mathbf{0.00}^{+29.77}_{-0.00}$
				&$28.06^{+0.70}_{-1.65}$ &$0.05^{+0.12}_{-0.05}$ &$\mathbf{0.00}^{+14.27}_{-0.00}$ 		\\
				
				2x2x32   
				&$36.57^{+0.92}_{-0.92}$  &$0.07^{+0.09}_{-0.07}$& $1.99^{+3.54}_{-1.44}$ 
				&$36.20^{+0.52}_{-0.50}$ &$\mathbf{0.00}^{+0.00}_{-0.00}$ &$3.20^{+10.24}_{-0.59}$ \\
				
				2x2x128   
				&$41.94^{+0.45}_{-0.47}$  &$0.20^{+0.42}_{-0.20}$ &$10.10^{+6.24}_{-3.51}$  
				&$41.79^{+0.46}_{-0.49}$  &$0.06^{+0.11}_{-0.06}$ &$5.51^{+6.79}_{-5.32}$  \\
				
				2x2x256   
				&$44.69^{+0.69}_{-0.41}$  &$0.75^{+0.38}_{-0.22}$ &$11.14^{+5.32}_{-2.72}$   
				&$45.42^{+0.55}_{-0.62}$   &$0.69^{+0.43}_{-0.69}$ &$5.17^{+2.98}_{-2.16}$ \\
				
				2x2x512   
				&$48.13^{+0.58}_{-0.88}$  &$0.43^{+0.38}_{-0.43}$ &$5.28^{+1.93}_{-2.98}$   
				&$49.04^{+0.29}_{-0.29}$  &$0.36^{+0.75}_{-0.36}$  &$1.25^{+1.81}_{-1.19}$  \\
				
				2x2x1024   
				&$\mathbf{51.42}^{+0.43}_{-0.38}$  &$\mathbf{0.24}^{+0.32}_{-0.24}$ &$\mathbf{0.25}^{+1.26}_{-0.17}$     
				&$\mathbf{53.82}^{+0.57}_{-0.82}$  &$\mathbf{0.03}^{+0.06}_{-0.03}$  &$\mathbf{0.00}^{+0.04}_{-0.00}$ \\
				\midrule
				\textbf{\textit{Target Model}}:&&&&  \\
				FC   
				&$44.69^{+0.69}_{-0.41}$    &$0.75^{+0.38}_{-0.22}$ &$11.14^{+5.32}_{-2.72}$ 
				&$45.42^{+0.55}_{-0.62}$   &$0.69^{+0.43}_{-0.69}$ &$5.17^{+2.98}_{-2.16}$  \\
				
				2FC   
				&$56.72^{+0.50}_{-0.42}$  &$\mathbf{0.03}^{+0.10}_{-0.03}$ &$5.68^{+2.87}_{-3.04}$  
				&$57.26^{+0.68}_{-0.52}$  &$\mathbf{0.00}^{+0.00}_{-0.00}$  &$5.14^{+2.68}_{-1.98}$  \\
				
				3FC  
				&$\mathbf{63.05}^{+0.74}_{-0.72}$  &$\mathbf{0.03}^{+0.11}_{-0.03}$ &$\mathbf{3.94}^{+0.93}_{-1.61}$     
				&$\mathbf{63.33}^{+0.23}_{-0.39}$  &$\mathbf{0.00}^{+0.00}_{-0.00}$ &$\mathbf{3.17}^{+1.51}_{-0.89}$ \\
				\midrule
				\textbf{\textit{Augmentations}}: &&&& \\
				All   
				&$44.69^{+0.69}_{-0.41}$   &$\mathbf{0.75}^{+0.38}_{-0.22}$ &$11.14^{+5.32}_{-2.72}$   
				&$45.42^{+0.55}_{-0.62}$   &$0.69^{+0.43}_{-0.69}$ &$5.17^{+2.98}_{-2.16}$    \\
				
				NoJitter   
				&$46.30^{+0.50}_{-0.46}$  &$0.99^{+0.59}_{-0.47}$  &$\mathbf{10.99}^{+1.24}_{-3.25}$  
				&$47.27^{+0.46}_{-0.50}$  &$\mathbf{0.50}^{+0.45}_{-0.50}$  &$2.33^{+1.74}_{-1.57}$ \\
				
				NoJitterNoFlip  
				&$\mathbf{46.48}^{+0.51}_{-0.75}$ &$1.00^{+0.48}_{-0.60}$ &$12.51^{+1.46}_{-1.74}$   
				&$\mathbf{47.38}^{+0.36}_{-0.44}$  &$0.55^{+0.51}_{-0.55}$ &$\mathbf{1.73}^{+4.43}_{-1.36}$ \\
				\bottomrule
		\end{tabular}}
	\end{center}
	\vspace{-2ex}
\end{table}

\begin{table}[t]
	\RawFloats
	\caption[Detailed version of \acs{CCAE} (Cifar100) and \acs{RCAE} (PCam) from \autoref{tab:ofm:ablation_rep_tm_augs}.]{Detailed version of \acs{CCAE} (Cifar100) and \acs{RCAE} (PCam) from \autoref{tab:ofm:ablation_rep_tm_augs}.} 
	\vspace{-1ex}
	\label{tab:ofm:app:ablation_rep_tm_augs_details_2}
	\begin{center}
		\scalebox{0.8}{%
			\setlength{\tabcolsep}{0.4em}
			\begin{tabular}{lccccccc}
				\toprule
				&\multicolumn{3}{c}{\acs{CCAE} (Cifar100)} &\multicolumn{4}{c}{\acs{RCAE} (PCam)}\\	
				\cmidrule(lr){2-4}
				\cmidrule(lr){5-8}
				& $\mathrm{ACC}$ & $\mathrm{cSM3}$ & $\mathrm{MOFM}$ &  $\mathrm{ACC}$ & $\mathrm{cSM3}$ &  $\mathrm{MOFM}$ &$\mathrm{MM3}$  \\
				\midrule
				\textbf{\textit{Rep. Size}}: &&&& \\ 
				2x2x4   	
				&$9.66^{+0.30}_{-0.45}$&$0.28^{+0.19}_{-0.13}$&$1.54^{+2.09}_{-0.98}$
				&$67.62^{+1.18}_{-2.05}$&$5.38^{+2.15}_{-1.66}$&$4.15^{+37.30}_{-4.15}$&$-22.26^{+2.96}_{-3.02}$ \\
				
				2x2x32   
				&$17.63^{+0.36}_{-0.51}$&$0.65^{+0.28}_{-0.64}$&$3.64^{+4.14}_{-2.04}$
				&$72.83^{+1.17}_{-1.72}$&$3.34^{+0.74}_{-0.52}$&$7.92^{+18.08}_{-1.46}$&$-21.09^{+1.90}_{-2.48}$\\
				
				2x2x128   
				&$24.36^{+0.56}_{-0.47}$&$0.51^{+0.24}_{-0.24}$&$0.81^{+1.60}_{-0.45}$
				&$78.17^{+0.63}_{-0.80}$&$1.03^{+0.88}_{-0.96}$&$4.04^{+4.35}_{-0.70}$&$-23.47^{+0.55}_{-0.65}$ \\
				
				2x2x256   
				&$28.36^{+0.78}_{-0.68}$&$0.17^{+0.37}_{-0.17}$&$\mathbf{0.00}^{+0.00}_{-0.00}$
				&$79.55^{+1.18}_{-1.07}$&$0.44^{+0.79}_{-0.44}$&$\mathbf{0.00}^{+3.95}_{-0.85}$&$-27.60^{+1.22}_{-1.00}$\\
				
				2x2x512   
				&$32.02^{+0.51}_{-0.74}$&$\mathbf{0.00}^{+0.02}_{-0.00}$&$\mathbf{0.00}^{+0.00}_{-0.00}$
				&$80.80^{+0.66}_{-0.63}$&$0.18^{+0.15}_{-0.18}$&$\mathbf{0.00}^{+1.13}_{-0.00}$&$\mathbf{-28.03}^{+1.28}_{-1.35}$  \\
				
				2x2x1024   
				&$\mathbf{34.89}^{+0.41}_{-0.83}$&$\mathbf{0.00}^{+0.00}_{-0.00}$&$\mathbf{0.00}^{+0.00}_{-0.00}$
				&$\mathbf{82.20}^{+0.76}_{-0.68}$&$\mathbf{0.09}^{+0.12}_{-0.09}$&$\mathbf{0.00}^{+0.34}_{-0.00}$&$-26.56^{+1.14}_{-1.18}$ \\
				\midrule
				\textbf{\textit{Target Model}}:&&&&  \\
				FC   
				&$32.02^{+0.51}_{-0.74}$&$\mathbf{0.00}^{+0.02}_{-0.00}$&$\mathbf{0.00}^{+0.00}_{-0.00}$
				&$78.17^{+0.63}_{-0.80}$&$1.03^{+0.88}_{-0.96}$&$4.04^{+4.35}_{-0.70}$&$-23.47^{+0.55}_{-0.65}$ \\
				
				2FC   
				&$36.01^{+0.31}_{-0.27}$&$0.08^{+0.32}_{-0.08}$&$\mathbf{0.00}^{+0.39}_{-0.00}$
				&$82.99^{+0.65}_{-0.62}$&$\mathbf{0.31}^{+0.31}_{-0.31}$&$0.76^{+1.76}_{-0.04}$&$-28.56^{+0.81}_{-0.72}$\\
				
				3FC  
				&$\mathbf{38.40}^{+0.34}_{-0.50}$&$0.12^{+0.28}_{-0.12}$&$0.02^{+0.44}_{-0.09}$
				&$\mathbf{84.18}^{+0.62}_{-0.45}$&$0.37^{+0.21}_{-0.37}$&$\mathbf{0.61}^{+3.68}_{-0.51}$&$\mathbf{-29.61}^{+0.71}_{-0.61}$ \\
				\midrule
				\textbf{\textit{Augmentations}}: &&&& \\
				All   
				&$28.36^{+0.78}_{-0.68}$&$\mathbf{0.17}^{+0.37}_{-0.17}$&$\mathbf{0.00}^{+0.00}_{-0.00}$
				&$78.17^{+0.63}_{-0.80}$&$1.03^{+0.88}_{-0.96}$&$4.04^{+4.35}_{-0.70}$&$-23.47^{+0.55}_{-0.65}$\\
				
				NoJitter 
				&-&-& -  &$\mathbf{80.88}^{+0.88}_{-0.94}$&$\mathbf{0.58}^{+0.34}_{-0.58}$&$\mathbf{0.10}^{+5.52}_{-0.04}$&$\mathbf{-28.60}^{+2.23}_{-1.18}$ \\
				NoJitterNoFlip 
				&-&-&-
				& $80.55^{+0.73}_{-0.55}$&$1.51^{+0.57}_{-1.09}$&$7.33^{+9.76}_{-1.78}$&$-10.60^{+1.21}_{-1.88}$
				\\
				NoFlip   
				&$\mathbf{29.33}^{+0.65}_{-0.61}$&$0.20^{+0.43}_{-0.20}$&$\mathbf{0.00}^{+0.22}_{-0.00}$ 
				&-&- &- &-\\
				\bottomrule
		\end{tabular}}
	\end{center}
	\vspace{-2ex}
\end{table}

\begin{table}[t]
	\RawFloats
	\caption[Detailed version of \acs{CCAE} (PCam) and \acs{SCLCAE} (3dshapes for object hue classification) from \autoref{tab:ofm:ablation_rep_tm_augs}.]{Detailed version of \acs{CCAE} (PCam) and \acs{SCLCAE} (3dshapes for object hue classification) from \autoref{tab:ofm:ablation_rep_tm_augs}.} 
	\vspace{-1ex}
	\label{tab:ofm:app:ablation_rep_tm_augs_details_3}
	\begin{center}
		\scalebox{0.75}{%
			\setlength{\tabcolsep}{0.4em}
			\begin{tabular}{lccccccc}
				\toprule
				&\multicolumn{3}{c}{\acs{CCAE} (PCam)}&\multicolumn{4}{c}{\acs{SCLCAE} (3dshapes)} \\
				\cmidrule(lr){2-4}
				\cmidrule(lr){5-8}	
				& $\mathrm{ACC}$ & $\mathrm{cSM3}$ & $\mathrm{MOFM}$  &   $\mathrm{ACC}$ & $\mathrm{cSM3}$ & $\mathrm{MOFM}$ &  $\mathrm{MM3}$\\
				\midrule
				\textbf{\textit{Rep. Size}}: &&&&& \\ 
				2x2x4   &$63.39^{+3.80}_{-1.58}$&$4.98^{+5.53}_{-3.53}$&$9.28^{+ \infty}_{-2.09}$ &$38.07^{+11.18}_{-10.06}$&$26.34^{+11.38}_{-9.69}$&$\infty$&$-7.99^{+1.83}_{-1.28}$\\
				2x2x32    &$72.73^{+2.98}_{-2.37}$&$5.17^{+2.95}_{-2.80}$&$34.30^{+16.31}_{-10.21}$ &$85.25^{+2.77}_{-2.36}$&$12.98^{+10.56}_{-12.24}$&$36.39^{+26.87}_{-29.73}$&$-57.67^{+6.98}_{-6.46}$\\
				2x2x128    &$78.66^{+0.46}_{-0.22}$&$0.32^{+0.47}_{-0.32}$&$0.10^{+3.51}_{-0.48}$ &$96.54^{+0.72}_{-0.83}$&$8.14^{+1.29}_{-2.53}$&$\mathbf{22.65}^{+13.03}_{-3.92}$&$\mathbf{-66.19}^{+1.24}_{-1.71}$\\
				2x2x256    &$79.97^{+0.68}_{-0.51}$&$0.43^{+0.51}_{-0.43}$&$0.87^{+4.56}_{-0.41}$
				&$98.65^{+0.55}_{-0.80}$&$6.52^{+0.62}_{-0.84}$&$27.65^{+5.66}_{-4.87}$&$-65.77^{+2.19}_{-1.12}$\\
				2x2x512  &  $82.34^{+0.66}_{-0.87}$&$0.20^{+0.59}_{-0.20}$&$0.07^{+0.24}_{-0.07}$ &$99.41^{+0.20}_{-0.25}$&$5.96^{+1.13}_{-0.40}$&$27.78^{+8.80}_{-7.10}$&$-63.61^{+1.75}_{-1.50}$\\
				2x2x1024     &$\mathbf{83.67}^{+0.45}_{-0.61}$&$\mathbf{0.00}^{+0.01}_{-0.00}$&$\mathbf{0.00}^{+0.20}_{-0.00}$    
				&$\mathbf{99.75}^{+0.04}_{-0.04}$&$\mathbf{4.76}^{+1.84}_{-1.36}$&$32.70^{+14.59}_{-7.15}$&$-57.92^{+2.29}_{-2.25}$\\
				\midrule
				\textbf{\textit{Target Model}}:&&&&&&  \\
				FC &$83.67^{+0.45}_{-0.61}$&$\mathbf{0.00}^{+0.01}_{-0.00}$&$\mathbf{0.00}^{+0.20}_{-0.00}$ &$38.07^{+11.18}_{-10.06}$&$6.52^{+0.62}_{-0.84}$&$\mathbf{27.65}^{+5.66}_{-4.87}$&$-65.77^{+2.19}_{-1.12}$\\
				2FC     &$89.17^{+0.50}_{-0.38}$&$\mathbf{0.00}^{+0.00}_{-0.00}$&$\mathbf{0.00}^{+0.12}_{-0.00}$ &$85.25^{+2.77}_{-2.36}$&$1.84^{+0.33}_{-0.66}$&$103.30^{+56.89}_{-21.22}$&$-70.49^{+1.41}_{-0.75}$\\
				3FC    &$\mathbf{90.59}^{+0.48}_{-0.60}$&$0.08^{+0.19}_{-0.08}$&$\mathbf{0.00}^{+0.44}_{-0.00}$ &$\mathbf{96.54}^{+0.72}_{-0.83}$&$\mathbf{0.91}^{+0.18}_{-0.30}$&$258.18^{+308.48}_{-88.99}$&$\mathbf{-71.26}^{+1.20}_{-0.77}$\\
				\midrule
				\textbf{\textit{Augmentations}}: &&&&&& \\
				All     &$79.97^{+0.68}_{-0.51}$&$0.43^{+0.51}_{-0.43}$&$0.87^{+4.56}_{-0.41}$ &$38.07^{+11.18}_{-10.06}$&$6.52^{+0.62}_{-0.84}$&$27.65^{+5.66}_{-4.87}$&$\mathbf{-65.77}^{+2.19}_{-1.12}$\\
				NoJitter     &- &-&- &$85.25^{+2.77}_{-2.36}$&$\mathbf{0.00}^{+0.02}_{-0.00}$&$0.01^{+0.07}_{-0.01}$&$-37.92^{+0.72}_{-0.89}$\\
				NoJitterNoFlip &-&-&-
				&$\mathbf{96.54}^{+0.72}_{-0.83}$&$\mathbf{0.00}^{+0.00}_{-0.00}$&$\mathbf{0.00}^{+0.04}_{-0.01}$&$-35.95^{+0.63}_{-0.82}$\\
				NoFlip   
				&$\textbf{80.70}^{+0.36}_{-0.57}$&$\textbf{0.41}^{+0.21}_{-0.30}$&$\textbf{0.04}^{+1.03}_{-0.20}$
				&-&- &- &-\\
				\bottomrule
		\end{tabular}}
	\end{center}
	\vspace{-2ex}
\end{table}

\begin{table}[t]
	\RawFloats
	\caption[Detailed version of \acs{CAE} and \acs{DCAE} from \autoref{tab:ofm:quantitativ_results_3dshapes}.]{Detailed version of \acs{CAE} and \acs{DCAE} from \autoref{tab:ofm:quantitativ_results_3dshapes}.} 
	\vspace{-1ex}
	\label{tab:ofm:app:quantitativ_results_3dshapes_details_1}
	\begin{center}
		\scalebox{0.8}{%
			\setlength{\tabcolsep}{0.4em}
			\begin{tabular}{lcccccc}
				\toprule
				&\multicolumn{3}{c}{\acs{CAE}} &\multicolumn{3}{c}{\acs{DCAE}} 
				\\	
				\cmidrule(lr){2-4}
				\cmidrule(lr){5-7}
				&  $\mathrm{ACC}$&  $\mathrm{cSM3}$ & $\mathrm{MOFM}$ &  
				$\mathrm{ACC}$&  $\mathrm{cSM3}$ & $\mathrm{MOFM}$ \\
				\midrule
				floor\_hue 	&$99.96^{+0.02}_{-0.02}$&$0.01^{+0.02}_{-0.01}$&$0.95^{+0.71}_{-0.51}$
				&$99.97^{+0.02}_{-0.06}$&$\mathbf{0.00}^{+0.00}_{-0.00}$&$1.28^{+1.11}_{-0.81}$\\
				wall\_hue   &$\mathbf{100.00}^{+0.00}_{-0.00}$&$0.02^{+0.04}_{-0.02}$&$32.03^{+6.38}_{-8.59}$ 
				&$\mathbf{99.99}^{+0.00}_{-0.01}$&$\mathbf{0.00}^{+0.01}_{-0.00}$&$24.43^{+7.56}_{-6.97}$\\
				object\_hue &$99.23^{+0.18}_{-0.35}$&$0.38^{+0.82}_{-0.37}$&$22.71^{+14.77}_{-6.45}$   
				&$99.22^{+0.26}_{-0.31}$&$0.43^{+0.61}_{-0.43}$&$24.55^{+7.04}_{-11.17}$\\
				scale 	&$74.17^{+6.07}_{-6.42}$&$0.41^{+1.64}_{-0.41}$&$\mathbf{0.00}^{+0.00}_{-0.00}$   
				&$68.27^{+2.36}_{-2.04}$&$0.27^{+0.59}_{-0.27}$&$\mathbf{0.00}^{+0.00}_{-0.00}$\\
				shape 	&$98.38^{+0.82}_{-1.19}$&$0.07^{+0.14}_{-0.07}$&$\mathbf{0.00}^{+0.04}_{-0.00}$   
				&$97.45^{+0.46}_{-0.57}$&$0.08^{+0.20}_{-0.08}$&$\mathbf{0.00}^{+0.06}_{-0.00}$\\
				orientation   &$81.63^{+3.23}_{-2.89}$&$\mathbf{0.00}^{+0.00}_{-0.00}$&$\mathbf{0.00}^{+0.00}_{-0.00}$
				&$74.27^{+1.07}_{-2.34}$&$\mathbf{0.00}^{+0.00}_{-0.00}$&$\mathbf{0.00}^{+0.00}_{-0.00}$\\
				\midrule
				average
				&$\mathbf{92.22}$ &$0.15$&$9.28$
				&$89.86$ &$\mathbf{0.13}$&$8.21$\\ 
				\bottomrule
		\end{tabular}}
	\end{center}
	\vspace{-2ex}
\end{table}

\begin{table}[t]
	\RawFloats
	\caption[Detailed version of CCAE and RCAE from \autoref{tab:ofm:quantitativ_results_3dshapes}.]{Detailed version of CCAE and RCAE from \autoref{tab:ofm:quantitativ_results_3dshapes}.} 
	\vspace{-1ex}
	\label{tab:ofm:app:quantitativ_results_3dshapes_details_2}
	\begin{center}
		\scalebox{0.8}{%
			\setlength{\tabcolsep}{0.4em}
			\begin{tabular}{lccccccc}
				\toprule
				&\multicolumn{3}{c}{CCAE} &\multicolumn{4}{c}{\acs{RCAE}}
				\\	
				\cmidrule(lr){2-4}
				\cmidrule(lr){5-8}
				&  $\mathrm{ACC}$ &   $\mathrm{cSM3}$& $\mathrm{MOFM}$
				&  $\mathrm{ACC}$ &  $\mathrm{cSM3}$& $\mathrm{MOFM}$ &$\mathrm{MM3}$\\
				\midrule
				floor\_hue   &$99.94^{+0.03}_{-0.05}$&$\mathbf{0.02}^{+0.03}_{-0.02}$&$\mathbf{0.00}^{+0.00}_{-0.00}$    
				&$92.08^{+1.79}_{-2.89}$&$56.68^{+0.94}_{-1.55}$&$\infty$&$44.67^{+4.10}_{-3.78}$  \\
				wall\_hue      &$\mathbf{99.98}^{+0.01}_{-0.02}$&$0.10^{+0.11}_{-0.07}$&$\mathbf{0.00}^{+0.78}_{-0.00}$
				&$\mathbf{99.96}^{+0.04}_{-0.04}$&$25.17^{+9.67}_{-8.15}$&$\infty$&$7.80^{+0.59}_{-0.85}$\\
				object\_hue      &$98.96^{+0.38}_{-0.30}$&$1.55^{+0.73}_{-0.63}$&$0.63^{+5.58}_{-0.59}$   
				&$98.79^{+0.35}_{-0.41}$&$59.65^{+8.12}_{-10.39}$&$\infty$&$40.11^{+0.99}_{-1.88}$\\
				scale      &$66.72^{+3.20}_{-2.20}$&$0.10^{+0.41}_{-0.10}$&$\mathbf{0.00}^{+0.00}_{-0.00}$    
				&$67.10^{+6.77}_{-3.96}$&$2.60^{+4.10}_{-2.60}$&$0.13^{+3.54}_{-0.13}$&$31.78^{+1.91}_{-2.44}$\\
				shape 	    &$98.75^{+0.39}_{-0.50}$&$0.03^{+0.10}_{-0.03}$&$\mathbf{0.00}^{+0.00}_{-0.00}$    
				&$98.17^{+0.43}_{-0.39}$&$\mathbf{0.20}^{+0.45}_{-0.20}$&$0.06^{+0.82}_{-0.06}$&$\mathbf{-2.48}^{+0.65}_{-1.28}$\\
				orientation     &$72.55^{+3.17}_{-2.81}$&$0.23^{+0.91}_{-0.23}$&$\mathbf{0.00}^{+0.00}_{-0.00}$
				&$80.69^{+2.55}_{-2.68}$&$0.48^{+1.91}_{-0.48}$&$\mathbf{0.00}^{+1.23}_{-0.00}$&$22.26^{+0.74}_{-1.19}$\\
				\midrule
				average
				&$89.48$ &$0.34$&$\mathbf{0.11}$
				&$89.47$ &$24.13$&$\infty$&$24.02$\\ 
				\bottomrule
		\end{tabular}}
	\end{center}
	\vspace{-2ex}
\end{table}

\begin{table}[t]
	\RawFloats
	\caption[Detailed version of \acs{SCLCAE} from \autoref{tab:ofm:quantitativ_results_3dshapes}.]{Detailed version of \acs{SCLCAE} from \autoref{tab:ofm:quantitativ_results_3dshapes}.} 
	\vspace{-1ex}
	\label{tab:ofm:app:quantitativ_results_3dshapes_details_3}
	\begin{center}
		\scalebox{0.8}{%
			\setlength{\tabcolsep}{0.4em}
			\begin{tabular}{lcccc}
				\toprule
				&\multicolumn{4}{c}{\acs{SCLCAE}}
				\\	
				\cmidrule(lr){2-5}
				&  $\mathrm{ACC}$ &  $\mathrm{cSM3}$& $\mathrm{MOFM}$ &$\mathrm{MM3}$\\
				\midrule
				floor\_hue &$93.53^{+2.62}_{-1.90}$&$28.18^{+11.12}_{-6.30}$&$268.27^{+161.92}_{-44.43}$&$-48.38^{+2.75}_{-2.15}$\\   
				wall\_hue &$\mathbf{99.96}^{+0.03}_{-0.02}$&$\mathbf{0.29}^{+0.07}_{-0.11}$&$0.46^{+5.04}_{-0.44}$&$\mathbf{-76.40}^{+2.97}_{-3.06}$\\     
				object\_hue &$98.67^{+0.82}_{-1.06}$&$2.87^{+0.69}_{-1.36}$&$8.69^{+3.20}_{-0.64}$&$-73.17^{+3.16}_{-2.77}$\\
				scale  &$83.94^{+1.15}_{-0.57}$&$2.43^{+4.02}_{-2.43}$&$\mathbf{0.00}^{+1.78}_{-0.00}$&$-44.80^{+2.06}_{-1.55}$\\     
				shape &$95.06^{+0.92}_{-0.86}$&$1.67^{+1.54}_{-1.67}$&$2.16^{+1.48}_{-0.13}$&$-67.54^{+2.38}_{-1.63}$\\	    
				orientation &$45.32^{+4.14}_{-3.57}$&$2.50^{+2.55}_{-1.97}$&$6.68^{+5.21}_{-3.02}$&$-9.11^{+2.46}_{-1.71}$\\
				\midrule
				average
				&$86.08$&$6.32$&$47.71$&$\mathbf{-53.23}$\\ 
				\bottomrule
		\end{tabular}}
	\end{center}
	\vspace{-2ex}
\end{table}

\begin{table}[t]
	\RawFloats
	\caption[Mismatches of other models we have tested.]{Mismatches of other models we have tested. Values are obtained by cross-validation, please refer to \autoref{tab:ofm:app:train_eval_details} for more details.} 
	\vspace{-1ex}
	\label{tab:ofm:app:additional_quantitativ_results_1}
	\begin{center}
		\scalebox{0.8}{%
			\setlength{\tabcolsep}{0.4em}
			\begin{tabular}{lcccc}
				\toprule
				&\multicolumn{4}{c}{Cifar10}		\\	
				\cmidrule(lr){2-5}
				& $\mathrm{ACC}$  & $\mathrm{cSM3}$ & $\mathrm{MOFM}$&$\mathrm{MM3}$\\
				\midrule
				\acs{CAE}&$44.69^{+0.69}_{-0.41}$&$1.42^{+0.68}_{-0.70}$&$18.71^{+6.61}_{-4.01}$&-\\
				CAENoCrossVal 	&$44.24^{+0.24}_{-0.41}$&$1.26^{+0.15}_{-0.47}$&$17.40^{+2.91}_{-2.58}$ & - \\
				CAE100E &$45.37^{+1.09}_{-1.06}$&$1.22^{+0.56}_{-0.32}$&$5.84^{+7.59}_{-1.12}$   &-\\
				RResNet18 &$54.64^{+1.80}_{-2.01}$&$3.98^{+1.74}_{-3.60}$&$4.87^{+4.42}_{-3.11}$&$31.82^{+0.75}_{-0.68}$ \\
				SCLResNet18 &$87.14^{+0.37}_{-0.40}$&$0.29^{+0.23}_{-0.20}$&$0.12^{+0.07}_{-0.02}$&$-31.83^{+0.12}_{-0.23}$\\ 
				RResNet18R32 &$39.14^{+1.53}_{-1.69}$&$1.92^{+1.75}_{-1.29}$&$4.39^{+9.53}_{-1.58}$&$48.93^{+1.17}_{-2.19}$\\
				RResNet18R256 &$47.46^{+1.44}_{-1.76}$&$3.91^{+2.94}_{-1.71}$&$6.86^{+8.51}_{-2.09}$&$39.08^{+1.78}_{-2.07}$\\
				RResNet18R512 &$50.55^{+2.91}_{-2.84}$&$7.01^{+1.83}_{-3.73}$&$9.83^{+10.55}_{-3.90}$&$36.93^{+3.15}_{-1.96}$\\
				RResNet18R756 &$51.80^{+2.04}_{-2.38}$&$7.56^{+5.01}_{-3.17}$&$8.50^{+11.90}_{-5.00}$&$35.61^{+1.45}_{-1.74}$\\
				RResNet18R1024 &$52.99^{+1.54}_{-2.58}$&$9.89^{+4.51}_{-2.41}$&$15.12^{+11.12}_{-4.18}$&$36.86^{+2.76}_{-2.20}$\\
				\bottomrule
		\end{tabular}}
	\end{center}
	\vspace{-2ex}
\end{table}

\begin{table}[t]
	\RawFloats
	\caption[Mismatches of other models we have tested.]{Mismatches of other models we have tested. Values are obtained by cross-validation, please refer to \autoref{tab:ofm:app:train_eval_details} for more details.} 
	\vspace{-1ex}
	\label{tab:ofm:app:additional_quantitativ_results_2}
	\begin{center}
		\scalebox{0.8}{%
			\setlength{\tabcolsep}{0.4em}
			\begin{tabular}{lcccccccc}
				\toprule
				&\multicolumn{4}{c}{PCam}  &\multicolumn{4}{c}{Cifar100}  	\\	
				\cmidrule(lr){2-5}
				\cmidrule(lr){6-9}
				&   $\mathrm{ACC}$ & $\mathrm{cSM3}$&  $\mathrm{MOFM}$ & $\mathrm{MM3}$ &   $\mathrm{ACC}$ & $\mathrm{cSM3}$ & $\mathrm{MOFM}$&$\mathrm{MM3}$	\\
				\midrule
				SCLResNet18 &$96.25^{+0.44}_{-0.23}$&$0.37^{+0.44}_{-0.37}$&$0.86^{+1.00}_{-0.60}$&$-53.26^{+0.52}_{-0.38}$ 
				&$59.20^{+0.19}_{-0.25}$&$0.41^{+0.10}_{-0.09}$&$0.06^{+0.05}_{-0.04}$&$0.84^{+0.06}_{-0.03}$\\ 
				\bottomrule
		\end{tabular}}
	\end{center}
	\vspace{-2ex}
\end{table}
\begin{figure}[t]
	\begin{center}
		\begin{tabular}{cc}
			\includegraphics[width=0.45\columnwidth]{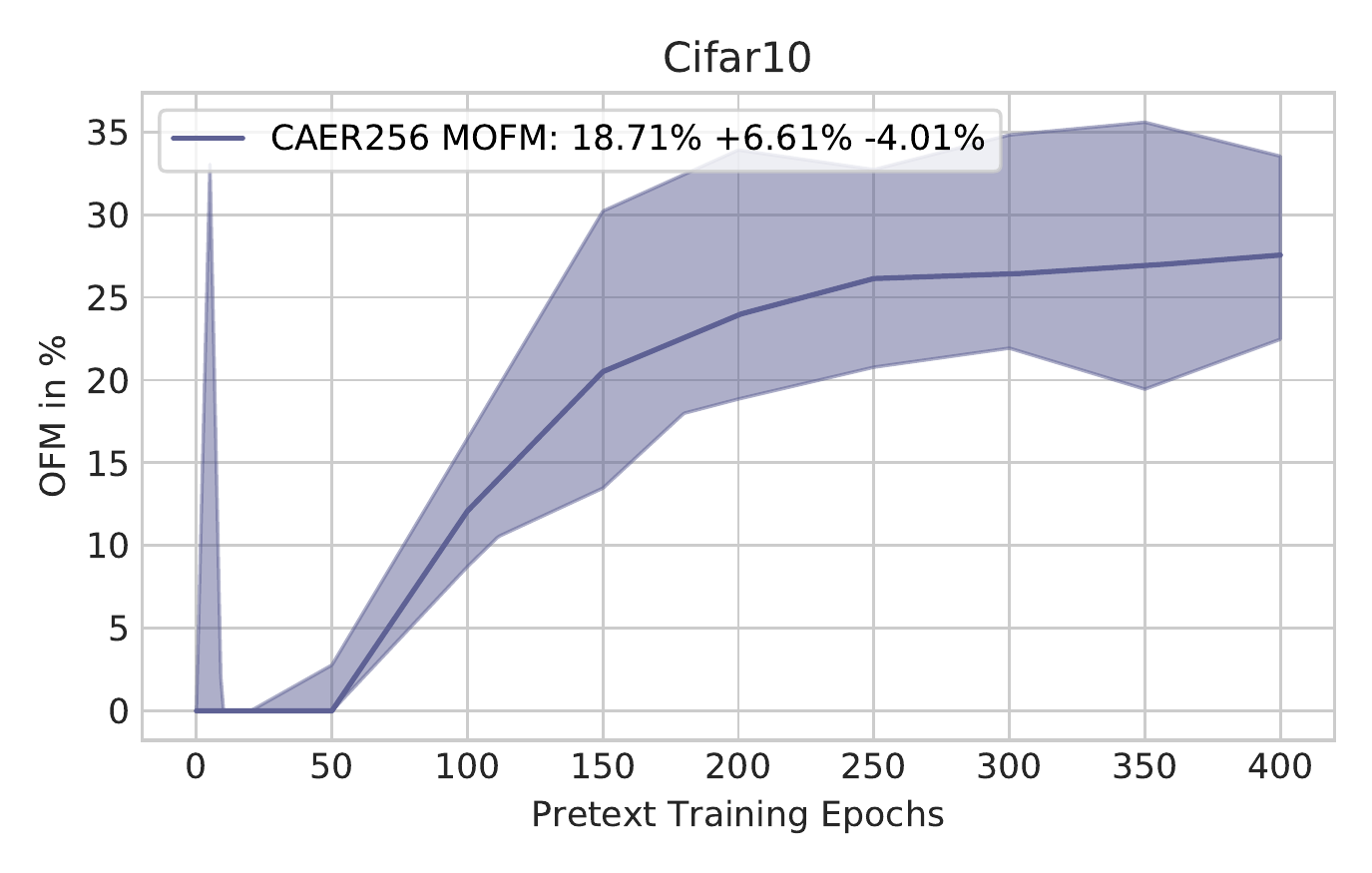} & 
			\includegraphics[width=0.45\columnwidth]{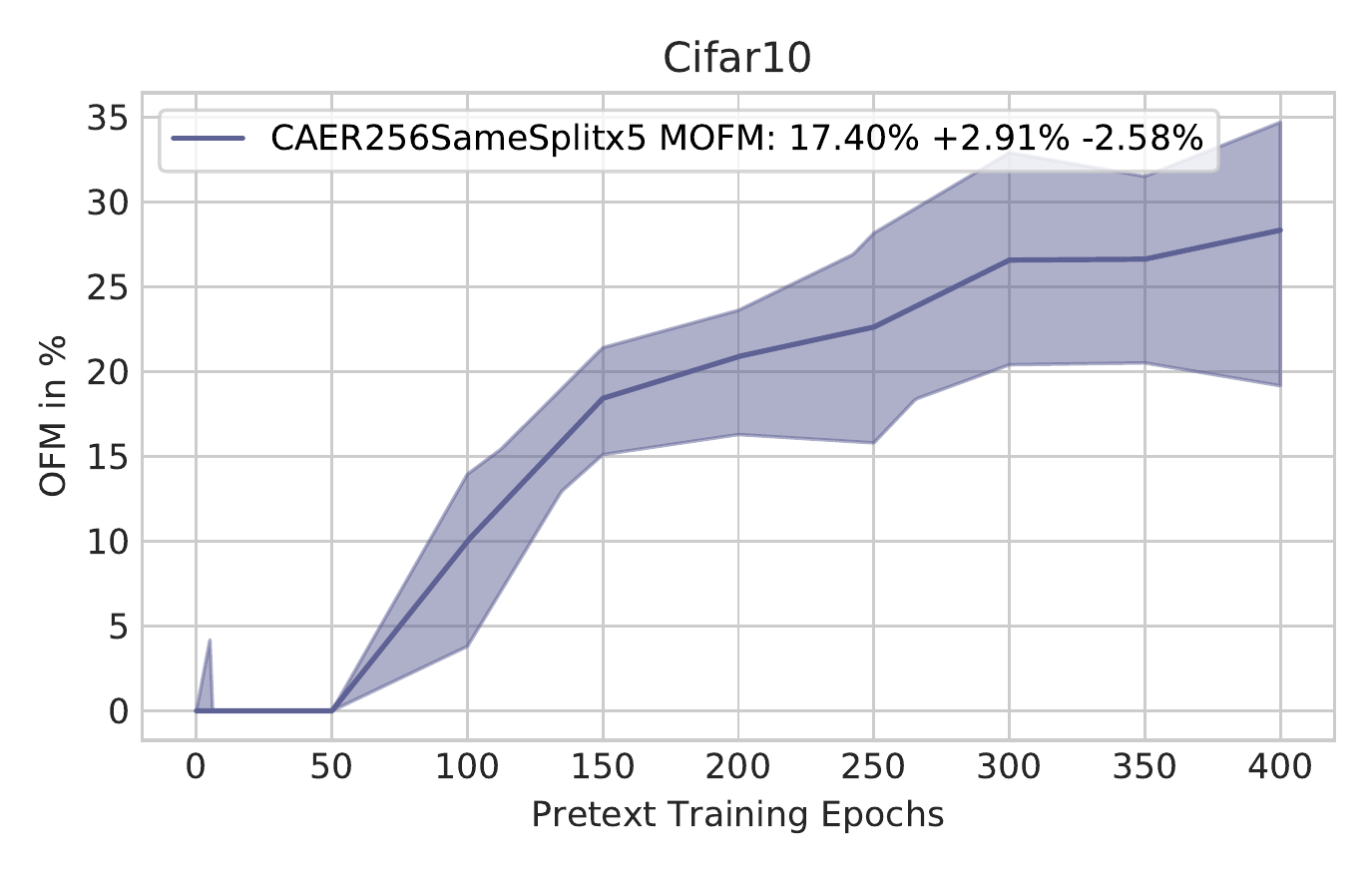} \\
		\end{tabular}
	\end{center}
	\vspace{-3ex}
	\caption[Stability of the partially measured $\mathrm{OFM}$.]{Stability of the partially measured $\mathrm{OFM}$. (Left) $\mathrm{OFM}$ of a \acs{CAE} trained for 400 epochs. Stability is measured by 5-fold cross-validation. (Right) $\mathrm{OFM}$ of a \acs{CAE} trained for 400 epochs. Stability is measured by training the \acs{CAE} five times on the same dataset split. Unsurprisingly, the stability of the $\mathrm{OFM}$ is higher when the \acs{CAE} is trained on the same split instead of the different splits from the 5-fold cross-validation.}
	\label{fig:ofm:app:cae_xfold_partial_stability}
\end{figure}

\begin{figure}[t]
	\begin{center}
		\begin{tabular}{cc}
			\includegraphics[width=0.45\columnwidth]{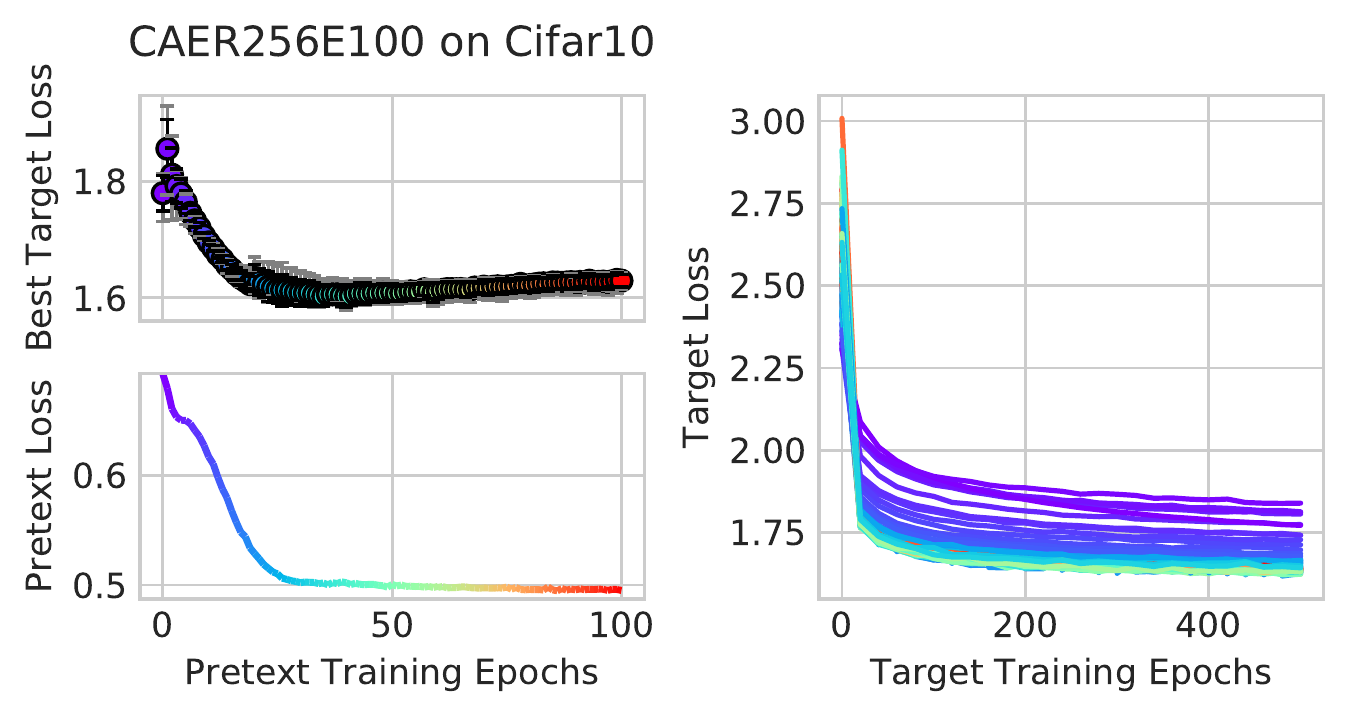} & 	\includegraphics[width=0.45\columnwidth]{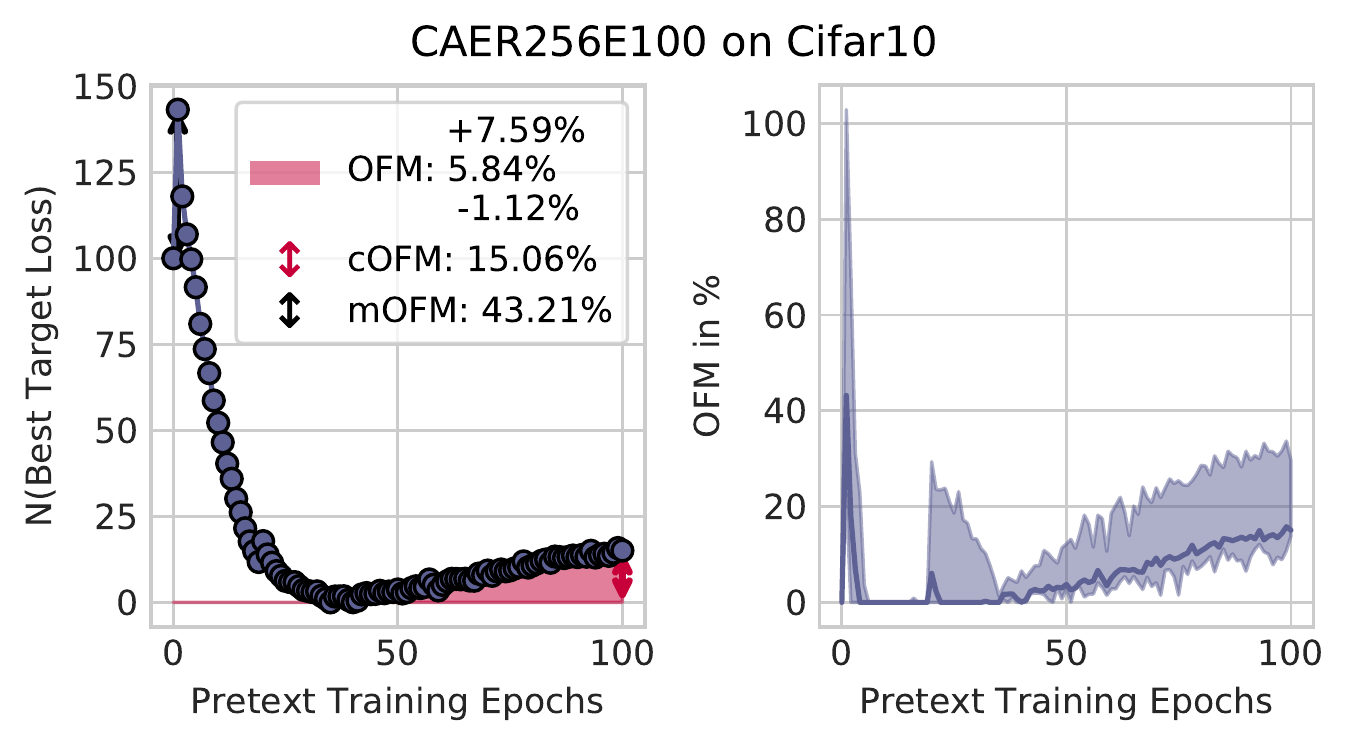} \\
		\end{tabular}
	\end{center}
	\vspace{-3ex}
	\caption[Stability of the fully measured $\mathrm{OFM}$.]{Stability of the fully measured $\mathrm{OFM}$. (Left) Losses of a simple \acs{CAE} measured for every pretext training epoch. The curve formed by the target models represents a typical target training curve in our setup. (Right) The $\mathrm{OFM}$ and its stability measured for every pretext training epoch of the \acs{CAE}. When we compare the stability to the partially measured \acs{CAE} in \autoref{fig:ofm:app:cae_xfold_partial_stability}, we observe a similar instability. Best viewed in color.}
	\label{fig:ofm:app:cae_xfold_full_stability}
\end{figure}

\begin{figure}[t]
	\begin{center}
		\begin{tabular}{cc}
			\includegraphics[width=0.45\columnwidth]{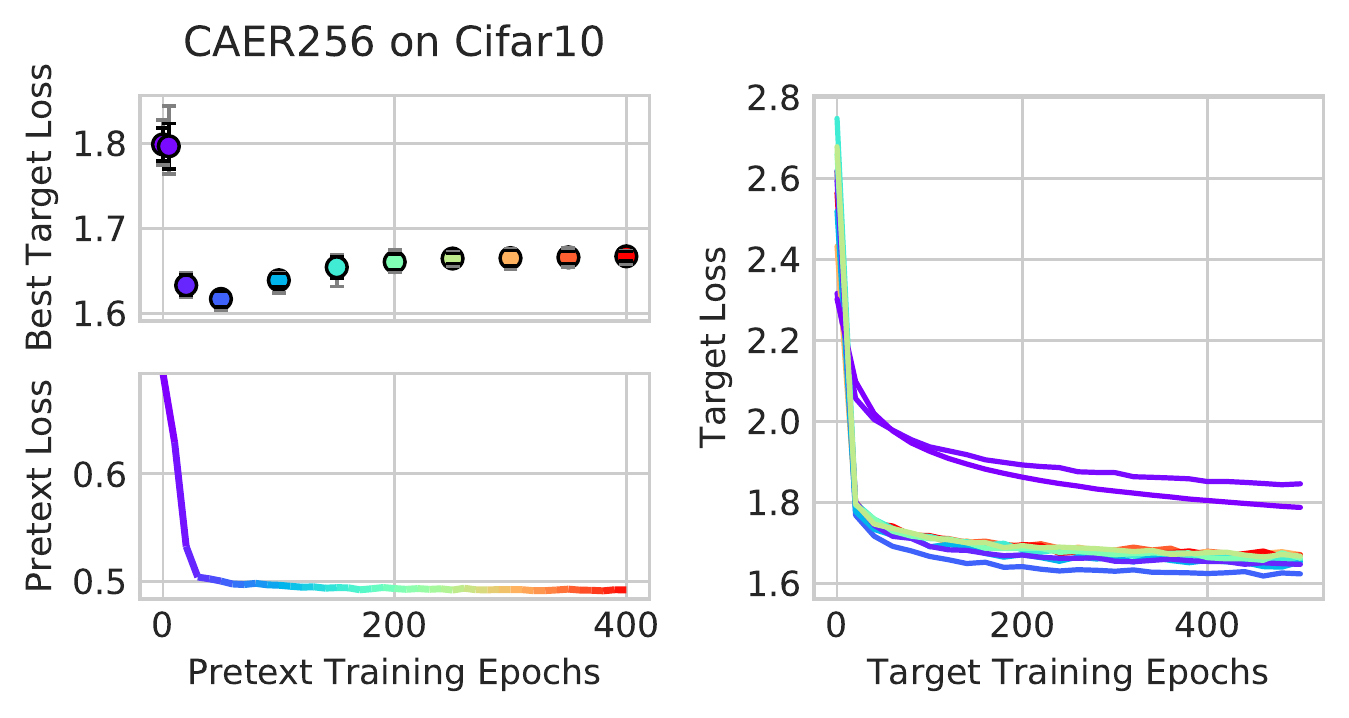}&  
			\includegraphics[width=0.45\columnwidth]{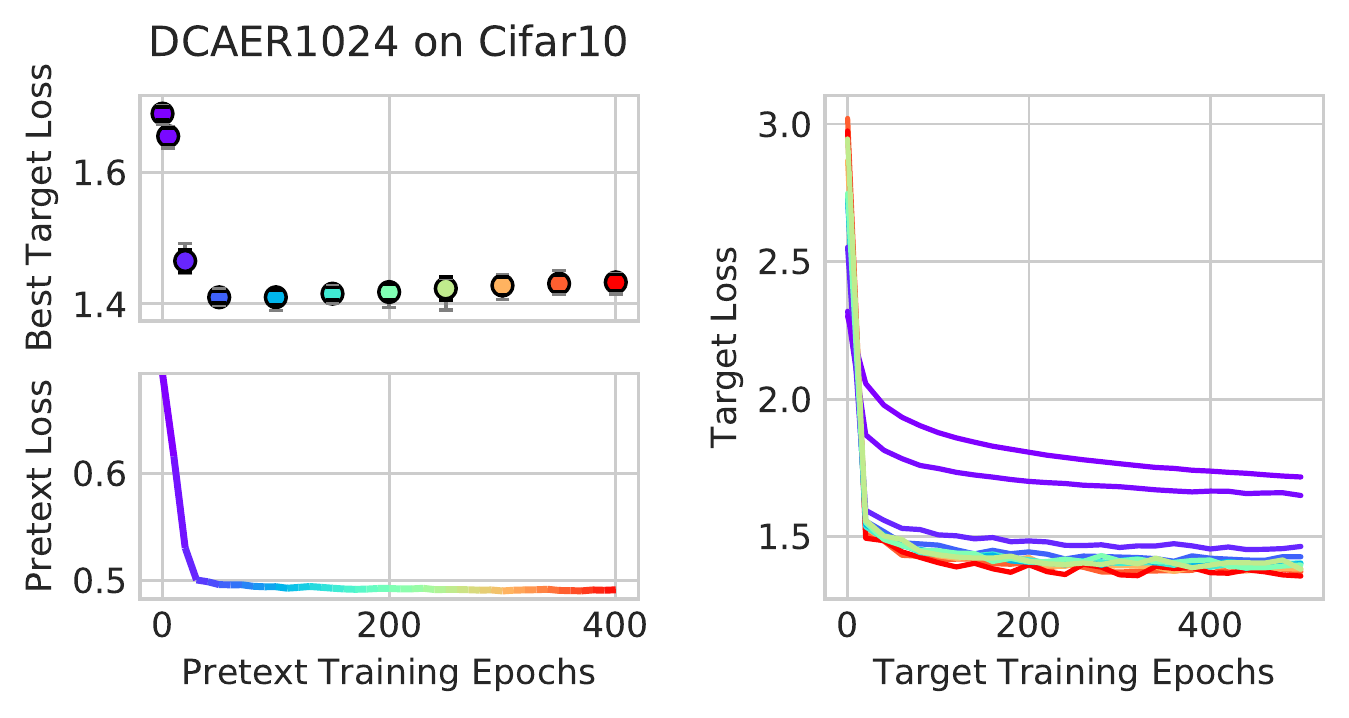}\\ 
			\includegraphics[width=0.45\columnwidth]{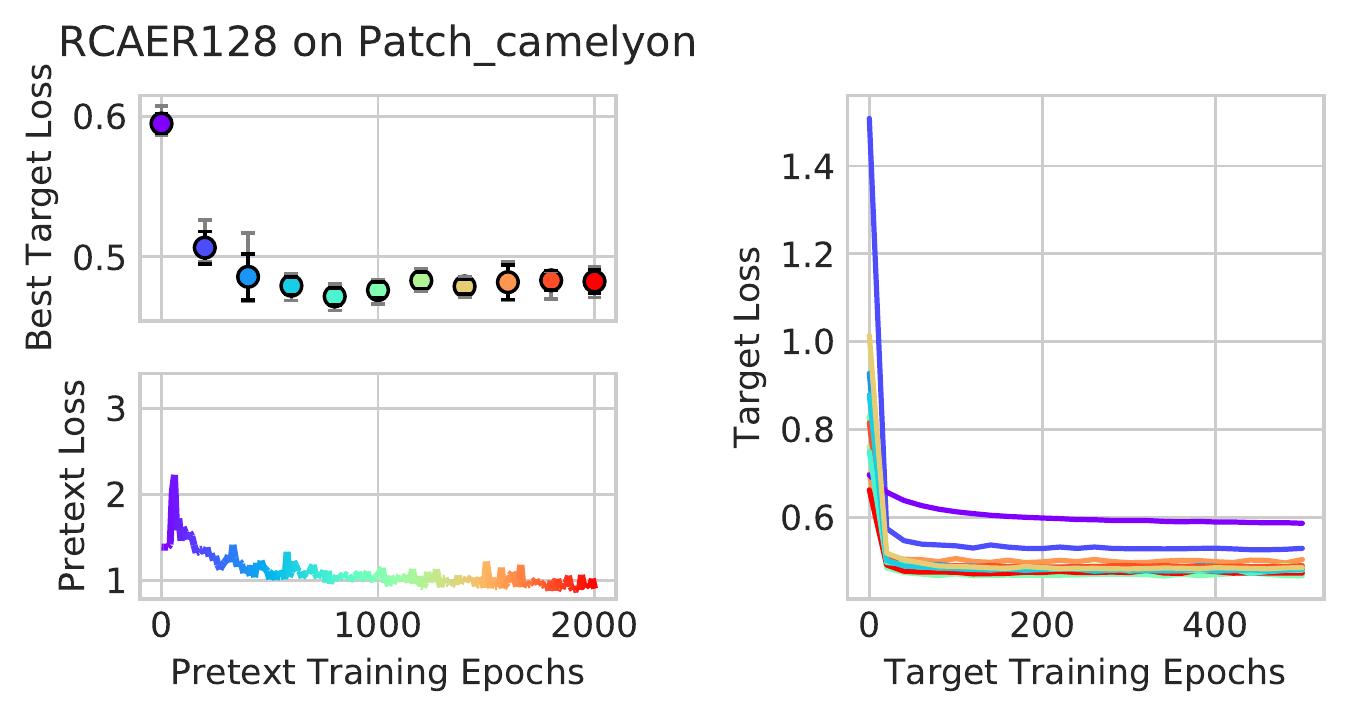}&	\includegraphics[width=0.45\columnwidth]{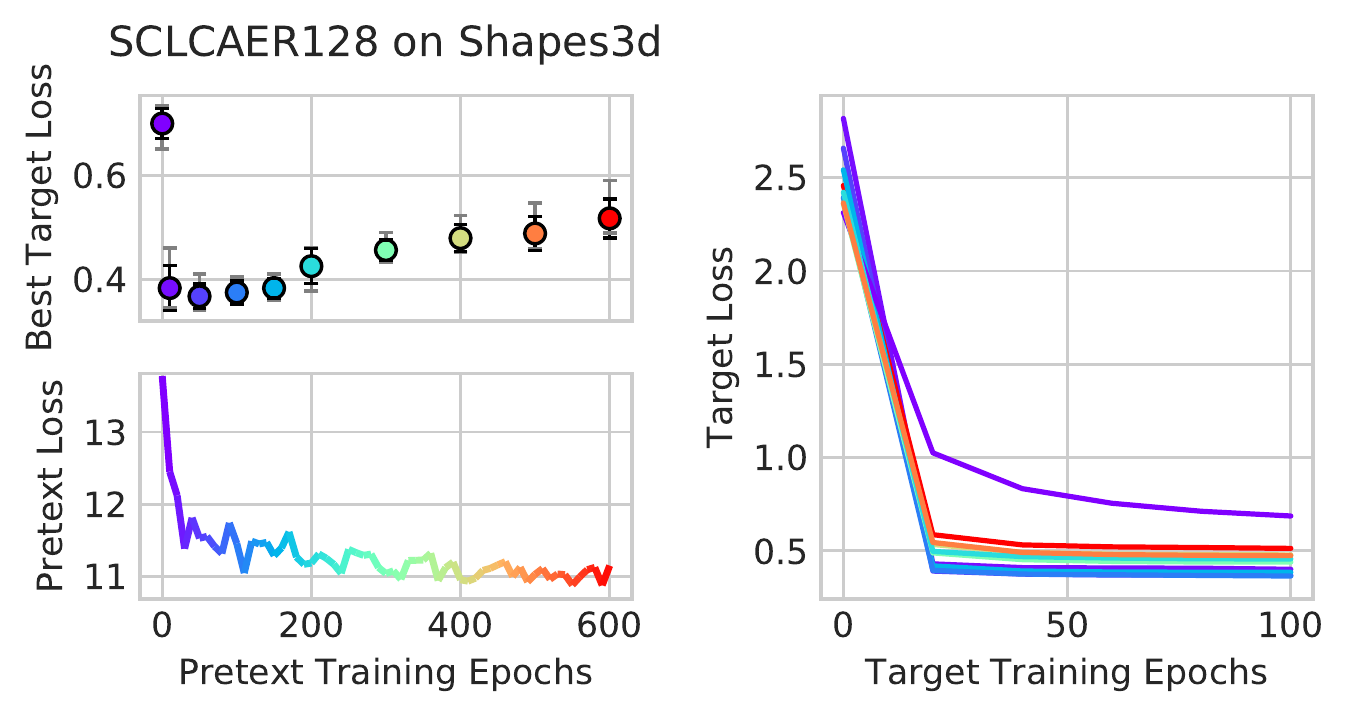}\\
		\end{tabular}
	\end{center}
	\vspace{-3ex}
	\caption[Additional evidence for the mismatch and convergence section \ref{sec:mandc}.]{Additional evidence for the mismatch and convergence section \ref{sec:mandc}. \textit{Longer training of the pretext task tends to create easier separable representations, which may mismatch with the class label}. We observe that the target loss curves converge faster for target models trained on the pretext model's representation from later pretext training epochs. Especially the purple curves show this behavior clearly. Best viewed in color.}
	\label{fig:ofm:app:additional_evidence_mis_conv_sec}	
\end{figure}

\begin{figure}[t]
	\begin{center}
		\includegraphics[width=1.0\columnwidth]{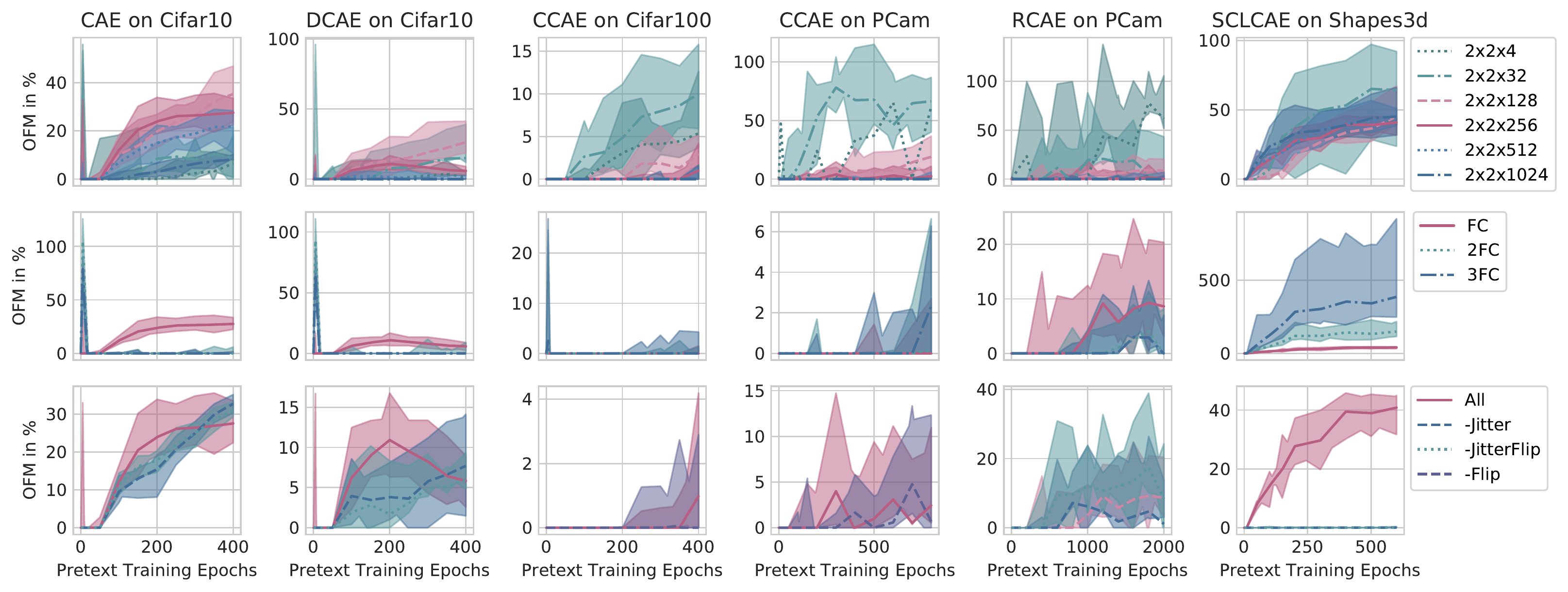} 
	\end{center}
	\vspace{-2ex}
	\caption[Version of \autoref{fig:ofm:ofm_ablation_study} without convergence criterium.]{Version of \autoref{fig:ofm:ofm_ablation_study} without convergence criterium. We observe similar behaviors of the mismatches as in \autoref{fig:ofm:ofm_ablation_study} in most cases. One exception is the color jitter, where the $\mathrm{OFM}$ starts to converge or decrease late in training. (top) Impact of different pretext model representation sizes on the $\mathrm{OFM}$ for our model. (middle) The $\mathrm{OFM}$ for the linear and nonlinear target models trained on our pretext model. (bottom) The $\mathrm{OFM}$ for the linear target model and for the pretext models trained on fewer augmentations. First we have removed the color jitter and then the vertical flip from the augmentations. For the CCAE, we only have removed the vertical flip. The target models of SCLCAE have been trained on 3dshapes to predict the object hue. Best viewed in color.}
	\label{fig:ofm:app:target_and_augmentations_and_rep_size_ext_nocc}
\end{figure}

\begin{figure}[t]
	\begin{center}
		\includegraphics[width=1.0\columnwidth]{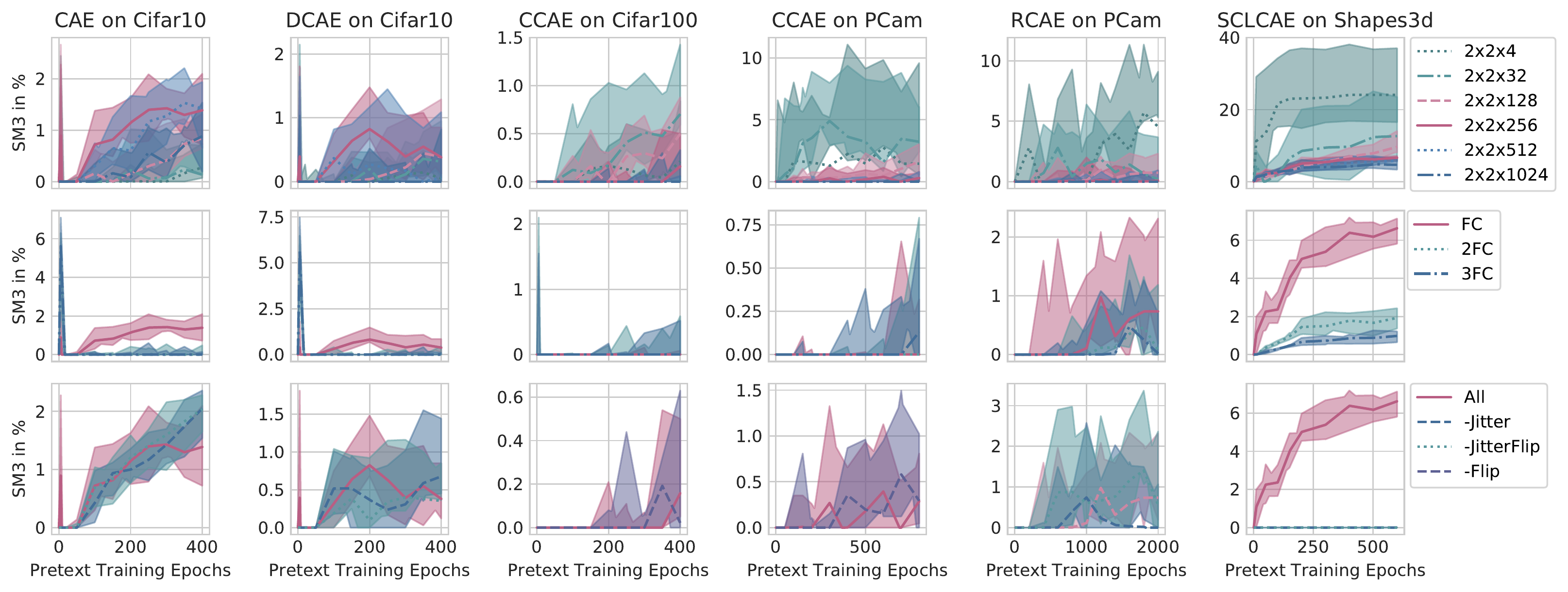} 
	\end{center}
	\vspace{-2ex}
	\caption[Version of \autoref{fig:ofm:ofm_ablation_study} for $\mathrm{SM3}$ on accuracies, without convergence criterium.]{Version of \autoref{fig:ofm:ofm_ablation_study} for $\mathrm{SM3}$ on accuracies, without convergence criterium. Again, we observe similar behaviors of the mismatches as in \autoref{fig:ofm:ofm_ablation_study} in most cases. Best viewed in color.}
	\label{fig:ofm:app:target_and_augmentations_and_rep_size_ext_sm3_nocc}
\end{figure}

\begin{figure}[t]
	\begin{center}
		\begin{tabular}{cc}
			\includegraphics[width=0.45\columnwidth]{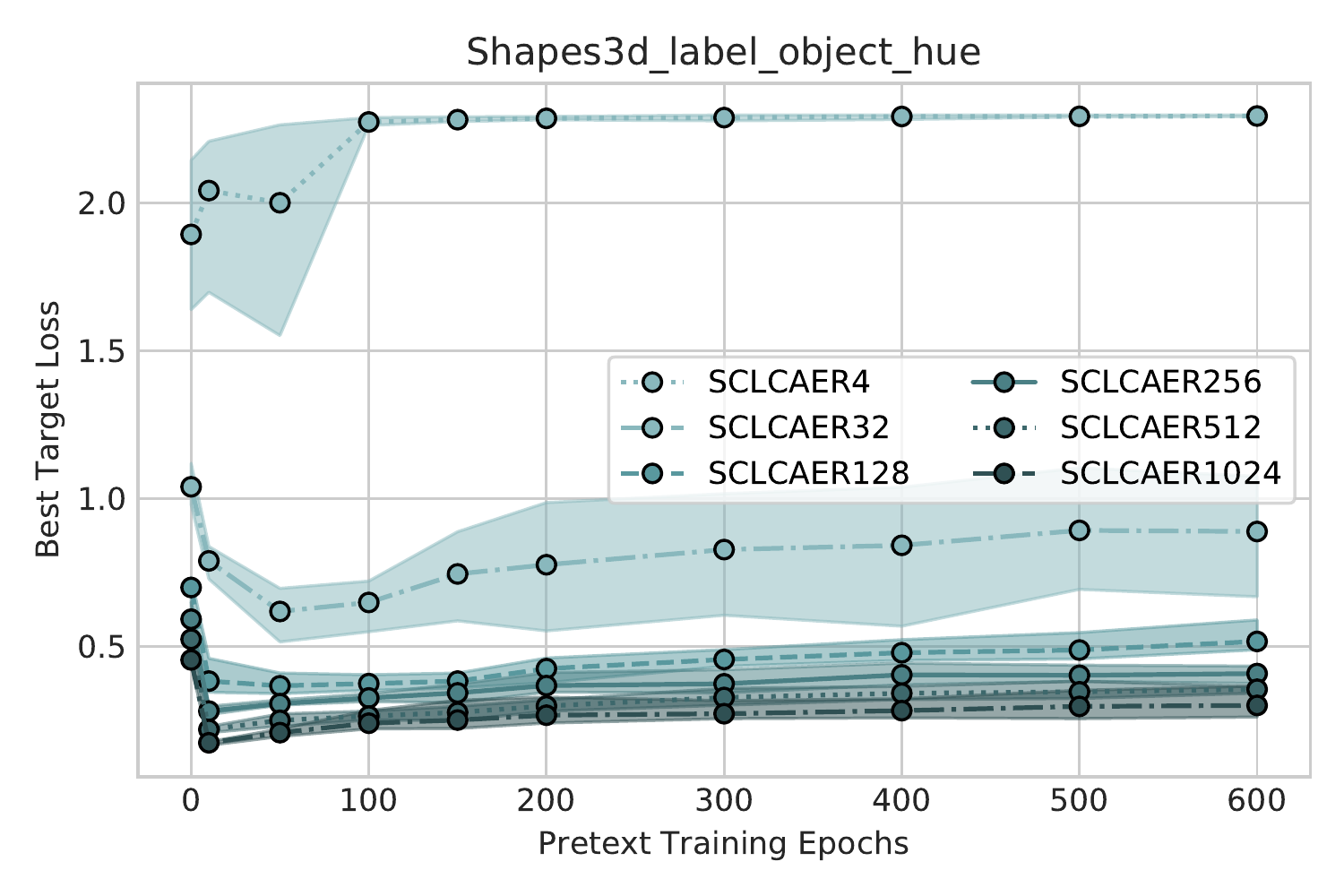}&
			\includegraphics[width=0.45\columnwidth]{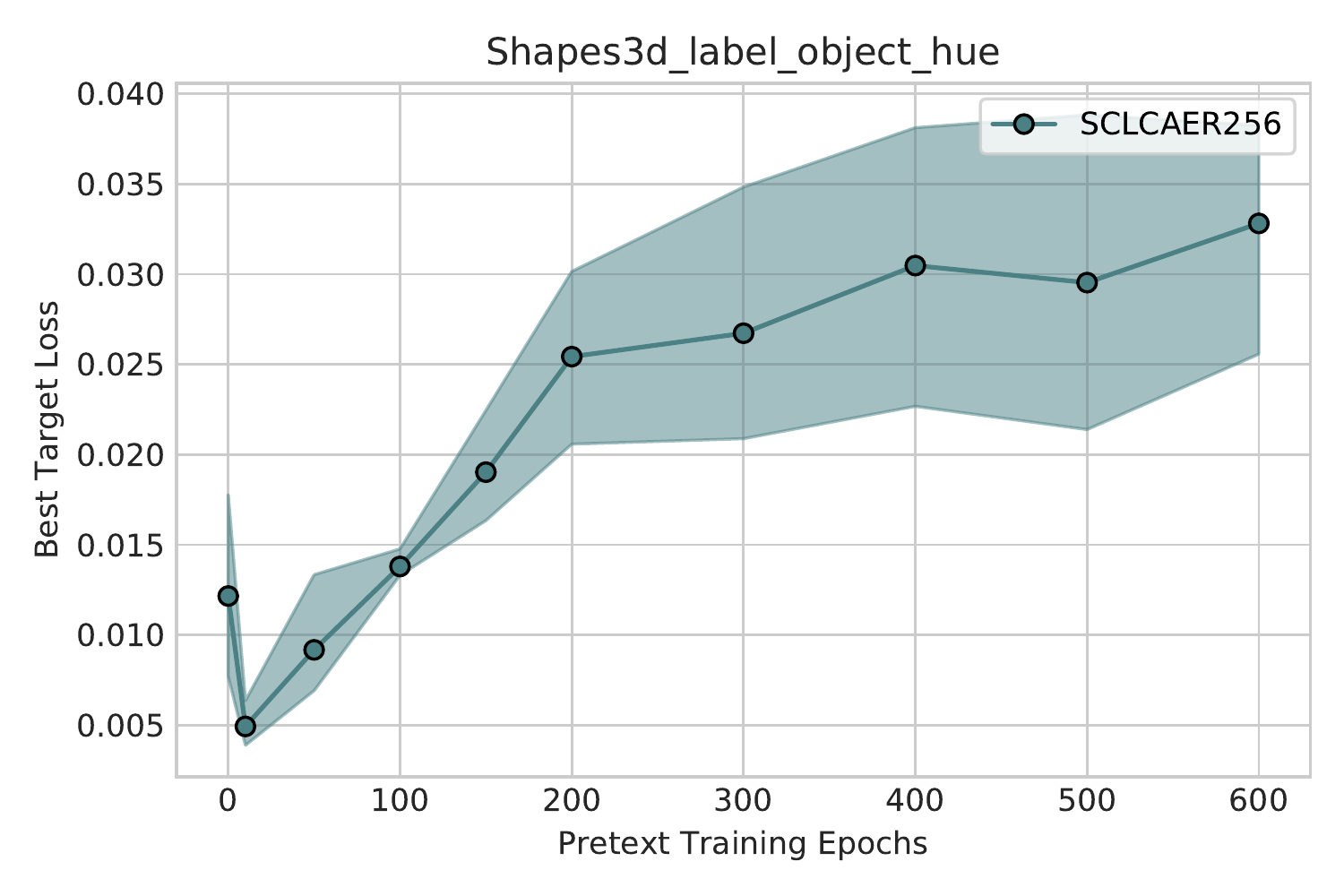}\\
		\end{tabular} 
	\end{center} 
	\vspace{-3ex}
	\caption[SCLCAE target losses for the object hue class of 3dshapes.]{SCLCAE target losses for the object hue class of 3dshapes. (left) We describe why the $OFM$ for larger representation sizes does not decrease in the setup where we train the contrastive pretext model (SCLCAE) on 3dshapes and the target model to predict the object hue. Here, the target models trained on the untrained pretext models with larger representation sizes already achieve high performance due to a higher amount of color-selective, random features. Additionally, learning the pretext model does not lead to a high performance gain, which leads again to a small interval for normalization. Therefore, forgetting useful features for the target task later in training leads to a high mismatch. (right) We describe why the $OFM$ does not decrease for more complex target models in the same setup. The nonlinear target model can make better sense of specific random pretext features for classification, which leads to a very low target loss at pretext model initialization. Since the pretext model does not learn many useful features for the target task later, this leads again to a small interval for normalization. Therefore, the $OFM$ gets very large later in training when the pretext model starts to forget useful features for the target task.}
	\label{fig:ofm:app:SCLCAER_target_losses}
\end{figure}

\begin{figure}[t]
	\begin{center}
		\begin{tabular}{cc}
			\includegraphics[width=0.45\columnwidth]{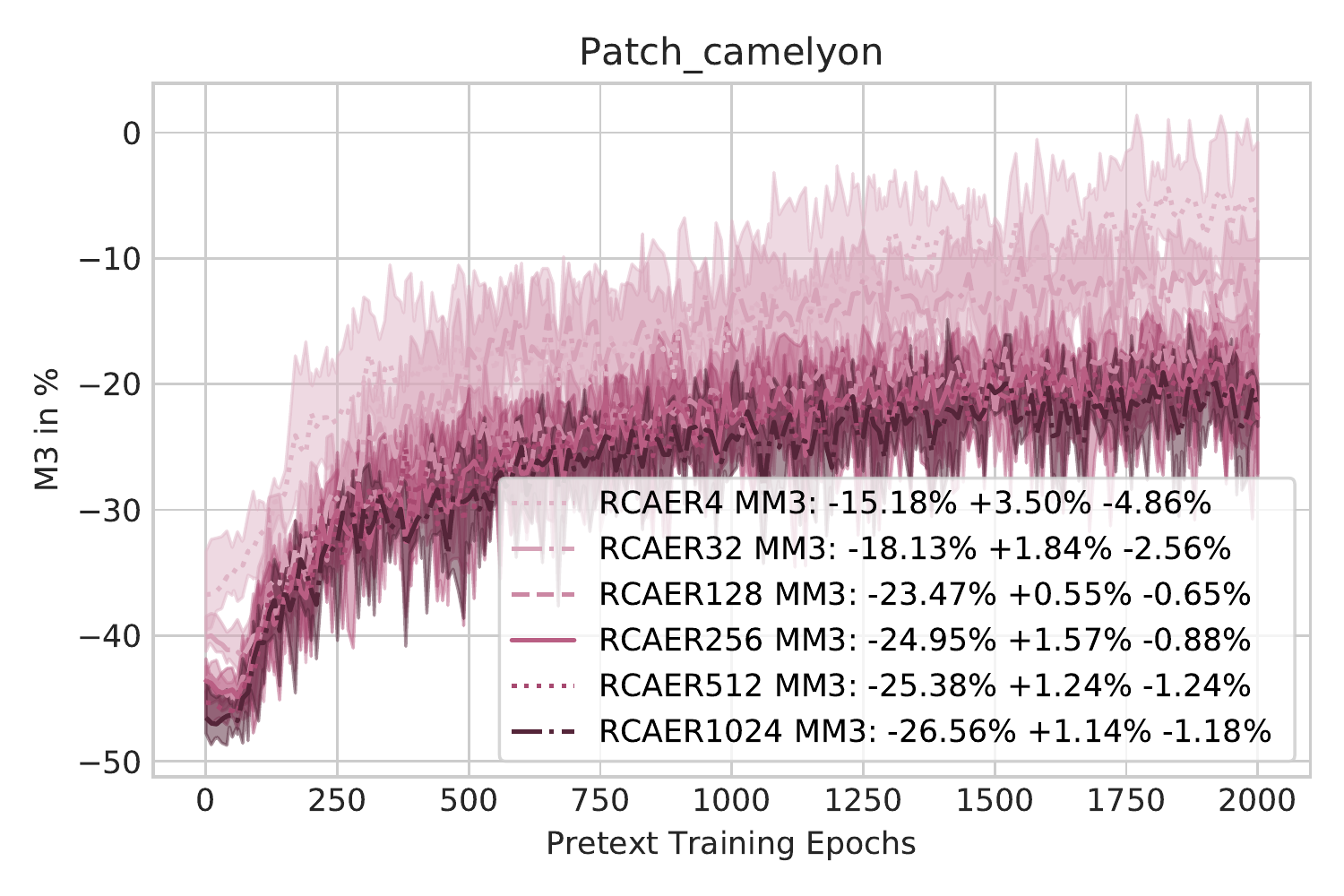} &    
			\includegraphics[width=0.45\columnwidth]{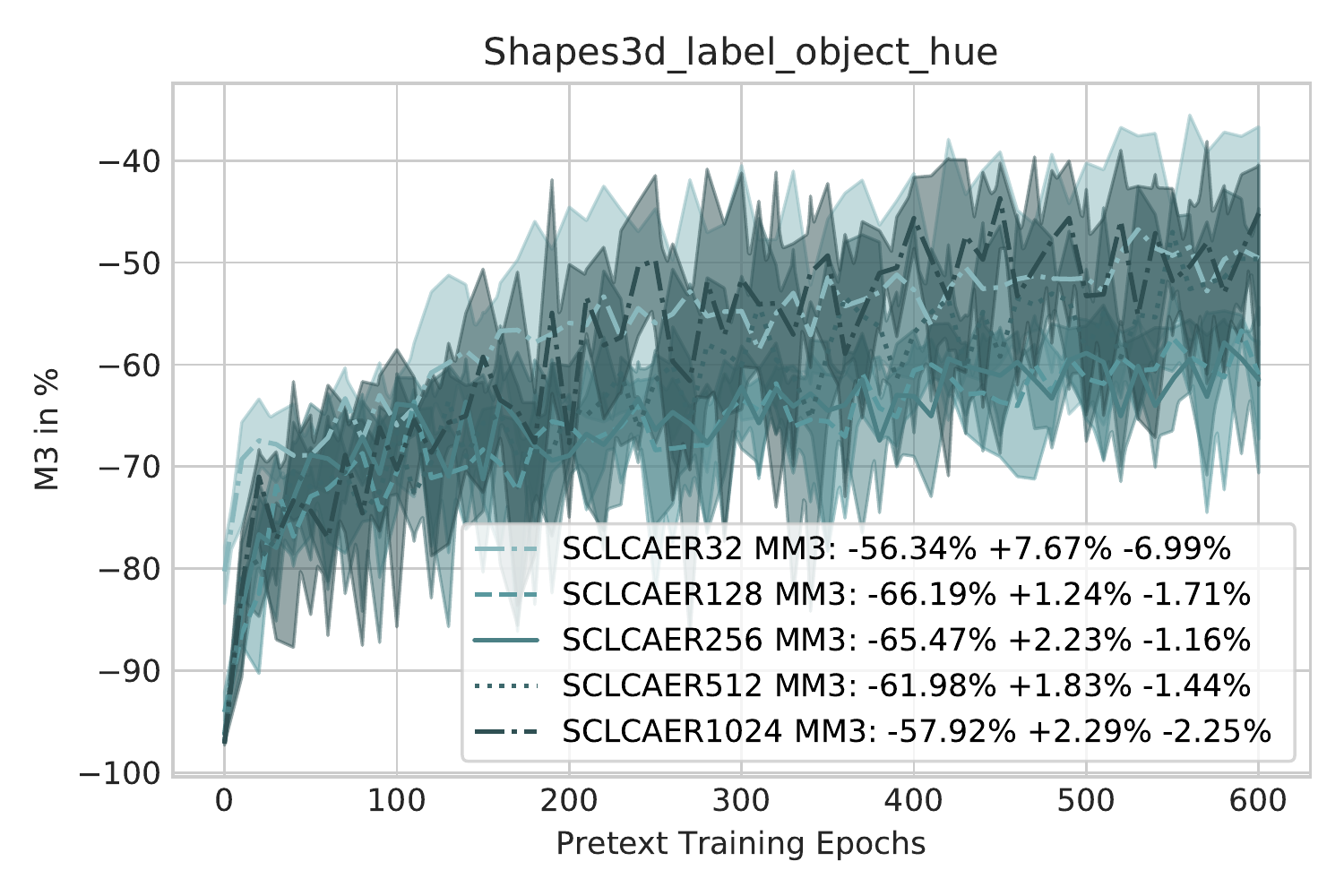} \\
			\includegraphics[width=0.45\columnwidth]{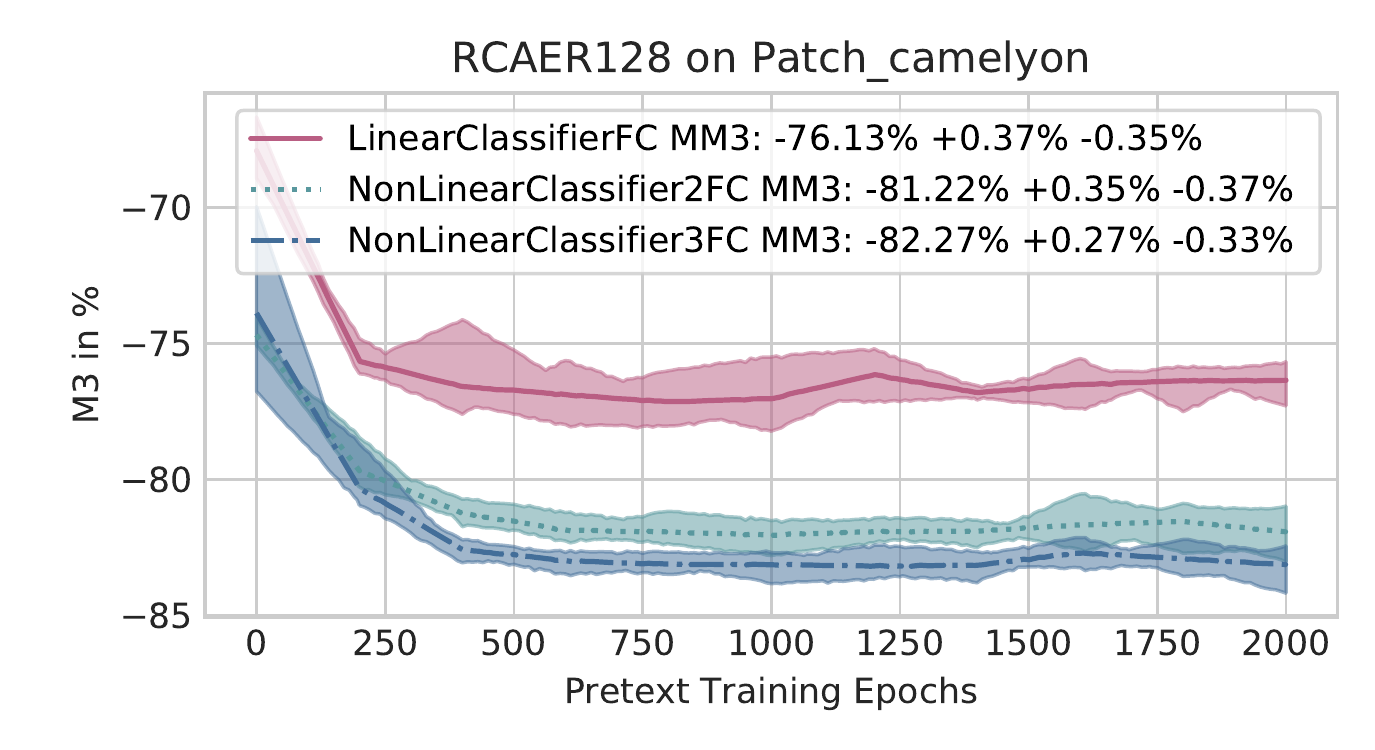} &    
			\includegraphics[width=0.45\columnwidth]{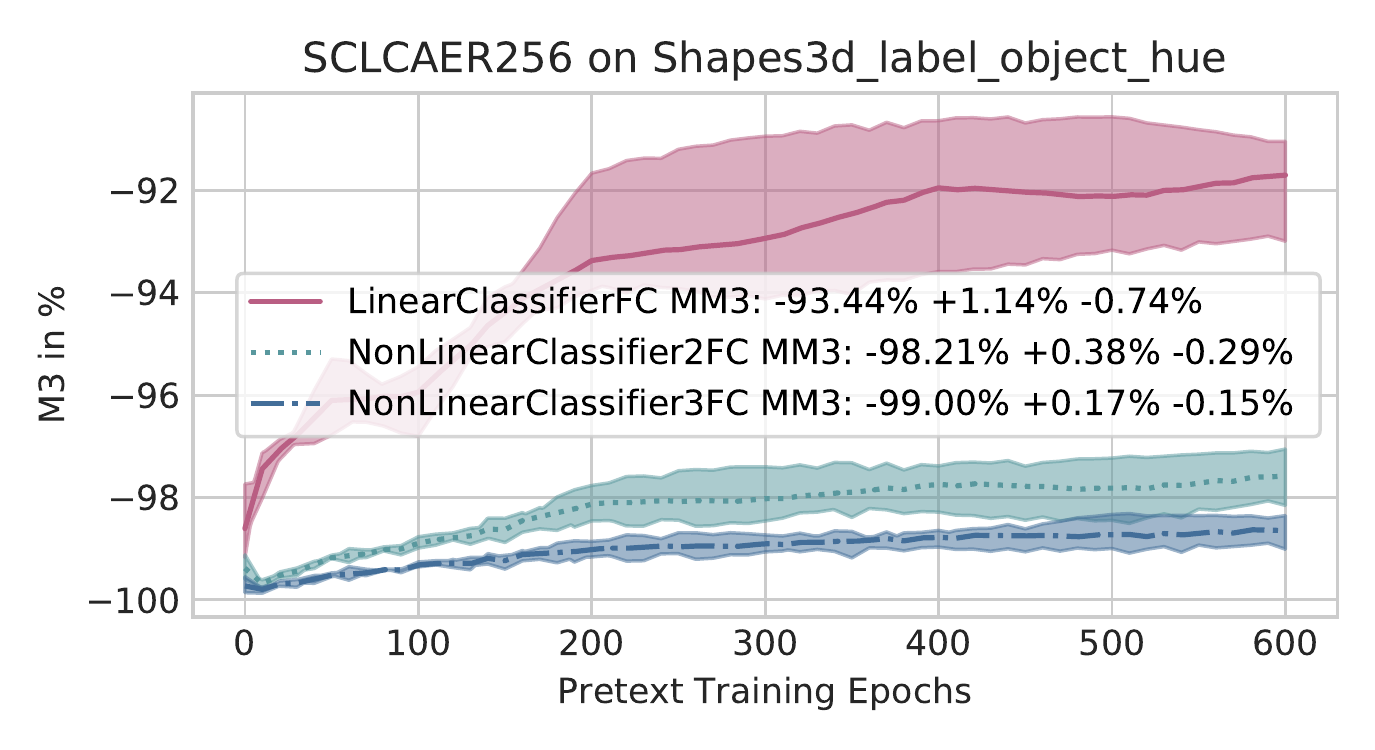} \\
			\includegraphics[width=0.45\columnwidth]{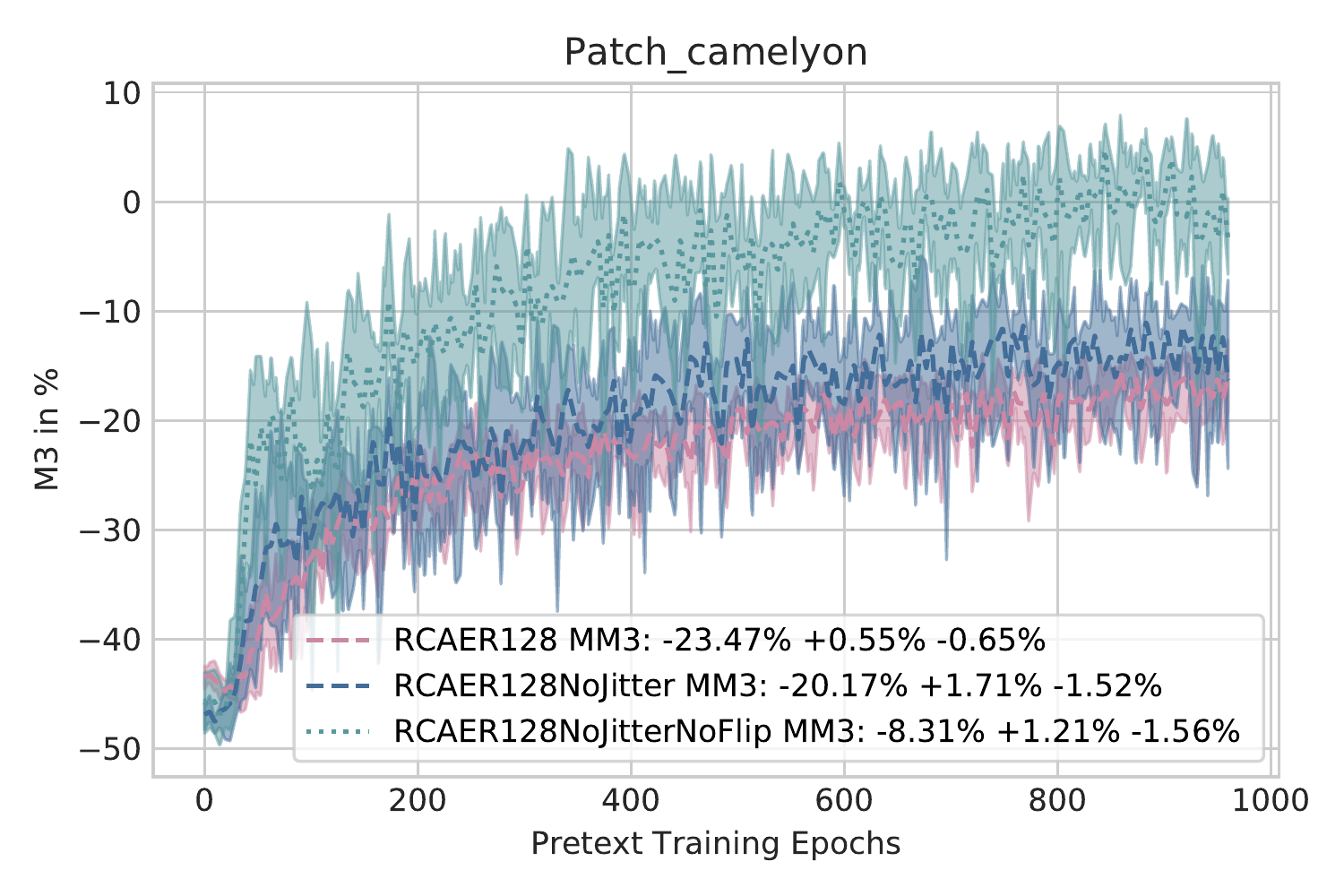} &    
			\includegraphics[width=0.45\columnwidth]{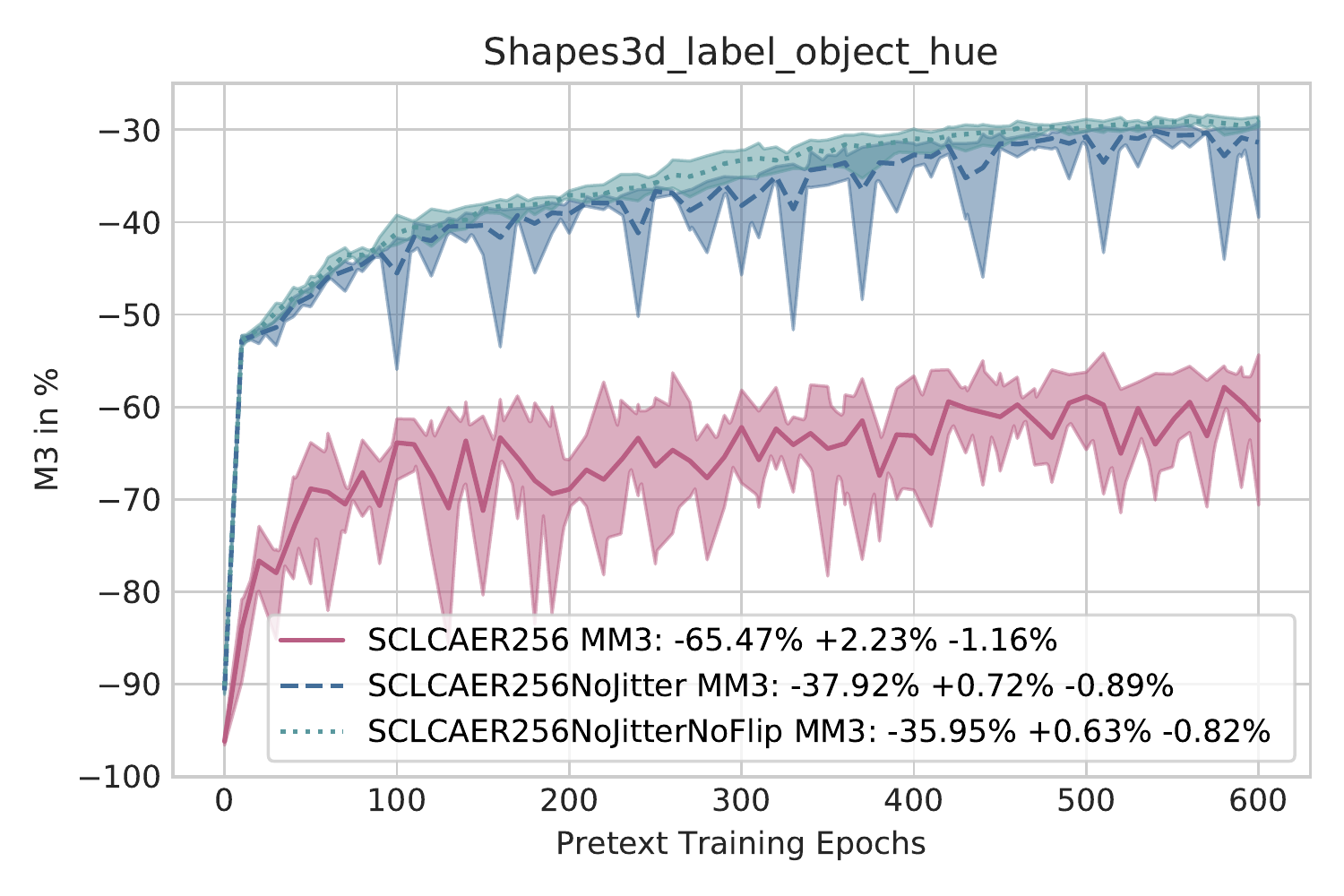} \\
		\end{tabular}
	\end{center}
	\vspace{-3ex}
	\caption[The version of \autoref{fig:ofm:ofm_ablation_study} for $\mathrm{M3}$ on accuracies.]{The version of \autoref{fig:ofm:ofm_ablation_study} for $\mathrm{M3}$ on accuracies. We observe that besides early spikes, $\mathrm{M3}$ decreases when we add complexity to the target model. Additionally, we observe that $\mathrm{M3}$ decreases when we add augmentations in this case. We measure the $\mathrm{M3}$ for RCAE between the classification error of the target task and the classification error of predicting the rotations of rotated images from PCam (pretext task). For SCLCAE, the pretext task metric measures the ability of the model to correctly detect the representation of each given image in a batch of representations of transformed images. Here we show $\mathrm{M3}$ without a convergence criterium. Best viewed in color.}
\end{figure}

\begin{figure}[t]
	\begin{center}
		\includegraphics[width=1.0\columnwidth]{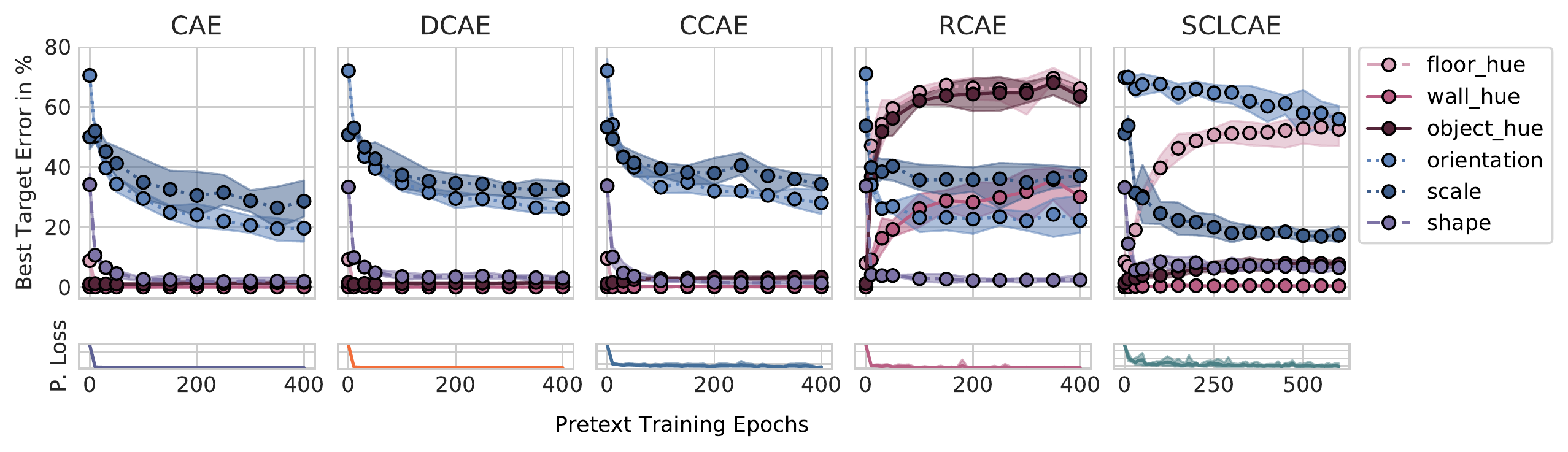} 
	\end{center}
	\vspace{-2ex}
	\caption[The version of \autoref{fig:ofm:pretext_target_losses_3dshapes_big} for $\mathrm{SM3}$ on accuracies.]{The version of \autoref{fig:ofm:pretext_target_losses_3dshapes_big} for $\mathrm{SM3}$ on accuracies. Best viewed in color.}
	\label{fig:ofm:app:m3_3dshapes_big}
\end{figure}

\begin{figure}[t]
	\begin{center}
		\begin{tabular}{cc}
			\includegraphics[width=0.45\columnwidth]{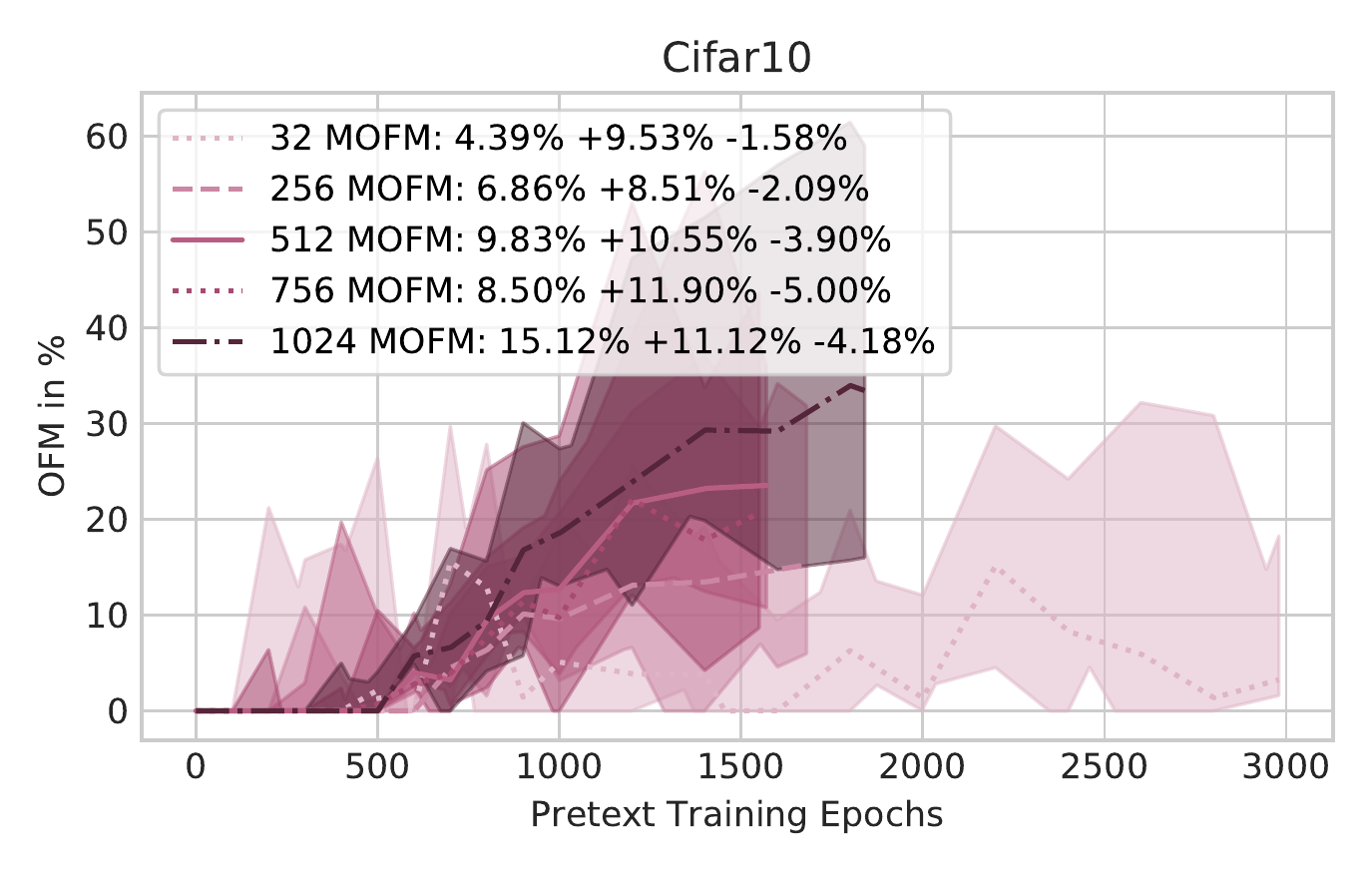}&
			\includegraphics[width=0.45\columnwidth]{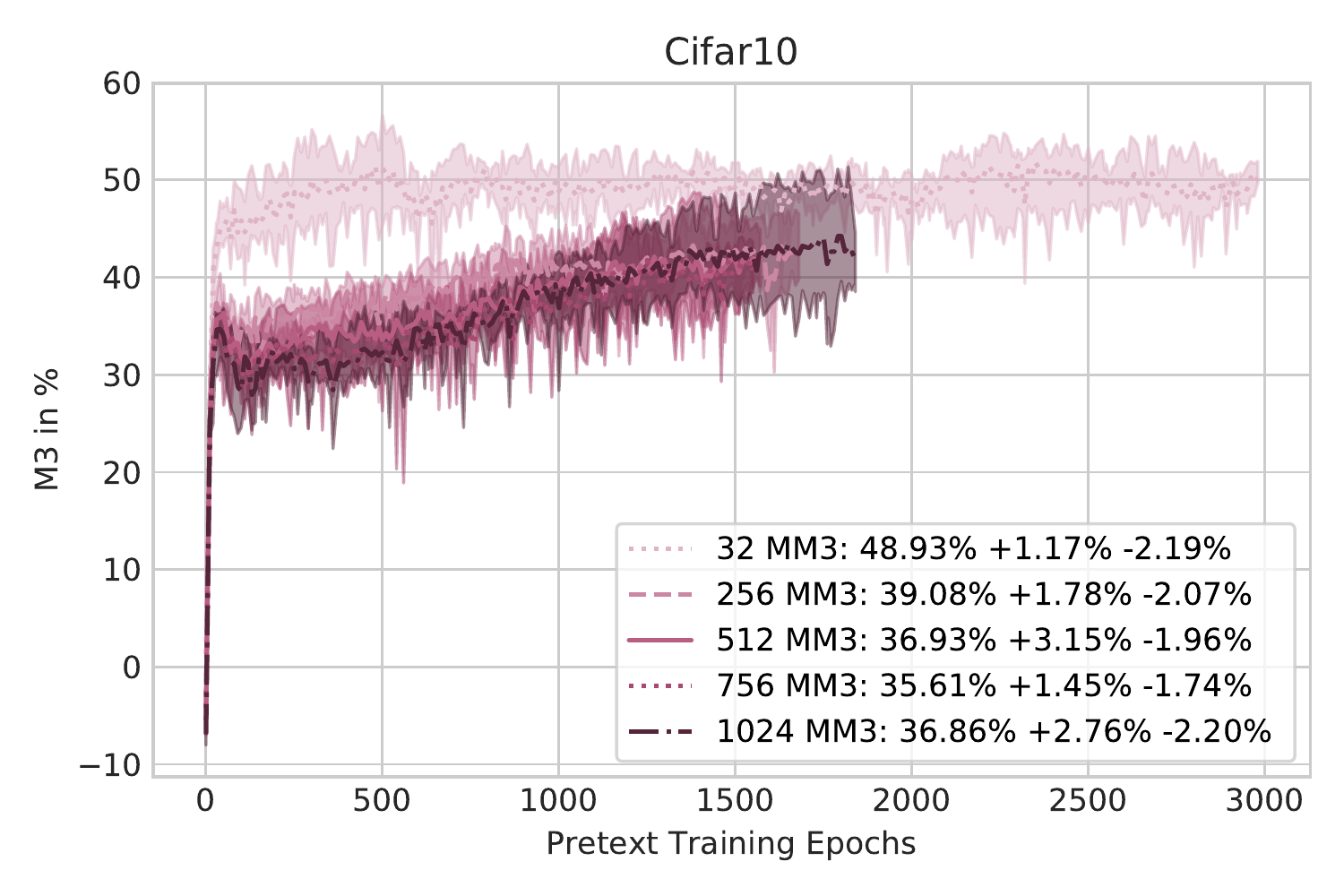}\\
		\end{tabular}  
	\end{center}
	\vspace{-3ex}
	\caption[$\mathrm{OFM}$ and $\mathrm{MM3}$ for other ResNets.]{$\mathrm{OFM}$ and $\mathrm{MM3}$ for other ResNets. (left) The $\mathrm{OFM}$ of ResNets with different representation sizes trained on the pretext task of predicting rotations on Cifar10. Target models are trained for Cifar10 classification. (right) $\mathrm{MM3}$ of those ResNets. We vary representation sizes in $[32,256,512,756,1024]$ by adding a $1\times1$ convolution layer on top of the ResNet18. Thereby the number of filters corresponds to the representation size. In contrast to our observations on our small model, the largest representation we have tested leads to a high $\mathrm{OFM}$. A reason for that could be that the larger representation size helps the model to solve the pretext task, and since there is a mismatch with the target task, a better understanding of this task leads to a higher mismatch. We note that a representation size of $1024$ is still very small for unsupervised learning. Therefore, an even larger representation size could still lead to a lower mismatch.}
	\label{fig:ofm:app:rep_size}
\end{figure}

\begin{figure}[t]
	\begin{center}
		\begin{tabular}{cc}   
			\includegraphics[width=0.45\columnwidth]{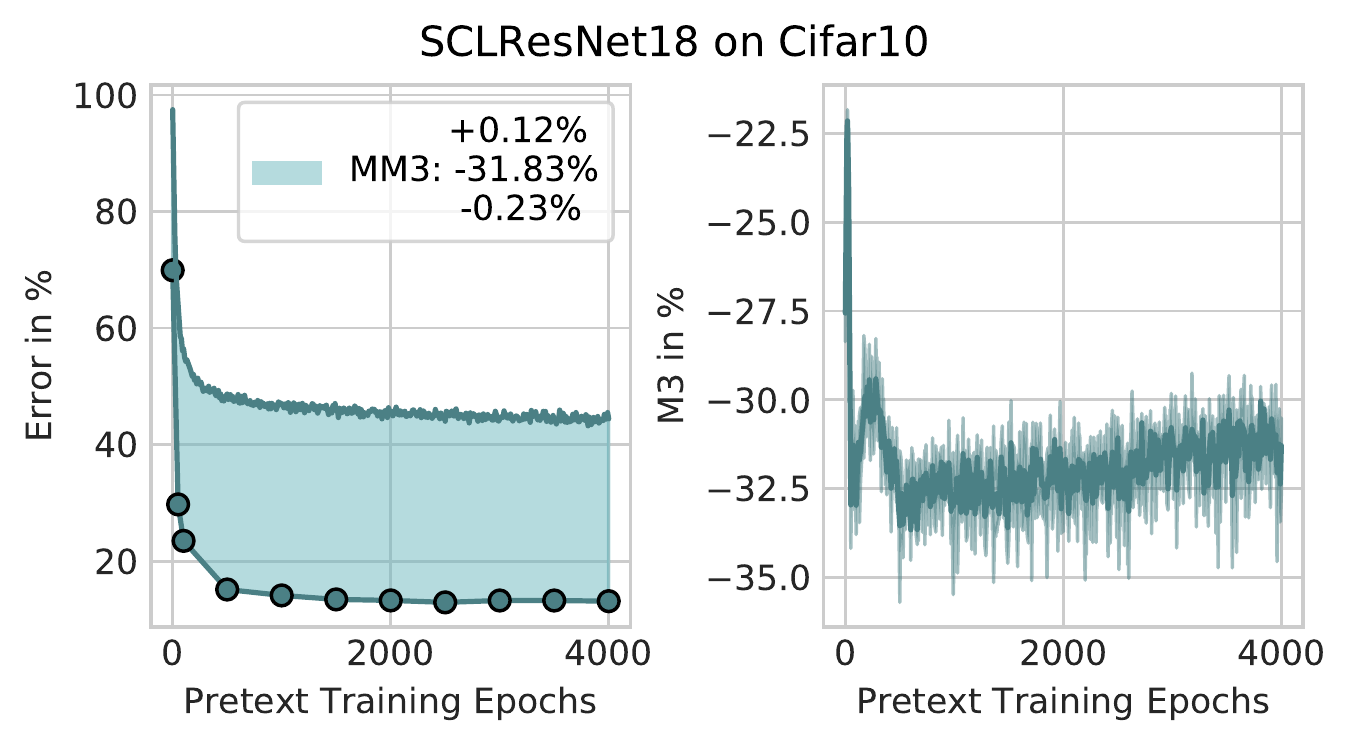} & 
			\includegraphics[width=0.45\columnwidth]{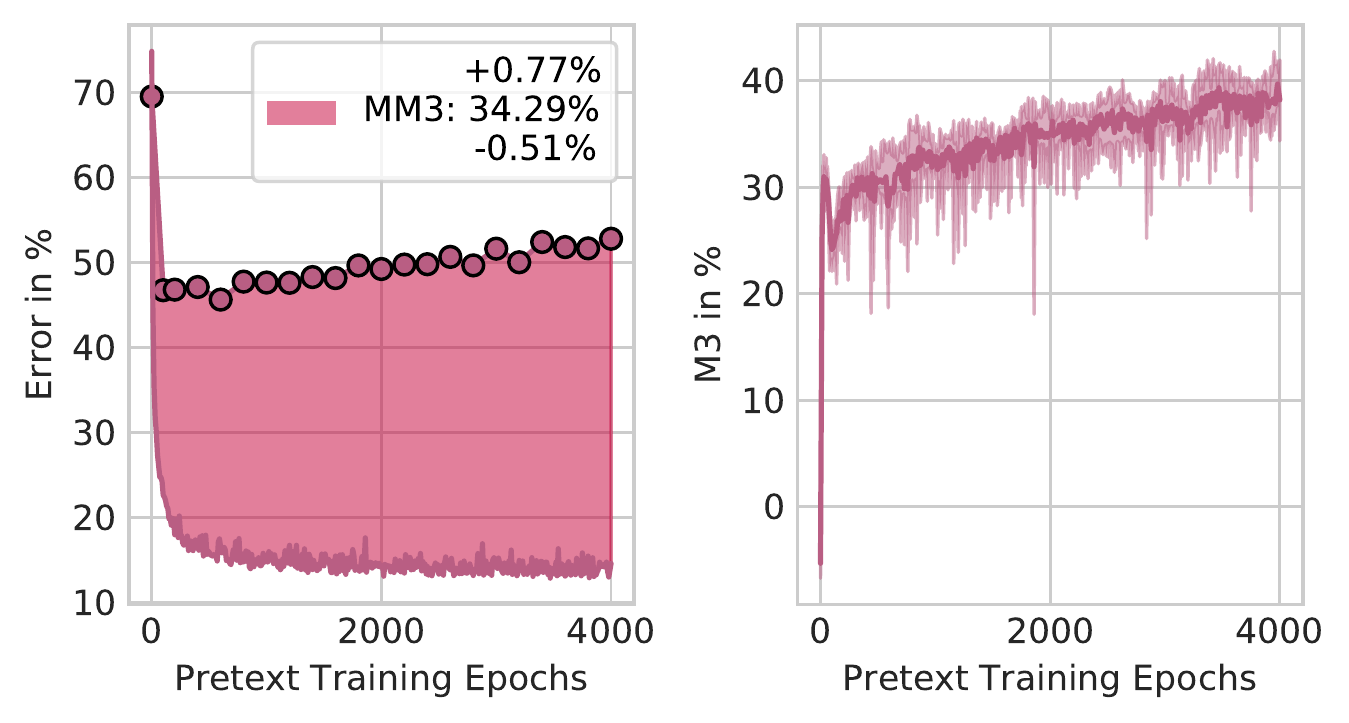}\\
			\includegraphics[width=0.45\columnwidth]{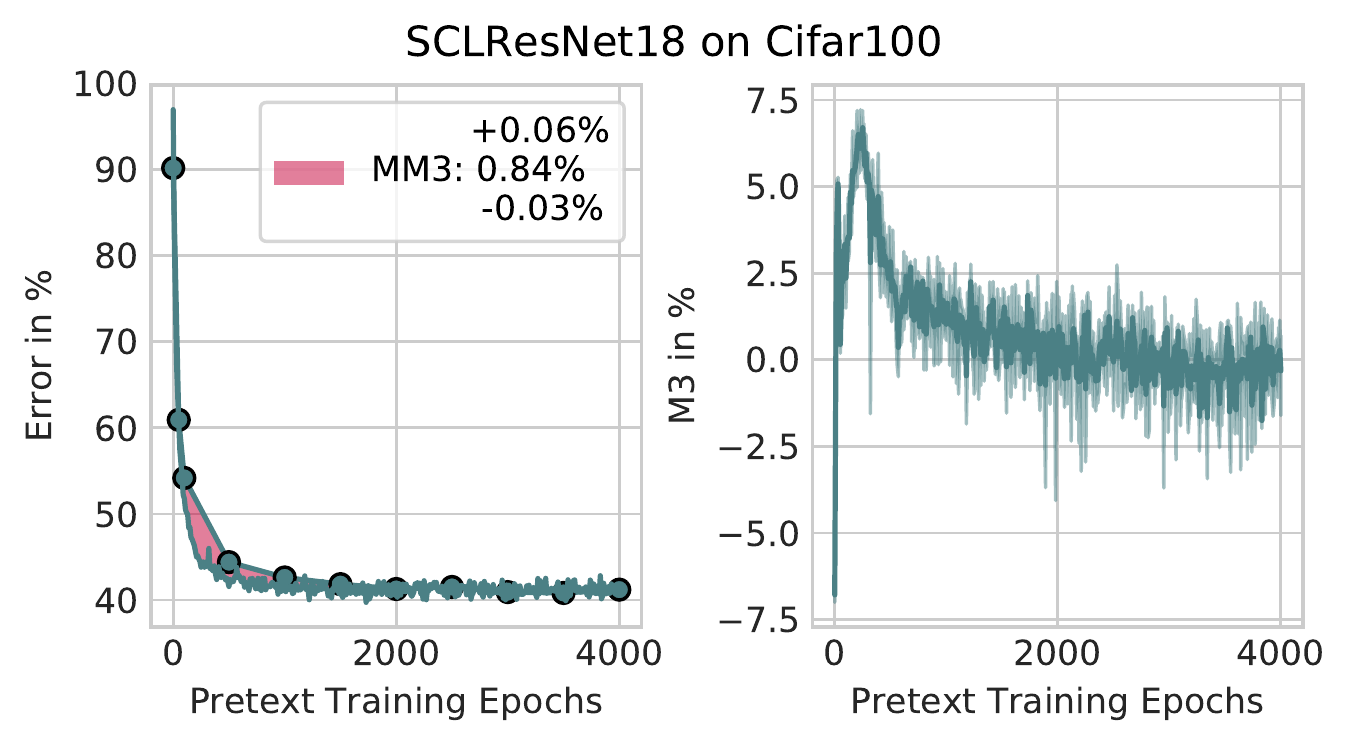} &
			\includegraphics[width=0.45\columnwidth]{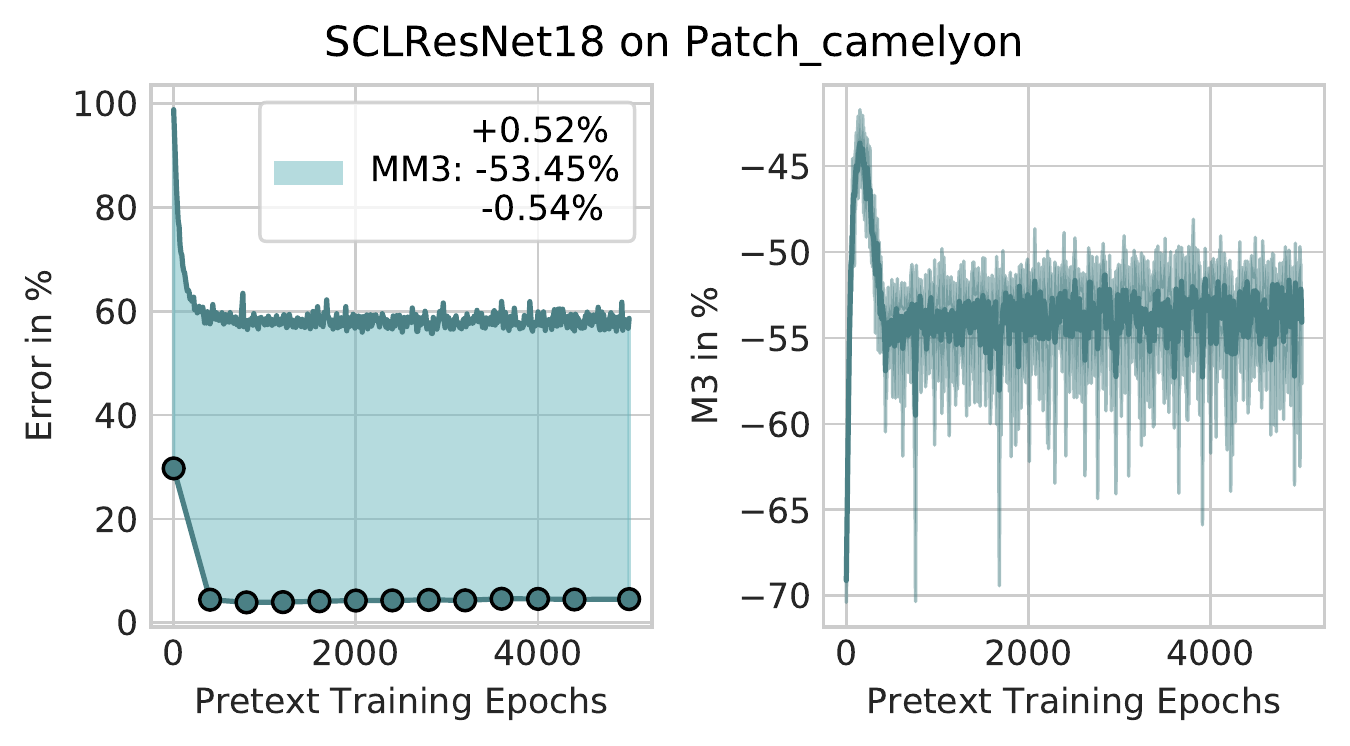}     
		\end{tabular}
	\end{center}
	\vspace{-3ex}
	\caption[$\mathrm{MM3}$ for different pretext tasks trained with a ResNet18 model as backbone.]{$\mathrm{MM3}$ for different pretext tasks trained with a ResNet18 model as the backbone. (top right) The mismatches are shown for the entire training process. In contrast to the prediction of rotations, SCLResNet18 has a high negative $\mathrm{MM3}$ for Cifar10. This indicates that learning the contrastive pretext task is better suited for distinguishing Cifar10 classes than the prediction of rotations. Furthermore, the error of the contrastive pretext task is significant, which indicates that the model still underfits the pretext task with this setup, and there is more room for improvement. For the 100 classes of Cifar100, $\mathrm{MM3}$ becomes slightly positive in the contrastive learning setup. For contrastive learning on the PCam dataset and rotation prediction on Cifar10, we observe an increasing mismatch during training.}
	\label{fig:ofm:app:resnet_mismatches_mm3}
\end{figure}
\chapter{Supplementary Material:\\A Lane Detection Benchmark for Multi-Target Domain Adaptation}
\label{app:02}

\section{Example Usage of the CARLANE Benchmark}
A Jupyter notebook with a tutorial on how to read the datasets for use in PyTorch can be found at \href{https://carlanebenchmark.github.io}{https://carlanebenchmark.github.io}. 

\section{Model Vehicle Description}
\begin{figure}[ht]
	\begin{center}
		\includegraphics[width=0.3\linewidth]{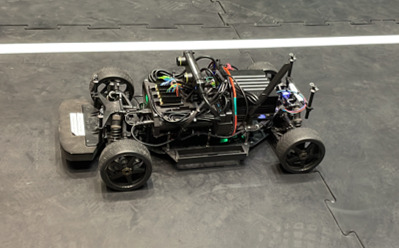}
	\end{center}
	\vspace{-1ex}
	\caption[Picture of the 1/8th model vehicle we built to capture images in our 1/8th target domain.]{Picture of the 1/8th model vehicle we built to capture images in our 1/8th target domain.} 
	\label{fig:carlane:app:model_vehicle}
\end{figure}

In \autoref{fig:carlane:app:model_vehicle}, the self-built 1/8th model vehicle is shown, which we have used to gather the images for the 1/8th scaled target domain. An NVIDIA Jetson AGX is the central computation unit powered by a separate Litionite Tanker Mini 25000mAh battery. For image collection, we utilize the software framework ROS Melodic and a Stereolabs ZEDM stereo camera with an integrated IMU. The camera is directly connected to the AGX and captures images with a resolution of $1280 \times 720$ pixels and a rate of $30$ FPS.

\section{Reproducibility of the Baselines}
To ensure reproducibility, we strictly follow UFLD \cite{qin2020ultra} and the corresponding unsupervised domain adaptation method for model architecture and hyperparameters. Thereby, we utilize UFLD as an encoder for the unsupervised domain adaptation method. We provide a detailed table of the tuned hyperparameters, architecture changes, and objectives in the main text. In addition, the trained weights of our baselines, their entire implementation, and the configuration files of our baselines are made publicly available at \href{https://carlanebenchmark.github.io}{https://carlanebenchmark.github\\.io}. 
\\\\
\textbf{Initialization}. We initialize convolutional layer weights with kaiming normal and their biases with 0.0. Linear layer weights are initialized with normal (mean = 0.0, std = 0.01), batch normalization weights, and biases are initialized with 1.0.

\section{Additional Results}
\begin{figure}
	\small
	\begin{center}
		\begin{tabular}{c@{}c@{}c@{}c@{}c}
			UFLD-SO & DANN & ADDA & SGADA & SGPCS \\
			\includegraphics[width=0.2\linewidth,valign=m]{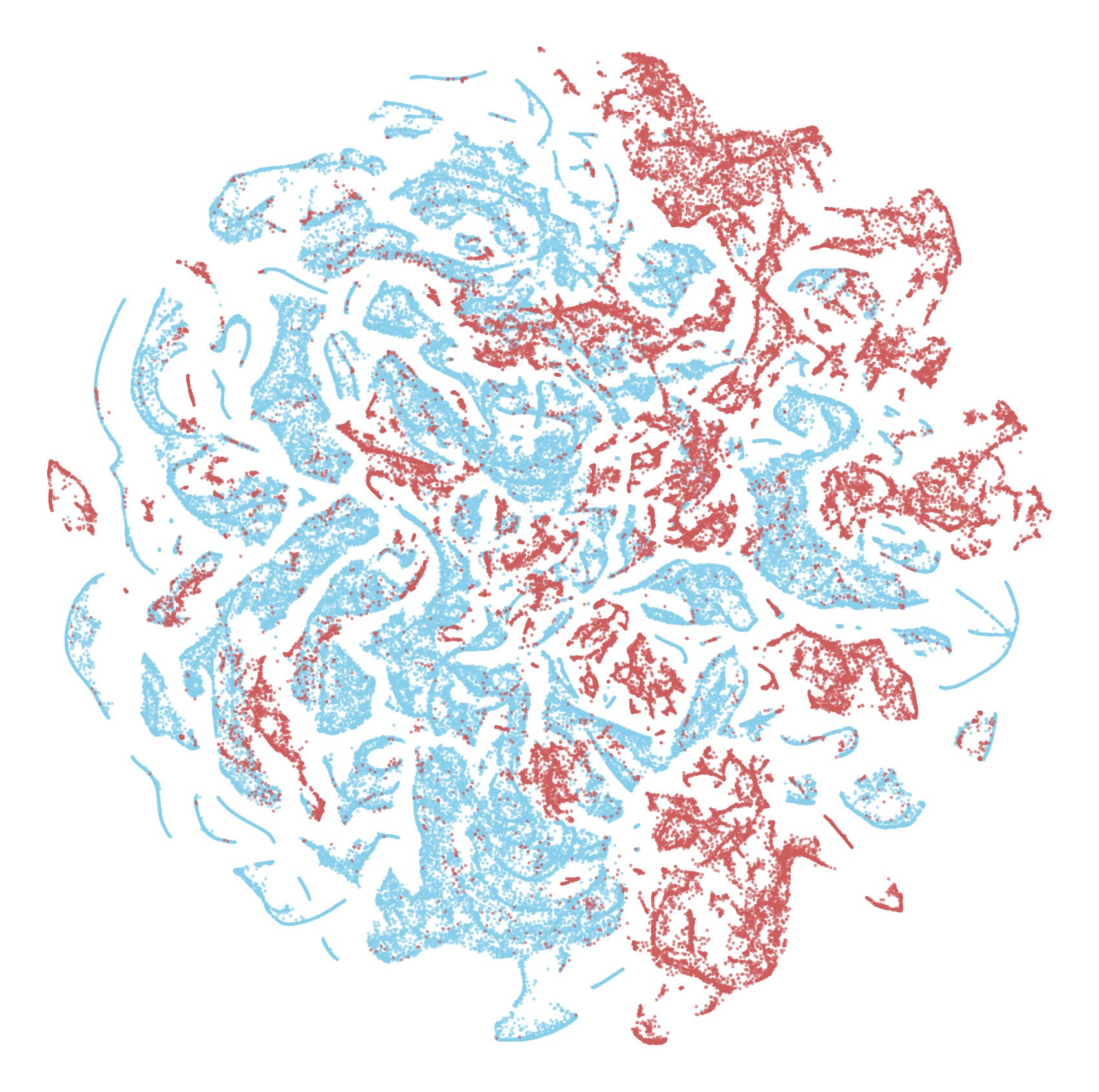} & \includegraphics[width=0.2\linewidth,valign=m]{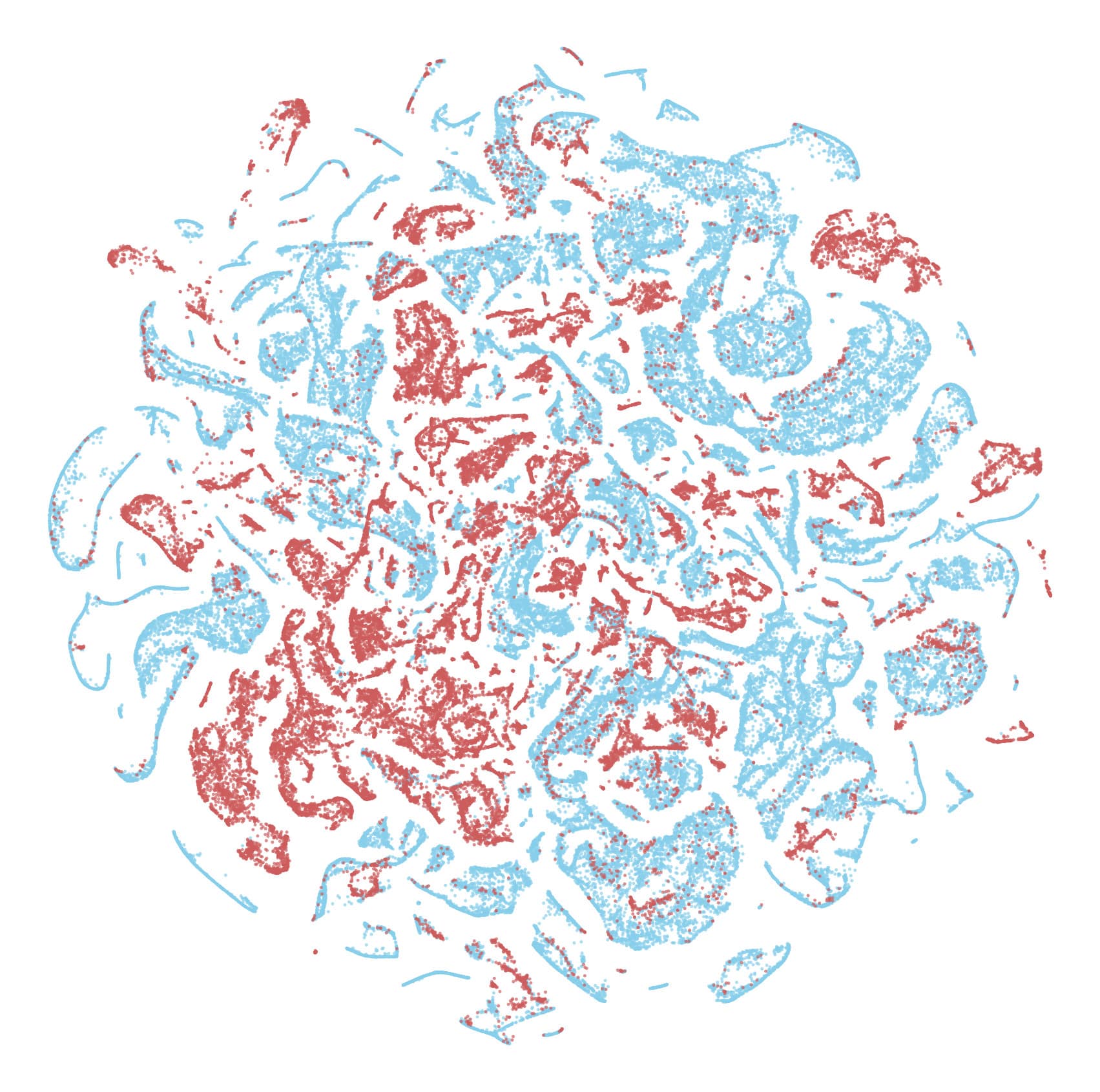} &
			\includegraphics[width=0.2\linewidth,valign=m]{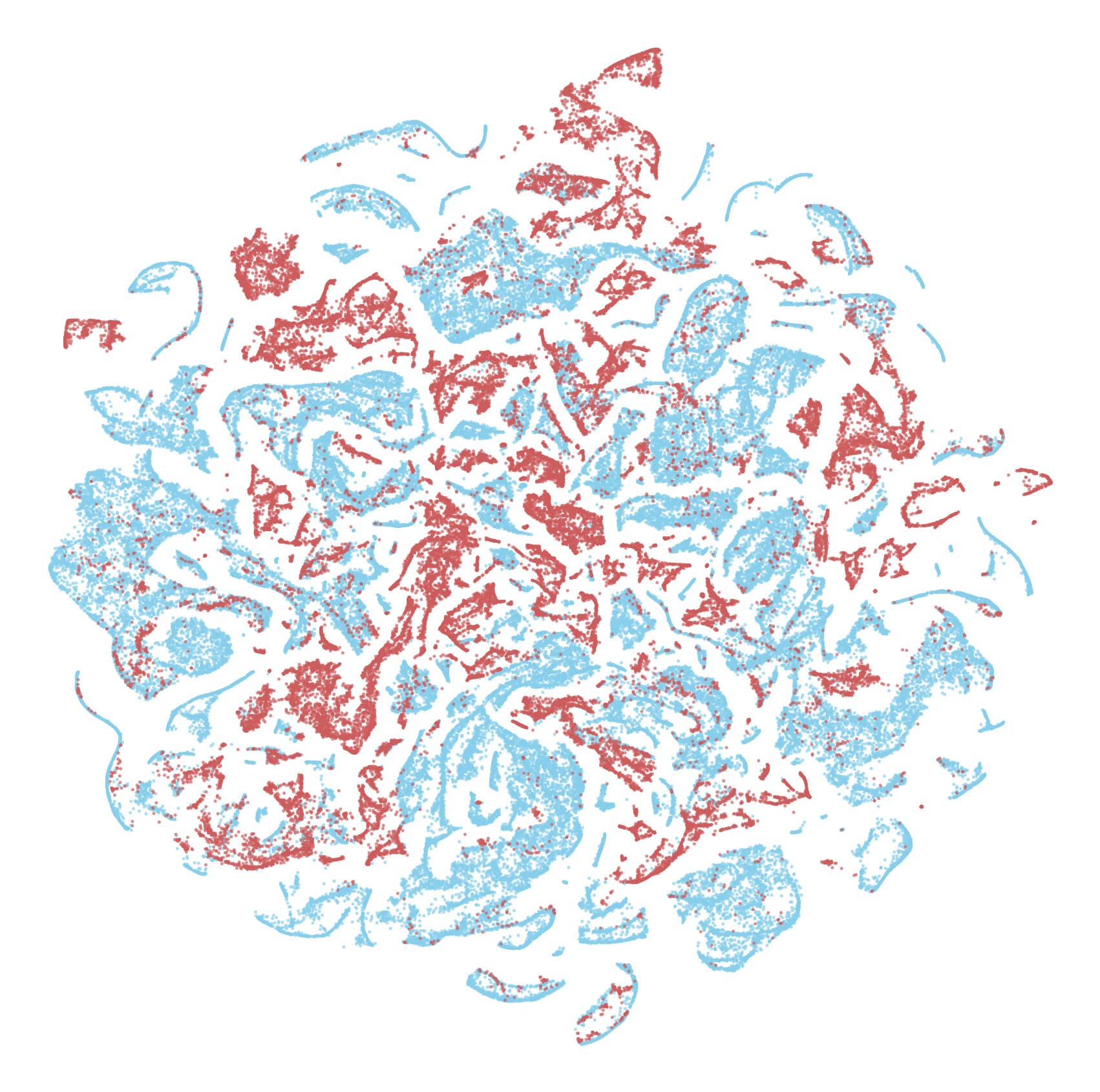} &
			\includegraphics[width=0.2\linewidth,valign=m]{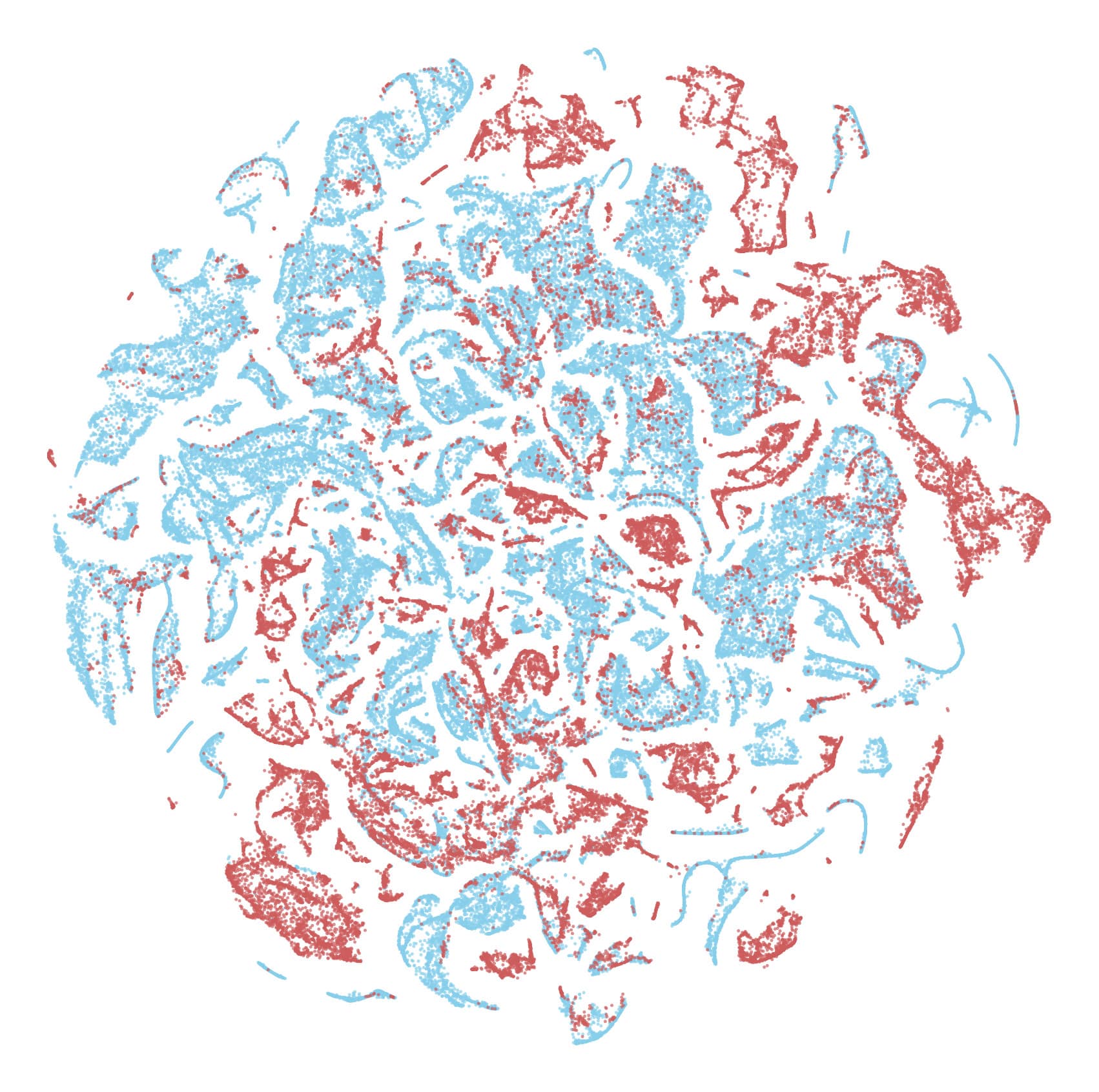} & \includegraphics[width=0.2\linewidth,valign=m]{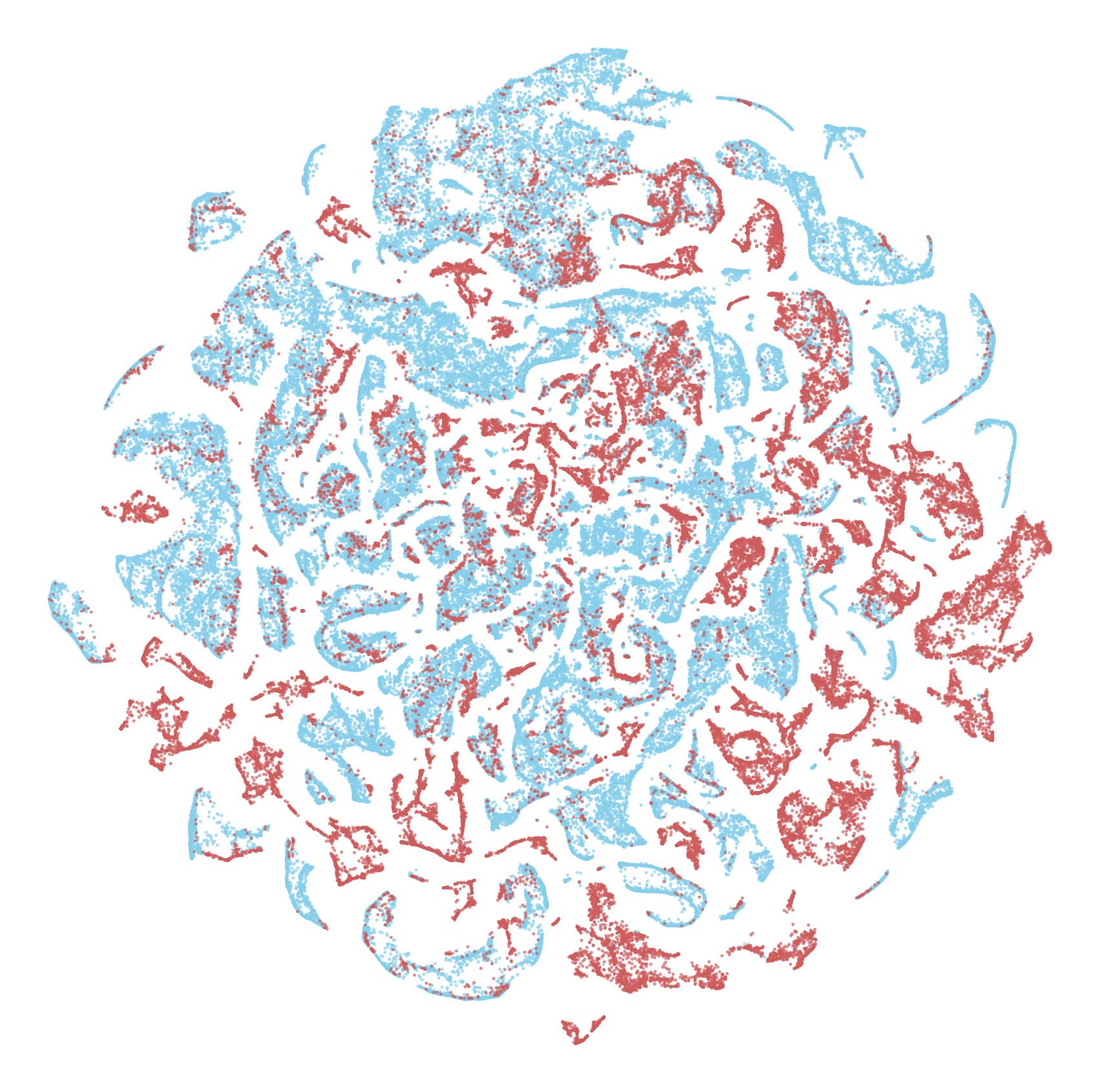}\\
			\includegraphics[width=0.2\linewidth,valign=m]{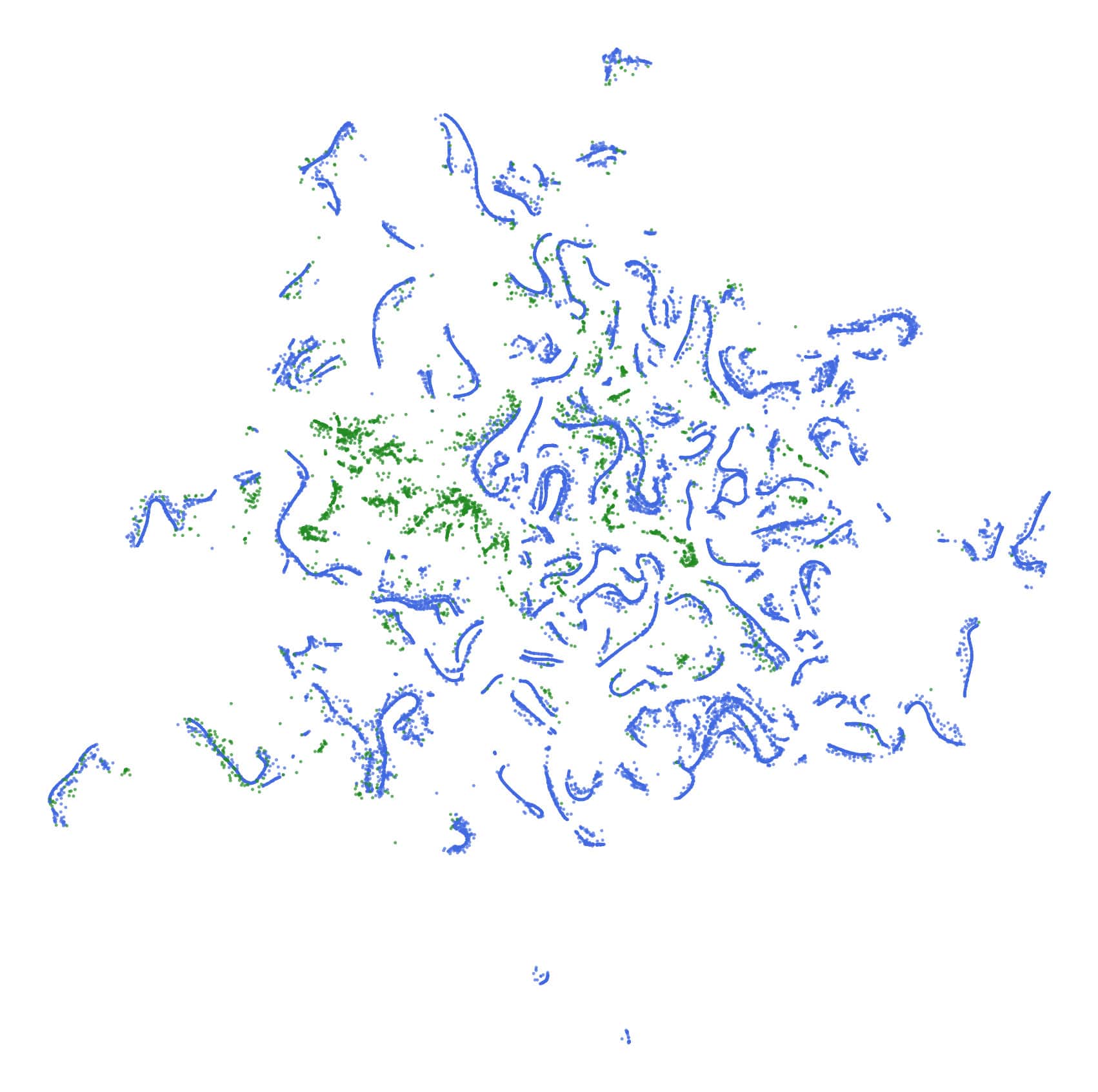} & \includegraphics[width=0.2\linewidth,valign=m]{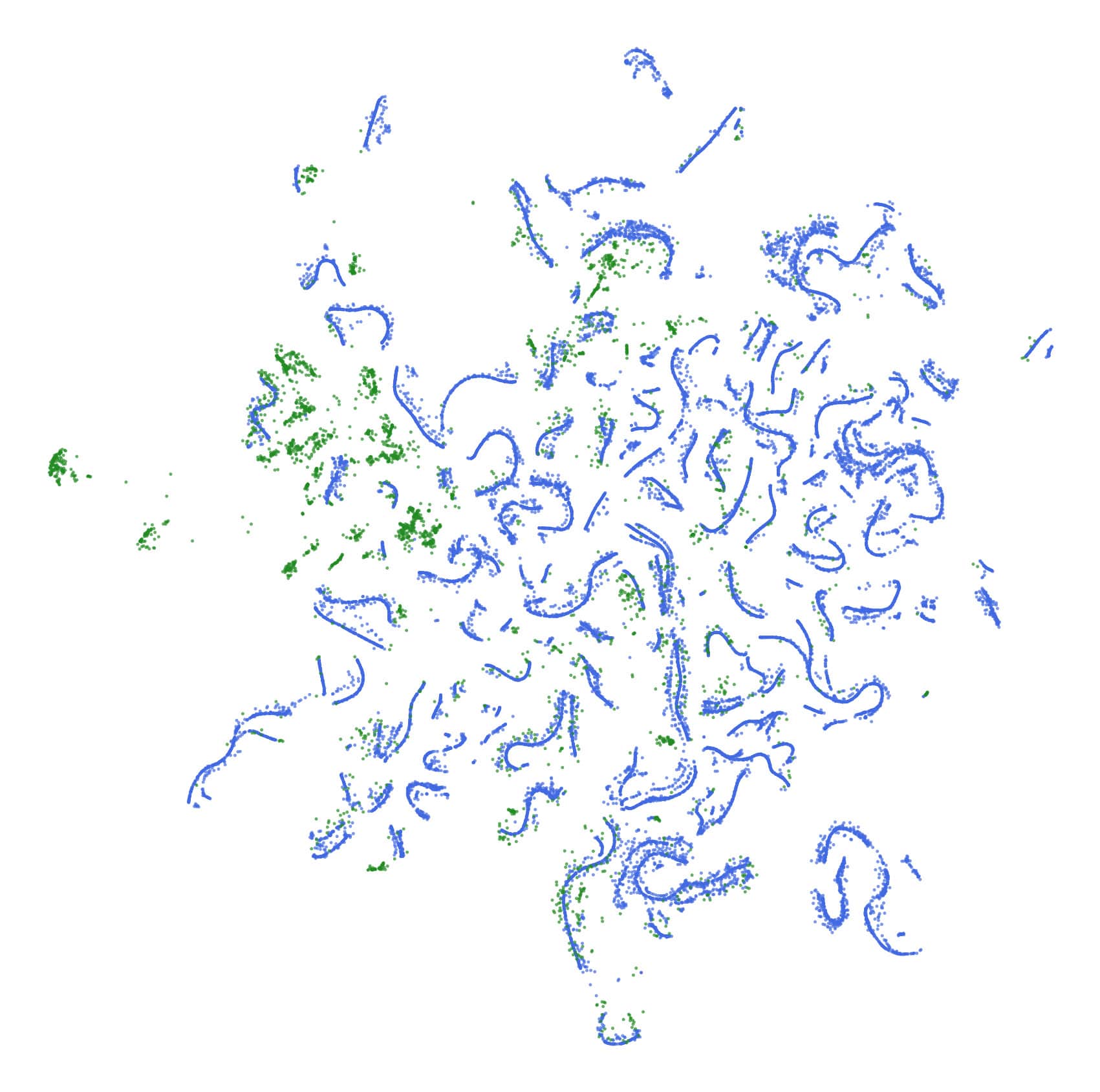} &
			\includegraphics[width=0.2\linewidth,valign=m]{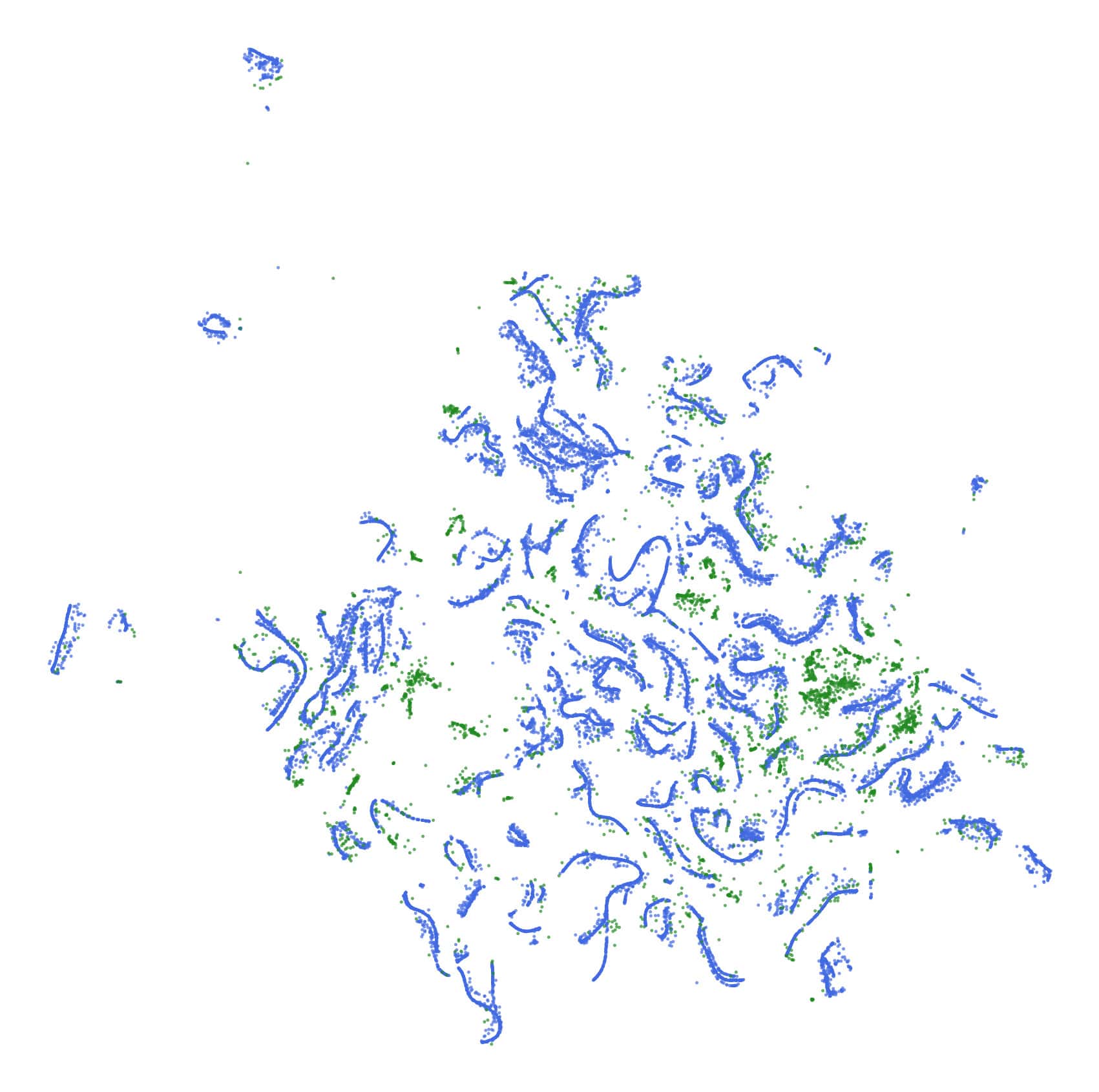} &
			\includegraphics[width=0.2\linewidth,valign=m]{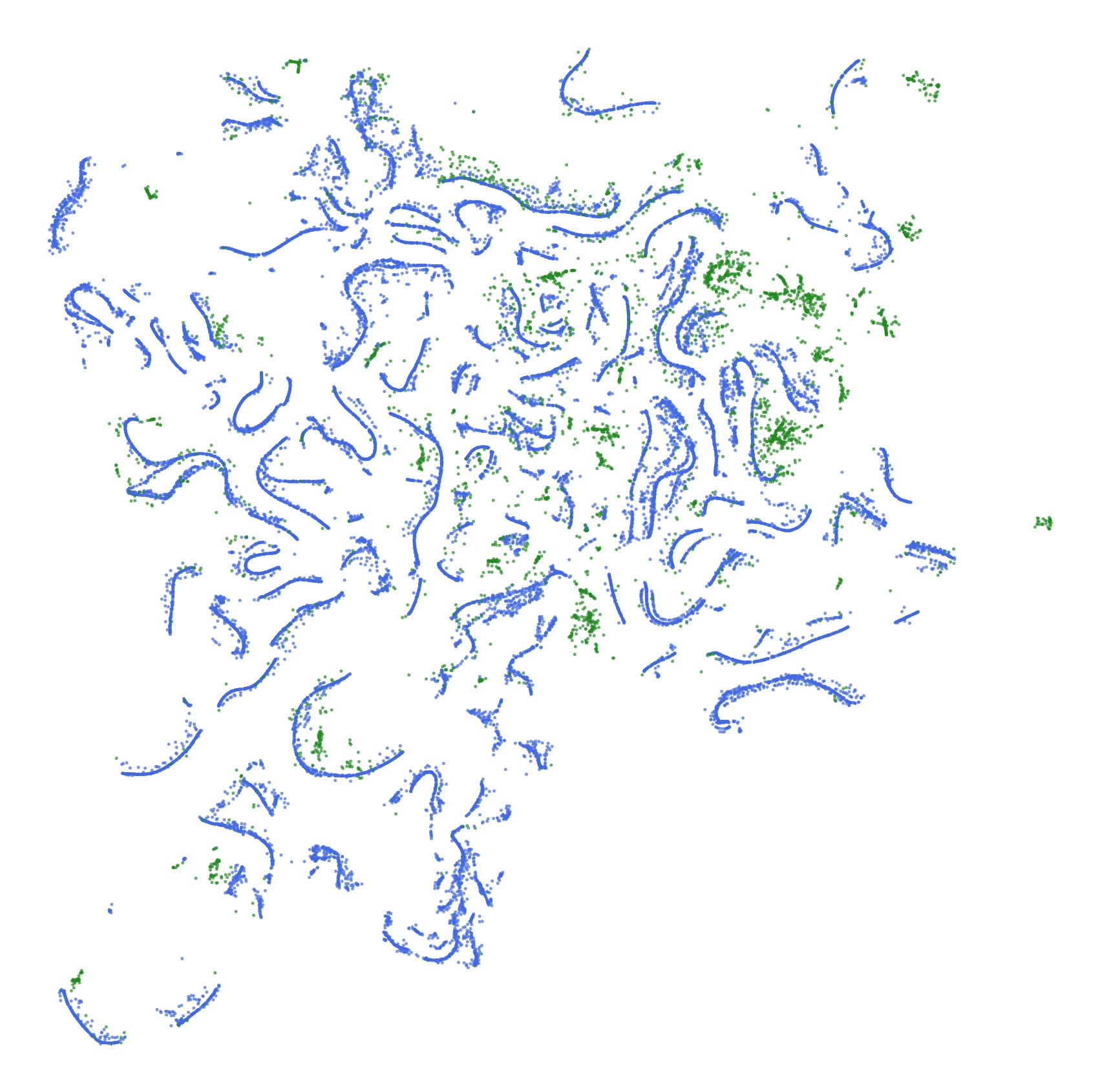} & \includegraphics[width=0.2\linewidth,valign=m]{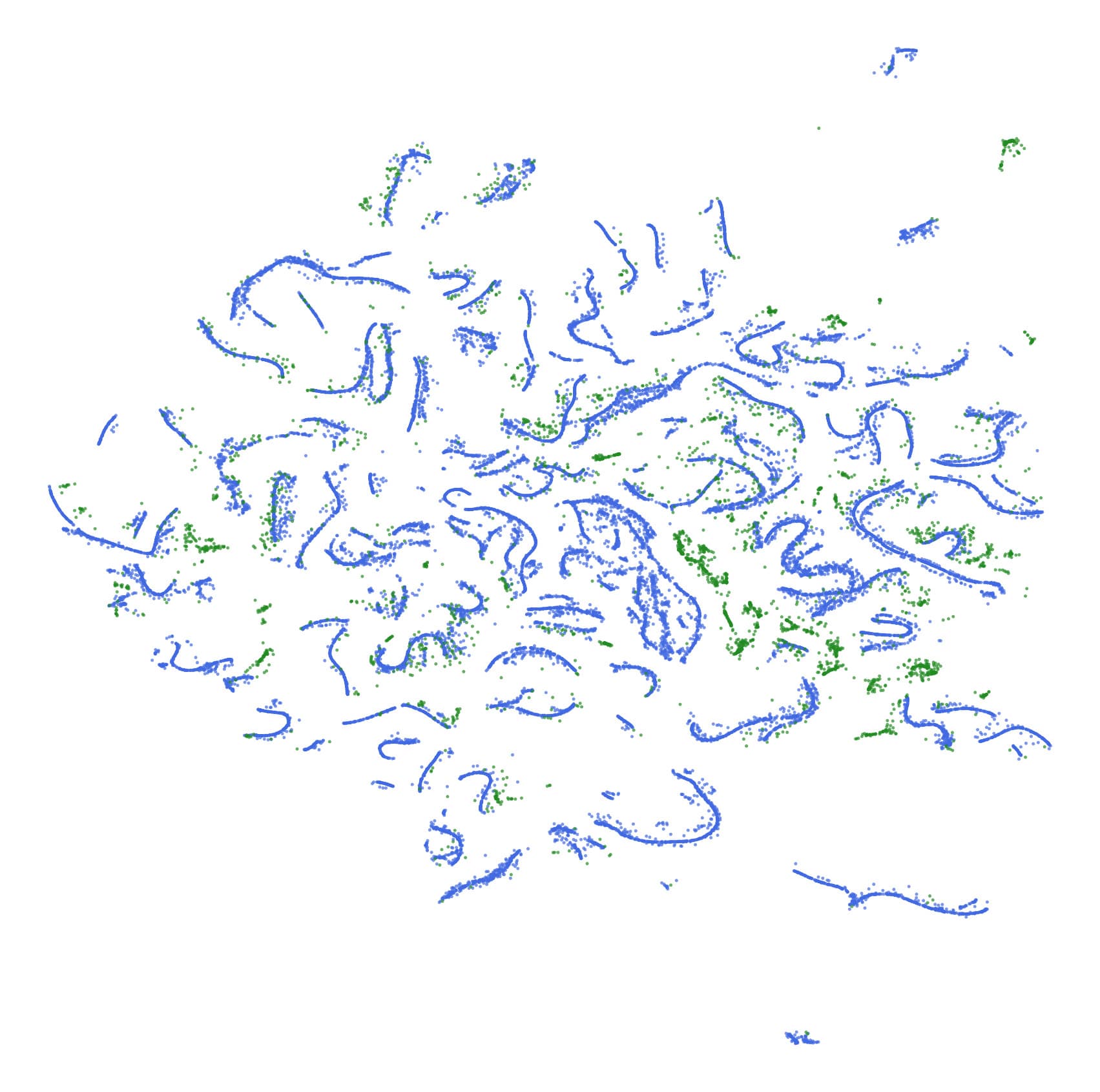}\\
		\end{tabular}
	\end{center}
	\vspace{-1ex}
	\caption[\acs{t-SNE} visualizations of the MoLane dataset and the TuLane dataset.]{\acs{t-SNE} visualizations of the MoLane dataset (top) and the TuLane dataset (bottom). The source domain is marked in blue, the real-world model vehicle target domain in red, and TuLane's target domain in green. Best viewed in color.}
	\label{fig:carlane:app:TSNE_plot_molane_tulane}
\end{figure}

\begin{figure}
	\small
	\begin{center}
		\begin{tabular}{rc@{}c@{}c@{}c}
			~ & MoLane & TuLane & \multicolumn{2}{c}{MuLane} \\
			UFLD-SO & 
			\includegraphics[width=0.18\linewidth,valign=m]{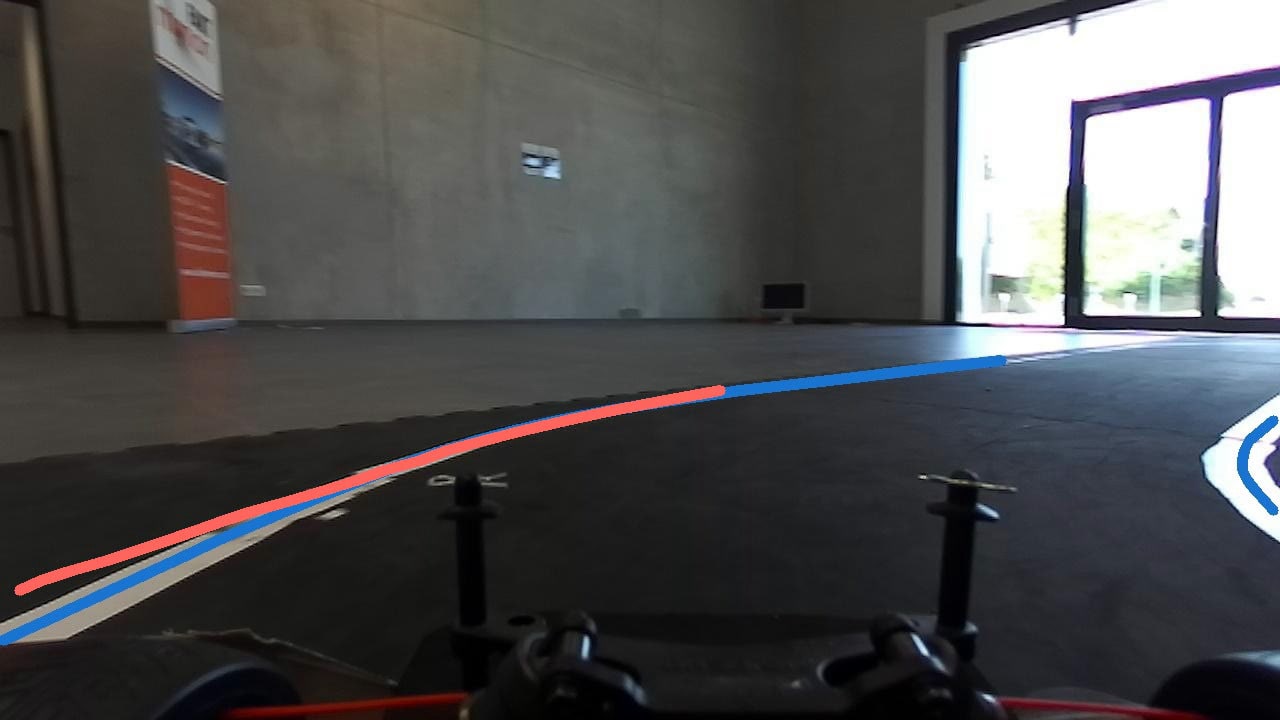} & \includegraphics[width=0.18\linewidth,valign=m]{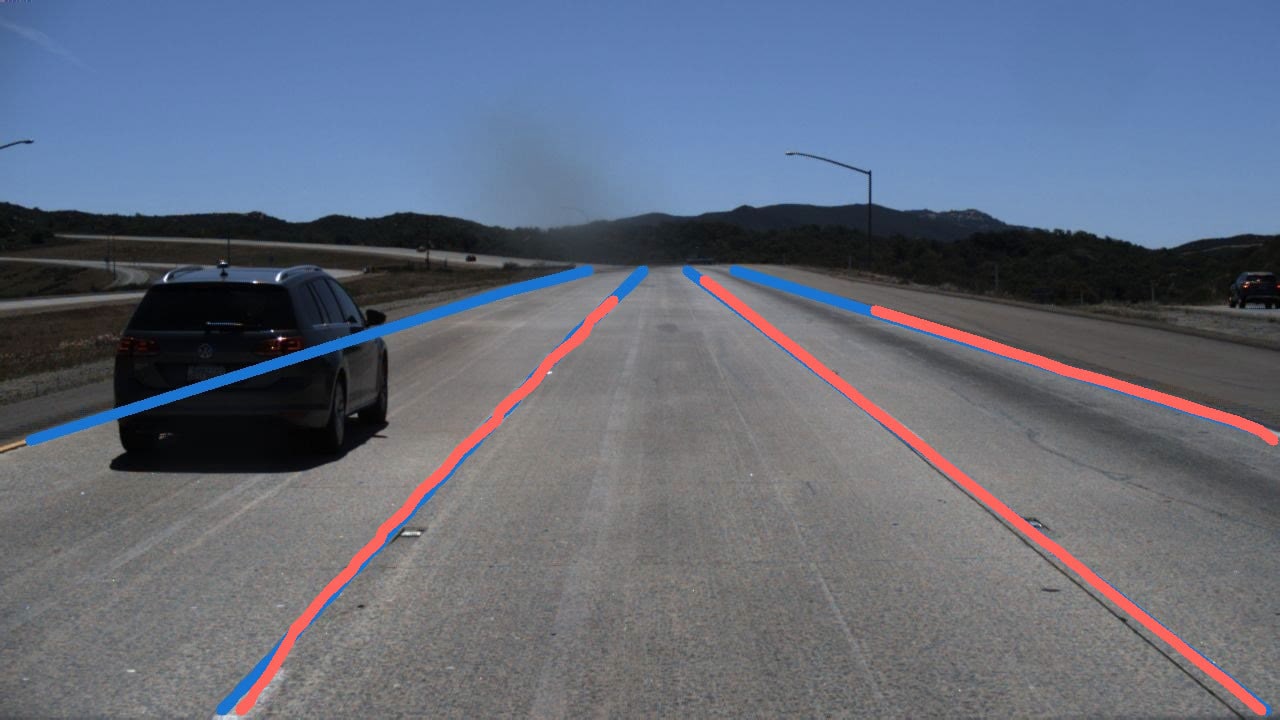} &
			\includegraphics[width=0.18\linewidth,valign=m]{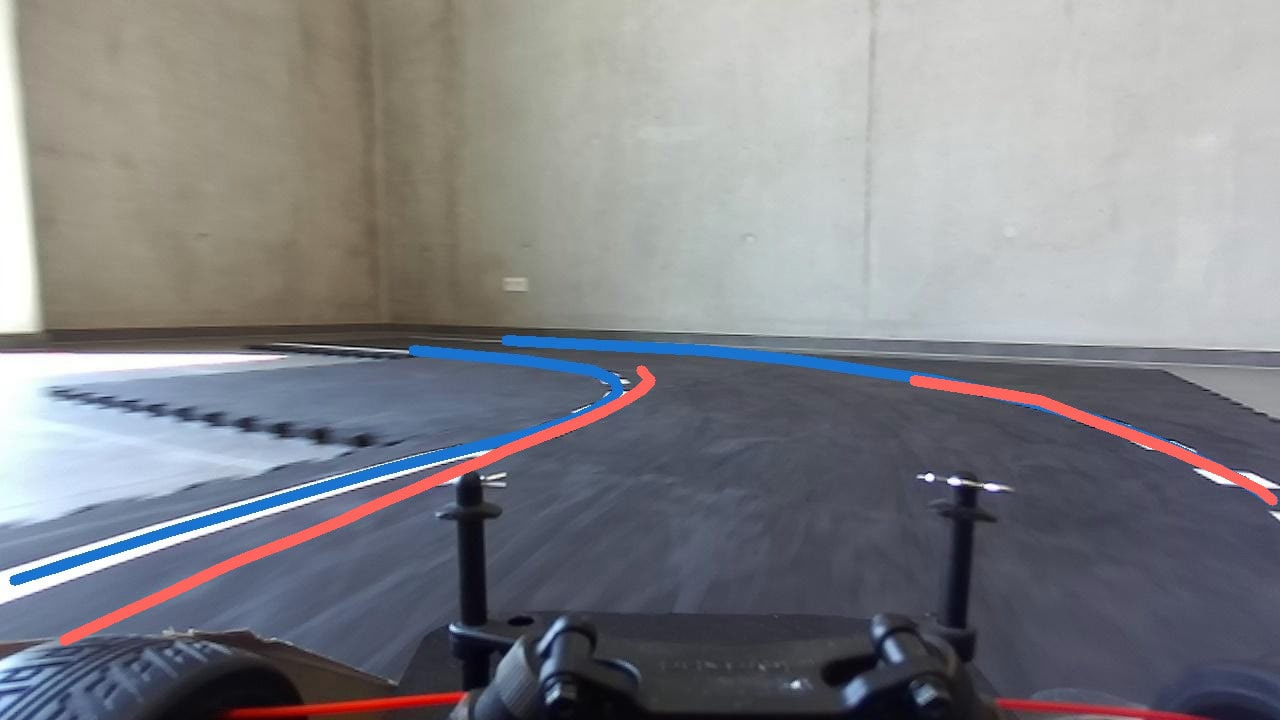} & \includegraphics[width=0.18\linewidth,valign=m]{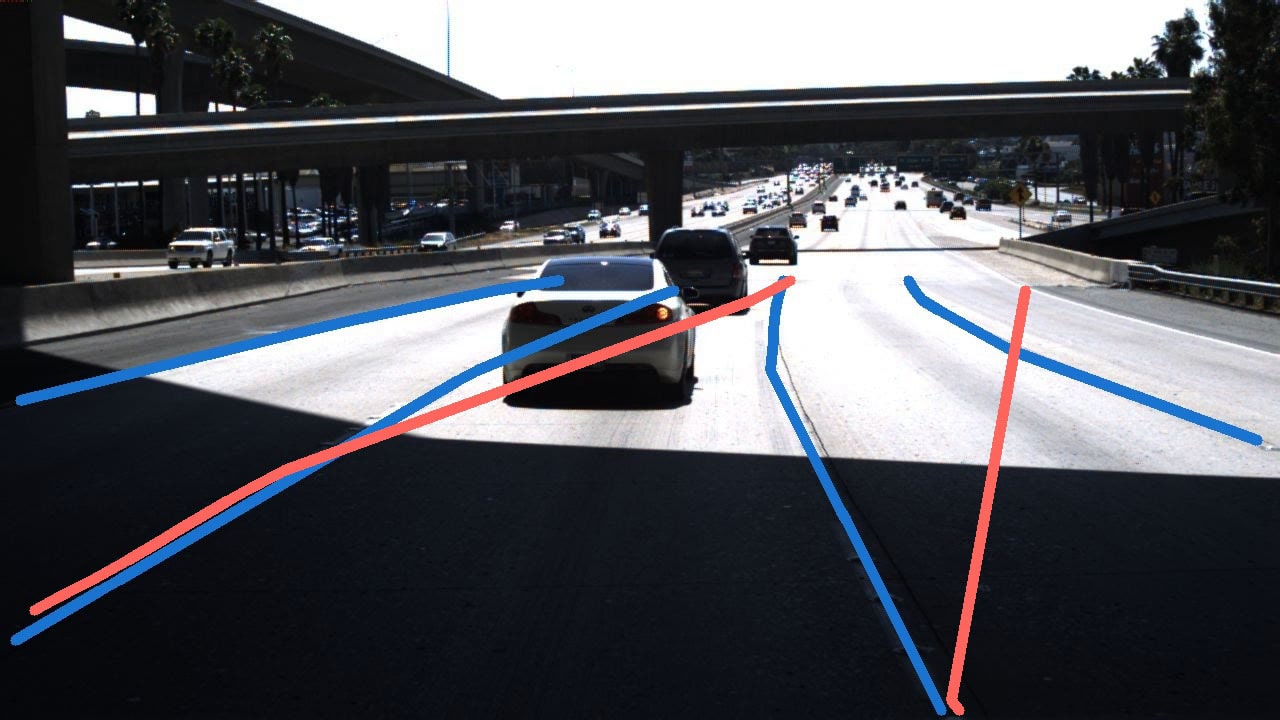}\\
			DANN & 
			\includegraphics[width=0.18\linewidth,valign=m]{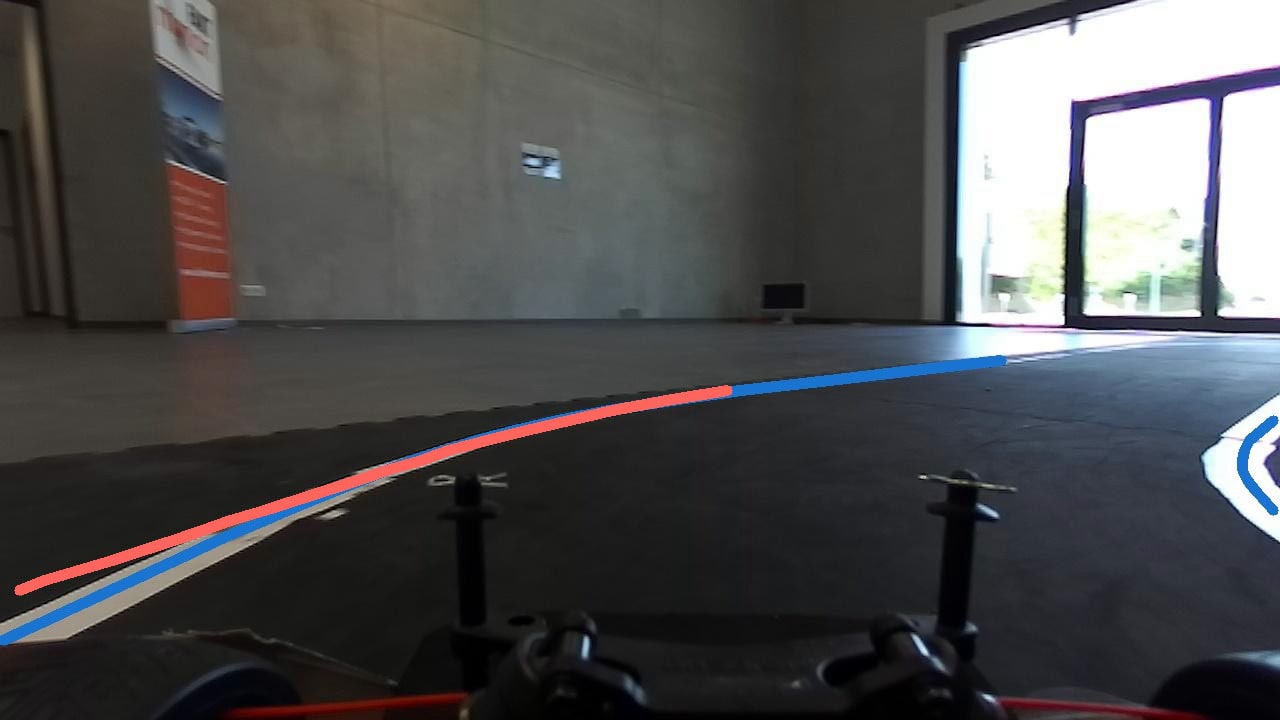} & 
			\includegraphics[width=0.18\linewidth,valign=m]{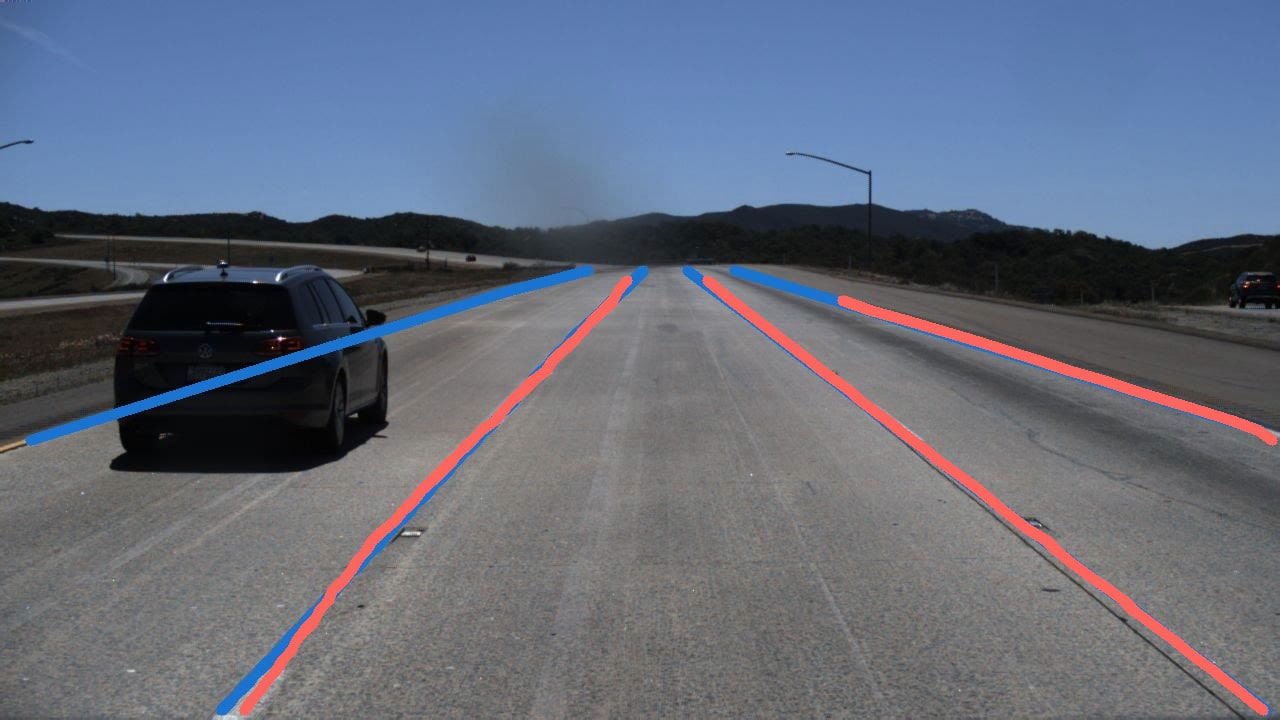} &
			\includegraphics[width=0.18\linewidth,valign=m]{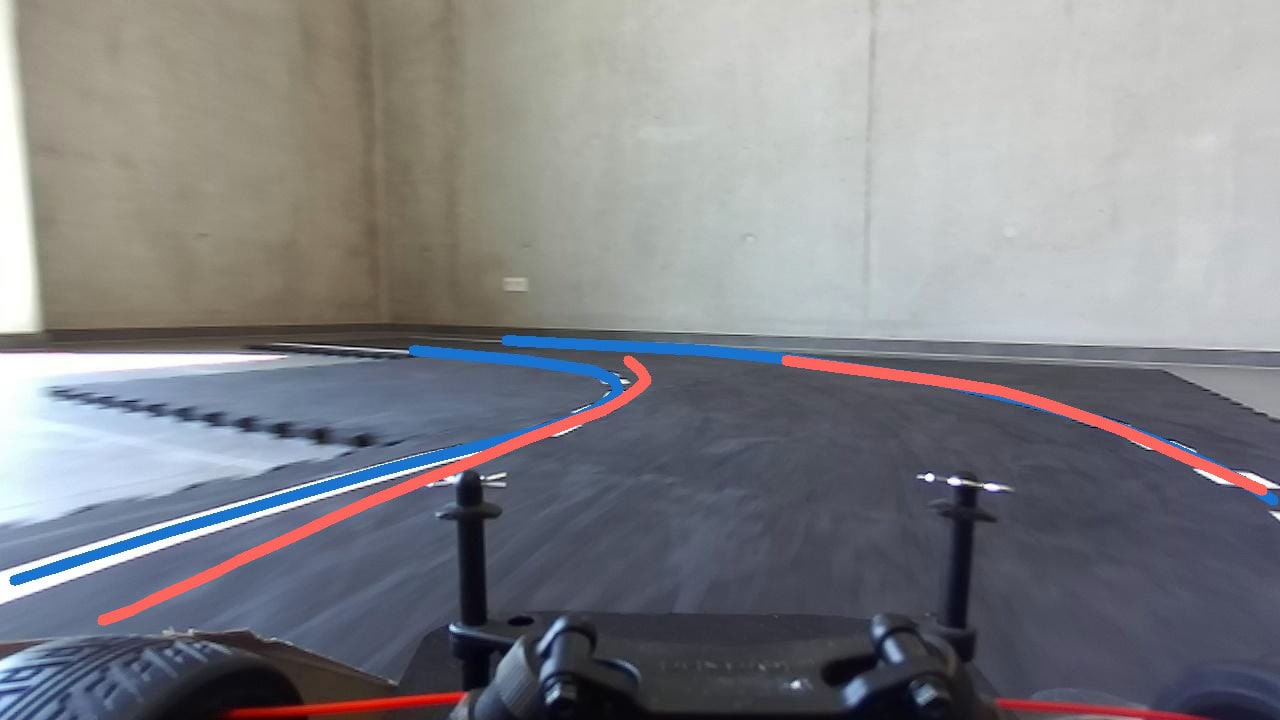} & \includegraphics[width=0.18\linewidth,valign=m]{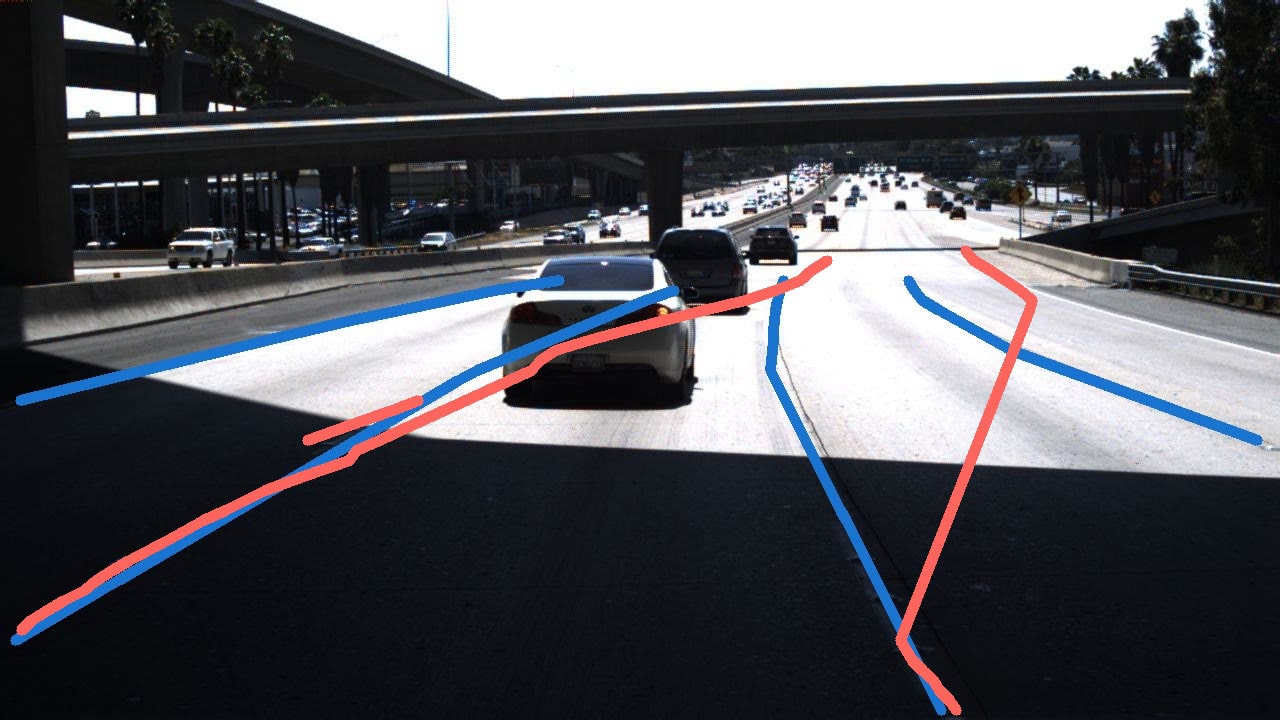}\\
			ADDA & 
			\includegraphics[width=0.18\linewidth,valign=m]{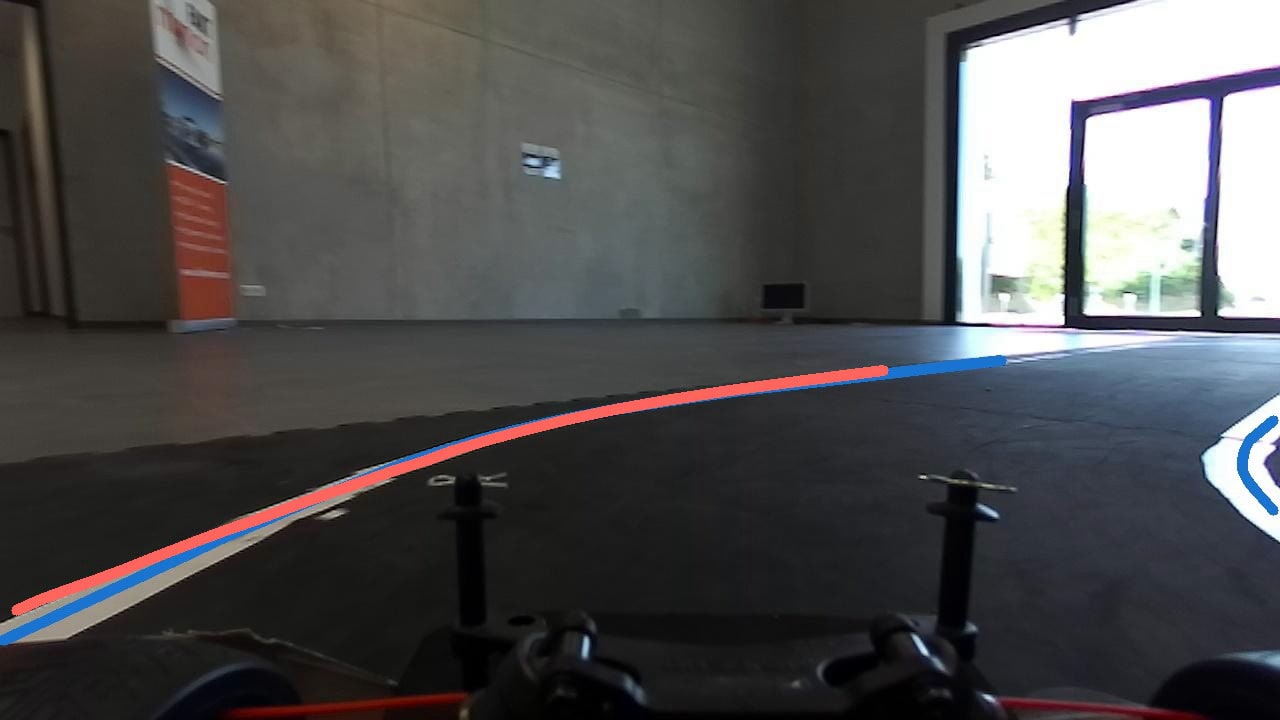} & 
			\includegraphics[width=0.18\linewidth,valign=m]{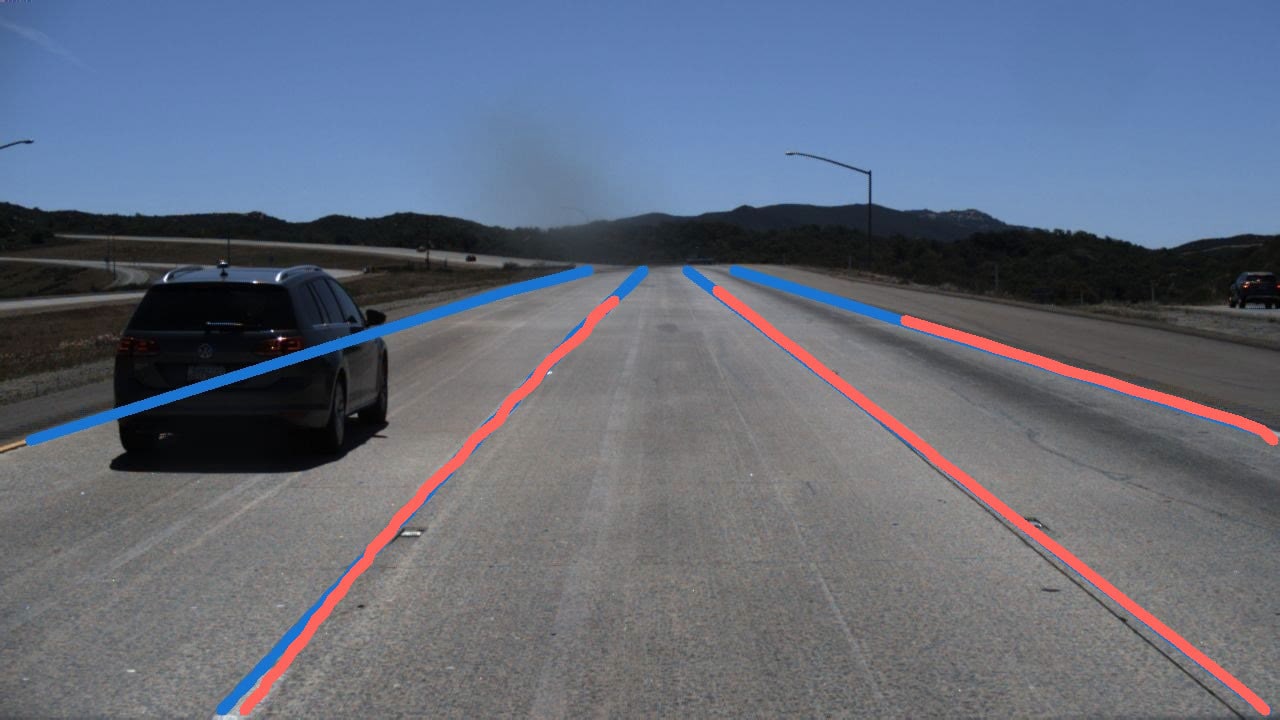} &
			\includegraphics[width=0.18\linewidth,valign=m]{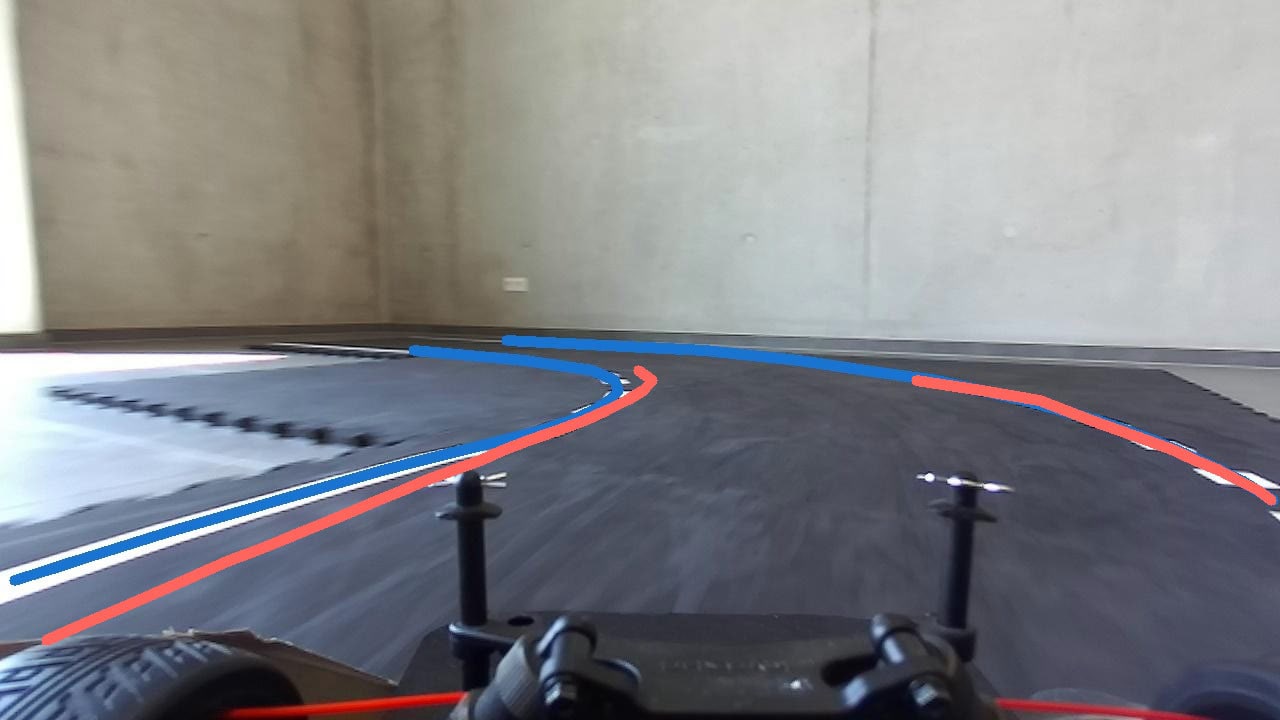} & \includegraphics[width=0.18\linewidth,valign=m]{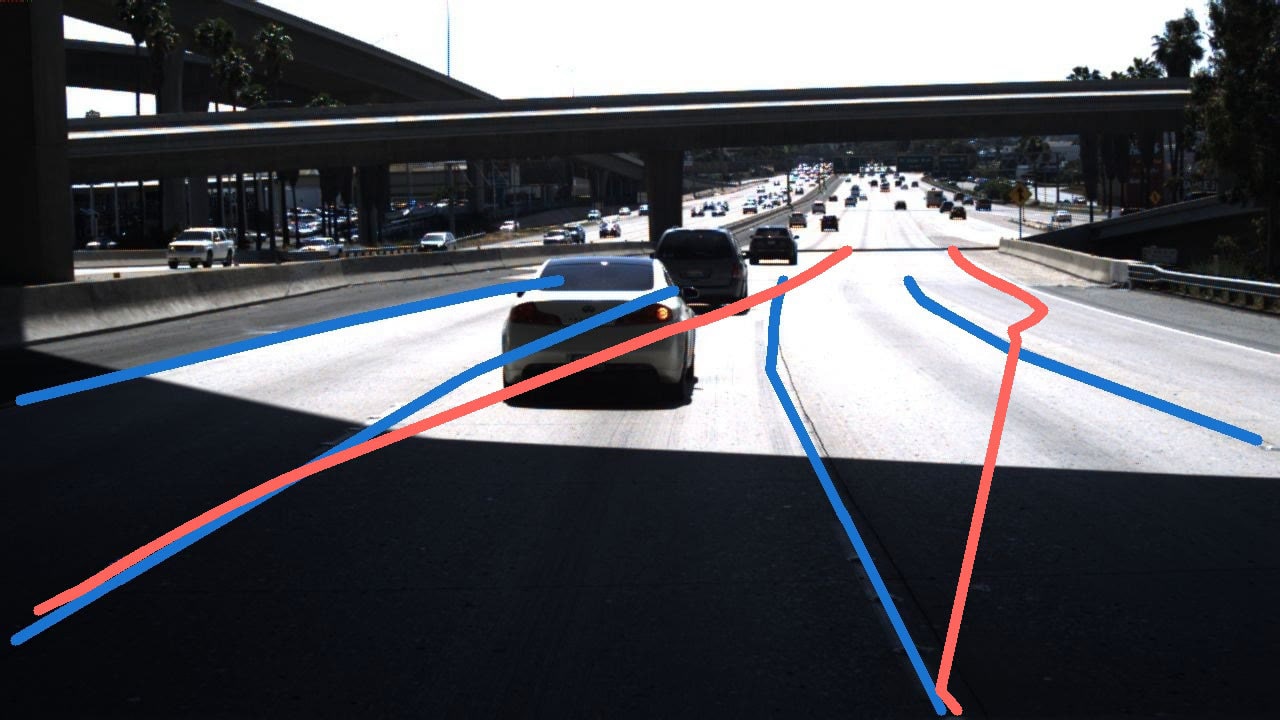}\\
			SGADA & 
			\includegraphics[width=0.18\linewidth,valign=m]{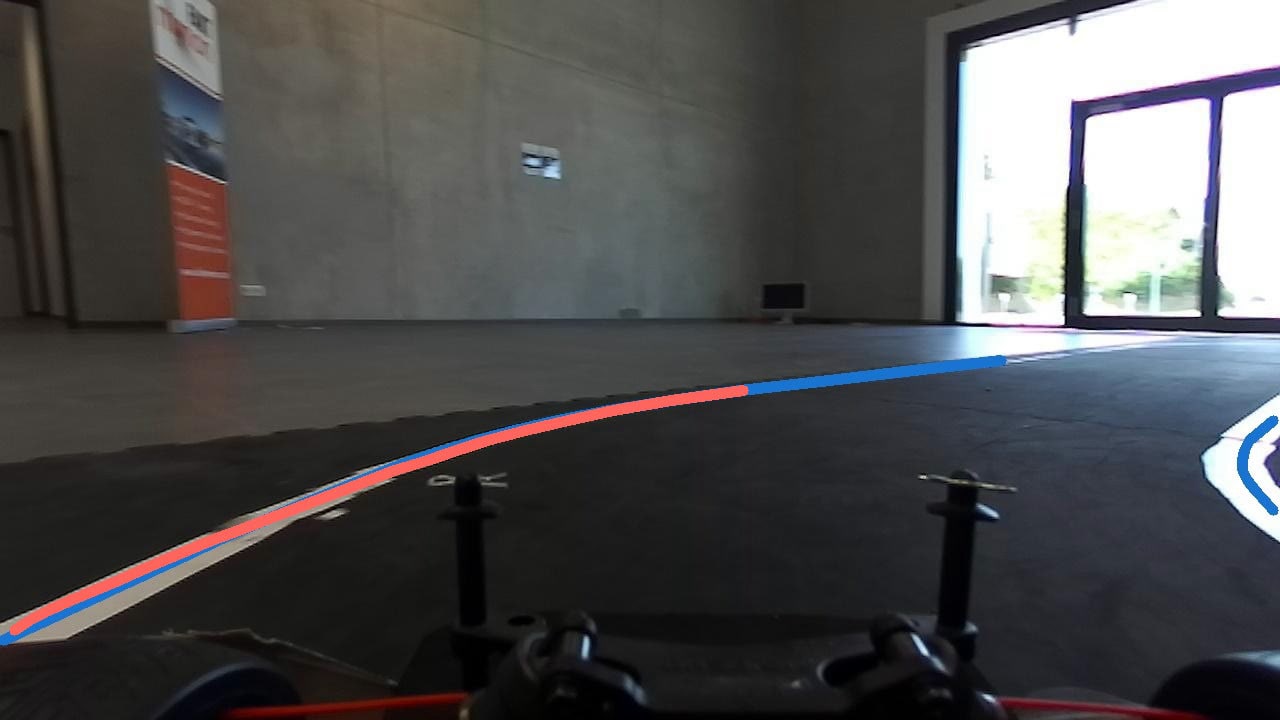} & 
			\includegraphics[width=0.18\linewidth,valign=m]{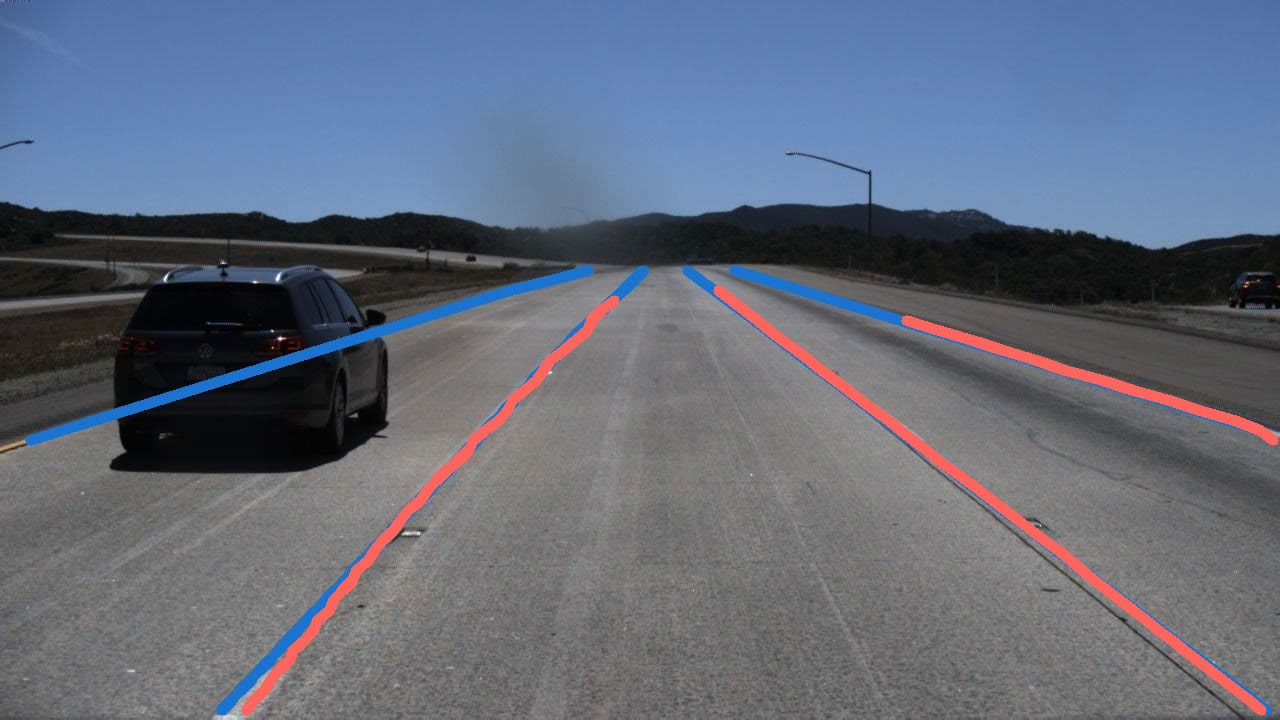} &
			\includegraphics[width=0.18\linewidth,valign=m]{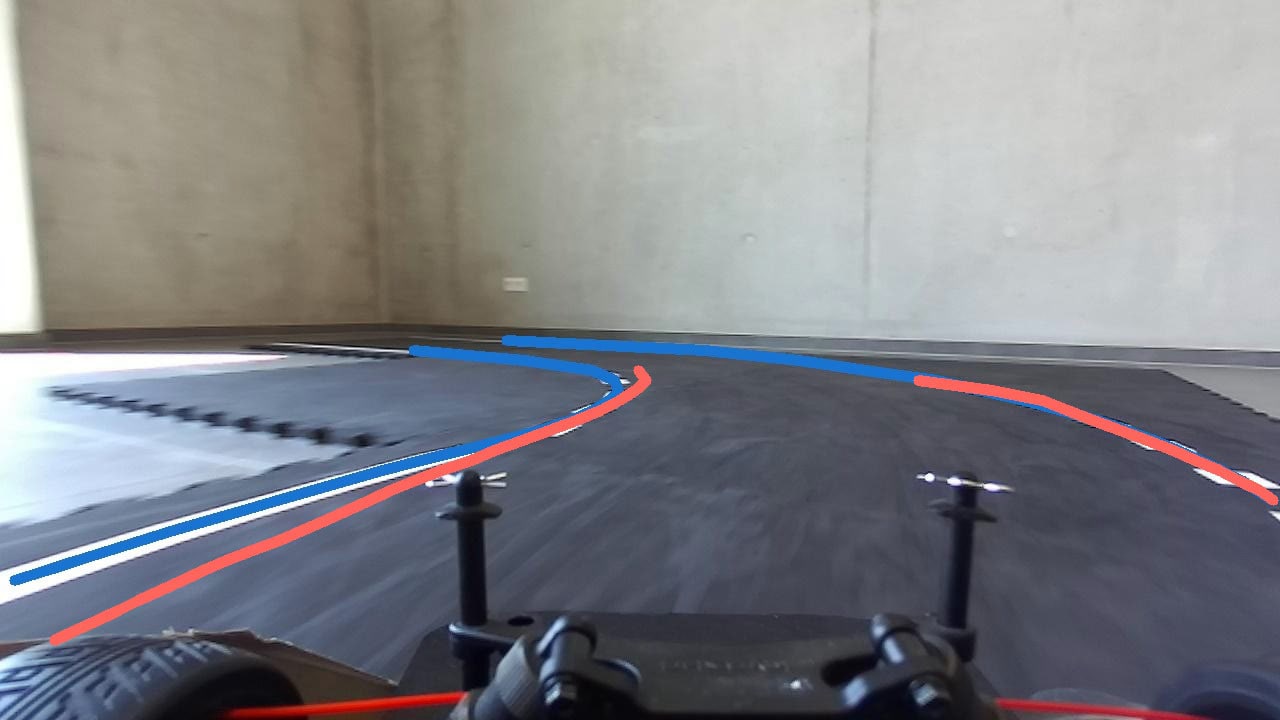} & \includegraphics[width=0.18\linewidth,valign=m]{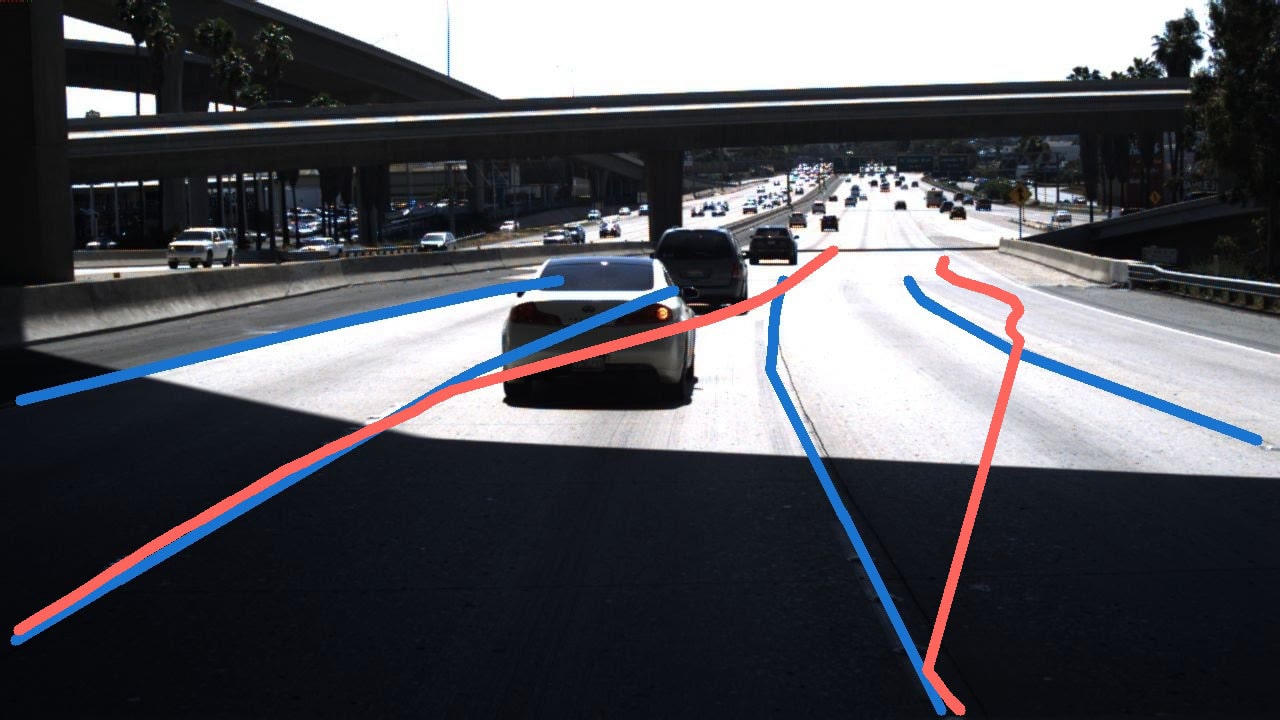}\\
			SGPCS & 
			\includegraphics[width=0.18\linewidth,valign=m]{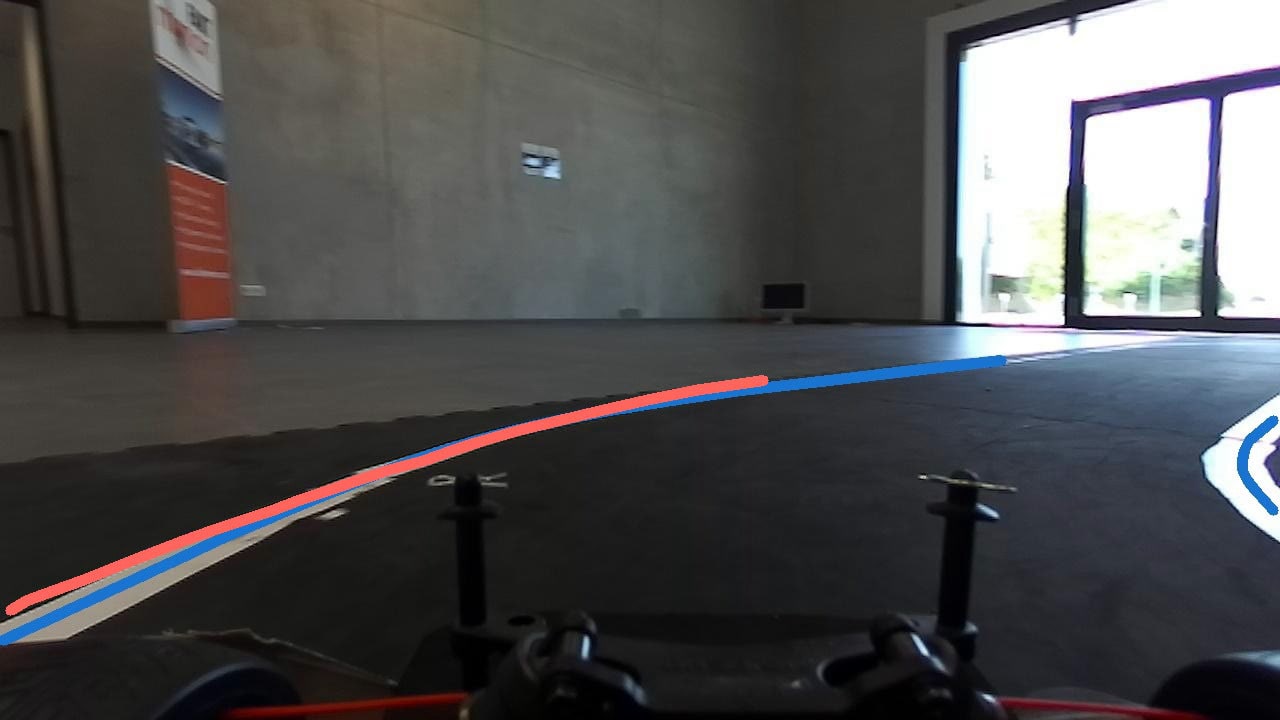} & \includegraphics[width=0.18\linewidth,valign=m]{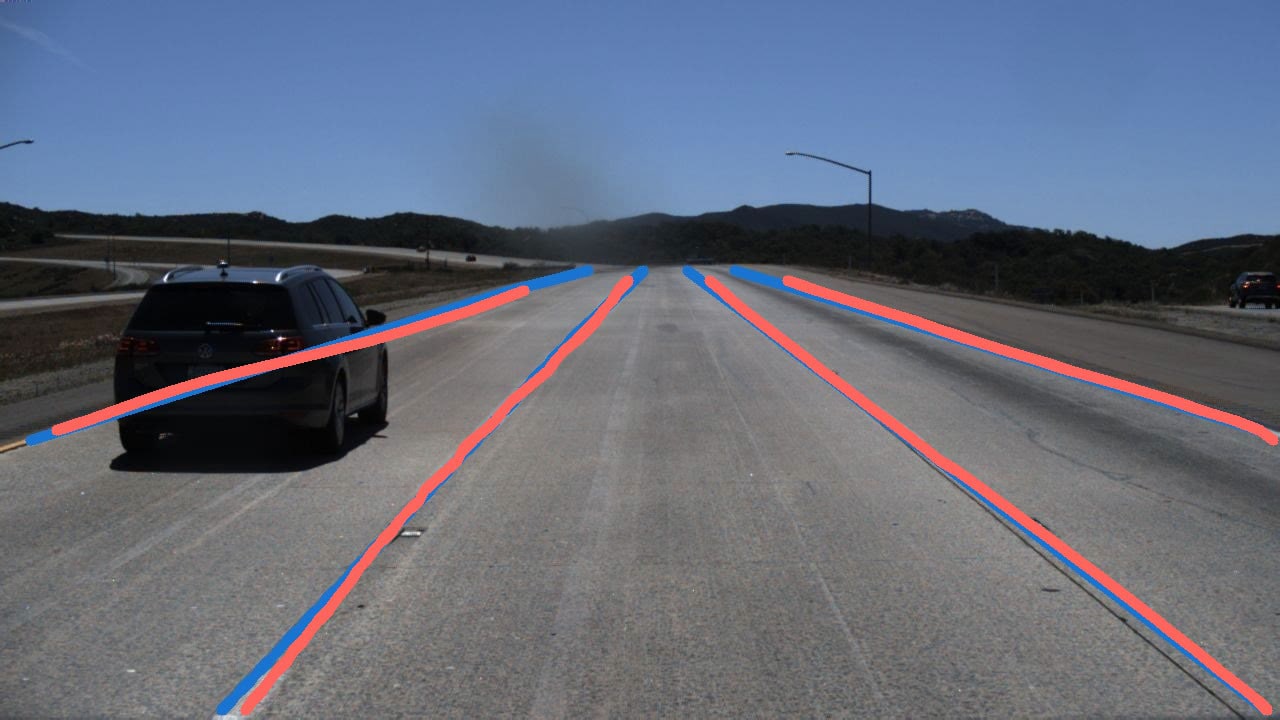} &
			\includegraphics[width=0.18\linewidth,valign=m]{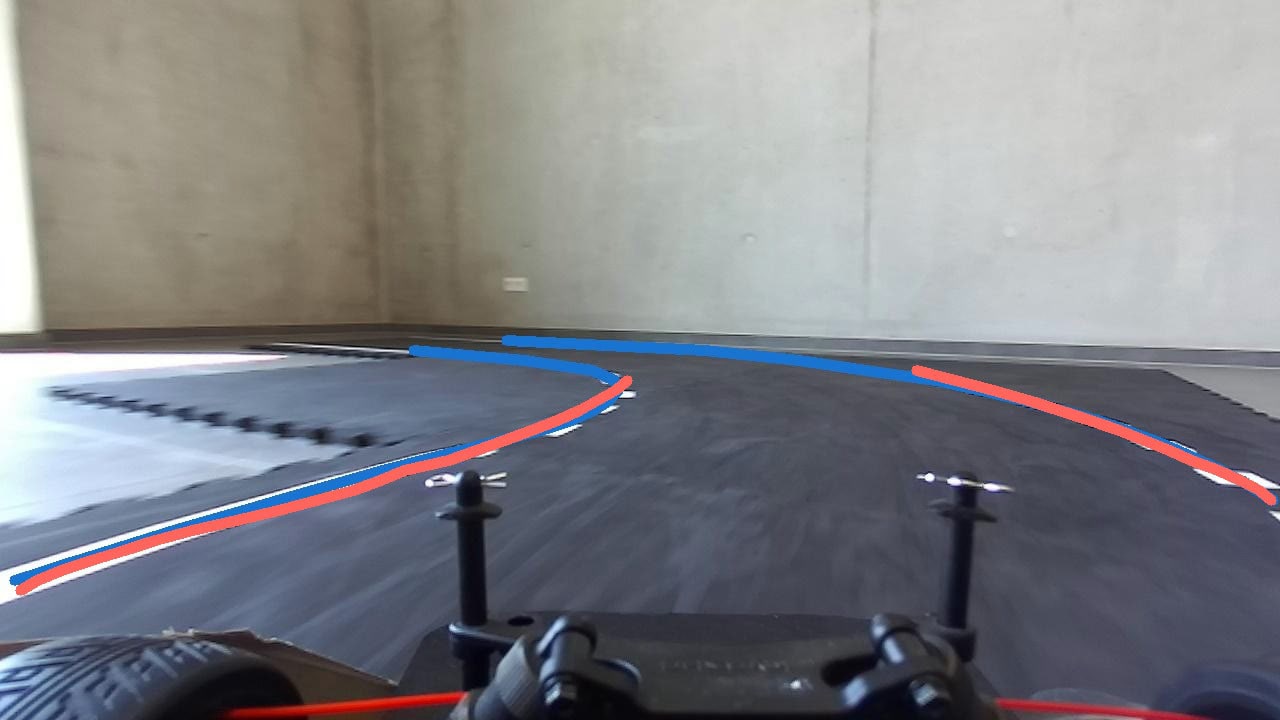} & \includegraphics[width=0.18\linewidth,valign=m]{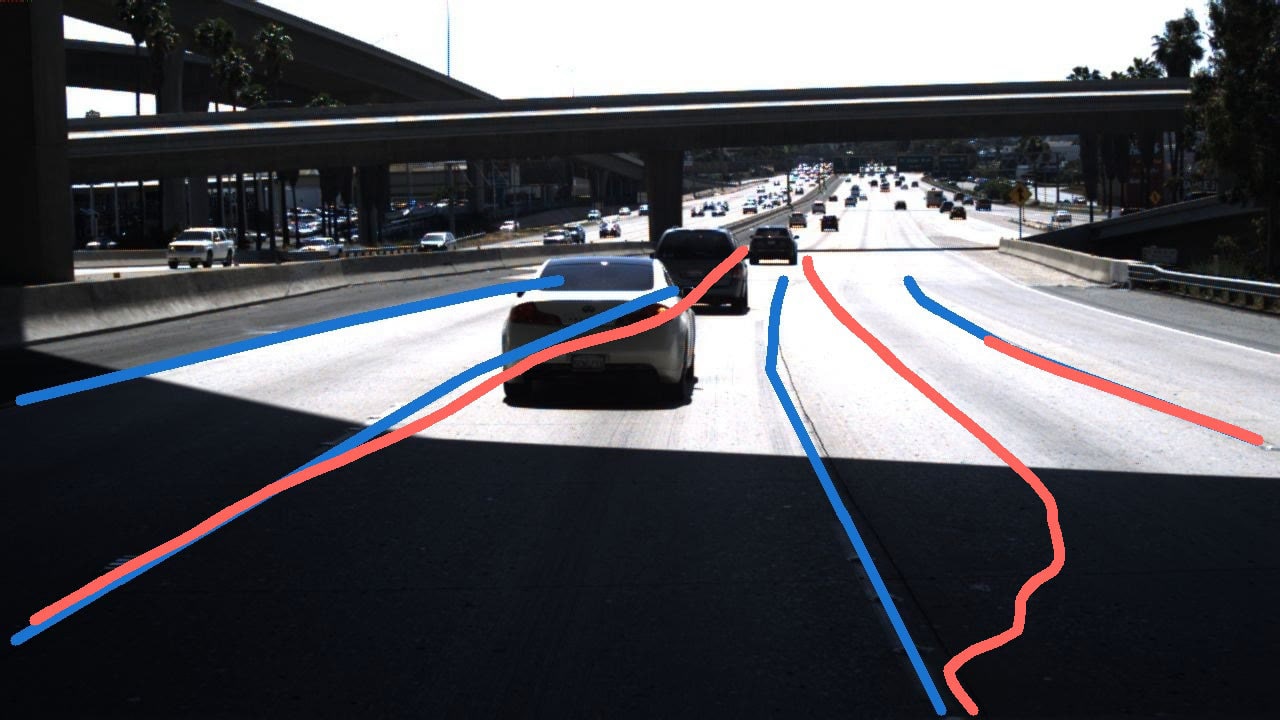}\\
			UFLD-TO & 
			\includegraphics[width=0.18\linewidth,valign=m]{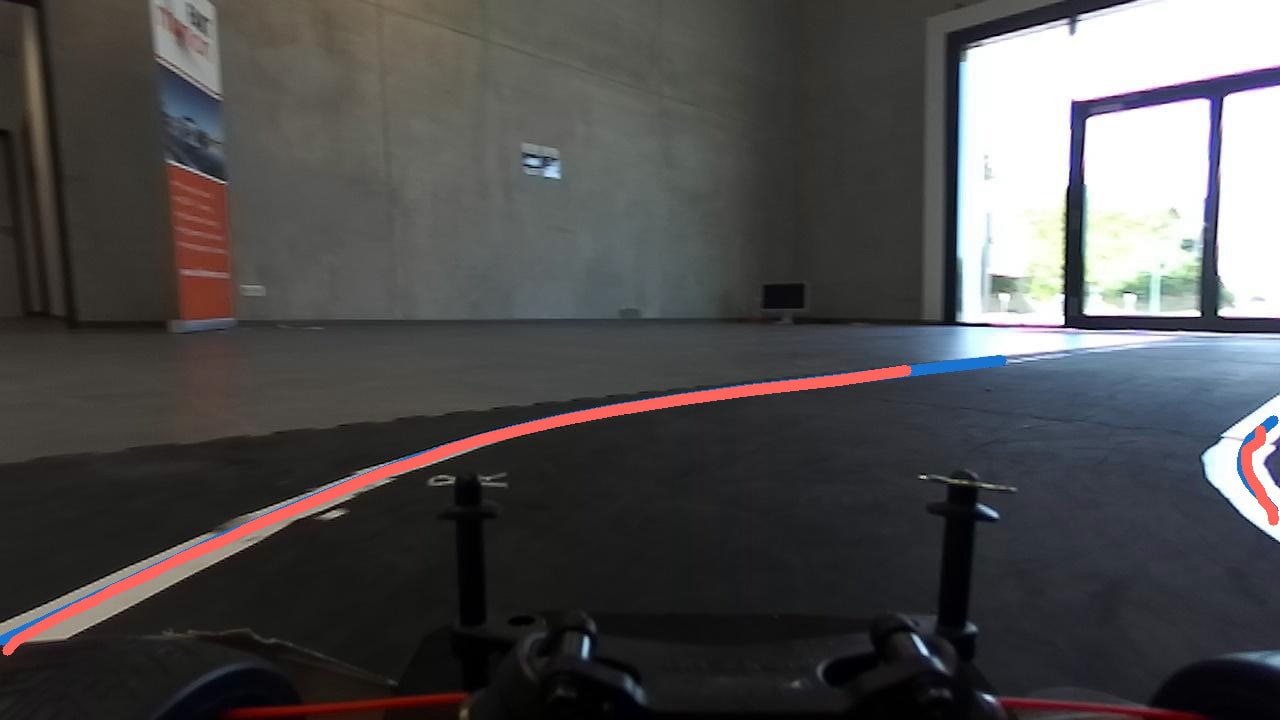} & \includegraphics[width=0.18\linewidth,valign=m]{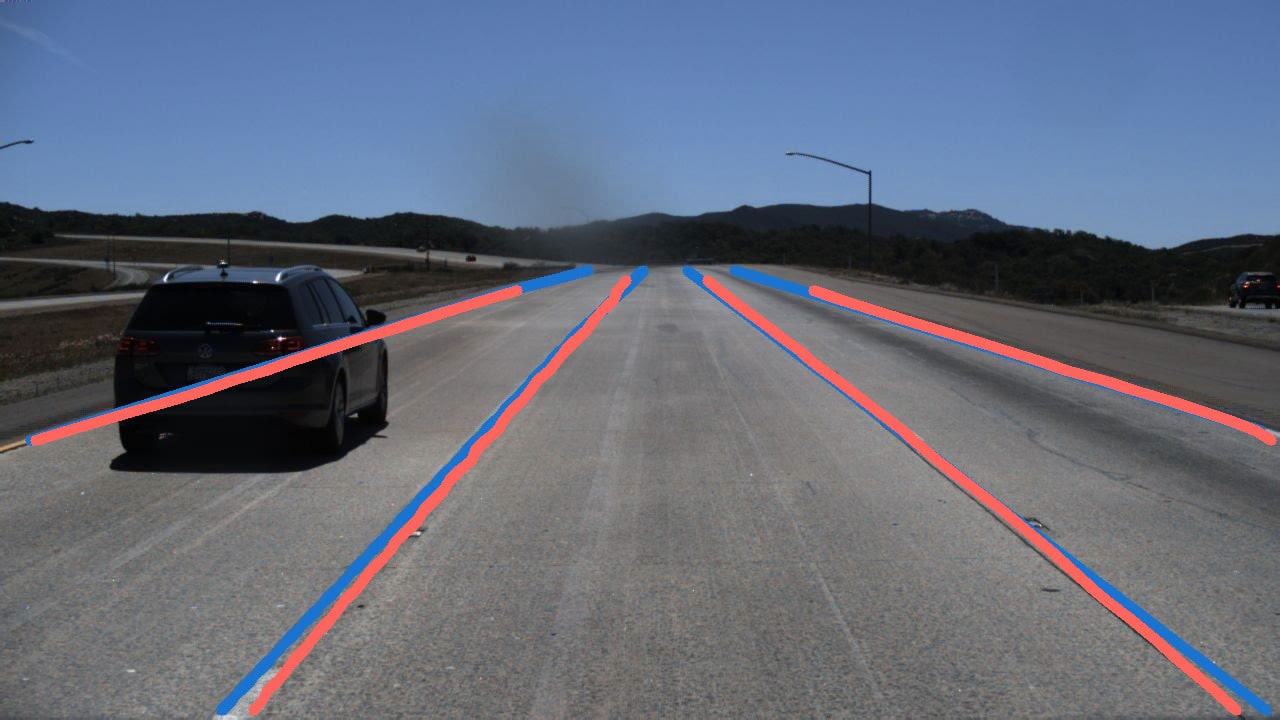} &
			\includegraphics[width=0.18\linewidth,valign=m]{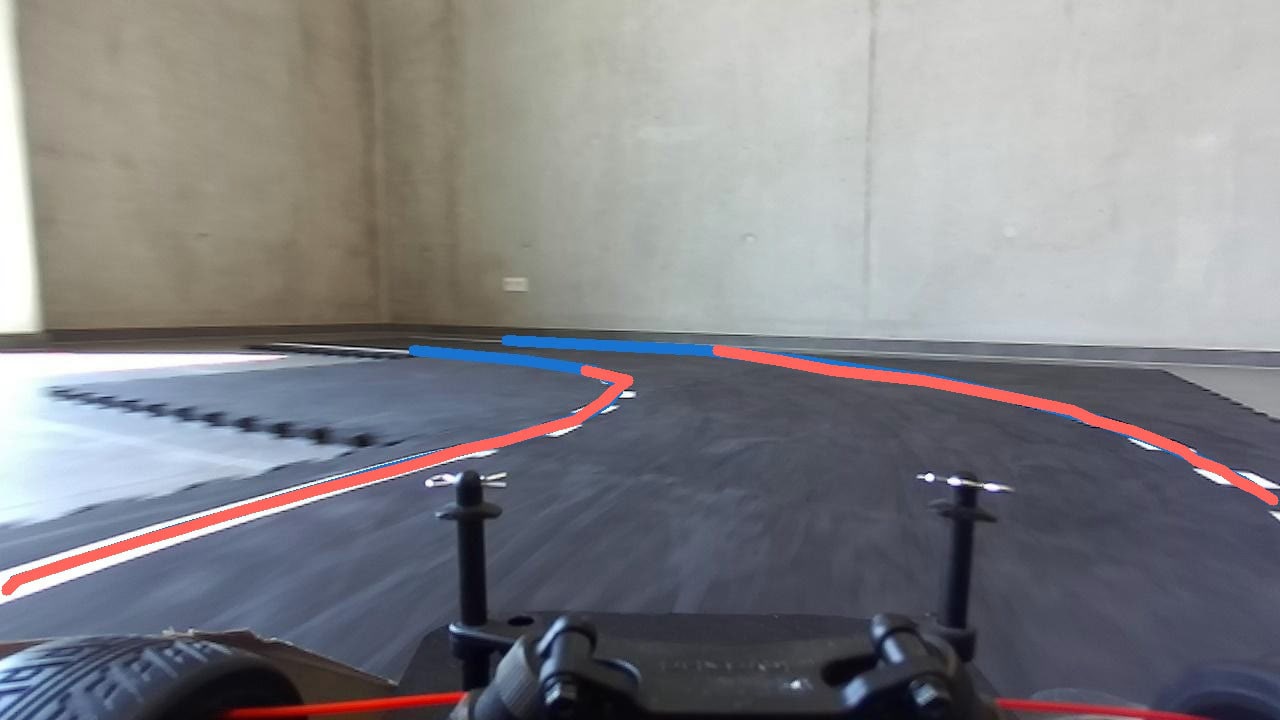} & \includegraphics[width=0.18\linewidth,valign=m]{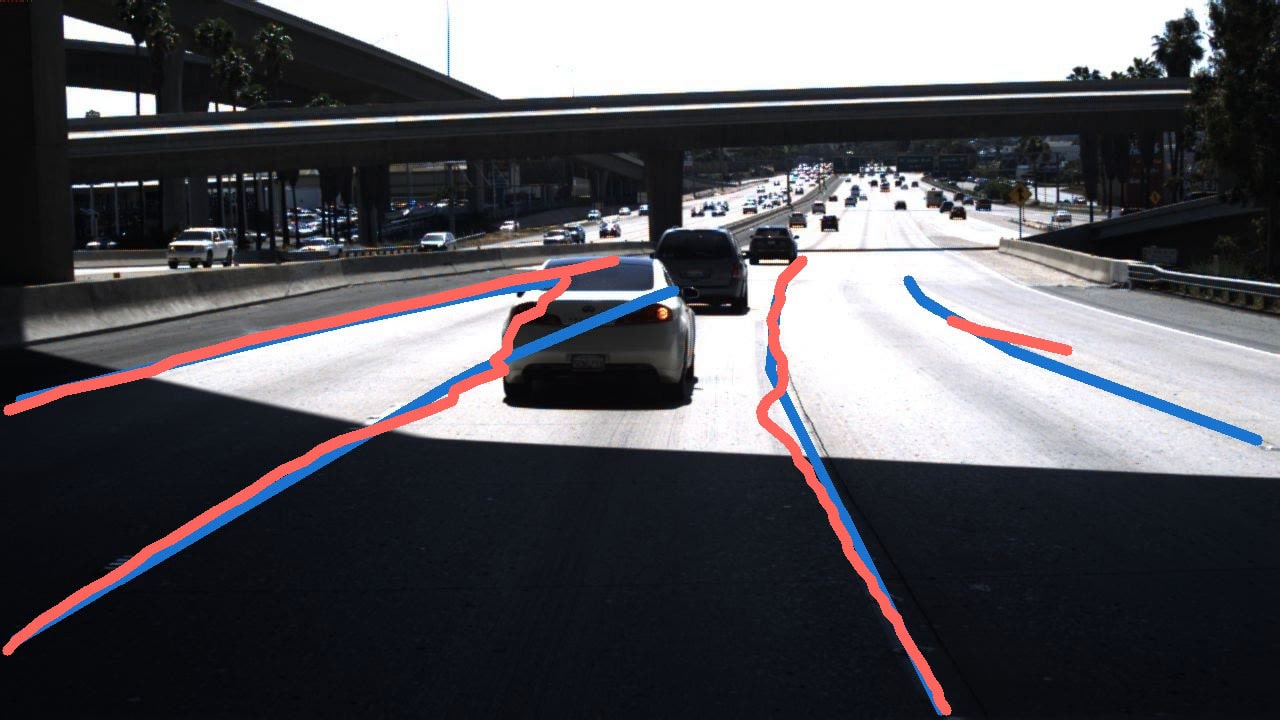}\\
		\end{tabular}
	\end{center}
	\vspace{-1ex}
	\caption[More qualitative results of target domain predictions.]{More qualitative results of target domain predictions. Images are randomly sampled. Ground truth lane annotations are marked in blue, and predictions in red. Best viewed in color.}
	\label{fig:carlane:app:inference_samples_1}
\end{figure}

\begin{figure}
	\small
	\begin{center}
		\begin{tabular}{rc@{}c@{}c@{}c}
			~ & MoLane & TuLane & \multicolumn{2}{c}{MuLane} \\
			UFLD-SO & 
			\includegraphics[width=0.18\linewidth,valign=m]{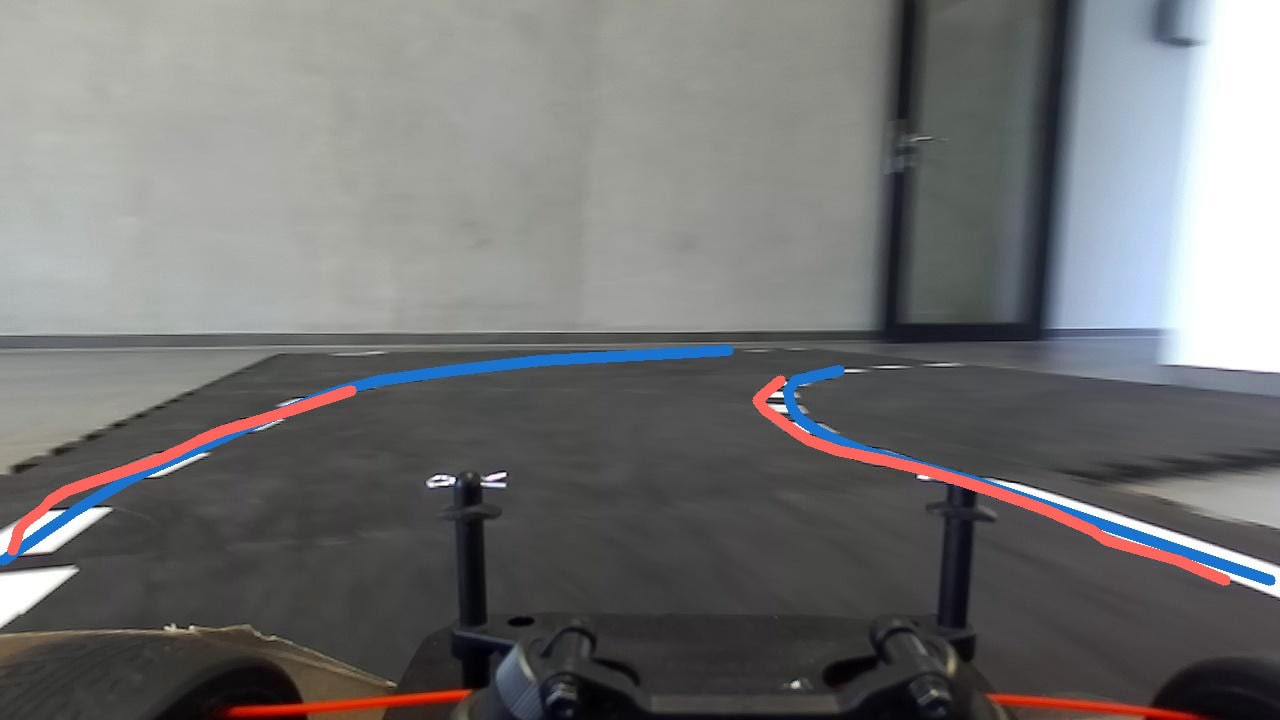} & \includegraphics[width=0.18\linewidth,valign=m]{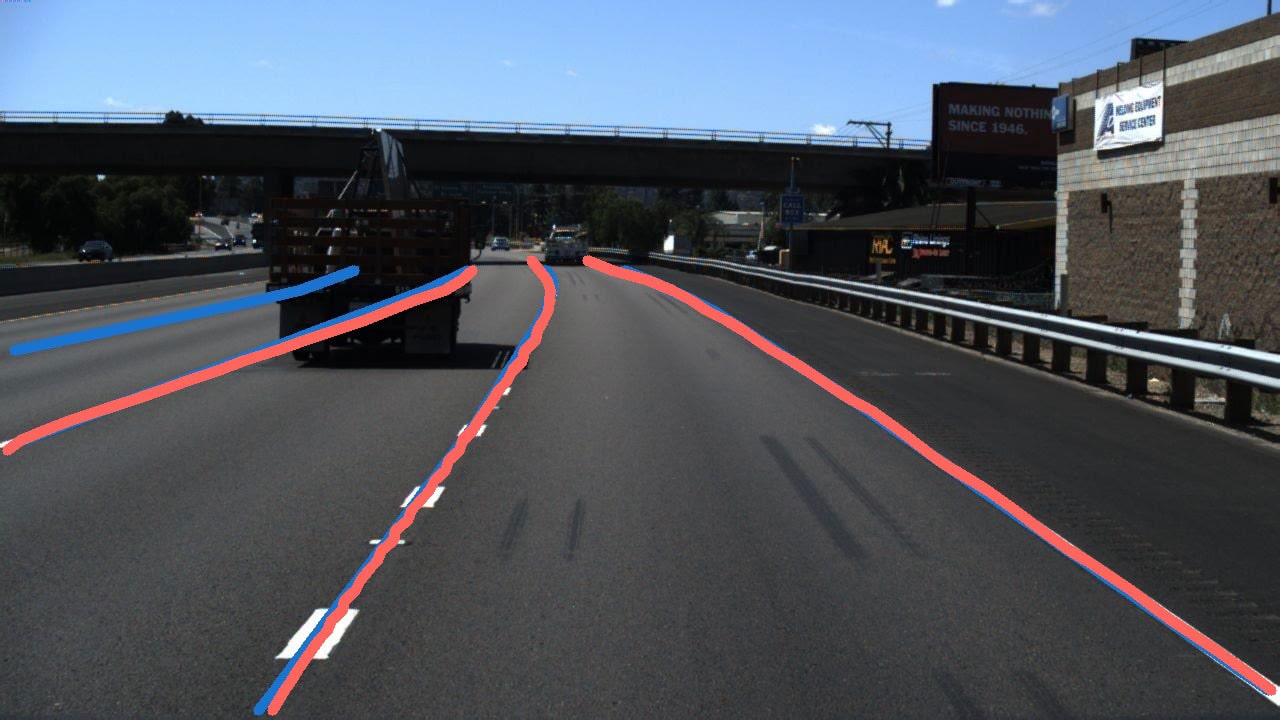} &
			\includegraphics[width=0.18\linewidth,valign=m]{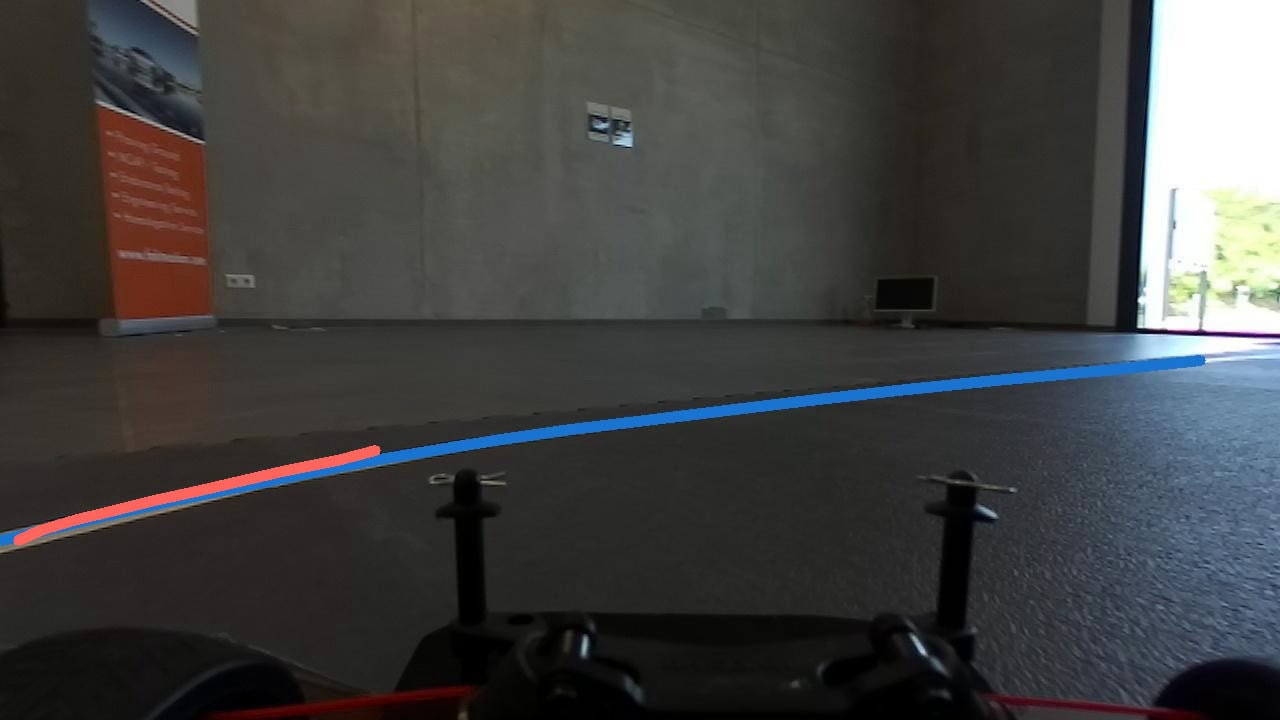} & \includegraphics[width=0.18\linewidth,valign=m]{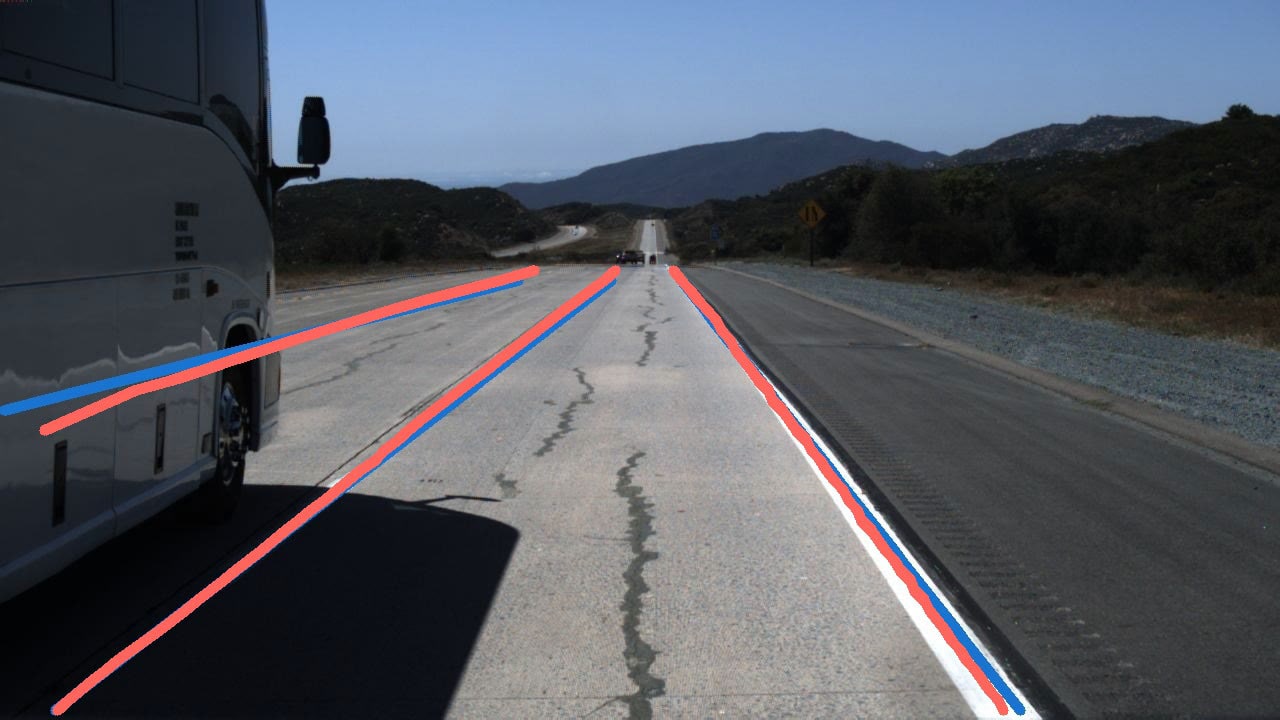}\\
			DANN & 
			\includegraphics[width=0.18\linewidth,valign=m]{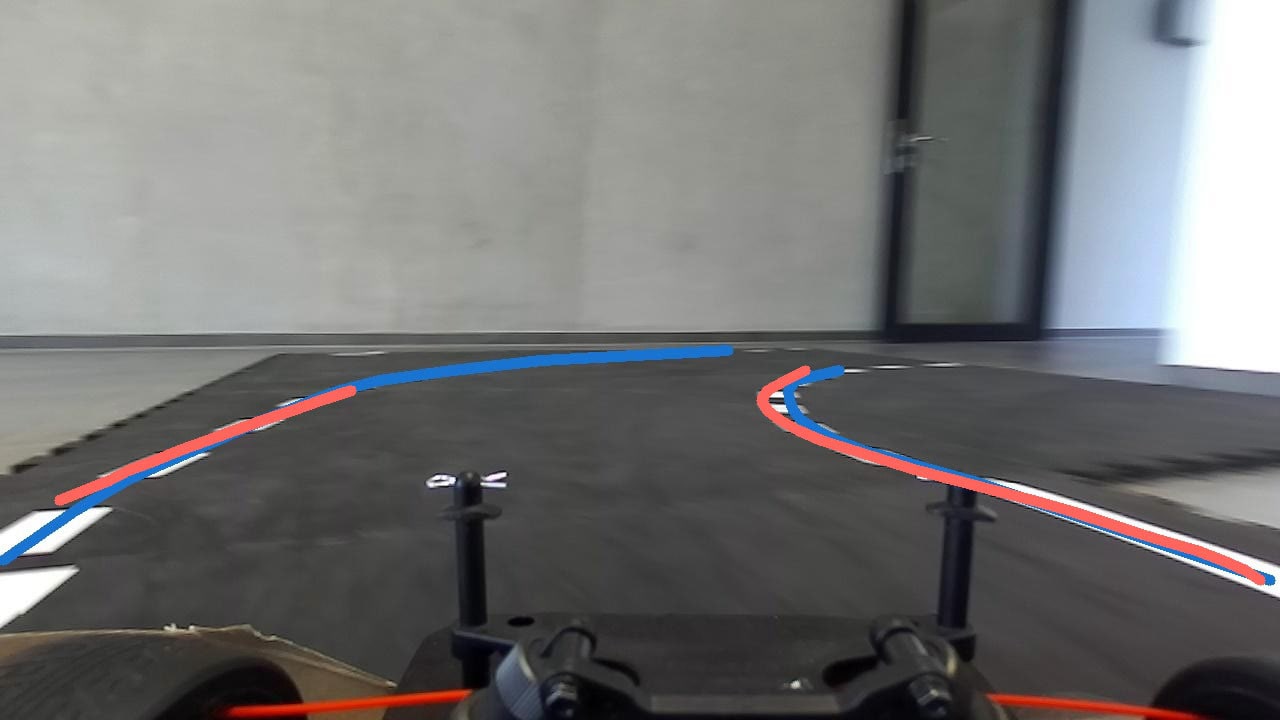} & 
			\includegraphics[width=0.18\linewidth,valign=m]{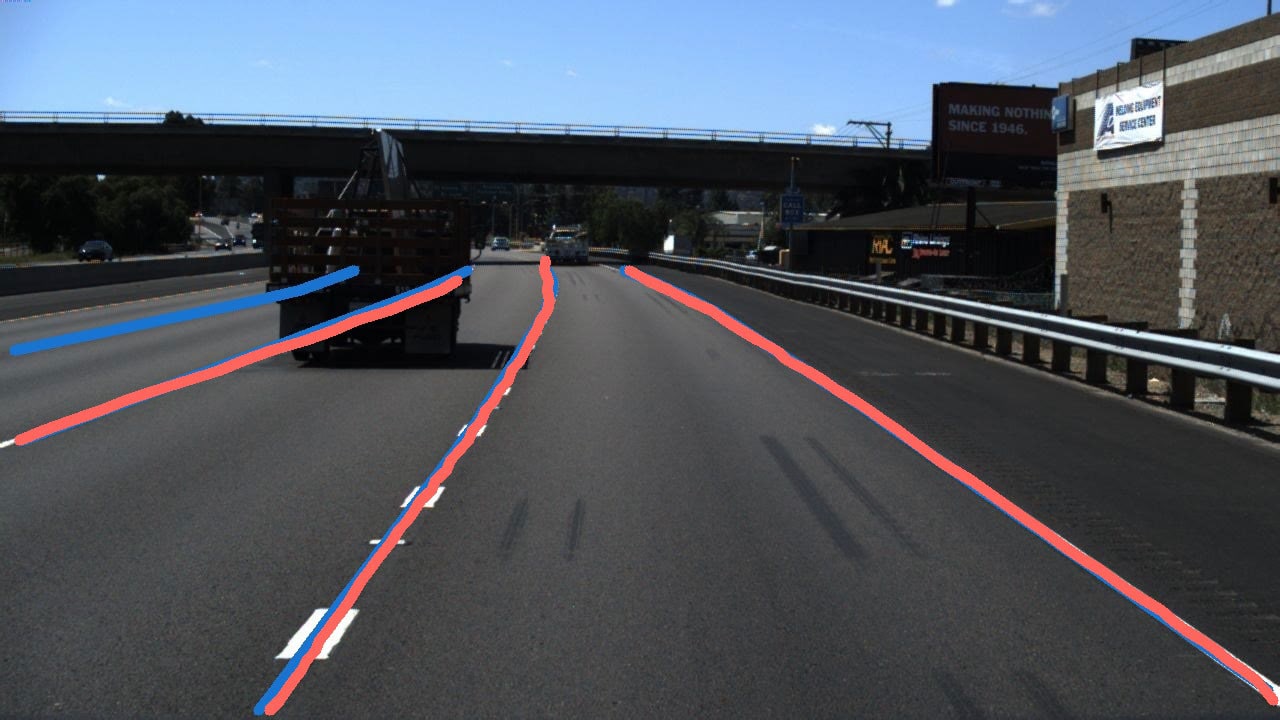} &
			\includegraphics[width=0.18\linewidth,valign=m]{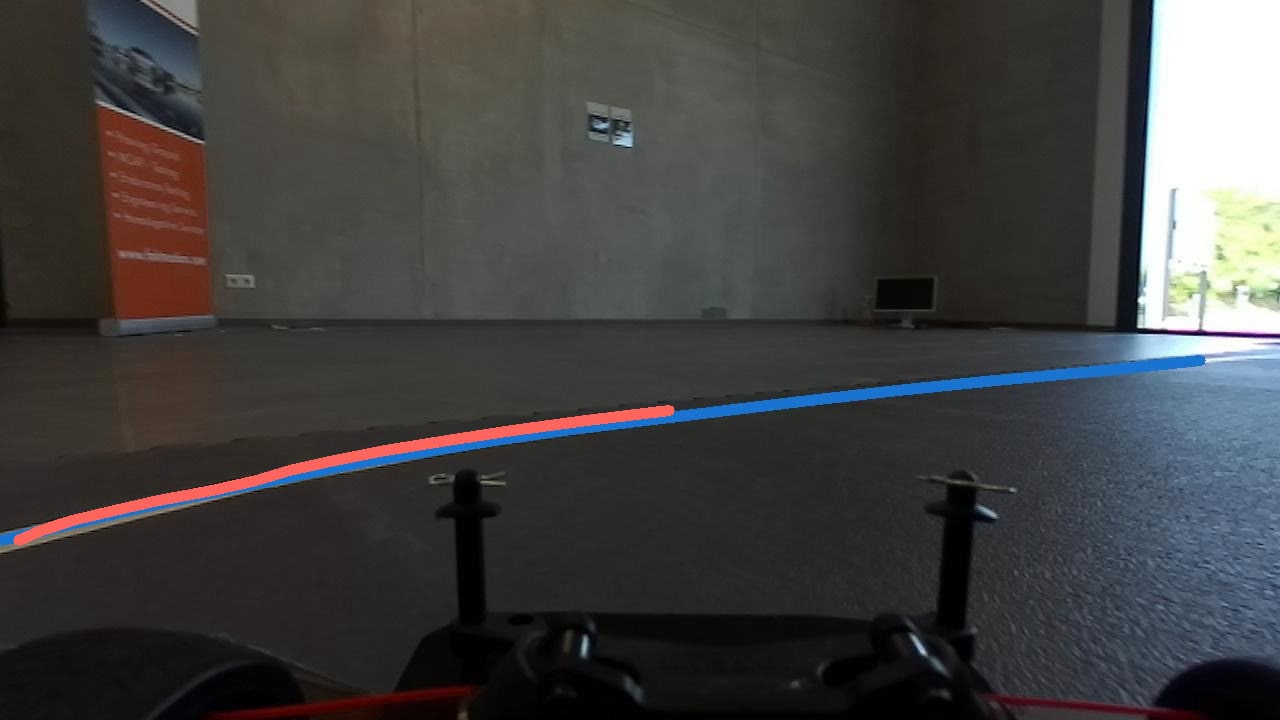} & \includegraphics[width=0.18\linewidth,valign=m]{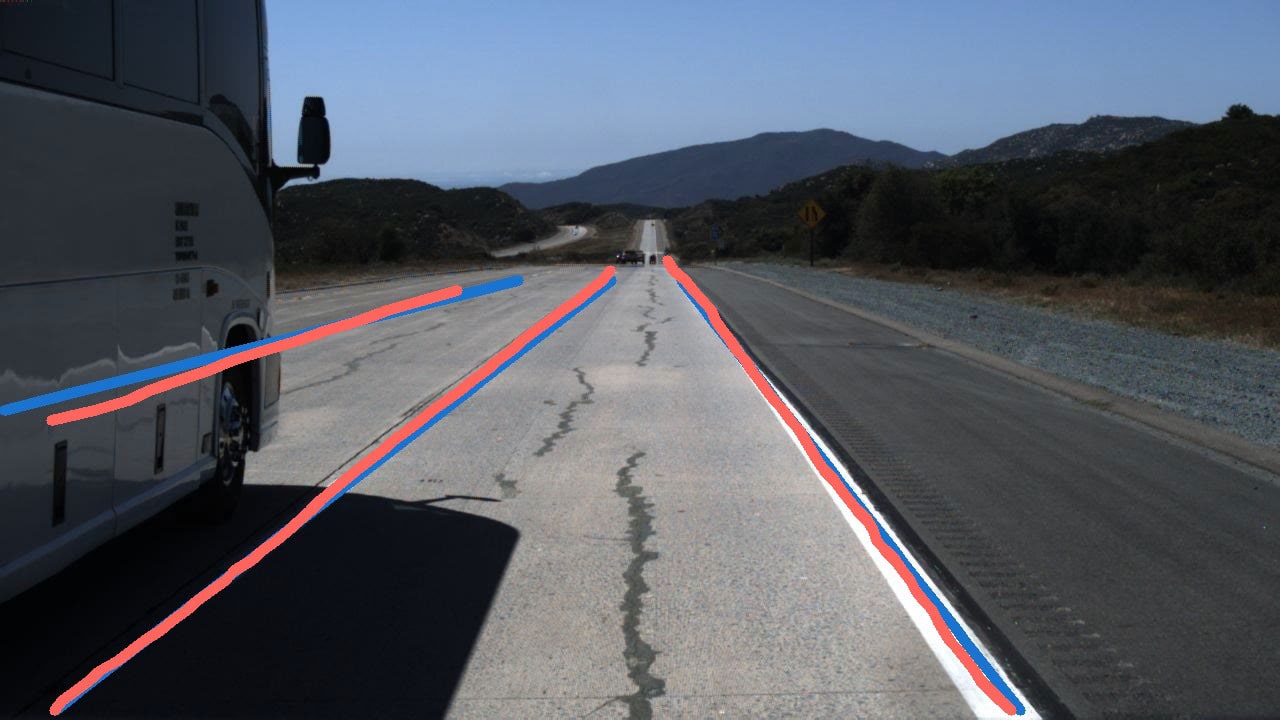}\\
			ADD & 
			\includegraphics[width=0.18\linewidth,valign=m]{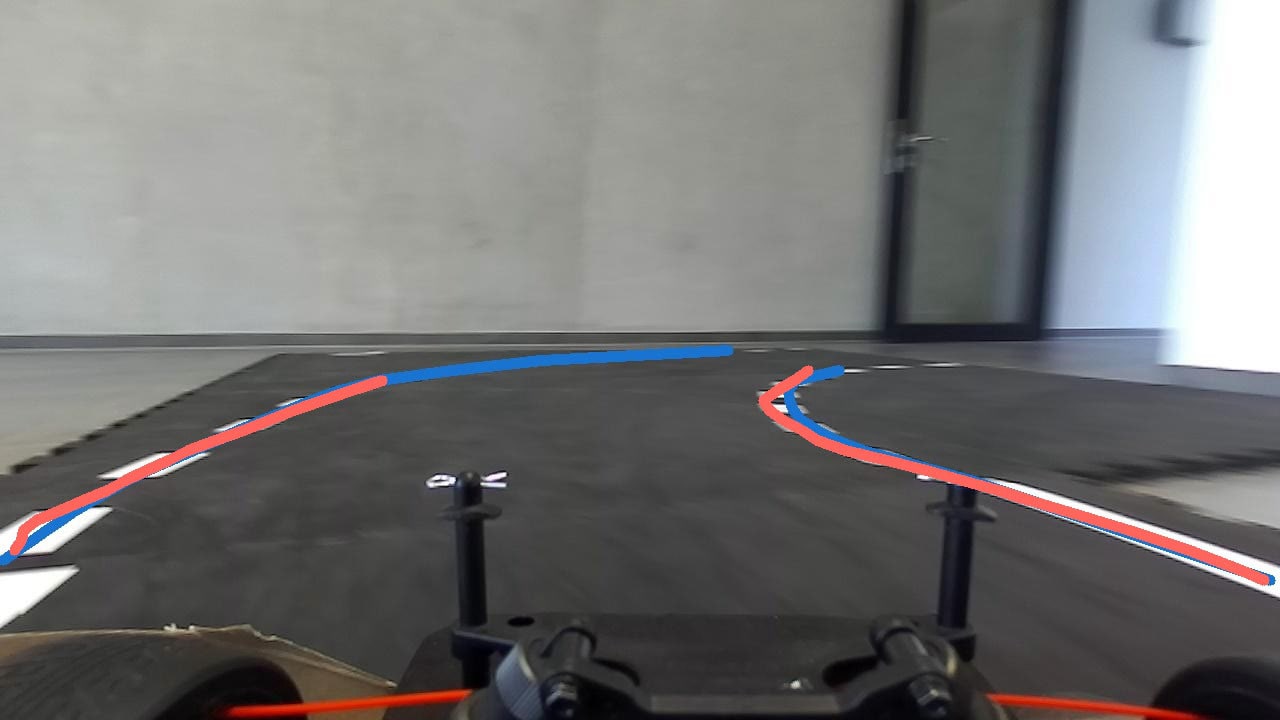} & 
			\includegraphics[width=0.18\linewidth,valign=m]{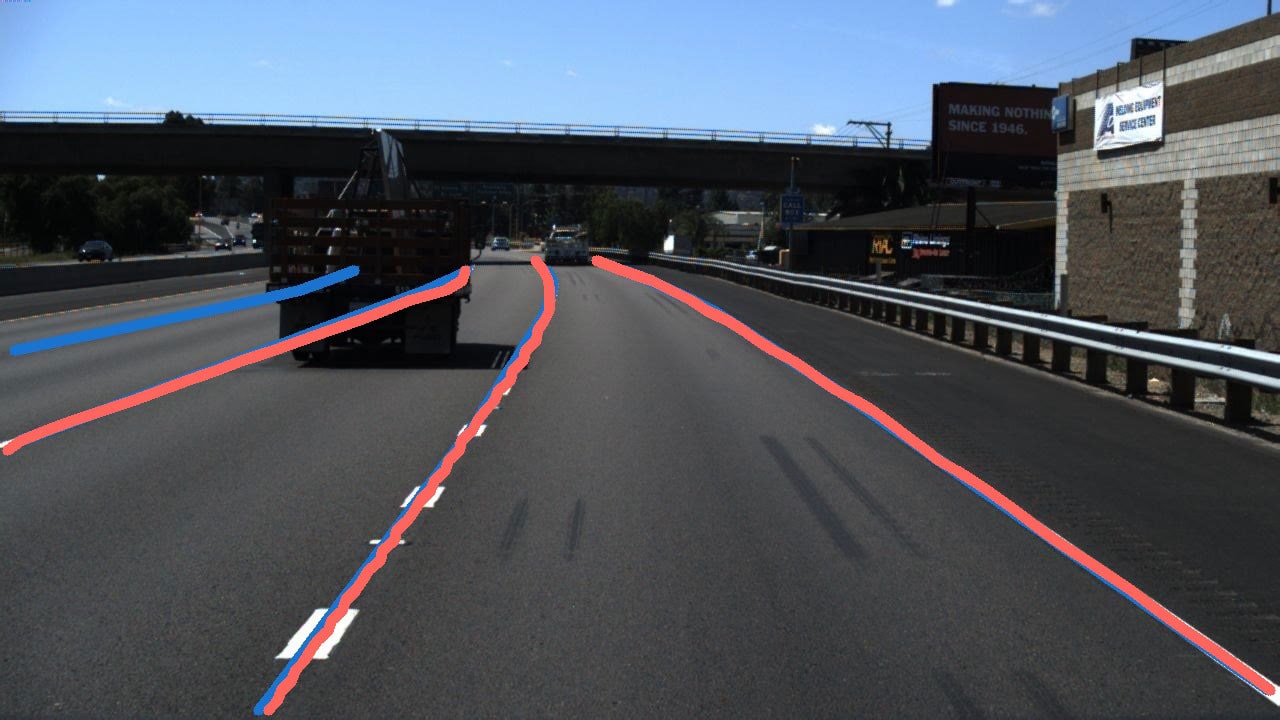} &
			\includegraphics[width=0.18\linewidth,valign=m]{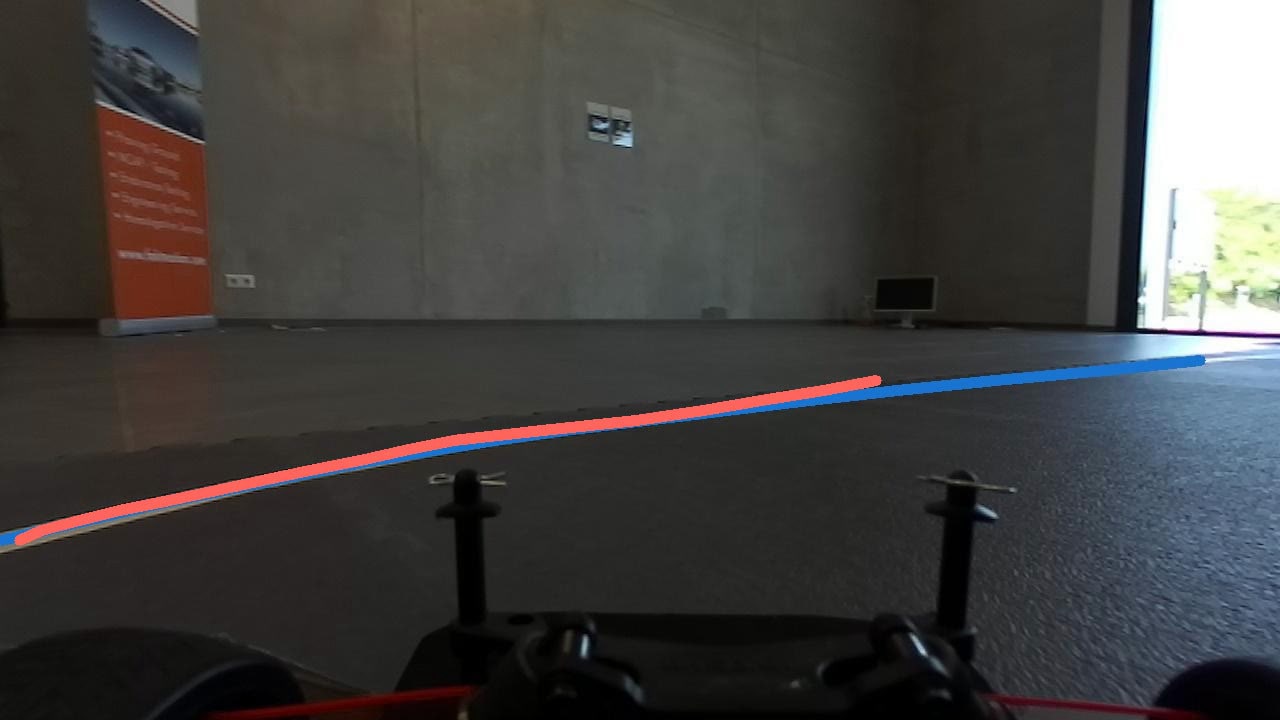} & \includegraphics[width=0.18\linewidth,valign=m]{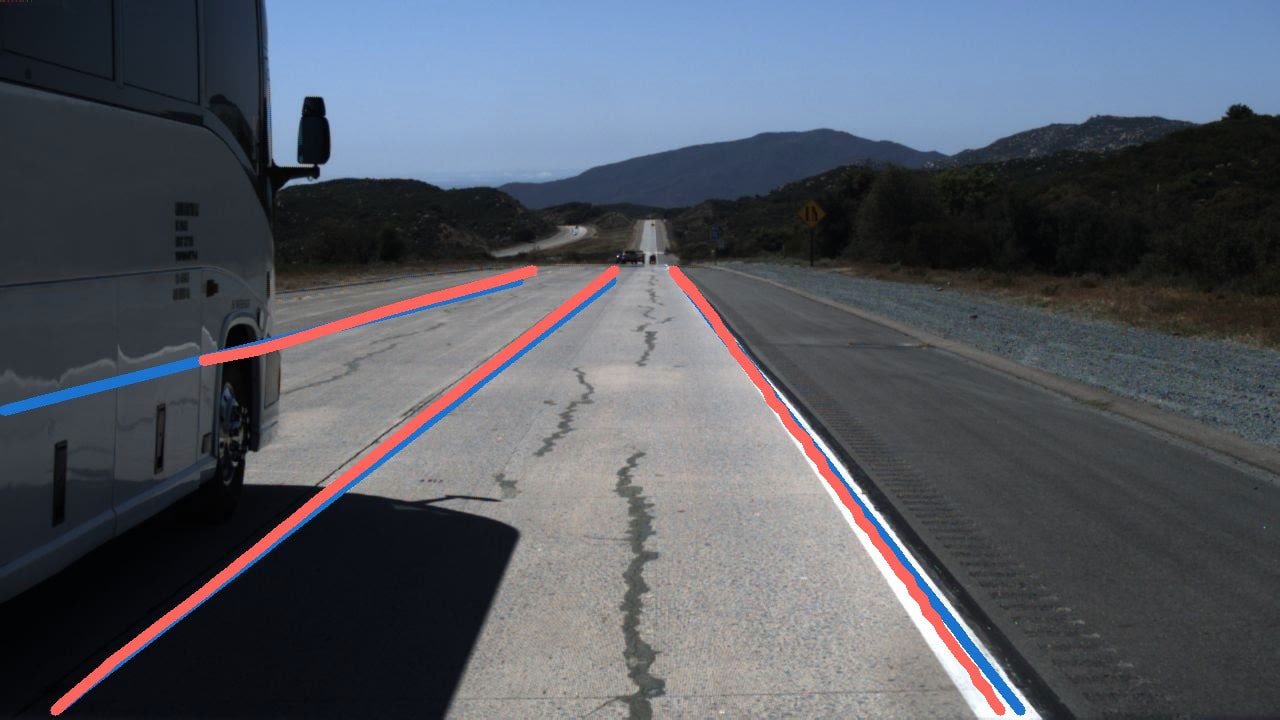}\\
			SGADA & 
			\includegraphics[width=0.18\linewidth,valign=m]{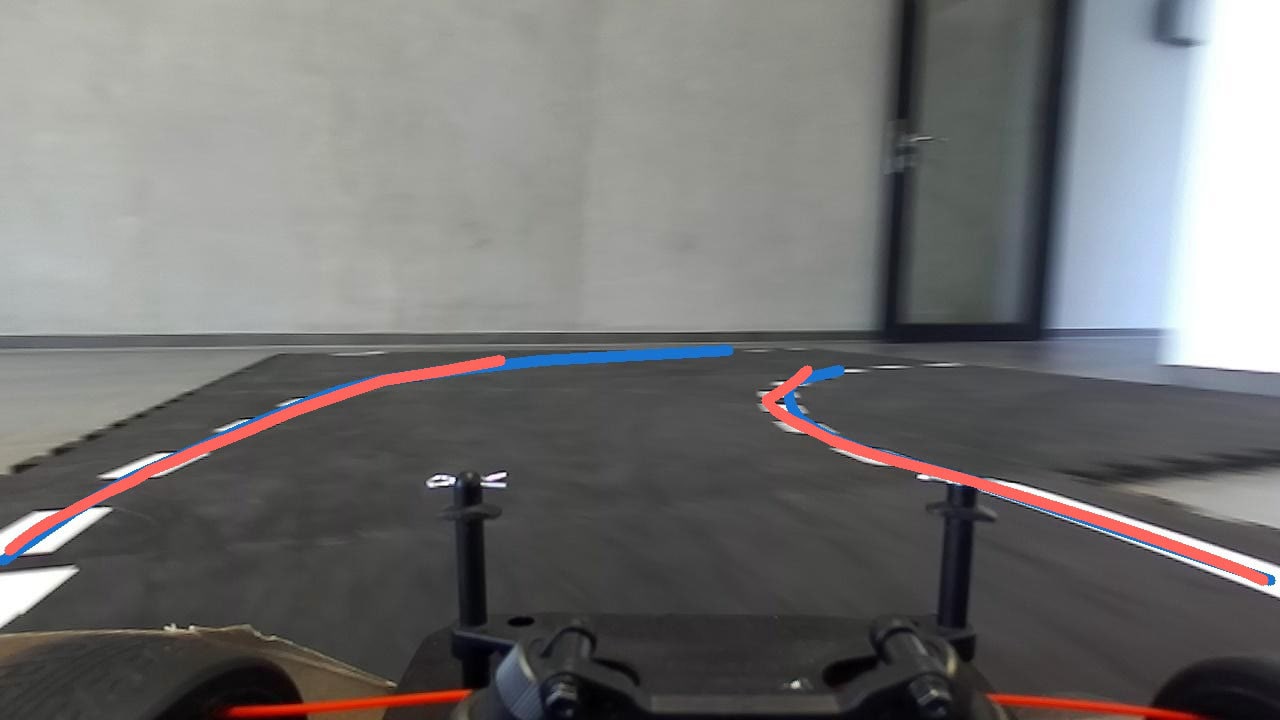} & 
			\includegraphics[width=0.18\linewidth,valign=m]{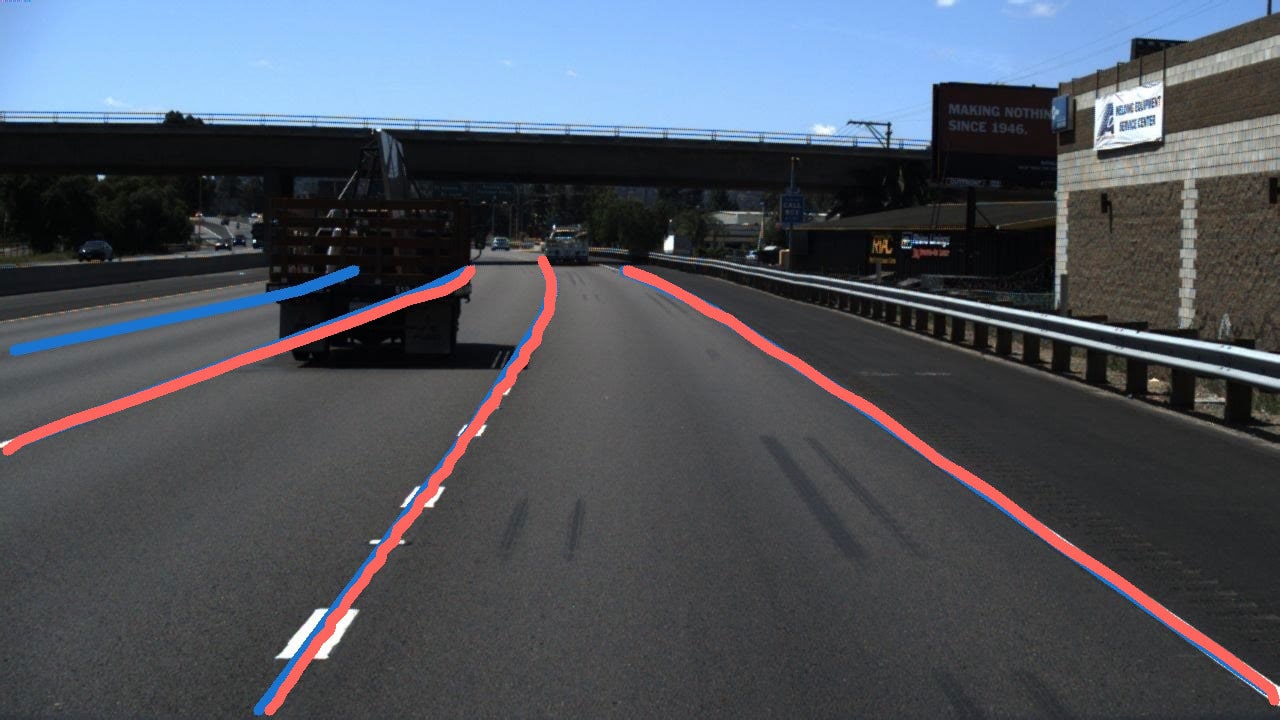} &
			\includegraphics[width=0.18\linewidth,valign=m]{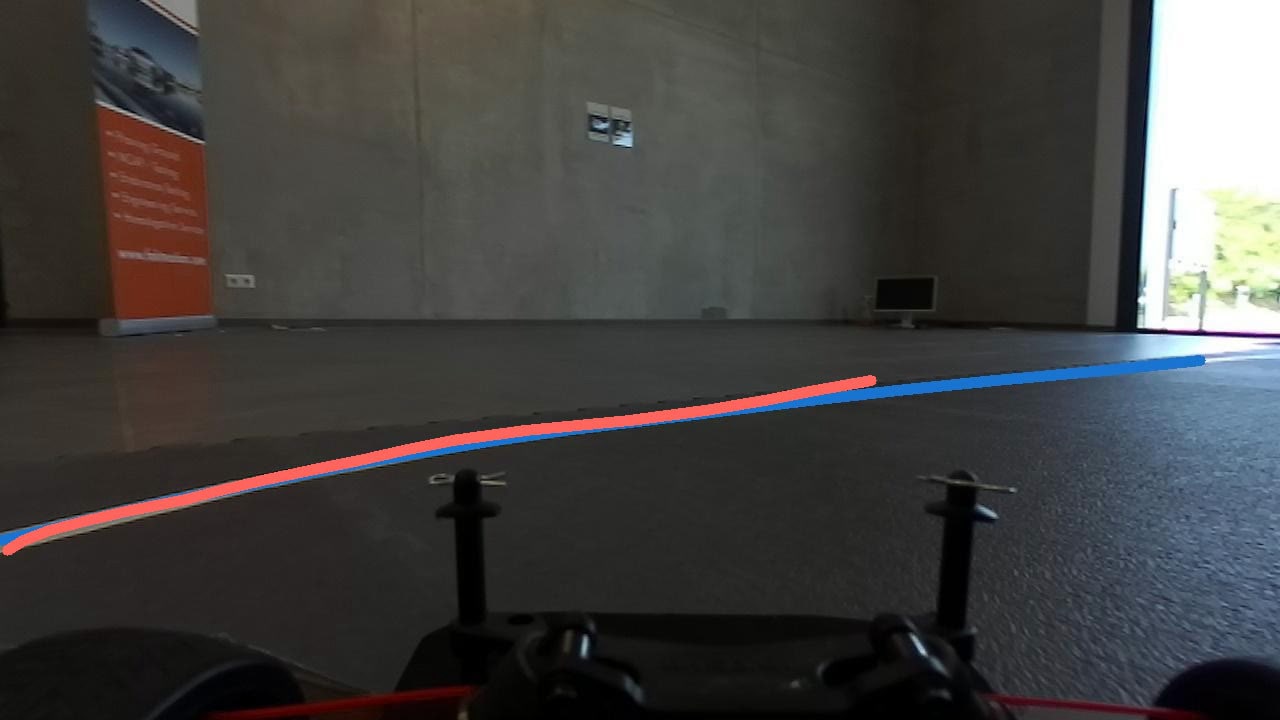} & \includegraphics[width=0.18\linewidth,valign=m]{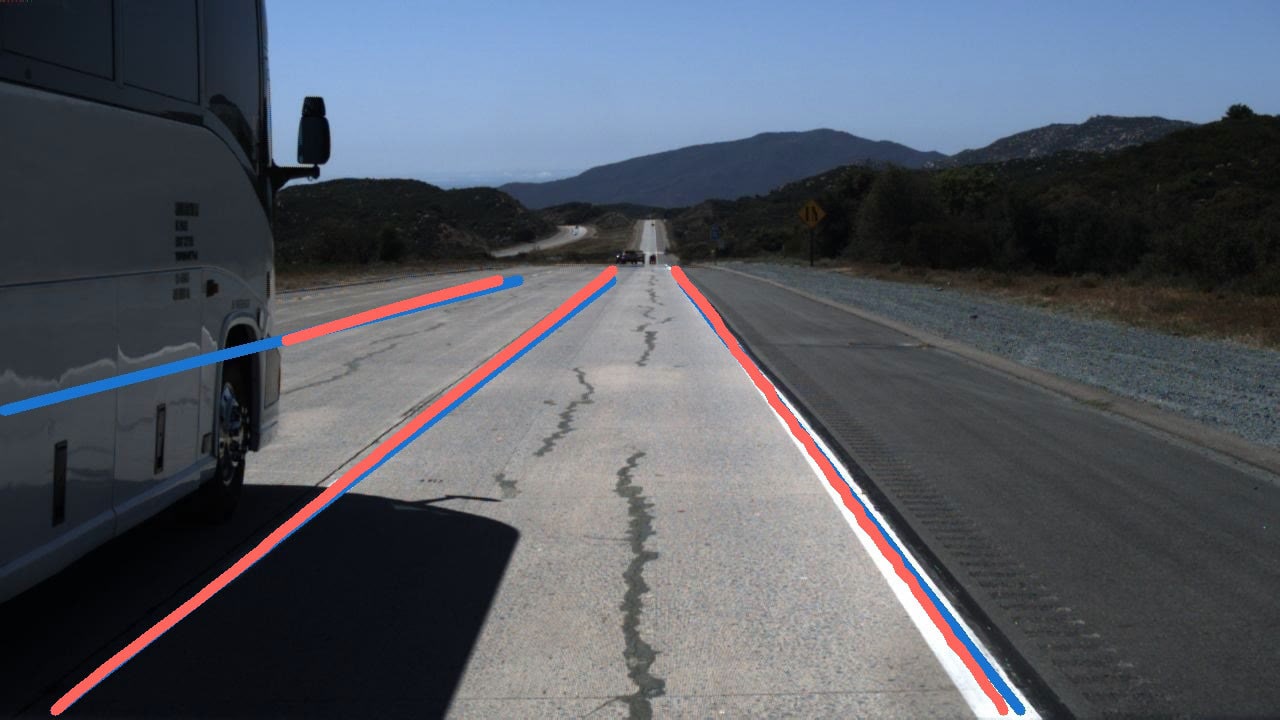}\\
			SGPCS & 
			\includegraphics[width=0.18\linewidth,valign=m]{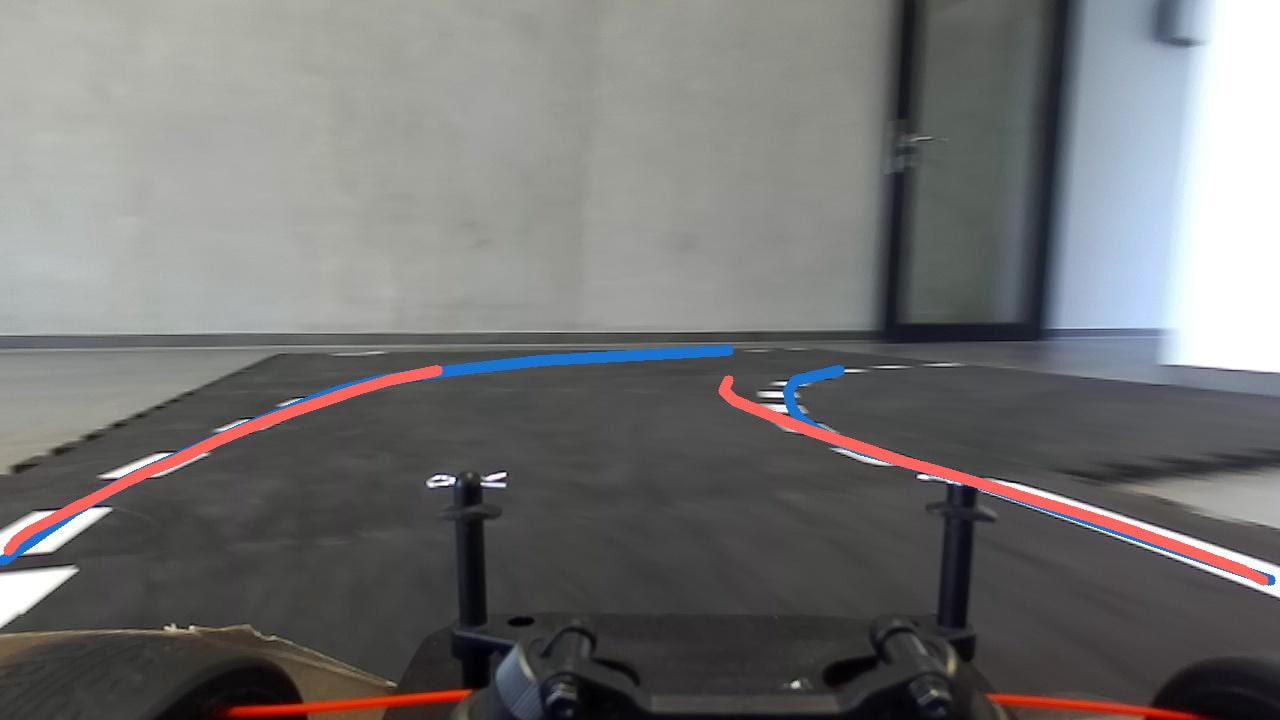} & \includegraphics[width=0.18\linewidth,valign=m]{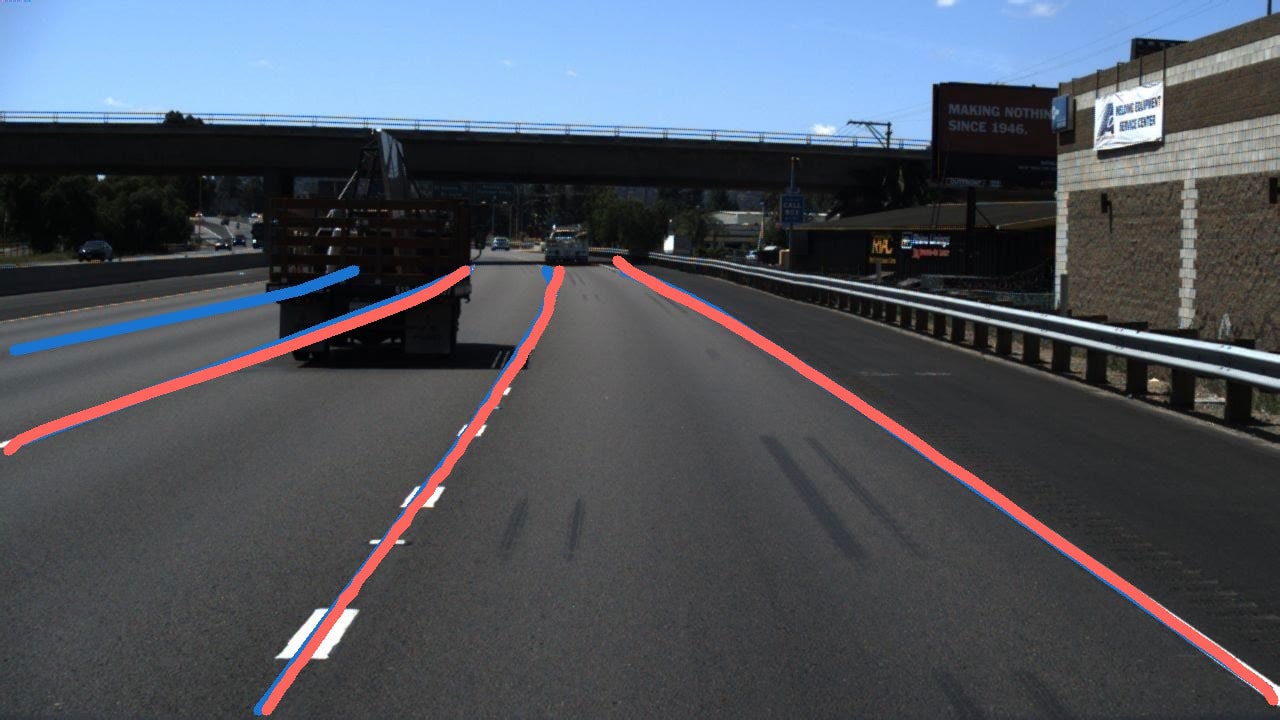} &
			\includegraphics[width=0.18\linewidth,valign=m]{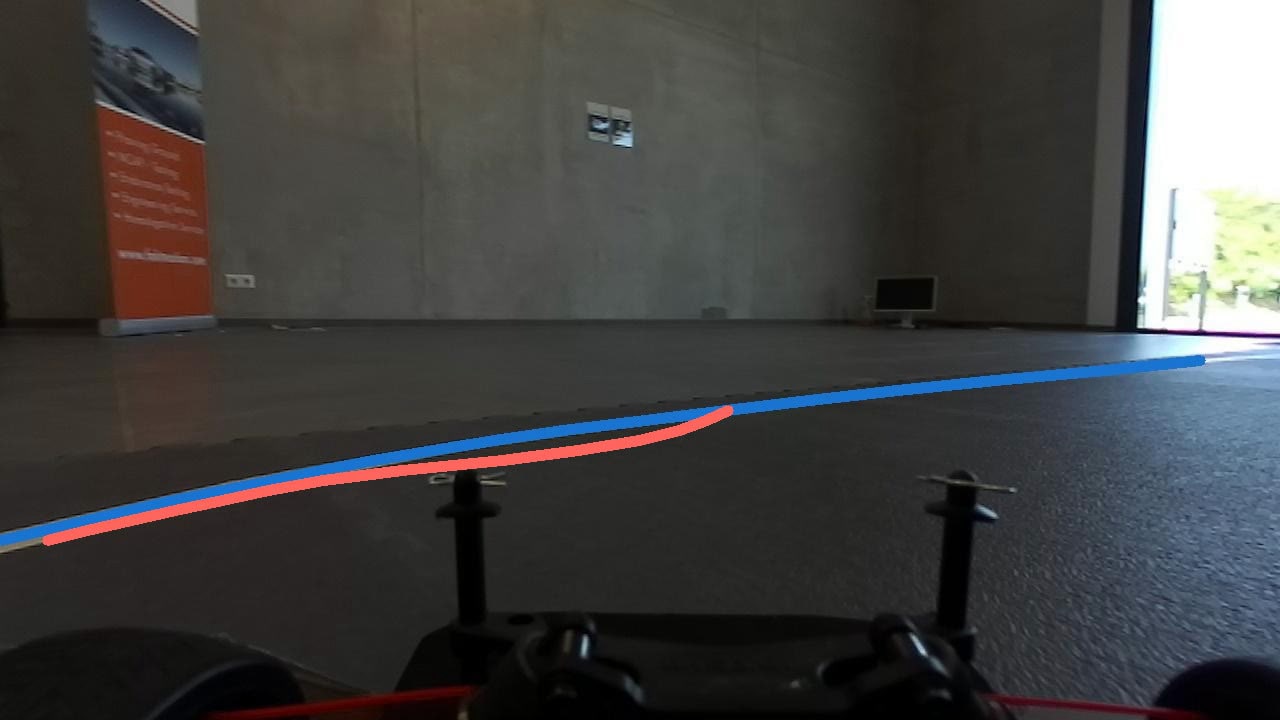} & \includegraphics[width=0.18\linewidth,valign=m]{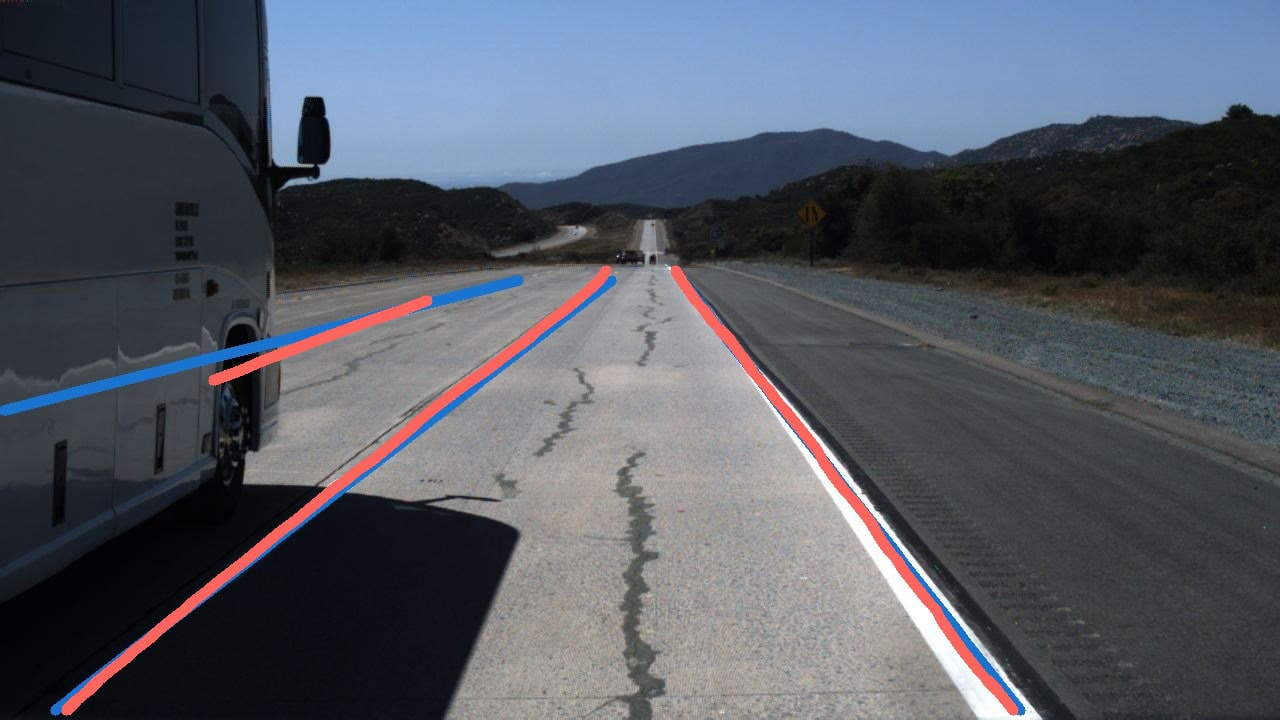}\\
			UFL & 
			\includegraphics[width=0.18\linewidth,valign=m]{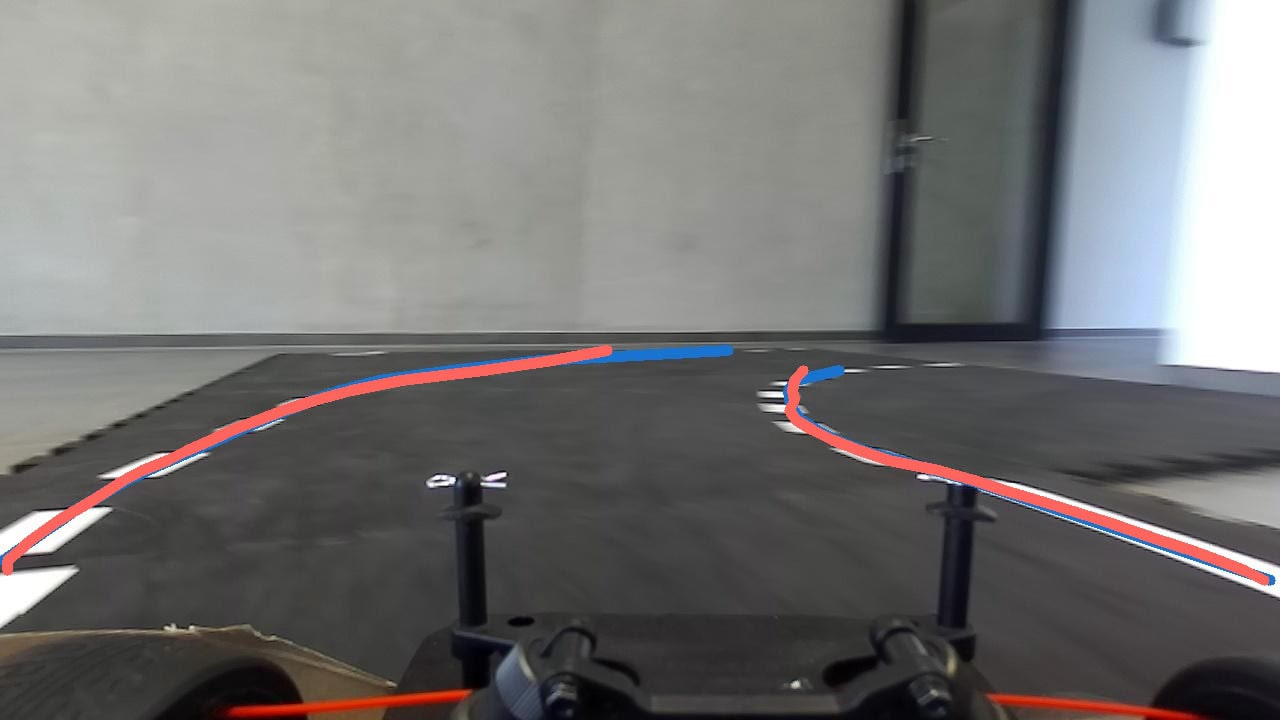} & \includegraphics[width=0.18\linewidth,valign=m]{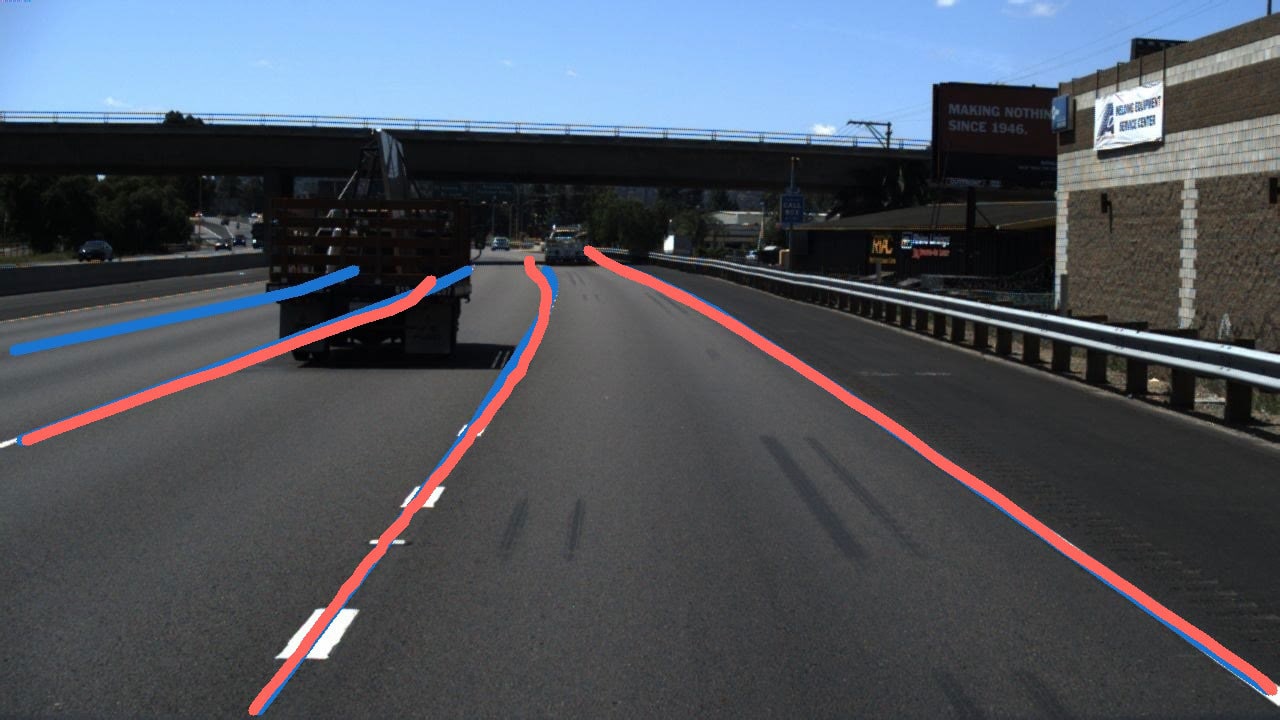} &
			\includegraphics[width=0.18\linewidth,valign=m]{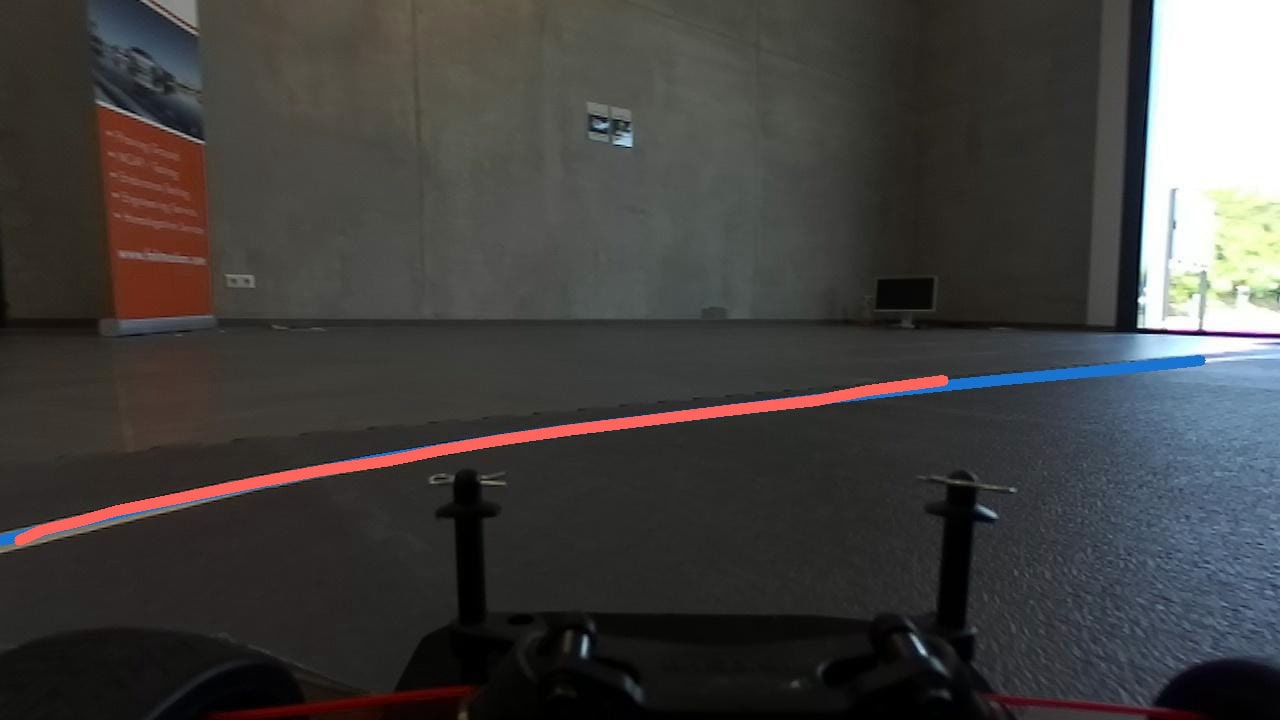} & \includegraphics[width=0.18\linewidth,valign=m]{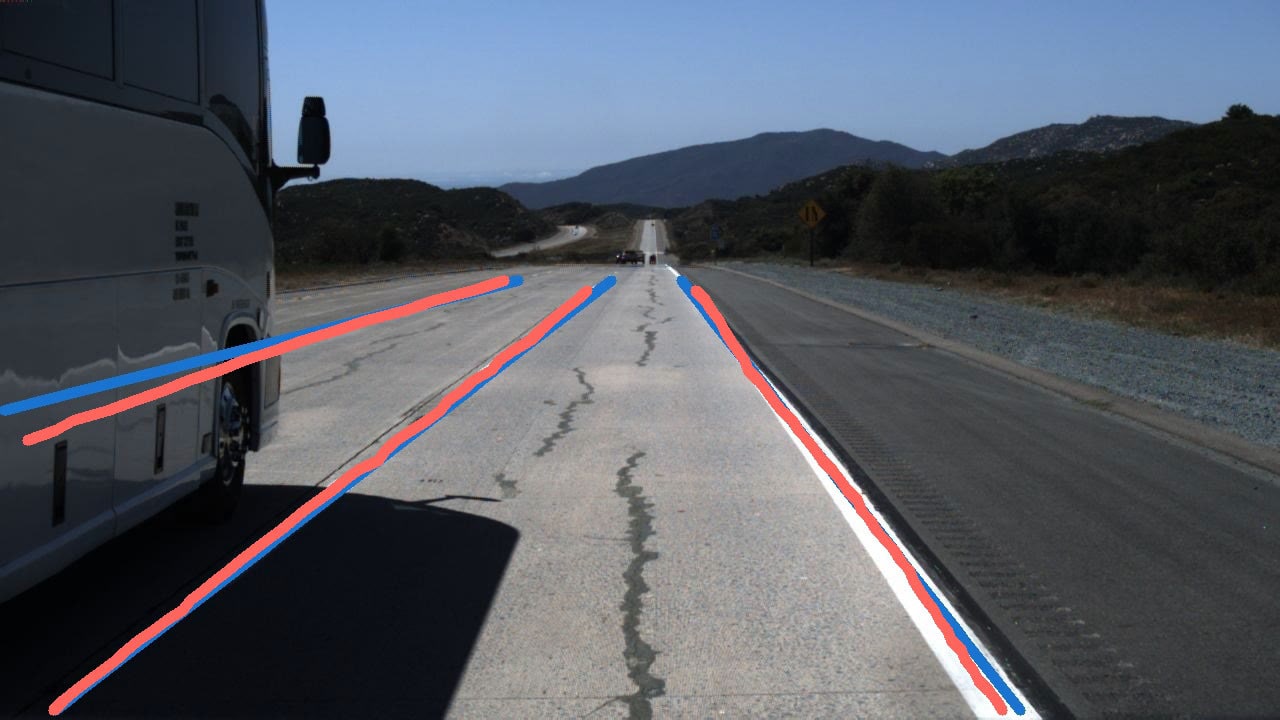}\\
		\end{tabular}
	\end{center}
	\vspace{-1ex}
	\caption[More qualitative results of target domain predictions.]{More qualitative results of target domain predictions. Images are randomly sampled. Ground truth lane annotations are marked in blue, and predictions in red. Best viewed in color.}
	\label{fig:carlane:app:inference_samples_2}
\end{figure}

\begin{figure}
	\small
	\begin{center}
		\begin{tabular}{rc@{}c@{}c@{}c}
			~ & MoLane & TuLane & \multicolumn{2}{c}{MuLane} \\
			UFLD-SO & 
			\includegraphics[width=0.18\linewidth,valign=m]{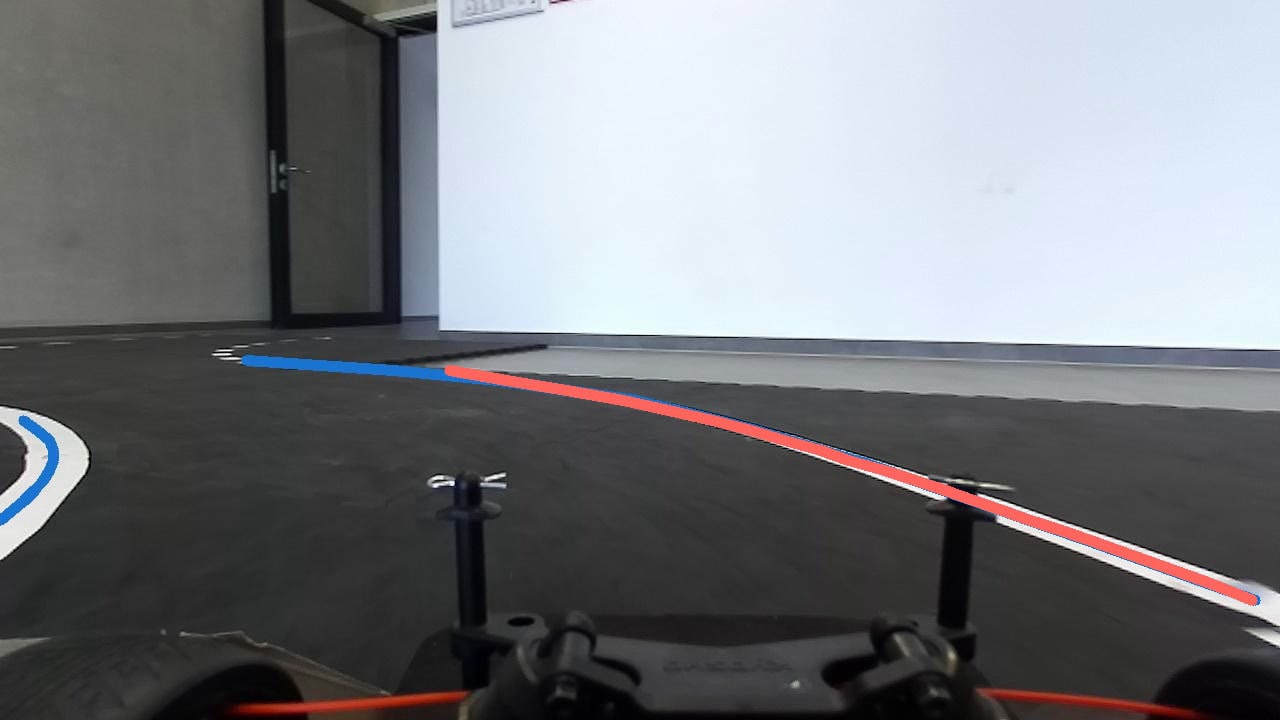} & \includegraphics[width=0.18\linewidth,valign=m]{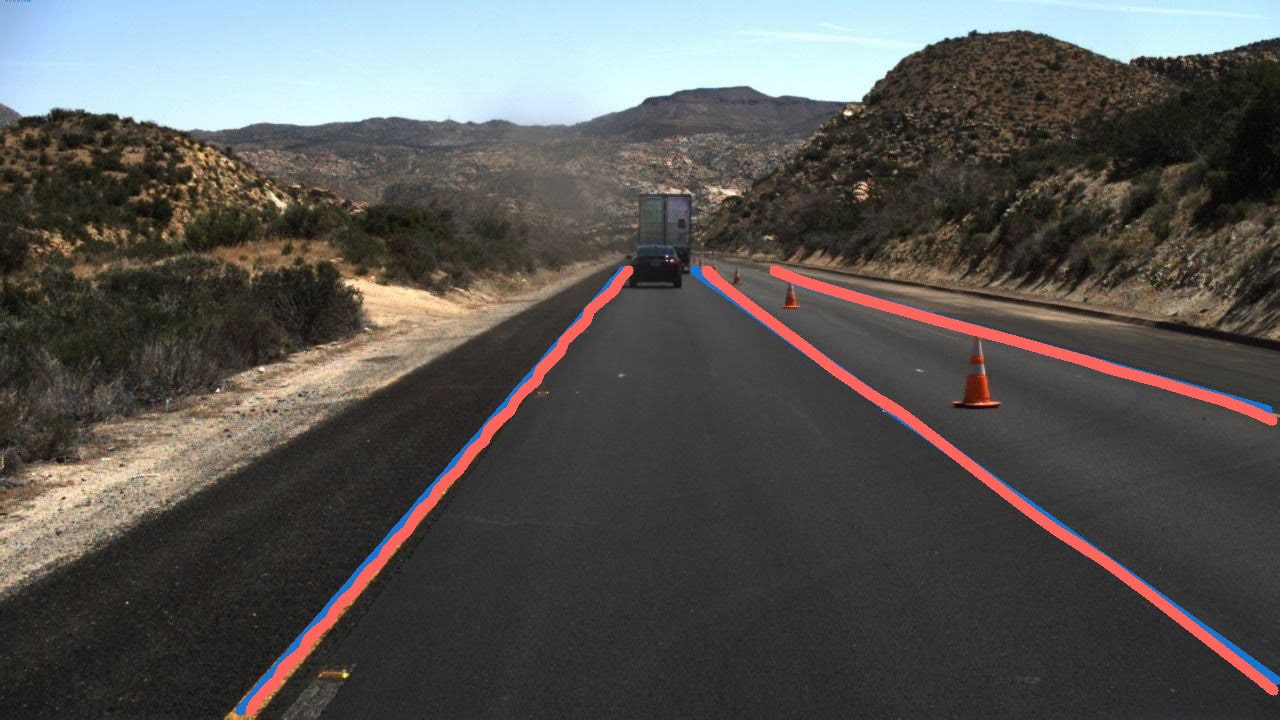} &
			\includegraphics[width=0.18\linewidth,valign=m]{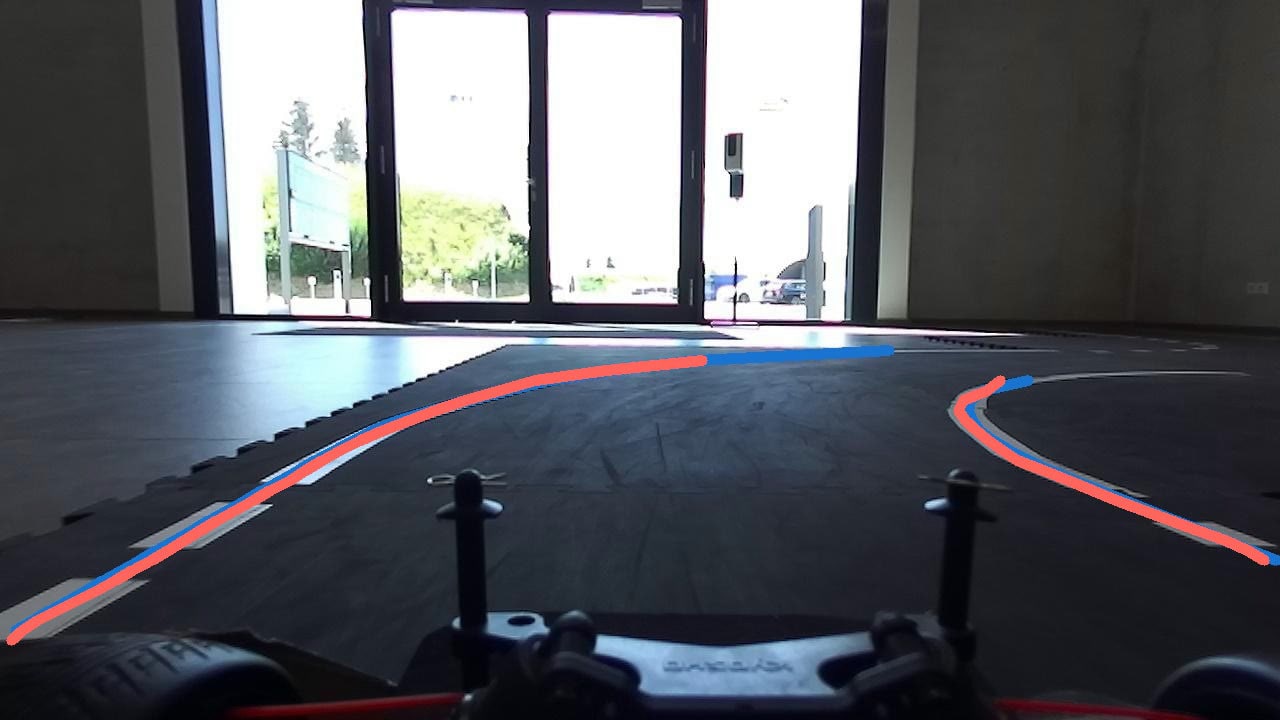} & \includegraphics[width=0.18\linewidth,valign=m]{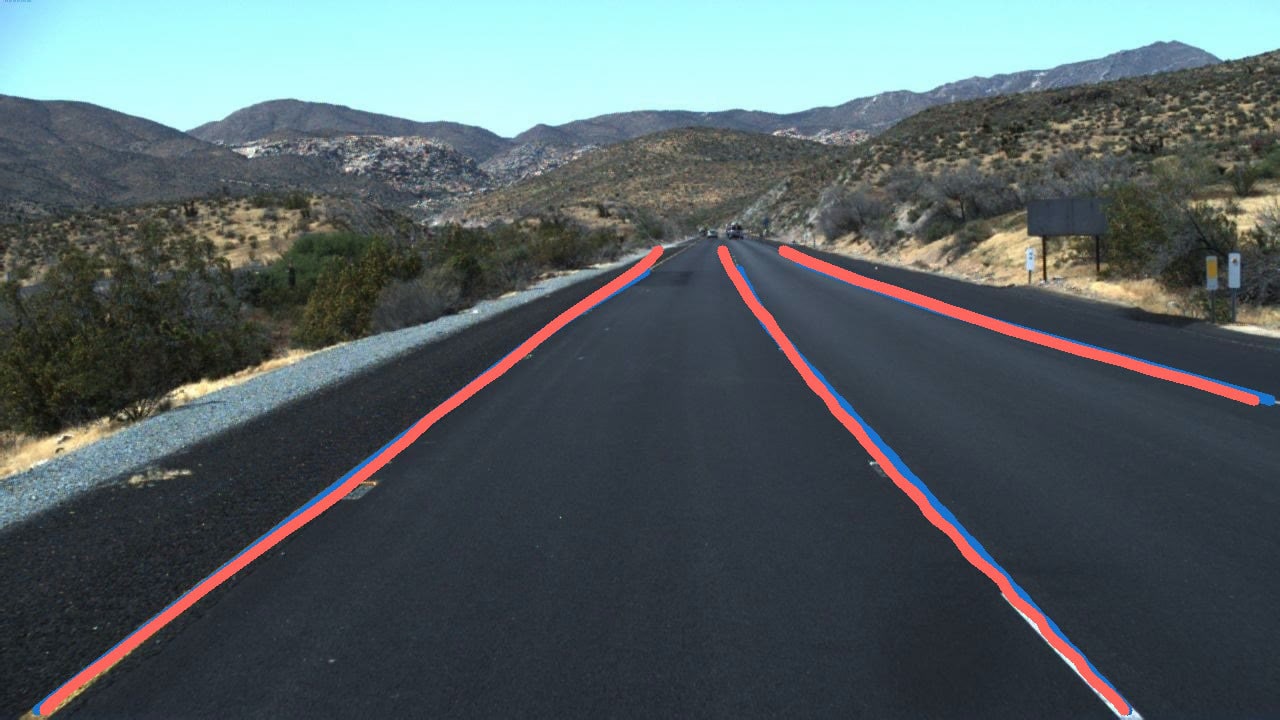}\\
			DANN & 
			\includegraphics[width=0.18\linewidth,valign=m]{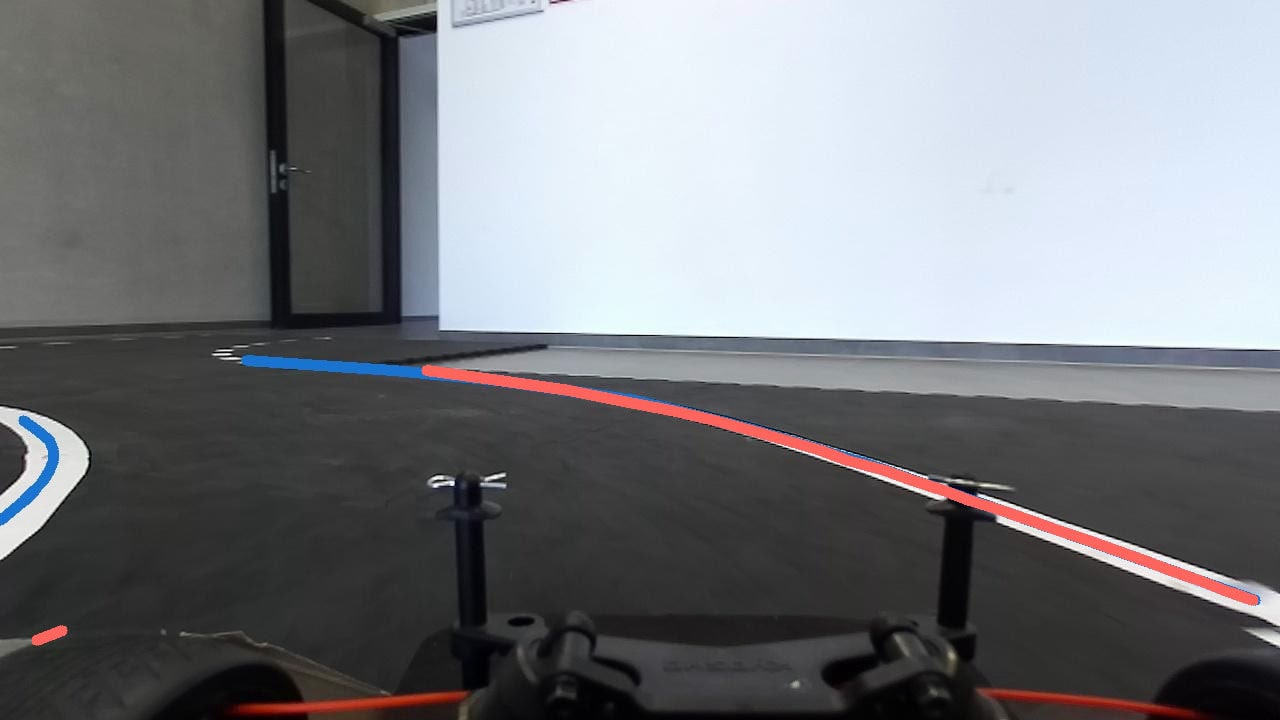} & 
			\includegraphics[width=0.18\linewidth,valign=m]{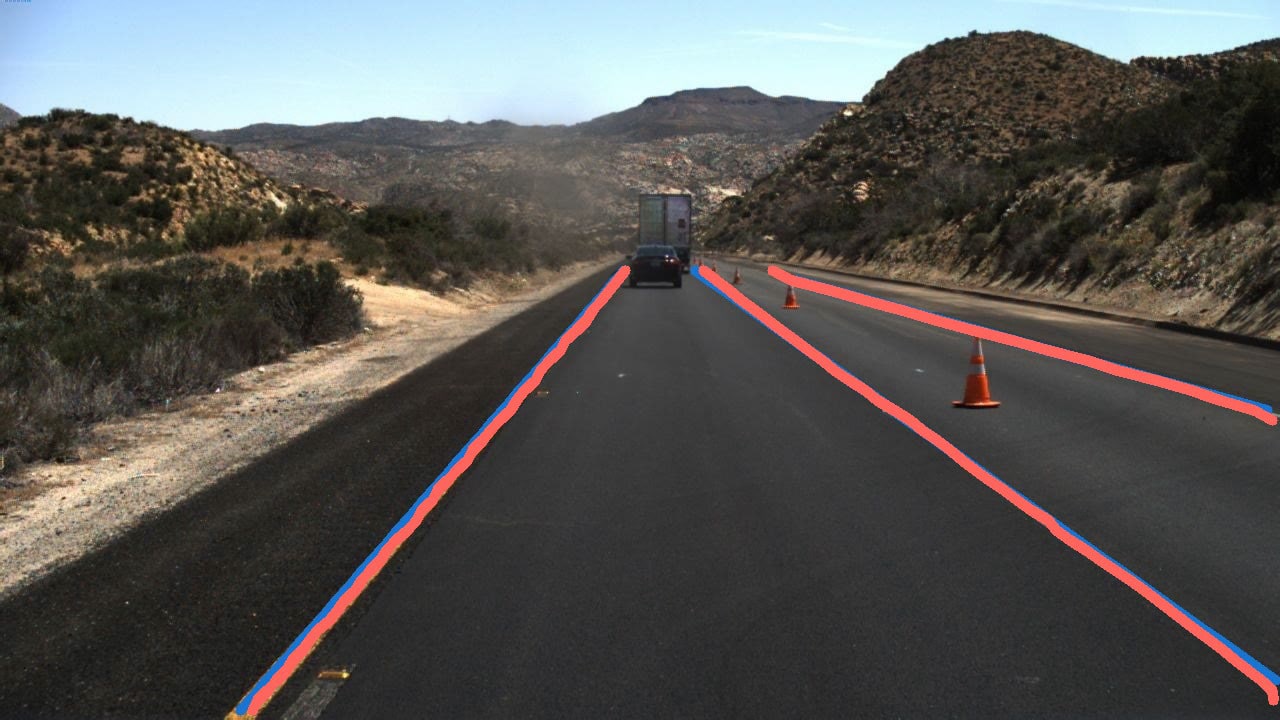} &
			\includegraphics[width=0.18\linewidth,valign=m]{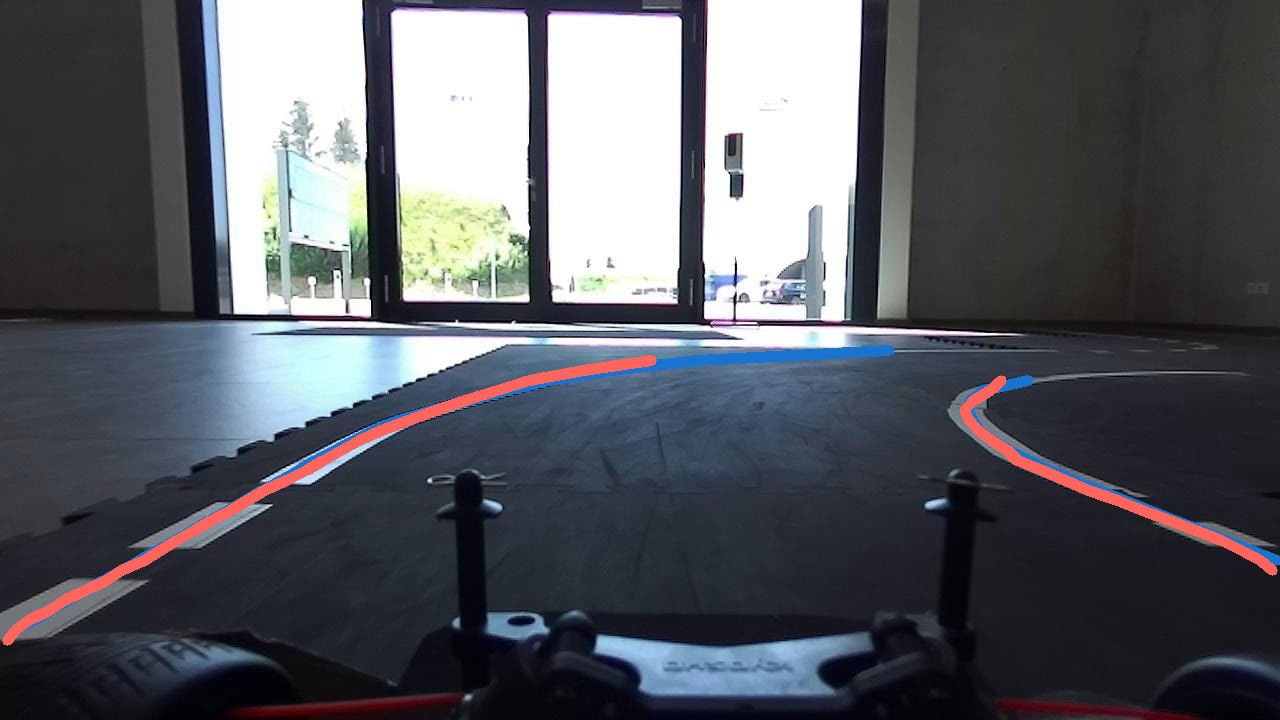} & \includegraphics[width=0.18\linewidth,valign=m]{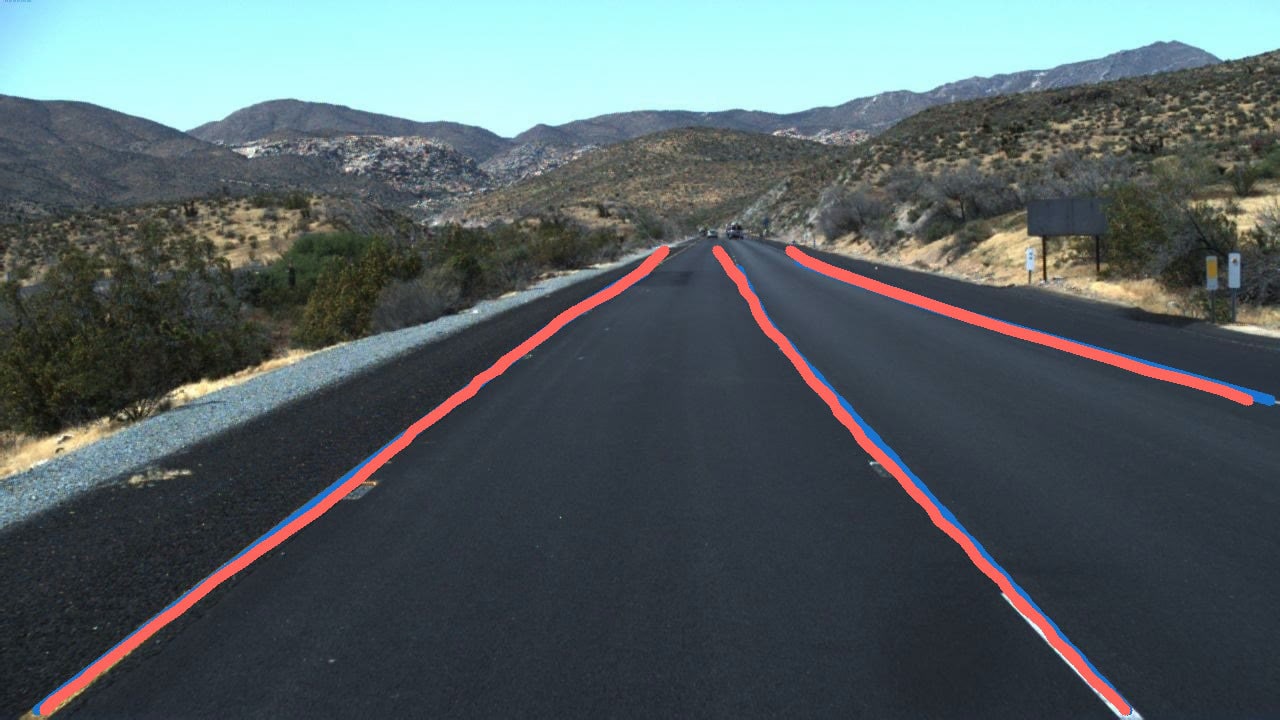}\\
			ADDA & 
			\includegraphics[width=0.18\linewidth,valign=m]{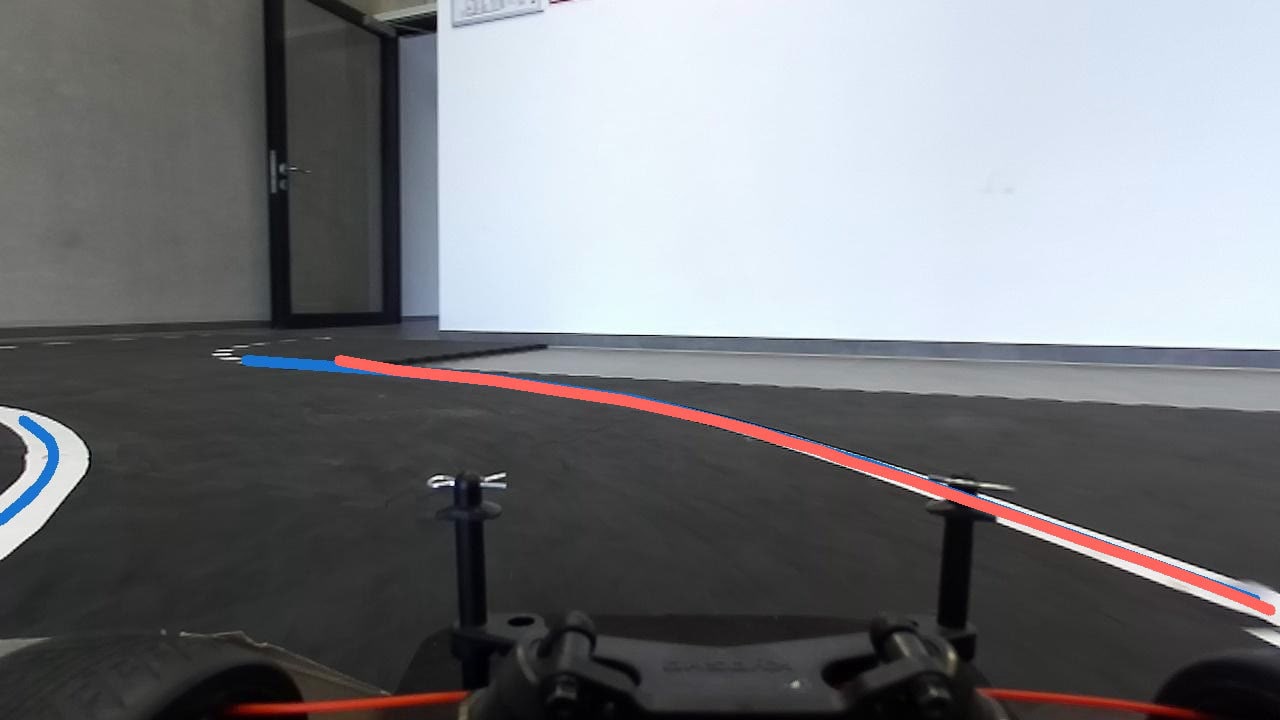} & 
			\includegraphics[width=0.18\linewidth,valign=m]{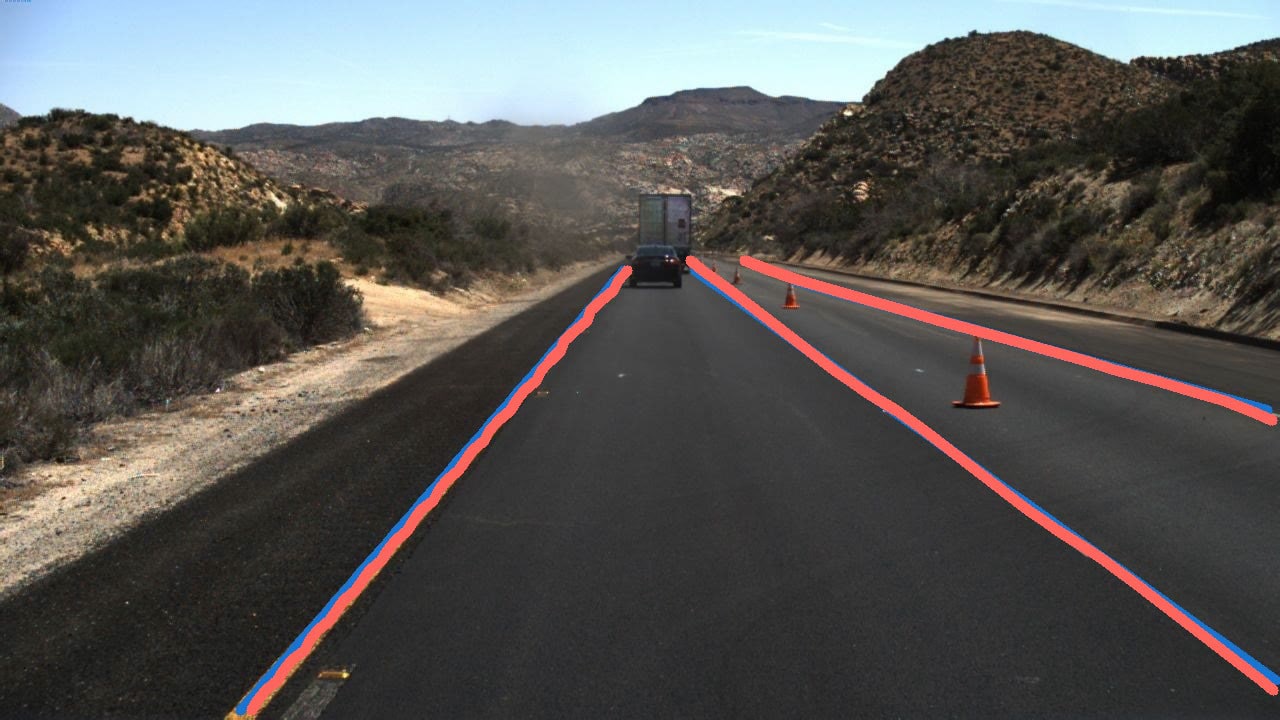} &
			\includegraphics[width=0.18\linewidth,valign=m]{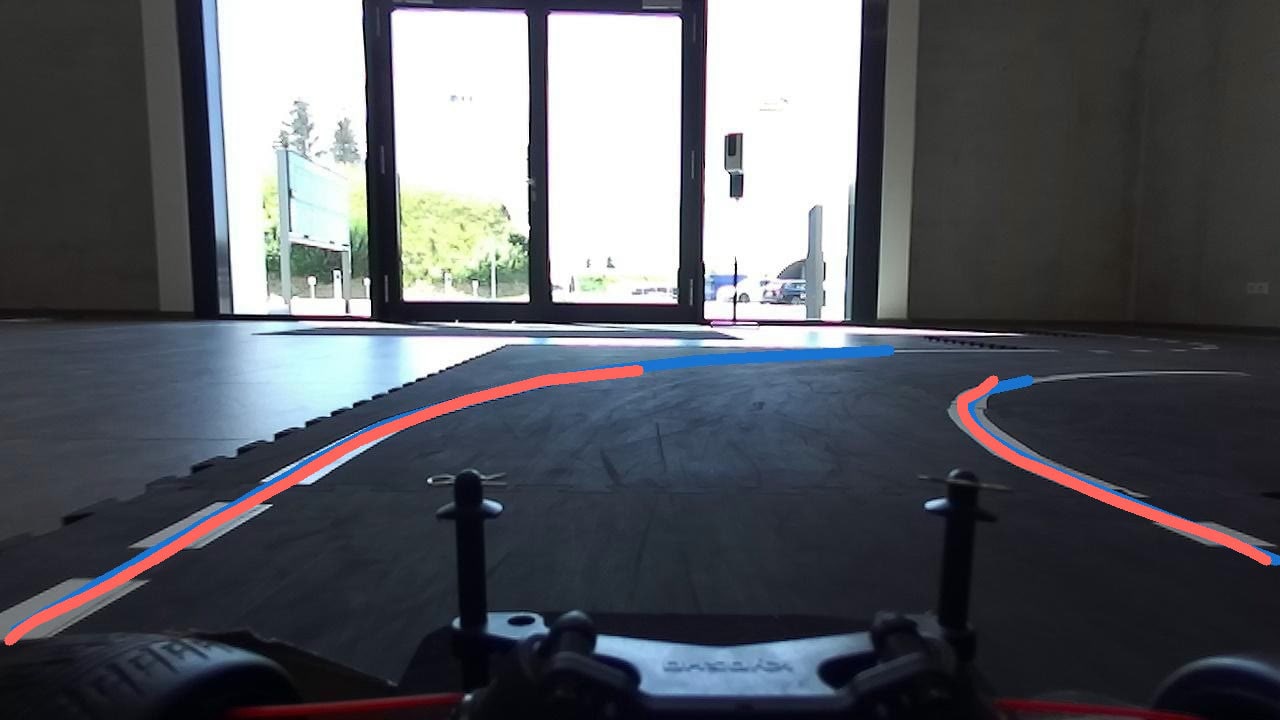} & \includegraphics[width=0.18\linewidth,valign=m]{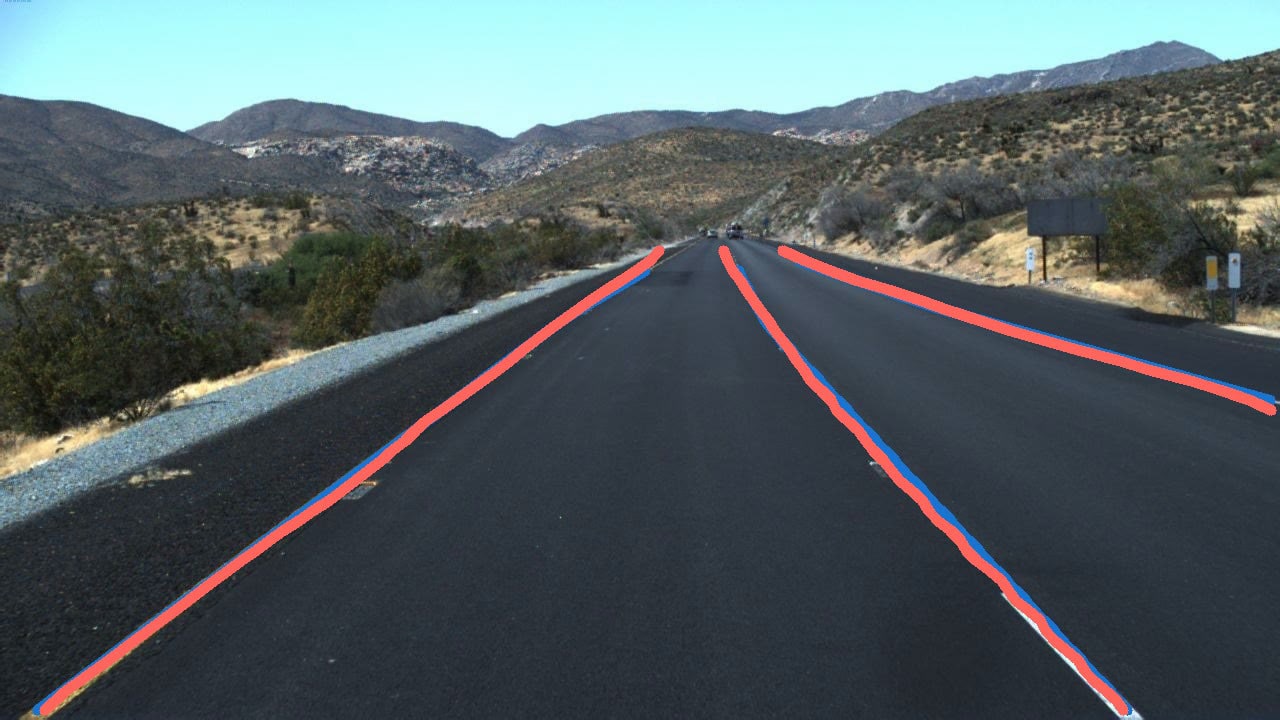}\\
			SGADA & 
			\includegraphics[width=0.18\linewidth,valign=m]{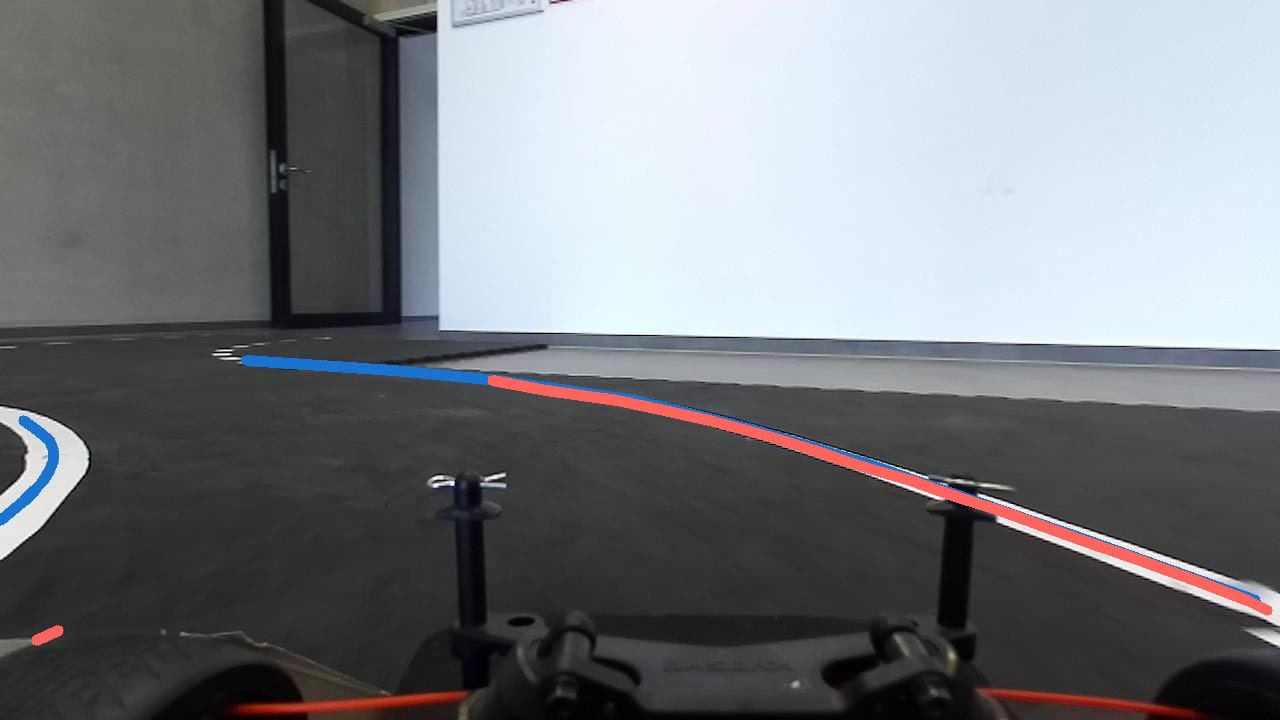} & 
			\includegraphics[width=0.18\linewidth,valign=m]{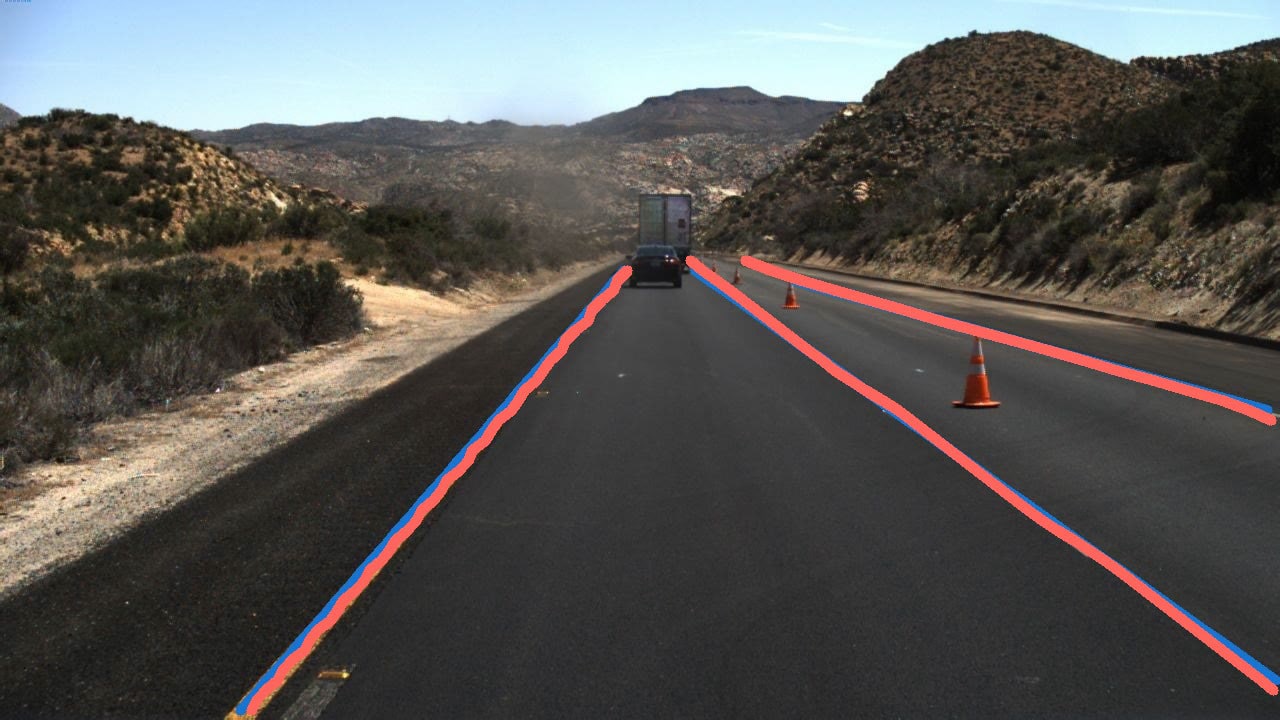} &
			\includegraphics[width=0.18\linewidth,valign=m]{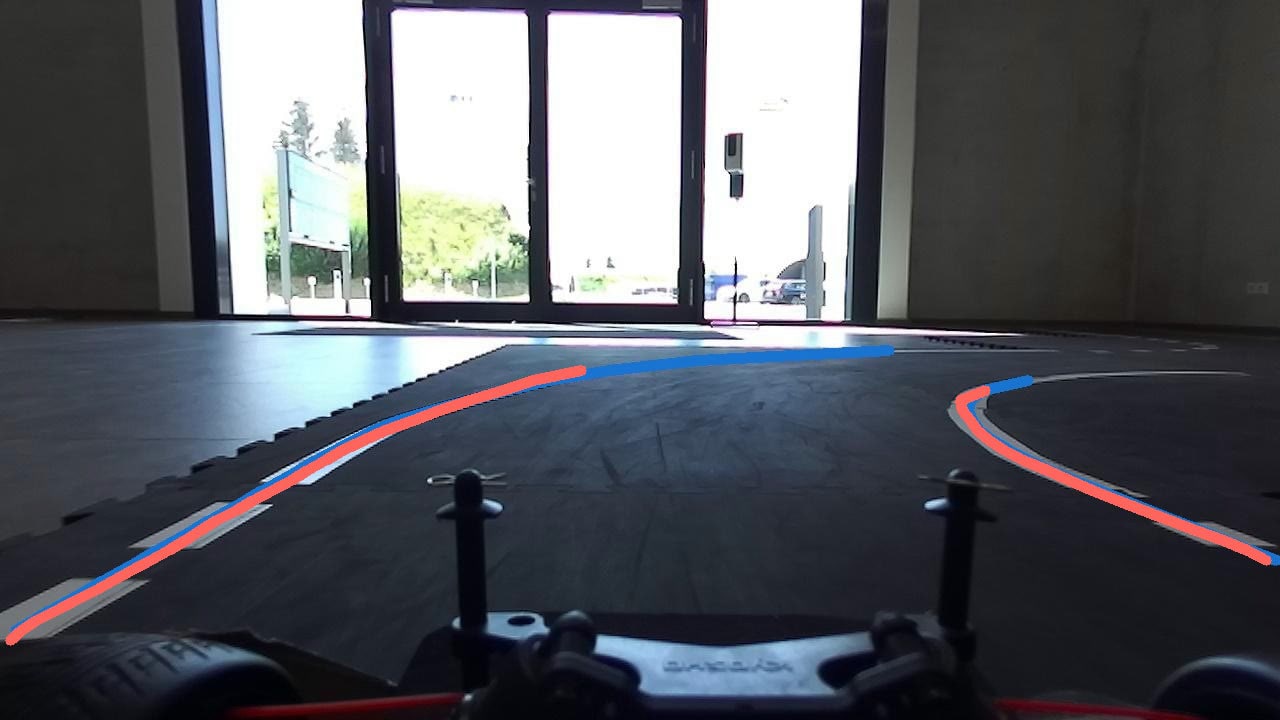} & \includegraphics[width=0.18\linewidth,valign=m]{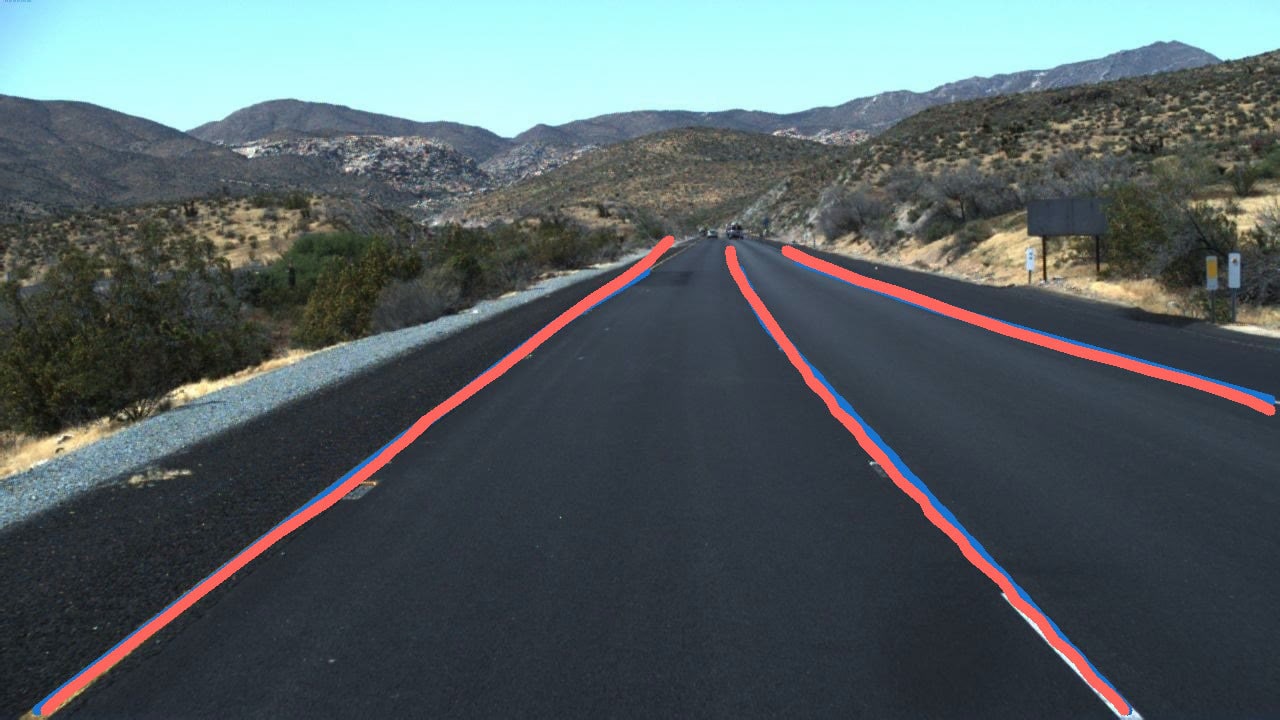}\\
			SGPCS & 
			\includegraphics[width=0.18\linewidth,valign=m]{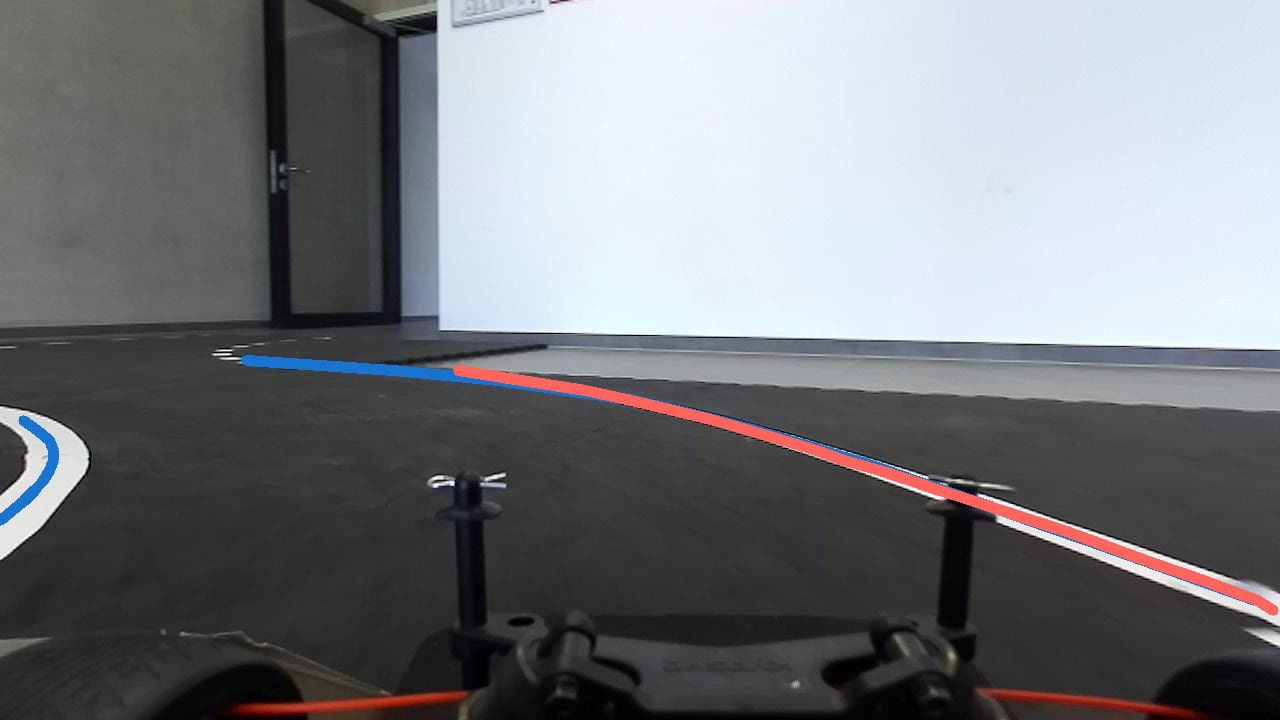} & \includegraphics[width=0.18\linewidth,valign=m]{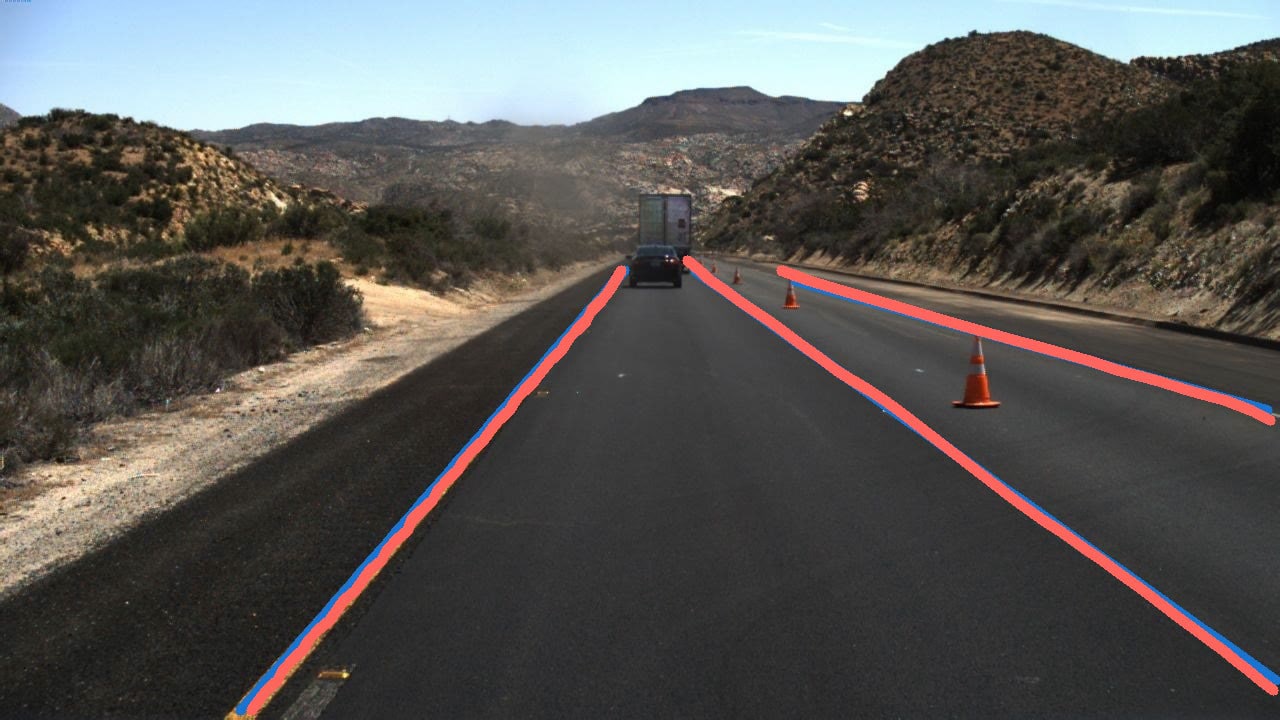} &
			\includegraphics[width=0.18\linewidth,valign=m]{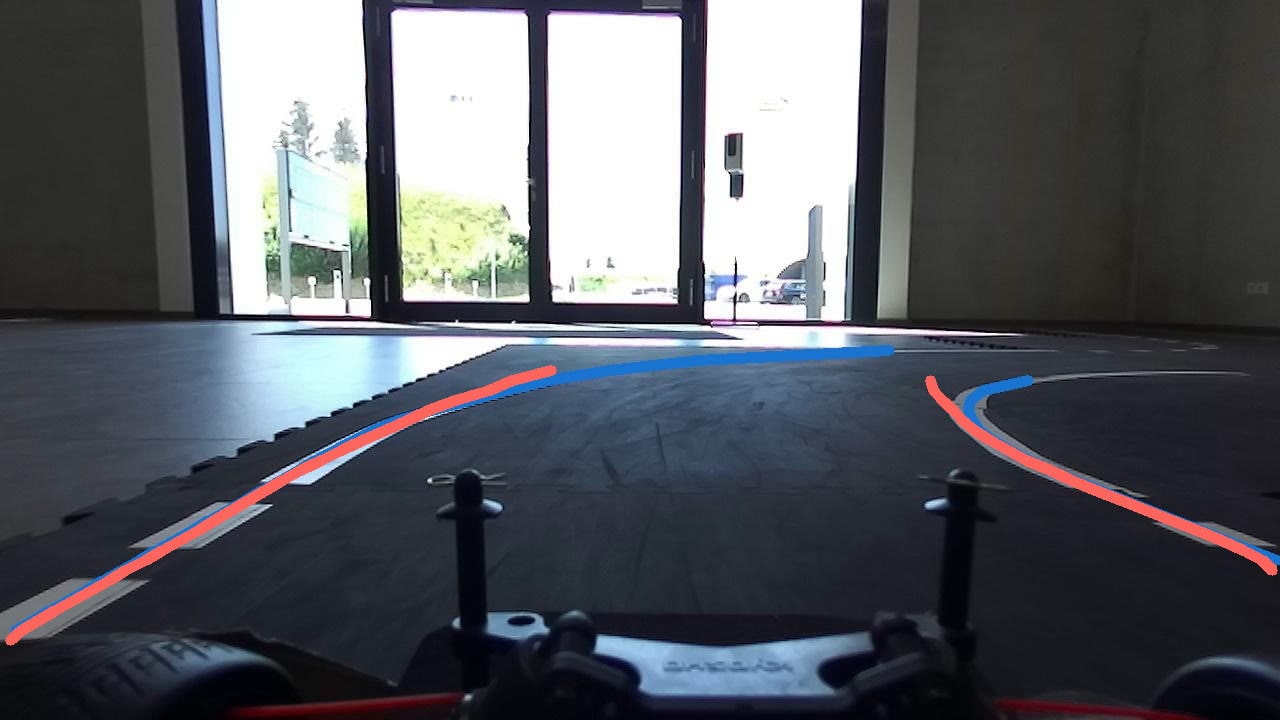} & \includegraphics[width=0.18\linewidth,valign=m]{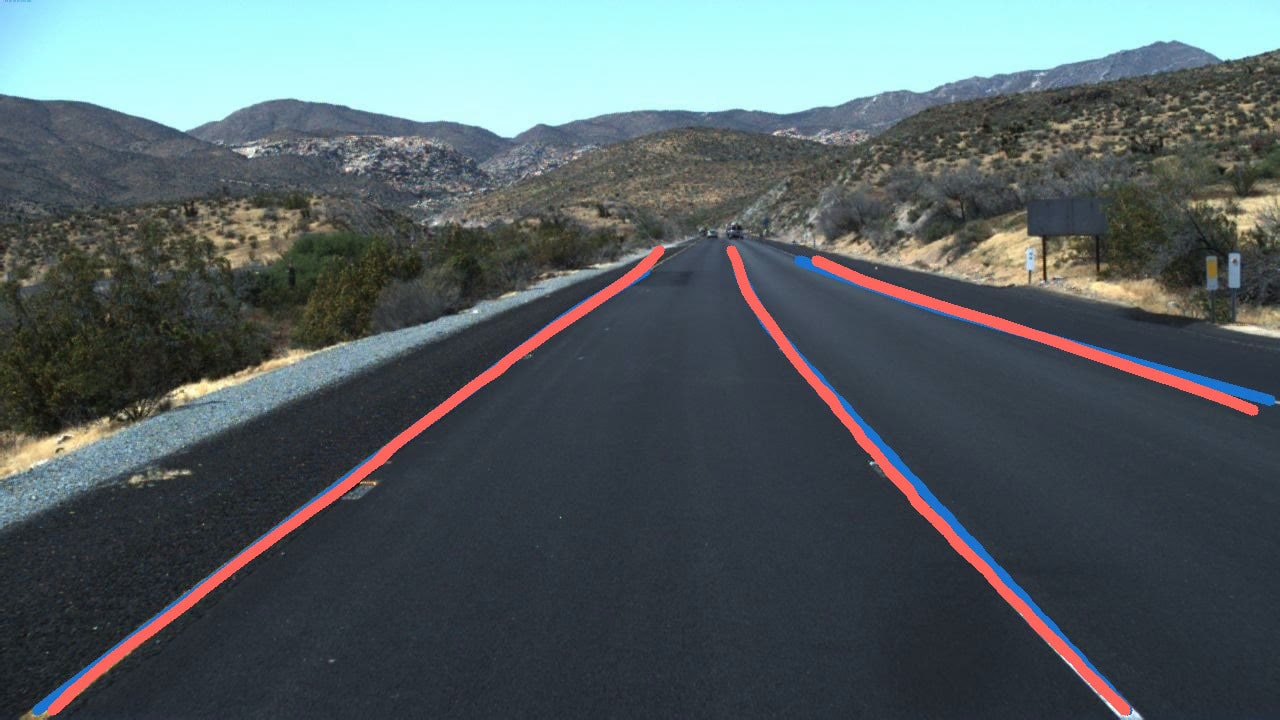}\\
			UFLD-TO & 
			\includegraphics[width=.18\linewidth,valign=m]{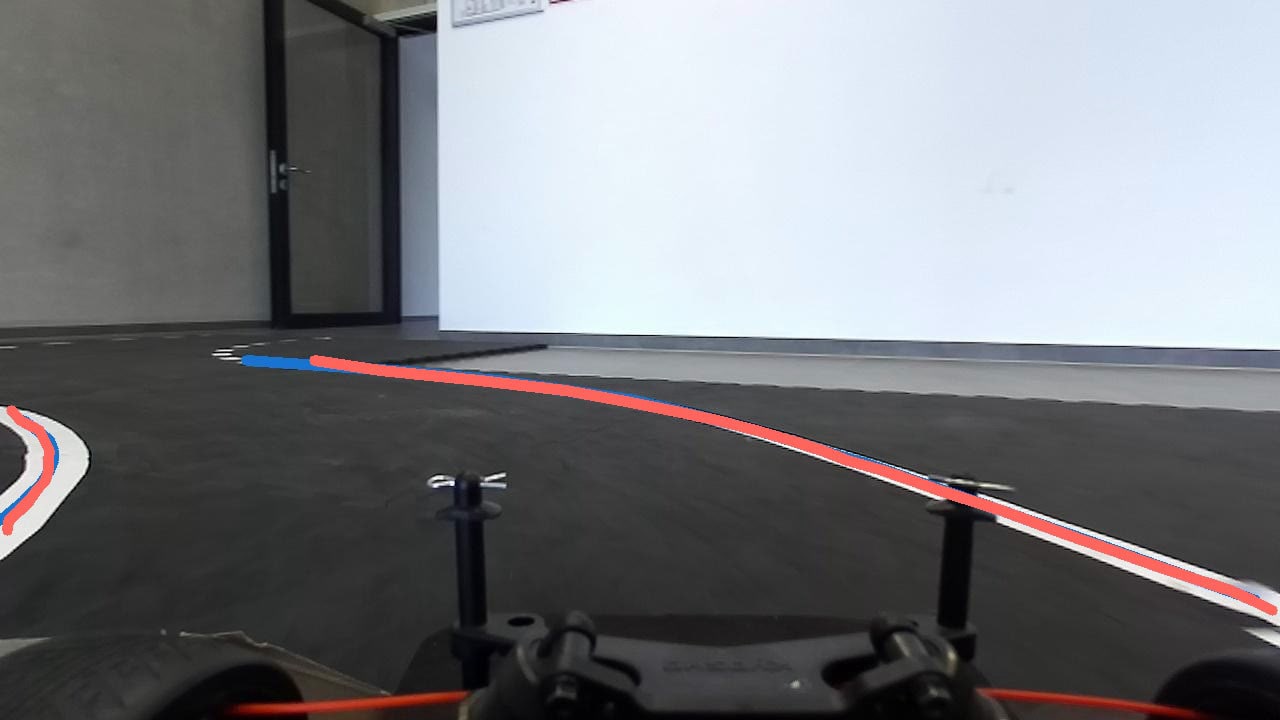} & \includegraphics[width=.18\linewidth,valign=m]{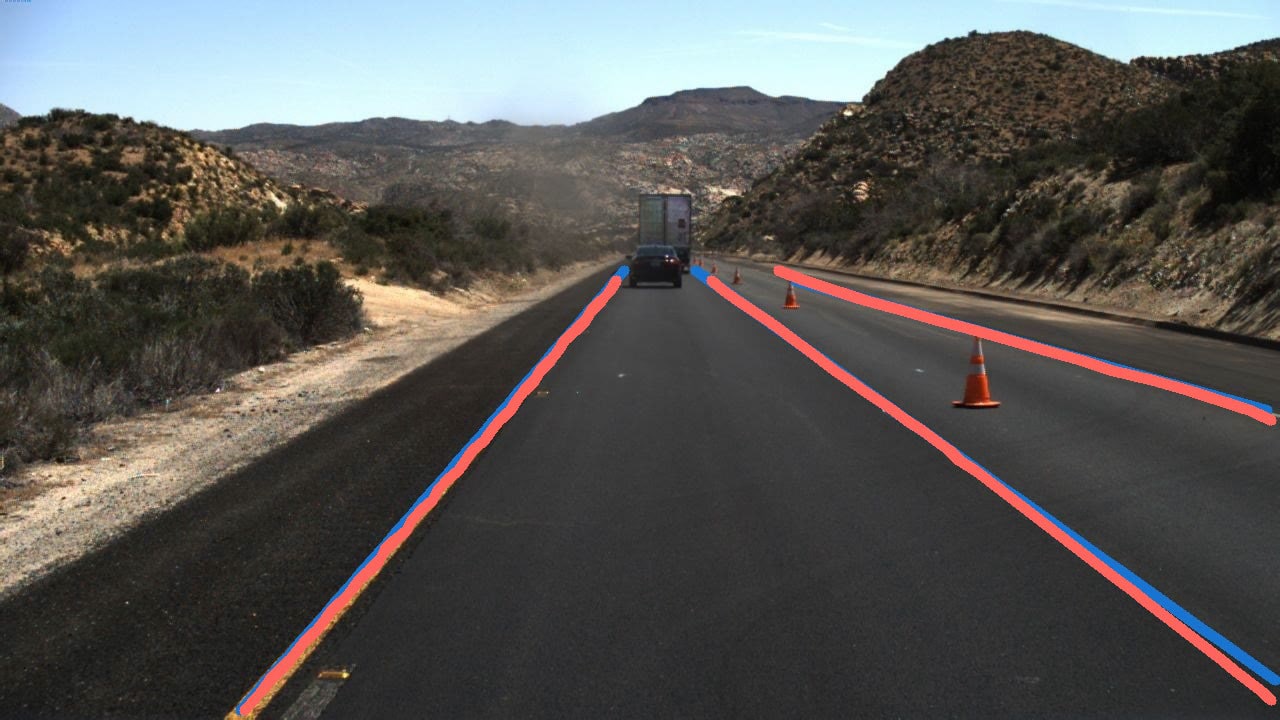} &
			\includegraphics[width=.18\linewidth,valign=m]{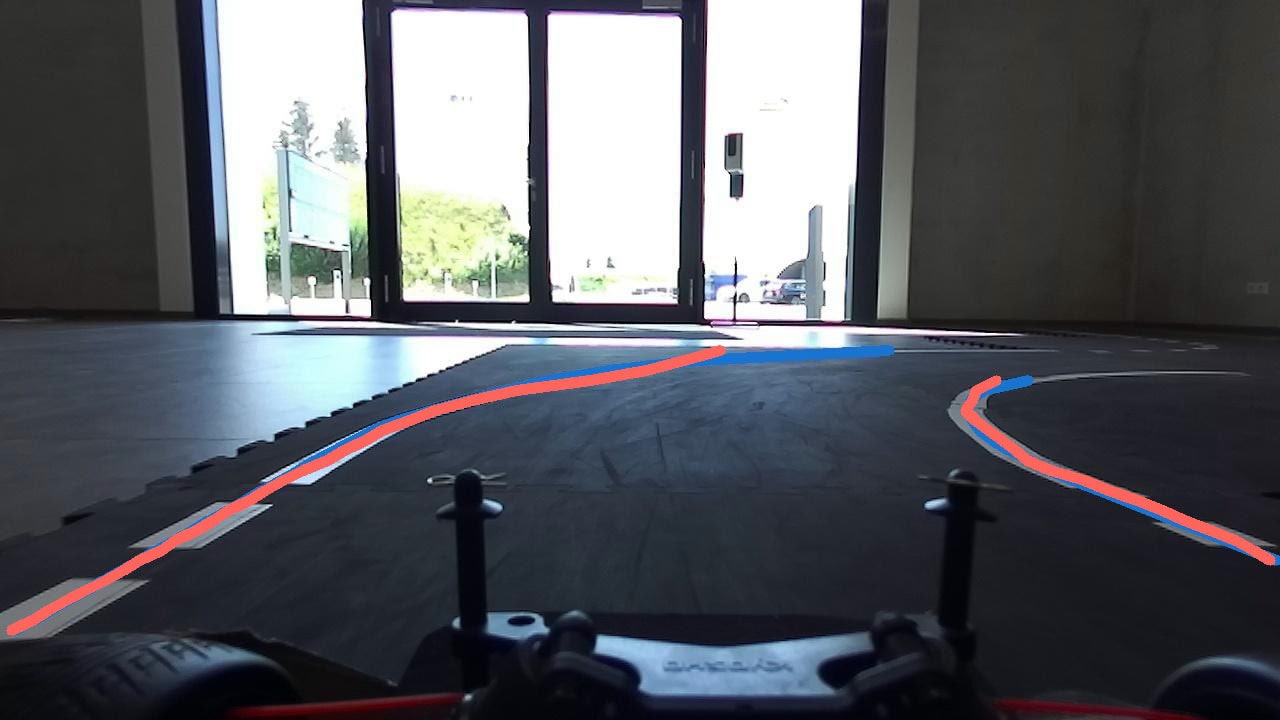} & \includegraphics[width=.18\linewidth,valign=m]{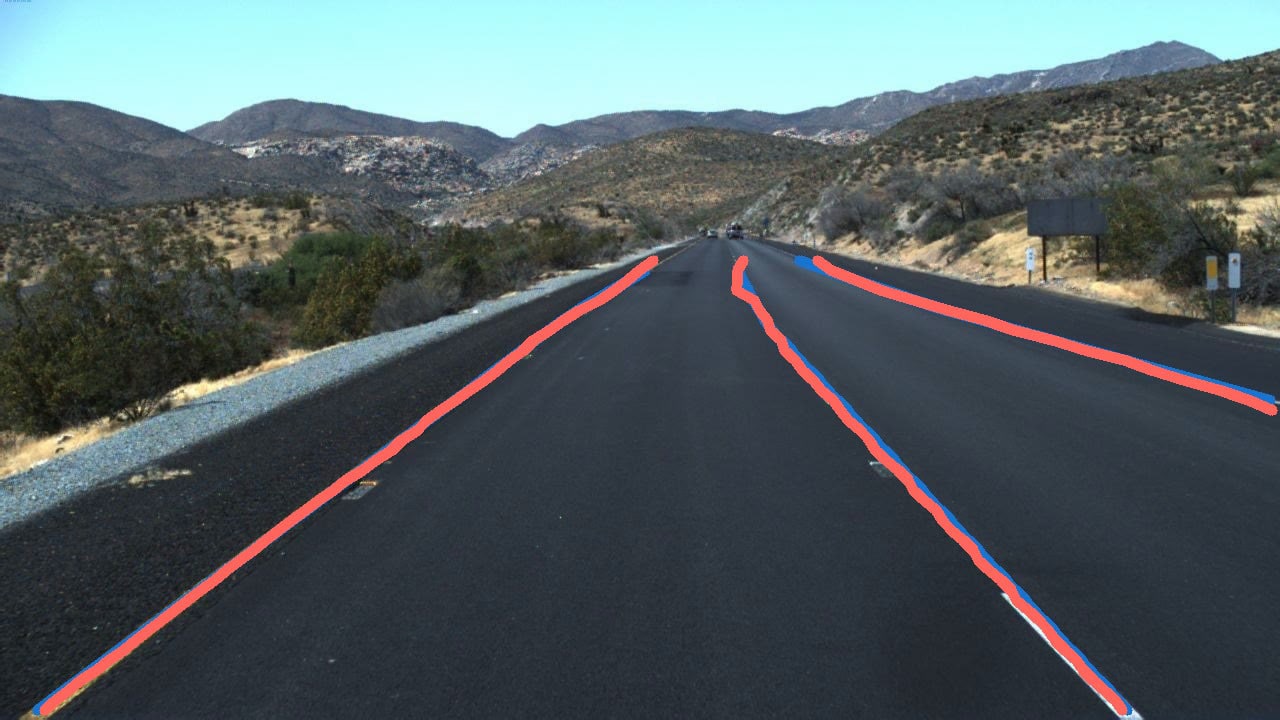}\\
		\end{tabular}
	\end{center}
	\vspace{-1ex}
	\caption[More qualitative results of target domain predictions.]{More qualitative results of target domain predictions. Images are randomly sampled. Ground truth lane annotations are marked in blue, and predictions in red. Best viewed in color.}
	\label{fig:carlane:app:inference_samples_3}
\end{figure}

\textbf{t-SNE feature clustering.} \autoref{fig:carlane:app:TSNE_plot_molane_tulane} shows the t-SNE feature clustering of the trained baselines for the MoLane and TuLane dataset, respectively. We observe that few features of both domains spread over the entire plot for higher-performing unsupervised domain adaptation methods. However, there are still large clusters of features from one domain, indicating that the domain adaptation only occurred slightly.
\\\\
\textbf{Qualitative results.} We randomly sample results from our baselines and show them in Figures \ref{fig:carlane:app:inference_samples_1}, \ref{fig:carlane:app:inference_samples_2}, and \ref{fig:carlane:app:inference_samples_3}. Compared to UFLD-SO, the unsupervised domain adaptation baselines ADDA, SGADA, and SGPCS increase performance consistently. UFLD-TO samples show the best results on the target domain. 

\section{Comparison to Related Work}
\begin{figure}[t]
	\small
	\begin{center}
		\begin{tabular}{rc@{}c@{}c}
			\cite{garnett20193d} & 
			\includegraphics[width=0.27\linewidth,valign=m]{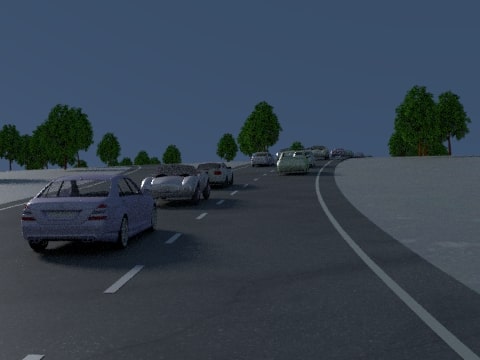} & 
			\includegraphics[width=0.27\linewidth,valign=m]{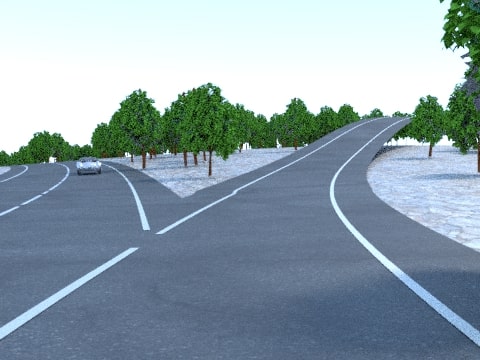} &
			\includegraphics[width=0.27\linewidth,valign=m]{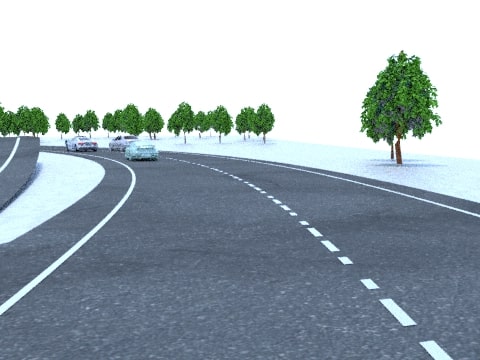}\\
			ours & 
			\includegraphics[width=0.27\linewidth,valign=m]{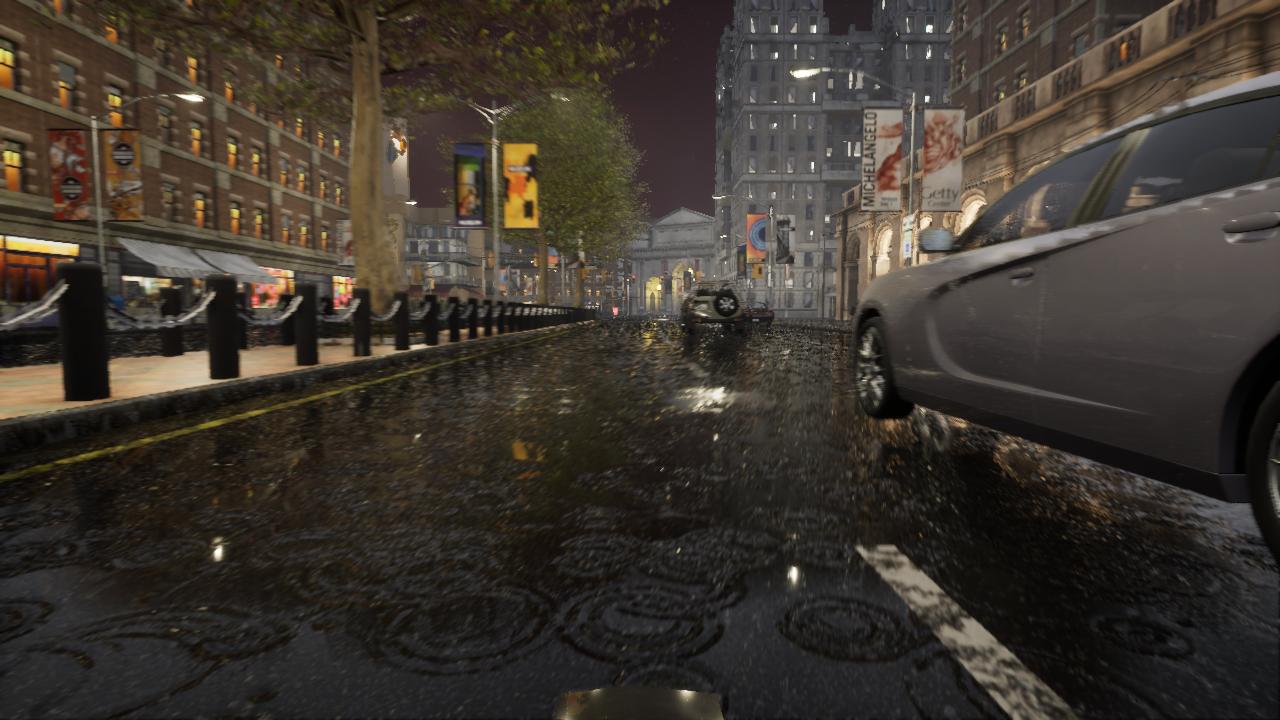} & 
			\includegraphics[width=0.27\linewidth,valign=m]{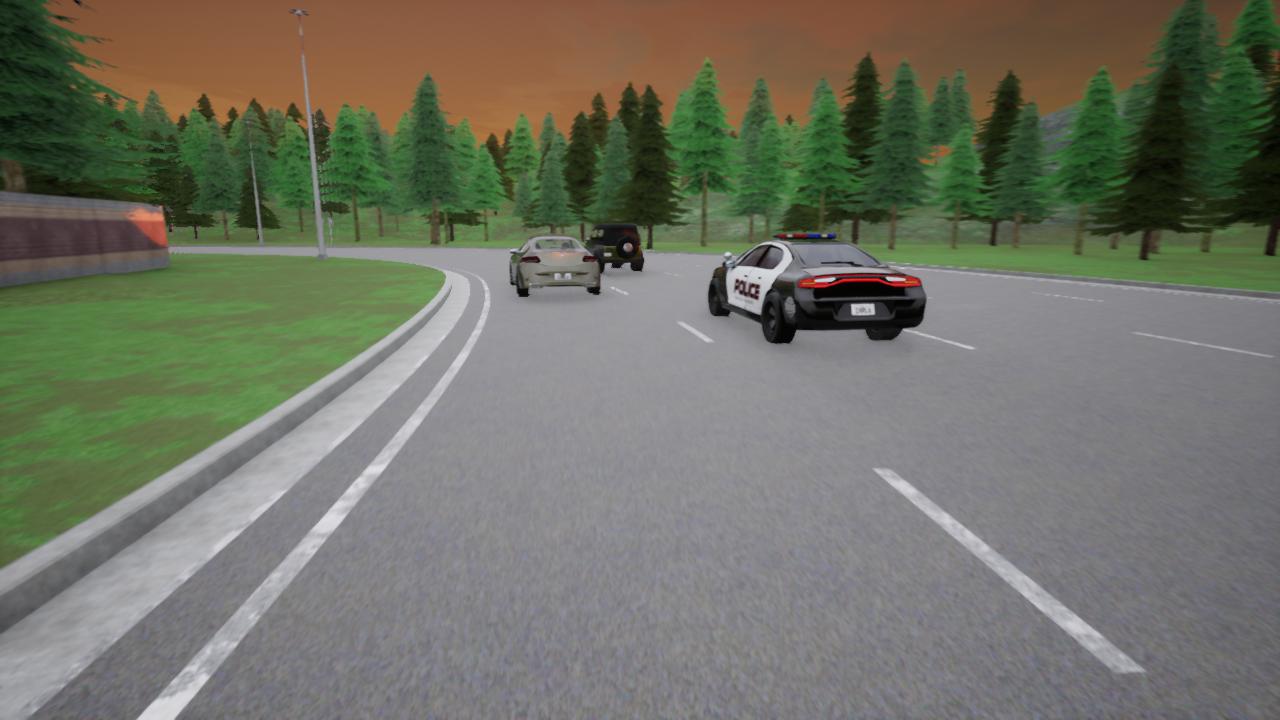} & 
			\includegraphics[width=0.27\linewidth,valign=m]{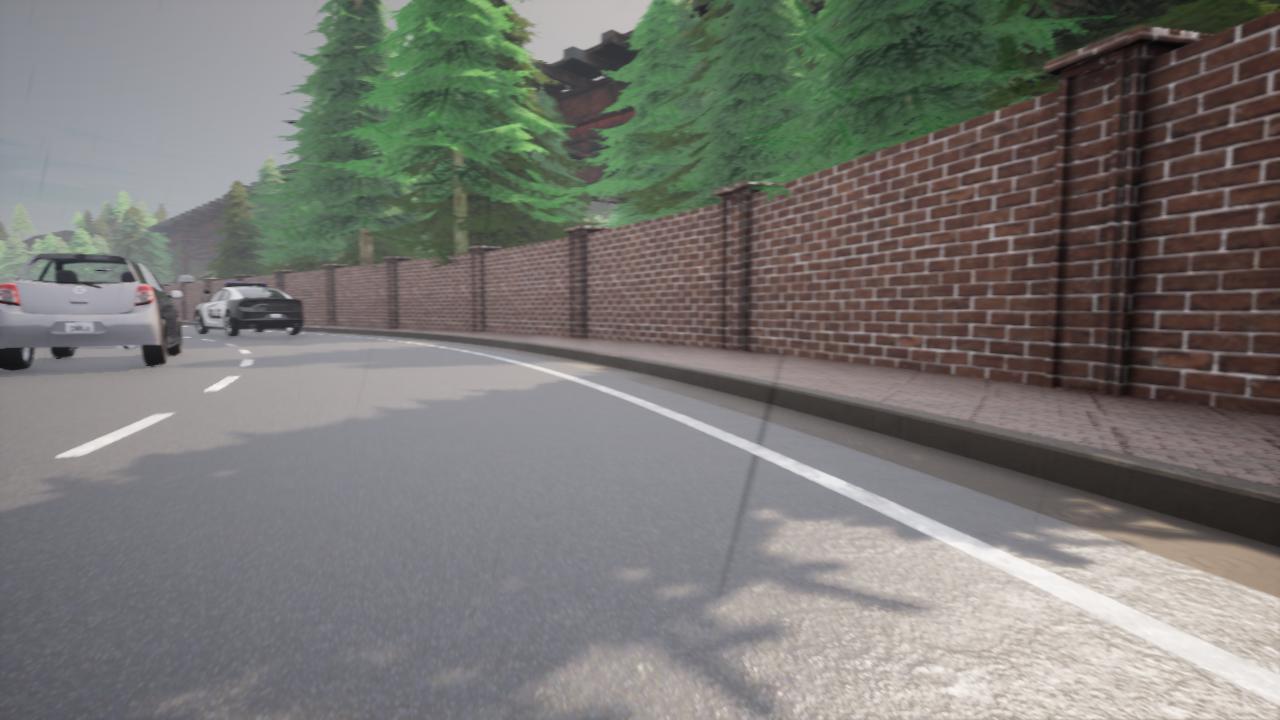} \\
		\end{tabular}
	\end{center}
	\vspace{-1ex}
	\caption[Visual comparison of simulation images.]{Visual comparison of simulation images from the custom blender simulation used in \cite{garnett20193d, garnett2020synthetic} and the CARLA simulation used by \cite{hu2022sim} and our work. We observe that scenes captured in CARLA are more detailed and realistic. Best viewed in color.}
	\label{fig:carlane:app:compare_sim_images}
\end{figure}

In \autoref{tab:carlane:app:dataset_comparison}, we compare CARLANE with the datasets created by related work. The main differentiators are that our dataset contains three distinct domains, including a scaled model vehicle, and is publicly available. To further compare our synthetic datasets with related work, the applied variations during the data collection process are summarized in \autoref{tab:carlane:app:dataset_variations_comparison}. Additionally, we highlight noticeable differences in the visual quality of the simulation engines in \autoref{fig:carlane:app:compare_sim_images}. Scenes captured in CARLA are more realistic and detailed.

\begin{table}[t]
	\RawFloats
	\caption[Comparison of CARLANE (ours) with datasets created by related work.]{Comparison of CARLANE (ours) with datasets created by related work.} 
	\vspace{-1ex}
	\label{tab:carlane:app:dataset_comparison}
	\begin{center}
		\scalebox{0.8}{%
			\setlength{\tabcolsep}{0.4em}
			\begin{tabular}{cccccccc}
				\toprule
				Dataset & Year  & \begin{tabular}[c]{@{}c@{}}Publicly\\ Available\end{tabular} &  Domains & Simulation & Resolution & \begin{tabular}[c]{@{}c@{}}Total\\ Images\end{tabular} &  Annotations  \\ 
				\midrule
				\cite{garnett20193d}                    & 2019 & \xmark & sim, real  & blender & $480 \times 360$ & $391$K    & 3D  \\
				\cite{garnett2020synthetic}      & 2020 & \xmark & sim, real  & blender & $480 \times 360$  & $586$K    & 3D \\
				\cite{hu2022sim}      & 2022 & \xmark & sim, real  & Carla & $1280 \times 720$  & $23$K    & 2D \\
				ours & 2022 & \cmark & sim, real, scaled  & Carla & $1280 \times 720$ & $163$K  & 2D \\
				\bottomrule
		\end{tabular}}
	\end{center}
	\vspace{-2ex}
\end{table}

\begin{table}[t]
	\RawFloats
	\caption[Comparison of applied variations for the collection of the synthetic datasets.]{Comparison of applied variations for the collection of the synthetic datasets.} 
	\vspace{-1ex}
	\label{tab:carlane:app:dataset_variations_comparison}
	\begin{center}
		\scalebox{0.9}{%
			\setlength{\tabcolsep}{0.4em}
			\begin{tabular}{ccccccccccccccc}
				\toprule
				Dataset & \head{Ego Vehicle}  & \head{Camera Position} &  \head{Lane Deviation} & \head{Traffic} & \head{Pedestrians} &  \head{World Objects}  & \head{Daytime} & \head{Weather} & \head{City} & \head{Rural} & \head{Highway} & \head{Terrain} & \head{Lane Topology} & \head{Road Appearance}\\ 
				\midrule
				\cite{garnett20193d}              & \xmark & \cmark & \cmark & \cmark & \xmark & \cmark & \cmark & \xmark & \xmark & \cmark & \xmark & \cmark & \cmark & \cmark \\
				\cite{garnett2020synthetic}  & \xmark & \cmark & \cmark & \cmark & \xmark & \cmark & \cmark & \xmark & \xmark & \cmark & \xmark & \cmark & \cmark & \cmark \\
				\cite{hu2022sim}                  & \xmark & \xmark & \cmark & \cmark & \cmark & \xmark & \cmark & \cmark & \cmark & \cmark & \cmark & \cmark & \cmark & \cmark \\
				ours                            & \cmark & \cmark & \cmark & \cmark & \xmark & \cmark & \cmark & \cmark & \cmark & \cmark & \cmark & \cmark & \cmark & \cmark \\
				\bottomrule
		\end{tabular}}
	\end{center}
	\vspace{-2ex}
\end{table}

\clearpage
\newpage
\section{NeurIPS Checklist for the CARLANE Benchmark}
\begin{enumerate}
	
	\item For all authors...
	\begin{enumerate}
		\item Do the main claims made in the abstract and introduction accurately reflect the paper's contributions and scope?
		{\color{Blue} [Yes]} See Section~\ref{sec:carlane:conclusion}.
		\item Did you describe the limitations of your work?
		{\color{Blue} [Yes]} See Section~\ref{sec:carlane:conclusion}.
		\item Did you discuss any potential negative societal impacts of your work?
		{\color{Blue} [Yes]} See Section~\ref{sec:carlane:conclusion}.
		\item Have you read the ethics review guidelines and ensured that your paper conforms to them?
		{\color{Blue} [Yes]}
	\end{enumerate}
	
	\item If you are including theoretical results...
	\begin{enumerate}
		\item Did you state the full set of assumptions of all theoretical results?
		{\color{Gray} [N/A]}
		\item Did you include complete proofs of all theoretical results?
		{\color{Gray} [N/A]}
	\end{enumerate}
	
	\item If you ran experiments (e.g. for benchmarks)...
	\begin{enumerate}
		\item Did you include the code, data, and instructions needed to reproduce the main experimental results (either in the supplemental material or as a URL)?
		{\color{Blue} [Yes]} See Abstract of \autoref{chap:02}, Section~\ref{sec:carlane:implementation_details} and the supplemental material.
		\item Did you specify all the training details (e.g., data splits, hyperparameters, how they were chosen)?
		{\color{Blue} [Yes]} See Section~\ref{sec:carlane:CARLANE}, Section~\ref{sec:carlane:experiments}, Section~\ref{sec:carlane:implementation_details}, Table~\ref{tab:carlane:dataset_overview}, Table~\ref{tab:carlane:method_hyperparameters} and the supplemental material.
		\item Did you report error bars (e.g., with respect to the random seed after running experiments multiple times)?
		{\color{Blue} [Yes]} Table~\ref{tab:carlane:quantitativ_results_comparison}.
		\item Did you include the total amount of compute and the type of resources used (e.g., type of GPUs, internal cluster, or cloud provider)?
		{\color{Blue} [Yes]} Section~\ref{sec:carlane:implementation_details}.
	\end{enumerate}
	
	\item If you are using existing assets (e.g., code, data, models) or curating/releasing new assets...
	\begin{enumerate}
		\item If your work uses existing assets, did you cite the creators?
		{\color{Blue} [Yes]} We cited the TuSimple dataset \cite{TuSimple2017}.
		\item Did you mention the license of the assets?
		{\color{Blue} [Yes]} See Section~\ref{sec:carlane:CARLANE}.
		\item Did you include any new assets either in the supplemental material or as a URL?
		{\color{Blue} [Yes]}
		\item Did you discuss whether and how consent was obtained from people whose data you're using/curating?
		{\color{Blue} [Yes]} TuSimple is open-source and licensed under the Apache License, Version 2.0 (January 2004).
		\item Did you discuss whether the data you are using/curating contains personally identifiable information or offensive content?
		{\color{Blue} [Yes]} See Section~\ref{sec:carlane:conclusion}.
	\end{enumerate}
	
	\item If you used crowdsourcing or conducted research with human subjects...
	\begin{enumerate}
		\item Did you include the full text of instructions given to participants and screenshots, if applicable?
		{\color{Gray} [N/A]}
		\item Did you describe any potential participant risks, with links to Institutional Review Board (IRB) approvals, if applicable?
		{\color{Gray} [N/A]}
		\item Did you include the estimated hourly wage paid to participants and the total amount spent on participant compensation?
		{\color{Gray} [N/A]}
	\end{enumerate}
	
\end{enumerate}

\clearpage
\newpage
\twocolumn

\section{Datasheet for the\\CARLANE Benchmark}
\datasheetsection{Motivation}
\begin{datasheetitem}{For what purpose was the dataset created? \normalfont Was there a specific task in mind? Was there a specific gap that needed to be filled? Please provide a description.}
	\noindent CARLANE was created to be the first publicly available single- and multi-target Unsupervised Domain Adaptation (UDA) benchmark for lane detection to facilitate future research in these directions. However, in a broader sense, the datasets of CARLANE were also created for unsupervised and semi-supervised learning and partially for supervised learning. Furthermore, a real-to-real transfer can be performed between the target domains of our datasets.
\end{datasheetitem}
\begin{datasheetitem}{Who created the dataset (e.g., which team, research group) and on behalf of which entity (e.g., company, institution, organization)?}
	\noindent As released on June 17, 2022, the initial version of CARLANE was created by Julian Gebele, Bonifaz Stuhr, and Johann Haselberger from the Institute for Driver Assistance Systems and Connected Mobility (IFM). The IFM is a part of the University of Applied Sciences Kempten. Furthermore, CARLANE was created by Bonifaz Stuhr as part of his Ph.D. at the Autonomous University of Barcelona (UAB) and by Johann Haselberger as part of his Ph.D. at the Technische Universität Berlin (TU Berlin).
\end{datasheetitem}
\begin{datasheetitem}{Who funded the creation of the dataset? \normalfont If there is an associated grant, please provide the name of the grantor and the grant name and number.}
	\noindent There is no specific grant for the creation of the CARLANE Benchmark. The datasets were created as part of the work at the IFM and the University of Applied Sciences Kempten.
\end{datasheetitem}
\datasheetsection{Composition}	
\begin{datasheetitem}{What do the instances that comprise the dataset represent (e.g., documents, photos, people, countries)? \normalfont Are there multiple types of instances (e.g., movies, users, and ratings; people and interactions between them; nodes and edges)? Please provide a description.}
	\noindent The instances are drives on diverse roads in simulation, in an abstract 1/8th real world, and in full-scale real-world scenarios, along with lane annotations of the up to four nearest lanes to the vehicle.
\end{datasheetitem}
\begin{datasheetitem}{How many instances are there in total (of each type, if appropriate)?}
	\begin{table*}[t]
		\RawFloats
		\caption[Dataset overview.]{Dataset overview. Unlabeled images are denoted by *, partially labeled images are denoted by **.} 
		\vspace{-1ex}
		\label{tab:carlane:app:dataset_overview}
		\begin{center}
			\scalebox{0.8}{%
				\setlength{\tabcolsep}{0.4em}
				\begin{tabular}{lcccccc}
					\toprule
					Dataset                  & domain             & total images & train   & validation  & test  & lanes       \\ \midrule
					\multirow{2}{*}{MoLane}  & CARLA simulation   & 84,000       & 80,000  & 4,000       & -     & \(\leq\) 2  \\ 
					& model vehicle      & 46,843       & 43,843* & 2,000       & 1,000 & \(\leq\) 2  \\ \midrule
					\multirow{2}{*}{TuLane}  & CARLA simulation   & 26,400       & 24,000  & 2,400       & -     & \(\leq\) 4  \\ 
					& TuSimple \cite{TuSimple2017} & 6,408        & 3,268   & 358         & 2,782 & \(\leq\) 4  \\ \midrule
					\multirow{2}{*}{MuLane}  & CARLA simulation   & 52,800       & 48,000  & 4,800       & -     & \(\leq\) 4  \\ 
					& model vehicle + TuSimple \cite{TuSimple2017} & 12,536      & 6,536** & 4,000       & 2,000 & \(\leq\) 4  \\ 
					\bottomrule
			\end{tabular}}
		\end{center}
		\vspace{-2ex}
	\end{table*}
	\noindent \autoref{tab:carlane:app:dataset_overview} shows the per-domain and per-subset breakdown of CARLANE instances. TuSimple is available at\\
	\href{https://github.com/TuSimple/tusimple-benchmark}{https://github.com/TuSimple/tusimple-benchmark}
	under the Apache License Version 2.0, January 2004. 
\end{datasheetitem}
\begin{datasheetitem}{Does the dataset contain all possible instances or is it a sample (not necessarily random) of instances from a larger set? \normalfont If the dataset is a sample, then what is the larger set? Is the sample representative of the larger set (e.g., geographic coverage)? If so, please describe how this representativeness was validated/verified. If it is not representative of the larger set, please describe why not (e.g., to cover a more diverse range of instances, because instances were withheld or unavailable).}
	\noindent The datasets of CARLANE contain samples of driving scenarios and lane annotations encountered in simulation and the real world. The datasets are not representative of all these driving scenarios, as the distribution of the latter is highly dynamic and diverse. Instead, the motivation was to resemble the variety and shifts of different domains in which such scenarios occur to strengthen the systematic study of UDA methods for lane detection. Therefore, CARLANE should be considered as an UDA benchmark with step-by-step testing possibility across three domains, which makes it possible to include an additional safety mechanism for real-world testing.
\end{datasheetitem}
\begin{datasheetitem}{What data does each instance consist of? \normalfont “Raw” data (e.g., unprocessed text or images) or features? In either case, please provide a description.}
	\noindent Each labeled instance consists of the following components: \vspace{-2pt}
	\\\\
	\textit{(1)} A single $1280\times720$ image from a driving scenario. \vspace{-2pt}
	\\\\
	\textit{(2)} A .json file entry for the corresponding subset containing lane annotations following TuSimple. The lanes’ y-values discretized by 56 raw anchors, the lanes’ x-values to 101 gridding cells, with the last gridding cell representing the absence of a lane. The file path to the corresponding image is also stored in the .json file. \vspace{-2pt}
	\\\\
	\textit{(3)} A .png file containing lane segmentations following UFLD (\href{https://github.com/cfzd/Ultra-Fast-Lane-Detection}{https://github.com\\/cfzd/Ultra-Fast-Lane-Detection}), where each of the four lanes has a different label. \vspace{-2pt}
	\\\\
	\textit{(4)} A .txt file entry for the corresponding subset containing the linkage between the raw image and its segmentation as well as the presence and absence of a lane. \vspace{-2pt}
	\\\\
	Each unlabeled instance consists of an $1280\times720$ image from a driving scenario and a .txt file entry for the corresponding subset.
\end{datasheetitem}
\begin{datasheetitem}{Is there a label or target associated with each instance? \normalfont If so, please provide a description.}
	\noindent As described above, the labels per instance are discretized lane annotations and lane segmentations.
\end{datasheetitem}
\begin{datasheetitem}{Is any information missing from individual instances? \normalfont If so, please provide a description, explaining why this information is missing (e.g., because it was unavailable). This does not include intentionally removed information, but might include, e.g., redacted text.}
	\noindent Everything is included. No data is missing.
\end{datasheetitem}
\begin{datasheetitem}{Are relationships between individual instances made explicit (e.g., users’ movie ratings, social network links)? \normalfont If so, please describe how these relationships are made explicit.}
	\noindent There are no relationships made explicit between instances. However, some instances are part of the same drive and therefore have an implicit relationship.
\end{datasheetitem}
\begin{datasheetitem}{Are there recommended data splits(e.g., training, development/validation, testing)? \normalfont If so, please provide a description of these splits, explaining the rationale behind them.}
	\noindent Each domain is split into training and validation subsets. Details are shown in \autoref{tab:carlane:app:dataset_overview}. The target domains for UDA additionally include test sets, which were recorded from separate tracks (model vehicle) or driving scenarios (TuSimple). Since UDA aims to adapt models to target domains, only the target domains include a test set.
\end{datasheetitem}
\begin{datasheetitem}{Are there any errors, sources of noise, or redundancies in the dataset? \normalfont If so, please provide a description.}
	\noindent CARLANE was recorded from different drives through simulation and real-world domains. Therefore there are images captured from the same drive, which result in similar scenarios for consecutive images. Target domain samples were annotated by hand and may include human labeling errors. However, we double-checked labels and cleaned TuSimple's test set with our labeling tool.  
\end{datasheetitem}
\begin{datasheetitem}{Is the dataset self-contained, or does it link to or otherwise rely on external resources (e.g., websites, tweets, other datasets)? \normalfont If it links to or relies on external resources, a) are there guarantees that they will exist, and remain constant, over time; b) are there official archival versions of the complete dataset (i.e., including the external resources as they existed at the time the dataset was created); c) are there any restrictions (e.g., licenses, fees) associated with any of the external resources that might apply to a dataset consumer? Please provide descriptions of all external resources and any restrictions associated with them, as well as links or other access points, as appropriate.}
	\noindent CARLANE is entirely self-contained.
\end{datasheetitem}
\begin{datasheetitem}{Does the dataset contain data that might be considered confidential (e.g., data that is protected by legal privilege or by doctor-patient confidentiality, data that includes the content of individuals’ non-public communications)? \normalfont If so, please provide a description.}
	\noindent The full-scale real-world target domain contains open-source images with unblurred license plates and people from the TuSimple dataset. This data should be treated with respect and in accordance with privacy policies. The other domains do not contain data that might be considered confidential since there where recorded in simulations or a controlled 1/8th real-world environment.
\end{datasheetitem}
\begin{datasheetitem}{Does the dataset contain data that, if viewed directly, might be offensive, insulting, threatening, or might otherwise cause anxiety? \normalfont If so, please describe why.}
	\noindent CARLANE includes driving scenarios; therefore, its datasets could cause anxiety in people with driving anxiety.
\end{datasheetitem}
\begin{datasheetitem}{Does the dataset identify any subpopulations (e.g., by age, gender)? \normalfont If so, please describe how these subpopulations are identified and provide a description of their respective distributions within the dataset.}
	\noindent No.
\end{datasheetitem}
\begin{datasheetitem}{Is it possible to identify individuals (i.e., one or more natural persons), either directly or indirectly (i.e., in combination with other data) from the dataset? \normalfont If so, please describe how.}
	\noindent Yes, individuals could be identified in the full-scale real-world target domain from TuSimple, since it contains unblurred license plates and people. However, the remaining domains do not contain identifiable individuals. 
\end{datasheetitem}
\begin{datasheetitem}{Does the dataset contain data that might be considered sensitive in anyway(e.g., data that reveals race or ethnic origins, sexual orientations, religious beliefs, political opinions or union memberships, or locations; financial or health data; biometric or genetic data; forms of government identification, such as social security numbers; criminal history)? \normalfont If so, please provide a description.}
	\noindent The full-scale real-world target domain from TuSimple could implicitly reveal sensitive information printed or put on the vehicles or people's wearings.
\end{datasheetitem}
\datasheetsection{Collection Process}	
\begin{datasheetitem}{How was the data associated with each instance acquired? \normalfont Was the data directly observable (e.g., raw text, movie ratings), reported by subjects (e.g., survey responses), or indirectly inferred/derived from other data (e.g., part-of-speech tags, model-based guesses for age or language)? If the data was reported by subjects or indirectly inferred/derived from other data, was the data validated/verified? If so, please describe how.}
	\noindent The source domain images of driving scenarios and the corresponding lane annotations were directly recorded from the simulation. Lanes were manually labeled for the directly recorded real-world images. For the images collected from the model vehicle, the authors annotated the data with a labeling tool created for this task. The labeling tool is publicly available at \url{https://carlanebenchmark.github.io}. The labeling tool is utilized to clean up the annotations of the test set in the real-world domain. The authors do not have information about the labeling process of the full-scale target domain since its data is derived from the TuSimple dataset.
\end{datasheetitem}
\begin{datasheetitem}{What mechanisms or procedures were used to collect the data (e.g., hardware apparatuses or sensors, manual human curation, software programs, software APIs)? \normalfont How were these mechanisms or procedures validated?}
	\noindent The source domain data was collected using the CARLA simulator and its APIs with a resolution of $1280\times720$ pixels. The real-world 1/8th target domain was collected with a Stereolabs ZEDM camera with 30 FPS and a resolution of $1280\times720$ pixels. The lane distributions were additionally balanced with a bagging approach, and lanes were annotated with a labeling tool. More information can be found in the corresponding paper and the implementation. The implementation and all used tools are publicly available at \url{https://carlanebenchmark.github.io}.
\end{datasheetitem}
\begin{datasheetitem}{If the dataset is a sample from a larger set, what was the sampling strategy (e.g., deterministic, probabilistic with specific sampling probabilities)?}
	\noindent Source domain dataset entries are sampled based on the relative angle $\beta$ of the agent to the center lane. For MoLane, five lane classes are defined for the bagging approach: strong left curve ($\beta\leq$\ang{-45}), soft left curve (\ang{-45} $ < \beta \leq $ \ang{-15}), straight (\ang{-15} $ < \beta <$ \ang{15}), soft right curve (\ang{15} $ \leq \beta < $ \ang{45}) and strong right curve (\ang{45}$\leq \beta$). 
	\\\\
	For TuLane, three lane classes are defined for the bagging approach: left curve (\ang{-12} $ < \beta \leq$ \ang{5}), straight (\ang{-5} $ < \beta <$ \ang{5}) and right curve (\ang{5} $ \leq \beta < $ \ang{12}). 
\end{datasheetitem}
\begin{datasheetitem}{Who was involved in the data collection process (e.g., students, crowdworkers, contractors) and how were they compensated (e.g., how much were crowdworkers paid)?}
	\noindent Only the authors were involved in the collection process. The authors do not have information about the people involved in collecting the TuSimple dataset.
\end{datasheetitem}
\begin{datasheetitem}{Over what timeframe was the data collected? \normalfont Does this timeframe match the creation timeframe of the data associated with the instances (e.g., recent crawl of old news articles)? If not, please describe the timeframe in which the data associated with the instances was created.}
	\noindent MoLane's data was collected and annotated from June 2021 to August 2021. Data for TuLane's source domain was collected in February 2022.
\end{datasheetitem}
\begin{datasheetitem}{Were any ethical review processes conducted (e.g., by an institutional review board)? \normalfont If so, please provide a description of these review processes, including the outcomes, as well as a link or other access point to any supporting documentation.}
	\noindent No ethical reviews have been conducted to date. However, an ethical review may be conducted as part of the paper review process.
\end{datasheetitem}
\datasheetsection{Preprocessing/cleaning\\/labeling}	
\begin{datasheetitem}{Was any preprocessing/cleaning/labeling of the data done (e.g., discretization or bucketing, tokenization, part-of-speech tagging, SIFT feature extraction, removal of instances, processing of missing values)? \normalfont If so, please provide a description. If not, you may skip the remaining questions in this section.}
	\noindent As described above, lane annotations were labeled or cleaned using a labeling tool and sampled based on the relative angle $\beta$ of the agent to the center lane.
\end{datasheetitem}
\begin{datasheetitem}{Was the “raw” data saved in addition to the preprocessed/cleaned/labeled data (e.g., to support unanticipated future uses)? \normalfont If so, please provide a link or other access point to the “raw” data.}
	\noindent No.
\end{datasheetitem}
\begin{datasheetitem}{Is the software that was used to preprocess/clean/label the data available? \normalfont If so, please provide a link or other access point.}
	\noindent Yes, the software is available at \url{https://carlanebenchmark.github.io}.
\end{datasheetitem}
\datasheetsection{Uses}	
\begin{datasheetitem}{Has the dataset been used for any tasks already? \normalfont If so, please provide a description.}
	\noindent The datasets were used to create UDA baselines for the corresponding paper presenting the CARLANE Benchmark.
\end{datasheetitem}
\begin{datasheetitem}{Is there a repository that links to any or all papers or systems that use the dataset? \normalfont If so, please provide a link or other access point.}
	\noindent Yes, the baselines presented in the corresponding paper are available at \url{https://carlanebenchmark.github.io}.
\end{datasheetitem}
\begin{datasheetitem}{What(other) tasks could the dataset be used for?}
	\noindent In a broader sense, the datasets of CARLANE can also be used for unsupervised and semi-supervised learning and partially for supervised learning.
\end{datasheetitem}
\begin{datasheetitem}{Is there anything about the composition of the dataset or the way it was collected and preprocessed/cleaned/labeled that might impact future uses? \normalfont For example, is there anything that a dataset consumer might need to know to avoid uses that could result in unfair treatment of individuals or groups (e.g., stereotyping, quality of service issues) or other risks or harms (e.g., legal risks, financial harms)? If so, please provide a description. Is there anything a dataset consumer could do to mitigate these risks or harms?}
	\noindent Yes, TuLane and MuLane contain open-source images with unblurred license\\plates and people. This data should be treated with respect and in accordance with privacy policies. In general, CARLANE contributes to the research in the field of autonomous driving, in which many unresolved ethical and legal questions are still being discussed. The step-by-step testing possibility across three domains makes it possible for our benchmark to include an additional safety mechanism for real-world testing. This can help the consumer to mitigate the risks and harms to some extent.
\end{datasheetitem}
\begin{datasheetitem}{Are there tasks for which the dataset should not be used? \normalfont If so, please provide a description.}
	\noindent Since CARLANE focuses on UDA for lane detection and spans a limited number of driving scenarios, consumers should not solely really on this dataset to train models for fully autonomous driving. 
\end{datasheetitem}
\datasheetsection{Distribution}	
\begin{datasheetitem}{Will the dataset be distributed to third parties outside of the entity (e.g., company, institution, organization) on behalf of which the dataset was created? \normalfont If so, please provide a description.}
	\noindent Yes, CARLANE is publicly available on the internet for anyone interested in using it.
\end{datasheetitem}
\begin{datasheetitem}{How will the dataset will be distributed (e.g., tarball on website, API, GitHub)? \normalfont Does the dataset have a digital object identifier (DOI)?}
	\noindent CARLANE is distributed through kaggle at \href{https://www.kaggle.com/datasets/carlanebenchmark/carlane-benchmark}{https://www.kaggle.com/datasets/\\carlanebenchmark/carlane-benchmark}
	\\\\
	DOI: 10.34740/kaggle/dsv/3798459
\end{datasheetitem}
\begin{datasheetitem}{When will the dataset be distributed?}
	\noindent The datasets have been available on kaggle since June 17, 2022.
\end{datasheetitem}
\begin{datasheetitem}{Will the dataset be distributed under a copyright or other intellectual property (IP) license, and/or under applicable terms o fuse (ToU)? \normalfont If so, please describe this license and/or ToU, and provide a link or other access point to, or otherwise reproduce, any relevant licensing terms or ToU, as well as any fees associated with these restrictions.}
	\noindent CARLANE is licensed under the Apache License Version 2.0, January 2004.
\end{datasheetitem}
\begin{datasheetitem}{Have any third parties imposed IP-based or other restrictions on the data associated with the instances? \normalfont If so, please describe these restrictions, and provide a link or other access point to, or otherwise reproduce, any relevant licensing terms, as well as any fees associated with these restrictions.}
	\noindent TuSimple, which is used for TuLanes and MuLanes target domains, is licensed under the Apache License Version 2.0, January 2004.
\end{datasheetitem}
\begin{datasheetitem}{Do any export controls or other regulatory restrictions apply to the dataset or to individual instances? \normalfont If so, please describe these restrictions, and provide a link or other access point to, or otherwise reproduce, any supporting documentation.}
	\noindent Unknown to authors of the datasheet.
\end{datasheetitem}
\datasheetsection{Maintenance}	
\begin{datasheetitem}{Who will be supporting/hosting/maintaining the dataset?}
	\noindent CARLANE is hosted on kaggle and supported and maintained by the authors.
\end{datasheetitem}
\begin{datasheetitem}{How can the owner/curator/manager of the dataset be contacted (e.g., email address)?}
	\noindent The curators of the datasets can be contacted under\\carlane.benchmark@gmail.com.
\end{datasheetitem}
\begin{datasheetitem}{Is there an erratum? \normalfont If so, please provide a link or other access point.}
	\noindent No. 
\end{datasheetitem}
\begin{datasheetitem}{Will the dataset be updated (e.g., to correct labeling errors, add new instances, delete instances)? \normalfont If so, please describe how often, by whom, and how updates will be communicated to dataset consumers (e.g., mailing list, GitHub)?}
	\noindent New versions of CARLANE's datasets will be shared and announced on our homepage (\url{https://carlanebenchmark.github.io}) and at kaggle if corrections are necessary.
\end{datasheetitem}
\begin{datasheetitem}{Will older versions of the dataset continue to be supported/hosted/maintained? \normalfont If so, please describe how. If not, please describe how its obsolescence will be communicated to dataset consumers.}
	\noindent Yes, we plan to support versioning of the datasets so that all the versions are available to potential users. We maintain the history of versions via our homepage\\ (\url{https://carlanebenchmark.github.io})\\
	and at kaggle. Each version will have a unique DOI assigned.
\end{datasheetitem}
\begin{datasheetitem}{If others want to extend/augment/build on/contribute to the dataset, is there a mechanism for them to do so? \normalfont If so, please provide a description. Will these contributions be validated/verified? If so, please describe how. If not, why not? Is there a process for communicating/distributing these contributions to dataset consumers? If so, please provide a description.}
	\noindent Others can extend/augment/build on CARLANE with the support of the open-source tools provided on our homepage. Besides these tools, there will be no mechanism to validate or verify the extended datasets. However, others are free to release their extension of the CARLANE Benchmark or its datasets under the Apa-che License Version 2.0.
\end{datasheetitem}
\onecolumn

\chapter{Supplementary Material:\\Content-Consistent Translation with Masked Discriminators}
\label{app:03}

\section{The FeaMGAN Architecture}
\begin{figure}[!htb]
	\begin{center} 
		\includegraphics[width=1.0\linewidth]{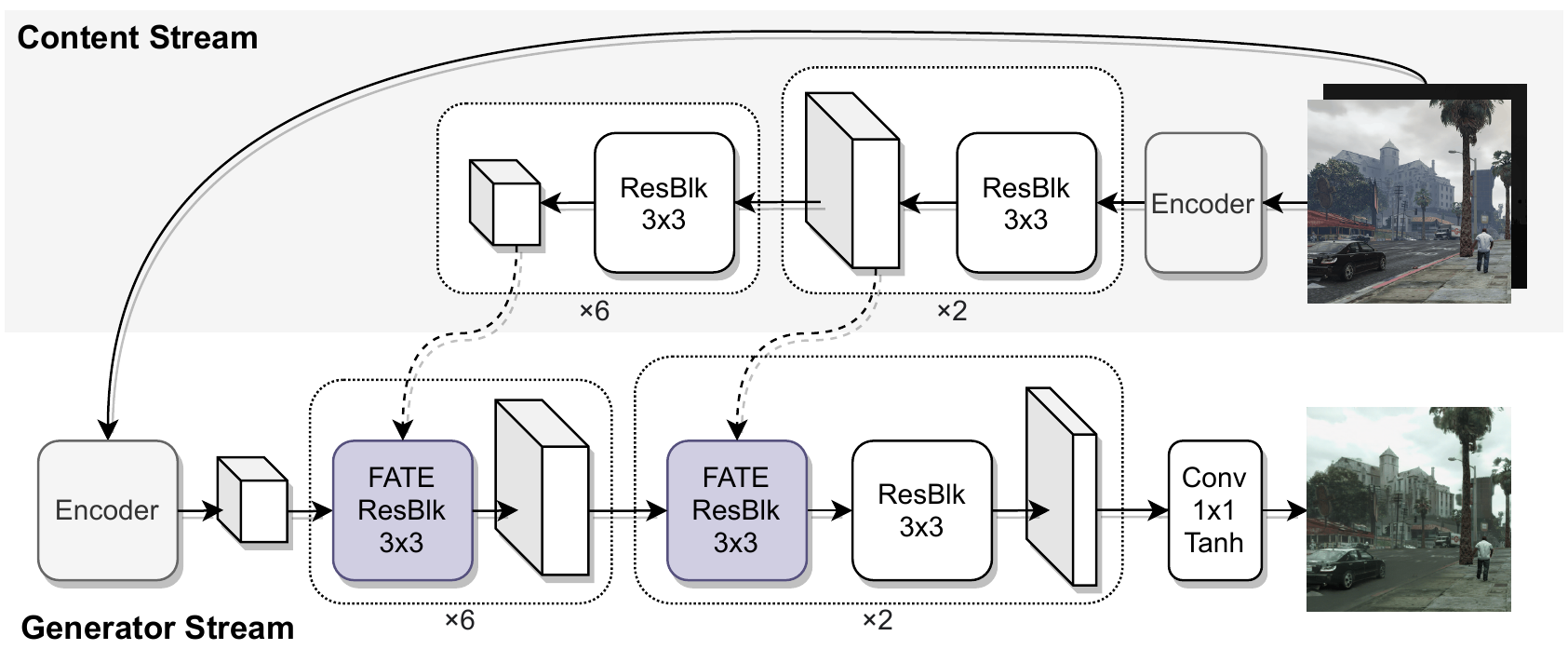}
	\end{center}	
	\vspace{-1ex}
	\caption[Generator architecture.]{Generator architecture. Arrows with dashed lines indicate connections at multiple levels between the two streams. Best viewed in color.}
	\label{fig:feamgan:app:feam_generator}
\end{figure}
\begin{figure}[!htb]
	\begin{center} 
		\includegraphics[width=0.45\linewidth]{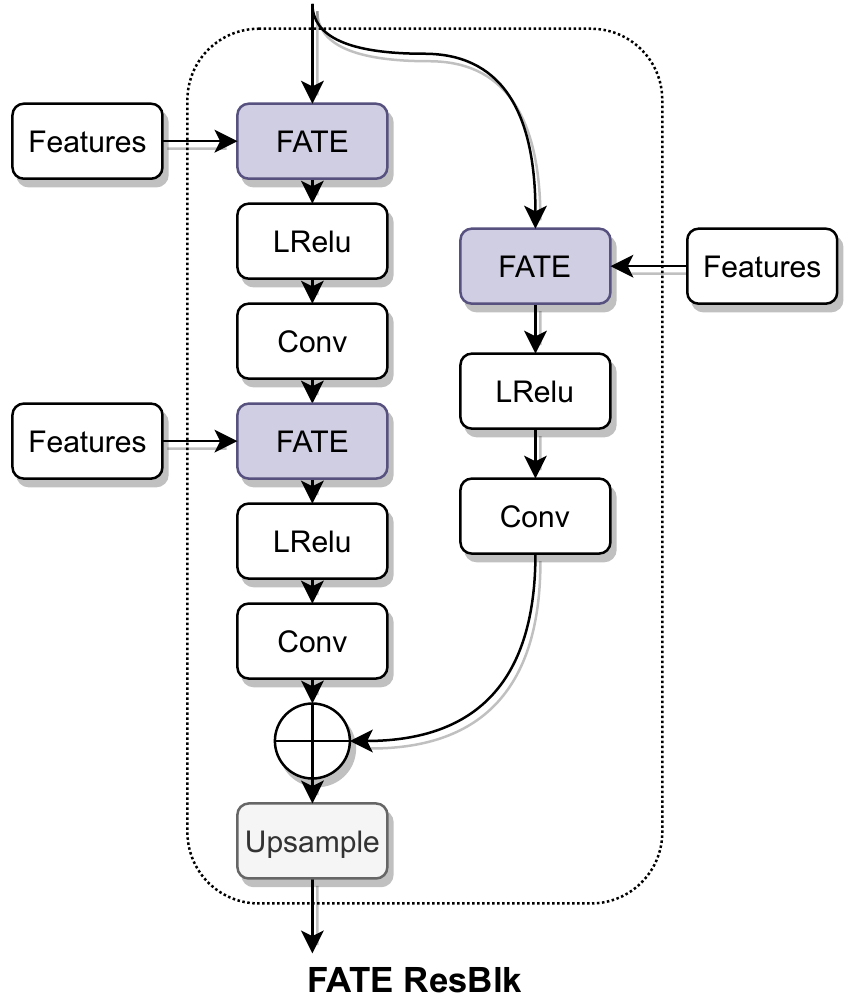}
	\end{center}
	\vspace{-1ex}
	\caption[FATE residual block.]{FATE residual block. The FATE residual block used in the generator stream.}
	\label{fig:feamgan:app:fate_res_blk}
\end{figure}
\begin{figure}[!htb]
	\begin{center} 
		\includegraphics[width=0.45\linewidth]{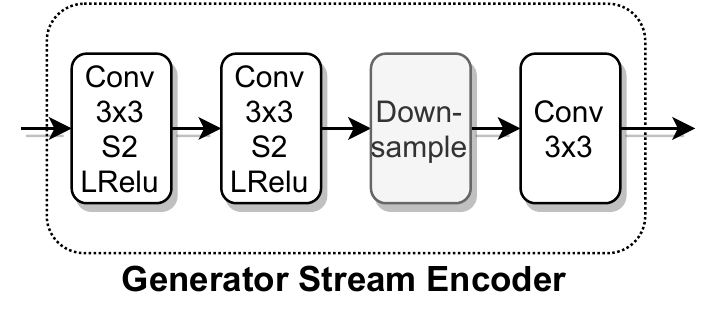}
	\end{center}
	\vspace{-1ex}
	\caption[Generator stream encoder.]{Generator stream encoder. The generator stream encoder used to encode the input image and condition for the generator stream.}
	\label{fig:feamgan:app:generator_stream_encoder}
\end{figure}
\begin{figure}[!htb]
	\begin{center} 
		\includegraphics[width=0.3\linewidth]{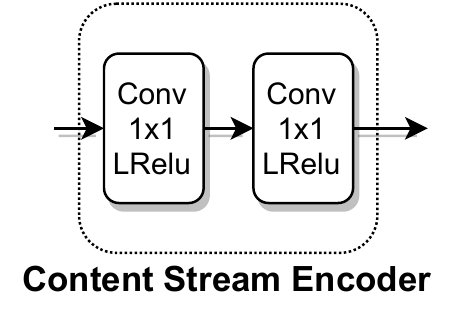}
	\end{center}		
	\vspace{-1ex}
	\caption[Content stream encoder.]{Content stream encoder. The content stream encoder used to encode the input image and condition for the content stream.}
	\label{fig:feamgan:app:content_stream_encoder}
\end{figure}

\noindent \textbf{Generator}. As shown in \autoref{fig:feamgan:app:feam_generator}, our generator consists of a content stream encoder, a content stream, a generator stream encoder, and a generator stream. The content stream encoder shown in \autoref{fig:feamgan:app:content_stream_encoder} is utilized to create the initial features of the source image and condition. These initial features are the input to the content stream, which creates features for multiple levels with residual blocks. The statistics of these features are then integrated into the generator at multiple levels utilizing the residual FATE blocks shown in \autoref{fig:feamgan:app:fate_res_blk}. The generator stream utilizes the encoder shown in \autoref{fig:feamgan:app:generator_stream_encoder} to create the initial latent from which the target image is generated. To further enforce content consistency, we do not use a variational autoencoder to obtain an deterministic latent. In addition, we found that utilizing additional residual blocks in the last layers of the generator stream improves performance, likely due to further refinement of the preceding upsampled features. We use spectral instance normalization for the residual blocks in the content stream and spectral batch normalization for the residual blocks in the generator stream. The convolutional layers in the generator stream encoder have the following numbers of filters: $[256,512,1024]$. The residual blocks in the generator have the following numbers of filters: $[1024,1024,1024,512,256,128,64,64,64,64]$. The numbers of filters of the convolutional layers in the content streams encoder are $[64,64]$. The numbers of filters in the content stream match those of the output of the preceding residual block in the generator stream at the respective level: $[64,128,256,512,1024,1024,1024,1024]$. For all residual blocks, we use $3\times3$ convolutions and $1\times1$ convolutions for the skip connections. $\gamma$ and $\beta$ in the FATE and FADE blocks are created with $3\times3$ convolutions. Throughout the generator, we use a padding of $1$ for the convolutions - we only downsample with strides and downsampling layers. We utilize the "nearest" upsampling and downsampling from Pytorch. For our small model, we halve the number of filters.

\begin{figure}[!htb]
	\begin{center} 
		\includegraphics[width=0.38\linewidth]{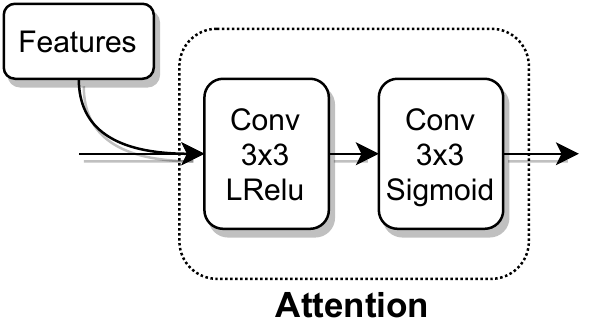}
	\end{center}
	\vspace{-1ex}
	\caption[Attention module.]{Attention module. The attention module used in the FATE block to attend to the statistics of the features.}
	\label{fig:feamgan:app:attention_module}
\end{figure}
\clearpage
\begin{figure}[!htb]
	\begin{center}
		\includegraphics[width=1.0\linewidth]{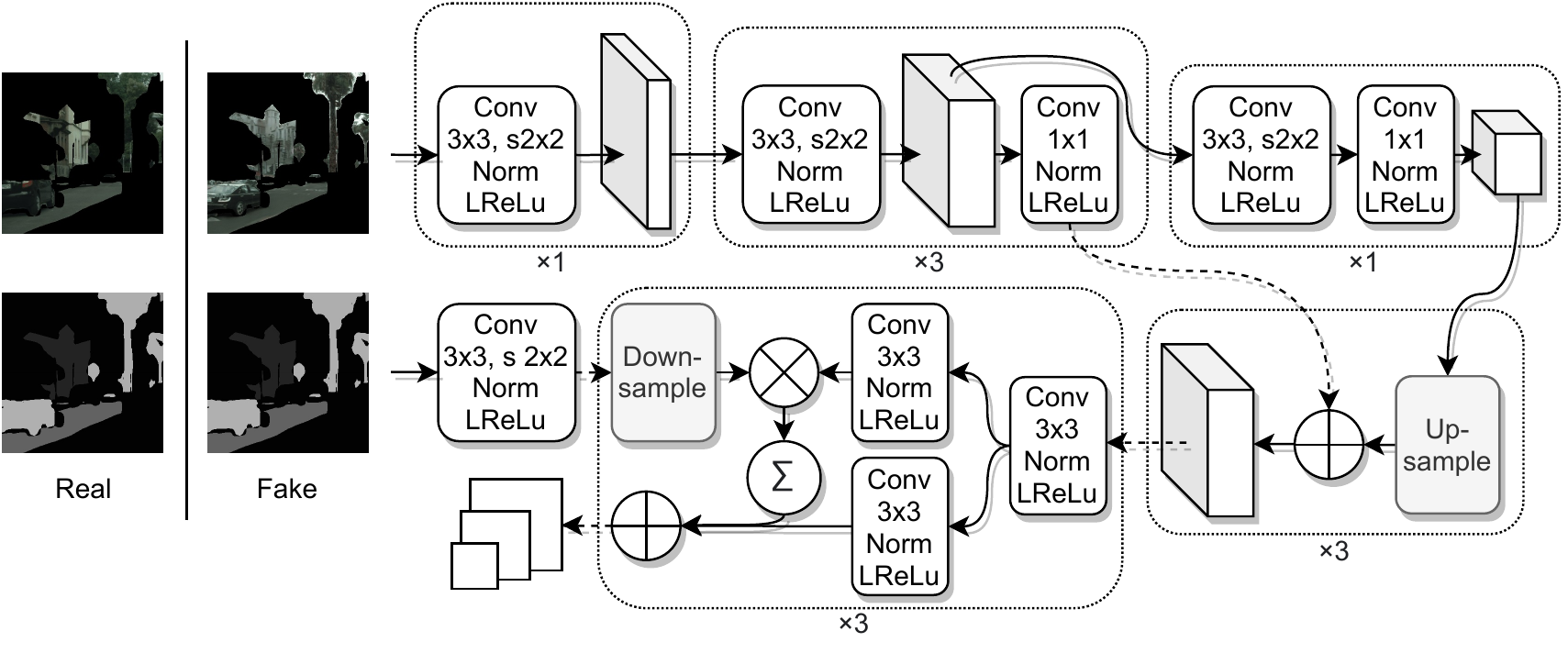}
	\end{center}
	\vspace{-1ex}
	\caption[Discriminator architecture.]{Discriminator architecture. Arrows with dashed lines indicate connections at multiple levels between the two components. Best viewed in color.}
	\label{fig:feamgan:app:feam_discriminator}
\end{figure}
\noindent \textbf{Discriminator}. As shown in \autoref{fig:feamgan:app:feam_discriminator}, our discriminator consists of downsampling, upsampling, and prediction components. First, the input images of the source or target domain are downsampled via $5$ stride $2$ convolutions. We transform the output feature map of the last $4$ downsampling convolutions with $1\times1$ convolutions. The last transformed feature map is used as input for the upsampling components, while the other transformed feature maps are added to the feature maps of the upsampling component for the receptive level. Then we utilize the feature maps of the $3$ upsampling levels to create the final prediction on $3$ levels. Thereby, we first apply a convolutional layer on the upsampled features. This convolution is followed by two convolutional layers: One is used to create the prediction feature map of depth $1$, and the feature map of the other convolutional layer is multiplied by the segmentation map. The resulting segmentation feature map is then collapsed into depth $1$ by adding the depth dimensions together. At last, the collapsed segmentation feature map is added to the prediction feature map to produce the final prediction. In this way, the discriminator is encouraged to produce class-specific predictions. We use spectral instance normalization for all convolutional layers. The $3\times3$ convolutions of the downsampling component have the following numbers of filters: $[64,128,256,512,512]$. The $1\times1$ convolutions of the downsampling component have the following numbers of filters: $[256,256,256,256]$. The first convolutions in the prediction component have the following numbers of filters: $[128,128,128]$. The convolutions that are multiplied by the downsampled segmentation maps have the following numbers of filters: $[128,128,128]$. The convolution used to create the downsampled segmentation map has $128$ filters. The convolutions to create the predictions have the following numbers of filters: $[1,1,1]$. Throughout the discriminator, we use a padding of $1$ for the convolutions - we only downsample with strides and downsampling layers. We utilize the "bilinear" upsampling and downsampling from Pytorch. For our small model, we halve the number of filters.\\\\

\section{Additional Dataset Details}
In \autoref{tab:feamgan:app:training_details}, we show additional details about the used datasets. Since we compare our method i.a. to the transferred PFD \cite{richter2016playing} images provided by EPE \cite{richter2022enhancing}, we use a base image height of $526$ throughout our experiments. The aspect ratio is preserved when resizing the input images. To match the input sizes of the images of both domains, we apply cropping to the image with the larger width if the image sizes of both domains do not align. The resulting images are randomly flipped before the sampling strategy is applied. 

\section{Additional Training Details}
In \autoref{tab:feamgan:app:datasets_details}, we show additional details about the hyperparameters used for training the four translation tasks. No tuning was performed for other translation tasks then PFD$\rightarrow$Cityscapes besides adapting the learning rate schedule for the dataset lengths of these tasks.

\section{Additional Results}
\begin{figure}[h] 
	\captionsetup[subfigure]{labelformat=empty}
	\begin{center}
		{\includegraphics[width=0.33\textwidth]{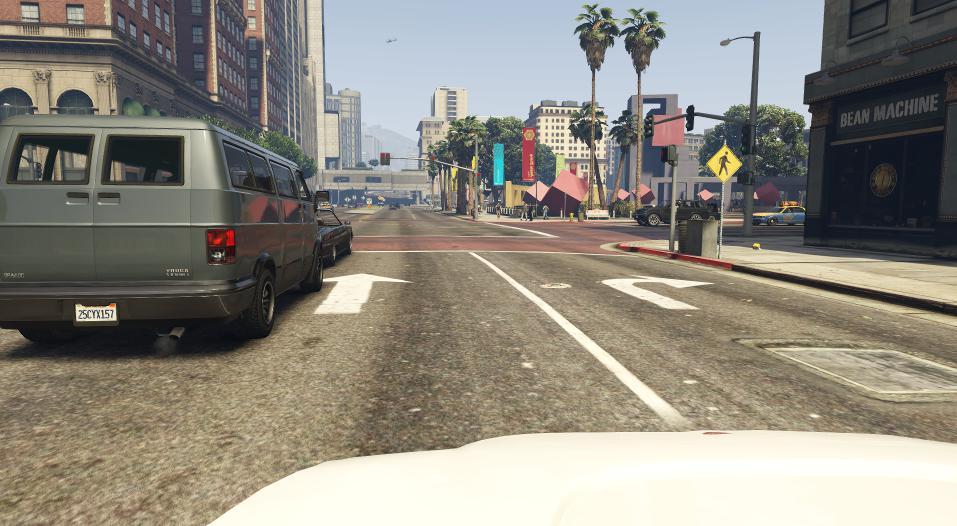}}\hfill
		{\includegraphics[width=0.33\textwidth]{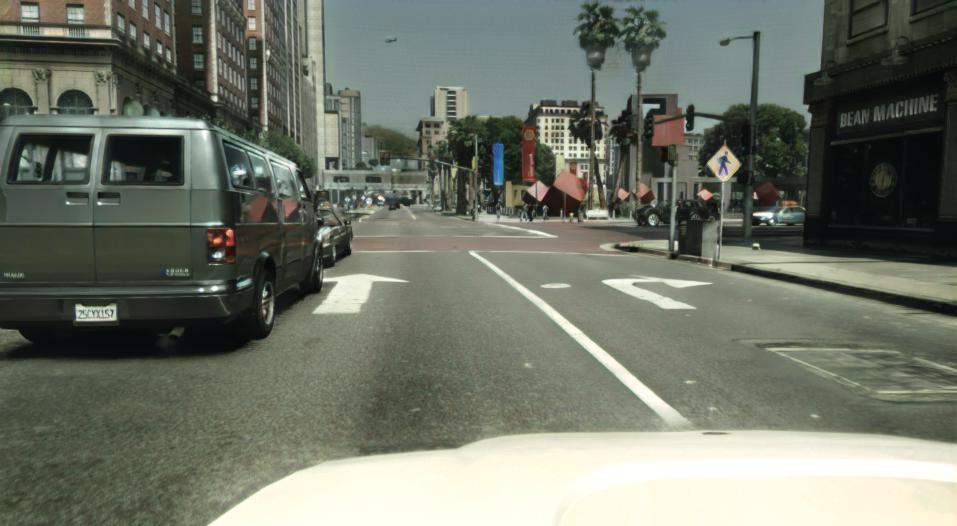}}\hfill
		{\includegraphics[width=0.33\textwidth]{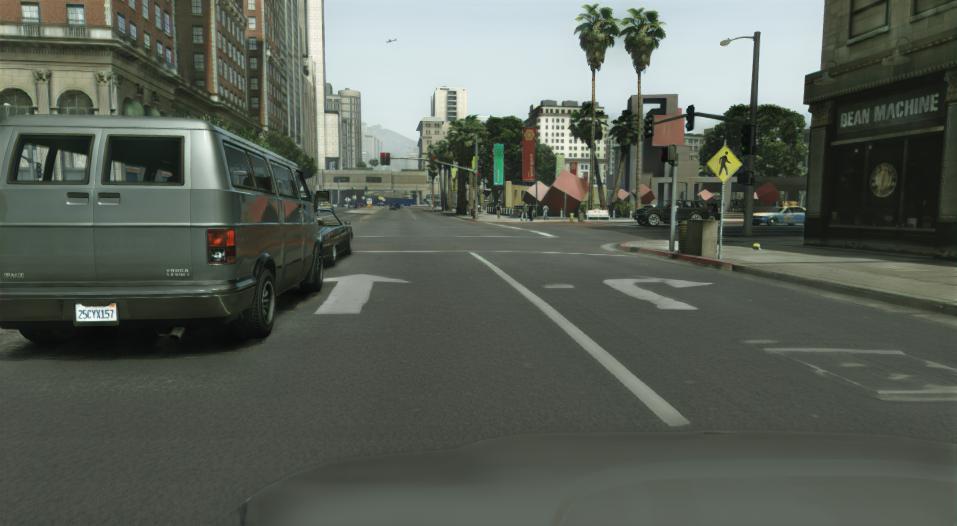}}\hfill\\\vspace{1pt}
		{\includegraphics[width=0.33\textwidth]{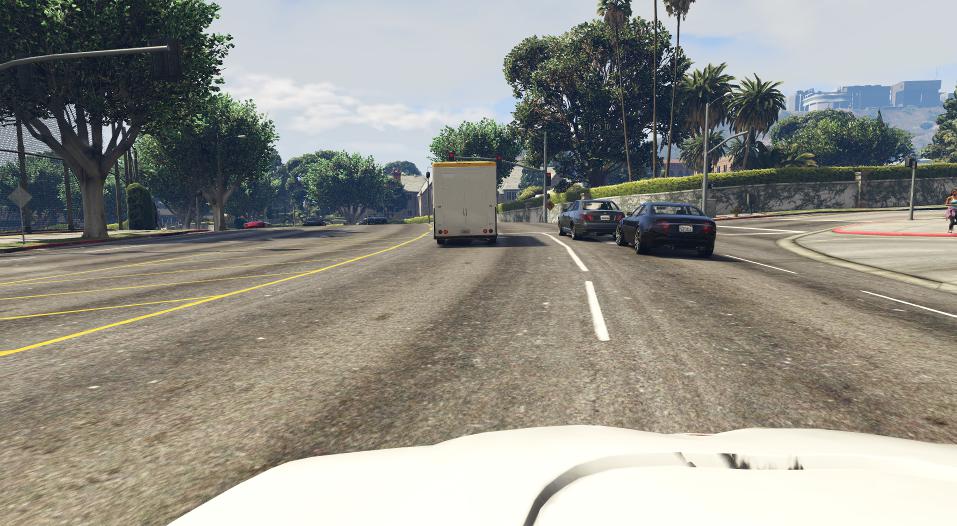}}\hfill
		{\includegraphics[width=0.33\textwidth]{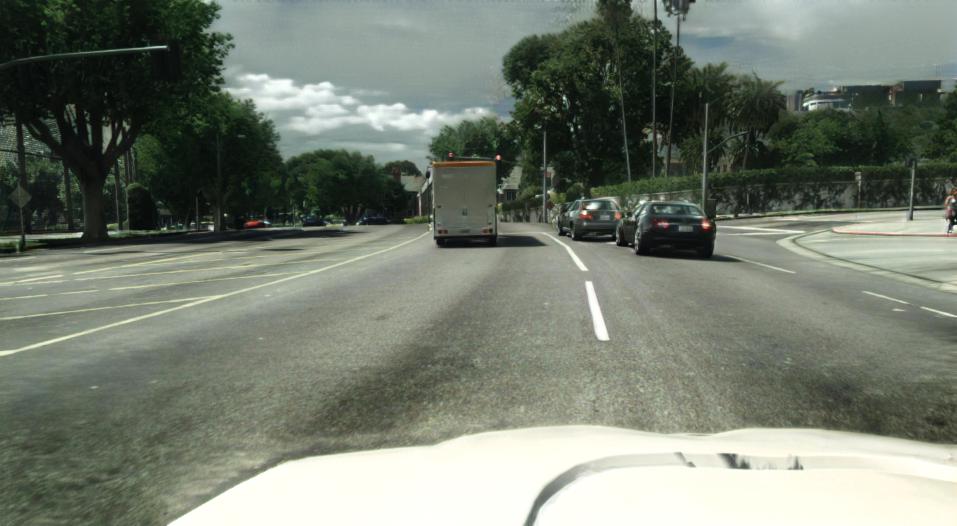}}\hfill
		{\includegraphics[width=0.33\textwidth]{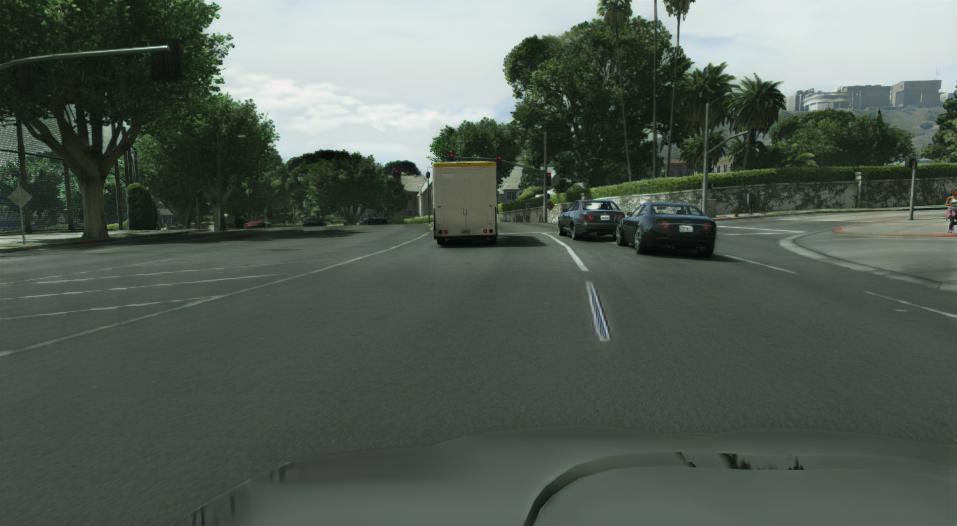}}\hfill\\\vspace{1pt}
		{\includegraphics[width=0.33\textwidth]{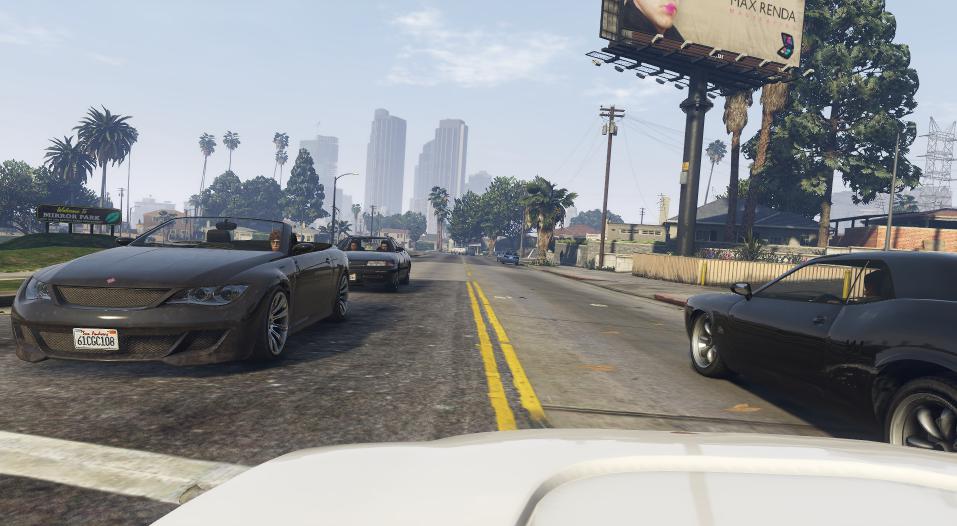}}\hfill
		{\includegraphics[width=0.33\textwidth]{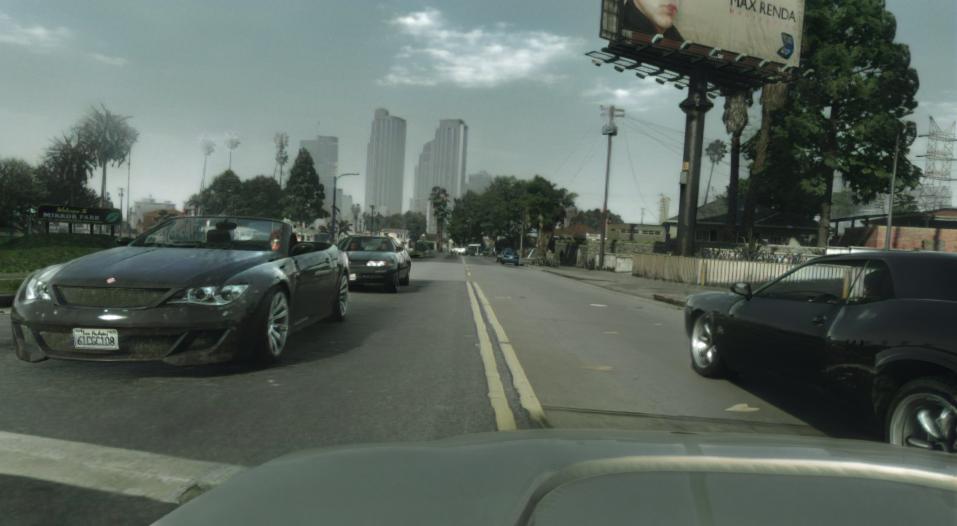}}\hfill
		{\includegraphics[width=0.33\textwidth]{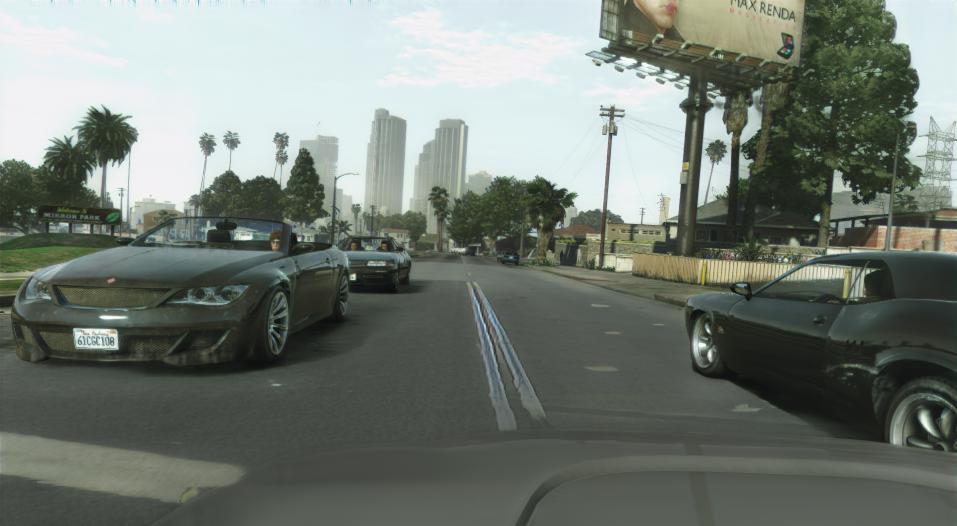}}\hfill\\\vspace{1pt}
		{\includegraphics[width=0.33\textwidth]{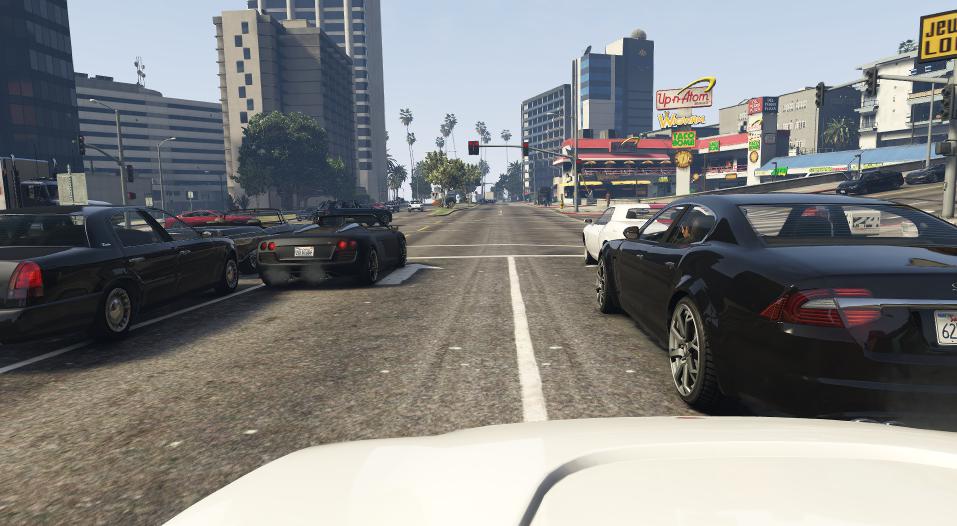}}\hfill
		{\includegraphics[width=0.33\textwidth]{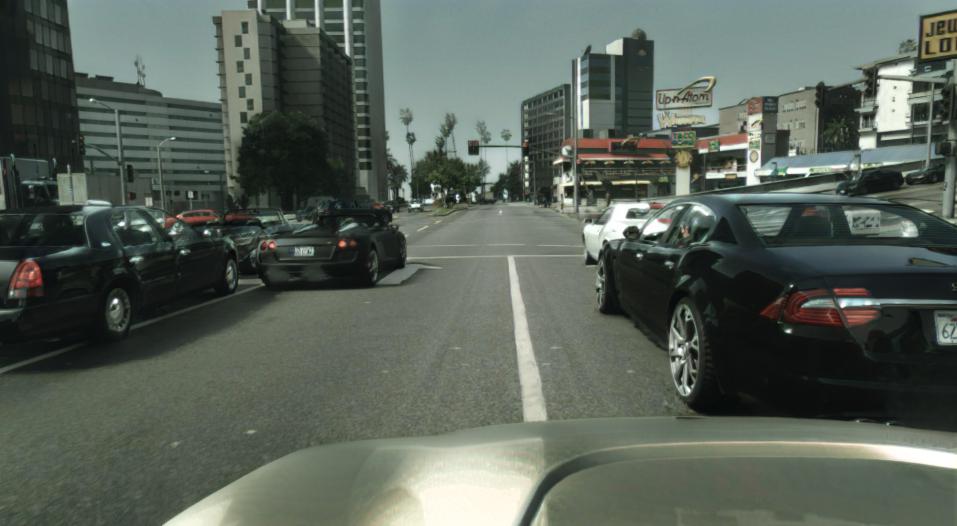}}\hfill
		{\includegraphics[width=0.33\textwidth]{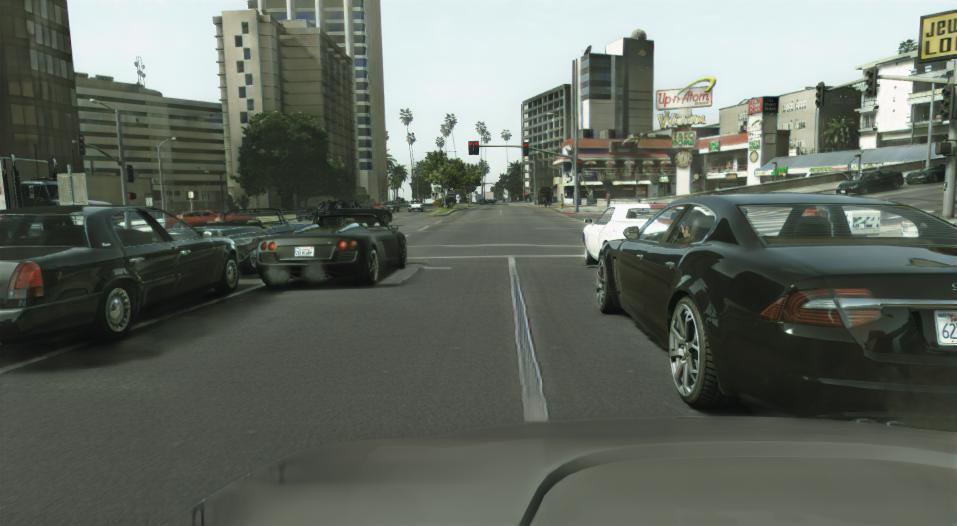}}\hfill\\\vspace{1pt}
		{\includegraphics[width=0.33\textwidth]{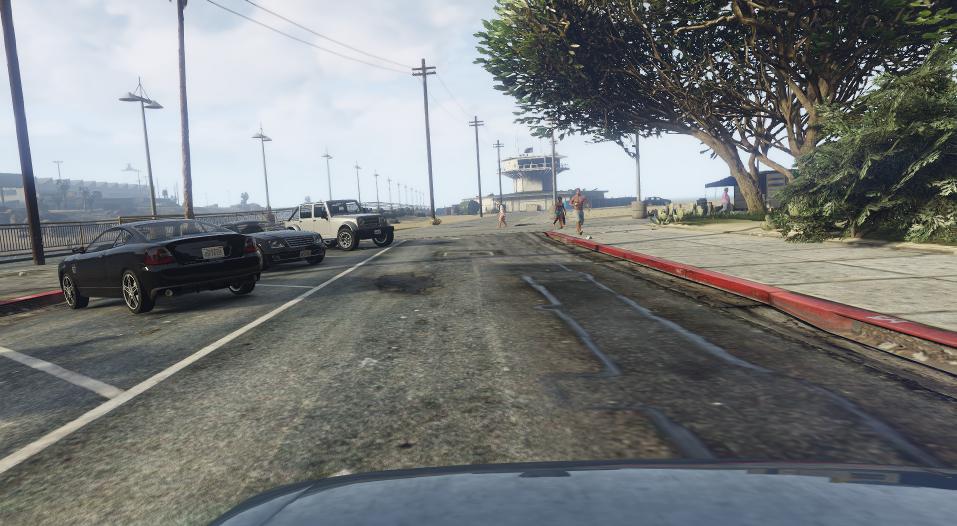}}\hfill
		{\includegraphics[width=0.33\textwidth]{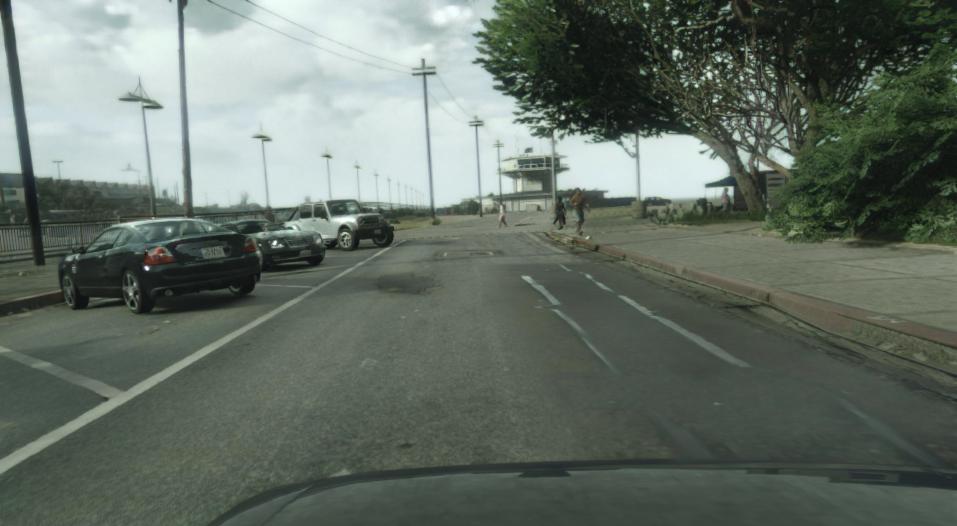}}\hfill
		{\includegraphics[width=0.33\textwidth]{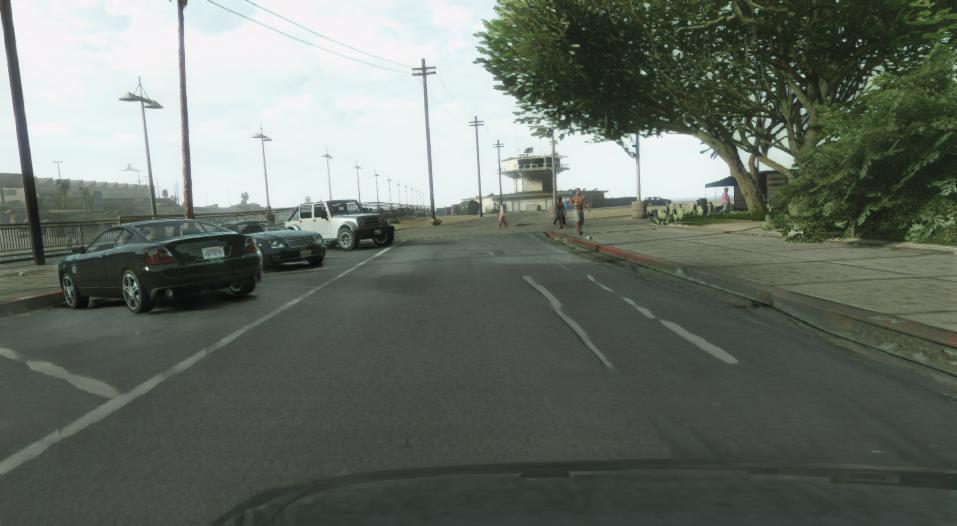}}\hfill\\\vspace{-9pt}
		\subfloat[Input]
		{\includegraphics[width=0.33\textwidth]{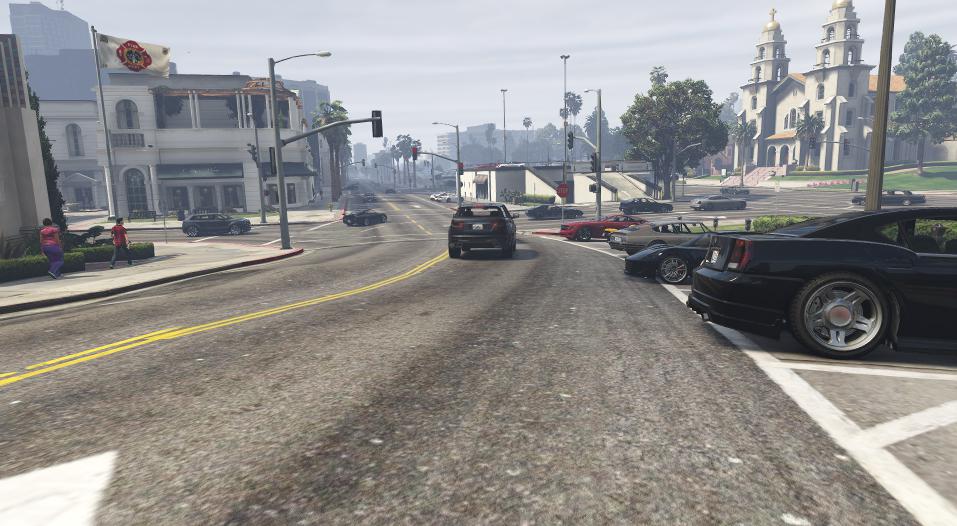}}\hfill
		\subfloat[EPE]
		{\includegraphics[width=0.33\textwidth]{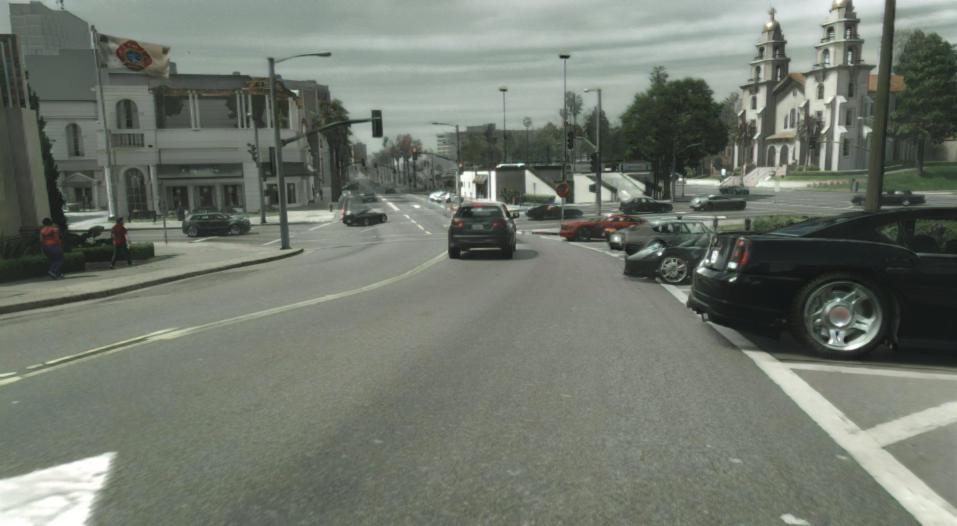}}\hfill
		\subfloat[FeaMGAN (ours)]
		{\includegraphics[width=0.33\textwidth]{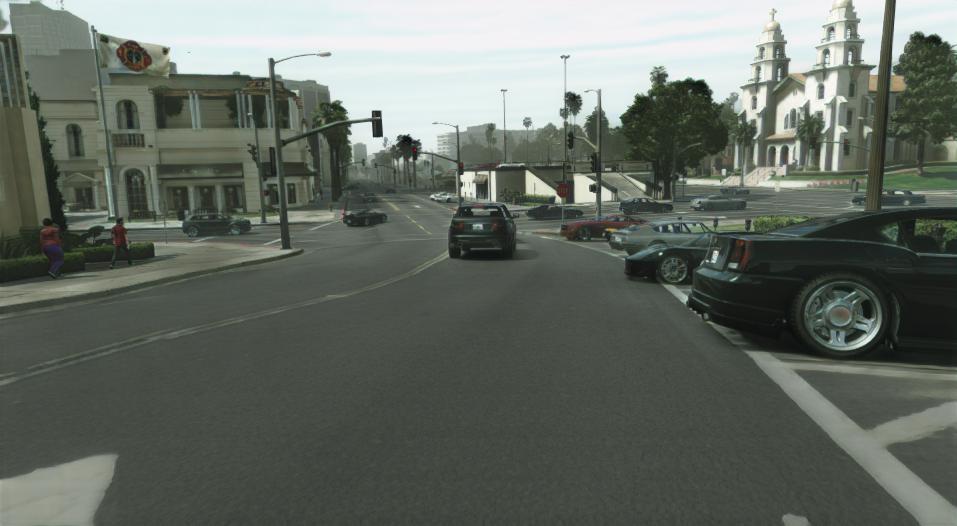}}\hfil 
	\end{center}
	\vspace{-1ex}
	\caption[Qualitative comparison to EPE.]{Qualitative comparison to EPE. We compare our method with the provided inferred images of EPE \cite{richter2022enhancing}. Best viewed in color.}
	\label{fig:feamgan:app:qualitative_comparison_epe_additional}
\end{figure}

\begin{figure}[h] 
	\captionsetup[subfigure]{labelformat=empty}
	\begin{center}
		{\includegraphics[width=0.33\textwidth]{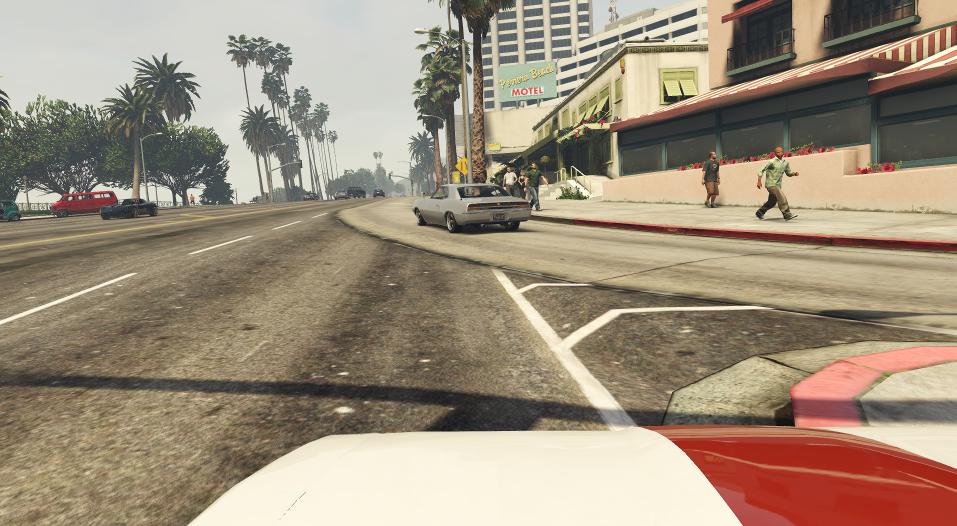}}\hfill
		{\includegraphics[width=0.33\textwidth]{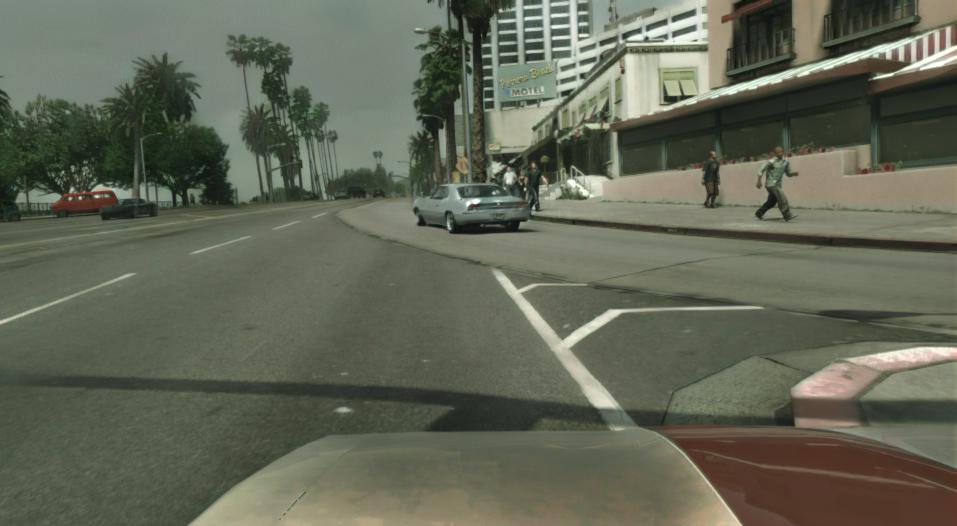}}\hfill
		{\includegraphics[width=0.33\textwidth]{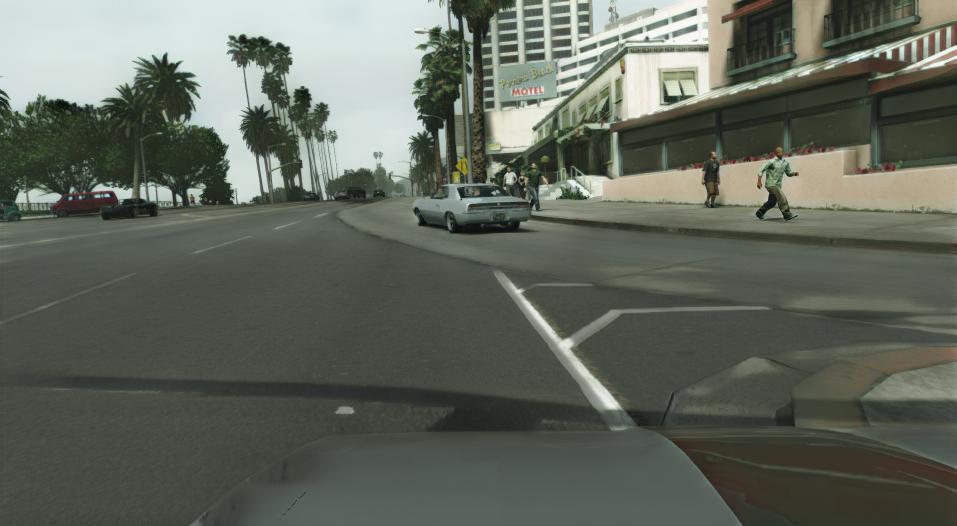}}\hfill\\\vspace{1pt}
		{\includegraphics[width=0.33\textwidth]{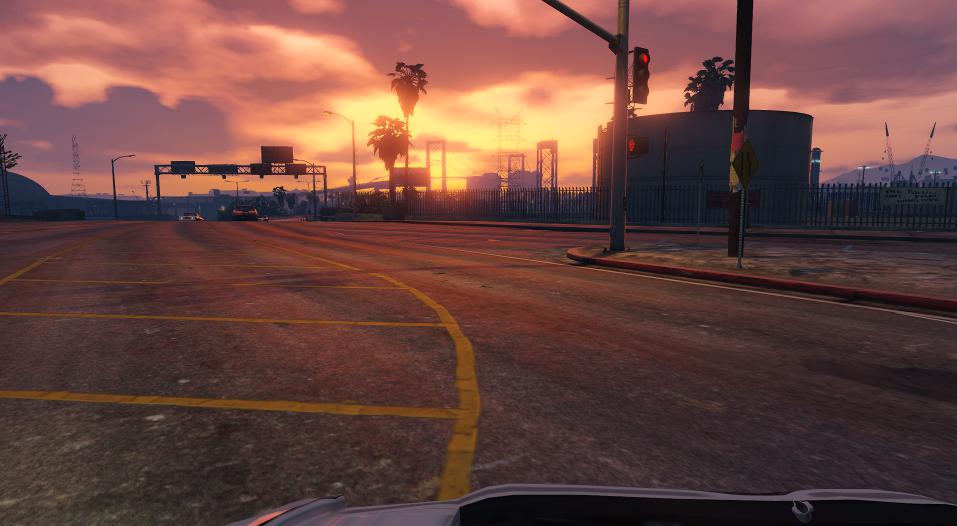}}\hfill
		{\includegraphics[width=0.33\textwidth]{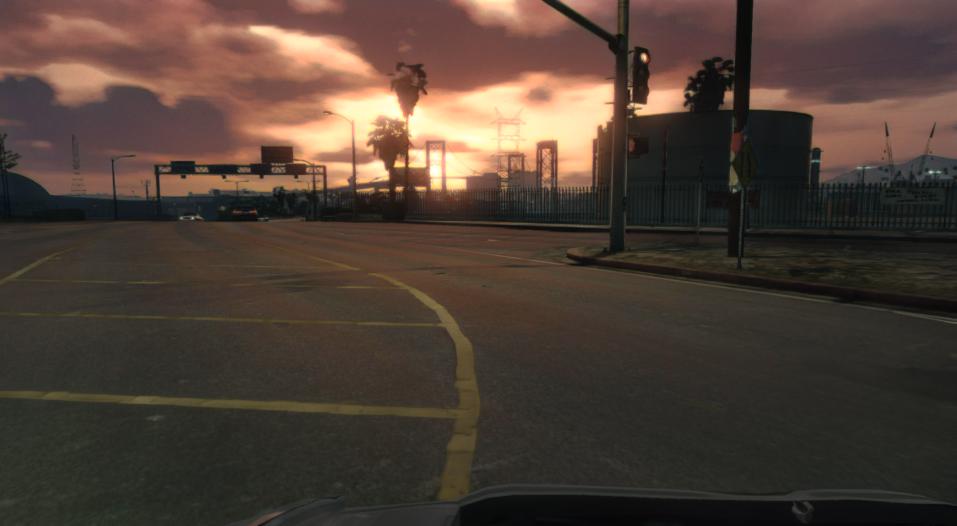}}\hfill
		{\includegraphics[width=0.33\textwidth]{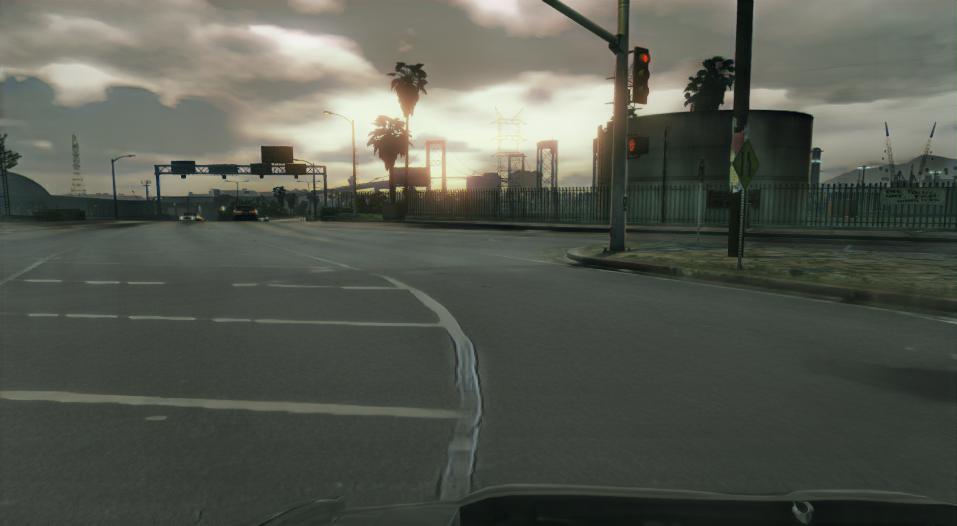}}\hfill\\\vspace{1pt}
		{\includegraphics[width=0.33\textwidth]{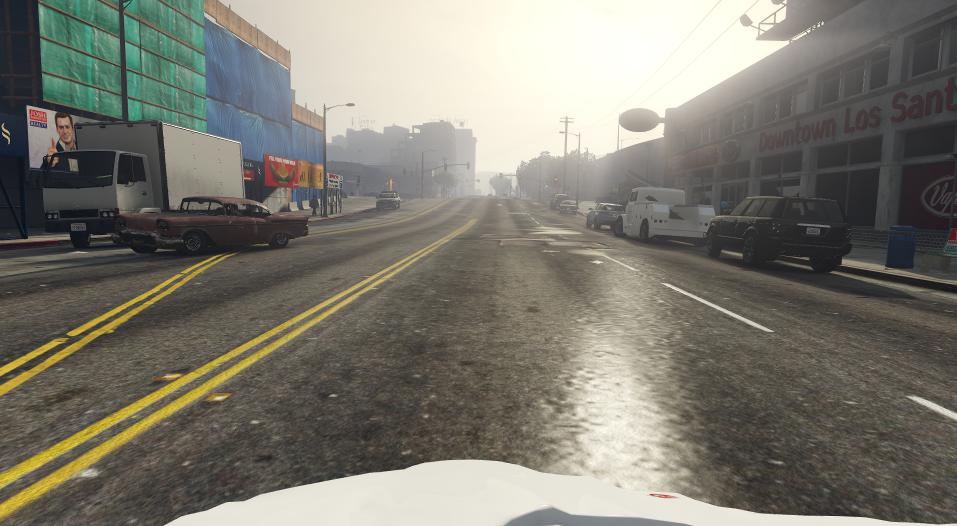}}\hfill
		{\includegraphics[width=0.33\textwidth]{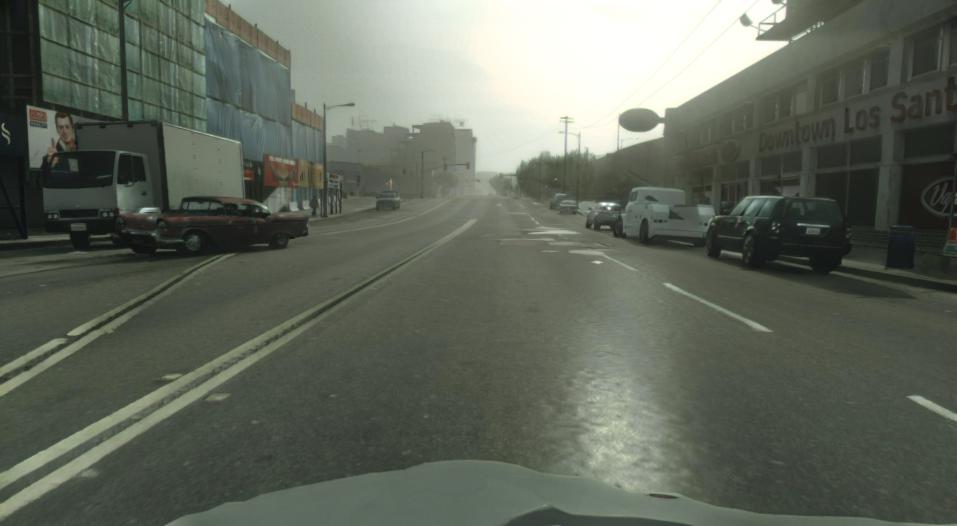}}\hfill
		{\includegraphics[width=0.33\textwidth]{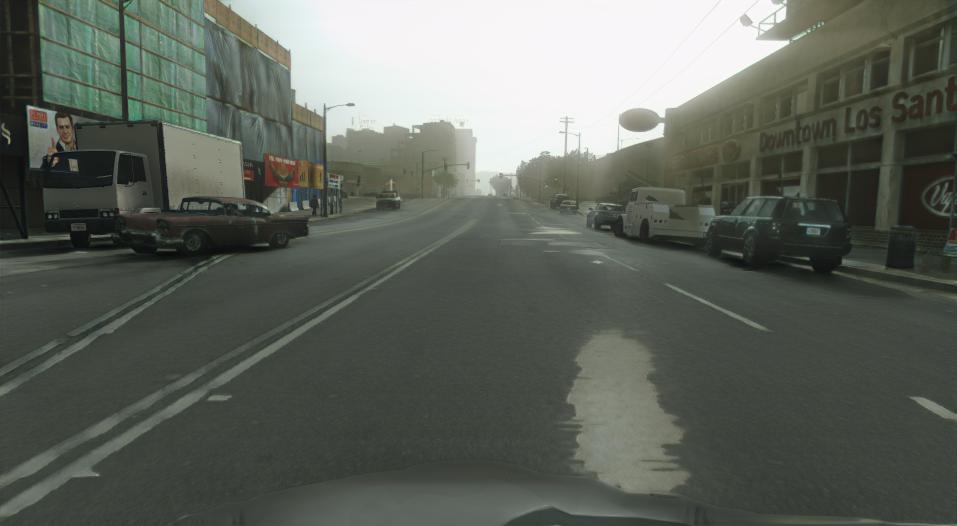}}\hfill\\\vspace{1pt}
		{\includegraphics[width=0.33\textwidth]{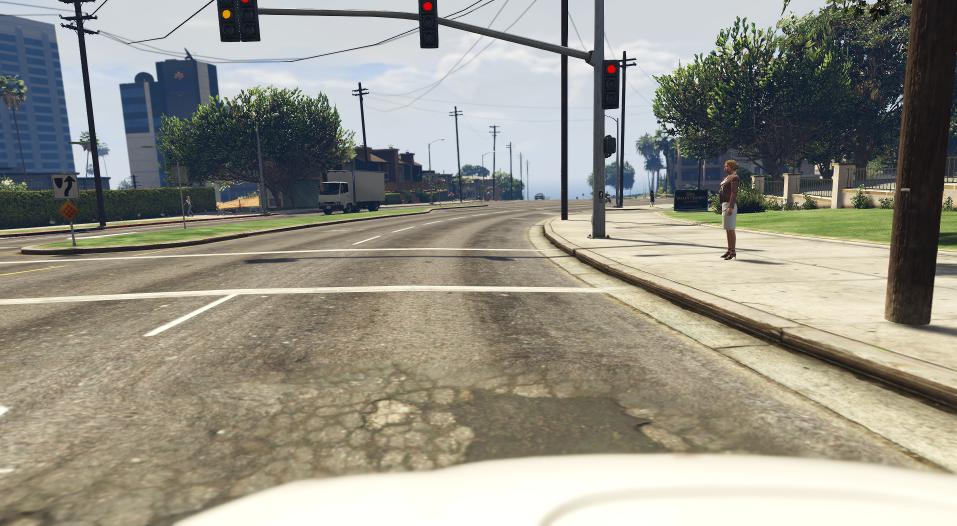}}\hfill
		{\includegraphics[width=0.33\textwidth]{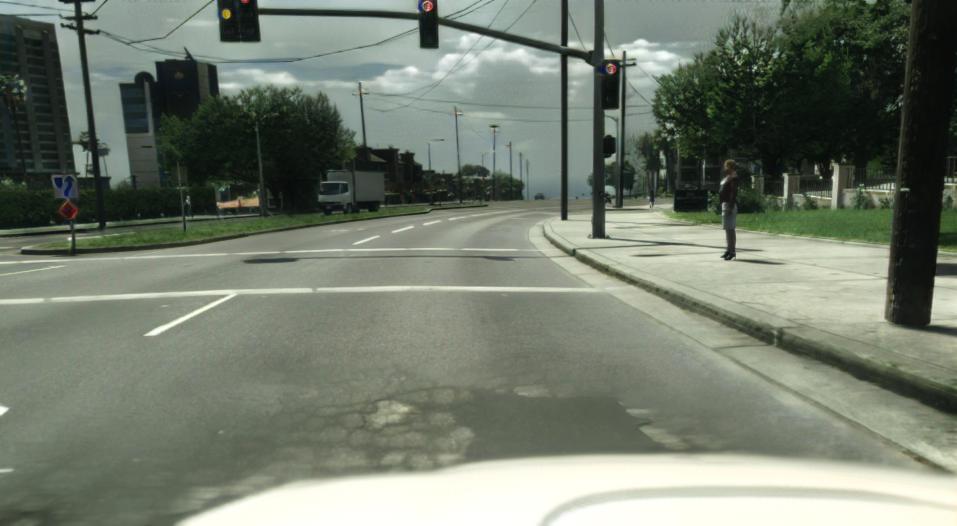}}\hfill
		{\includegraphics[width=0.33\textwidth]{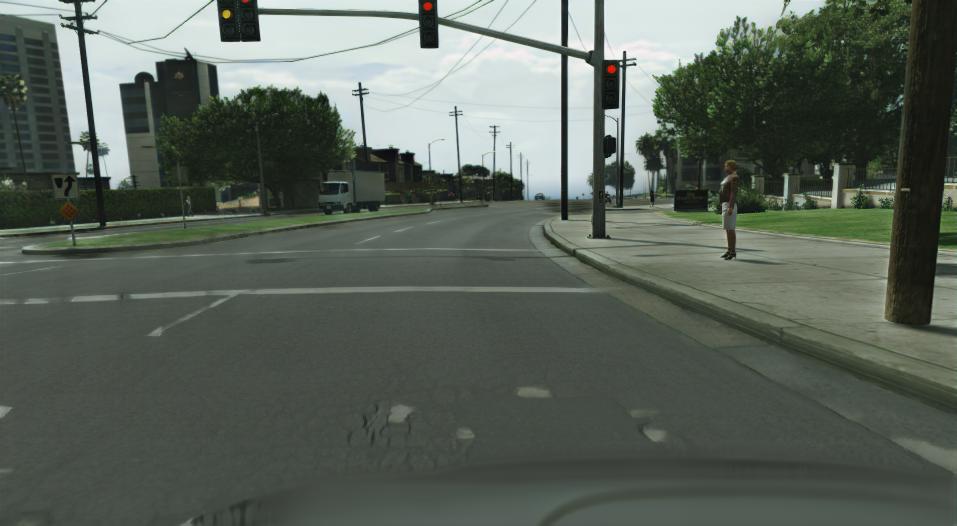}}\hfill\\\vspace{1pt}
		{\includegraphics[width=0.33\textwidth]{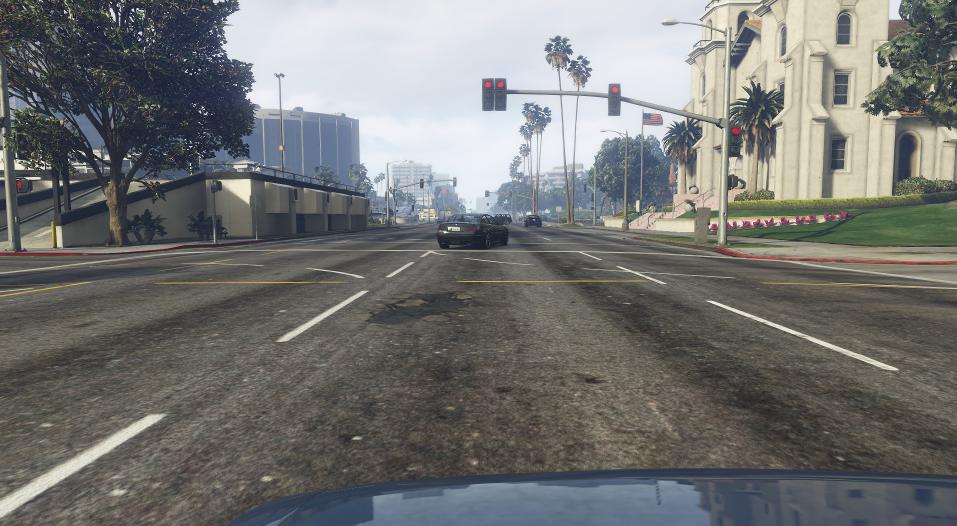}}\hfill
		{\includegraphics[width=0.33\textwidth]{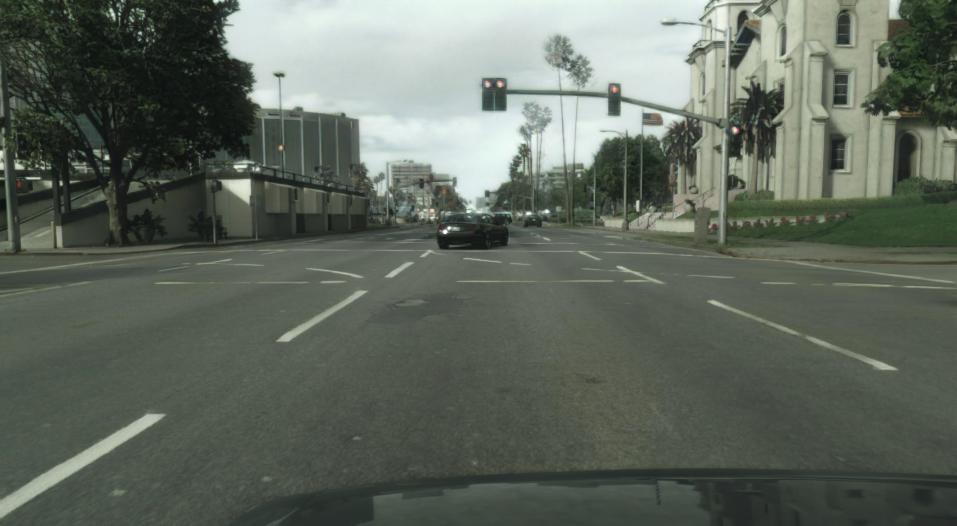}}\hfill
		{\includegraphics[width=0.33\textwidth]{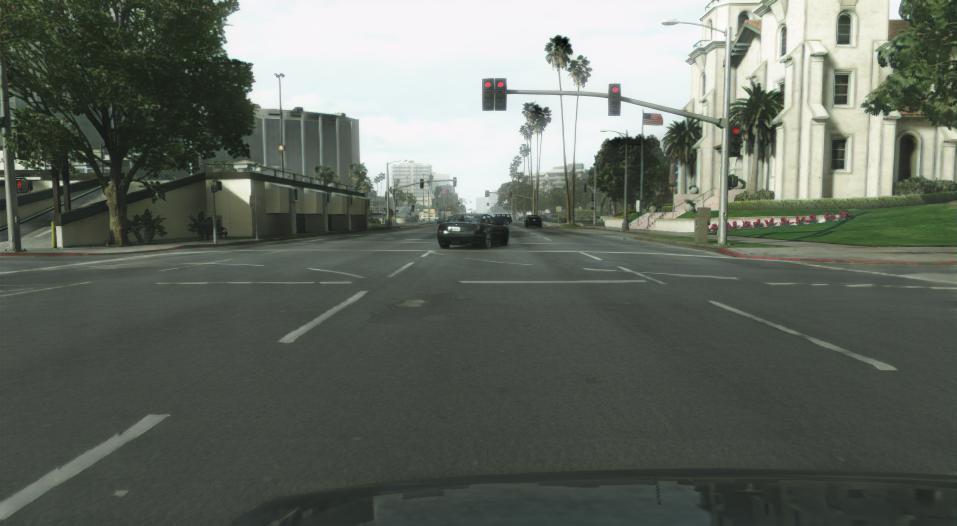}}\hfill\\\vspace{-9pt}
		\subfloat[Input]
		{\includegraphics[width=0.33\textwidth]{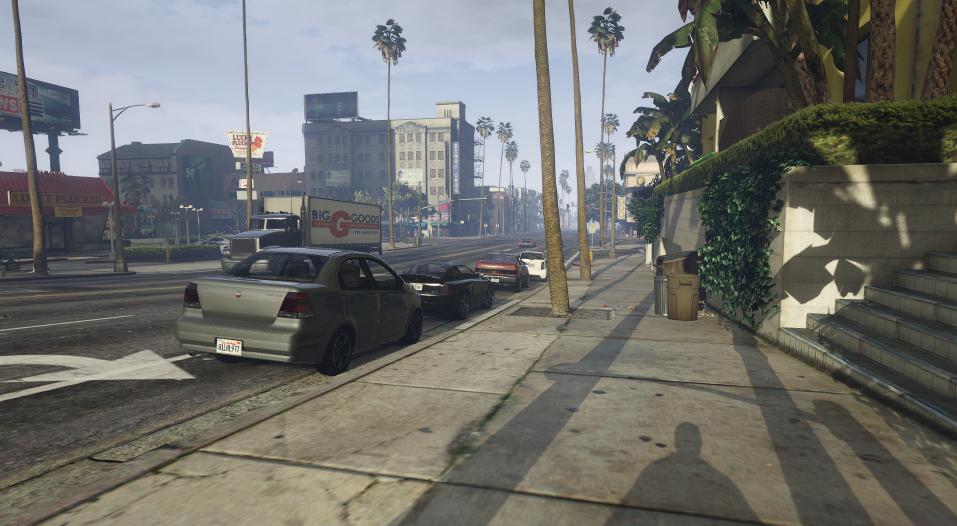}}\hfill
		\subfloat[EPE]
		{\includegraphics[width=0.33\textwidth]{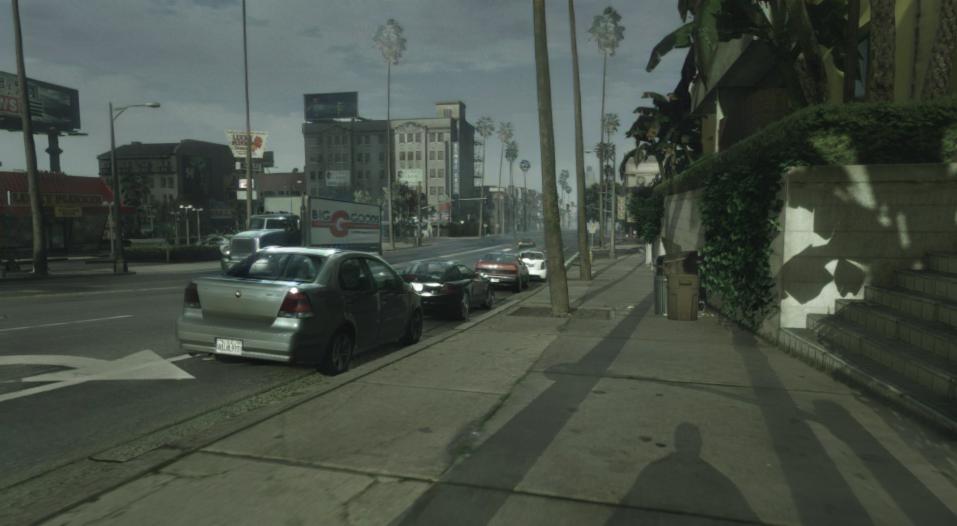}}\hfill
		\subfloat[FeaMGAN (ours)]
		{\includegraphics[width=0.33\textwidth]{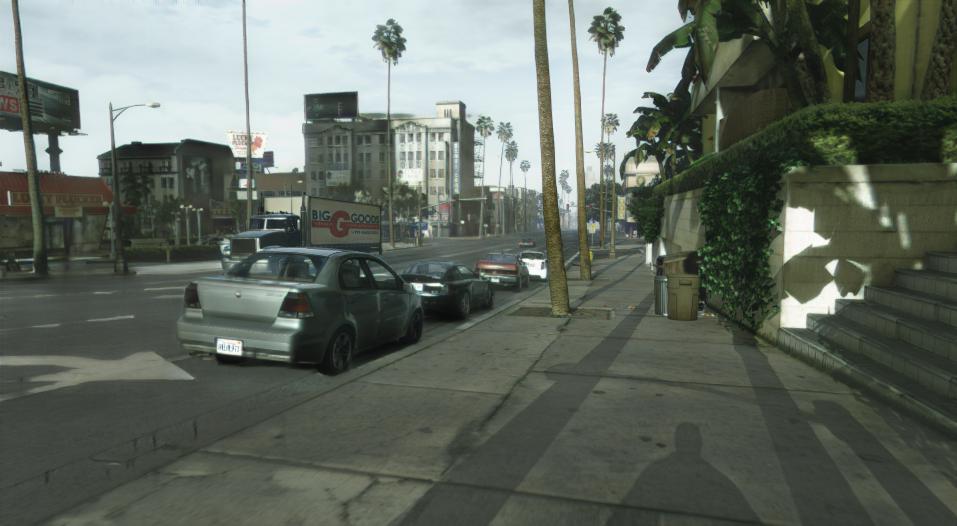}}\hfill 
	\end{center}
	\vspace{-1ex}
	\caption[Qualitative comparison to EPE.]{Qualitative comparison to EPE. We compare our method with the provided inferred images of EPE \cite{richter2022enhancing}. Results are randomly sampled from the best model. Best viewed in color.}
	\label{fig:feamgan:app:qualitative_comparison_epe_additional_random}
\end{figure}

\begin{figure}[h] 
	\captionsetup[subfigure]{labelformat=empty}
	\begin{center}
		{\scriptsize PFD$\rightarrow$Cityscapes} \hfill\\\vspace{1pt}
		{\includegraphics[width=0.165\textwidth]{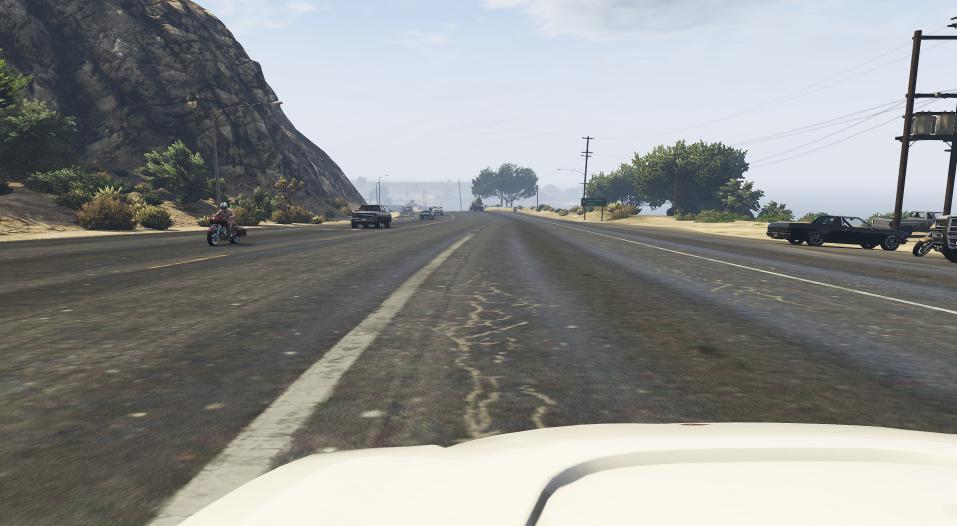}}\hfill
		{\includegraphics[width=0.165\textwidth]{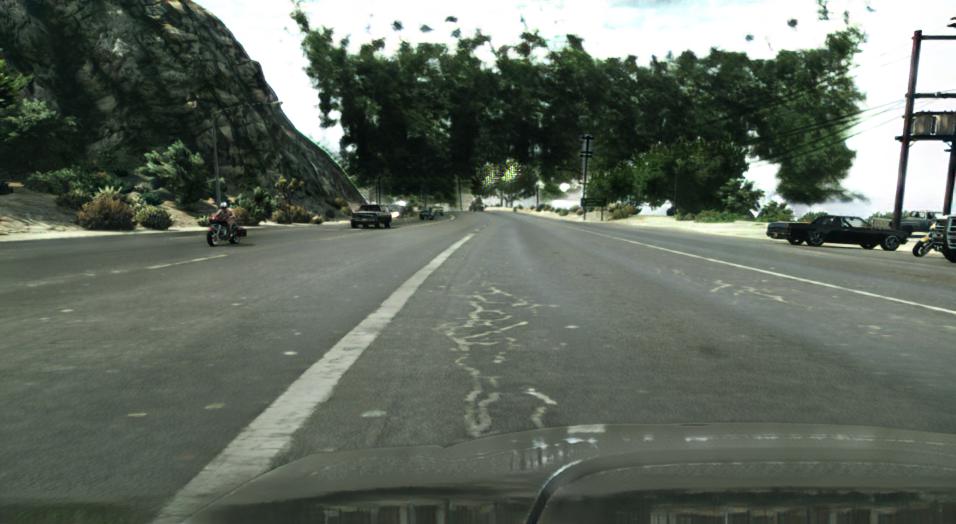}}\hfill
		{\includegraphics[width=0.165\textwidth]{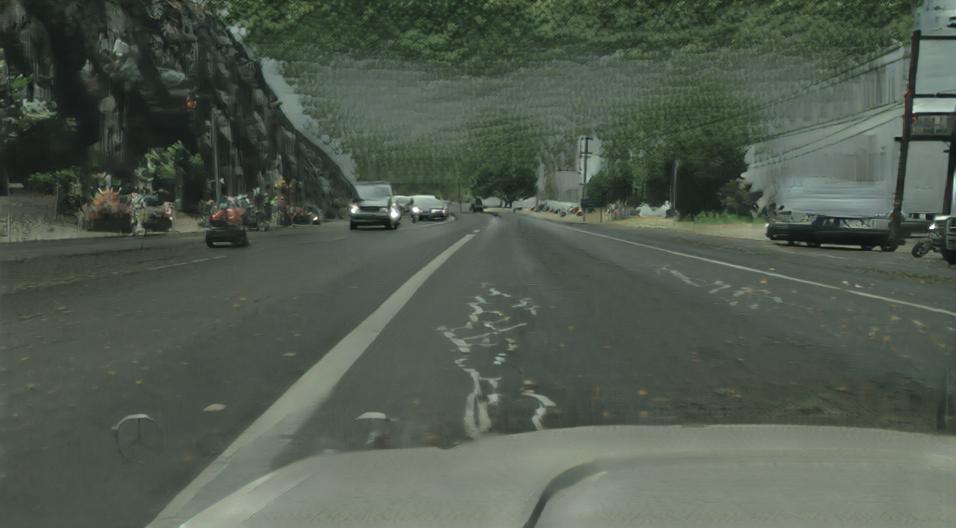}}\hfill
		{\includegraphics[width=0.165\textwidth]{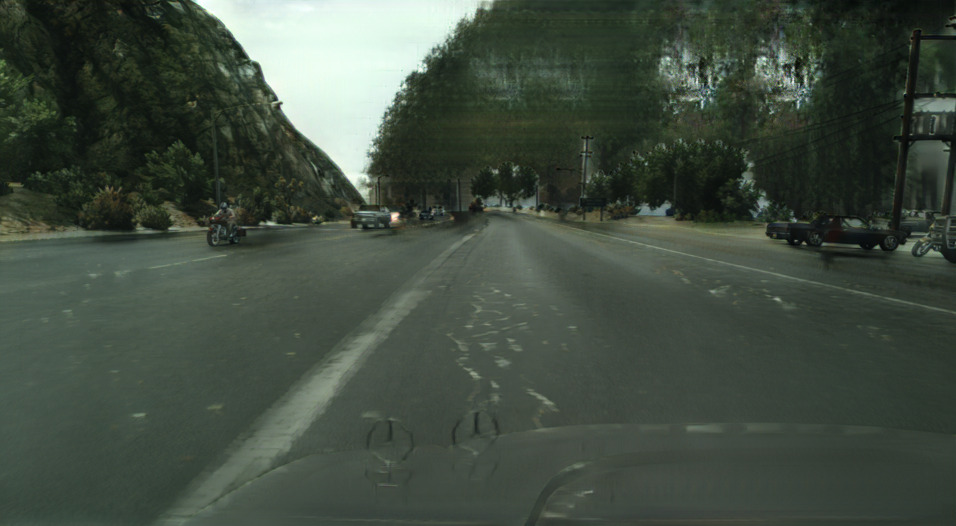}}\hfill
		{\includegraphics[width=0.165\textwidth]{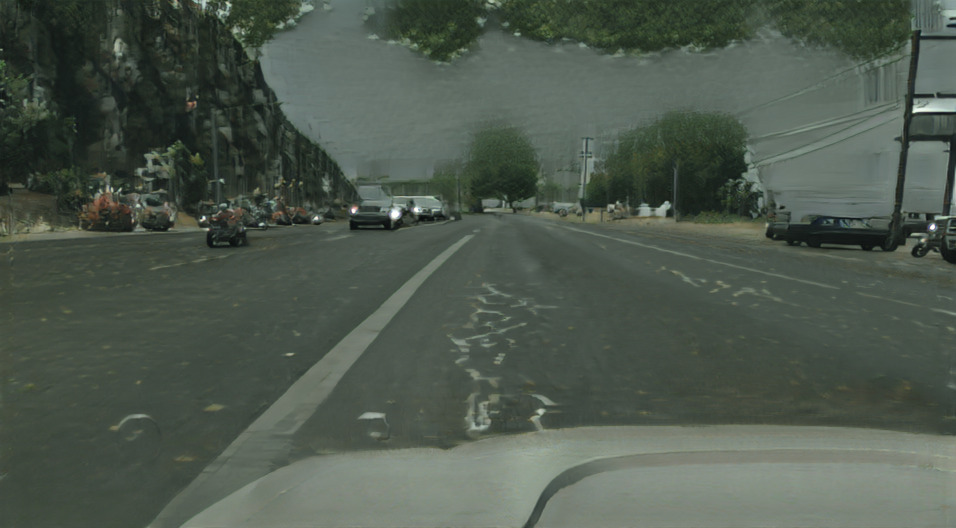}}\hfill
		{\includegraphics[width=0.165\textwidth]{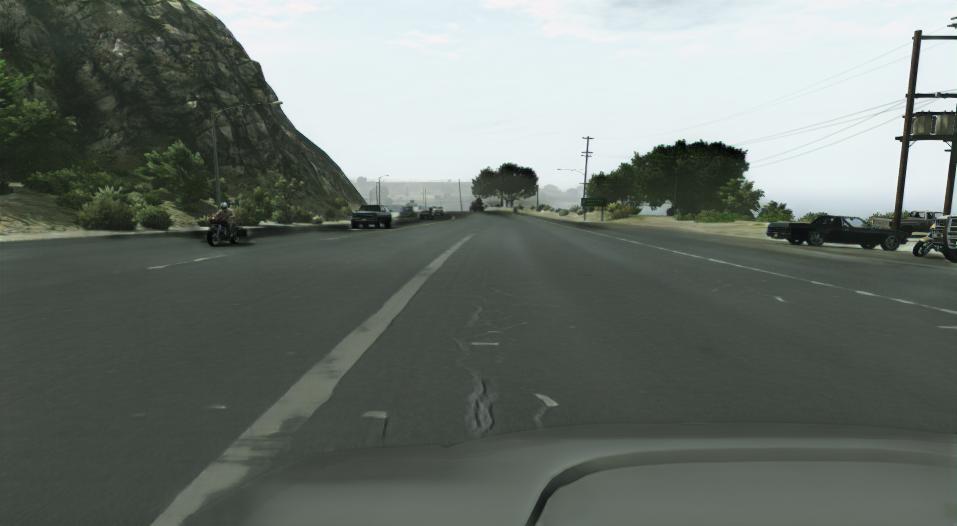}}\hfill\\
		{\includegraphics[width=0.165\textwidth]{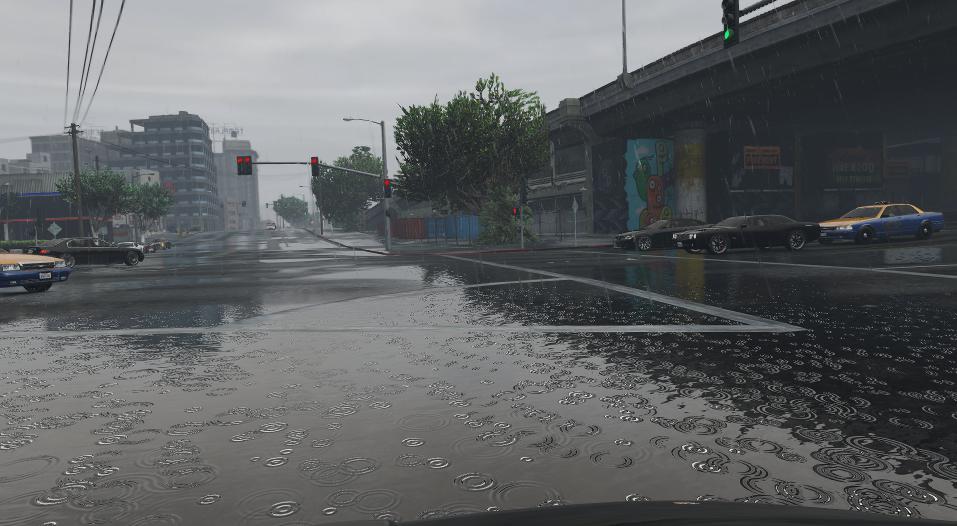}}\hfill
		{\includegraphics[width=0.165\textwidth]{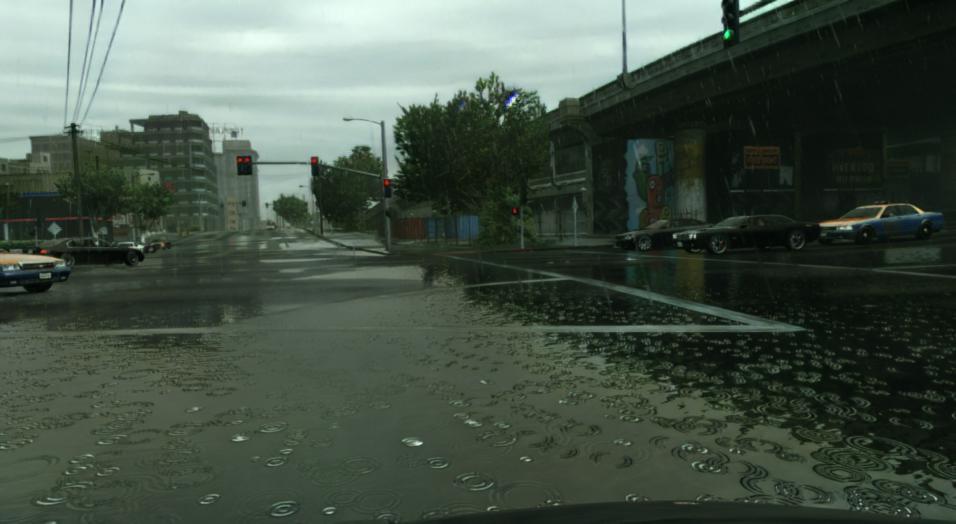}}\hfill
		{\includegraphics[width=0.165\textwidth]{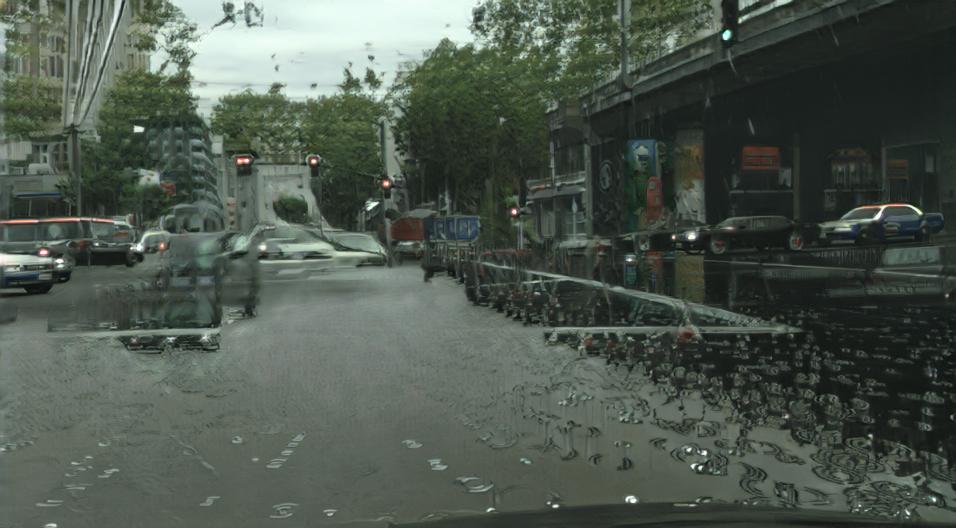}}\hfill
		{\includegraphics[width=0.165\textwidth]{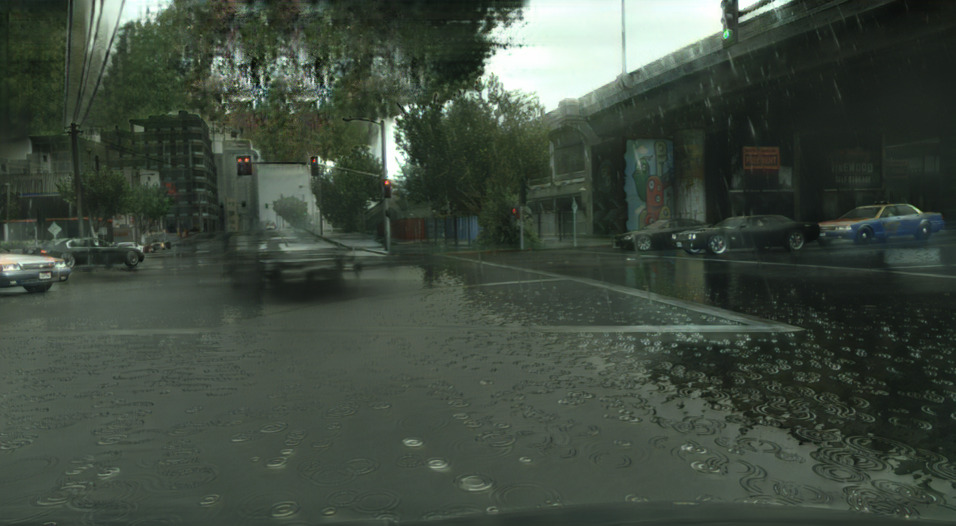}}\hfill
		{\includegraphics[width=0.165\textwidth]{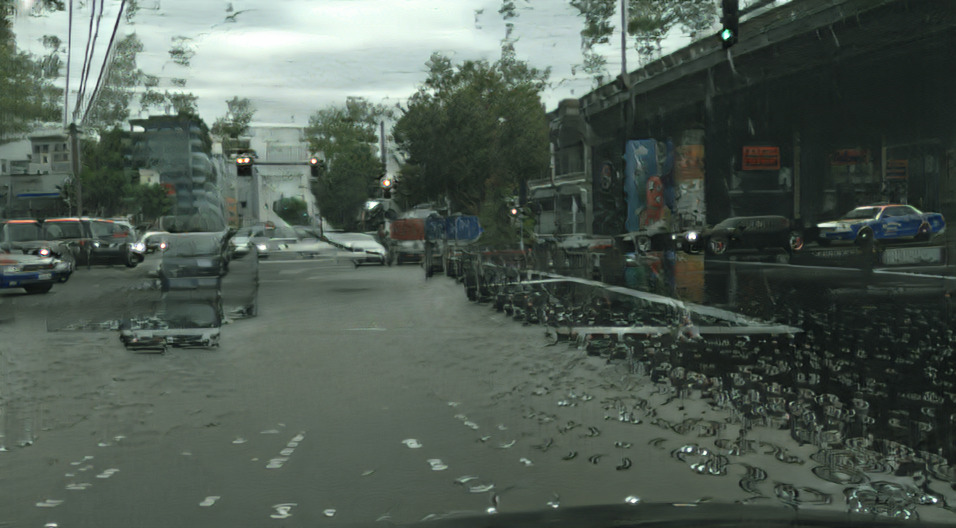}}\hfill
		{\includegraphics[width=0.165\textwidth]{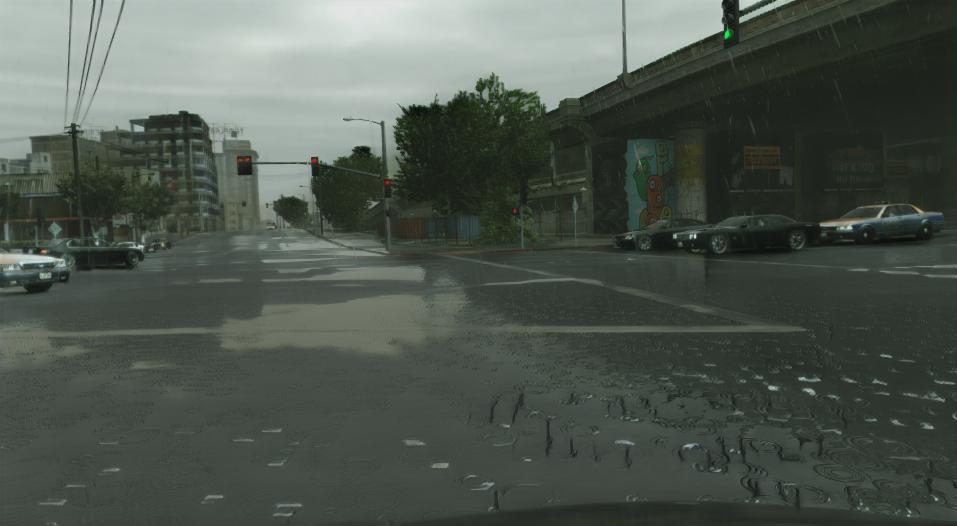}}\hfill\\
		{\includegraphics[width=0.165\textwidth]{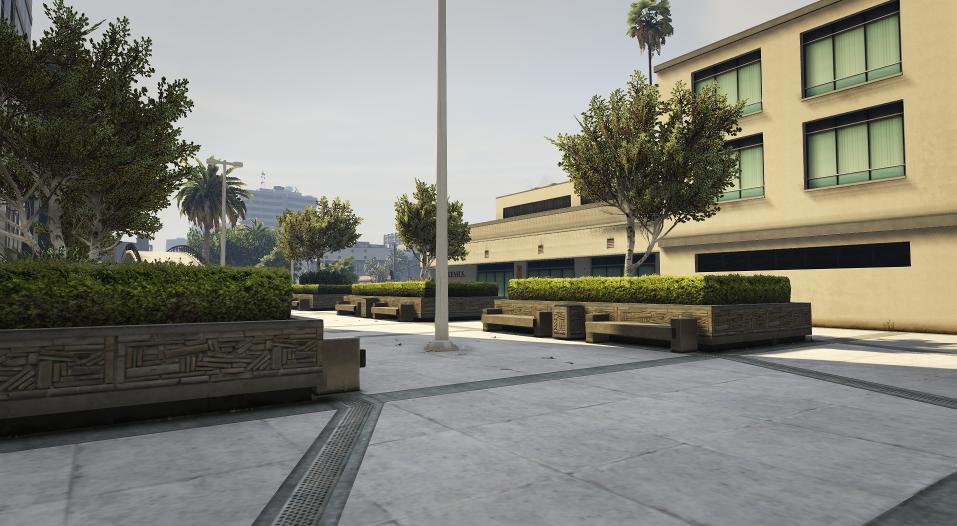}}\hfill
		{\includegraphics[width=0.165\textwidth]{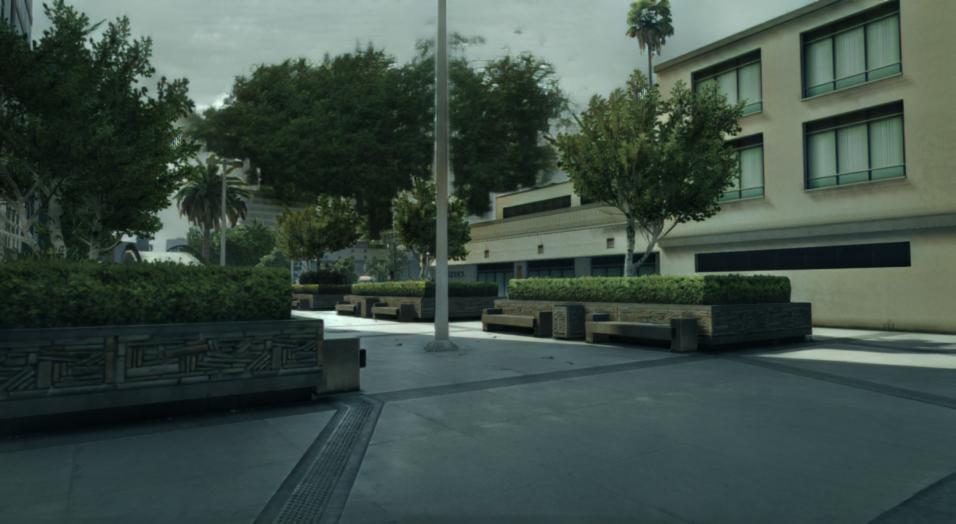}}\hfill
		{\includegraphics[width=0.165\textwidth]{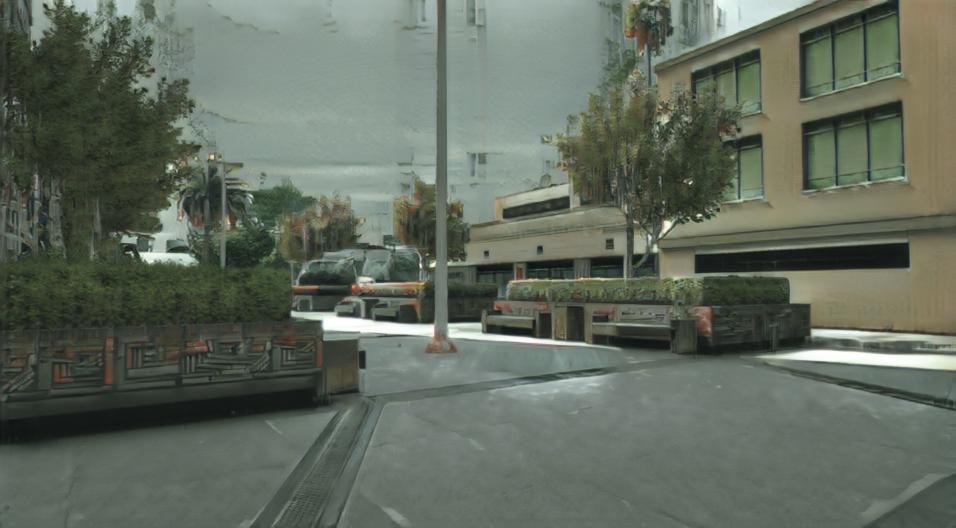}}\hfill
		{\includegraphics[width=0.165\textwidth]{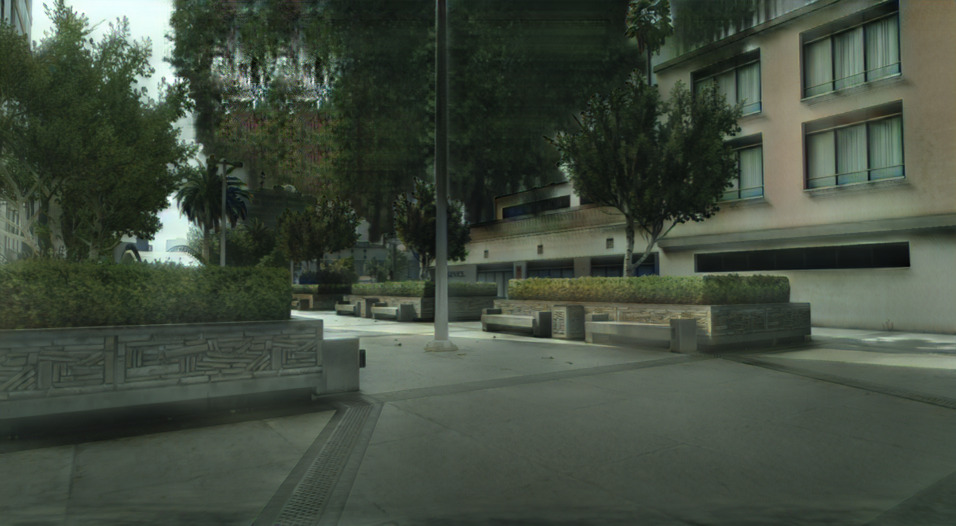}}\hfill
		{\includegraphics[width=0.165\textwidth]{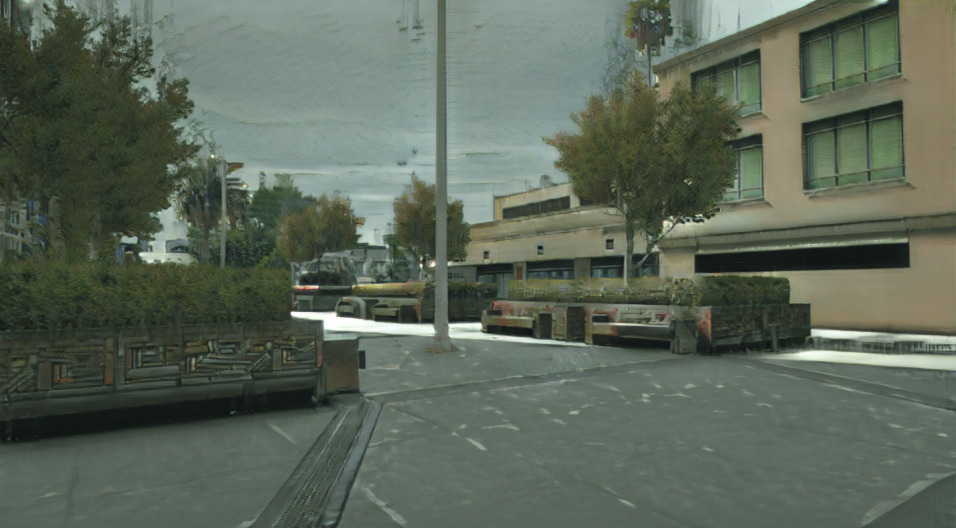}}\hfill
		{\includegraphics[width=0.165\textwidth]{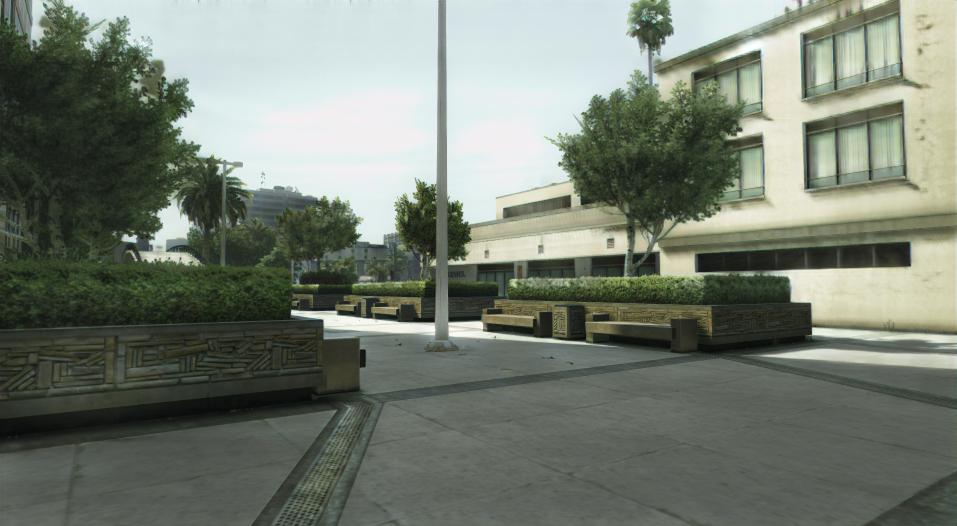}}\hfill\\\vspace{1pt}
		
		{\scriptsize Viper$\rightarrow$Cityscapes} \hfill\\\vspace{1pt}		
		{\includegraphics[width=0.165\textwidth]{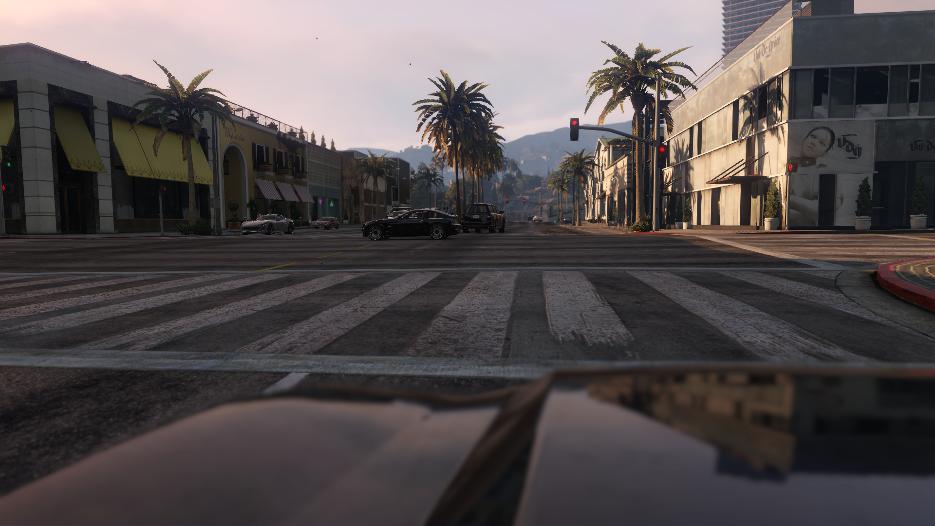}}\hfill
		{\includegraphics[width=0.165\textwidth]{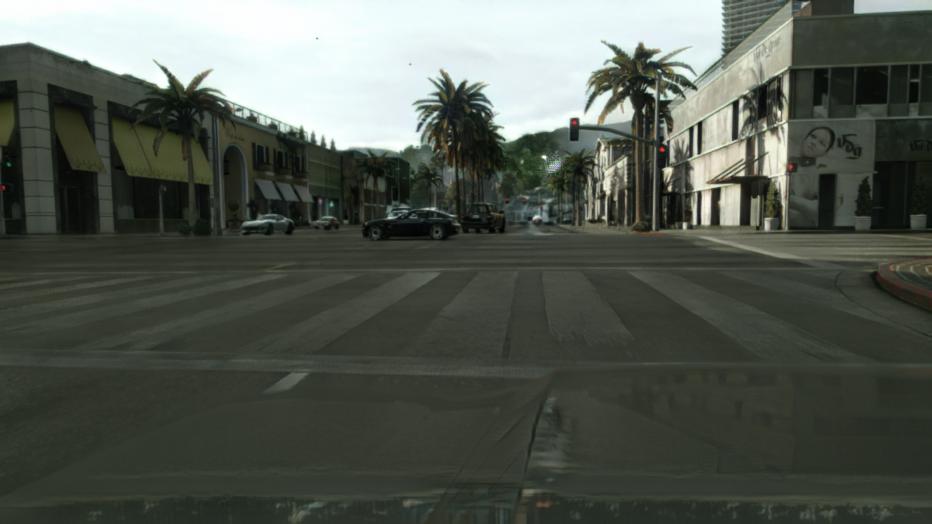}}\hfill
		{\includegraphics[width=0.165\textwidth]{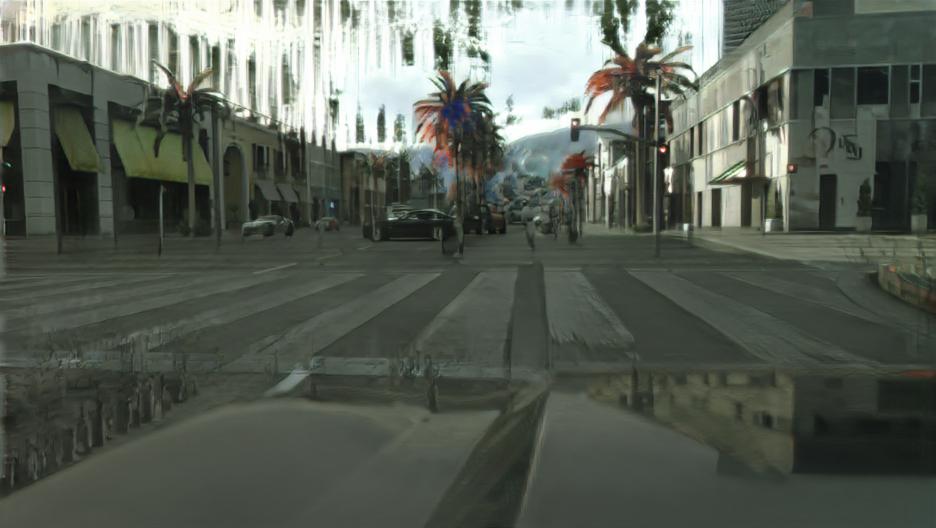}}\hfill
		{\includegraphics[width=0.165\textwidth]{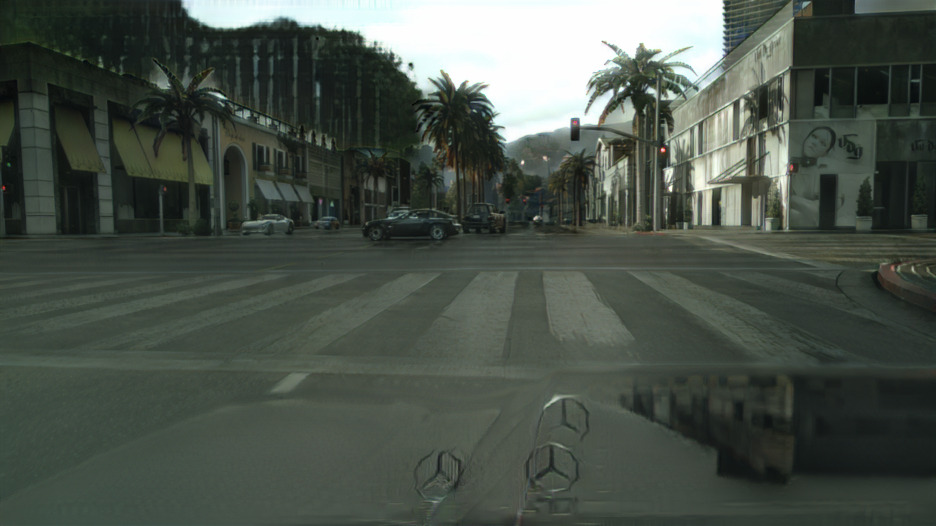}}\hfill
		{\includegraphics[width=0.165\textwidth]{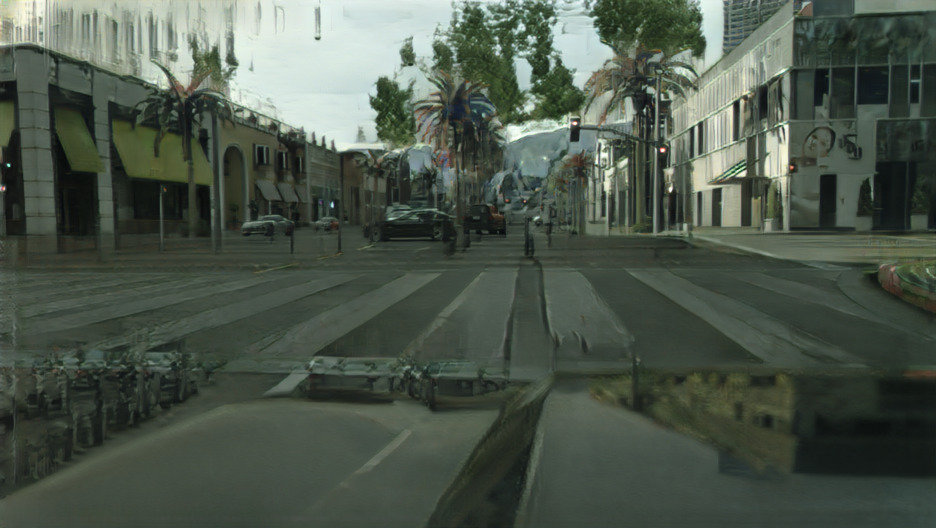}}\hfill
		{\includegraphics[width=0.165\textwidth]{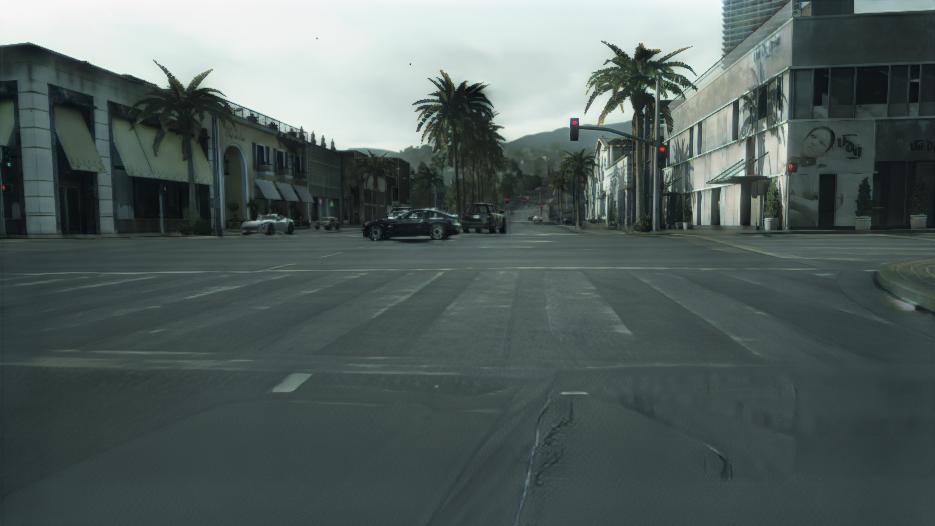}}\hfill\\
		{\includegraphics[width=0.165\textwidth]{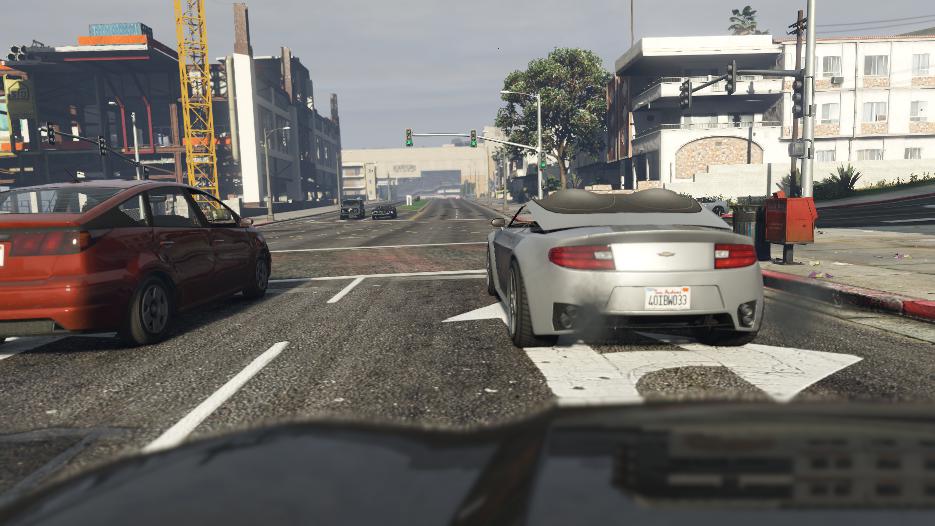}}\hfill
		{\includegraphics[width=0.165\textwidth]{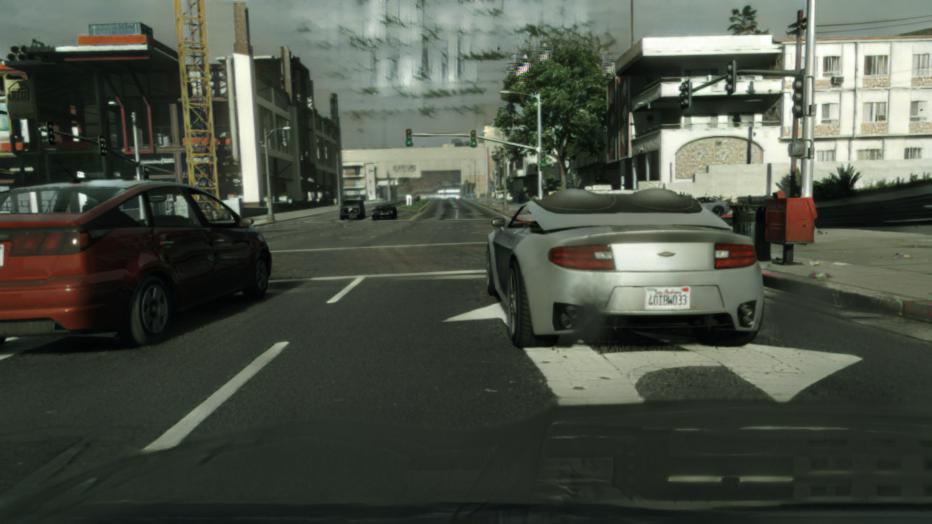}}\hfill
		{\includegraphics[width=0.165\textwidth]{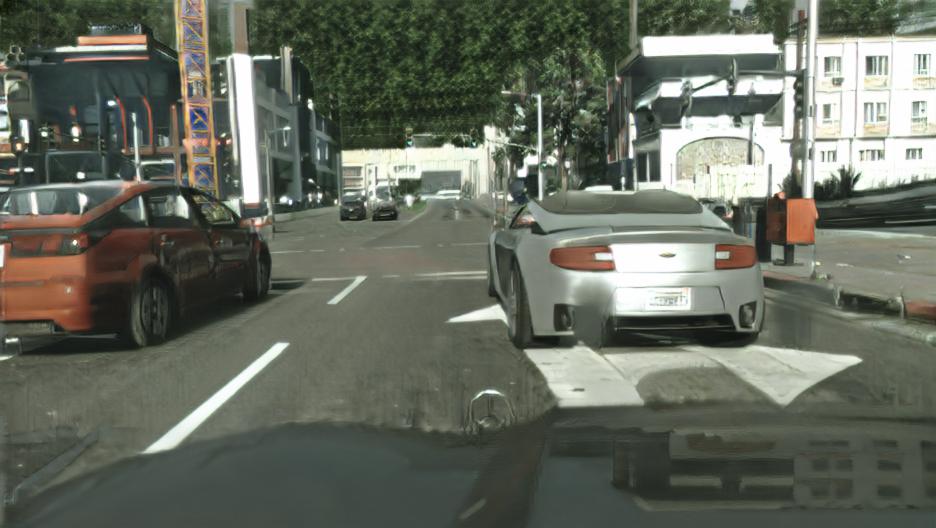}}\hfill
		{\includegraphics[width=0.165\textwidth]{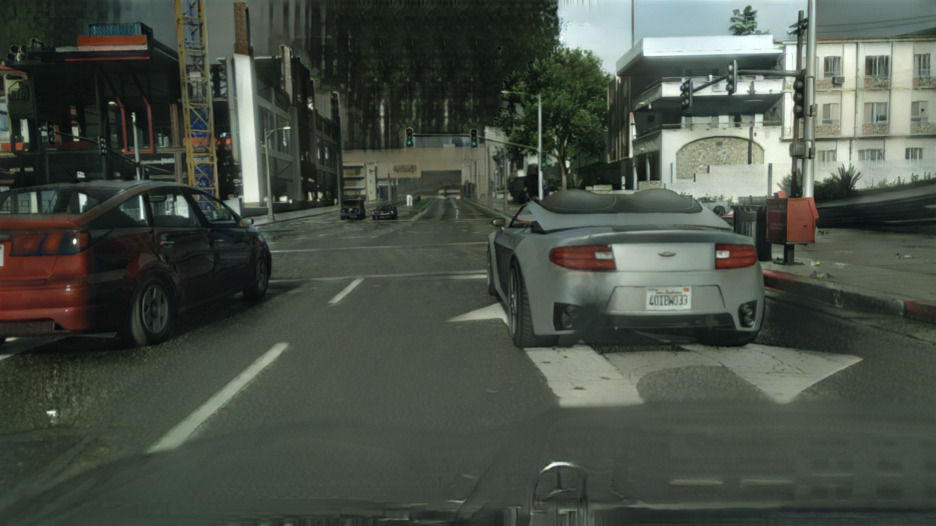}}\hfill
		{\includegraphics[width=0.165\textwidth]{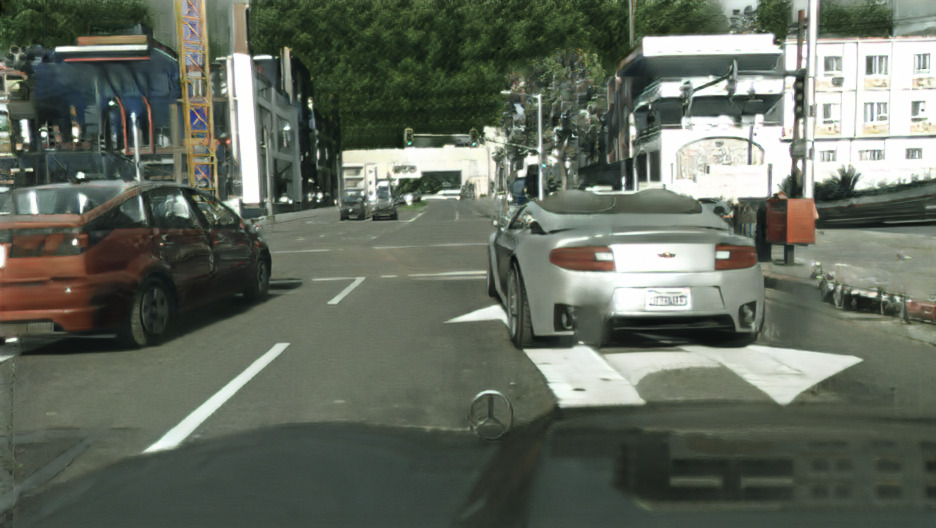}}\hfill
		{\includegraphics[width=0.165\textwidth]{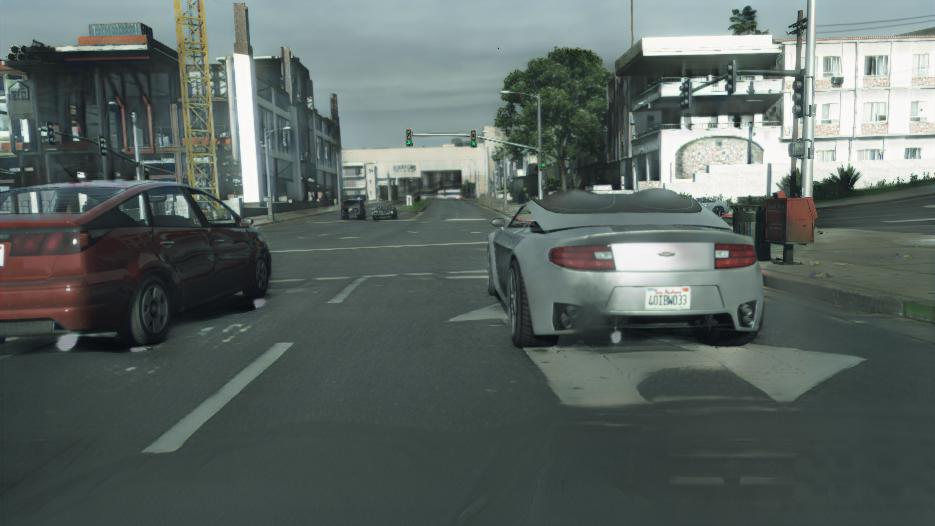}}\hfill\\
		{\includegraphics[width=0.165\textwidth]{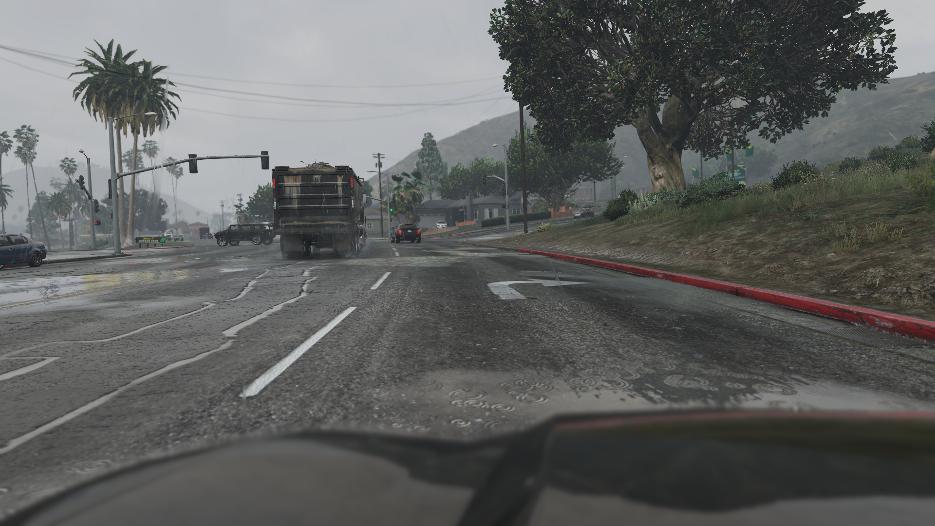}}\hfill
		{\includegraphics[width=0.165\textwidth]{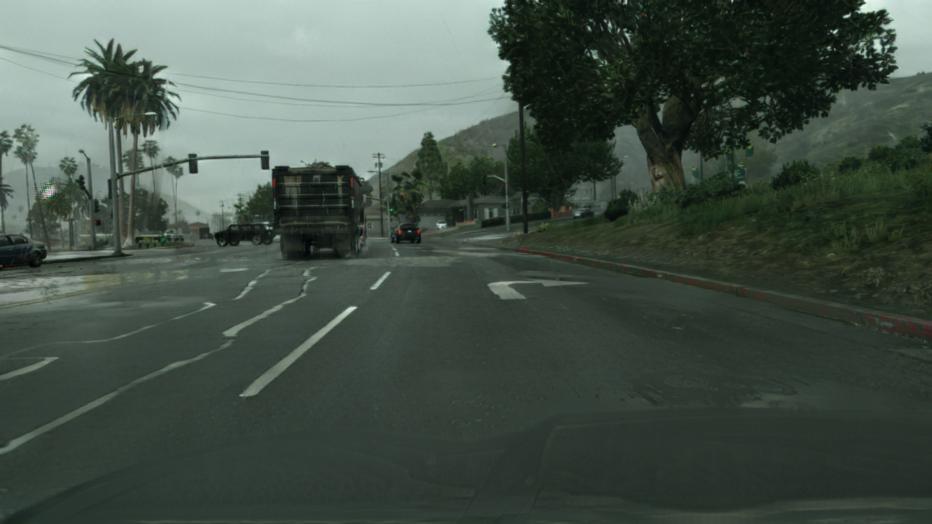}}\hfill
		{\includegraphics[width=0.165\textwidth]{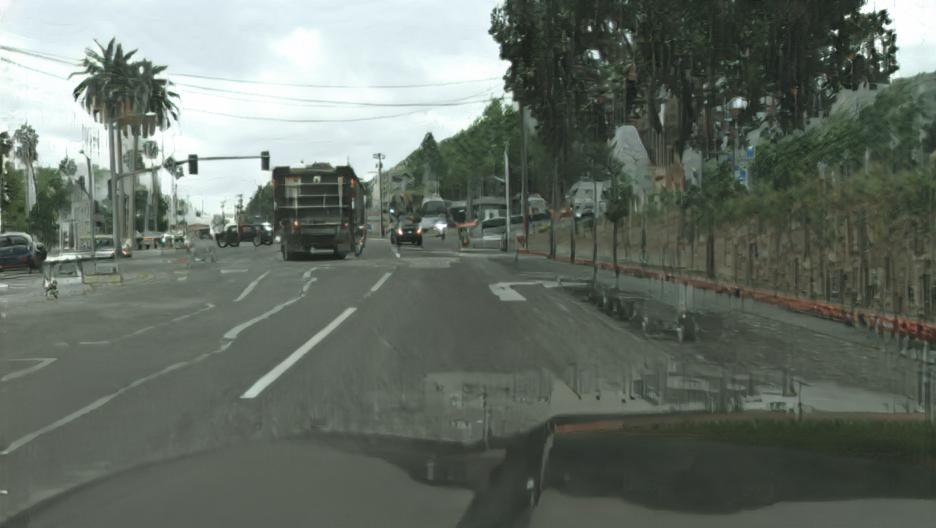}}\hfill
		{\includegraphics[width=0.165\textwidth]{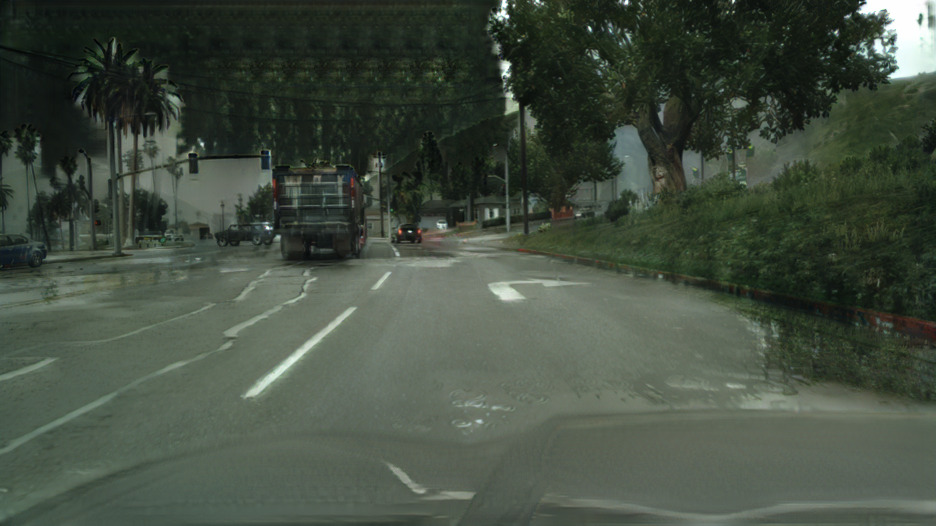}}\hfill
		{\includegraphics[width=0.165\textwidth]{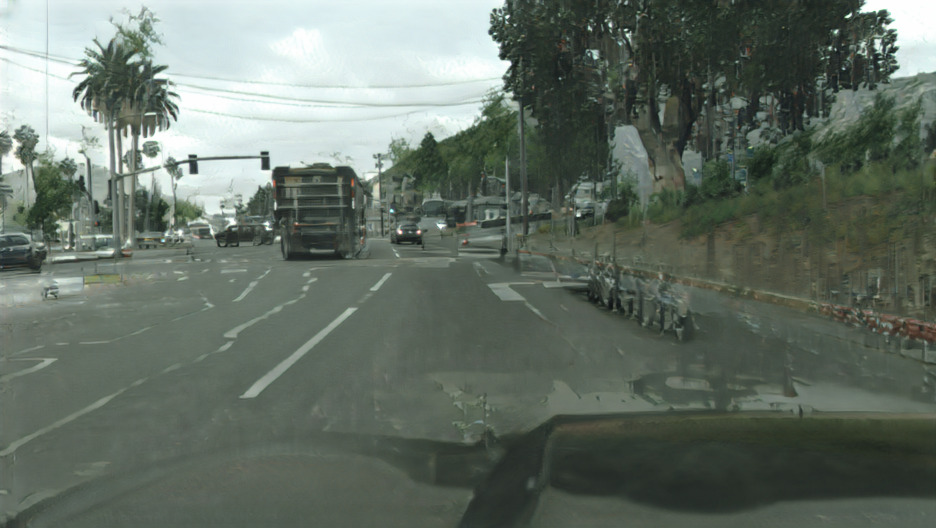}}\hfill
		{\includegraphics[width=0.165\textwidth]{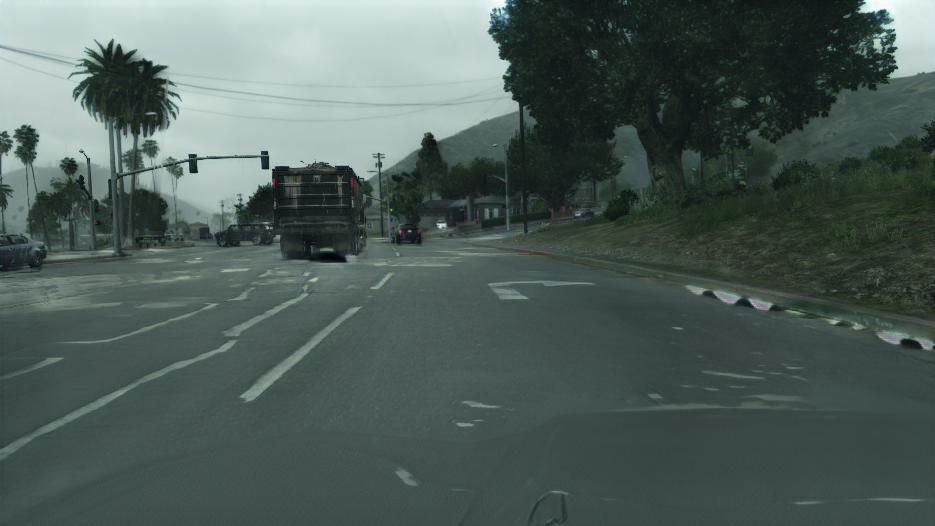}}\hfill\\\vspace{1pt}
		
		{\scriptsize Day$\rightarrow$Night} \hfill\\\vspace{1pt}
		{\includegraphics[width=0.165\textwidth]{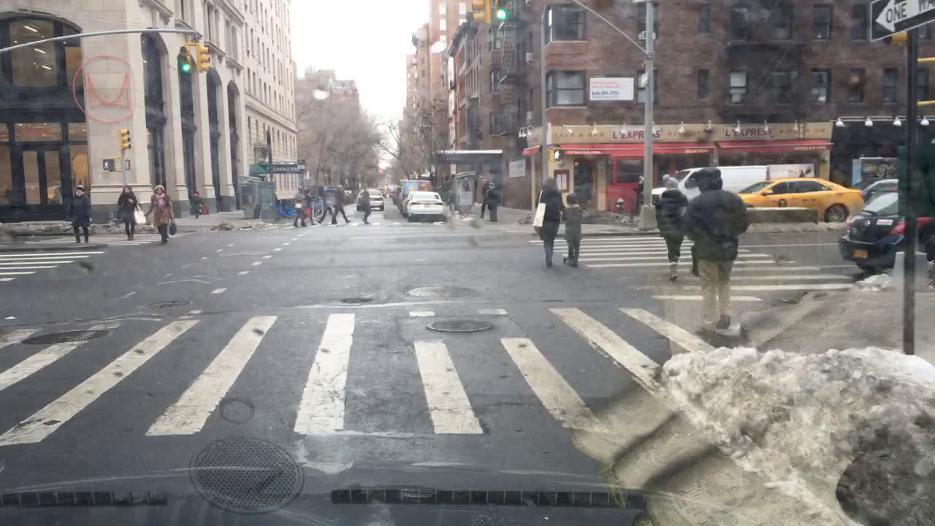}}\hfill
		{\includegraphics[width=0.165\textwidth]{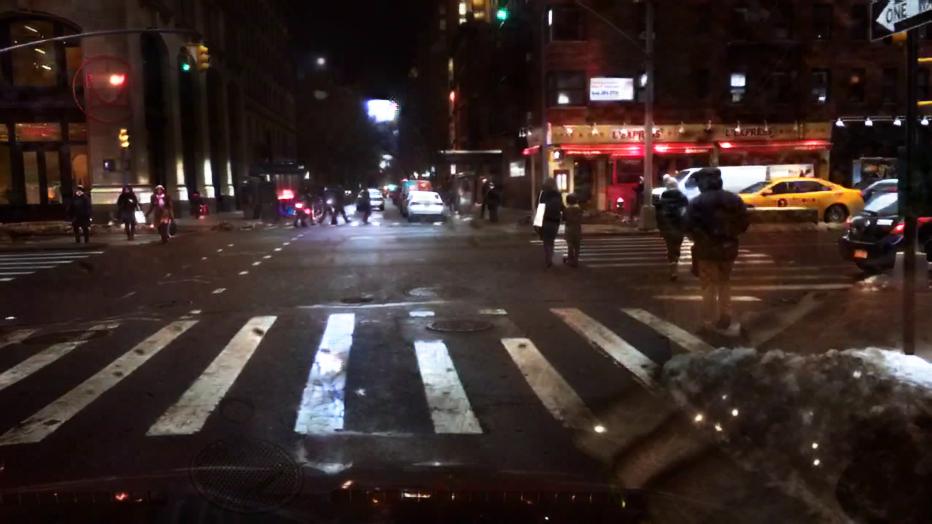}}\hfill
		{\includegraphics[width=0.165\textwidth]{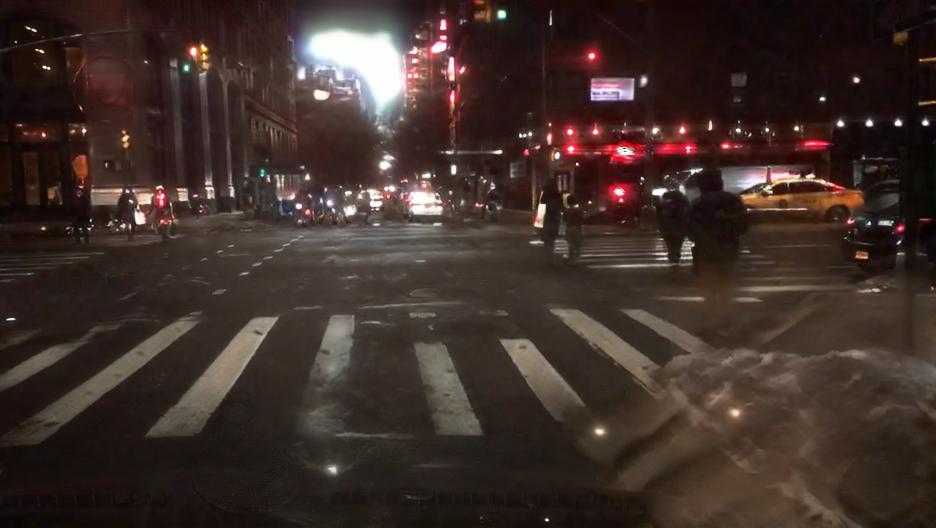}}\hfill
		{\includegraphics[width=0.165\textwidth]{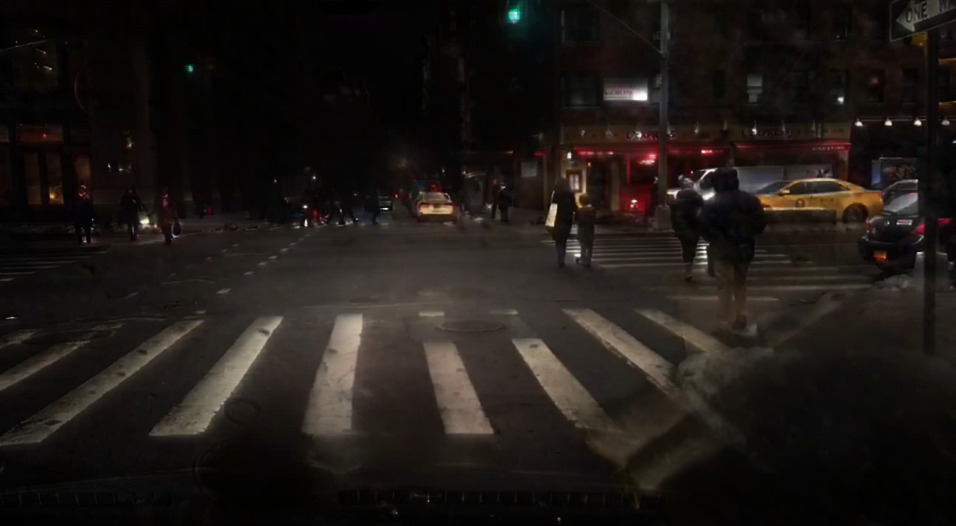}}\hfill
		{\includegraphics[width=0.165\textwidth]{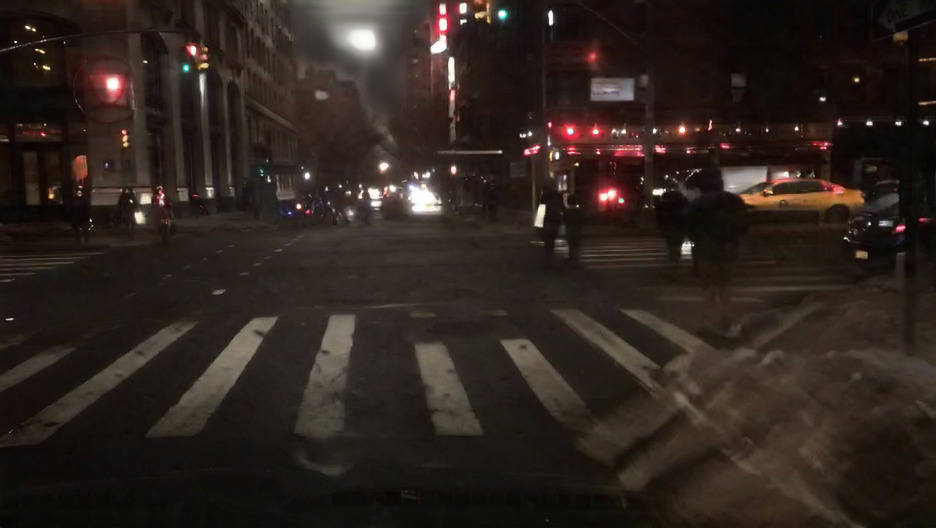}}\hfill
		{\includegraphics[width=0.165\textwidth]{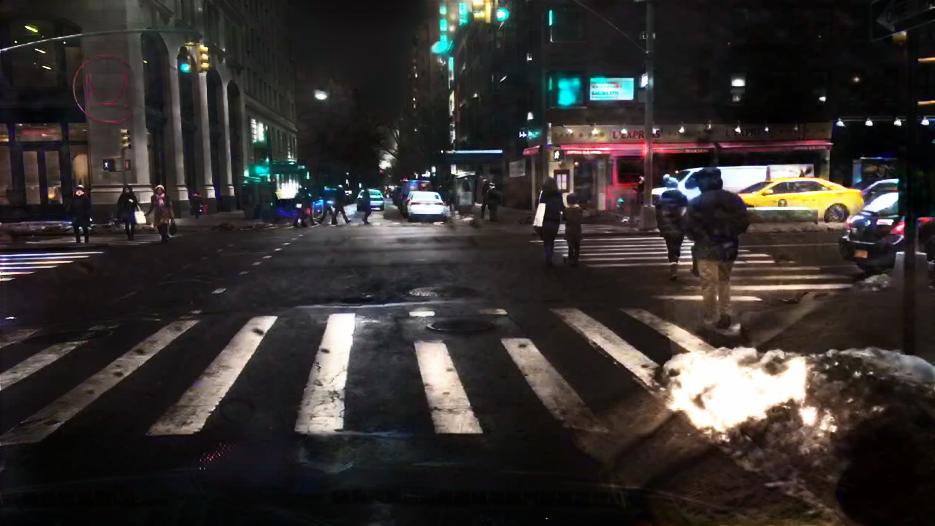}}\hfill\\
		{\includegraphics[width=0.165\textwidth]{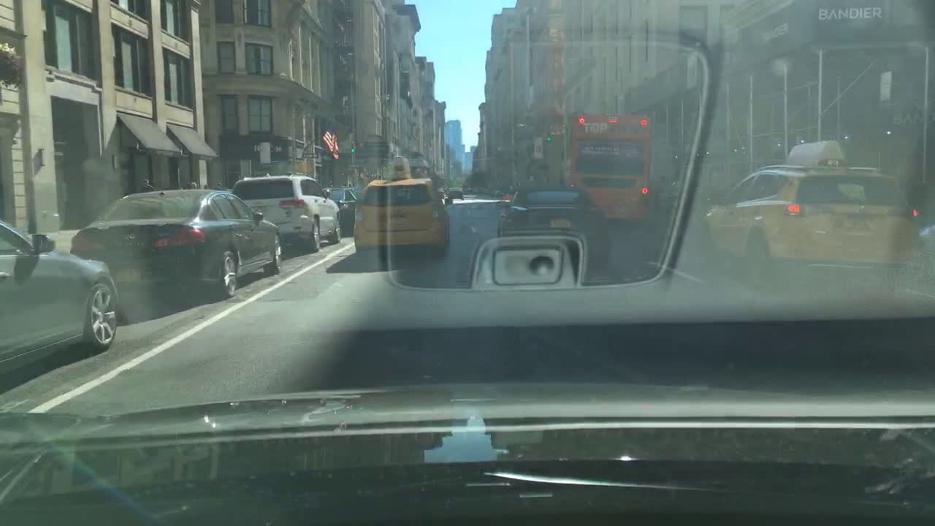}}\hfill
		{\includegraphics[width=0.165\textwidth]{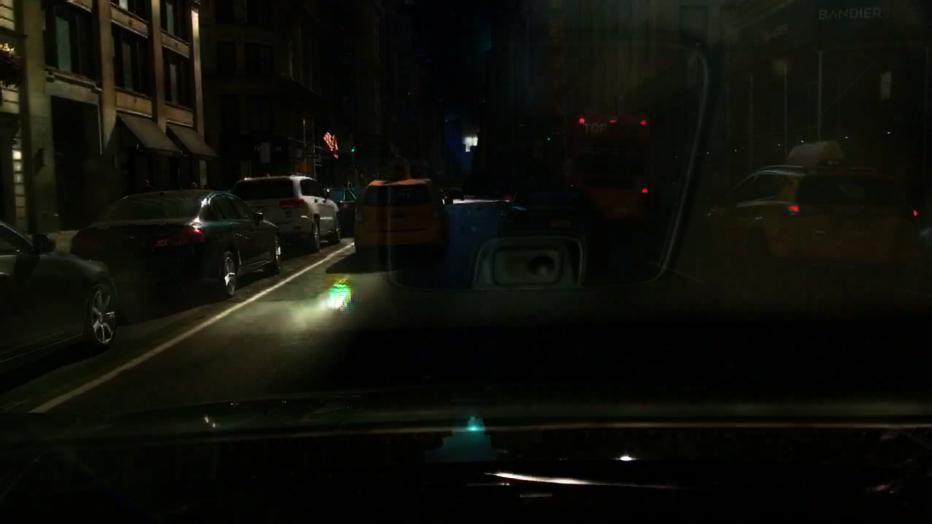}}\hfill
		{\includegraphics[width=0.165\textwidth]{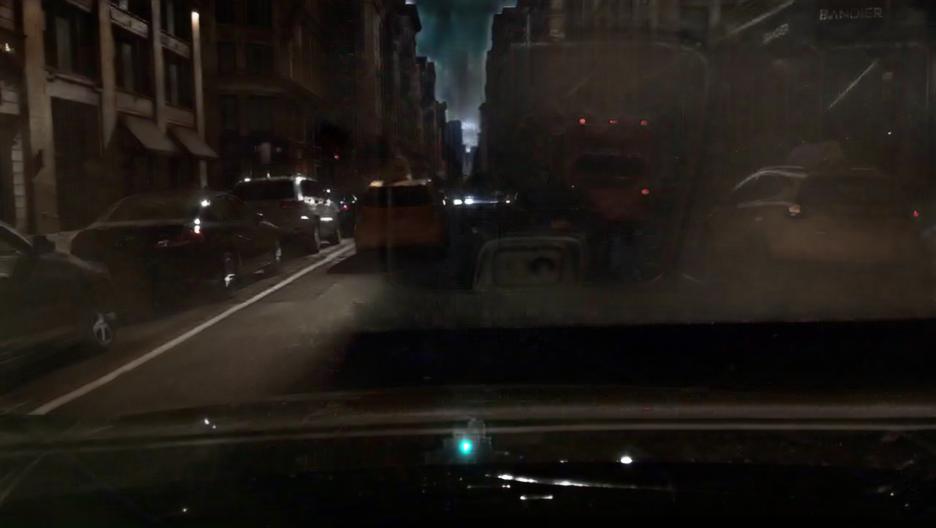}}\hfill
		{\includegraphics[width=0.165\textwidth]{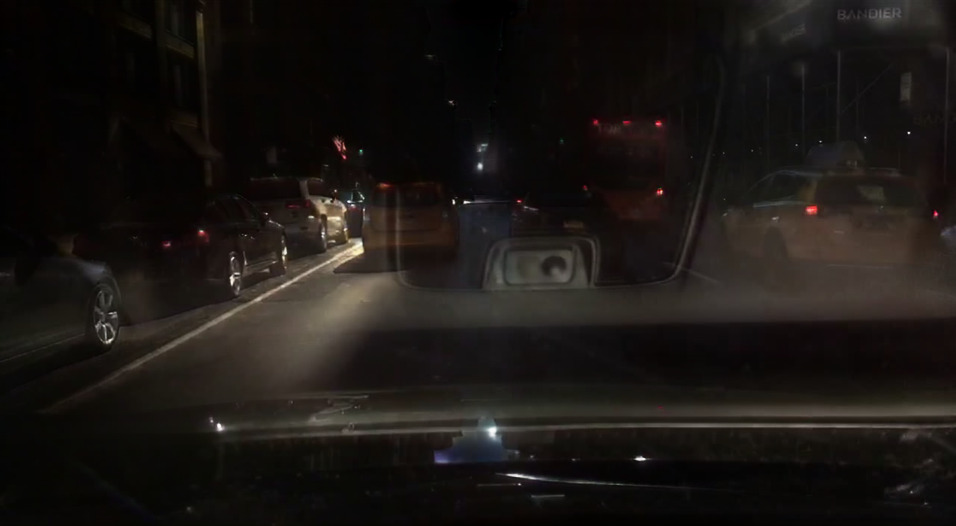}}\hfill
		{\includegraphics[width=0.165\textwidth]{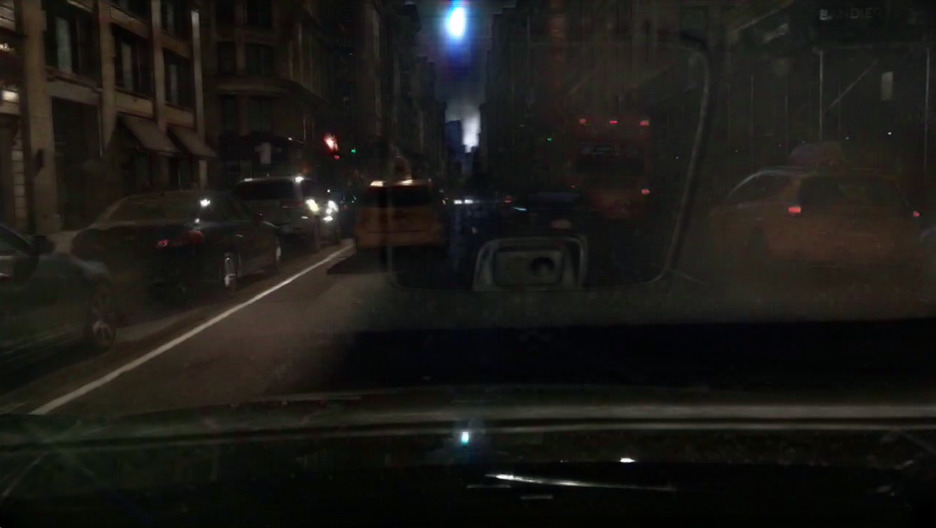}}\hfill
		{\includegraphics[width=0.165\textwidth]{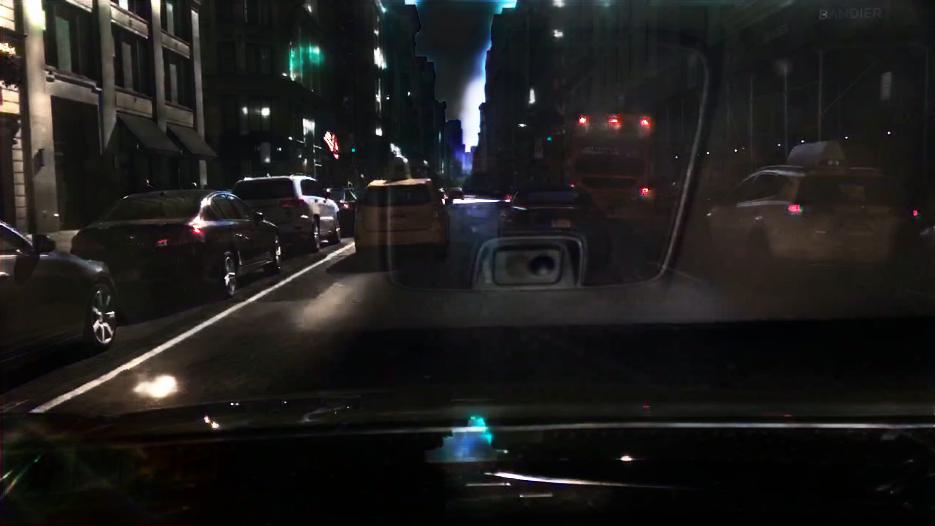}}\hfill\\
		{\includegraphics[width=0.165\textwidth]{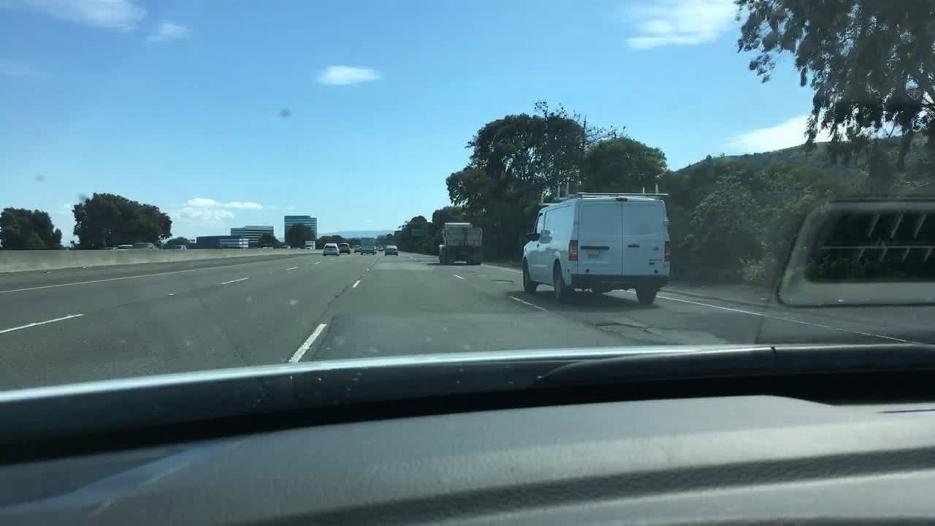}}\hfill
		{\includegraphics[width=0.165\textwidth]{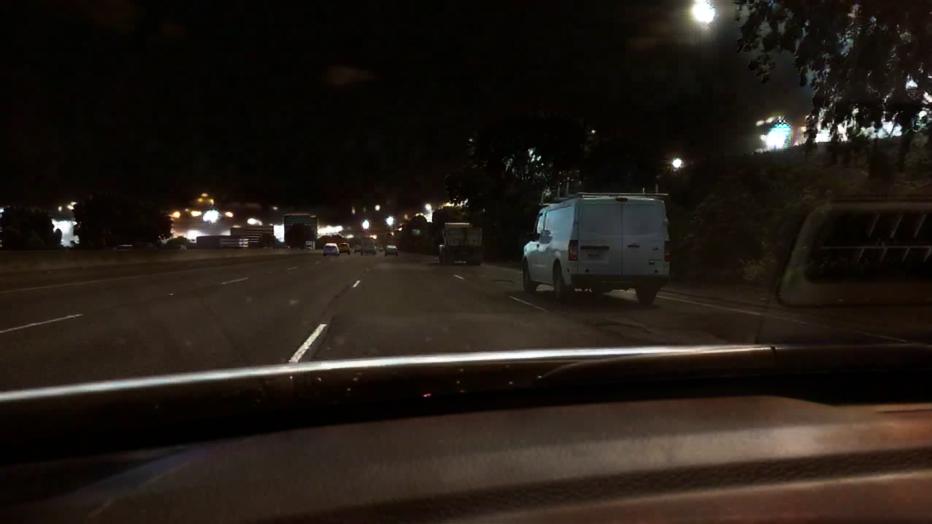}}\hfill
		{\includegraphics[width=0.165\textwidth]{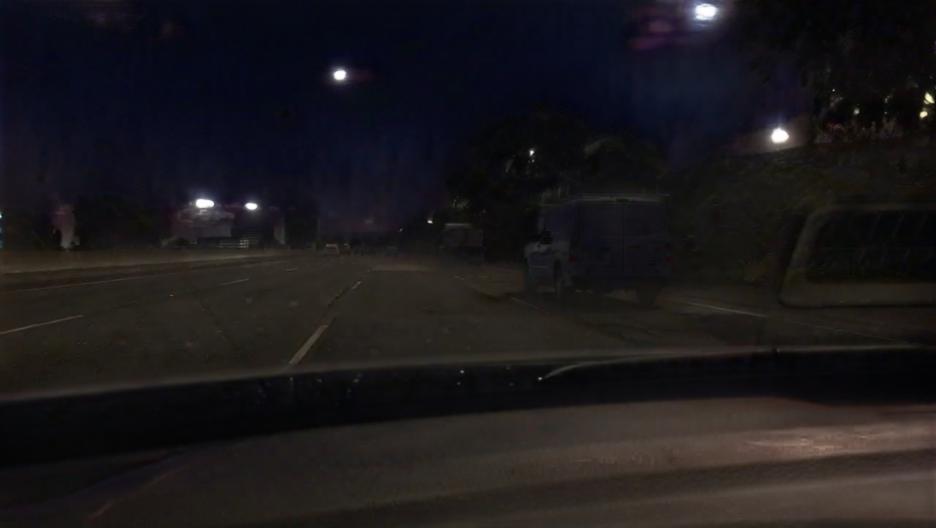}}\hfill
		{\includegraphics[width=0.165\textwidth]{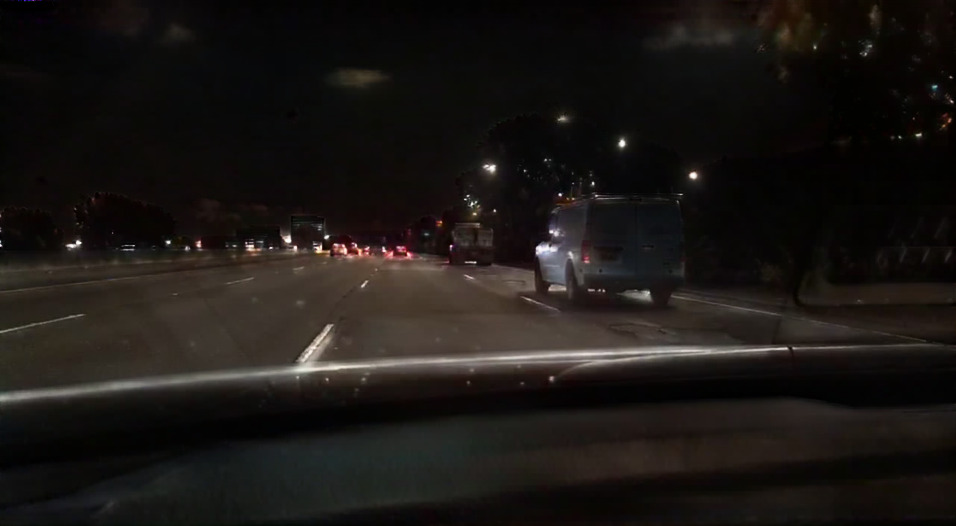}}\hfill
		{\includegraphics[width=0.165\textwidth]{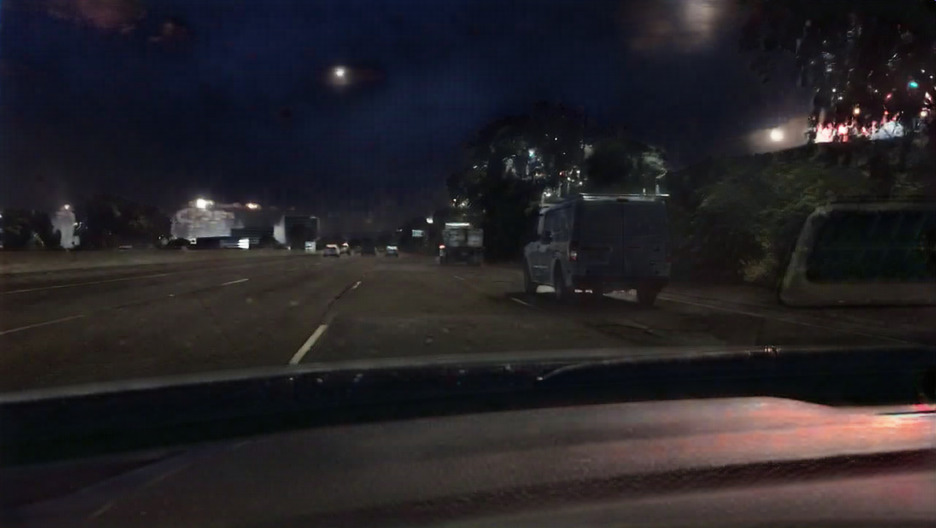}}\hfill
		{\includegraphics[width=0.165\textwidth]{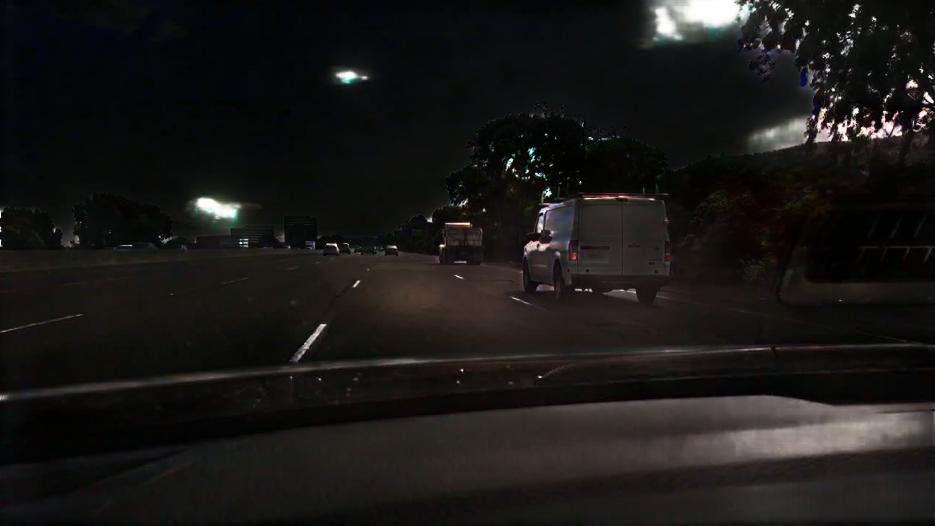}}\hfill\\\vspace{1pt}	
		{\scriptsize Clear$\rightarrow$Snowy} \hfill\\\vspace{1pt}
		{\includegraphics[width=0.165\textwidth]{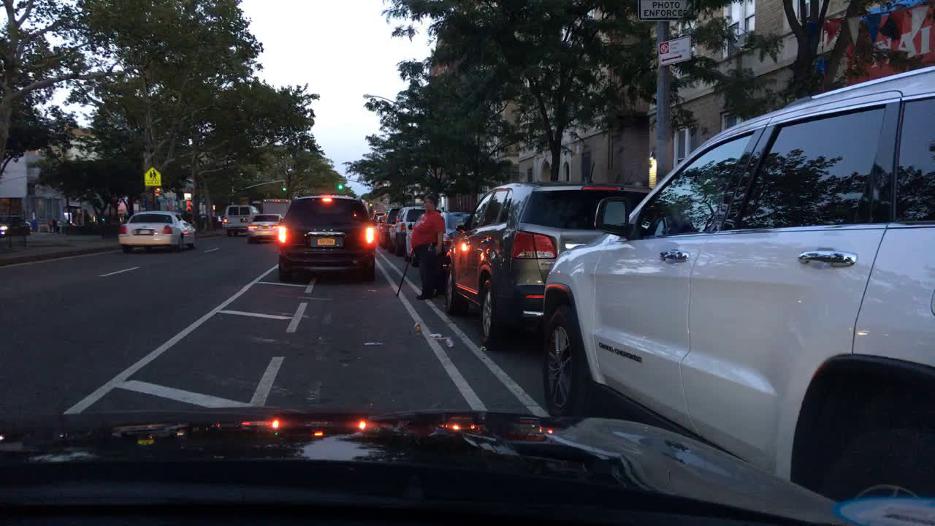}}\hfill
		{\includegraphics[width=0.165\textwidth]{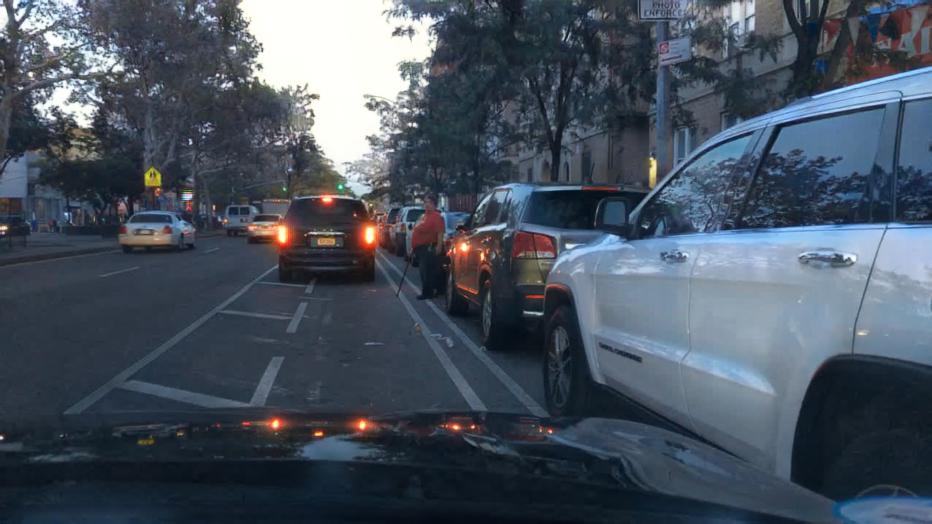}}\hfill
		{\includegraphics[width=0.165\textwidth]{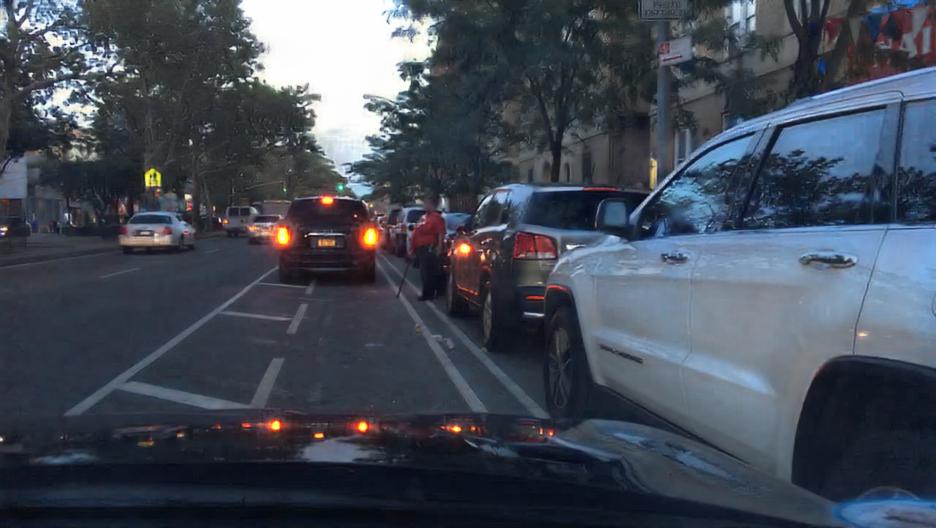}}\hfill
		{\includegraphics[width=0.165\textwidth]{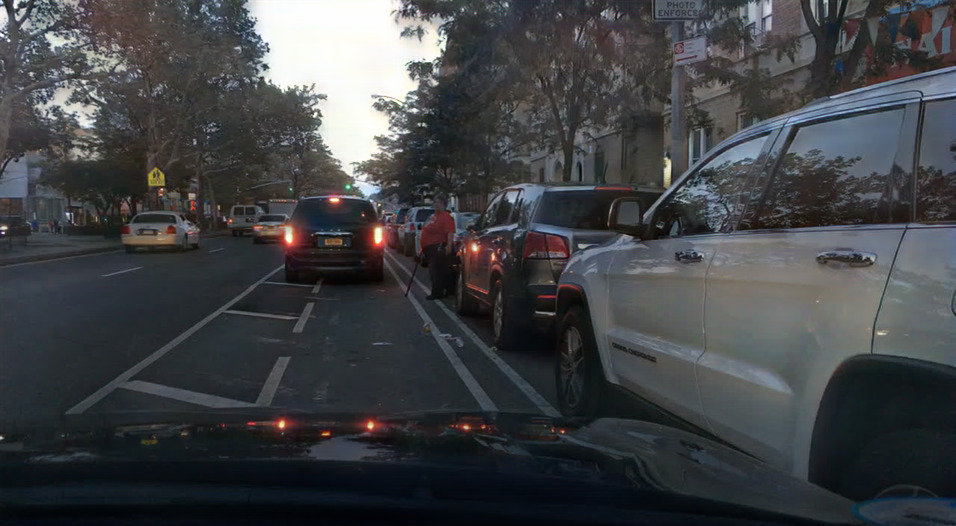}}\hfill
		{\includegraphics[width=0.165\textwidth]{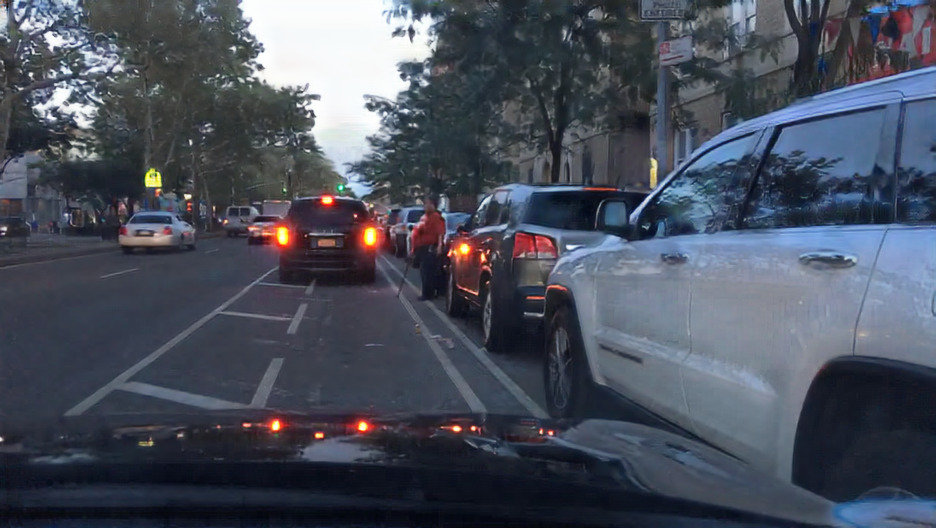}}\hfill
		{\includegraphics[width=0.165\textwidth]{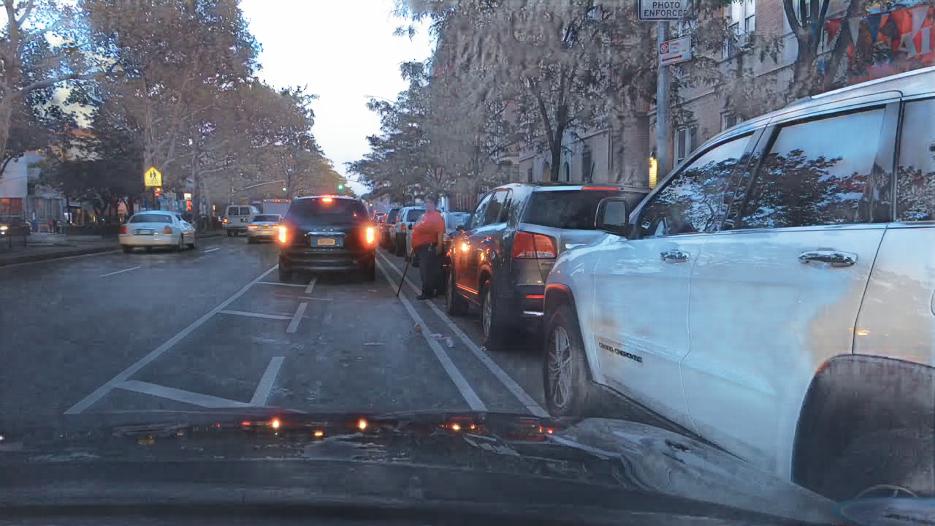}}\hfill\\
		{\includegraphics[width=0.165\textwidth]{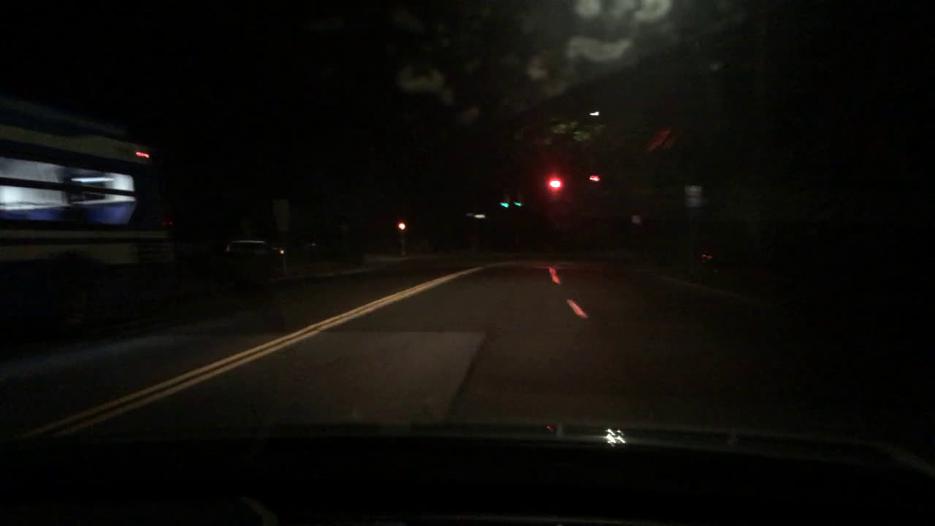}}\hfill
		{\includegraphics[width=0.165\textwidth]{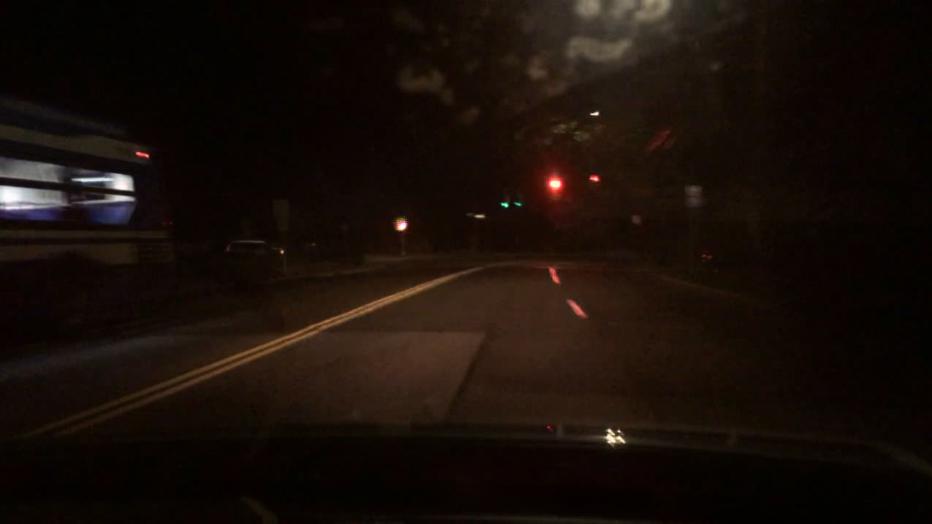}}\hfill
		{\includegraphics[width=0.165\textwidth]{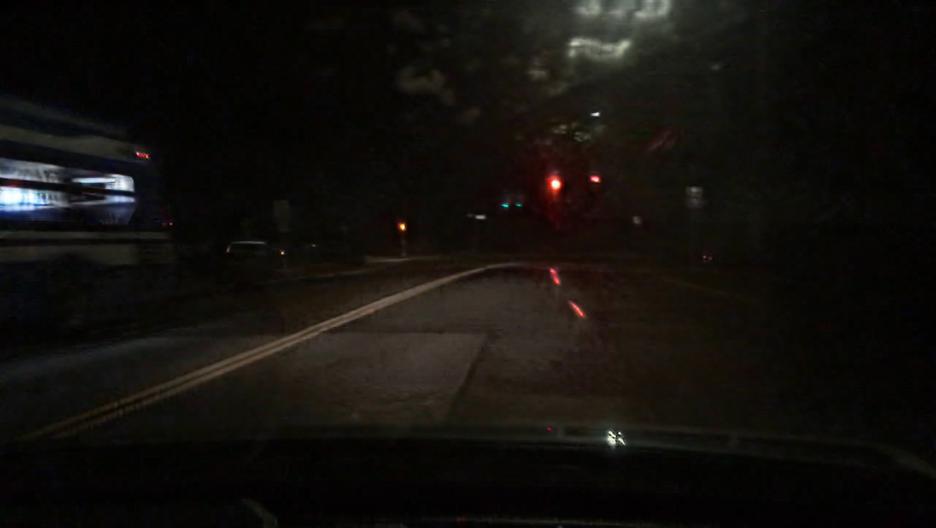}}\hfill
		{\includegraphics[width=0.165\textwidth]{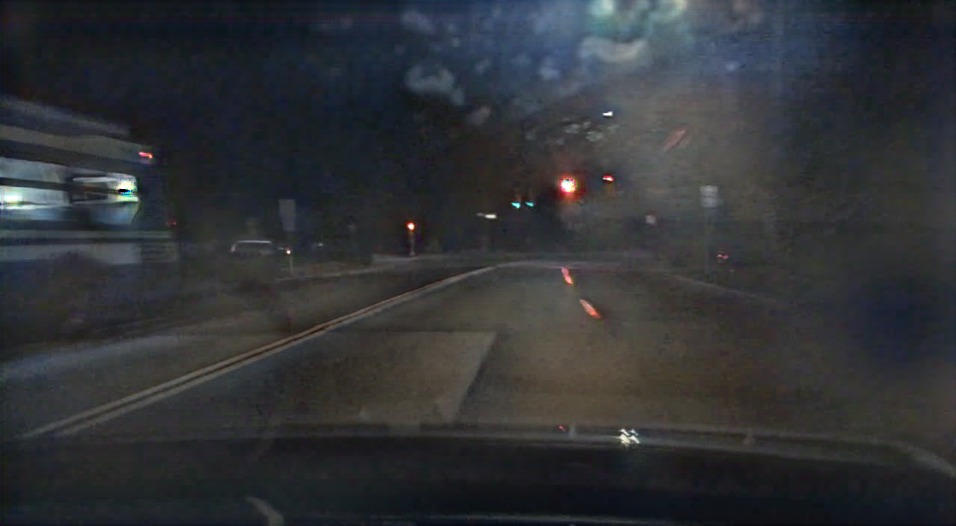}}\hfill
		{\includegraphics[width=0.165\textwidth]{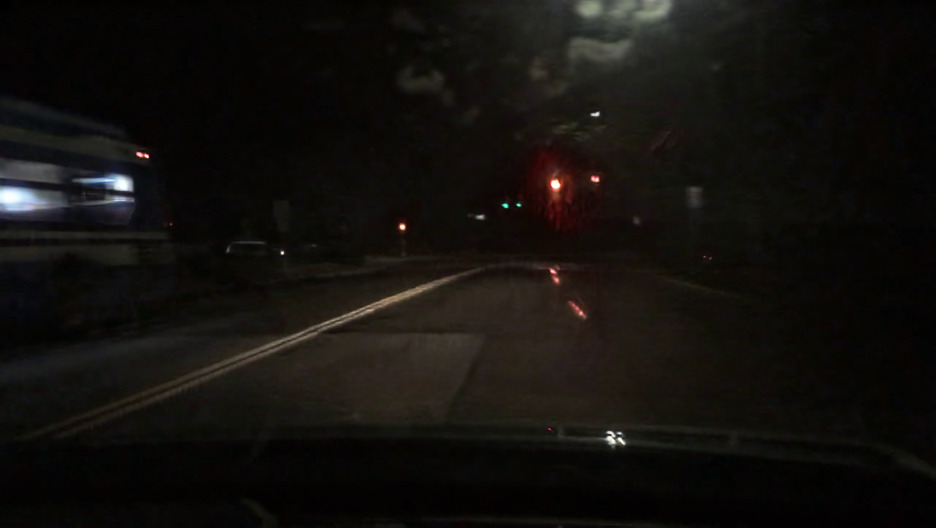}}\hfill
		{\includegraphics[width=0.165\textwidth]{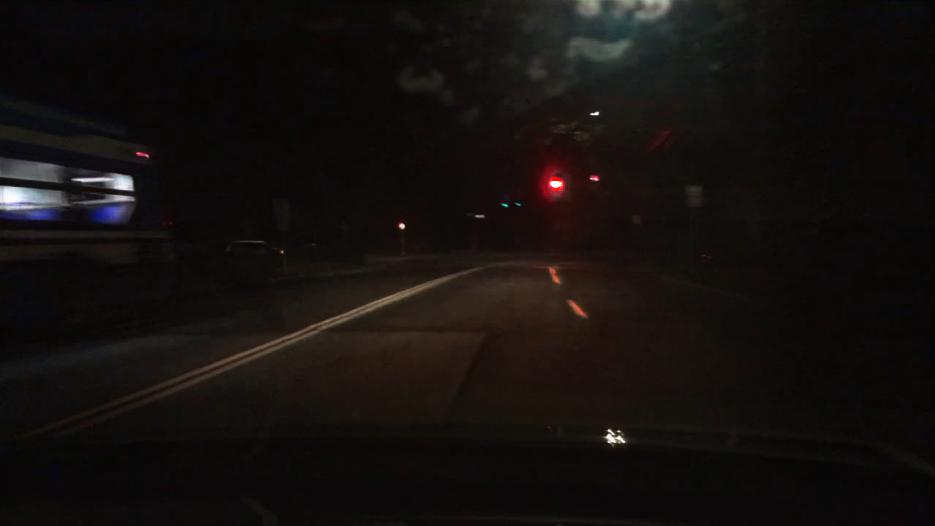}}\hfill\\
		\vspace{-10pt}
		\subfloat[Input]
		{\includegraphics[width=0.165\textwidth]{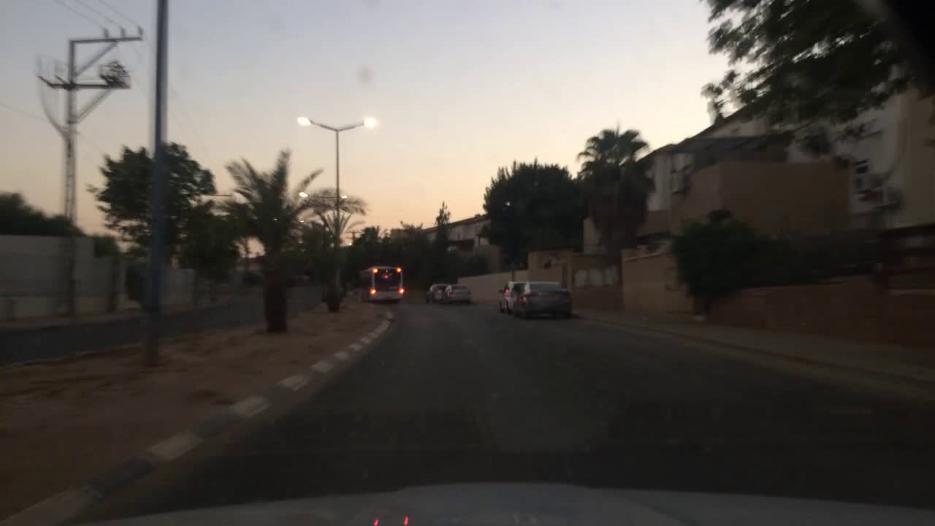}}\hfill
		\subfloat[MUNIT]
		{\includegraphics[width=0.165\textwidth]{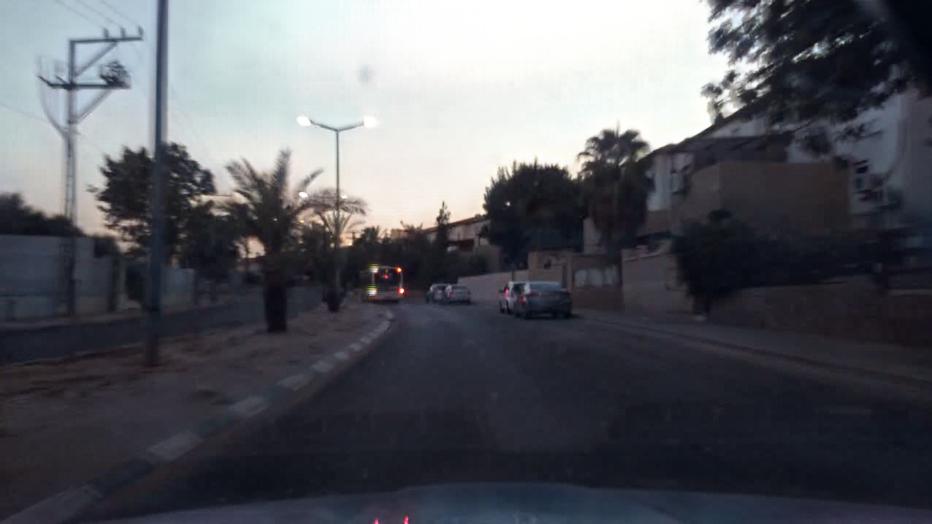}}\hfill
		\subfloat[CUT]
		{\includegraphics[width=0.165\textwidth]{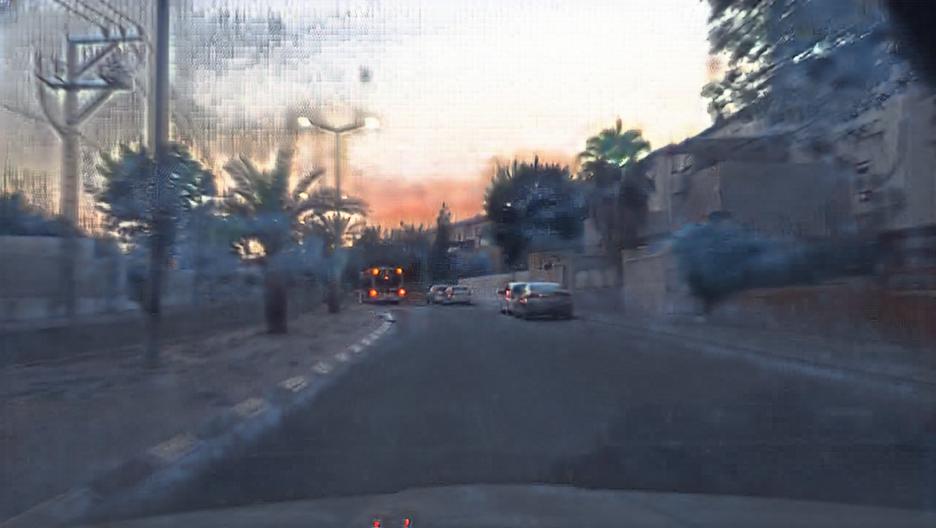}}\hfill
		\subfloat[TSIT]
		{\includegraphics[width=0.165\textwidth]{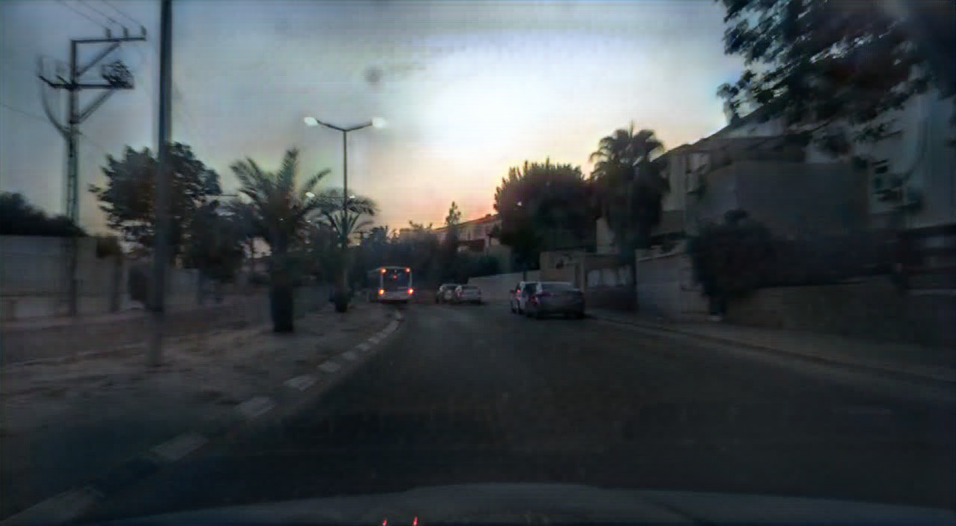}}\hfill
		\subfloat[QS-Attn]
		{\includegraphics[width=0.165\textwidth]{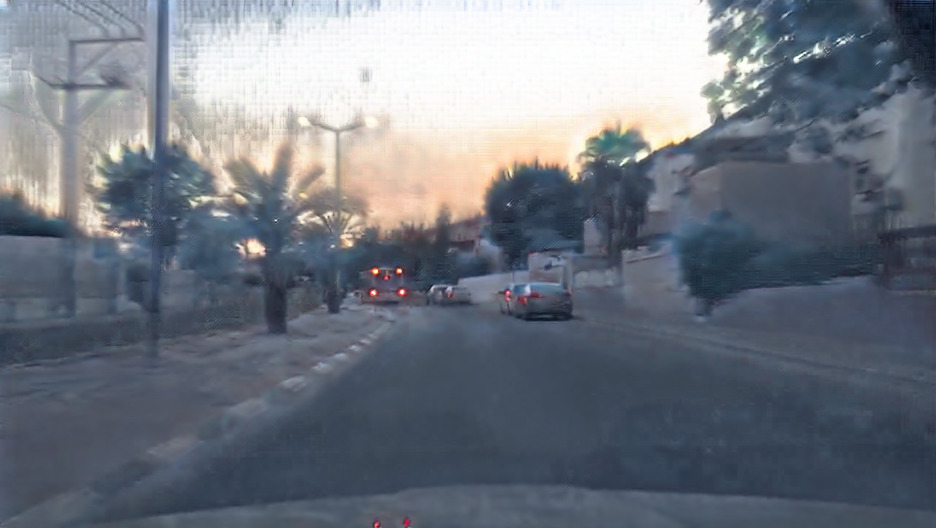}}\hfill
		\subfloat[FeaMGAN (ours)]
		{\includegraphics[width=0.165\textwidth]{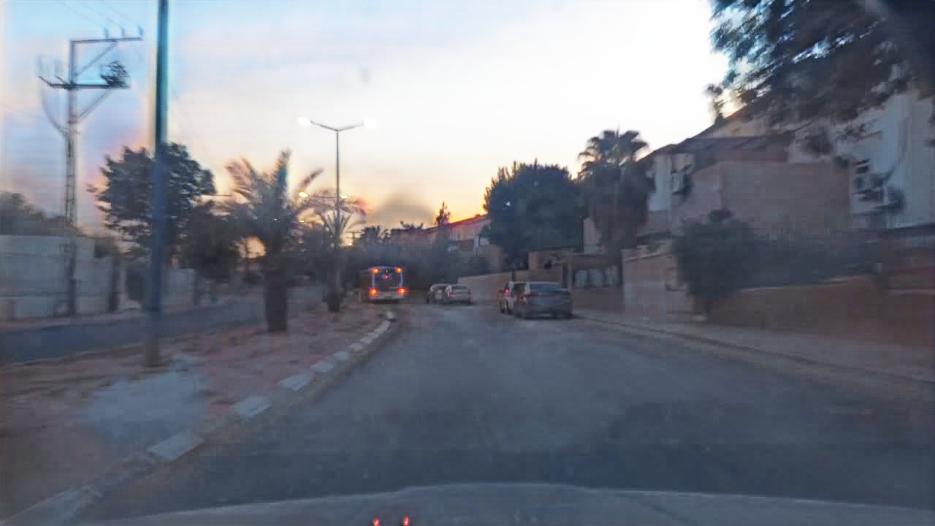}}\hfill
	\end{center}
	\vspace{-1ex}
	\caption[Qualitative comparison to prior work.]{Qualitative comparison to prior work. Results are randomly sampled from the best model. Best viewed in color.}
	\label{fig:feamgan:app:qualitative_comparison_additional_random}
\end{figure}

\begin{figure}[h] 
	\captionsetup[subfigure]{labelformat=empty}
	\begin{center}
		{\includegraphics[width=0.198\textwidth]{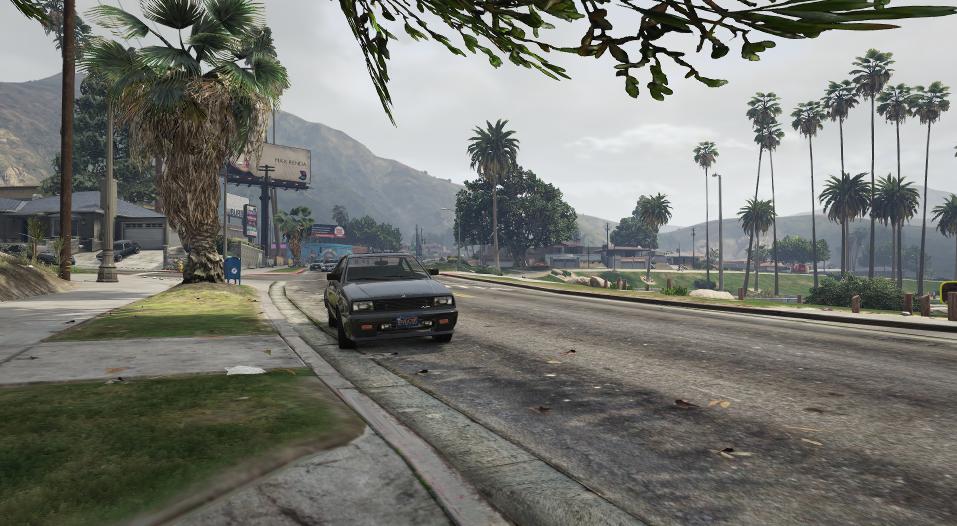}}\hfill
		{\includegraphics[width=0.198\textwidth]{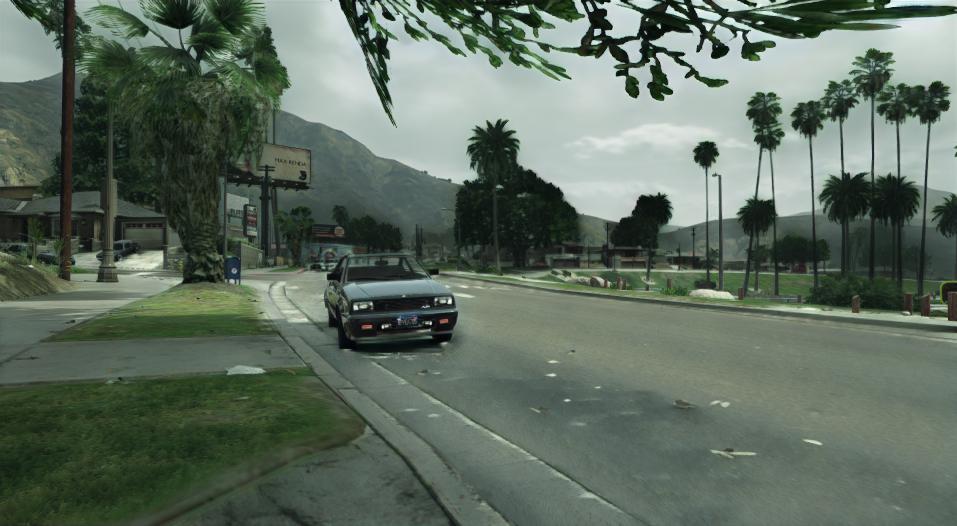}}\hfill
		{\includegraphics[width=0.198\textwidth]{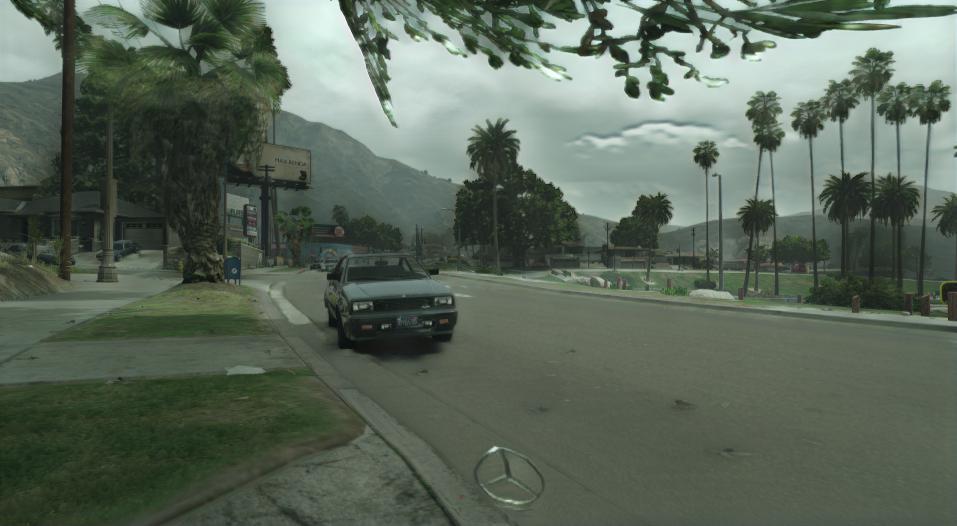}}\hfill
		{\includegraphics[width=0.198\textwidth]{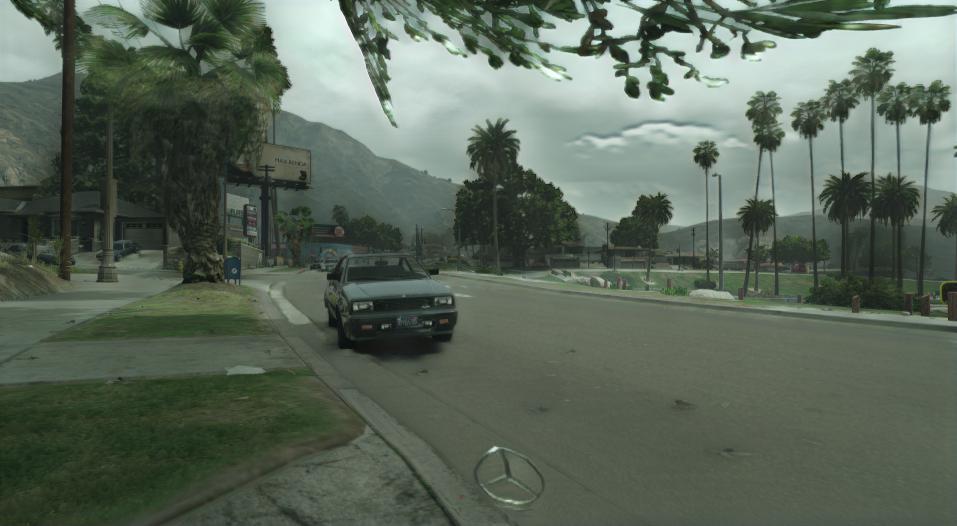}}\hfill
		{\includegraphics[width=0.198\textwidth]{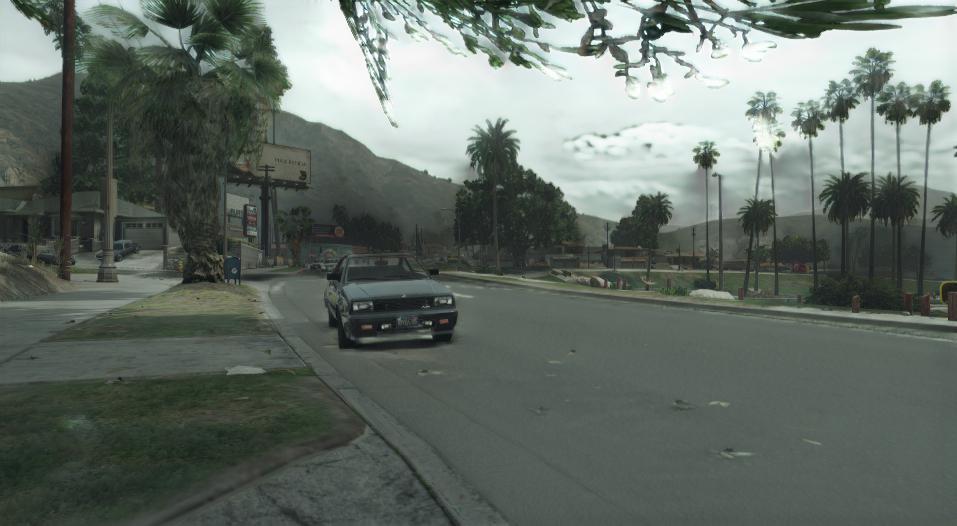}}\hfill \\ 
		
		{\includegraphics[width=0.198\textwidth]{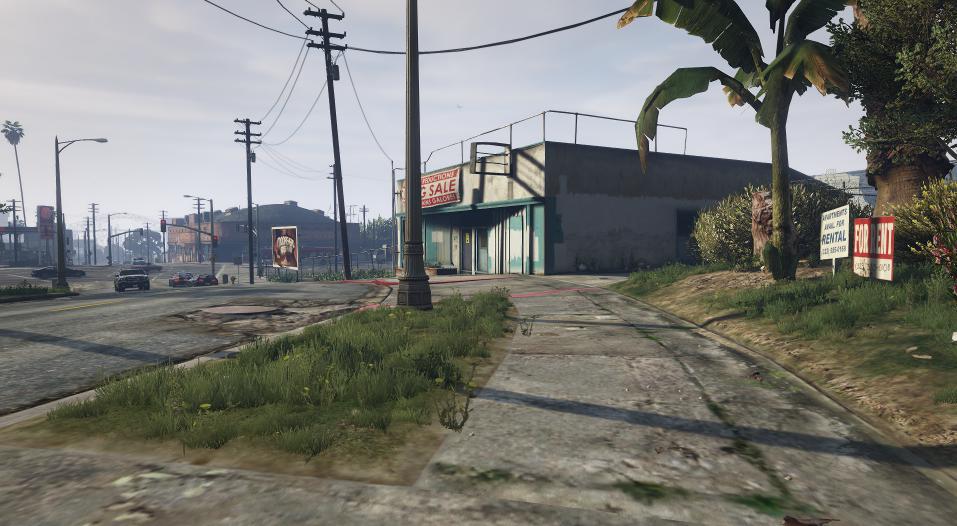}}\hfill
		{\includegraphics[width=0.198\textwidth]{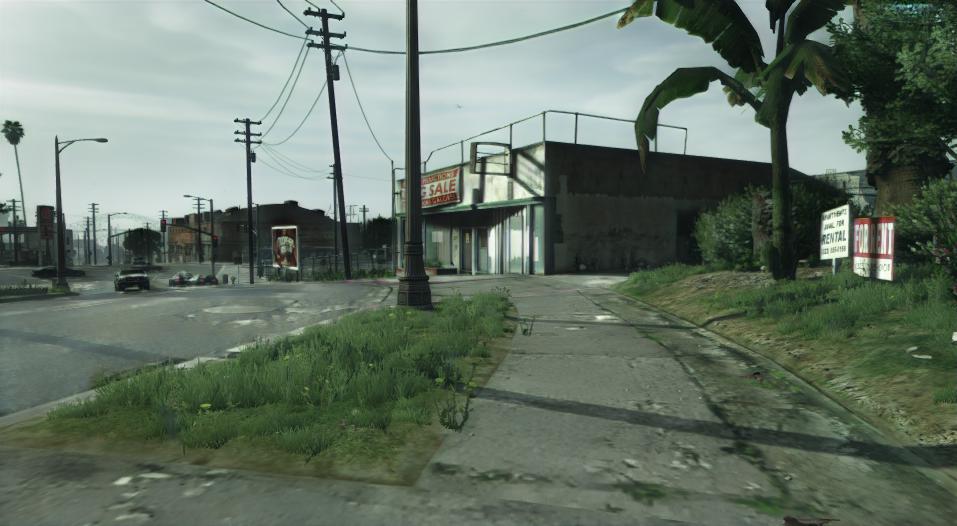}}\hfill
		{\includegraphics[width=0.198\textwidth]{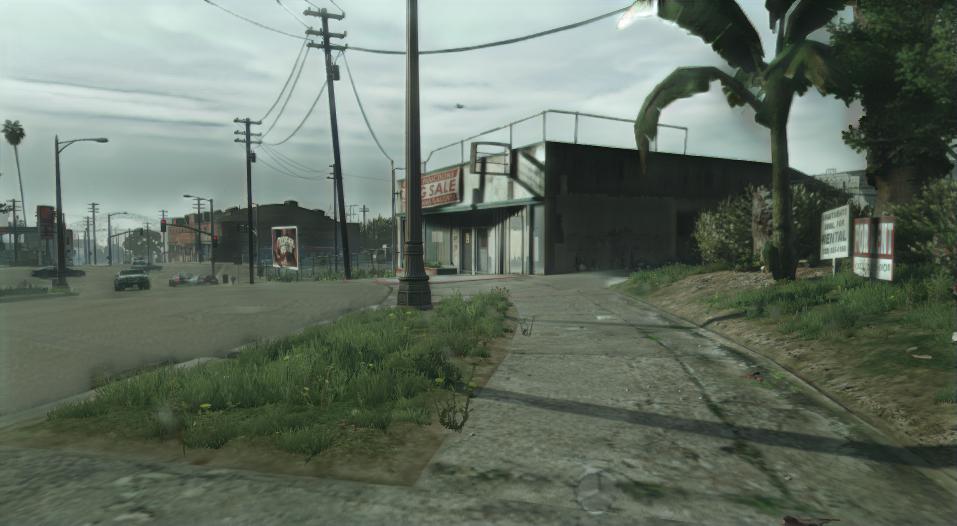}}\hfill
		{\includegraphics[width=0.198\textwidth]{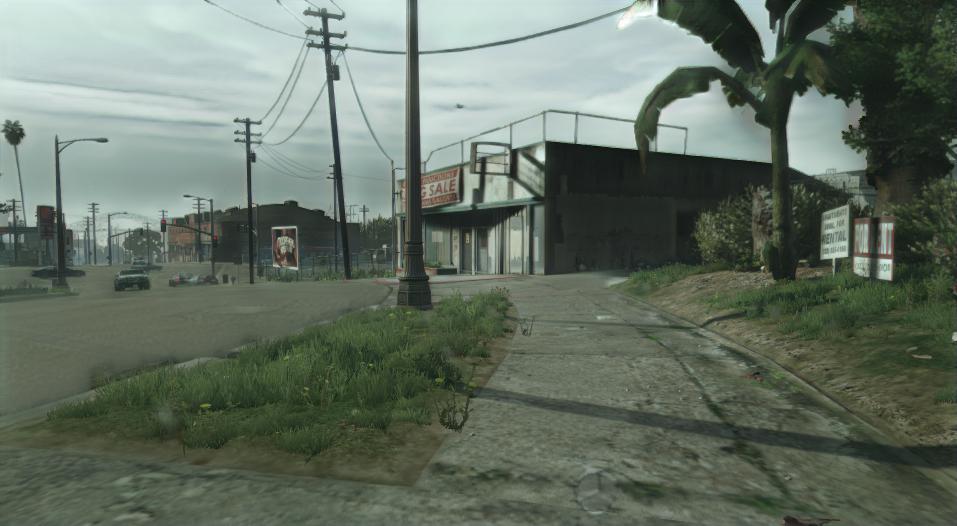}}\hfill
		{\includegraphics[width=0.198\textwidth]{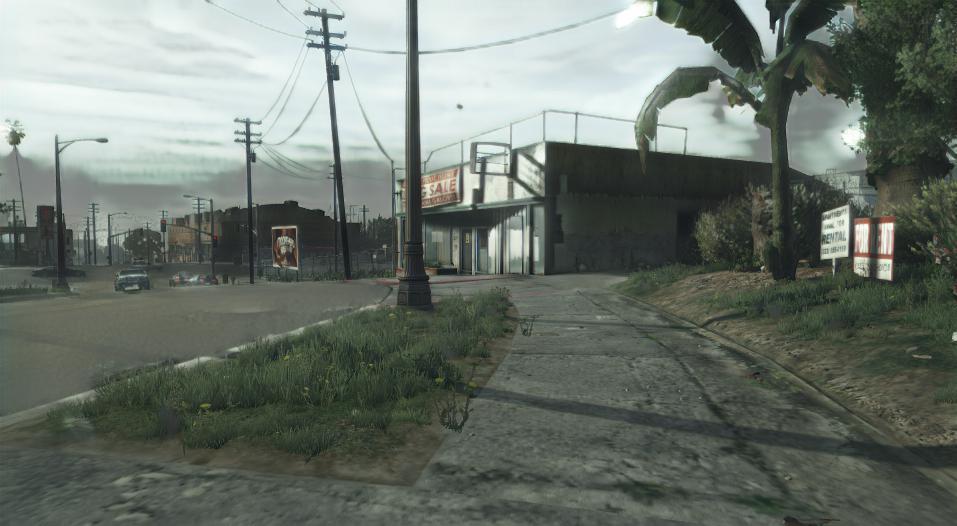}}\hfill \\ 
		
		{\includegraphics[width=0.198\textwidth]{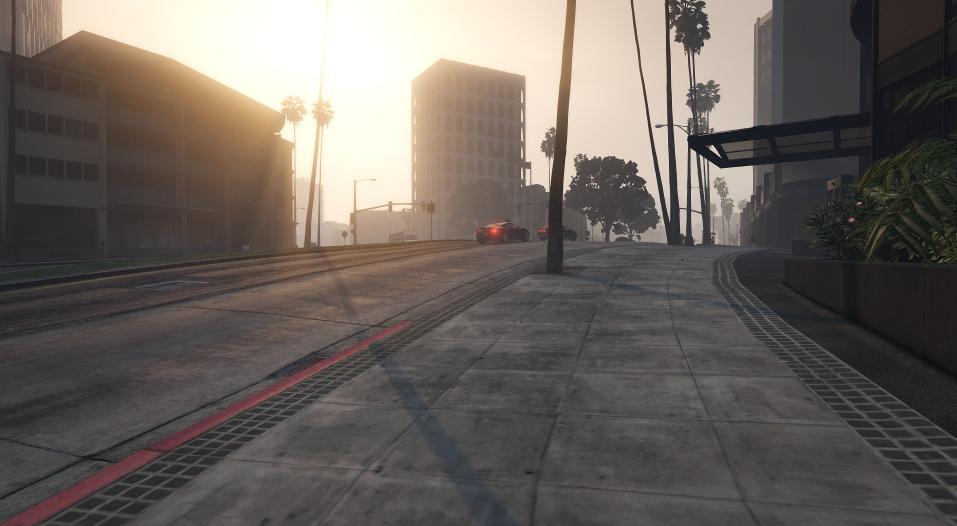}}\hfill
		{\includegraphics[width=0.198\textwidth]{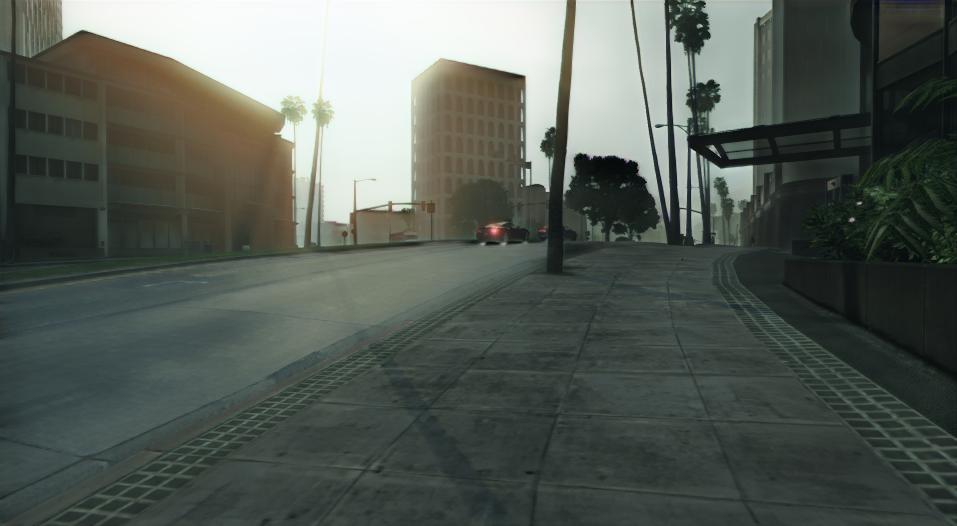}}\hfill
		{\includegraphics[width=0.198\textwidth]{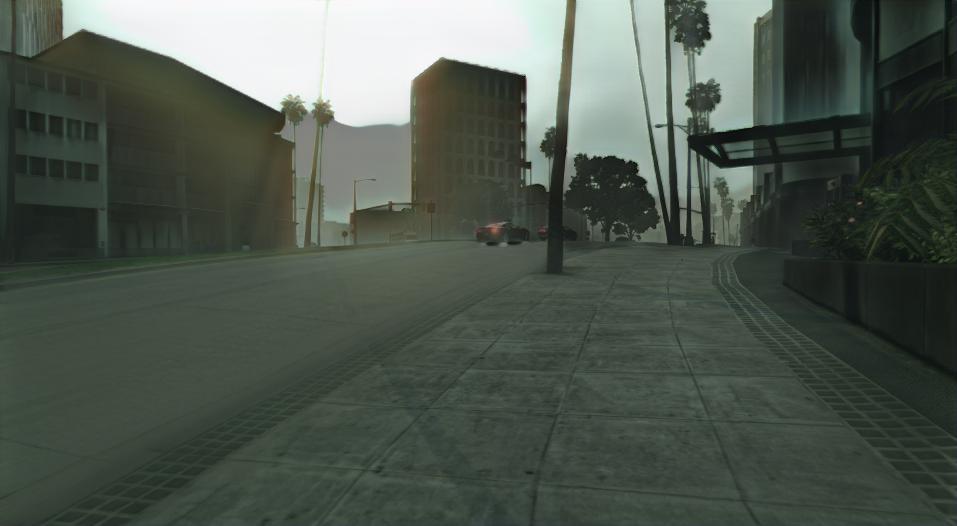}}\hfill
		{\includegraphics[width=0.198\textwidth]{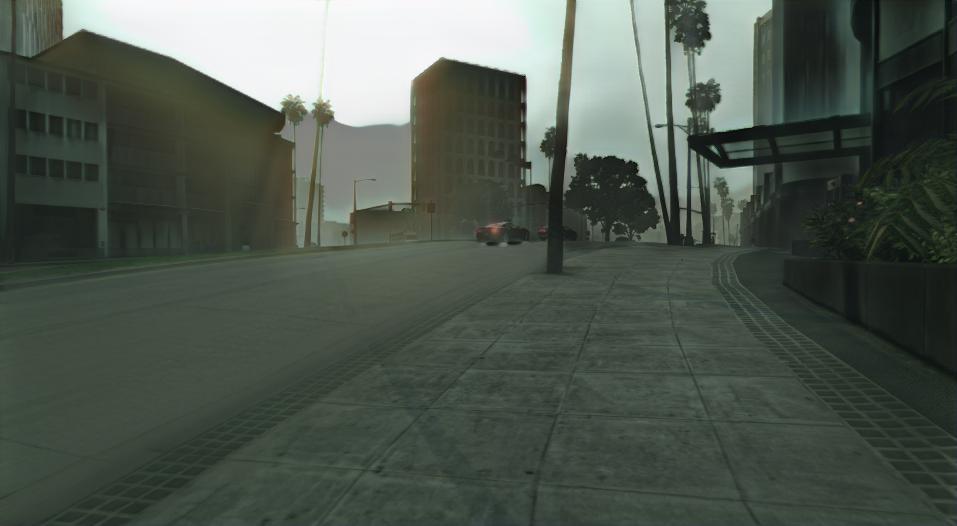}}\hfill
		{\includegraphics[width=0.198\textwidth]{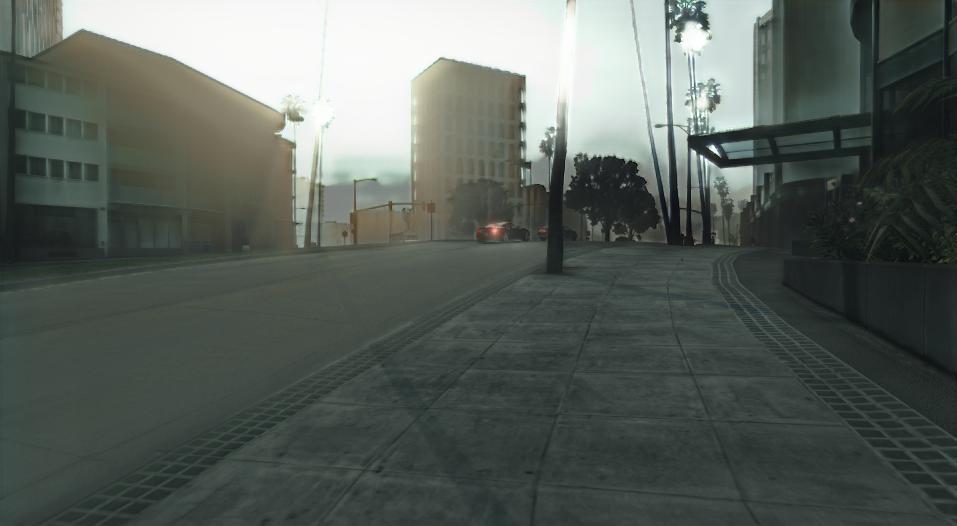}}\hfill \\ 
		
		{\includegraphics[width=0.198\textwidth]{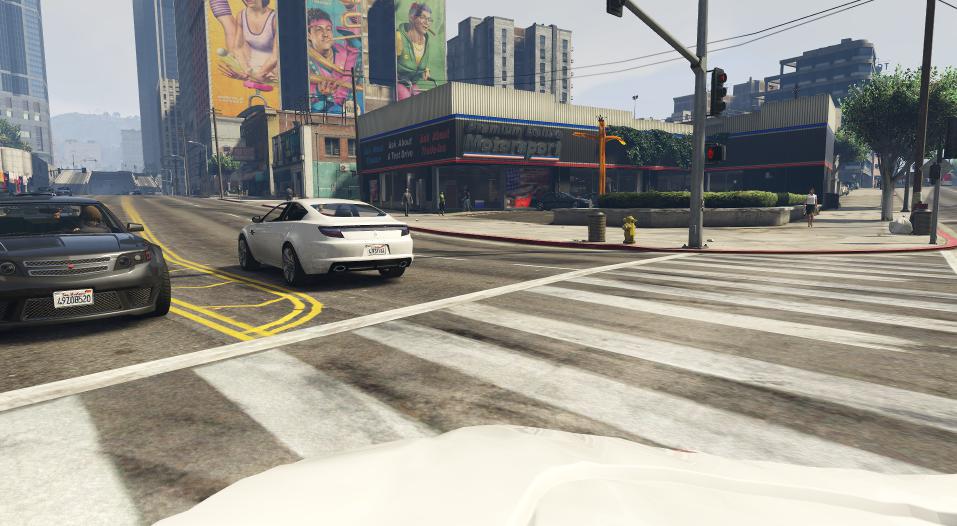}}\hfill
		{\includegraphics[width=0.198\textwidth]{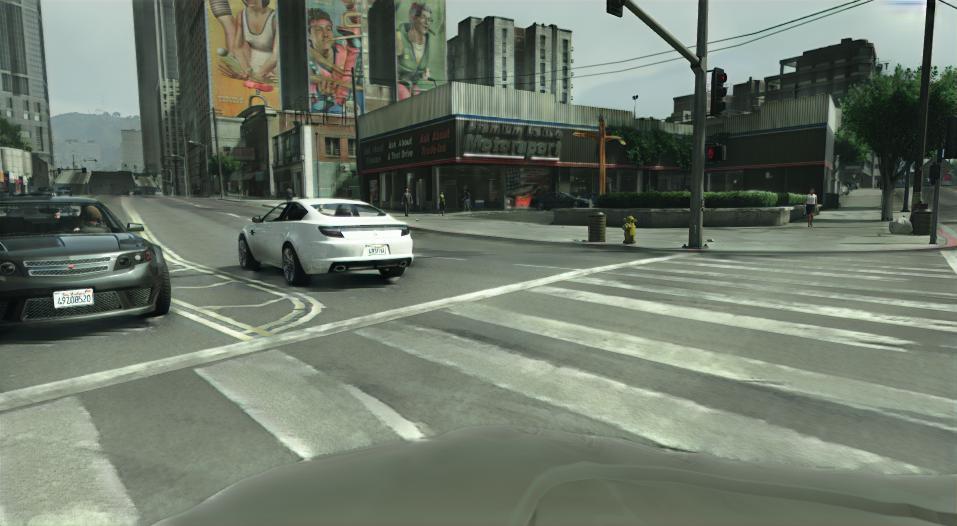}}\hfill
		{\includegraphics[width=0.198\textwidth]{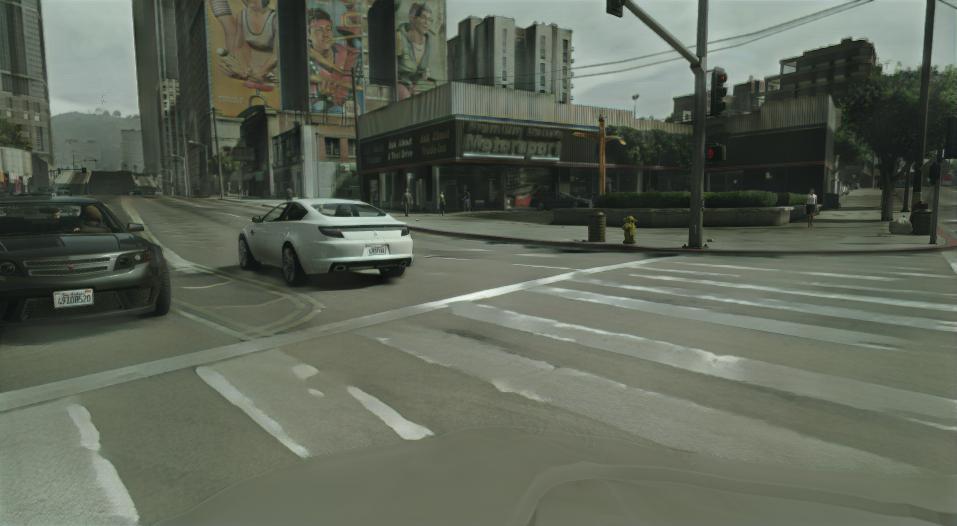}}\hfill
		{\includegraphics[width=0.198\textwidth]{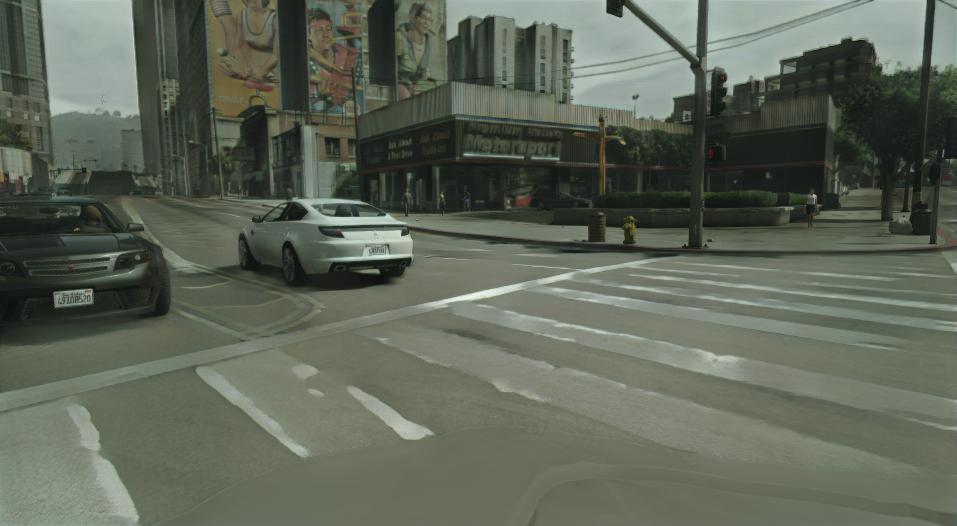}}\hfill
		{\includegraphics[width=0.198\textwidth]{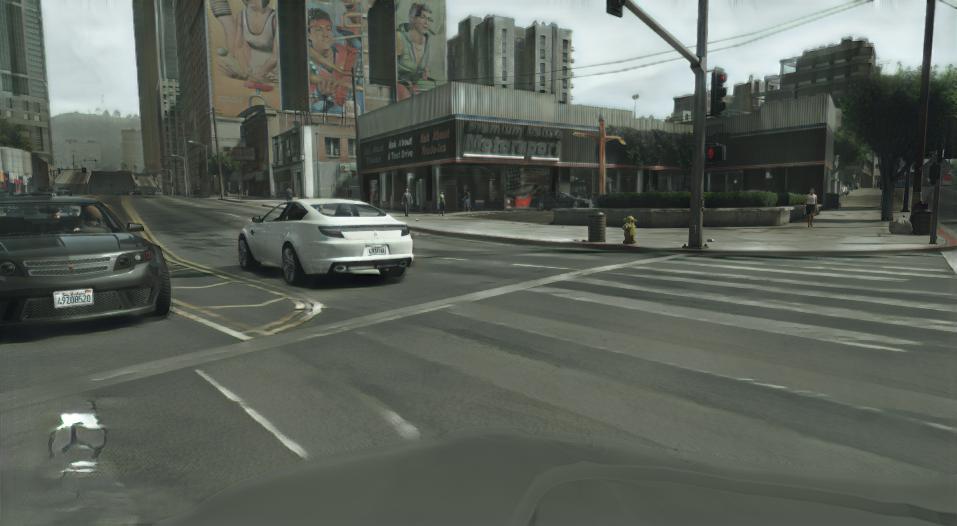}}\hfill \\ 
		
		{\includegraphics[width=0.198\textwidth]{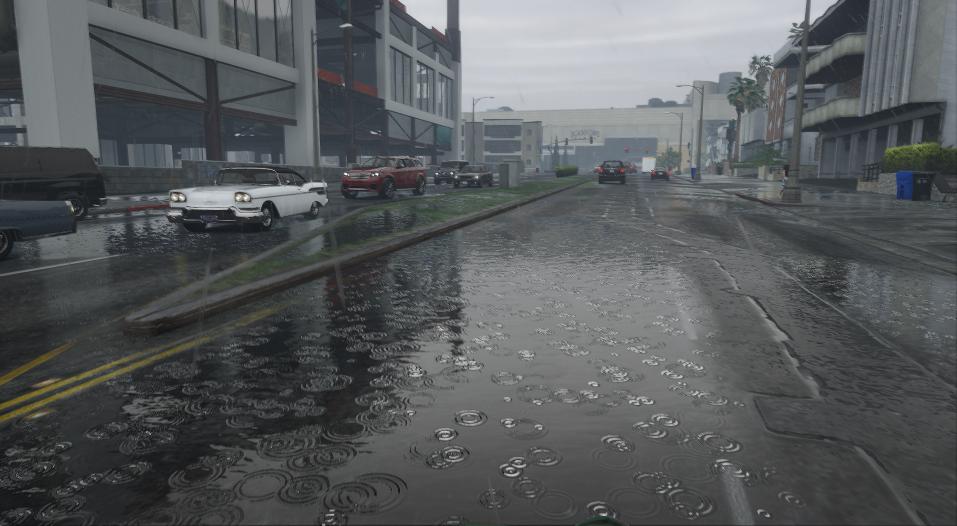}}\hfill
		{\includegraphics[width=0.198\textwidth]{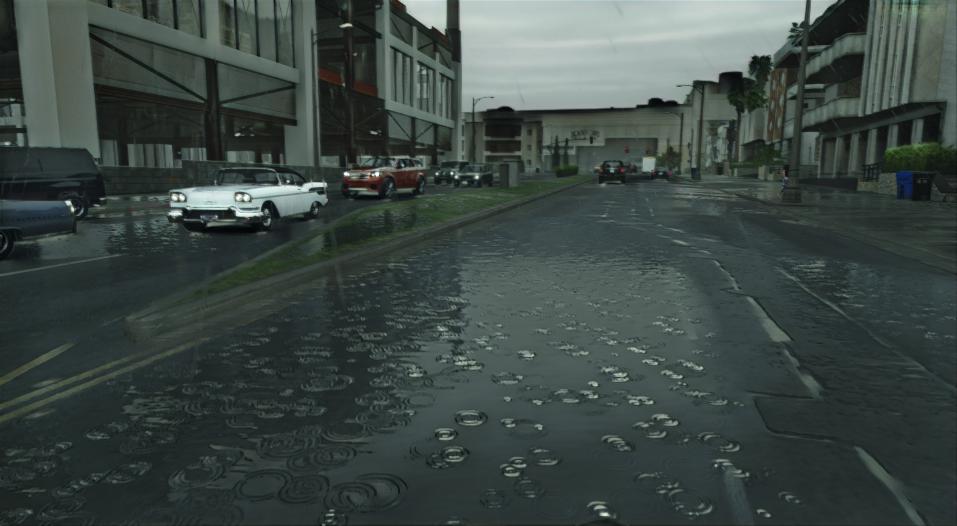}}\hfill
		{\includegraphics[width=0.198\textwidth]{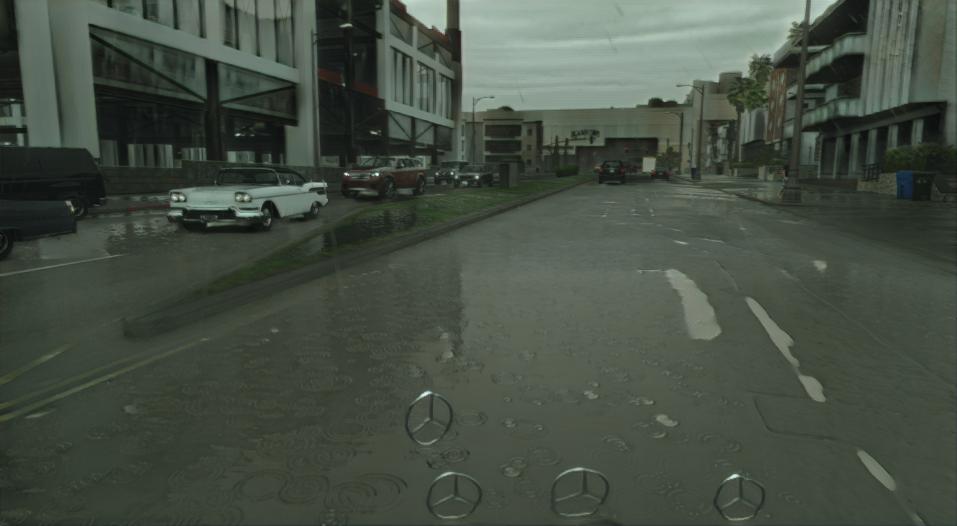}}\hfill
		{\includegraphics[width=0.198\textwidth]{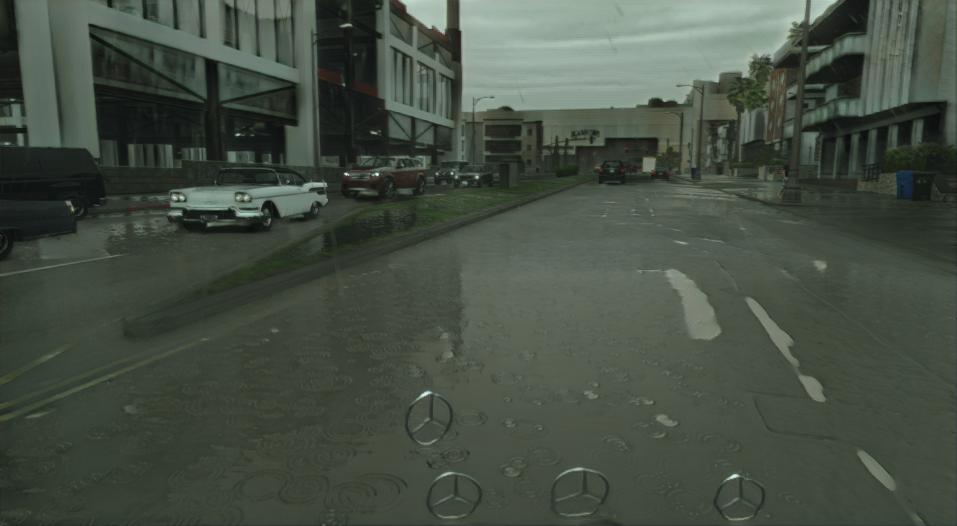}}\hfill
		{\includegraphics[width=0.198\textwidth]{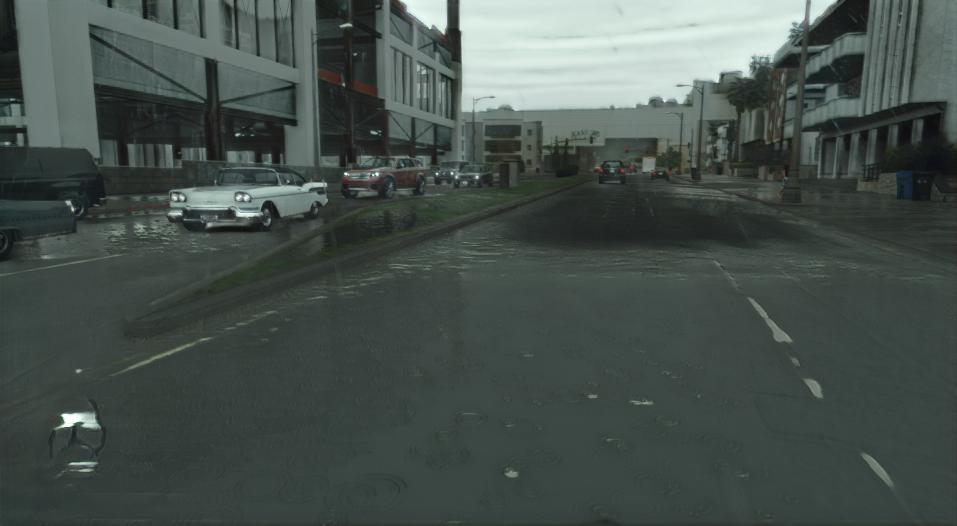}}\hfill \\ 
		
		{\includegraphics[width=0.198\textwidth]{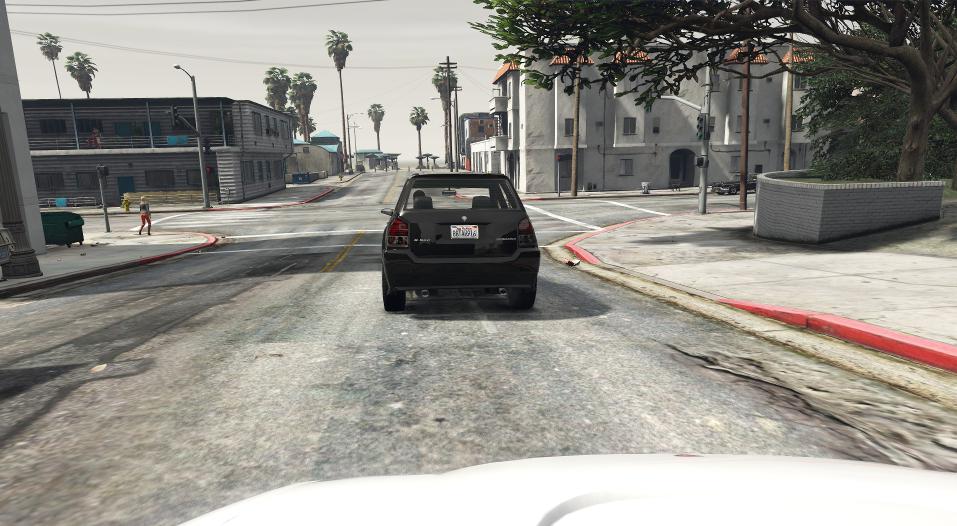}}\hfill
		{\includegraphics[width=0.198\textwidth]{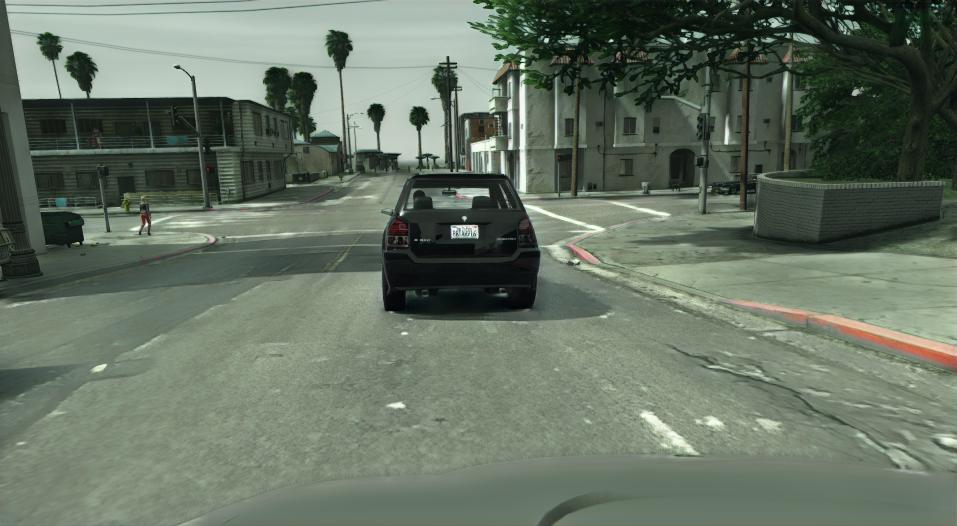}}\hfill
		{\includegraphics[width=0.198\textwidth]{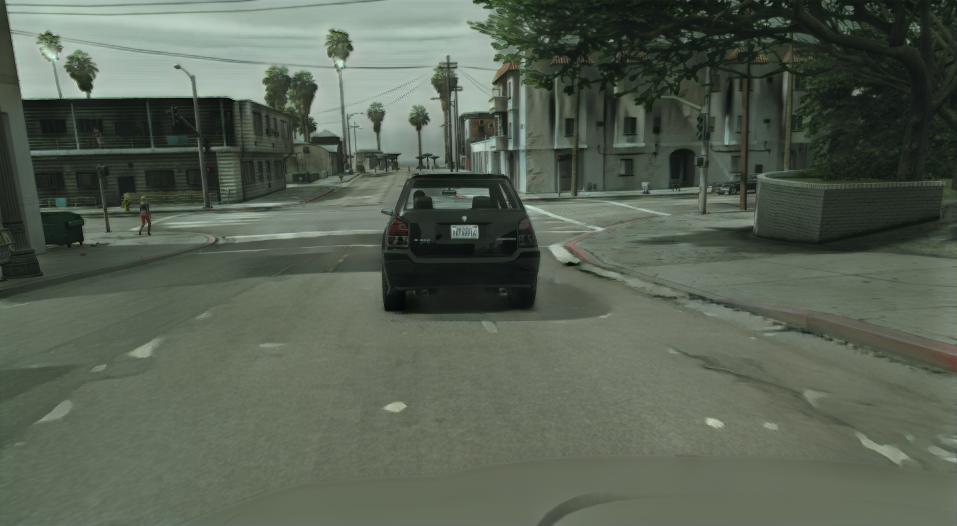}}\hfill
		{\includegraphics[width=0.198\textwidth]{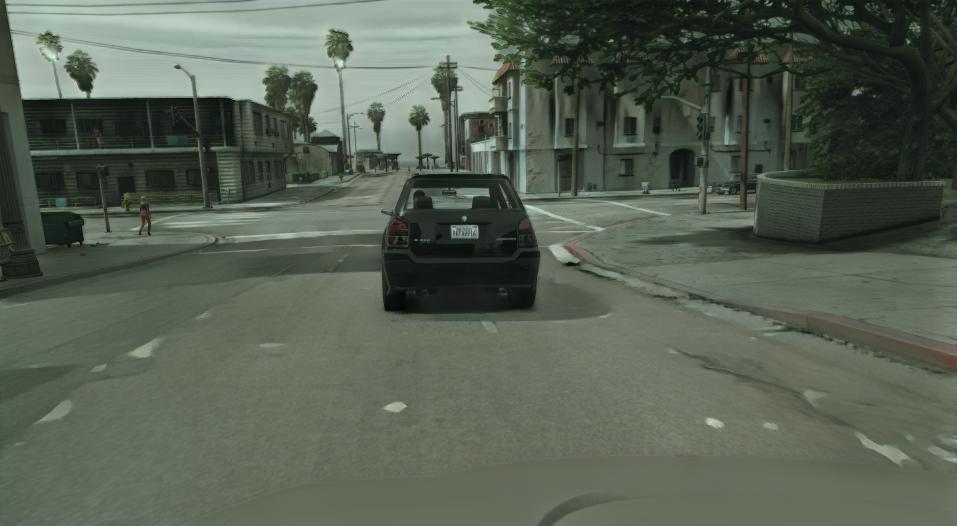}}\hfill
		{\includegraphics[width=0.198\textwidth]{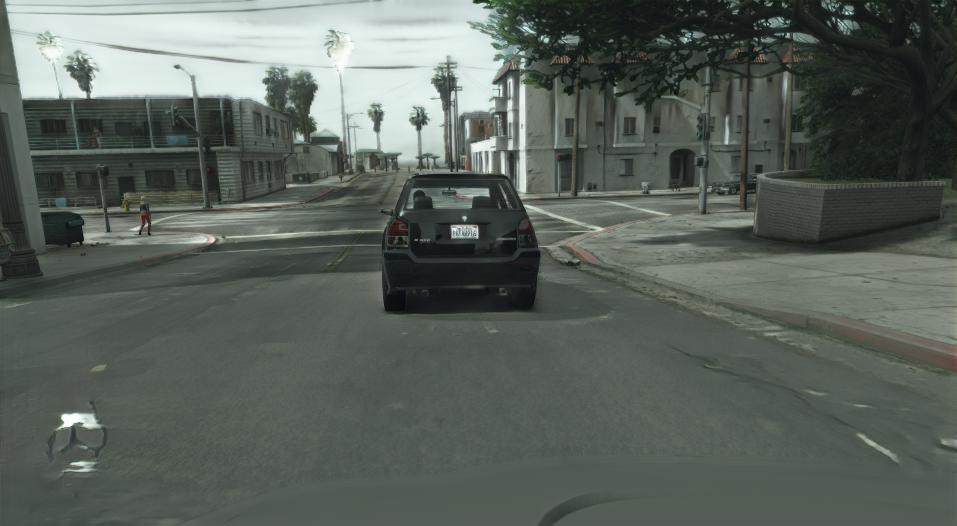}}\hfill \\ 
		
		{\includegraphics[width=0.198\textwidth]{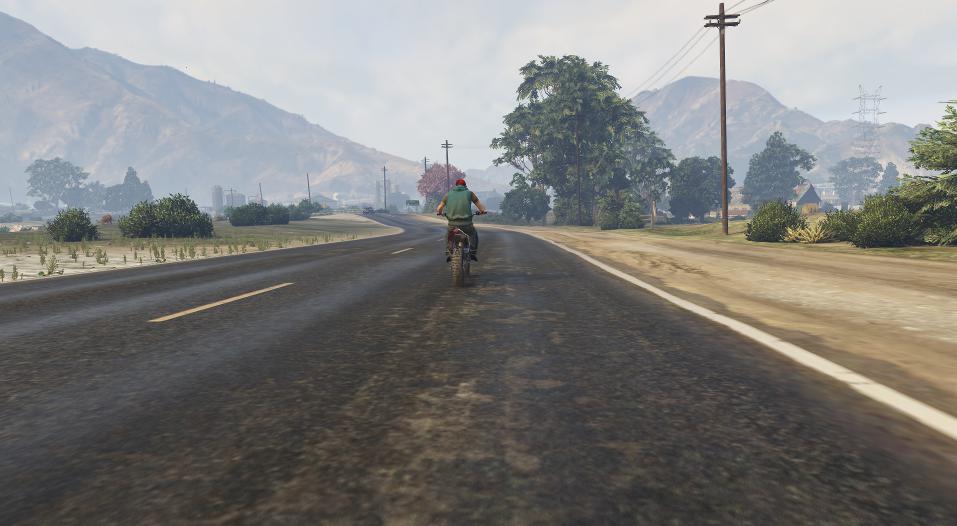}}\hfill
		{\includegraphics[width=0.198\textwidth]{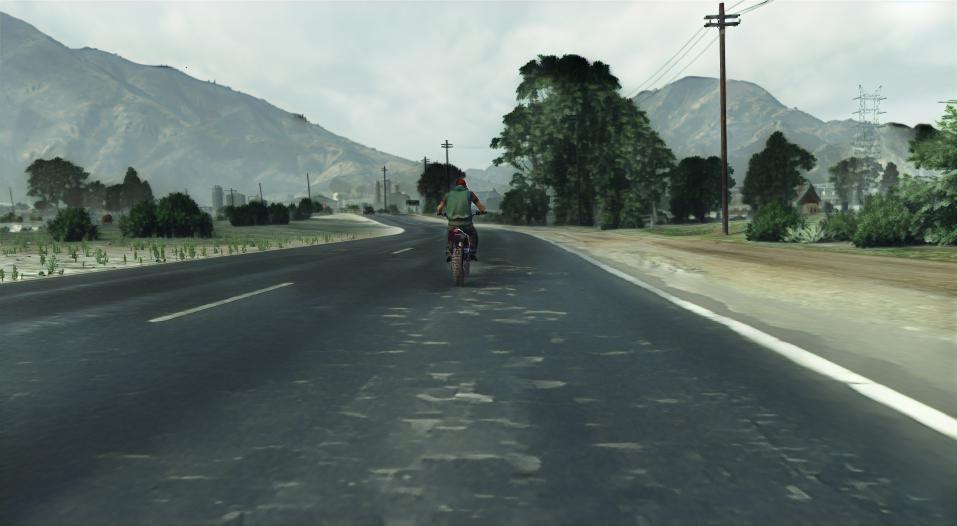}}\hfill
		{\includegraphics[width=0.198\textwidth]{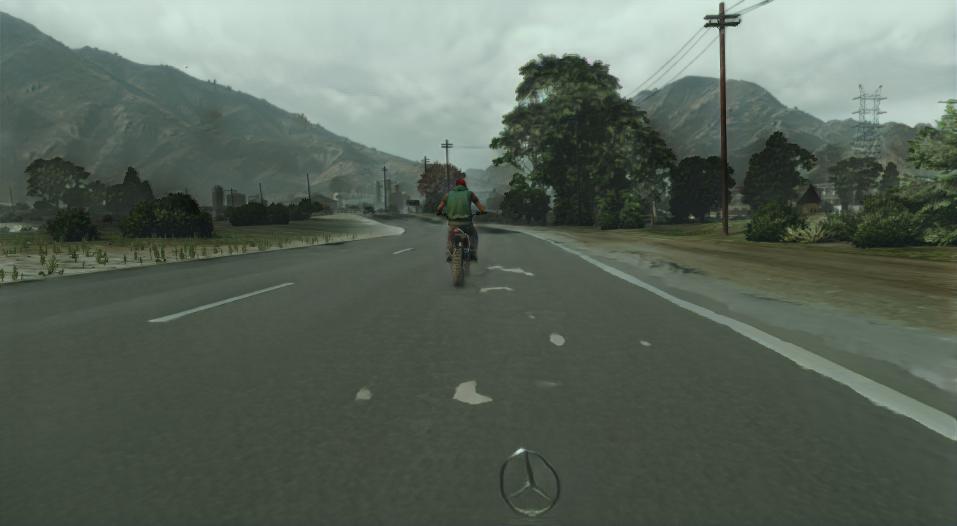}}\hfill
		{\includegraphics[width=0.198\textwidth]{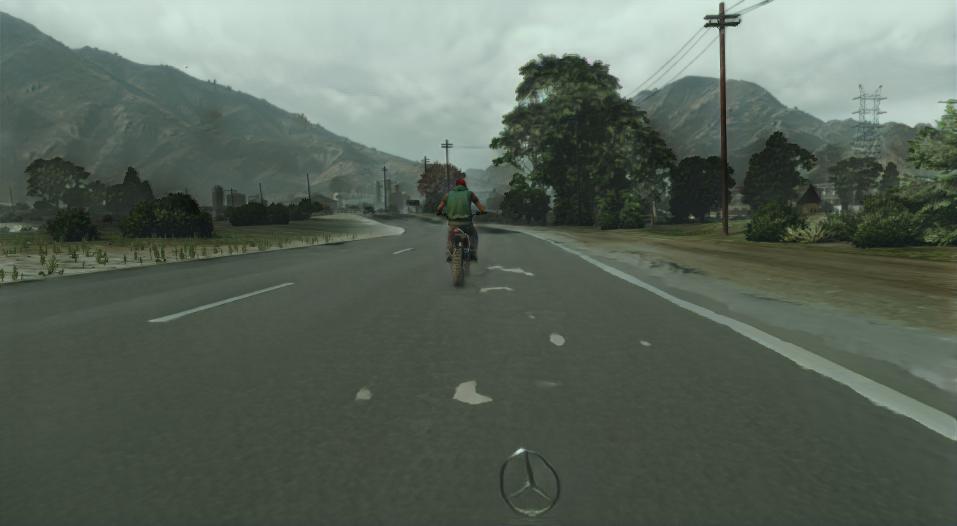}}\hfill
		{\includegraphics[width=0.198\textwidth]{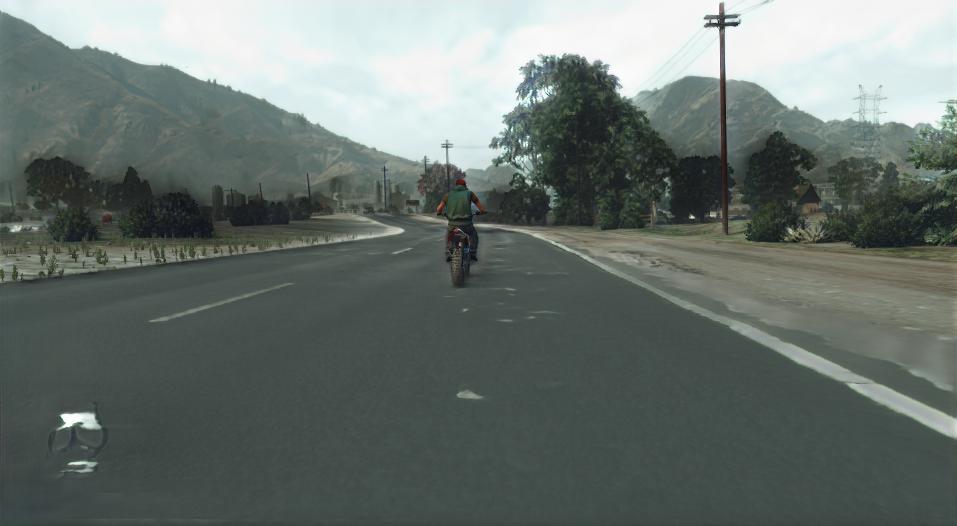}}\hfill \\ 
		
		{\includegraphics[width=0.198\textwidth]{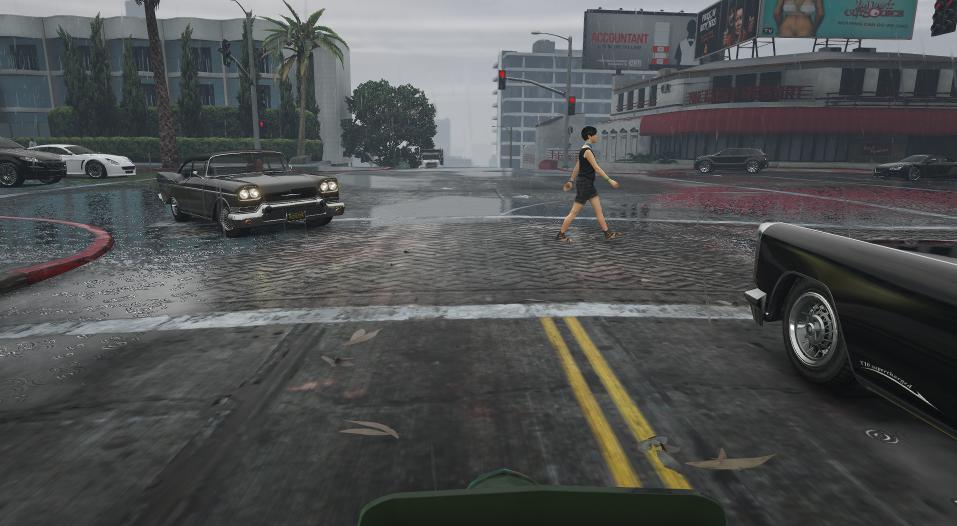}}\hfill
		{\includegraphics[width=0.198\textwidth]{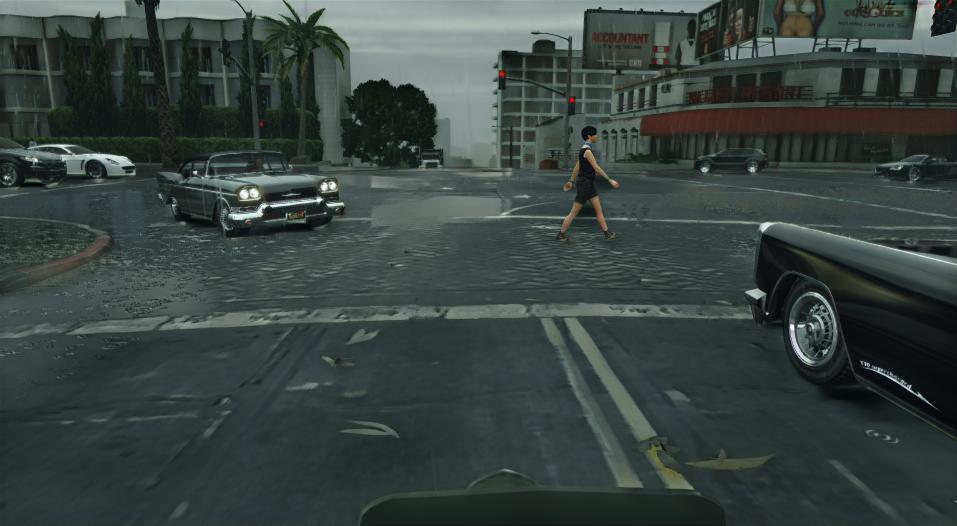}}\hfill
		{\includegraphics[width=0.198\textwidth]{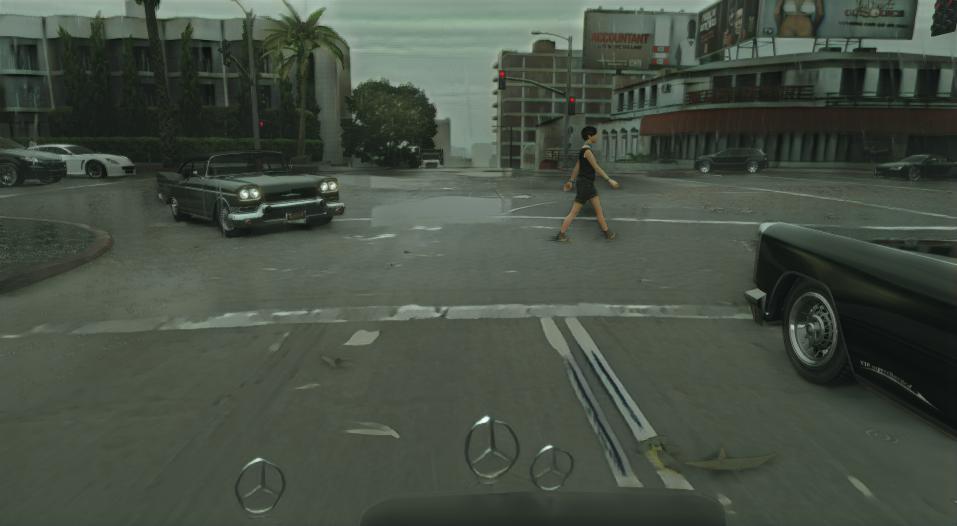}}\hfill
		{\includegraphics[width=0.198\textwidth]{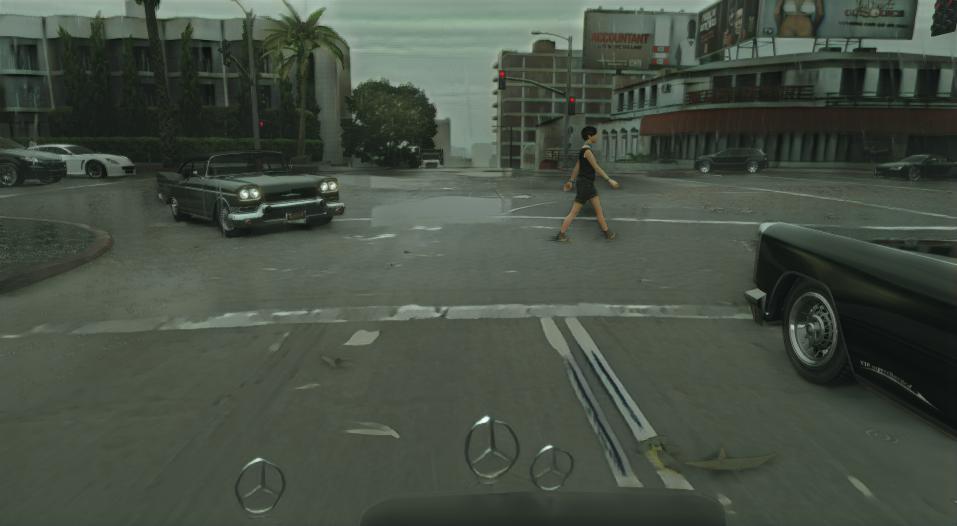}}\hfill
		{\includegraphics[width=0.198\textwidth]{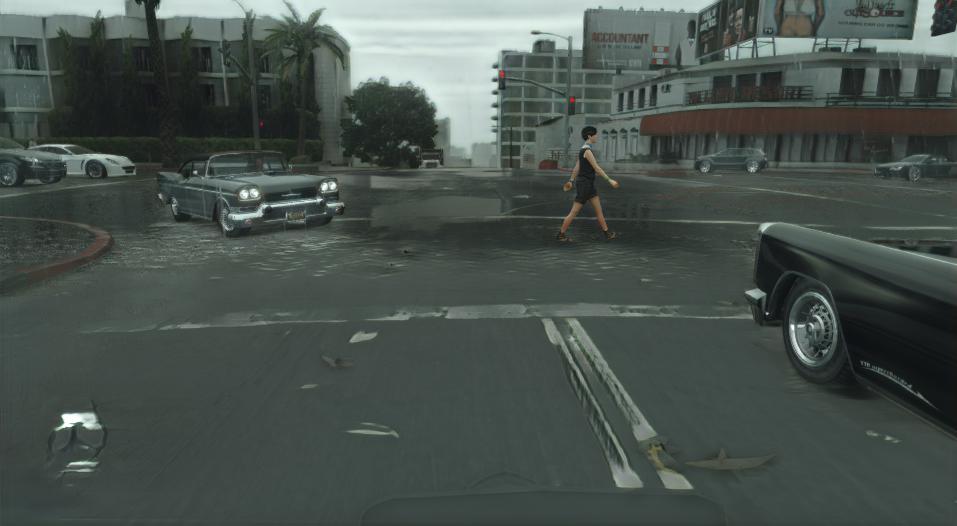}}\hfill \\ 
		
		{\includegraphics[width=0.198\textwidth]{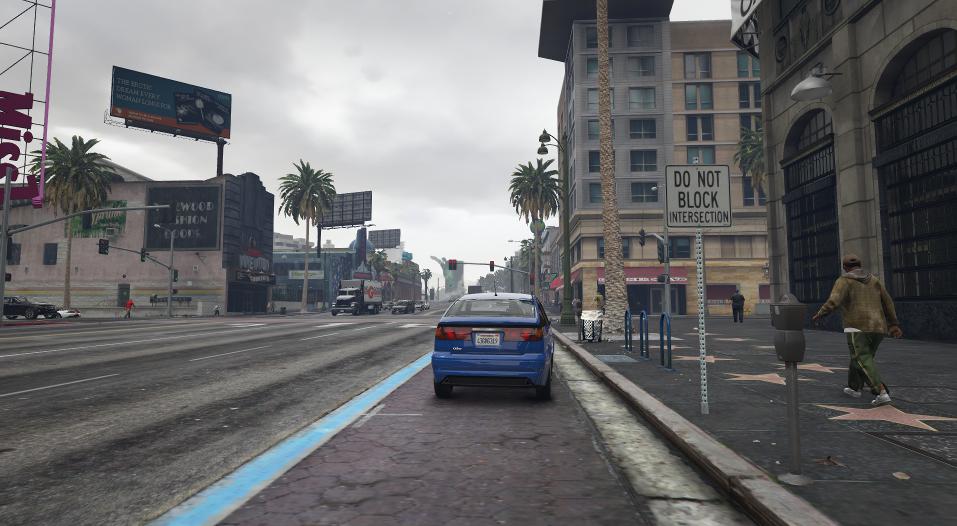}}\hfill
		{\includegraphics[width=0.198\textwidth]{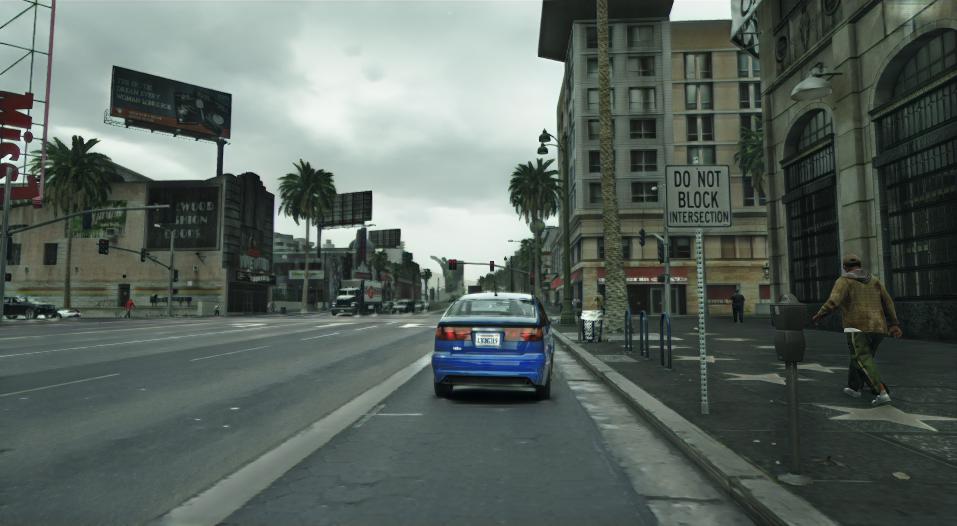}}\hfill
		{\includegraphics[width=0.198\textwidth]{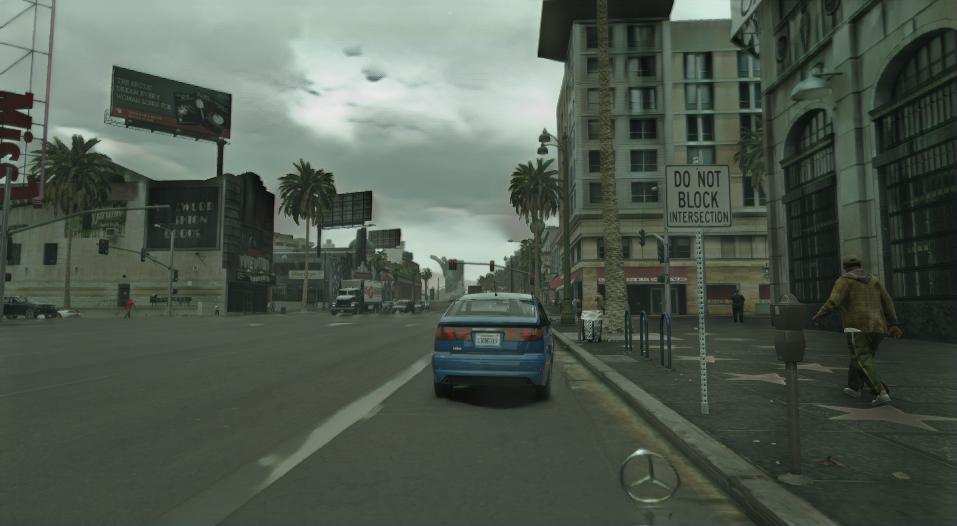}}\hfill
		{\includegraphics[width=0.198\textwidth]{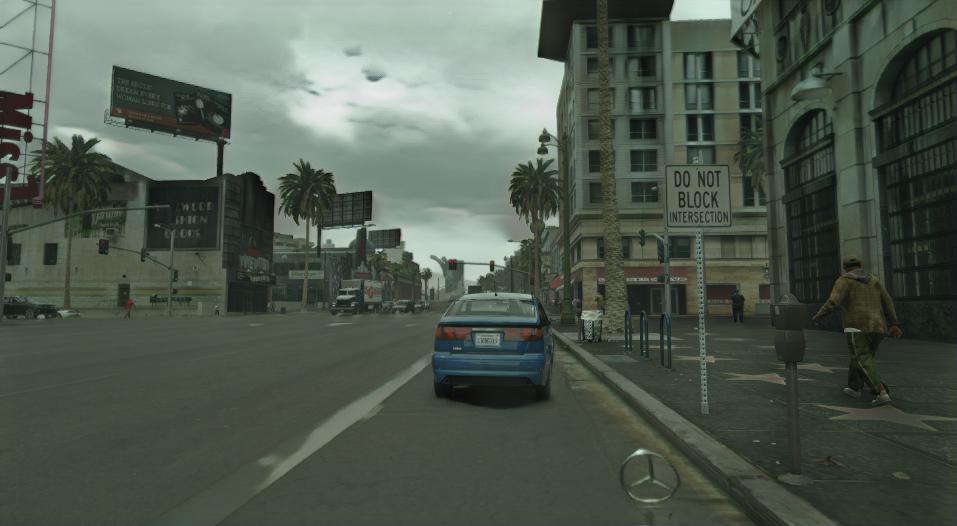}}\hfill
		{\includegraphics[width=0.198\textwidth]{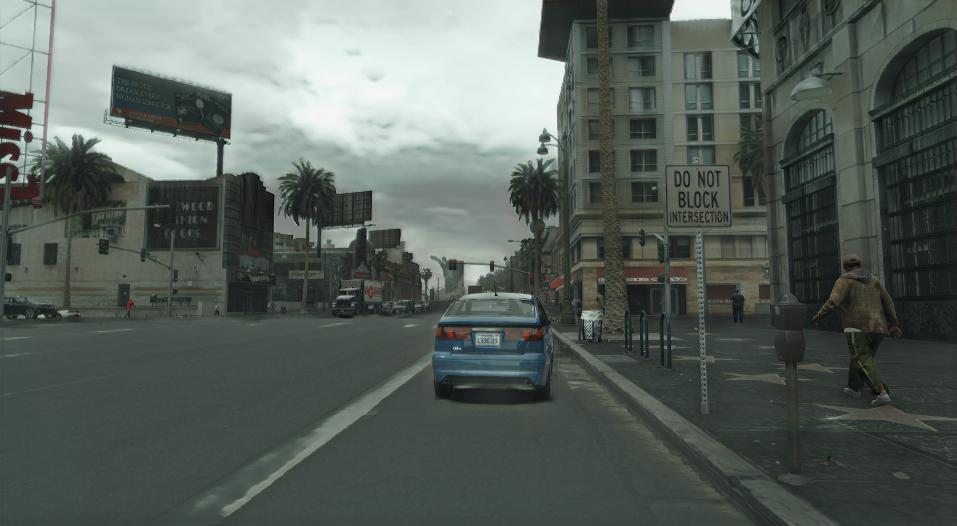}}\hfill \\ 
		
		{\includegraphics[width=0.198\textwidth]{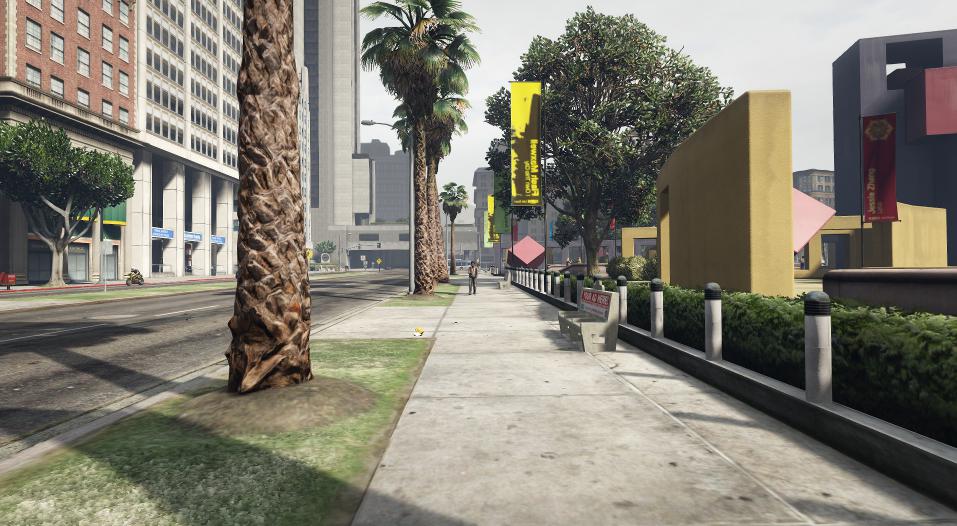}}\hfill
		{\includegraphics[width=0.198\textwidth]{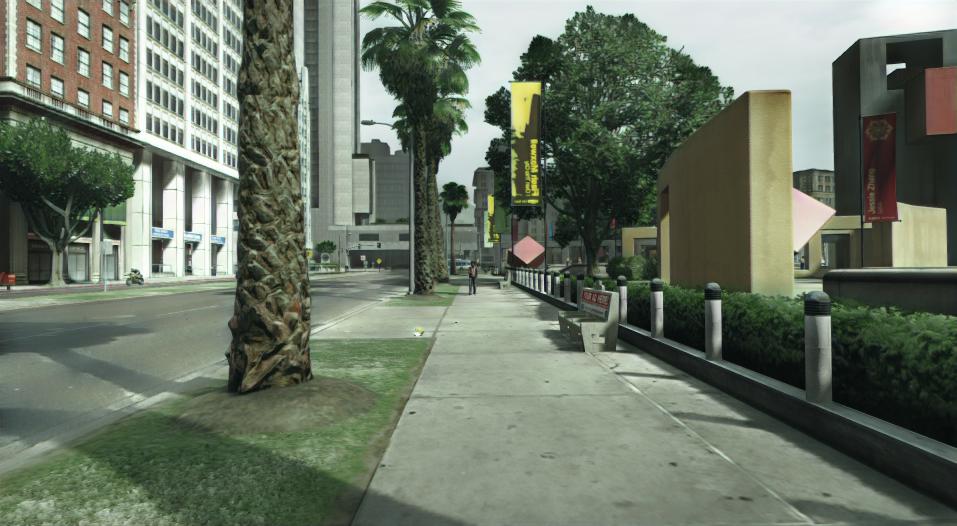}}\hfill
		{\includegraphics[width=0.198\textwidth]{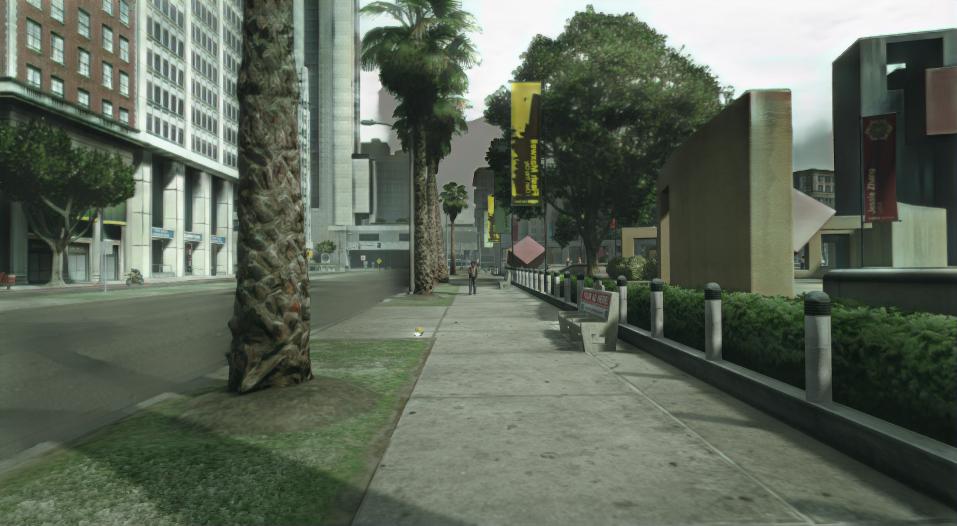}}\hfill
		{\includegraphics[width=0.198\textwidth]{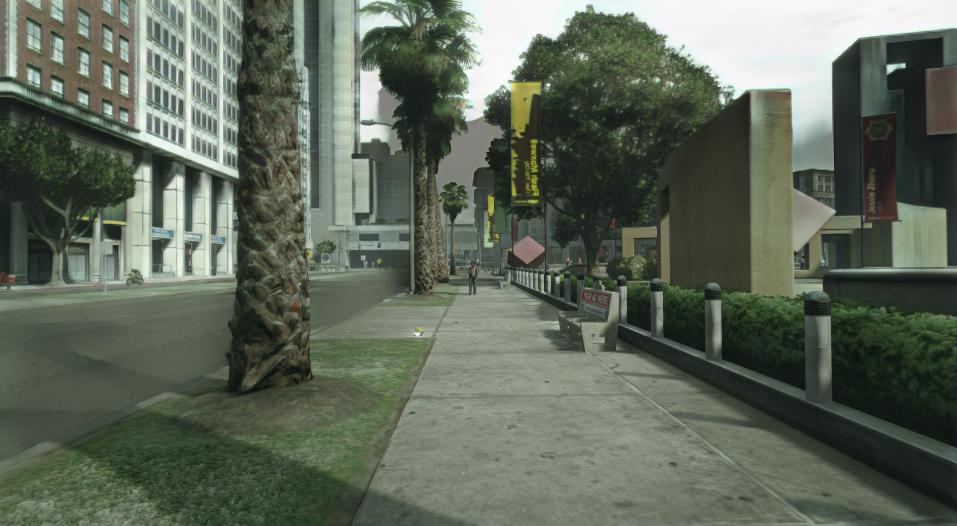}}\hfill
		{\includegraphics[width=0.198\textwidth]{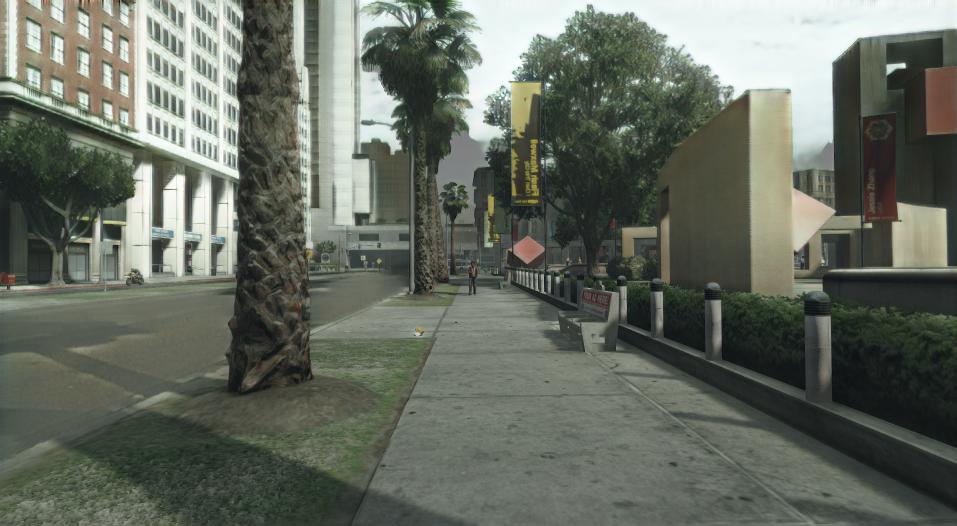}}\hfill \\
		\vspace{-10pt}
		\subfloat[Input]
		{\includegraphics[width=0.198\textwidth]{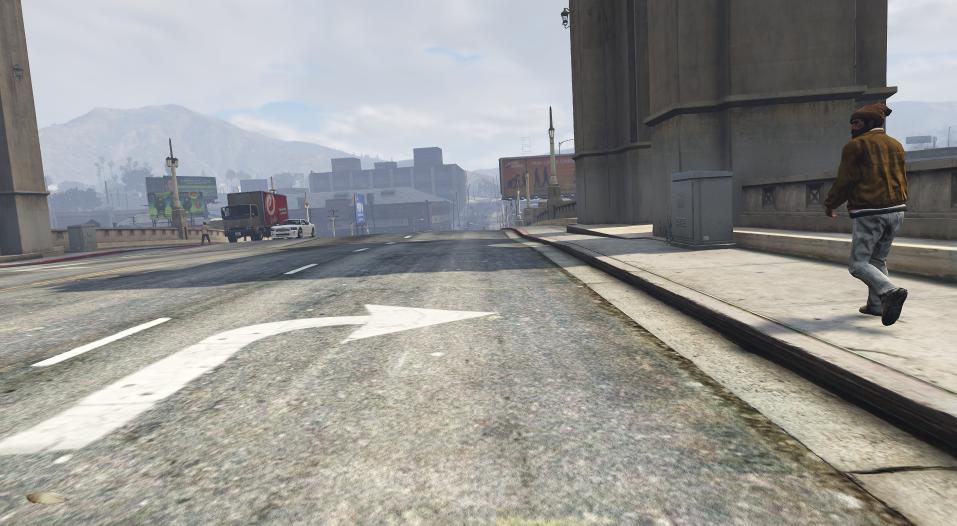}}\hfill
		\subfloat[252$\times$252]
		{\includegraphics[width=0.198\textwidth]{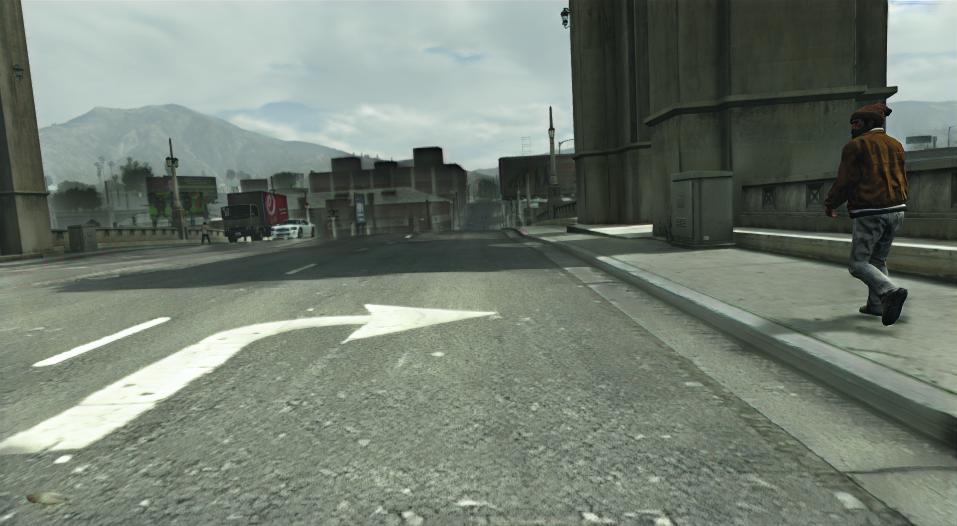}}\hfill
		\subfloat[352$\times$352]
		{\includegraphics[width=0.198\textwidth]{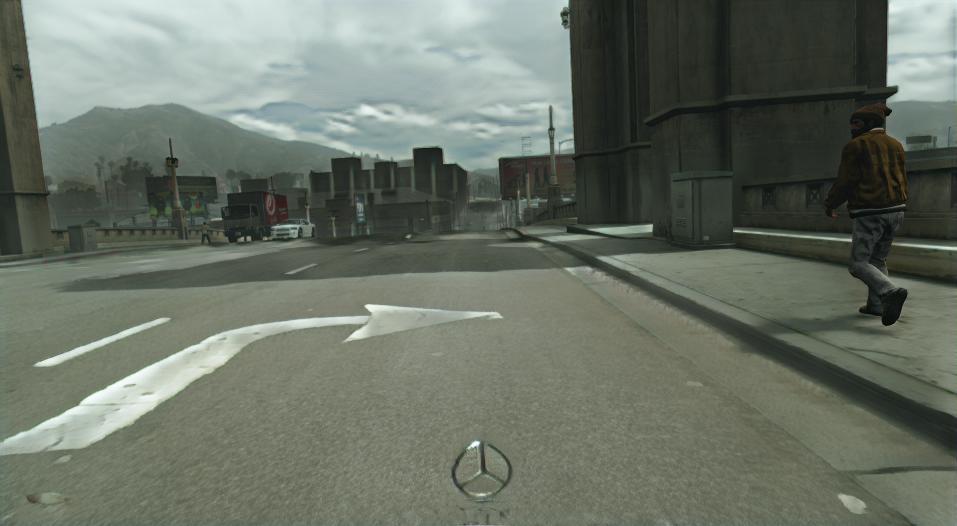}}\hfill
		\subfloat[464$\times$464]
		{\includegraphics[width=0.198\textwidth]{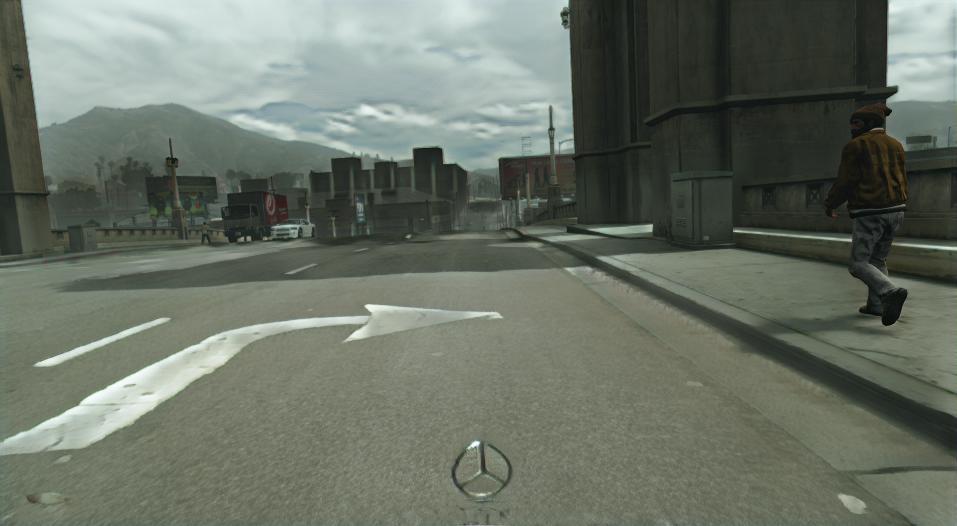}}\hfill
		\subfloat[512$\times$512]
		{\includegraphics[width=0.198\textwidth]{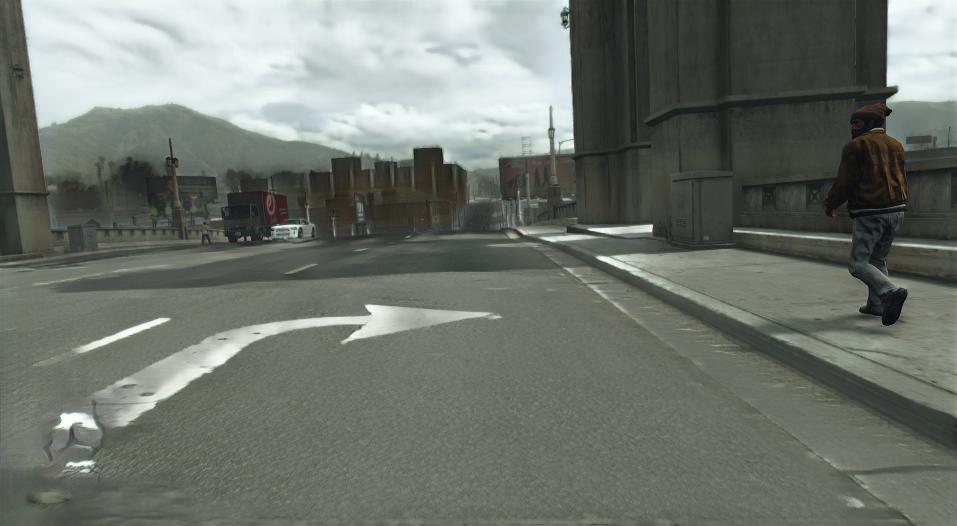}}\hfill 
	\end{center}
	\vspace{-1ex}
	\caption[Qualitative ablation of crop sizes.]{Qualitative ablation of crop sizes. For each crop size, results are randomly sampled from the best model. Best viewed in color.}
	\label{fig:feamgan:app:qualitative_ablation_crop_size_additional_random}
\end{figure}

\begin{figure}[h] 
	\captionsetup[subfigure]{labelformat=empty}
	\begin{center}
		{\includegraphics[width=0.198\textwidth]{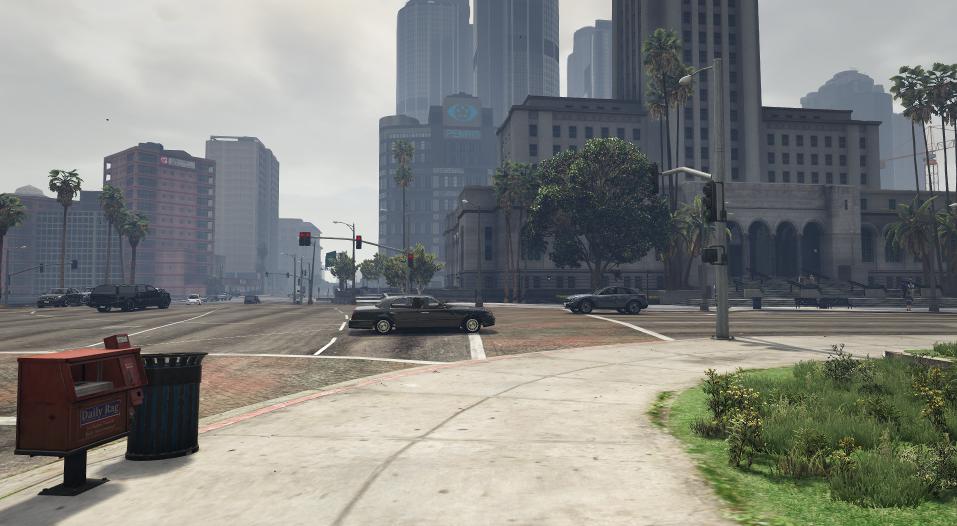}}\hfill
		{\includegraphics[width=0.198\textwidth]{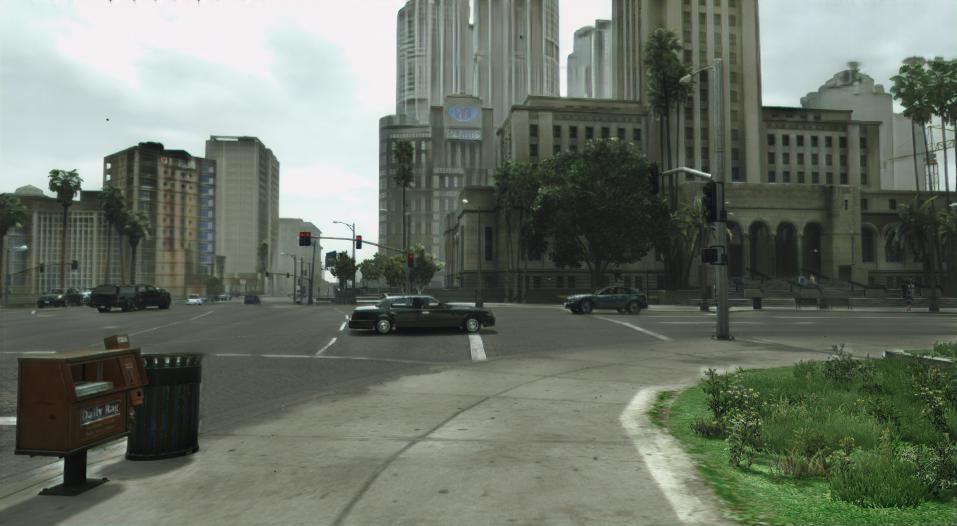}}\hfill
		{\includegraphics[width=0.198\textwidth]{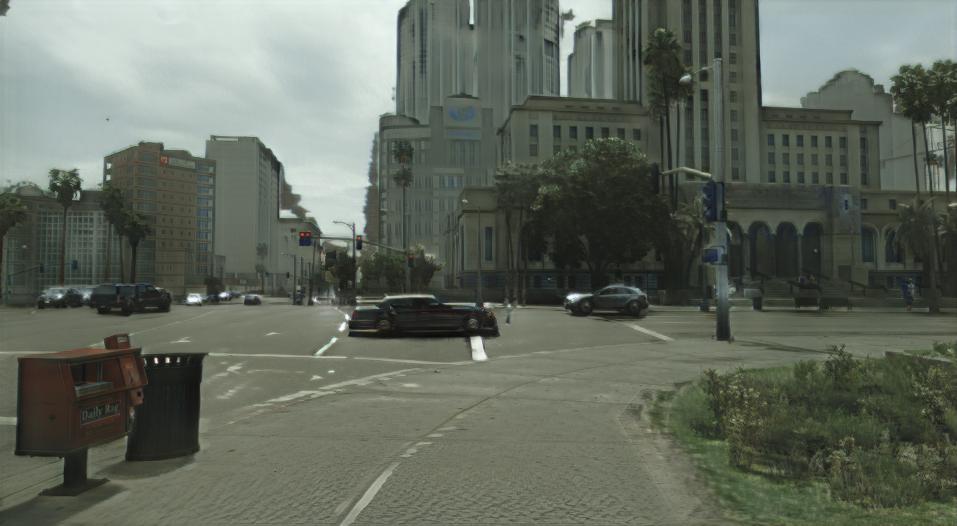}}\hfill
		{\includegraphics[width=0.198\textwidth]{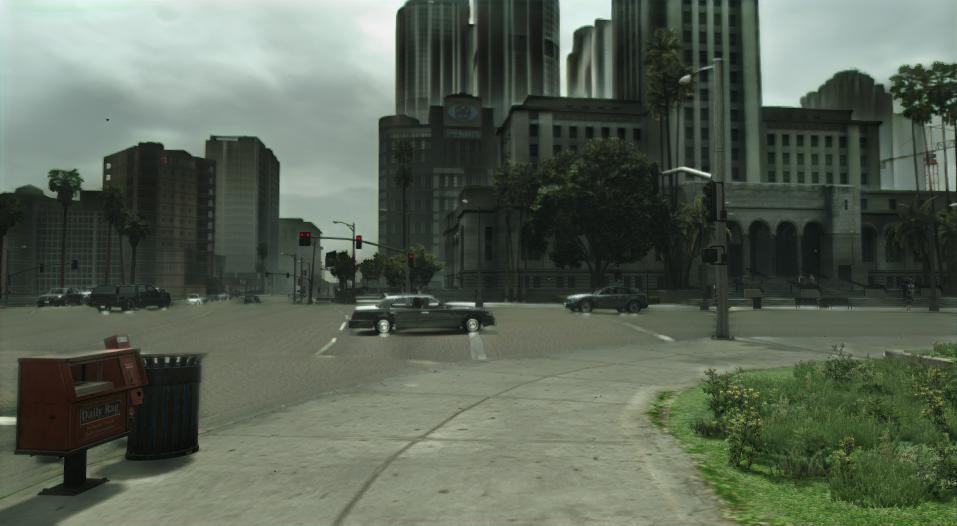}}\hfill
		{\includegraphics[width=0.198\textwidth]{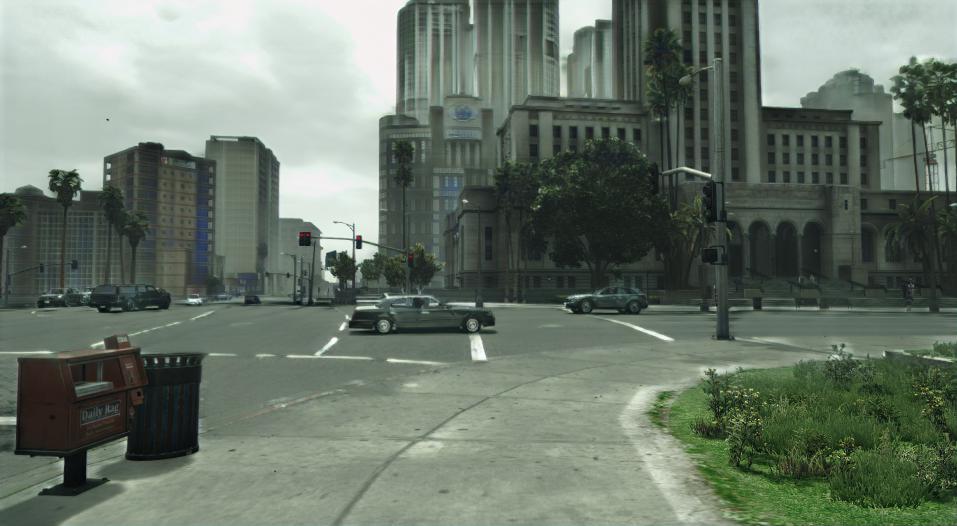}}\hfill \\ 
		{\includegraphics[width=0.198\textwidth]{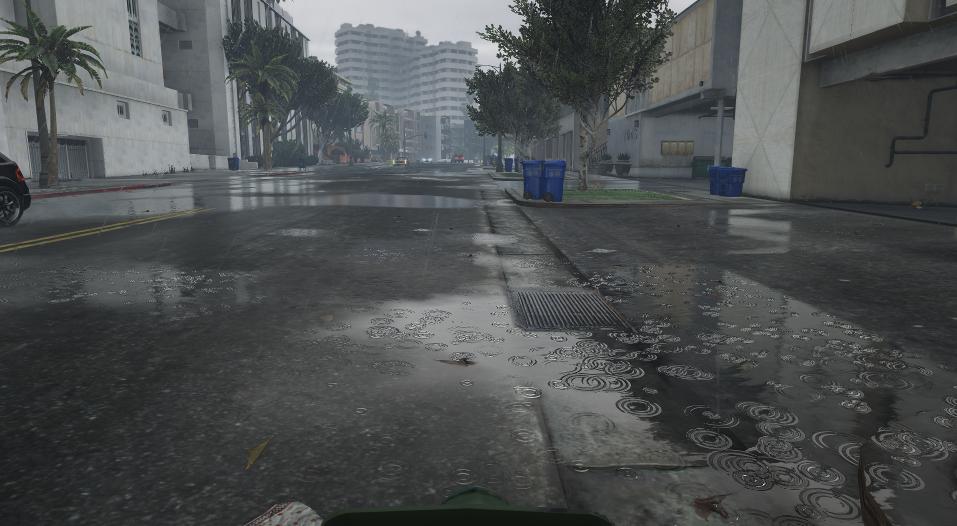}}\hfill
		{\includegraphics[width=0.198\textwidth]{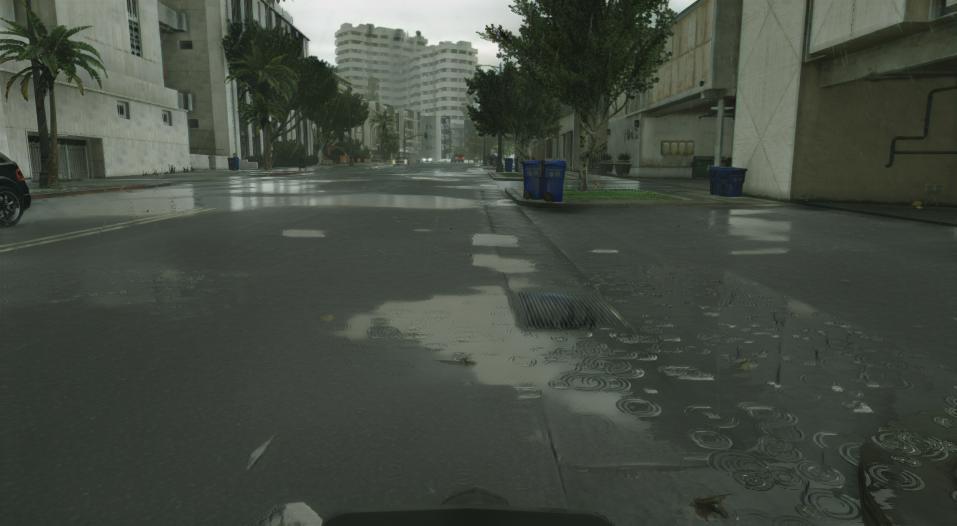}}\hfill
		{\includegraphics[width=0.198\textwidth]{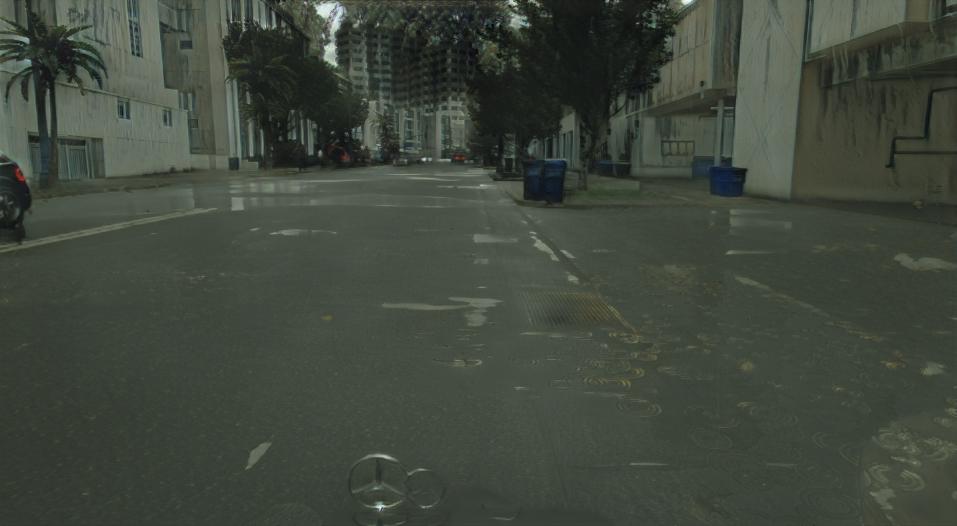}}\hfill
		{\includegraphics[width=0.198\textwidth]{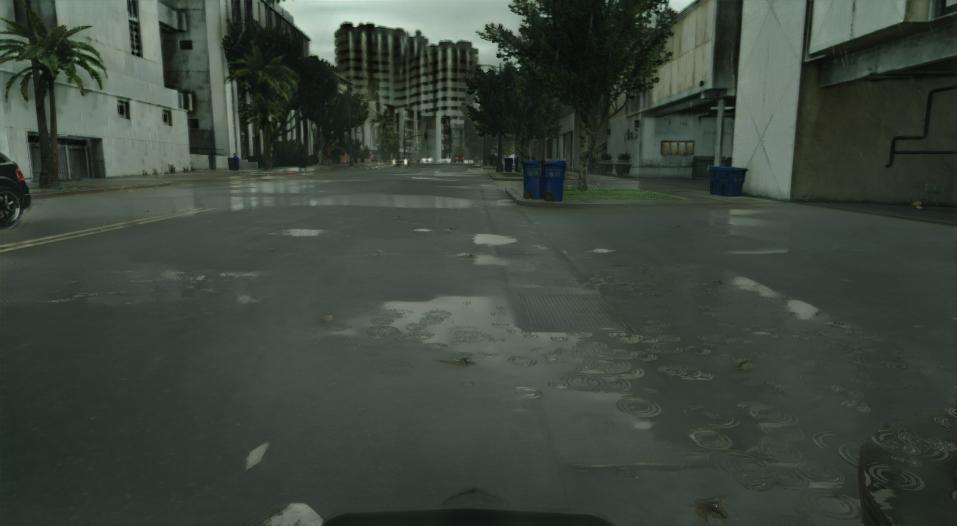}}\hfill
		{\includegraphics[width=0.198\textwidth]{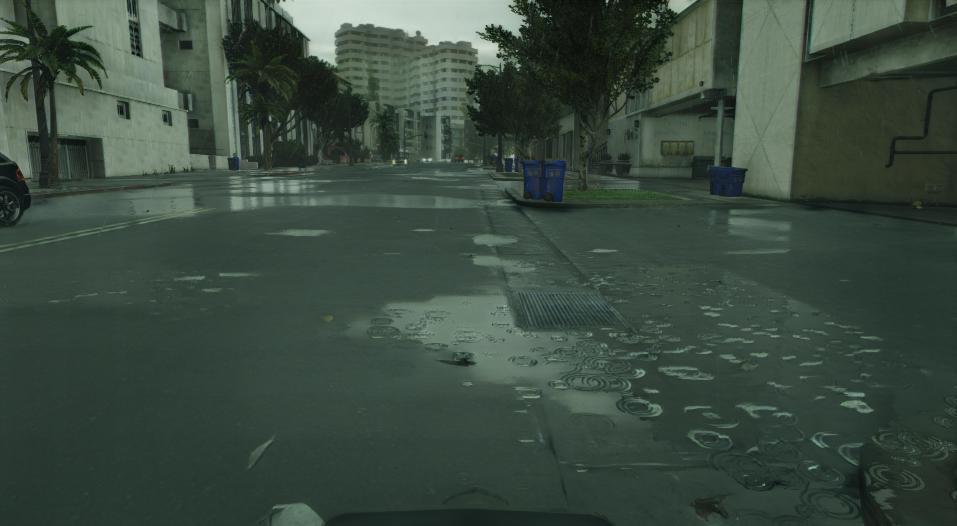}}\hfill \\ 
		{\includegraphics[width=0.198\textwidth]{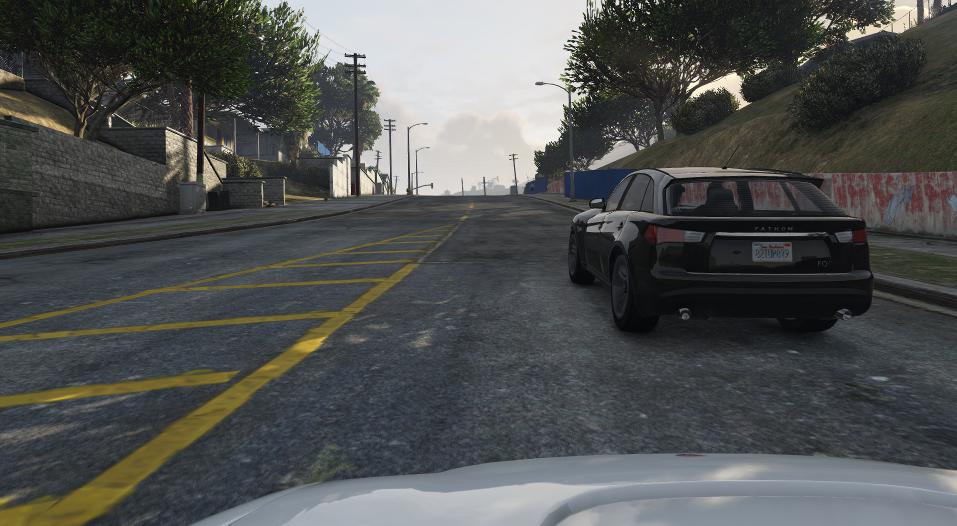}}\hfill
		{\includegraphics[width=0.198\textwidth]{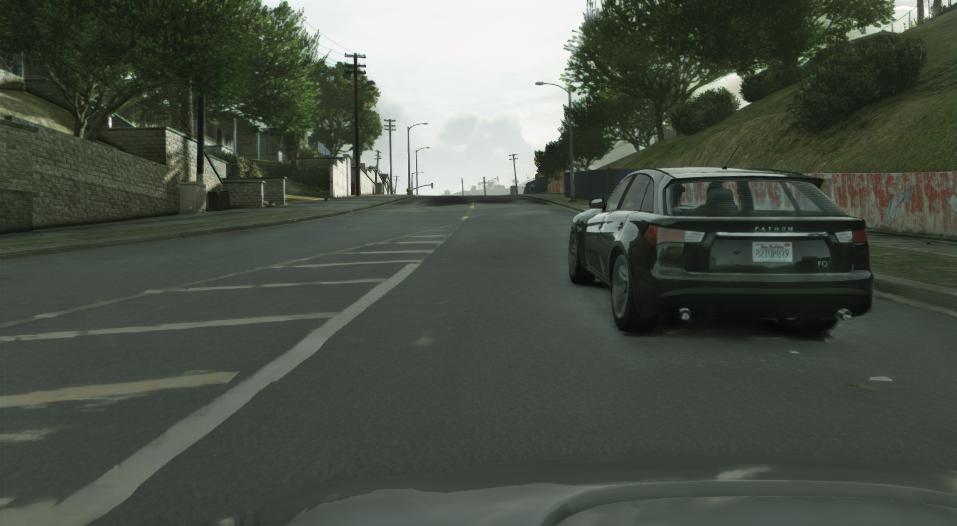}}\hfill
		{\includegraphics[width=0.198\textwidth]{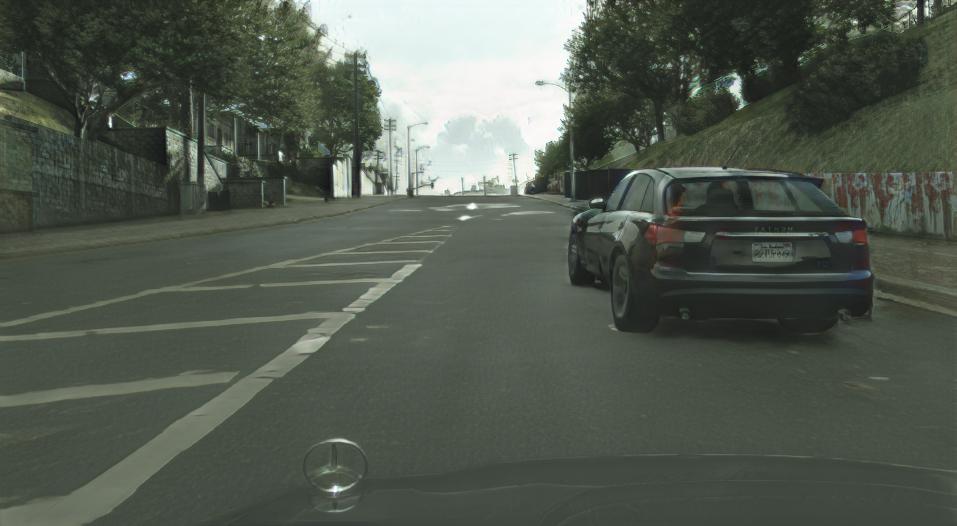}}\hfill
		{\includegraphics[width=0.198\textwidth]{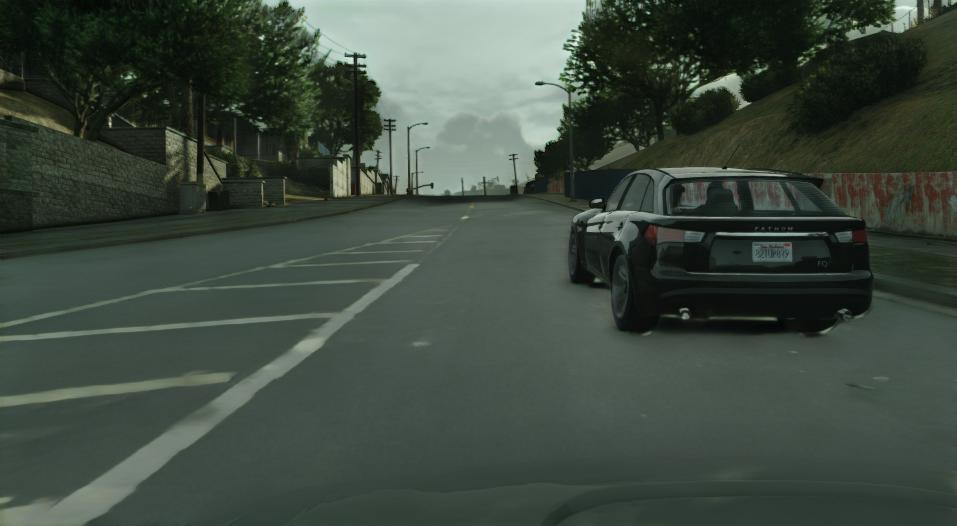}}\hfill
		{\includegraphics[width=0.198\textwidth]{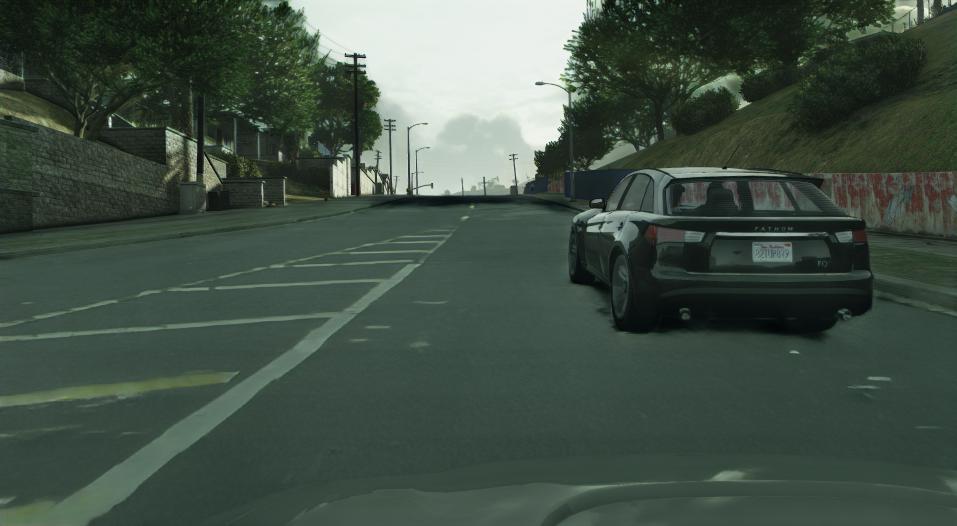}}\hfill \\ 
		{\includegraphics[width=0.198\textwidth]{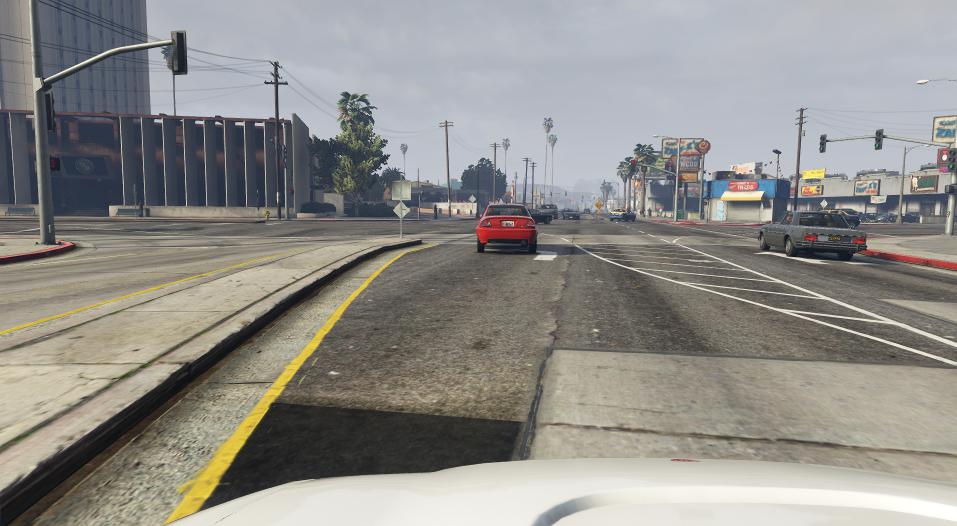}}\hfill
		{\includegraphics[width=0.198\textwidth]{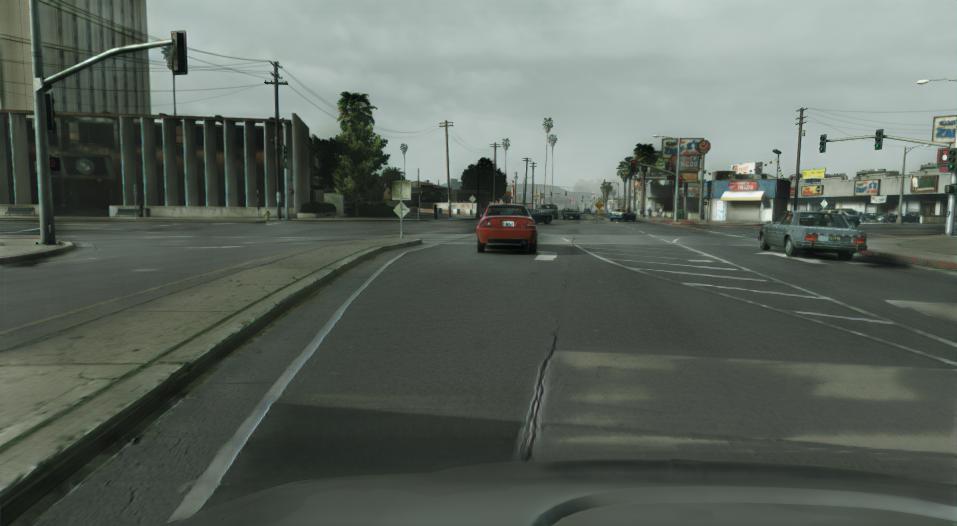}}\hfill
		{\includegraphics[width=0.198\textwidth]{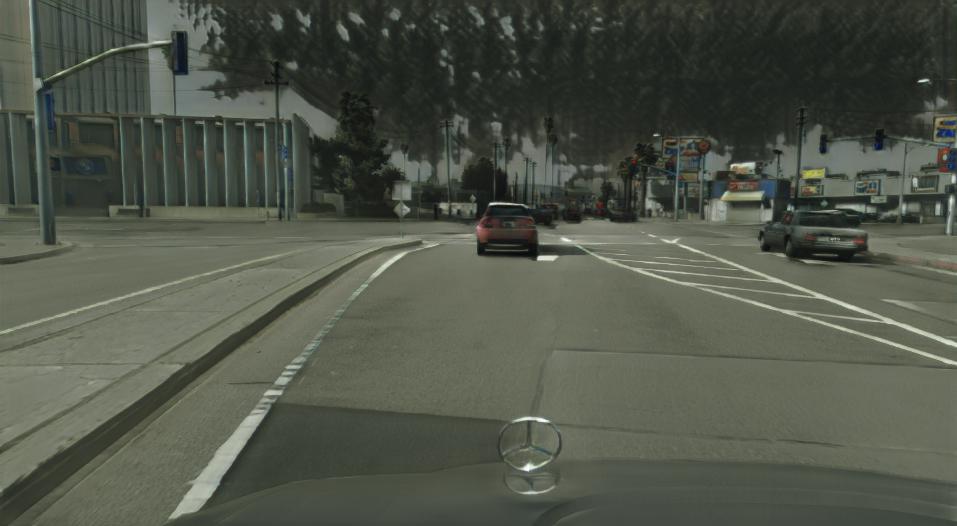}}\hfill
		{\includegraphics[width=0.198\textwidth]{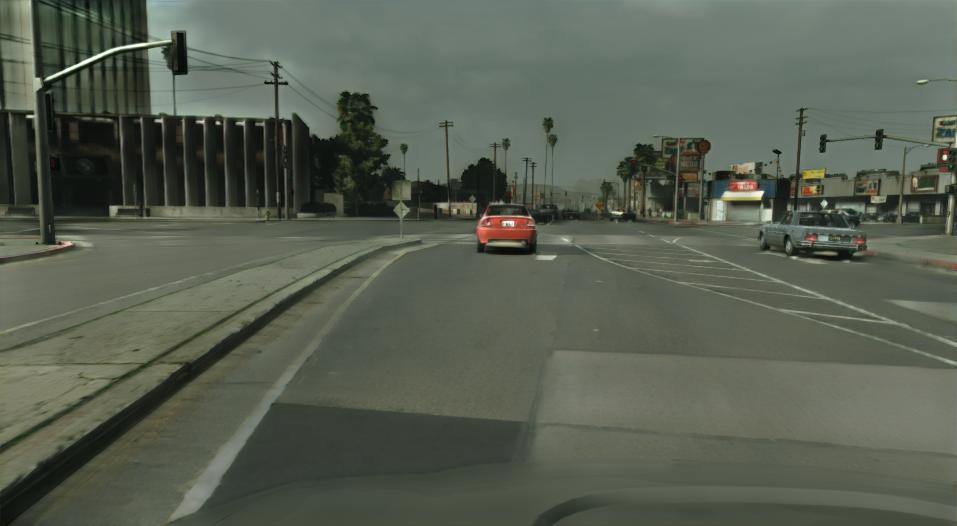}}\hfill
		{\includegraphics[width=0.198\textwidth]{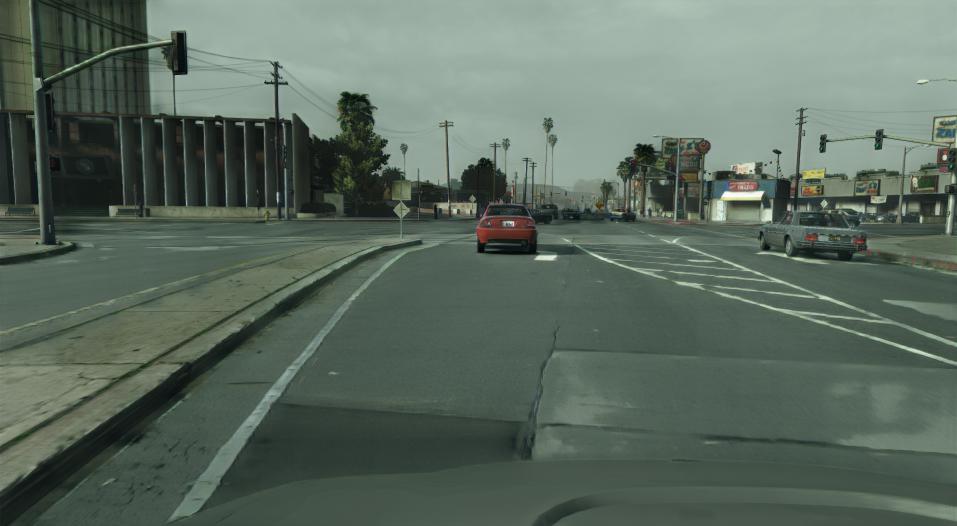}}\hfill \\ 
		{\includegraphics[width=0.198\textwidth]{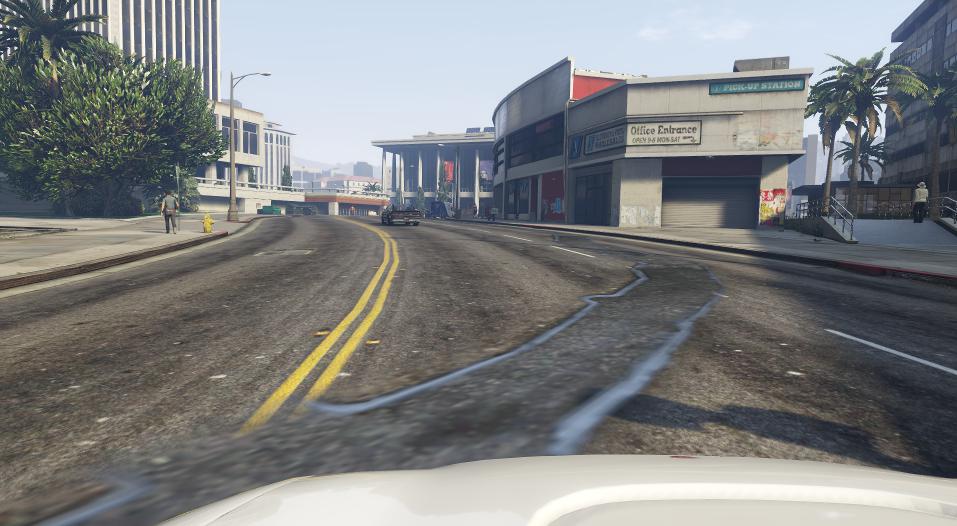}}\hfill
		{\includegraphics[width=0.198\textwidth]{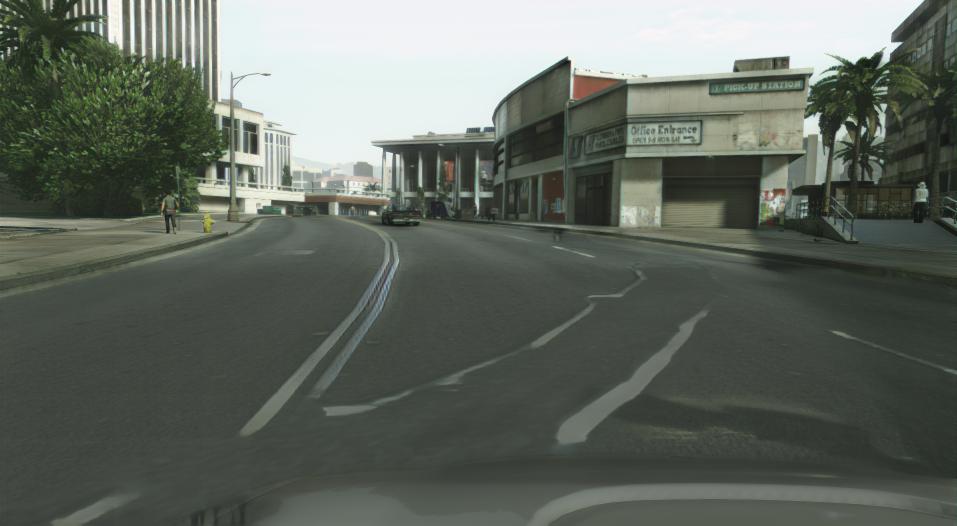}}\hfill
		{\includegraphics[width=0.198\textwidth]{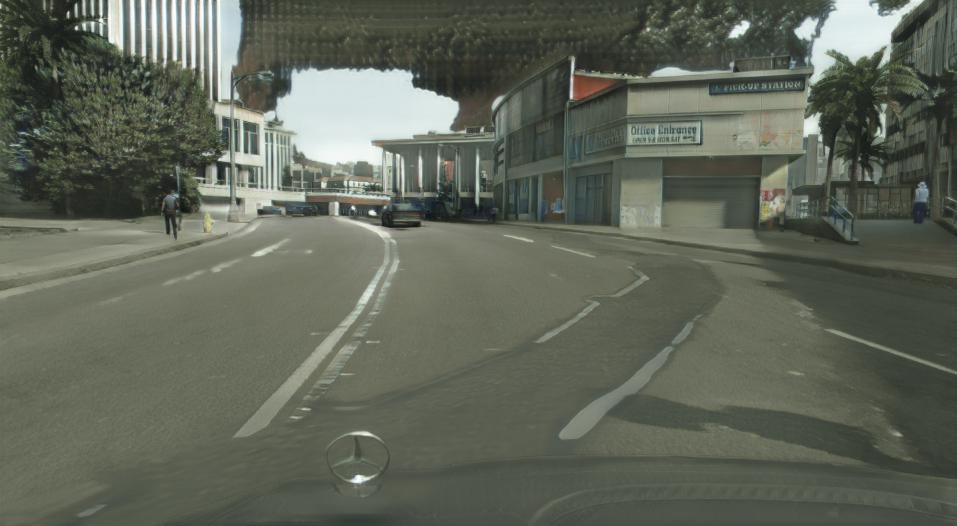}}\hfill
		{\includegraphics[width=0.198\textwidth]{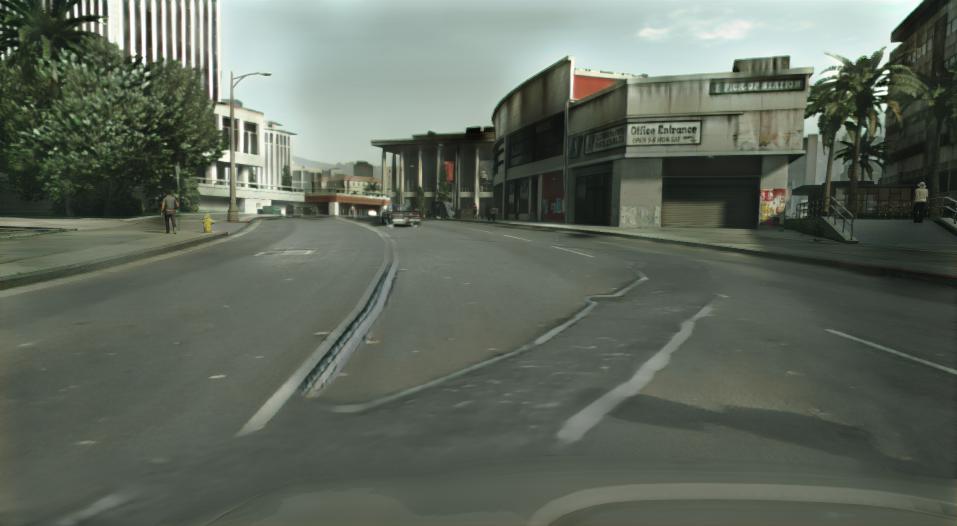}}\hfill
		{\includegraphics[width=0.198\textwidth]{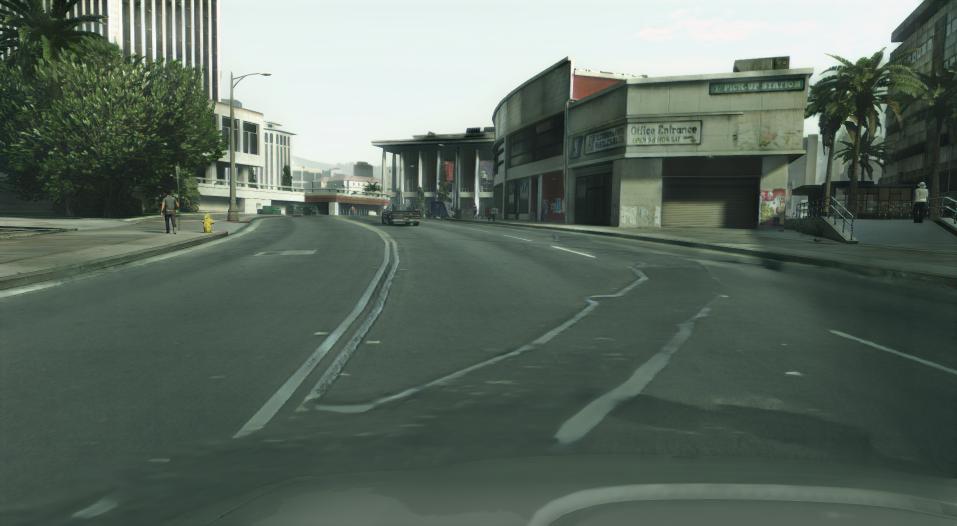}}\hfill \\ 
		{\includegraphics[width=0.198\textwidth]{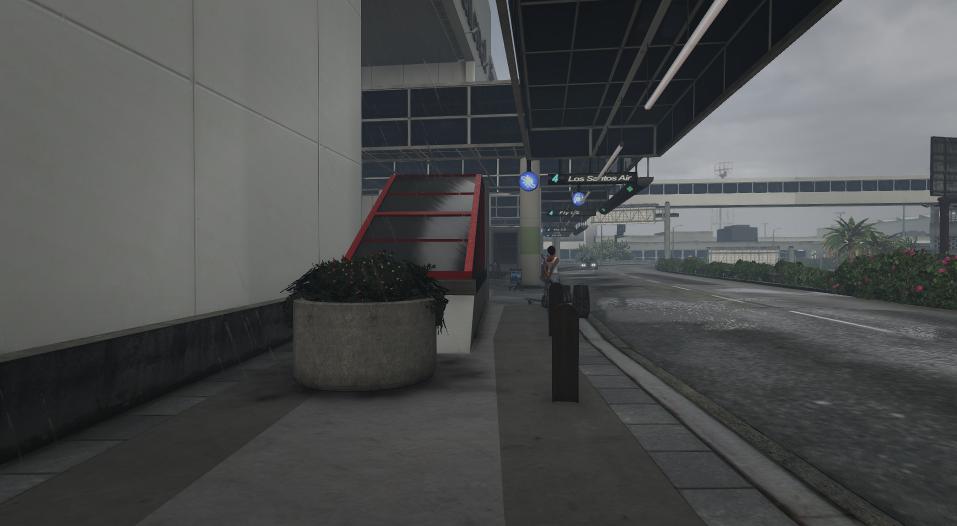}}\hfill
		{\includegraphics[width=0.198\textwidth]{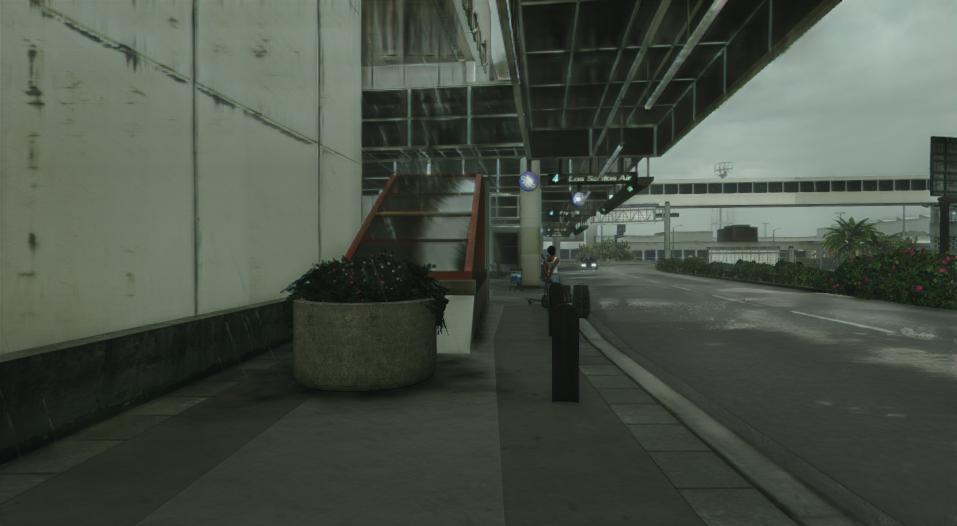}}\hfill
		{\includegraphics[width=0.198\textwidth]{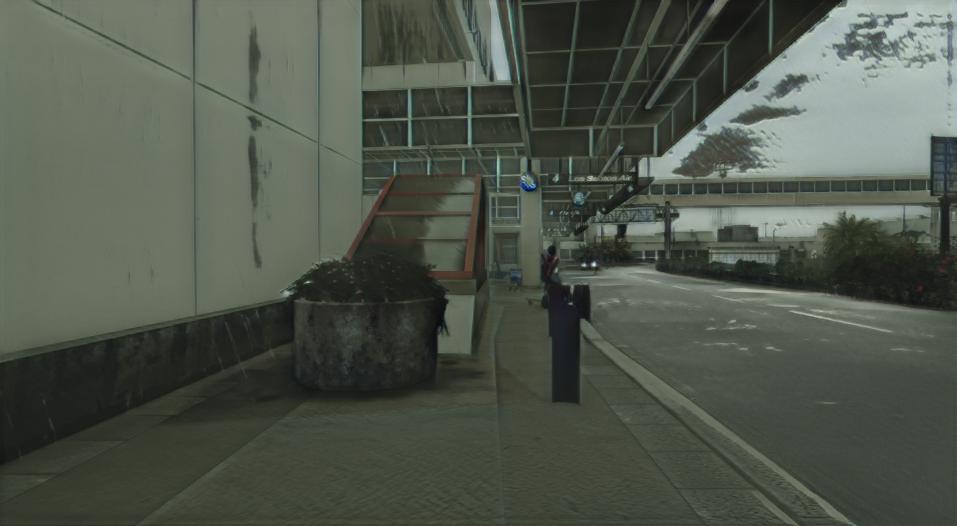}}\hfill
		{\includegraphics[width=0.198\textwidth]{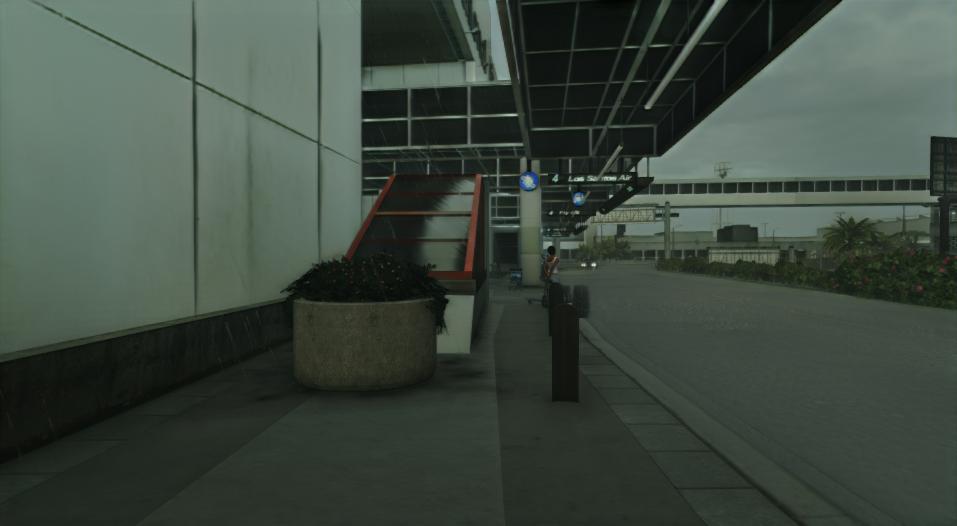}}\hfill
		{\includegraphics[width=0.198\textwidth]{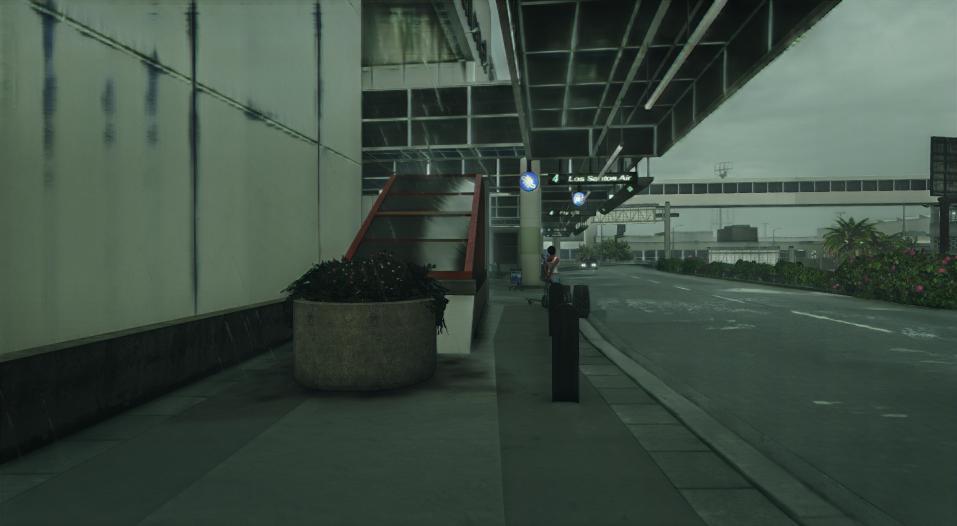}}\hfill \\ 
		{\includegraphics[width=0.198\textwidth]{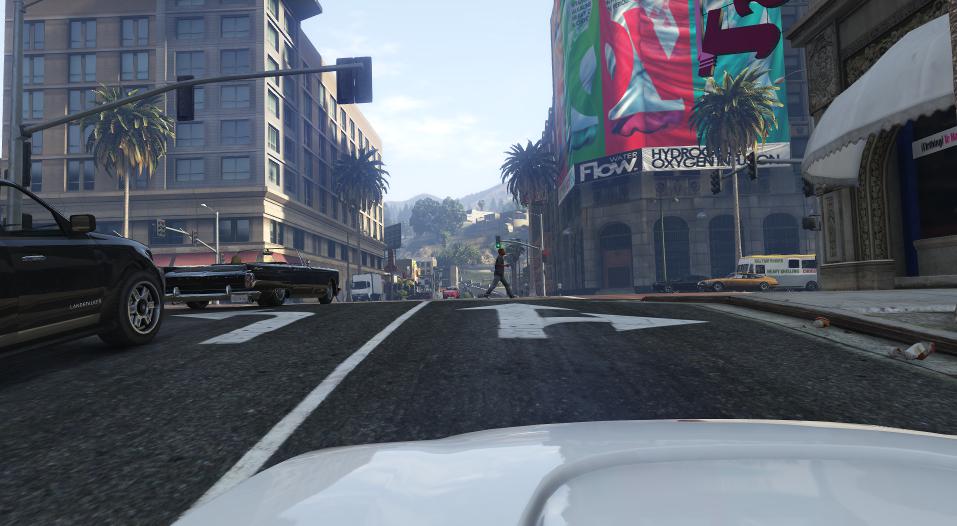}}\hfill
		{\includegraphics[width=0.198\textwidth]{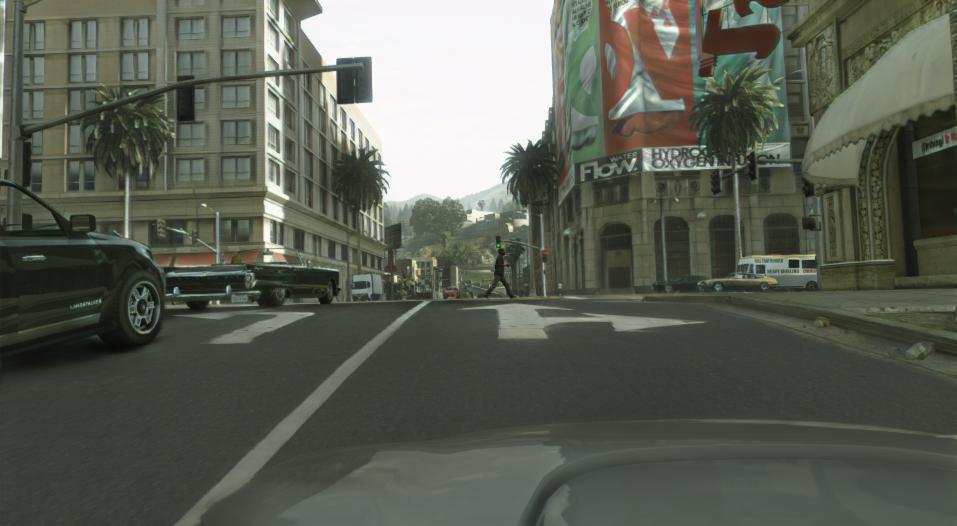}}\hfill
		{\includegraphics[width=0.198\textwidth]{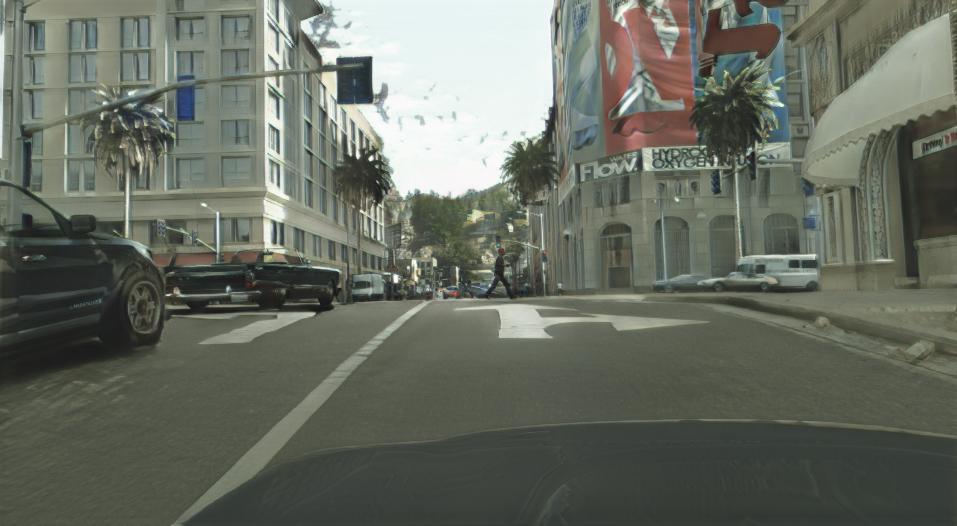}}\hfill
		{\includegraphics[width=0.198\textwidth]{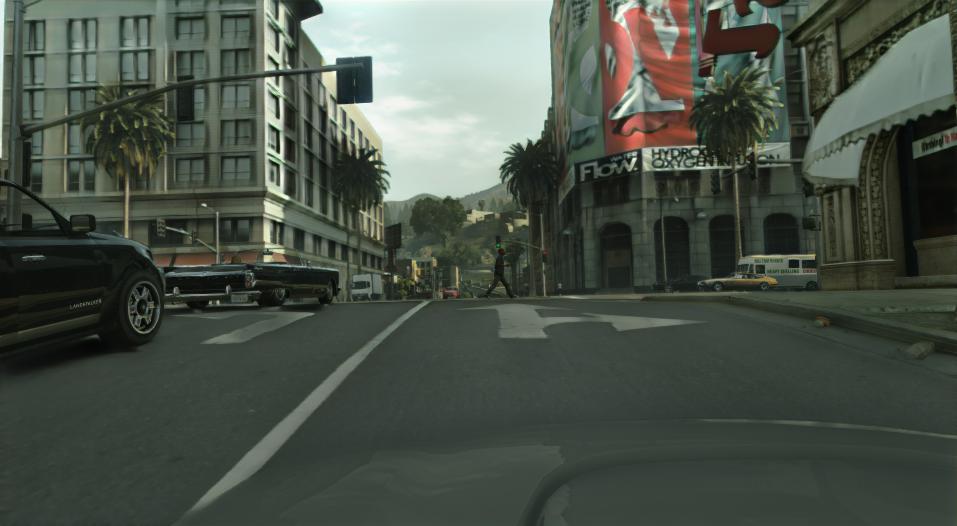}}\hfill
		{\includegraphics[width=0.198\textwidth]{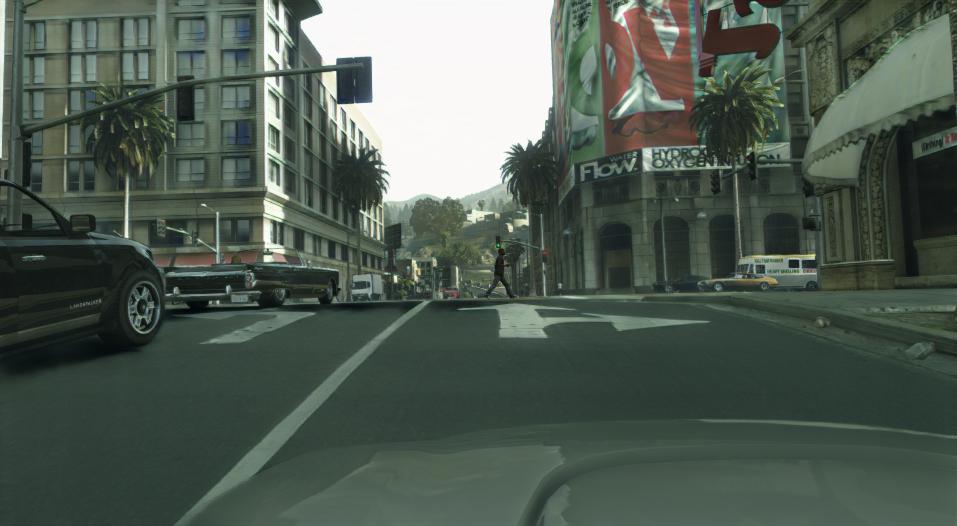}}\hfill \\ 
		{\includegraphics[width=0.198\textwidth]{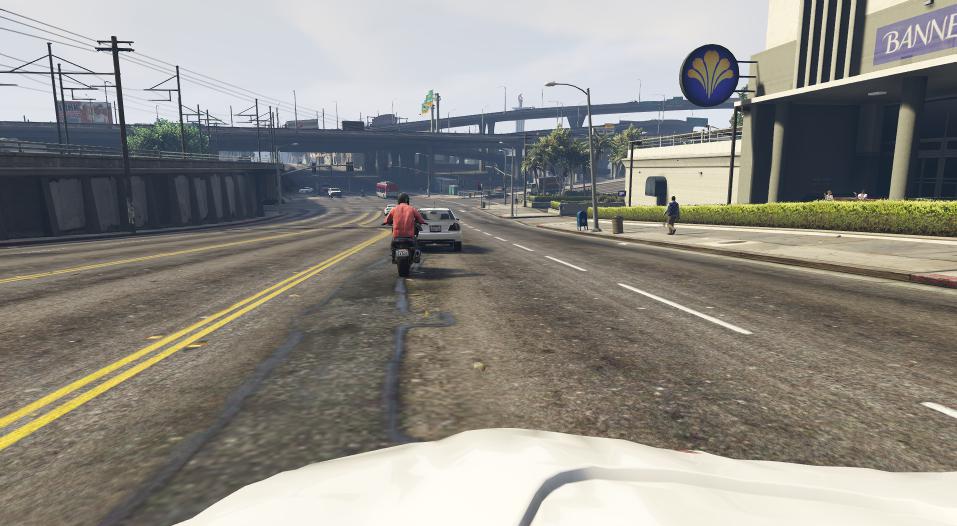}}\hfill
		{\includegraphics[width=0.198\textwidth]{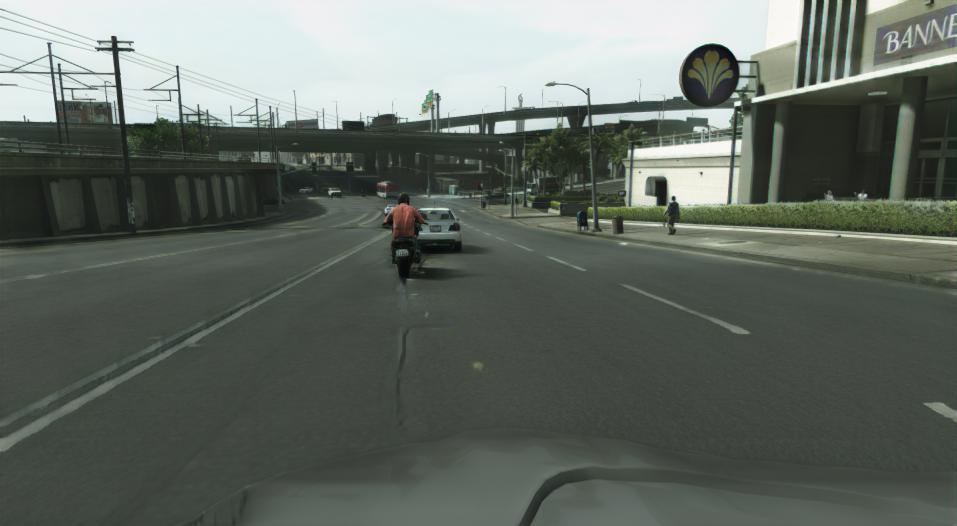}}\hfill
		{\includegraphics[width=0.198\textwidth]{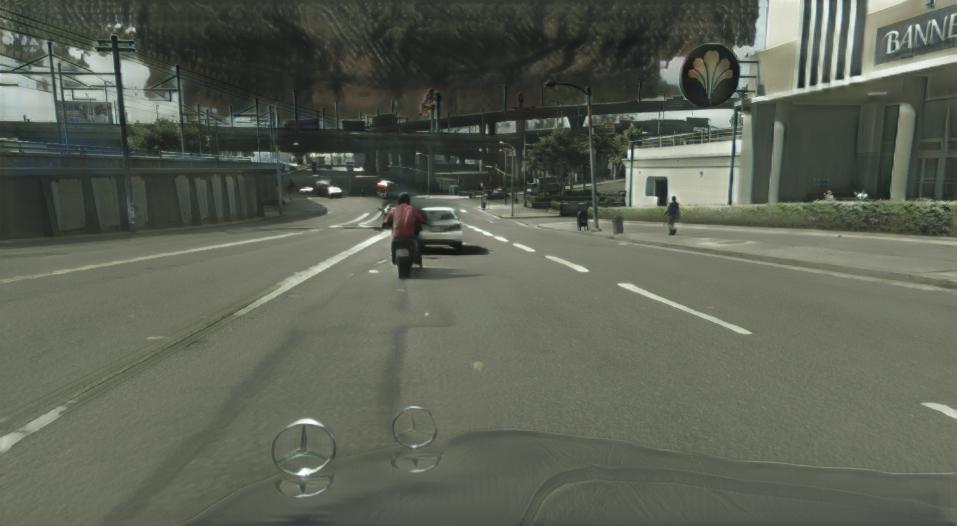}}\hfill
		{\includegraphics[width=0.198\textwidth]{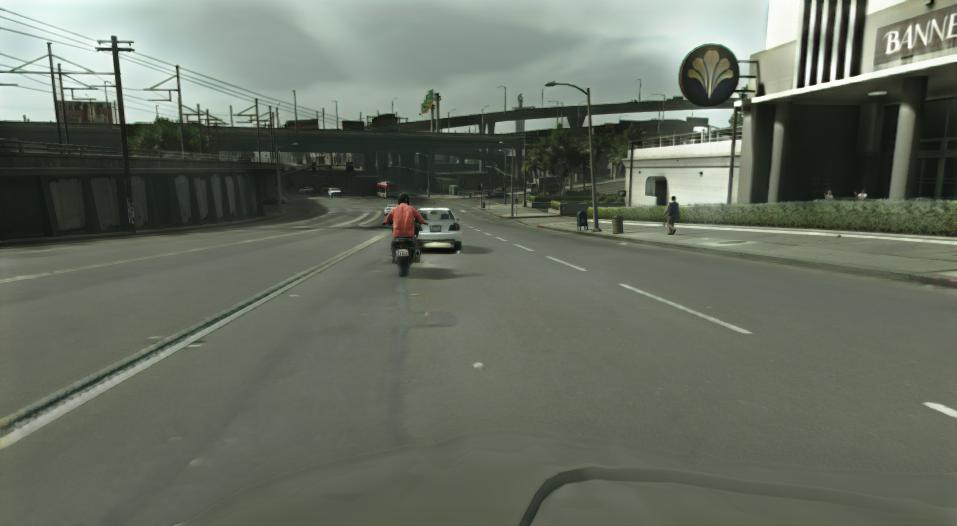}}\hfill
		{\includegraphics[width=0.198\textwidth]{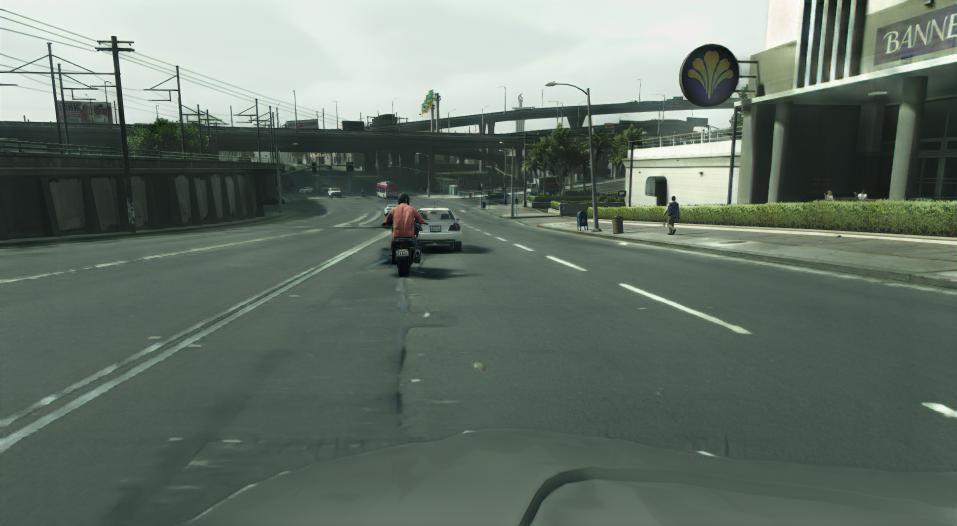}}\hfill \\ 
		{\includegraphics[width=0.198\textwidth]{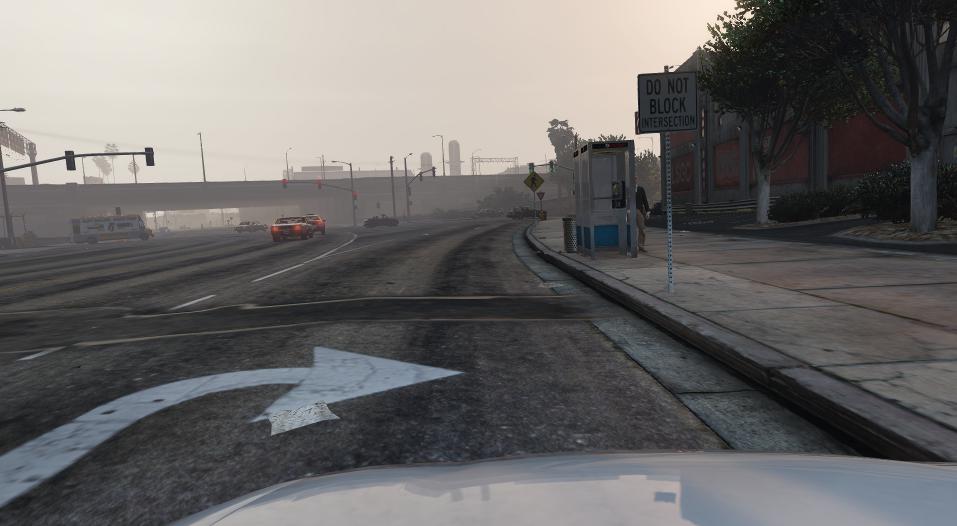}}\hfill
		{\includegraphics[width=0.198\textwidth]{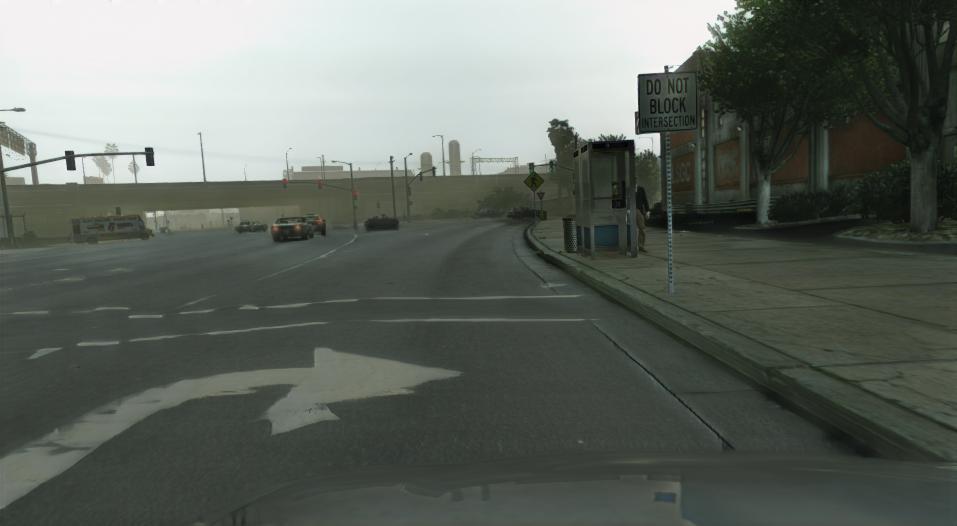}}\hfill
		{\includegraphics[width=0.198\textwidth]{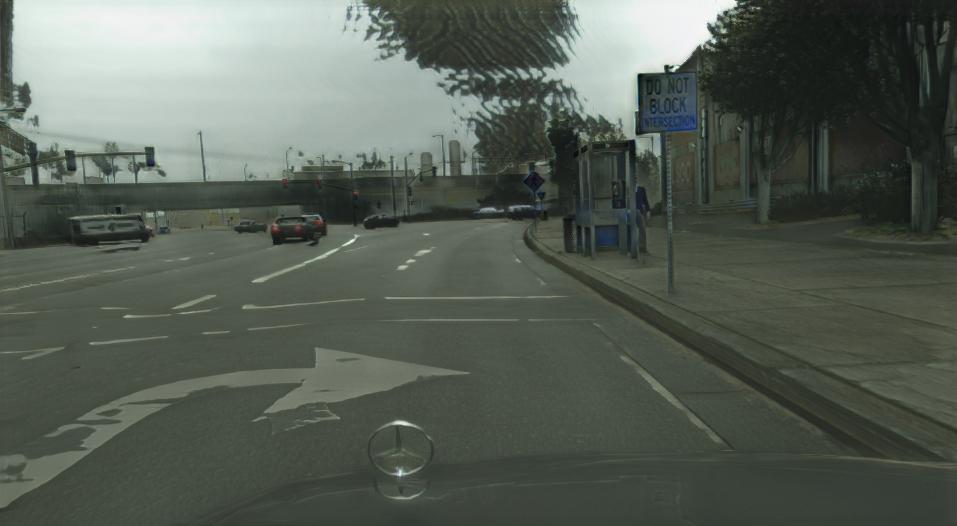}}\hfill
		{\includegraphics[width=0.198\textwidth]{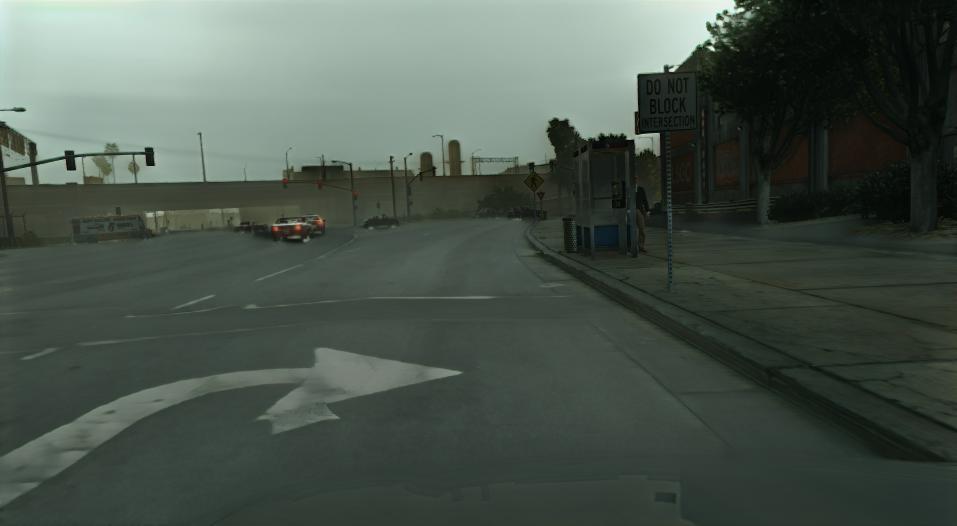}}\hfill
		{\includegraphics[width=0.198\textwidth]{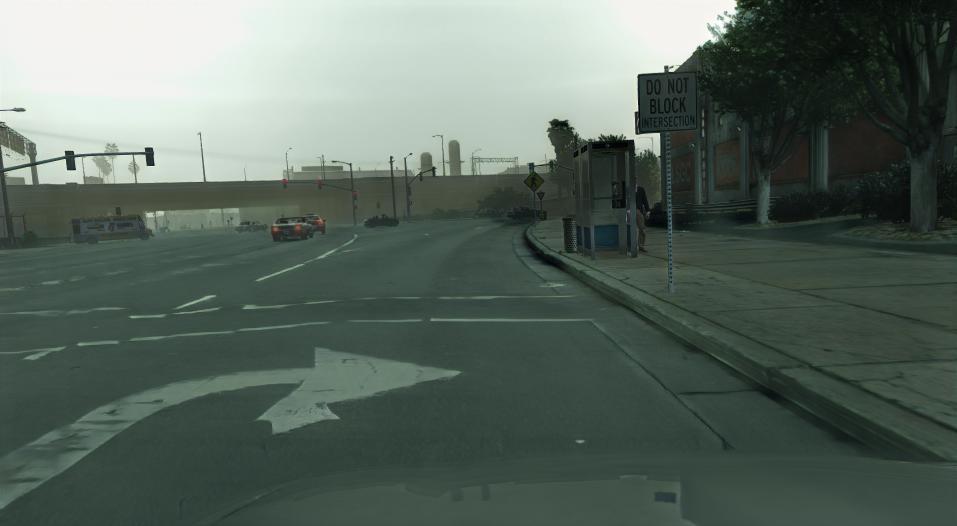}}\hfill \\ 
		{\includegraphics[width=0.198\textwidth]{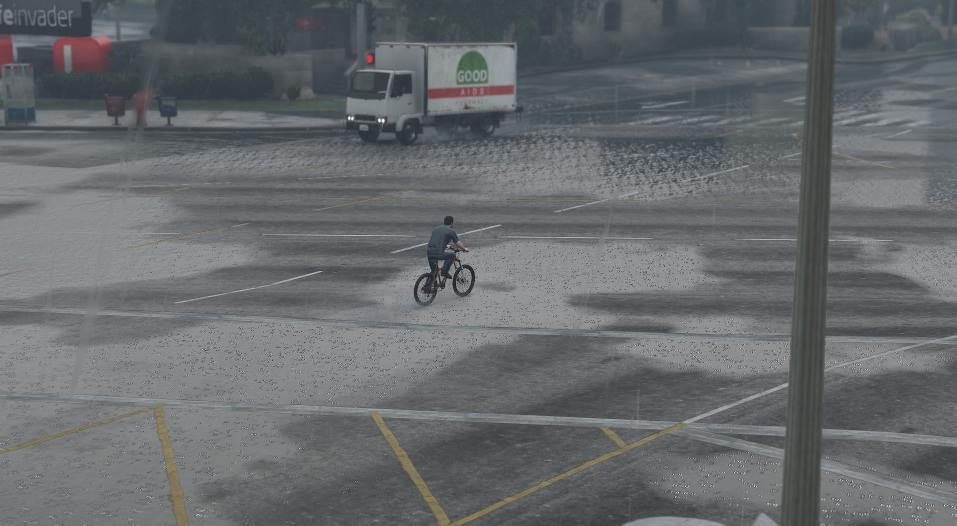}}\hfill
		{\includegraphics[width=0.198\textwidth]{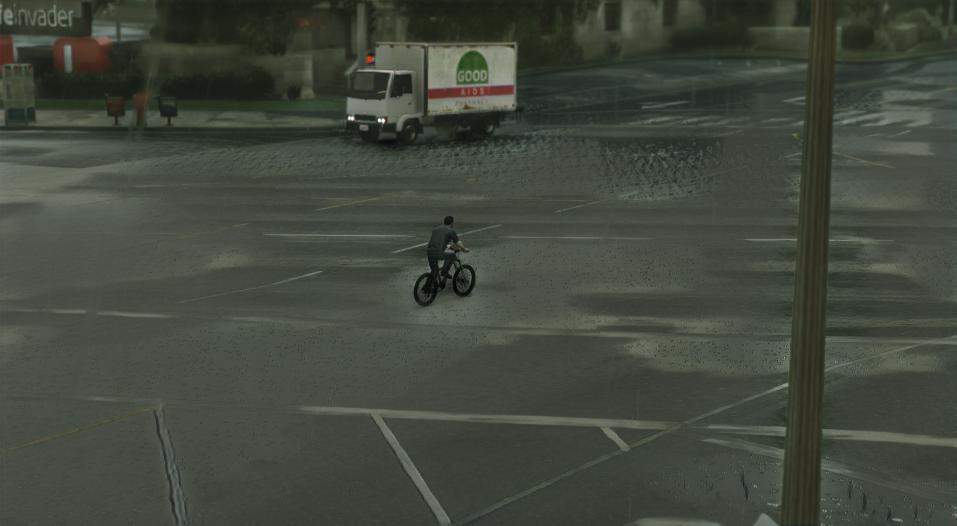}}\hfill
		{\includegraphics[width=0.198\textwidth]{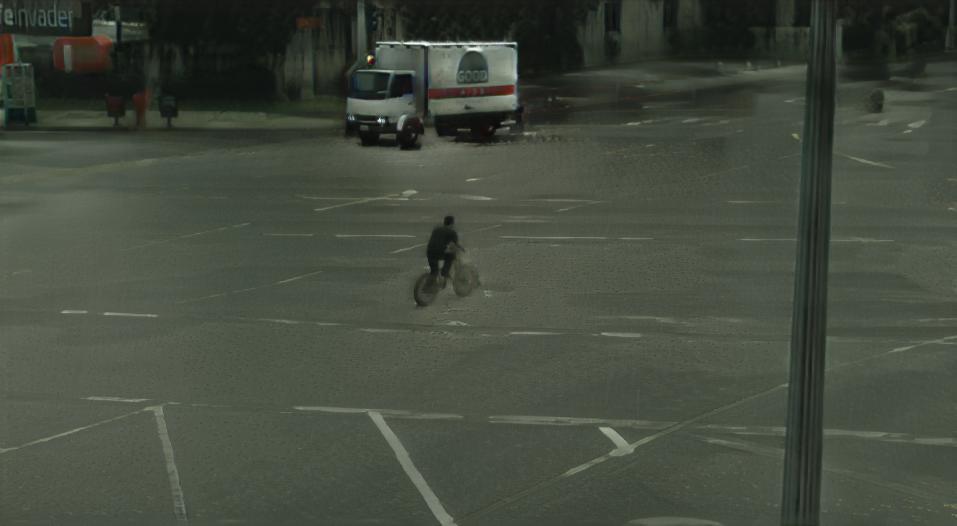}}\hfill
		{\includegraphics[width=0.198\textwidth]{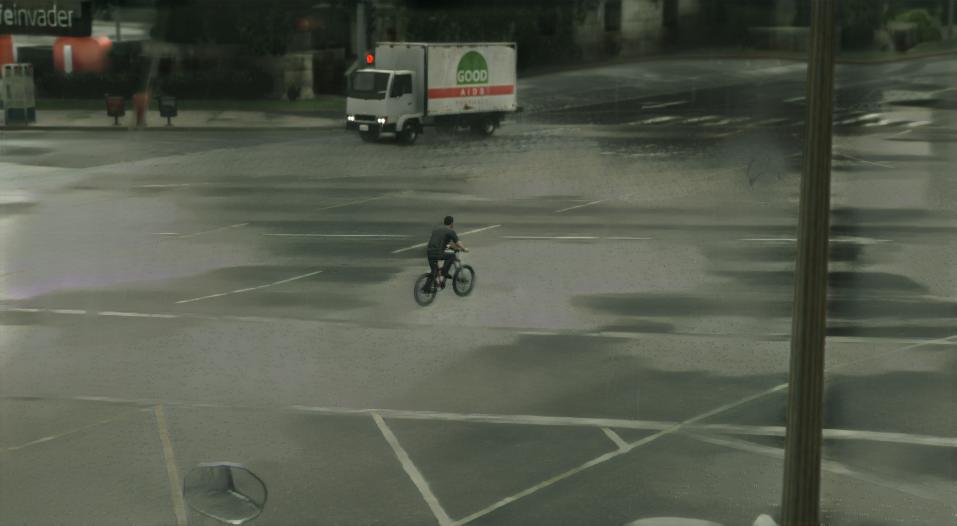}}\hfill
		{\includegraphics[width=0.198\textwidth]{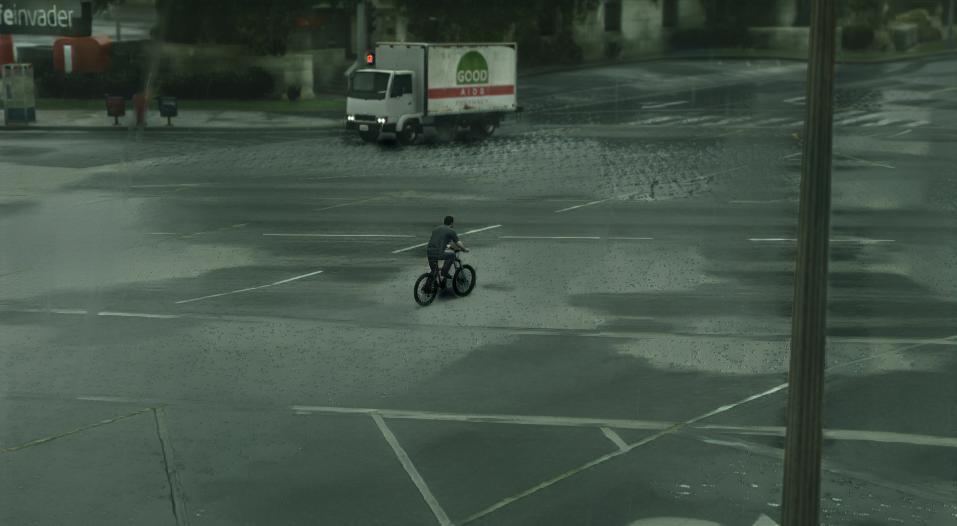}}\hfill \\ 
		\vspace{-10pt}
		\subfloat[Input]
		{\includegraphics[width=0.198\textwidth]{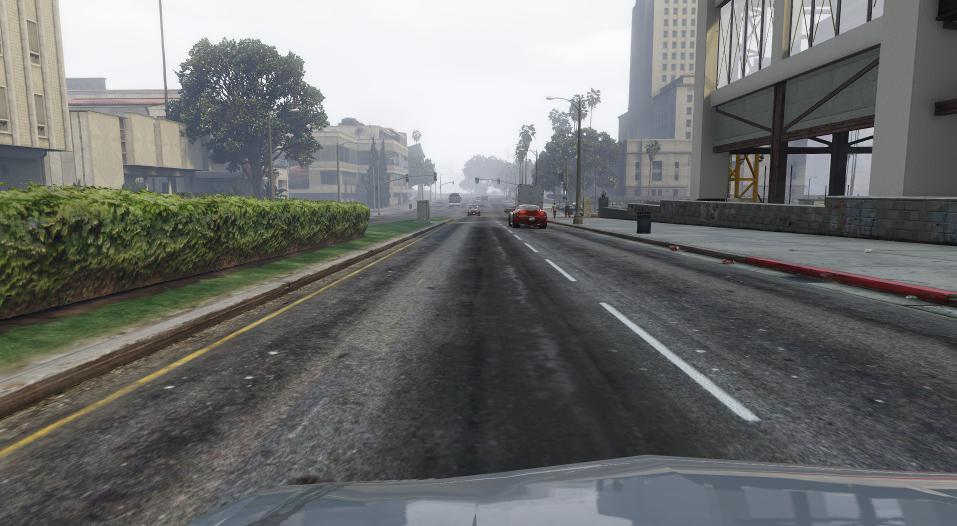}}\hfill
		\subfloat[Full]
		{\includegraphics[width=0.198\textwidth]{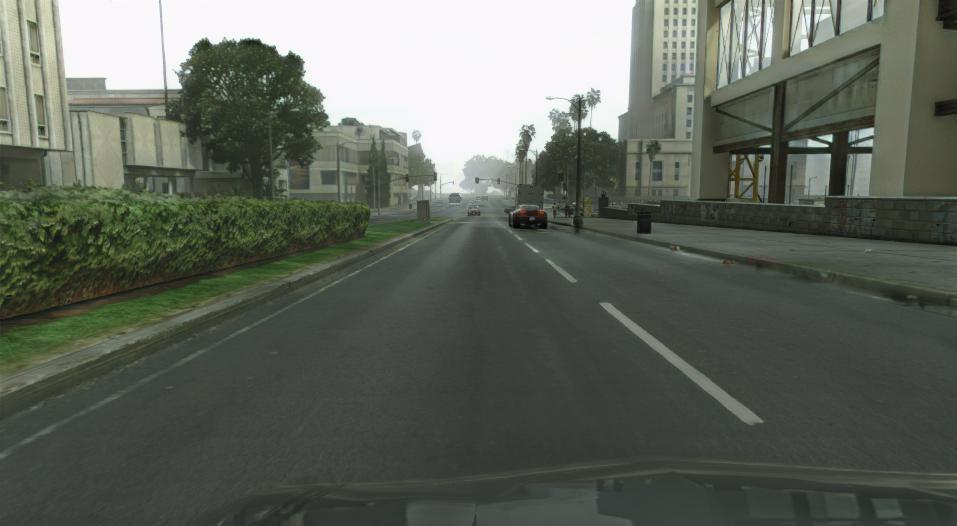}}\hfill
		\subfloat[w/o Dis. Mask]
		{\includegraphics[width=0.198\textwidth]{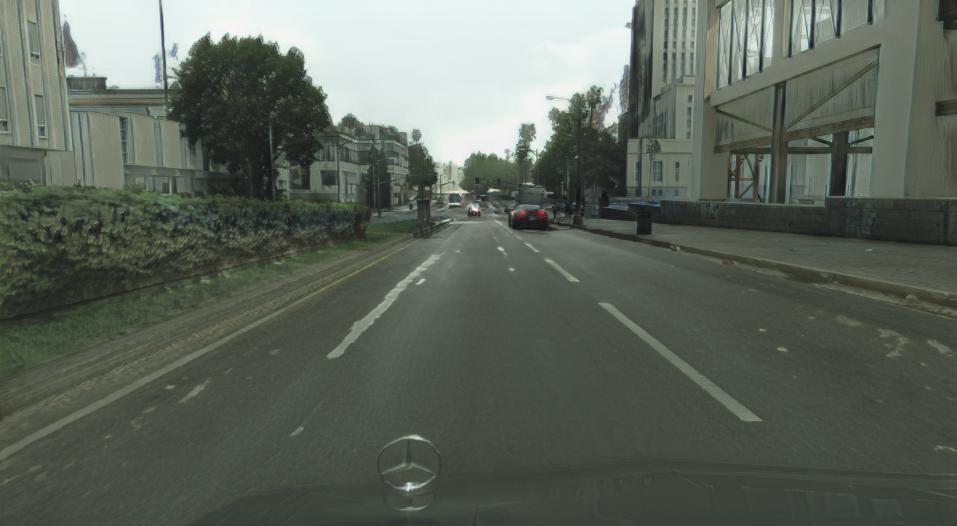}}\hfill
		\subfloat[w/o Local Dis.]
		{\includegraphics[width=0.198\textwidth]{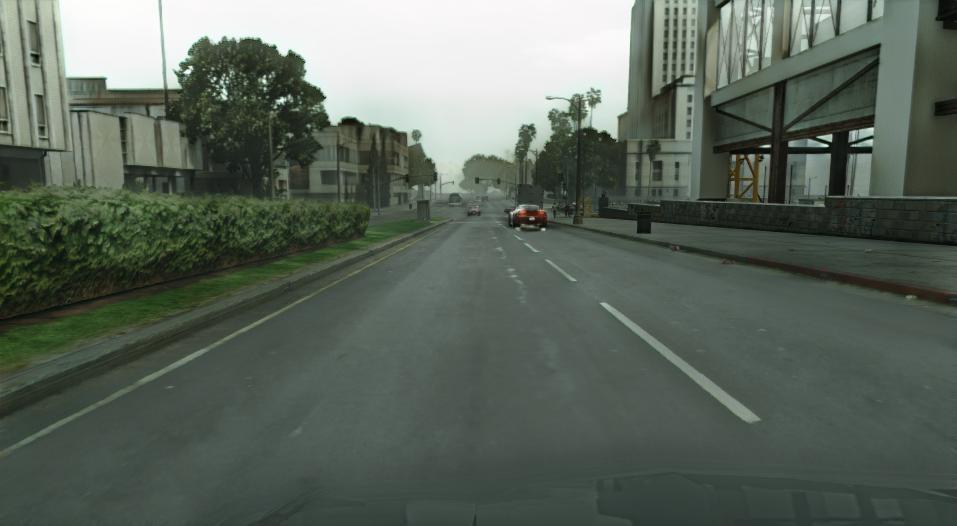}}\hfill
		\subfloat[w/ FADE w/o FATE]
		{\includegraphics[width=0.198\textwidth]{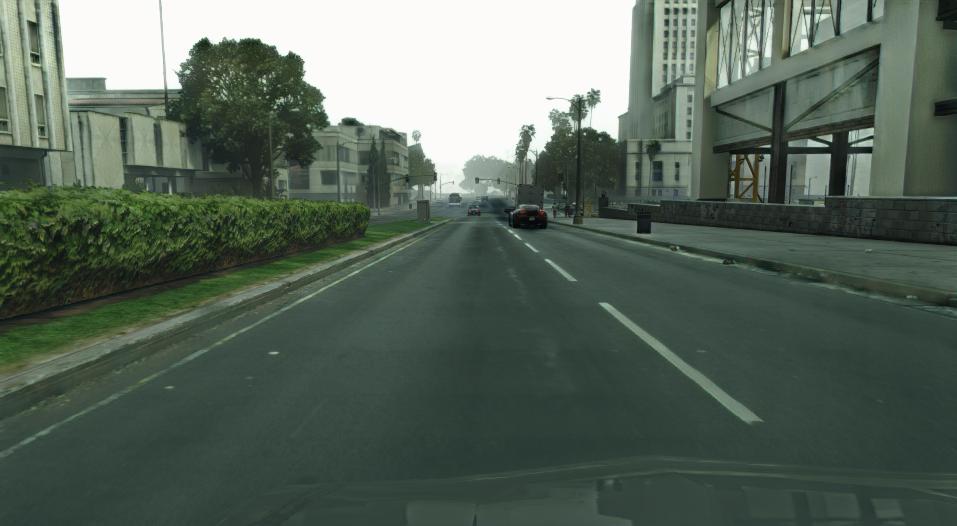}}\hfill 
	\end{center}
	\vspace{-1ex}
	\caption[Qualitative ablations.]{Qualitative ablations. Results are randomly sampled from the best model. Best viewed in color.}
	\label{fig:feamgan:app:qualitative_ablations_additional_random}
\end{figure}

We show additional results of our experiments in Figures \ref{fig:feamgan:app:qualitative_comparison_epe_additional}, \ref{fig:feamgan:app:qualitative_comparison_epe_additional_random}, \ref{fig:feamgan:app:qualitative_comparison_additional_random}, \ref{fig:feamgan:app:qualitative_ablation_crop_size_additional_random}, and \ref{fig:feamgan:app:qualitative_ablations_additional_random}. In \autoref{tab:feamgan:app:quantitative_comparison_extended}, we report additional results from our cKVD metric and the stability of all results over five runs. Furthermore, we report the stability of all results from the ablation study in \autoref{tab:feamgan:app:quantitative_ablation_extended}. We note that the results for most baselines and for our method show non-negligible deviations in many tasks.

\begin{table}[!htb]
	\RawFloats
	\caption[cKVD class mapping.]{cKVD class mapping.}
	\vspace{-1ex}
	\label{tab:feamgan:app:sKVD_class_mapping}
	\begin{center}
		\scalebox{0.6}{%
			\setlength{\tabcolsep}{0.4em}
			\begin{tabular}{ll}
				\toprule
				cKVD Class & MSeg-Id(Name)\\   
				\midrule	
				sky & 142(sky) \\ [+3pt] 
				\multirow{2}*{ground}  & 94(gravel), 95(platform), 97(railroad), \\
				& 100(pavement-merged), 101(ground) \\ [+3pt] 
				road & 98(road) \\ [+3pt] 
				terrain & 102(terrain) \\ [+3pt] 
				vegetation & 174(vegetation) \\ [+3pt] 
				\multirow{2}*{building} & 31(tunnel), 32(bridge), 33(building-parent), \\
				&  35(building), 36(ceiling-merged) \\ 			[+3pt] 			
				\multirow{4}*{roadside-obj.}  & 130(streetlight), 131(road\_barrier), 132(mailbox), \\
				& 133(cctv\_camera), 134(junction\_box), 135(traffic\_sign), \\
				& 136(traffic\_light), 137(fire\_hydrant), 138(parking\_meter), \\
				& 139(bench), 140(bike\_rack), 141(billboard) \\ 	[+3pt] 
				\multirow{2}*{person} & 125(person), 126(rider\_other), 127(bicyclist), \\
				& 128(motorcyclist) \\[+3pt] 
				\multirow{3}*{vehicle}  &  175(bicycle), 176(car), 177(autorickshaw), \\
				& 178(motorcycle), 180(bus), 181(train), \\
				& 182(truck), 183(trailer), 185(slow\_wheeled\_object) \\ [+3pt] 
				rest & all other MSeg classes \\ 
				\bottomrule	
		\end{tabular}}
	\end{center}
	\vspace{-2ex}
\end{table}
\begin{table}[!htb]
	\RawFloats
	\caption[Additional details of the used datasets.]{Additional details of the used datasets.} 
	\vspace{-1ex}
	\label{tab:feamgan:app:datasets_details}
	\begin{center}
		\scalebox{0.6}{%
			\setlength{\tabcolsep}{0.4em}
			\begin{tabular}{lcclcccc}
				\toprule
				Dataset & Resolution & fps & Used Train/Val Data & Task & Input Resolution & Input Cropping  \\   
				\midrule	
				PFD \cite{richter2016playing} & 1914$\times$1052& - & all images & \textit{PFD$\rightarrow$Cityscapes} & 957$\times$526 &  -  \\  [+3pt] 
				Viper \cite{richter2017playing} & 1920$\times$1080 & $\sim$15 & all train/val data, but no night sequences& \textit{Viper$\rightarrow$Cityscapes} & 935$\times$526 & - \\ 	[+3pt] 
				\multirow{2}*{Cityscapes \cite{cordts2016cityscapes}} & \multirow{2}*{2048$\times$1024} & \multirow{2}*{17} & \multirow{2}*{all sequences of the train/val data} & \textit{PFD$\rightarrow$Cityscapes}  & 1.052$\times$526 &  957$\times$526  \\  
				&  &  &  &  \textit{Viper$\rightarrow$Cityscapes}  & 1.052$\times$526 & 935$\times$526 \\  [+3pt] 
				\multirow{2}*{BDD100K \cite{yu2020bdd100k}} & \multirow{2}*{1280$\times$720} & \multirow{2}*{30} & train: first 100k, val: first 40k & \textit{Day$\rightarrow$Night}& \multirow{2}*{935$\times$526} & \multirow{2}*{-} \\ 
				& &  & train: first 50k, val: first 40k   & \textit{Clear$\rightarrow$Snowy} &  &  \\ 
				\bottomrule	
		\end{tabular}}
	\end{center}
	\vspace{-2ex}
\end{table}
\begin{table}[!htb]
	\RawFloats
	\caption[Additional training details.]{Additional training details.} 	
	\vspace{-3ex}
	\label{tab:feamgan:app:training_details}
	\begin{center}
		\scalebox{0.6}{%
			\setlength{\tabcolsep}{0.4em}
			\begin{tabular}{lcccc}
				Task & Epochs & Schedule & Decay & Local Discriminator Batch Size  \\   
				\midrule	
				\textit{PFD$\rightarrow$Cityscapes} & $20$ & half learning rate stepwise, learning rate $\geq 0.0000125$ & after each $3$rd epoch  & 32\\  
				\textit{Viper$\rightarrow$Cityscapes} & $5$ & half learning rate stepwise, learning rate $\geq 0.0000125$ & after each epoch  & 32\\    
				\textit{Day$\rightarrow$Night} & $5$ & half learning rate stepwise, learning rate $\geq 0.0000125$ & after each epoch  & 32\\    
				\textit{Clear$\rightarrow$Snowy} & $10$ & half learning rate stepwise, learning rate $\geq 0.0000125$ & after each epoch & 32\\    
		\end{tabular}}
	\end{center}
	\vspace{-2ex}
\end{table}

\begin{table}[t]
	\RawFloats
	\caption[Extended quantitative comparison to prior work.]{Extended quantitative comparison to prior work. Models were trained using their official implementations. Results are reported as the average across five runs.}
	\vspace{-4ex}
	\label{tab:feamgan:app:quantitative_comparison_extended}
	\begin{center}
		\rotatebox{270}{
		\scalebox{0.6}{%
		\setlength{\tabcolsep}{0.175em}
		\begin{tabular}{lcccccccccccccc}
			\toprule
			\multirow{2}{*}{Method}&\multirow{2}{*}{FID}&\multirow{2}{*}{KID} &\multirow{2}{*}{sKVD}&\multicolumn{11}{c}{cKVD}\\
			\cmidrule(lr){5-15}
			&    &  &  & AVG& 		sky& 	ground&	road&	terrain&	vegetation&	building&	roadside-obj.&	person&	vehicle&	rest	\\
			\midrule
			\textbf{PFD$\rightarrow$Cityscapes} &   &  &  &  &  &  &   &  &  &  &    &  &  &    \\	
			
			Color Transfer& $91.01^{+0.05}_{-0.03}$ &	$94.82^{+0.14}_{-0.11}$ & $18.16^{+0.20}_{-0.18}$ & $50.87^{+1.18}_{-1.36}$ &$58.05^{+2.28}_{-1.24}$ & $16.66^{+0.19}_{-0.26}$ & $16.38^{+0.14}_{-0.09}$ & $26.91^{+1.23}_{-1.29}$ & $28.18^{+0.44}_{-0.26}$ & $32.60^{+0.33}_{-0.45}$ & $58.36^{+5.56}_{-7.85}$ & $125.37^{+11.20}_{-7.66}$ & $55.12^{+1.38}_{-1.86}$ & $91.11^{+0.39}_{-0.84}$ \\
			
			MUNIT & $40.36^{+1.22}_{-1.18}$ &$29.98^{+1.43}_{-1.14}$ & $14.99^{+0.06}_{-0.05}$ & $43.24^{+0.67}_{-0.71}$ &$37.92^{+0.85}_{-0.55}$ & $13.23^{+0.29}_{-0.23}$ & $14.33^{+0.16}_{-0.12}$ & $22.70^{+0.81}_{-0.75}$ & $24.97^{+0.37}_{-0.35}$ & $27.52^{+0.10}_{-0.15}$ & $58.24^{+3.37}_{-3.87}$ & $108.61^{+5.56}_{-4.30}$ & $45.35^{+0.32}_{-0.24}$ & $79.54^{+0.25}_{-0.49}$ \\
			
			CUT & $49.55^{+5.19}_{-3.63}$ &	$44.25^{+7.57}_{-4.99}$ & $16.85^{+1.28}_{-1.82}$ & $\mathbf{37.53}^{+1.09}_{-1.13}$ &$\mathbf{28.76}^{+0.98}_{-0.39}$ & $\mathbf{11.17}^{+1.26}_{-0.84}$ & $\mathbf{13.92}^{+0.58}_{-0.80}$ & $13.49^{+1.21}_{-1.54}$ & $24.20^{+1.23}_{-1.39}$ & $24.69^{+0.92}_{-0.86}$ & $57.45^{+4.33}_{-2.22}$ & $\mathbf{90.52}^{+3.33}_{-6.67}$ & $40.92^{+2.77}_{-3.08}$ & $70.20^{+3.10}_{-0.99}$ \\
			
			TSIT & $\mathbf{38.70}^{+1.59}_{-1.16}$ &	$\mathbf{28.70}^{+1.81}_{-1.27}$ & $\mathbf{10.80}^{+0.57}_{-0.29}$ & $42.35^{+0.73}_{-0.89}$ &$40.13^{+1.58}_{-1.28}$ & $13.74^{+0.39}_{-1.14}$ & $14.09^{+0.27}_{-0.78}$ & $23.48^{+0.84}_{-1.17}$ & $23.74^{+0.36}_{-0.47}$ & $25.76^{+0.16}_{-0.22}$ & $\mathbf{51.95}^{+2.31}_{-2.72}$ & $107.98^{+4.67}_{-5.61}$ & $43.49^{+1.32}_{-1.01}$ & $79.13^{+2.72}_{-2.05}$ \\
			
			QS-Attn& $49.42^{+5.71}_{-6.93}$ &	$42.87	^{+7.34}_{-10.03}$ & $14.01^{+0.34}_{-0.38}$ & $38.57^{+2.85}_{-1.68}$ &$29.50^{+2.40}_{-1.88}$ & $11.69^{+1.51}_{-0.69}$ & $\mathbf{13.92}^{+0.68}_{-0.76}$ & $\mathbf{13.22}^{+2.40}_{-1.23}$ & $23.99	^{+1.34}_{-1.72}$ & $\mathbf{23.36}^{+1.32}_{-1.56}$ & $57.88^{+2.89}_{-1.88}$ & $100.32^{+14.23}_{-6.18}$ & $\mathbf{40.77}^{+3.32}_{-2.01}$ & $71.06^{+5.17}_{-3.29}$ \\
			\midrule
			
			FeaMGan-S (ours)& $45.16^{+5.24}_{-3.23}$ &	$34.93^{+6.92}_{-3.48}$ & $13.87^{+0.66}_{-0.89}$ & $40.50^{+2.83}_{-1.96}$ &$40.57^{+5.56}_{-4.83}$ & $13.32^{+2.35}_{-1.48}$ & $16.00^{+1.51}_{-2.76}$ & $24.55^{+2.88}_{-3.76}$ & $\mathbf{20.82}^{+5.94}_{-4.03}$ & $27.54^{+1.07}_{-0.34}$ & $63.09^{+8.99}_{-4.97}$ & $102.53^{+1.58}_{-0.79}$ & $42.58^{+3.36}_{-4.07}$ & $\mathbf{53.99}^{+5.41}_{-5.45}$ \\
			
			FeaMGan (ours)& $46.12^{+4.60}_{-5.80}$ &	$36.56^{+6.70}_{-7.96}$ & $13.69^{+1.13}_{-1.15}$ & $ 41.19^{+2.89}_{-2.81}$ & $42.69^{+4.00}_{-5.01}$ & $14.97^{+3.86}_{-3.23}$ & $17.35^{+5.09}_{-4.28}$ & $26.51^{+5.70}_{-3.27}$ & $\mathbf{20.25}^{+2.88}_{-2.56}$ & $26.34^{+1.36}_{-0.77}$ & $64.64^{+4.94}_{-5.07}$ & $102.23^{+10.58}_{-10.91}$ & ${42.38}^{+2.91}_{-3.01}$ & $\mathbf{54.52}^{+5.00}_{-3.09}$ \\		
			\midrule	
			\midrule
			\textbf{Viper$\rightarrow$Cityscapes} &   &  &  &  &  &  &   &  &  &  &    &  &  &    \\	
			
			Color Transfer& $89.30^{+0.06}_{-0.05}$ &	$83.51^{+0.10}_{-0.10}$ & $20.20^{+0.32}_{-0.24}$ & $51.23^{+0.75}_{-0.75}$ &$65.74^{+1.33}_{-2.59}$ & $19.98^{+1.08}_{-0.44}$ & $16.87^{+0.16}_{-0.14}$ & $26.65^{+2.69}_{-2.39}$ & $28.79^{+0.45}_{-0.30}$ & $36.21^{+0.14}_{-0.14}$ & $41.97^{+1.52}_{-3.65}$ & $139.26^{+10.03}_{-10.10}$ & $57.10^{+1.04}_{-0.74}$ & $79.73^{+0.75}_{-0.82}$ \\
			
			MUNIT & $47.96^{+0.52}_{-1.11}$ &	$30.35^{+0.68}_{-1.29}$ & $14.14^{+0.10}_{-0.09}$ & $59.62^{+1,87}_{-2.34}$ &$46.44^{+2.83}_{-2.02}$ & $15.85^{+0.87}_{-0.71}$ & $14.11^{+0.31}_{-0.11}$ & $32.69^{+2.23}_{-3.94}$ & $25.75^{+0.27}_{-0.36}$ & $25.76^{+0.24}_{-0.24}$ & $\mathbf{39.99}^{+1.07}_{-1.26}$ & $274.68^{+15.46}_{-23.29}$ & $46.64^{+1.75}_{-1.70}$ & $74.33^{+0.40}_{-0.79}$ \\
			
			CUT & $60.35^{+6.50}_{-8.13}$ &	$49.48^{+7.19}_{-10.15}$ & $16.80^{+1.04}_{-1.11}$ & $51.02^{+3.71}_{-4.32}$ &$\mathbf{34.79}^{+6.87}_{-2.98}$ & $14.88^{+0.79}_{-0.76}$ & $16.80^{+2.50}_{-1.68}$ & $\mathbf{22.40}^{+2.41}_{-2.33}$ & $22.91^{+1.81}_{-0.84}$ & $\mathbf{23.34}^{+1.10}_{-1.04}$ & $45.00^{+5.23}_{-2.68}$ & $224.47^{+29.25}_{-28.29}$ & $42.29^{+2.56}_{-2.55}$ & $63.36^{+1.72}_{-3.17}$ \\
			
			TSIT & $\mathbf{45.26}^{+1.92}_{-1.39}$ &	$\mathbf{28.40}^{+2.55}_{-2.16}$ & $\mathbf{8.47}^{+0.25}_{-0.26}$ & $50.03^{+3.06}_{-2.12}$ &$46.25^{+0.69}_{-0.93}$ & $\mathbf{14.46}^{+2.11}_{-1.95}$ & $\mathbf{12.28}^{+0.98}_{-0.97}$ & $31.95^{+5.17}_{-4.96}$ & $24.86^{+1.50}_{-1.27}$ & $24.91^{+1.26}_{-1.43}$ & $45.19^{+2.35}_{-2.10}$ & $184.05^{+18.06}_{-10.62}$ & $\mathbf{44.59}^{+2.46}_{-1.55}$ & $71.72^{+3.90}_{-3.72}$ \\
			
			QS-Attn& $55.62^{+12.05	}_{-9.66}$ &	$39.31^{+11.87}_{-9.92}$ & $12.99^{+1.87}_{-1.60}$ & $63.22^{+17.47}_{-13.74}$ &$36.44^{+15.38}_{-4.97}$ & $16.04^{+1.56}_{-1.26}$ & $15.25^{+1.10}_{-2.49}$ & $25.20^{+4.27}_{-2.30}$ & $26.09^{+1.88}_{-2.02}$ & $24.24^{+1.26}_{-0.89}$ & $46.54^{+1.63}_{-1.84}$ & $326.60^{+171.61}_{-128.90}$ & $46.44^{+3.78}_{-5.51}$ & $69.33^{+5.26}_{-5.06}$ \\
			\midrule	
			
			FeaMGan-S (ours)& $52.79^{+2.50}_{-2.79}$ &  $35.92^{+3.88}_{-3.18}$ &	$14,34^{+0.65}_{-0.73}$ &  $\mathbf{45.38}^{+1.53}_{-1.63}$ &$56.75^{+5.13}_{-8.76}$ & $18.51^{+1.49}_{-1.08}$ & $16.68^{+1.90}_{-3.18}$ & $42.85^{+1.59}_{-1.97}$ & $\mathbf{22.70}^{+1.41}_{-1.40}$ & $26.82^{+0.74}_{-1.12}$ & $45.27^{+1.37}_{-1.11}$ & $\mathbf{130.76}^{+6.27}_{-11.95}$ & $45.25^{+1.49}_{-3.29}$ & $\mathbf{48.19}^{+2.45}_{-1.49}$ \\
			
			FeaMGan (ours)& $51.56^{+1.97}_{-3.56}$ &	$34.63^{+3.32}_{-5.48}$ & $14.01^{+0.58}_{-0.73}$ & $\mathbf{47.21}^{+1.29}_{-1.10}$ &$58.87^{+3.48}_{-1.62}$ & $21.20^{+0.72}_{-0.93}$ & $18.03^{+1.38}_{-0.62}$ & $50.01^{+7.63}_{-3.72}$ & $23.55^{+3.08}_{-2.42}$ & $26.67^{+0.46}_{-0.66}$ & $45.55^{+1.11}_{-1.79}$ & $\mathbf{132.77}^{+2.96}_{-3.46}$ & $45.13^{+1.72}_{-1.51}$ & $\mathbf{50.32}^{+1.55}_{-1.24}$ \\
			\midrule	
			\midrule
			\textbf{Day$\rightarrow$Night} &   &  &  &  &  &  &   &  &  &  &    &  &  &    \\	
			
			Color Transfer& $125.90^{+0.13}_{-0.10}$ &	$140.60^{+0.10}_{-0.10}$ & $32.58^{+0.32}_{-0.52}$ & $56.52^{+1.76}_{-1.26}$ &$47.62^{+0.50}_{-0.78}$ & $27.41^{+1.37}_{-1.27}$ & $15.89^{+0.46}_{-0.23}$ & $\mathbf{32.60}^{+1.76}_{-2.27}$ & $44.24^{+0.30}_{-0.25}$ & $32.61^{+0.68}_{-1.07}$ & $128.57^{+11.25}_{-13.18}$ & $108.52^{+8.17}_{-6.36}$ & $25.65^{+0.43}_{-0.37}$ & $102.06^{+0.39}_{-0.31}$ \\
			
			MUNIT & $42.53^{+1.65}_{-1.27}$ &	$31.83^{+1.73}_{-0.98}$ & $15.02^{+0.64}_{-0.65}$ & $50.83^{+1.25}_{-0.88}$ &$\mathbf{29.25}^{+0.23}_{-0.29}$ & $28.00^{+0.72}_{-0.50}$ & $13.49^{+0.30}_{-0.16}$ & $36.57^{+1.16}_{-0.53}$ & $44.86^{+0.31}_{-0.35}$ & $\mathbf{24.96}^{+0.69}_{-0.59}$ & $115.00^{+6.94}_{-5.17}$ & $101.70^{+4.53}_{-4.06}$ & $19.66^{+0.34}_{-0.33}$ & $94.82^{+1.31}_{-1.37}$ \\
			
			CUT & $\mathbf{34.36}^{+3.71}_{-6.12}$ &	$\mathbf{20.54}^{+4.81}_{-7.05}$ & $10.16^{+1.98}_{-1.14}$ & $53.55^{+3.05}_{-3.22}$ &$31.89^{+0.92}_{-2.06}$ & $27.44^{+1.58}_{-1.46}$ & $\mathbf{13.14}^{+0.64}_{-0.73}$ & $40.93^{+6.70}_{-8.59}$ & $49.79^{+3.41}_{-2.78}$ & $25.52^{+1.69}_{-1.71}$ & $104.26^{+7.21}_{-10.28}$ & $122.50^{+11.90}_{-10.06}$ & $27.30^{+2.60}_{-2.87}$ & $92.76^{+0.52}_{-1.51}$ \\
			
			TSIT & $54.979^{6.83}_{-7.99}$ &	$33.21^{+5.26}_{-6.21}$ & $12.71^{+5.77}_{-3.49}$ & $57.91^{+2.92}_{-2.28}$ &$36.27^{+2.32}_{-1.39}$ & $31.56^{+1.18}_{-2.21}$ & $16.93^{+2.20}_{-1.07}$ & $45.23^{+9.74}_{-4.86}$ & $54.82^{+4.55}_{-4.89}$ & $29.09^{+3.19}_{-1.65}$ & $143.47^{+14.01}_{-10.47}$ & $99.30^{+3.07}_{-5.53}$ & $27.43^{+2.98}_{-4.22}$ & $94.98^{+2.46}_{-2.60}$ \\
			
			QS-Attn& $46.68^{+2.73}_{-2.03}$ &	$21.47^{+3.94}_{-2.55}$ & $\mathbf{7.58}^{+1.27}_{-1.77}$ & $52.02^{+4.14}_{-3.29}$ &$31.62^{+1.73}_{-1.66}$ & $\mathbf{26.73}^{+2.64}_{-3.41}$ & $13.26^{+0.99}_{-0.92}$ & $38.25^{+5.42}_{-3.84}$ & $47.26^{+3.31}_{-3.13}$ & $25.42^{+2.05}_{-1.80}$ & $\mathbf{100.84}^{+11.90}_{-7.85}$ & $123.79^{+18.23}_{-14.72}$ & $26.67^{+4.28}_{-3.72}$ & $86.39^{+5.62}_{-4.37}$ \\
			\midrule
			
			FeaMGan-S (ours)& $70.40^{+15.29}_{-4.76}$ &	$51.30^{+21.06}_{-6.09}$ & $14.68^{+3.45}_{-1.84}$ & $\mathbf{46.66}^{+2.63}_{-2.20}$ &$30.35^{+1.05}_{-0.84}$ & $35.47^{+3.71}_{-3.90}$ & $17.26^{+1.46}_{-1.08}$ & $47.29^{+10.32}_{-8.77}$ & $\mathbf{27.12}^{+1.14}_{-1.10}$ & $25.22^{+1.36}_{-1.38}$ & $116.25^{+4.47}_{-4.70}$ & $\mathbf{70.75}^{+4.87}_{-5.91}$ & $\mathbf{19.29}^{+0.52}_{-0.87}$ & $\mathbf{77.60}^{+4.37}_{-2.09}$ \\
			
			FeaMGan (ours)& $66.39^{+6.39}_{-8.43}$ & $46.96^{+10.07}_{-10.41}$ & $13.14^{+3.34}_{-2.01}$ & $\mathbf{46.88}^{+2.52}_{-2.83}$ & $29.72^{+2.56}_{-1.42}$ & $35.94^{+1.75}_{-1.77}$ &	$17.48^{+0.98}_{-1.44}$ & $49.78^{+7.45}_{-7.78}$ & $28.78^{+1.94}_{-3.03}$ &$25.65^{+0.71}_{-0.61}$ &  $115.66^{+2.72}_{-5.03}$ & $\mathbf{70.94}^{+9.44}_{-11.48}$ & $\mathbf{19.23}^{+0.64}_{-0.78}$ & $\mathbf{75.57}^{+4.86}_{-3.73}$ \\
			\midrule	
			\midrule
			\textbf{Clear$\rightarrow$Snowy} &   &  &  &  &  &  &   &  &  &  &    &  &  &    \\
			
			Color Transfer& $46.85^{+0.12}_{-0.38}$ &	$19.44^{+0.43}_{-1.43}$ & $14.91^{+0.86}_{-2.97}$ & $42.89^{+13.81}_{-3.96}$ &$25.78^{+1.19}_{-2.03}$ & $22.99^{+3.22}_{-2.15}$ & $16.01^{+0.37}_{-0.35}$ & $21.54^{+1.72}_{-1.52}$ & $41.13^{+7.40}_{-2.60}$ & $24.20^{+1.99}_{-0.78}$ & $\mathbf{57.67}^{+18.52}_{-11.79}$ & $128.26^{+94.11}_{-29.48}$ & $25.95^{+7.36}_{-2.36}$ & $65.39^{+5.38}_{-1.91}$ \\
			
			MUNIT & $\mathbf{44.74}^{+1.23}_{-0.79}$ &	$17.48^{+0.59}_{-0.86}$ & $11.65^{+0.34}_{-0.22}$ & $48.10^{+0.49}_{-0.73}$ &$28.47^{+0.78}_{-0.73}$ & $25.64^{+0.30}_{-0.46}$ & $15.27^{+0.21}_{-0.13}$ & $26.21^{+0.60}_{-0.48}$ & $40.31^{+0.58}_{-0.37}$ & $24.26^{+0.13}_{-0.09}$ & $101.98^{+1.73}_{-4.77}$ & $116.14^{+3.89}_{-4.34}$ & $21.63^{+0.41}_{-0.18}$ & $81.08^{+0.31}_{-0.63}$ \\
			
			CUT & $46.03^{+1.08}_{-0.85}$ &	$15.70^{+0.77}_{-0.94}$ & $14.71^{+1.15}_{-0.94}$ & $43.91^{+0.97}_{-0.78}$ &$26.74^{+0.50}_{-0.63}$ & $\mathbf{21.96}^{+0.75}_{-0.96}$ & $\mathbf{13.15}^{+0.31}_{-0.50}$ & $21.49^{+1.02}_{-1.08}$ & $35.20^{+0.47}_{-0.28}$ & $25.31^{+0.50}_{-0.39}$ & $76.67^{+4.18}_{-6.21}$ & $119.13^{+16.19}_{-10.00}$ & $23.91^{+0.59}_{-1.18}$ & $75.51^{+1.17}_{-0.60}$ \\
			
			TSIT & $79.29^{+5.08}_{-6.69}$ &	$40.02^{+7.17}_{-7.14}$ & $12.97^{+0.37}_{-0.51}$ & $41.52^{+3.31}_{-2.59}$ &$28.02^{2.81}_{-1.34}$ & $22.72^{+2.06}_{-1.40}$ & $14.32^{+0.64}_{-0.48}$ & $\mathbf{18.92}^{+2.42}_{-2.13}$ & $34.54^{+2.08}_{-2.52}$ & $23.02^{+0.69}_{-0.66}$ & $72.13^{+7.37}_{-4.90}$ & $104.05^{+12.49}_{-12.64}$ & $21.64^{+2.06}_{-1.56}$ & $75.84^{+4.73}_{-5.53}$ \\
			
			QS-Attn& $60.91^{+0.79}_{-1.02}$ &	$18.85^{+1.05}_{-1.36}$ & $14.19^{+1.70}_{-1.01}$ & $44.00^{+1.95}_{-1.80}$ &$25.60^{+0.50}_{-0.75}$ & $22.04^{+1.20}_{-1.41}$ & $13.24^{+0.17}_{-0.16}$ & $22.71^{+0.46}_{-0.71}$ & $36.02^{+2.20}_{-1.40}$ & $26.45^{+1.42}_{-1.75}$ & $78.58^{+4.70}_{-4.53}$ & $114.07^{+13.76}_{-8.81}$ & $25.17^{+3.07}_{-1.42}$ & $76.08^{+1.84}_{-2.29}$ \\
			\midrule
			
			FeaMGan-S (ours)& $57.93^{+1.37}_{-1.68}$ &	$16.24^{+1.19}_{-0.59}$ & $11.88^{+0.55}_{-0.52}$ & $\mathbf{38.28}^{+2.86}_{-2.83}$ &$\mathbf{22.69}^{+0.67}_{-1.34}$ & $25.71^{+1.14}_{-1.76}$ & $15.82^{+0.82}_{-1.27}$ & $37.47^{+3.98}_{-4.13}$ & $\mathbf{25.94}^{+1.50}_{-2.69}$ & $\mathbf{21.80}^{+0.53}_{-0.61}$ & $75.32^{+4.53}_{-6.80}$ & $\mathbf{81.47}^{+22.52}_{-14.46}$ & $\mathbf{19.10}^{+0.55}_{-0.30}$ & $\mathbf{57.46}^{+2.18}_{-3.98}$ \\
			
			FeaMGan (ours)& $56.78^{+0.81}_{-0.32}$ &	$\mathbf{14.77}^{+0.98}_{-1.59}$ & $\mathbf{11.36}^{+0.21}_{-0.43}$ & $41.72^{+2.61}_{-1.76}$ &$\mathbf{22.71}^{+1.17}_{-1.09}$ & $26.64^{+3.01}_{-2.07}$ & $16.19^{+1.74}_{-1.13}$ & $38.00^{+3.46}_{-2.77}$ & $\mathbf{27.78}^{+2.95}_{-2.21}$ & $\mathbf{21.41}^{+0.61}_{-0.69}$ & $79.35^{+3.72}_{-1.97}$ & $105.08^{+35.80}_{-25.18}$ & $\mathbf{19.50}^{+0.67}_{-1.04}$ & $\mathbf{60.59}^{+1.99}_{-1.90}$ \\
			\bottomrule
		\end{tabular}}}
	\end{center}
	\vspace{-2ex}
\end{table}

\begin{table}[t]
	\RawFloats
	\caption[Extended quantitative evaluation for ablation study.]{Results are reported as the average across five runs.}
	\vspace{-1ex}
	\label{tab:feamgan:app:quantitative_ablation_extended}
	\begin{center}
		\rotatebox{90}{
		\scalebox{0.6}{%
		\setlength{\tabcolsep}{0.3em}
		\begin{tabular}{lcccccccccccccc}
			\toprule
			\multirow{2}{*}{Method}&\multirow{2}{*}{FID}&\multirow{2}{*}{KID} &\multirow{2}{*}{sKVD}&\multicolumn{11}{c}{cKVD}\\
			\cmidrule(lr){5-15}
			&    &  &  & AVG& 		sky& 	ground&	road&	terrain&	vegetation&	building&	roadside-obj.&	person&	vehicle&	rest	\\
			\midrule
			FeaMGan (Full)& $46.11^{+4.60}_{-5.80}$ &	$36.66^{+6.70}_{-7.96}$ & $13.69^{+1.13}_{-1.15}$ & $ 41.19^{+2.89}_{-2.81}$ & $42.69^{+4.00}_{-5.01}$ & $14.97^{+3.86}_{-3.23}$ & $17.35^{+5.09}_{-4.28}$ & $26.51^{+5.70}_{-3.27}$ & $\mathbf{20.25}^{+2.88}_{-2.56}$ & $26.34^{+1.36}_{-0.77}$ & $64.64^{+4.94}_{-5.07}$ & $102.23^{+10.58}_{-10.91}$ & $\mathbf{42.38}^{+2.91}_{-3.01}$ & $54.52^{+5.00}_{-3.09}$ \\
			
			w/o Dis. Mask & $ \mathbf{37.10}^{+3,03}_{-5,46}$ & $ \mathbf{25.88}^{+3.60}_{-8.65}$ & $ 14.73^{+1.27}_{-1.66}$ & $\mathbf{39.65}^{+4.73}_{-4.16}$ & $\mathbf{26.70}^{+4.09}_{-5.99}$ & $15.81^{+5.28}_{-3.67}$ & $16.65^{+4.43}_{-2.85}$ & $31.02^{+11.19}_{-8.23}$ & $22.97^{+4.15}_{-6.89}$ & $\mathbf{25.39}^{+2.75}_{-1.24}$ & $67.01^{+6.63}_{-7.99}$ & $\mathbf{93.78}^{+10.71}_{-7.90}$ & $44.23^{+2.88}_{-4.95}$ & $\mathbf{52.91}^{+5.34}_{-5.61}$\\
			
			w/ FADE w/o FATE & $45.46^{+2.68}_{-2.65}$ & $ 35.73^{+4.44}_{-3.53}$ & $ \mathbf{13.17}^{+0.54}_{-0.60}$ & $40.90^{+4.40}_{-2.04}$ & $41.49^{+8.66}_{-8.86}$ & $13.78^{+3.22}_{-1.70}$ & $16.78^{+4.99}_{-2.30}$ & $25.30^{+2.41}_{-2.12}$ & $20.58^{+1.62}_{-1.88}$ & $27.21^{+1.93}_{-0.68}$ & $\mathbf{63.12}^{+3.77}_{-5.25}$ & $104.43^{+13.22}_{-7.73}$ & $42.44^{+1.27}_{-0.91}$ & $53.83^{+5.46}_{-1.96}$\\
			
			w/ Random Crop & $47.88^{+5.10}_{-3.82}$ & $38.48^{+7.27}_{-4.14}$ & $	13.37^{+0.61}_{-1.14}$ & $ 40.18^{+1.55}_{-1.66}$ & $39.88^{+3.92}_{-7.65}$ & $\mathbf{12.90}^{+0.94}_{-0.68}$ & $\mathbf{14.65}^{+1.62}_{-2.00}$ & $\mathbf{25.09}^{+2.01}_{-2.93}$ & $21.89^{+5.21}_{-3.36}$ & $27.32^{+2.05}_{-1.87}$ & $64.32^{+2.86}_{-1.79}$ & $98.81^{+7.63}_{-5.61}$ & $43.08^{+3.81}_{-2.90}$ & $53.86^{+1.66}_{-1.56}$ \\
			
			w/ VGG Crop& $51.23^{+5.12}_{-2.11}$ & $	42.46^{+6.65}_{-3.42}$ & $	13.56^{+0.78}_{-0.81}$ & $ 40.62^{+2.22}_{-1.99}$ & $40.32^{+4.36}_{-5.72}$ & $13.3^{+2.06}_{-1.83}$ & $15.67^{+2.15}_{-2.14}$ & $26.47^{+4.07}_{-1.25}$ & $21.09^{+1.61}_{-2.41}$ & $27.28^{+0.98}_{-1.46}$ & $65.23^{+8.64}_{-3.75}$ & $99.61^{+9.27}_{-7.66}$ & $43.19^{+2.16}_{-3.12}$ & $53.94^{+3.81}_{-5.04}$ \\
			\midrule			
			w/o Local Dis. &   &  &  &  &  &  &   &  &  &  &    &  &  &    \\			
			- w/ 256$\times$256 Crop& $48.57^{+5.16}_{-2.30}$ & $ 38.89^{+6.82}_{-3.47}$ & $ \mathbf{12.89}^{+1.32}_{-0.90}$ & $ 41.26^{+3.00}_{-2.64}$ & $	42.31^{+2.92}_{-3.88}$ & $	13.57^{+1.65}_{-2.15}$ & $ 15.98^{+1.44}_{-1.58}$ & $	\mathbf{25.28}^{+4.26}_{-2.75}$ & $	22.18^{+8.57}_{-4.62}$ & $ \mathbf{26.56}^{+2.06}_{-1.46}$ & $\mathbf{61.13}^{+2.92}_{-2.21}$ & $	107.48^{+6.40}_{-11.62}$ & $ 42.44^{+4.20}_{-2.30}$ & $	55.62^{+5.79}_{-4.61}$\\
			
			- w/ 352$\times$352 Crop& $47.26^{+3.44}_{-2.31}$ & $ 37.75^{+5.30}_{-3.23}$ & $ 14.38^{+0.76}_{-1.00}$ & $ 39.30^{+2.59}_{-1.40}$ & $34.44^{+7.70}_{-5.20}$ & $\mathbf{13.09}^{+1.62}_{-1.17}$ & $15.84^{+1.79}_{-1.04}$ & $ 25.83^{+1.56}_{-2.01}$ & $21.50^{+3.84}_{-1.63}$ & $27.20^{+1.47}_{-1.24}$ & $61.24^{+3.62}_{-5.82}$ & $98.25^{+4.82}_{-3.34}$ & $42.24^{+2.34}_{-1.79}$ & $53.38^{+3.71}_{-2.18}$\\
			
			- w/ 464$\times$464 Crop& $\mathbf{46.61}^{+2.57}_{-3.25}$ & $ \mathbf{37.25}^{+3.17}_{-4.48}$ & $ 15.04^{+0.79}_{-0.45}$ & $\mathbf{38.62}^{+1.52}_{-1.68}$ & $\mathbf{31.60}^{+4.89}_{-3.80}$ & $13.13^{+1.08}_{-0.99}$ & $\mathbf{15.38}^{+1.77}_{-1.93}$ & $27.06^{+1.37}_{-2.76}$ & $22.23^{+3.99}_{-3.40}$ & $29.67^{+0.67}_{-1.10}$ & $63.38^{+2.10}_{-3.73}$ & $\mathbf{87.51}^{+3.83}_{-4.99}$ & $44.41^{+2.76}_{-2.76}$ & $51.77^{+2.19}_{-3.06}$\\
			
			- w/ 512$\times$512 Crop& $55.89^{+19.35}_{-12.54}$ & $ 49.12^{+26.93}_{-16.19}$ & $ 15.94^{+4.24}_{-2.24}$ & $ 39.35^{+4.47}_{-3.91}$ & $36.48^{+6.33}_{-12.08}$ & $14.68^{+2.82}_{-2.30}$ & $ 16.06^{+3.25}_{-3.12}$ & $26.87^{+8.35}_{-4.31}$ & $\mathbf{19.61}^{+4.76}_{-4.21}$ & $27.37^{+3.00}_{-2.34}$ & $62.40^{+3.93}_{-4.27}$ & $98.90^{+8.87}_{-5.05}$ & $\mathbf{40.32}^{+3.12}_{-4.21}$ & $\mathbf{50.86}^{+5.50}_{-7.19}$\\	
			\bottomrule
		\end{tabular}}}
	\end{center}
	\vspace{-2ex}
\end{table}

\backmatter

\cleardoublepage
\phantomsection
\bibliographystyle{plain}
\bibliography{bibliography}
\addcontentsline{toc}{chapter}{Bibliography}

\end{document}